\RequirePackage[l2tabu, orthodox]{nag}

\documentclass[12pt,phd,a4paper,oneside]{ucl_thesis}
\AtBeginDocument{\pagestyle{plain}}

\usepackage{emptypage}

\usepackage{gensymb}
\usepackage{textcomp}

\usepackage{setspace}

\usepackage[T1]{fontenc}
\usepackage[utf8]{inputenc}
\usepackage[english]{babel}
\usepackage{mathtools}
\usepackage{url}
\usepackage{nicefrac}
\usepackage{bm}
\usepackage{float}
\usepackage{bbold}

\usepackage{microtype}
\usepackage{graphicx}
\usepackage{subcaption}
\usepackage{booktabs}
\usepackage{ upgreek }
\usepackage{IEEEtrantools}

\usepackage[inline]{enumitem}
\usepackage{pgfplots}
\usepackage{tikz}

\usepackage{amsbsy, amsmath, amsthm, amsfonts, amssymb}
\usepackage[ruled, vlined,algo2e]{algorithm2e}
\usepackage{cases}
\usepackage[dvipsnames]{xcolor}

\usepackage{tabularx}

\usepackage{svg}
\usepackage{epigraph}
\usepackage{tcolorbox}

\usepackage{scalerel,stackengine}
\usepackage{pdfpages}
\stackMath

\usepackage{xr}
\makeatletter

\newcommand*{\addFileDependency}[1]{
\typeout{(#1)}
\@addtofilelist{#1}
\IfFileExists{#1}{}{\typeout{No file #1.}}
}\makeatother

\usepackage{algorithm,algorithmic}
\usepackage{dsfont}

\usepackage{natbib}

\makeatletter\let\saved@bibitem\@bibitem\makeatother
\usepackage{hyperref}
\usepackage[nameinlink]{cleveref}
\hypersetup{
  colorlinks=true,
  linkcolor=blue,
  filecolor=magenta,
  urlcolor=cyan,
  citecolor=blue
}
\newcommand*\isbnhref[1]{\href{https://openlibrary.org/isbn/#1}{\nolinkurl{#1}}}
\providecommand*\doi[1]{doi: \href{https://doi.org/#1}{\nolinkurl{#1}}}
\providecommand*\isbn[1]{\href{https://openlibrary.org/isbn/#1}{\nolinkurl{#1}}}

\makeatletter\let\@bibitem\saved@bibitem\makeatother
\makeatletter
\AtBeginDocument{
    \hypersetup{
        pdfsubject={Thesis Subject},
        pdfkeywords={Thesis Keywords},
        pdfauthor={Author},
        pdftitle={Title},
    }
}
\makeatother

\newcommand\reallywidehat[1]{%
\savestack{\tmpbox}{\stretchto{%
  \scaleto{%
    \scalerel*[\widthof{\ensuremath{#1}}]{\kern-.6pt\bigwedge\kern-.6pt}%
    {\rule[-\textheight/2]{1ex}{\textheight}}
  }{\textheight}%
}{0.5ex}}%
\stackon[1pt]{#1}{\tmpbox}%
}
\newenvironment{talign*}
 {\csname align*\endcsname}
 {\endalign}

\newtheorem{theorem}{Theorem}
\newtheorem{corollary}{Corollary}

\newtheorem{connection}{Connection}
\crefname{connection}{Connection}{Connections}
\newtheorem{conjecture}{Conjecture}
\crefname{conjecture}{Conjecture}{Conjectures}
\newtheorem{lemma}{Lemma}
\newtheorem{proposition}{Proposition}
\newtheorem{assumption}{Assumption}

\crefname{assumption}{}{}
\newtheorem{oldassumption}{Assumption}

\crefname{oldassumption}{}{}

\crefname{subsubsubappendix}{Section}{Sections}
\Crefname{subsubsubappendix}{Section}{Sections}

\theoremstyle{definition}
\newtheorem{definition}{Definition}

\theoremstyle{definition} 
\newtheorem{remark}[theorem]{Remark}

\DeclareMathOperator{\diag}{diag}

\def\bR{\mathbb{R}}
\def\bP{\mathbb{P}}
\def\bQ{\mathbb{Q}}

\def\bE{\mathbb{E}}

\def\bN{\mathbb{N}}

\def\bU{\mathbb{U}}
\def\bC{\mathbb{C}}

\def\calL{\mathcal{L}}
\def\calX{\mathcal{X}}

\def\calH{\mathcal{H}}

\def\calP{\mathcal{P}}

\def\calW{\mathcal{W}}
\def\calF{\mathcal{F}}
\def\calN{\mathcal{N}}
\def\calB{\mathcal{B}}
\def\calU{\mathcal{U}}
\def\calC{\mathcal{C}}

\def\given{\ |\ }

\def\GP{\mathrm{GP}}
\def\BQ{\mathrm{BQ}}

\def\Id{\mathrm{Id}}

\def\MMD{\mathrm{MMD}}
\def\ekqd{\text{e-KQD}}
\def\supkqd{\text{sup-KQD}}

\def\argmin{\mathop{\text{argmin}}}
\def\argmax{\mathop{\text{argmax}}}

\def\d{\text{d}}

\newcommand{\bigo}{\mathcal{O}}
\newcommand{\prt}{\mathrm{Prt}}

\def\bx{\mathbf{x}}

\newcommand{\spn}[1]{\mathrm{span}\{#1\}}

\crefname{enumi}{}{}
\crefname{enumii}{}{}
\crefname{corollary}{Corollary}{Corollaries}

\newcommand{\eqcolon}{\mathrel{\resizebox{\widthof{$\mathord{=}$}}{\height}{ $\!\!=\!\!\resizebox{1.2\width}{0.8\height}{\raisebox{0.23ex}{$\mathop{:}$}}\!\!$ }}}

\makeatletter
\newcommand{\customlabel}[2]{%
   \protected@write \@auxout {}{\string \newlabel {#1}{{#2}{\thepage}{#2}{#1}{}} }%
   \hypertarget{#1}{}
}
\makeatother

\def\tausqcvest{\hat{\tau}_\textup{CV}^2}
\def\tausqmlest{\hat{\tau}_\textup{ML}^2}
\def\tausqicvest{\hat{\tau}_\textup{ICV}^2}

\newcommand{\nocontentsline}[3]{}
\newcommand{\tocless}[2]{\bgroup\let\addcontentsline=\nocontentsline#1{#2}\egroup}

\begin{document}

\title{Scalable Kernel-Based Distances for Statistical Inference and Integration}

\author{Masha Naslidnyk}
\department{Department of Computer Science}

\maketitle

\begin{abstract} 
Representing, comparing, and measuring the distance between probability distributions is a key task in computational statistics and machine learning. The choice of representation and the associated distance determine properties of the methods in which they are used: for example, certain distances can allow one to encode robustness or smoothness of the problem. Kernel methods offer flexible and rich Hilbert space representations of distributions that allow the modeller to enforce properties through the choice of kernel, and estimate associated distances at efficient nonparametric rates. In particular, the maximum mean discrepancy (MMD), a kernel-based distance constructed by comparing Hilbert space mean functions, has received significant attention due to its computational tractability and is favoured by practitioners.

In this thesis, we conduct a thorough study of kernel-based distances with a focus on efficient computation, with core contributions in~\Cref{sec:mmdsbi,sec:mmdintegration,sec:gpcv,sec:kqe}. \Cref{sec:part1} of the thesis is focused on the MMD, specifically on improved MMD estimation. In~\Cref{sec:mmdsbi} we propose a theoretically sound, improved estimator for MMD in simulation-based inference. Then, in~\Cref{sec:mmdintegration}, we propose an MMD-based estimator for conditional expectations, a ubiquitous task in statistical computation. Closing \Cref{sec:part1}, in~\Cref{sec:gpcv} we study the problem of calibration when MMD is applied to the task of integration.

In~\Cref{sec:part2}, motivated by the recent developments in kernel embeddings beyond the mean, we introduce a family of novel kernel-based discrepancies: kernel quantile discrepancies. These address some of the pitfalls of MMD, and are shown through both theoretical results and an empirical study to offer a competitive alternative to MMD and its fast approximations. We conclude with a discussion on broader lessons and future work emerging from the thesis.
\end{abstract}

\begin{acknowledgements}
Pursuing a PhD was something I had long dreamed of. After seven years working in the industry, it still felt like the right next step; when I got my acceptance, it barely felt real. Everything shifted after that moment. Over the next four years, I was lucky to explore new ideas and join the academic community. There were ups and downs, but nothing matches the freedom and creativity I found in this work. I feel very fortunate.

First, I’m grateful to my supervisors: Carlo Ciliberto, for unwavering support and timely insight; Jeremias Knoblauch, for inspiring passion and radical honesty; and François-Xavier Briol, my first supervisor, for your consistent generosity with time, your deep investment in my development, and for showing, by example, that passion in research can be sustainable. I learned so much from you, and hope to have the privilege of passing it on.

Research is a collaborative adventure. I’m grateful to Ayush Bharti, Siu Lun (Alan) Chau, Zonghao (Hudson) Chen, Arthur Gretton, Toni Karvonen, Oscar Key, Maren Mahsereci, Motonobu Kanagawa, and Krikamol Muandet for their time and expertise; the work in this thesis would not have been possible without you. Thank you to Ti John, Antonin Schrab, Harita Dellaporta, Dimitri Meunier, Hugh Dance, Nathaël Da Costa, Takuo Matsubara, and Xing Liu for enlightening conversations. Being part of a team makes all the difference; to everyone in The Fundamentals of Statistical Machine Learning Group, thank you for four years of lively discussions, countless cross-reviews, and the little things that kept the days light.

Bianca, thank you for the many reality checks and for encouraging me to go for things, whether a PhD application or a swim under a waterfall. Maren, thank you for believing in me when there weren’t many reasons to, and for introducing me to the world of GPs. Prof. Danila Proskurin, your early guidance and encouragement to explore research changed my life.

To friends and family: thank you for making life joyful. I’ve been terrified of leaving someone out for all four years, so I’ll say this instead: if we shared a laugh, an adventure, or a slice of cake, thank you. I’m grateful for you.

To Sunil: thank you for sharing your life with me and making everything better. To Puff, my hairiest collaborator: woof woof, bark. Good boy.

To my parents, whose strength and love are unmatched, who taught me to always make space for lightness and laughter, especially when things grow heavy: I am lucky to be your daughter. This is all thanks to you.

\end{acknowledgements}

\setcounter{tocdepth}{1} 
\tableofcontents
\listoffigures
\listoftables

\chapter{Introduction}
\label{sec:introduction}
\setlength{\epigraphwidth}{0.66\textwidth}
\epigraph{The distinction between a `test' and a `measure' of group divergence is fundamental: a test merely tells us whether the two groups (from which the two given samples are drawn) are different or not, while a measure gives us a quantitative estimate of the magnitude of the difference (if any) between the two groups.}{\citet{mahalanobis1930}}

Comparing two probability distributions $\bP, \bQ$ is a keystone of many methods in computational statistics and machine learning. At their core, a wide range of fundamental tasks can be framed as questions about which distributions are closer or farther apart under some notion of discrepancy $D(\bP, \bQ)$:
\begin{itemize}
        \item  Hypothesis testing~\citep{salicru1994applications,sejdinovic2012hypothesis}: does the observed discrepancy between $\bP$ and $\bQ$ exceed what is expected under the null hypothesis of zero discrepancy? This includes two-sample (empirical vs.~empirical), goodness-of-fit (empirical vs.~model), and independence (joint vs.~product of marginals) testing.
        \item  Parameter estimation and inference~\citep{basu2011statistical,briol2019statistical}: which point estimate or posterior distribution over a parameter $\theta$ places weight on models $\bP_\theta$ that are close to the observed data distribution $\bQ$ under $D$?
        \item  Numerical integration~\citep{davis2007methods,karvonen2017classical}: which choice of points and weights $\{x_n, w_n\}_{n=1}^N$ in $\int_\calX f(x) \bP(\d x) \approx \sum_{n=1}^N w_n f(x_n)$ minimises the discrepancy between the weighted empirical distribution $\sum_{n=1}^N w_n \delta_{x_n}$ and the integration measure $\bP$?
        \item  Domain adaptation and generative modelling~\citep{gretton2009covariate,acuna2021f}: what transformation $T$ on the input space of the source distribution $\bP$ minimises the discrepancy between the transformed $T\#\bP$ and the target $\bQ$?
        \item   Dataset summarisation and thinning~\citep{dwivedi2024kernel,Mak2018}: which subset of samples $\{y_m\}_{m=1}^M \subseteq \{x_n\}_{n=1}^N$ minimises the discrepancy between the empirical distributions $\nicefrac{1}{M} \sum_{m=1}^M \delta_{y_m}$ and $\nicefrac{1}{N} \sum_{n=1}^N \delta_{x_n}$?
\end{itemize}

In each of these examples, the choice of discrepancy determines the properties of the method addressing the task. This raises a central question:
\textbf{what makes a `good' discrepancy?} At the very least, the discrepancy should be able to tell different distributions apart: $\bP \neq \bQ$ should imply $D(\bP, \bQ)>0$. It should also be computationally convenient, with efficient estimators that do not suffer from the `curse of dimensionality'. Ideally, to enable optimisation, the gradients of the discrepancy should be efficiently computable as well. The \emph{maximum mean discrepancy} (MMD,~\citet{borgwardt2006integrating,gretton2012kernel}), a kernel-based notion of discrepancy that compares means of certain distributional embeddings, satisfies these desiderata, leading to its widespread adoption. However, its practical application reveals key challenges: standard estimators can be inefficient, and the reliance on mean embeddings can be restrictive. This thesis addresses these limitations by developing kernel-based discrepancies and estimators that are more efficient, expressive, and robust. To situate these contributions, we begin with the historical context from which these methods emerged, before turning to the MMD and then outlining the challenges that motivate the thesis.

\section{From tests to measures of discrepancy}

The idea of discrepancy between probability distributions as a central object of study in statistical methodology is not new. In the early 20th century, classification problems in biology and anthropology motivated statisticians to draw a sharp line between \emph{tests of group divergence} (in today’s terms, hypothesis tests giving a reject/do not reject answer) and \emph{measures of group divergence} (discrepancies between empirical distributions), and explicitly argue for general-purpose, non-ad hoc discrepancies as a methodological foundation. \citet{mahalanobis1936} established this line of work with the introduction of what is now called the Mahalanobis distance, which measures how far a point lies from a distribution while accounting for correlations between variables; \citet{bhattacharyya1943} generalised this idea to define a discrepancy between two distributions.

Since then, many further notions of discrepancies and discrepancy-based methods in statistical and machine learning have been proposed: some novel, others extending existing statistics into discrepancy measures. For example,~\citet{csiszar1963informationstheoretische}
reinterpreted the statistic in Pearson's $\chi^2$ test as a discrepancy: specifically, a $\chi^2$-divergence, a kind of $f$-divergence. $f$-divergences quantify the discrepancy between distributions through their ratios: for a convex function $f$, the $f$-divergence between $\bP$ and $\bQ$ is defined as 
\begin{equation*}
    D_f(\bP, \bQ)=\bE_{X \sim \bQ} [f \circ \d \bP / \d \bQ ] (X),
\end{equation*}
the expected value under $\bQ$ of $f$ composed with the Radon-Nikodym derivative of $\bP$ with respect to $\bQ$. Important discrepancies, such as the total variation distance~\citep[Section 2.9]{vanDerVaart1998}, the Kullback--Leibler divergence~\citep{kullback1951information}, and the aforementioned $\chi^2$-divergence are all $f$-divergences for different choices of $f$. The main benefit of the $f$-divergences is the interpretability of the ratio $\d \bP / \d \bQ$; however, it is only well-defined when $\bP$ is absolutely continuous with respect to $\bQ$, which in turn requires the support of $\bP$ to lie within the support of $\bQ$. Additionally, estimating $f$-divergences is a notoriously challenging task: the $f$-divergence between empirical distributions, $D_f(\bP_N, \bQ_N)$, does not converge to the $f$-divergence between true distributions $D_f(\bP, \bQ)$, as $N \to \infty$~\citep[Section 8.4.2]{peyre2019computational}. The proposed alternative nonparametric and neural network-based estimators suffer from the curse of dimensionality, converging at rates as slow as $N^{-1/d}$ unless $f$ and $\d \bP / \d \bQ$ exhibit sufficient smoothness~\citep{rubenstein2019practical,mcallester2020formal,nguyen2010estimating}.

Another major family of discrepancies is motivated by a classic result on uniqueness of the representation of probability measures via integrals of bounded continuous functions~\citep[Lemma 9.3.2]{dudley2002}: any two distributions $\bP, \bQ$ on a metric space $(\calX, d)$ coincide if and only if $\bE_{X \sim \bP} f(X) = \bE_{Y \sim \bQ} f(Y)$ for all continuous and bounded functions $f: \calX \to \bR$. This class of functions, while expressive, is prohibitively large for practical applications. \emph{Integral probability pseudometrics} (IPMs,~\citet{muller1997integral}) are a generalisation of this idea to an arbitrary class of real-valued functions $\calF$, defined as
\begin{equation}
    \mathrm{IPM}_\calF(\bP, \bQ) = \sup_{f \in \calF} \left| \bE_{X \sim \bP} f(X) - \bE_{Y \sim \bQ} f(Y) \right|.
\end{equation}
Trivially, when $\calF$ contains all continuous functions bounded by some constant, the corresponding IPM is a metric; when $\calF$ is the class of all polynomials of order up to $p$, the IPM reduces to $p$-moment matching. Other choices of $\calF$ recover well-known discrepancies, including total variation distance (that metrises strong convergence)\footnote{Curiously, total variation distance is, up to a constant scaling factor, the only discrepancy that is both an $f$-divergence and an IPM~\citep{Sriperumbudur2012}.}, the 1-Wasserstein distance~\citep{Kantorovich1942,villani2009optimal}, and the canonical kernel-based distance, the MMD. Whenever the IPM metrises weak convergence, as is the case for the 1-Wasserstein distance and the MMD for a large class of kernels~\citep{simon2023metrizing}, the IPM between empirical distributions is a consistent estimator of the true IPM. Its sample complexity depends on the class $\calF$; for example, the 1-Wasserstein distance estimator converges at the slow rate of $N^{-1/d}$, unless the distributions are supported on a lower-dimensional manifold~\citep{Fournier2015}. 

\section{The MMD and its applications}

When $\calF=B_{\calH_k}$, the unit ball of some \emph{reproducing kernel Hilbert space} (RKHS,~\citet{Berlinet2004}) $\calH_k$ with inner product $\langle \cdot, \cdot \rangle_{\calH_k}$ induced by a \emph{reproducing kernel} $k: \calX \times \calX \to \bR$, the corresponding IPM, the MMD, is particularly computationally attractive. Unlike other common IPMs, the MMD~\citep{borgwardt2006integrating,gretton2012kernel} is available in closed functional form: owing to the rich structure of the RKHS, the MMD between $\bP$ and $\bQ$ is exactly the RKHS inner product norm of $\mu_{k, \bP} - \mu_{k, \bQ}$, the difference between \emph{kernel mean embeddings}~\citep{smola07hilbert} of $\bP$ and $\bQ$,
\begin{equation*}
    \MMD_k(\bP, \bQ) = \| \mu_{k, \bP} - \mu_{k, \bQ} \|_{\calH_k}.
\end{equation*}
$\mu_{k, \bP}$ represents a distribution $\bP$ as a mean function in the RKHS; when the mapping $\bP \mapsto \mu_{k, \bP}$ is injective, meaning the representation of a distribution as a kernel mean embedding is unique and captures all information about the distribution, the kernel is said to be `characteristic'~\citep{Sriperumbudur2009}. MMD has received significant attention due to its computational tractability: its most common estimator has cost $\bigo(N^2)$ and converges at the rate $\bigo(N^{-1/2})$ as the number of datapoints $N$ increases, with cheaper alternatives proposed~\citep{gretton2012kernel,chwialkowski2015fast,Bodenham2023,Schrab2022}.
For this reason, the MMD has been used to address a broad range of tasks from hypothesis testing~\citep{gretton2012kernel} to parameter estimation~\citep{briol2019statistical,cherief2020mmd}, causal inference~\citep{muandet2021counterfactual,Sejdinovic2024}, feature attribution~\citep{chau2022rkhs,chau2023explaining}, and learning on distributions~\citep{Muandet12:SMM,szabo2016learning}. A particularly elegant application, and one central to this thesis, is numerical integration, which we cover next.

\section{MMD for integration: kernel and Bayesian quadrature}

MMD-minimising integration methods are known as \emph{kernel quadrature}~\citep{sommariva2006numerical,dick2013high,bach2017equivalence,kanagawa2020convergence} or, in their probabilistic form, \emph{Bayesian quadrature}~\citep{Diaconis1988,OHagan1991,Hennig2022}. These approaches have been used primarily in statistical computation (see~\citet{briol2019statistical} and references therein), as well as in diverse scientific and engineering domains such as astronomy~\citep{Lin2025}, battery design~\citep{dlr201186}, weather modelling~\citep{Osborne2012}, reliability engineering~\citep{COUSIN2024102483}, computer graphics \citep{Marques2013,xi2018bayesian}, cardiac modelling \citep{Oates2017heart} and tsunami modelling \citep{li2022multilevel}.
Kernel quadrature replaces classical numerical integration rules (such as Monte Carlo and trapezoidal) with weighted schemes in which the weights are selected to minimise the MMD, which can also be shown to be the root mean squared integration error over functions sampled from a Gaussian process~\citep[Corollary 7 in p.40]{ritter2000}. This view naturally leads to Bayesian quadrature, a probabilistic interpretation in which the integrand is modelled as a Gaussian process. This approach produces not only an estimate of the integral, but also a principled posterior uncertainty quantification, and a way to estimate hyperparameters via empirical Bayes. These advantages make Bayesian quadrature particularly appealing when (1) function evaluations are expensive or scarce, since accurate estimates can be achieved with relatively few samples; (2) there is prior information available about the integrand that can be incorporated into the Gaussian process; and (3) uncertainty quantification is important for reliability of the estimate.

\section{Challenges and contributions}
\label{sec:challenges_and_contributions}

While the MMD is a widely used tool for comparing probability distributions, applying it effectively in practice can be surprisingly subtle. Its performance depends on both the structure of the problem and the choice of kernel, and naïve application may lead to inefficiencies or failures in capturing important distributional features. We now highlight two key practical challenges.

\paragraph{Challenge 1: Specialising MMD-based methods.}
In structured settings, such as simulation-based inference and conditional or multi-task integration, generic, task-agnostic uses of MMD are often outperformed by estimators that make use of problem structure. For example, for smooth simulators, kernel mean embedding integrals can be estimated more efficiently than with the V-statistic; similarly, methods modelling dependencies across integrands outperform MMD-minimising quadrature on each integral in isolation~\citep{Niu2023,gessner2020active}. These observations motivate tailored, structure-aware MMD estimators that are more data-efficient in such settings.

\paragraph{Challenge 2: Alternative kernel-based distances.} 
When the kernel $k$ is not characteristic, so that $\MMD_k(\bP,\bQ)=0$ does not imply $\bP=\bQ$, the kernel mean embedding fails to uniquely determine the distribution, limiting the reliability of MMD for distribution comparison. Further, even when $k$ is characteristic, alternative kernel-based discrepancies that probe information beyond the kernel mean may capture distributional differences more efficiently in the finite-sample regime.

These challenges motivate the core contributions of the thesis. Chapter~2 provides the necessary technical background and tools, introducing kernel methods, MMD-based numerical integration, and function smoothness. The remaining chapters address the two challenges in turn.

\begin{itemize}
\item In \textbf{\Cref{sec:part1}}, we focus on the first challenge: improving MMD estimators in structured settings. In~\Cref{sec:mmdsbi}, we propose an \emph{optimally-weighted} estimator of the MMD in simulation-based inference, by making use of Bayesian quadrature, an MMD-minimising integration method, to improve estimation of kernel means. Then, in~\Cref{sec:mmdintegration}, we propose conditional Bayesian quadrature, an MMD-based estimator extending Bayesian quadrature to computing conditional expectations, a task that arises frequently in statistical computation. Lastly, in~\Cref{sec:gpcv}, we turn to the issue of calibration in Bayesian quadrature, specifically amplitude hyperparameter selection and its impact on uncertainty quantification.
\item  In \textbf{\Cref{sec:part2}}, we move beyond the MMD and consider alternative general-purpose kernel-based discrepancies. Motivated by recent advances in kernel embeddings beyond the mean, we introduce a new family of kernel-based statistical distances, the kernel quantile discrepancies. These discrepancies mitigate some of the shortcomings of MMD, and we show through both theoretical analysis and empirical study that they offer a competitive alternative to MMD and its fast approximations.
\end{itemize}

\chapter{Background}
\label{sec:background}
In the previous chapter, we briefly surveyed discrepancies between probability distributions and outlined challenges addressed in this thesis. This chapter introduces key concepts and tools needed for our results, and sets the notation used in later chapters. First, however, a clarifying note on fundamental terminology.

\paragraph{Discrepancy vs. divergence vs. distance.} 
Let $\calP$ be some space of probability distributions, and let $\bP,\bQ,\bP'$ be some distributions in $\calP$.
In modern statistics and machine learning, a \emph{discrepancy} is any non-negative function $D : \calP \times \calP \to [0,\infty]$ intended to quantify distributional dissimilarity; it is typically required that $D(\bP,\bP)=0$. A \emph{divergence} is a discrepancy that also satisfies the \emph{identity of indiscernibles}, i.e., $D(\bP,\bQ)=0$ if and only if $\bP=\bQ$. A \emph{distance} is the strongest notion of dissimilarity: it is a divergence that is finite, symmetric, $D(\bP,\bQ)=D(\bQ,\bP)$, and satisfies the triangle inequality, $D(\bP,\bQ) \leq D(\bP,\bP') + D(\bP',\bQ)$, so that $D$ is a \emph{metric} on $\calP$. 
Terminology in the literature is not always consistent, owing to historical usage or notational simplifications. For example, the Bhattacharyya distance is, strictly speaking, only a divergence, and integral probability pseudometrics (IPMs) are conventionally called `metrics' despite being merely pseudometrics unless the function class is rich enough to guarantee the identity of indiscernibles. Moreover, whether a given discrepancy is a distance or a divergence may depend on the choice of domain $\calP$: for instance, the $p$-Wasserstein distance $W_p$ for $p \geq 1$ is a distance only on the space of probability distributions with finite $p$-th moment. In what follows, we will take care to keep these distinctions explicit. 

In the rest of the chapter, we introduce core concepts in kernel and reproducing kernel Hilbert space theory, as well as more specific details necessary to introduce the setting of our results. Throughout the work, $\calX$ denotes the input space or set; in the most general setting, $\calX$ is a set, and in the most specific setting, $\calX$ is a subset of $\bR^d$. We clarify the restrictions on $\calX$ throughout.

We aim to define all notation before using it, and give frequent explanations of its meaning in text. However, to aid readability, we give a short summary of notation used in this work in \Cref{tab:notation}. 

\begin{table}[!t]
\renewcommand{\arraystretch}{1.3}
\begin{tabularx}{\textwidth}{@{}p{0.35\textwidth}X@{}}
\toprule
  $\bN$ & The set of natural numbers, $\{0, 1, \dots\}$. \\
  $\bN_{\geq a}$ & The subset $\{a, a+1, \dots\} \subseteq \bN$, for $a \in \bN$. \\
  $\Id_d$ & The $d \times d$ identity matrix. \\
  $x_{1:N} \in \calX^N$ & Tuple $(x_1, x_2, \dots, x_N)$, for $x_1, \dots, x_N \in \calX$. \\
  $x_{1:N} \in \bR^{N \times d}$ & Vector (for $d=1$) or matrix $\begin{bmatrix} x_1 & x_2 & \dots & x_N \end{bmatrix}^\top$, for $x_1, \dots, x_N \in \calX \subseteq \bR^d$. \\
  $\| \cdot \|_{\calF}$, $\langle \cdot, \cdot \rangle_{\calF}$ & The norm and inner product of a space $\calF$. \\
  $f(x_{1:N}) \in \bR^N$ & Column vector $\begin{bmatrix} f(x_1) & f(x_2) & \dots & f(x_N) \end{bmatrix}^\top$, for $x_1, \dots, x_N \in \calX$ and $f: \calX \to \bR$. \\
  $k(x_{1:N}, x_{1:N}) \in \bR^{N \times N}$ & Gram matrix $\begin{bmatrix}
        k(x_1, x_1) & \dots & k(x_1, x_N) \\
        \vdots & \ddots & \vdots \\
        k(x_N, x_1) & \dots & k(x_N, x_N)
    \end{bmatrix}$, for $x_1, \dots, x_N \in \calX$ and $k: \calX \times \calX \to \bR$. \\
  $\|\cdot\|_2$ & Euclidean norm. For $a \in \bR^d$, $\|a\|_2 = \sqrt{\sum_{i=1}^d a_i^2}$. \\
  $\calH_k$ & RKHS induced by a kernel $k$ (see~\Cref{def:rkhs}). \\
  $\mu_{k,\bP} \in \calH_k$ & Kernel mean embedding of $\bP$ in $\calH_k$ (see~\Cref{def:kme}). \\
  $C^s(\calX)$ & Space of $s$-times continuously differentiable functions on $\calX$ (see~\Cref{def:cs_space}). \\
  $C^{s,\alpha}(\calX)$ & H\"older space (see~\Cref{def:hoelder_space}). \\
  $\calL^p(\calX, \bP)$ & $\calL^p$ space, for a measure $\bP$ on $\calX$ (see~\Cref{def:lp_space}). \\
  $\calL^p(\calX)$ & $\calL^p$ space, for the Lebesgue measure $\mu$ on $\calX$ (see~\Cref{def:lp_space}). \\
  $\calW^{s, p}(\calX)$ & Sobolev space (see~\Cref{def:sobolev_space}). \\
  $X,X' \sim \calN(\mu, \Sigma)$ & $X,X'$ are independent random variables; the law of each is the Gaussian distribution with mean $\mu$ and covariance $\Sigma$.\\
  $X \sim \nu$ & $X$ is a random variable the law of which is the measure $\nu$.\\
  $\bP_N = \frac{1}{N} \sum_{n=1}^N \delta_{x_n}$ & An equally-weighted empirical measure, $x_1, \dots, x_N \sim \bP$.\\
  $\bP^w_N = \sum_{n=1}^N w_n \delta_{x_n}$ & A weighted empirical measure, $x_1, \dots, x_N \sim \bP$.\\
\bottomrule
\end{tabularx}
\caption{Summary of notation used in the thesis.}
\label{tab:notation} 
\end{table}

\section{Kernels and reproducing kernel Hilbert spaces}

Recall that a symmetric $N \times N$ matrix $k(x_{1:N}, x_{1:N})$ is said to be positive definite if for any vector $\alpha_{1:N} \in \bR^N$, it holds that $\alpha_{1:N}^\top k(x_{1:N}, x_{1:N})\alpha_{1:N} \geq 0$.

\begin{definition}[Positive definite kernel]
\label{def:kernel_pd}
    A symmetric function $k:\calX \times \calX \to \bR$ on a set $\calX$ is a \emph{positive definite} kernel if the Gram matrix~\citep[Section 2]{Berlinet2004}
    \begin{equation*}
        k(x_{1:N}, x_{1:N}) \coloneqq \begin{bmatrix}
            k(x_1, x_1) & \dots & k(x_1, x_N) \\
            \vdots & \ddots & \vdots \\
            k(x_N, x_1) & \dots & k(x_N, x_N)
        \end{bmatrix}
    \end{equation*}
    is positive definite, for any $N \in \bN_{\geq 1}$ and any $x_{1:N} \in \calX^N$.
\end{definition}

We will simply refer to positive definite kernels as `kernels'. 
When $\alpha_{1:N}^\top k(x_{1:N}, x_{1:N})\alpha_{1:N} = 0$ for mutually distinct $x_{1:N} \in \calX^N$ implies $\alpha_1=\dots=\alpha_N=0$, the kernel is said to be \emph{strictly positive definite}. Commonly used kernels, such as the Gaussian kernel and the Mat\'ern family of kernels introduced below, are strictly positive definite. 

\begin{remark}
    The terminology in the literature can be inconsistent, with kernels as defined in~\Cref{def:kernel_pd} referred to as positive \emph{semi}definite, and reserving the term positive definite for kernels that we call strictly positive definite.
\end{remark}

The concepts and results introduced in this chapter, and the rest of the thesis, hold for kernels as defined in~\Cref{def:kernel_pd}; whenever a distinction with strictly positive definite kernels is helpful, we will make it clear.

\subsection{Examples and basic properties of kernels}
\label{sec:properties_of_kernels}

The kernel choice is central in kernel-based discrepancies. First, the kernel determines general task-agnostic properties: whether the discrepancy is a distance~\citep{sriperumbudur2011universality}, the topology it induces on probability distributions~\citep{simon2023metrizing,barp2024targeted}, and what differences between the distributions it emphasises~\citep{rahimi2007random,sriperumbudur2015optimal}. Second, the kernel may drive task-specific performance: as discussed in~\Cref{sec:kernel_and_bayesian_quadrature}, MMD-minimising integration can be viewed as Gaussian process regression with kernel $k$ on the integrand $f$; adapting $k$ to match properties of $f$ accelerates convergence and improves uncertainty calibration \citep{kanagawa2020convergence,wynne2021convergence}.

\Cref{tab:example_kernels} lists commonly used kernels that are invoked in this thesis. The \emph{amplitude parameter} $\tau^2>0$ scales the output of the kernel, and is instrumental in uncertainty quantification; this is discussed further in~\Cref{sec:gpcv}. The lengthscale parameter $l>0$ in the Mat\'ern and Gaussian kernels scales the input, as the distance $\|x -x' \|_2$ gets scaled by $l$.

\paragraph{Mat\'ern kernels} are of particular interest: the \emph{order} parameter $\nu$ determines the smoothness of $k_\nu$; this will be formalised further in~\Cref{sec:bg_sobolev_kernels}. Further, for half-integer orders $\nu \in \{\nicefrac{1}{2}, \nicefrac{3}{2}, \dots\}$, the kernel $k_\nu$ takes a convenient form: it is a product of an exponential and a polynomial of degree $\lfloor \nu \rfloor$~\citep{rassmussen2006gaussian}. For example, for the first three half-integers, and $\rho=\| x- x'\|_2$,
\begin{align*}
    \nu=\nicefrac{1}{2}&:\quad k_\nu(x, x') = \tau^2 \exp \left( - \frac{\rho}{l} \right)\\
    \nu=\nicefrac{3}{2}&:\quad k_\nu(x, x') = \tau^2 \left( 1 + \frac{\sqrt 3 \rho}{l} \right) \exp \left( - \frac{\sqrt 3 \rho}{l} \right)\\
    \nu=\nicefrac{5}{2}&:\quad k_\nu(x, x') = \tau^2 \left( 1 + \frac{\sqrt 5 \rho}{l} + \frac{ 5 \rho^2}{3 l^2} \right) \exp \left( - \frac{\sqrt 5 \rho}{l} \right) \\
\end{align*}
At the limit $\nu \to \infty$, $k_\nu$ goes to the infinitely smooth Gaussian kernel $k_\mathrm{Gauss}$.

\paragraph{Polynomial kernels} illustrate two phenomena. First, a polynomial kernel is \emph{not} strictly positive definite: even more strongly, a Gram matrix for \emph{any} pairwise distinct $x_{1:N}$ for a large enough $N$ is not strictly positive definite~\citep[Section 4.2.2]{rassmussen2006gaussian}. Second, the MMD with the polynomial kernel is not a distance on the space of Borel probability measures; this example is instrumental in proofs in~\Cref{sec:kqe}.

\paragraph{The Brownian motion kernel} is well-studied~\citep{morters2010brownian} and is integral to many fields of study, from calculus on general stochastic processes to modelling physics and biological phenomena. The fractional Brownian motion kernel $k_{0,H}$~\citep[Chapter IX]{mandelbrot1982fractal}, with its smoothness parameterised by the Hurst parameter $H$, generalises the Brownian motion kernel: for $H>1/2$, the fractional Brownian motion kernel is smoother than the regular Brownian motion kernel; for $H<1/2$ it is less smooth. It is easy to see that $k_{0,1/2}$ is the Brownian motion kernel $k_\mathrm{BM}$. We make `smoothness' precise and discuss the properties of the (fractional) Brownian motion kernel in~\Cref{sec:fbm}.

All kernels in~\Cref{tab:example_kernels}, except for the polynomial ones, are strictly positive definite; this property is central to MMD-minimising integration in~\Cref{sec:mmdintegration}. The Mat\'ern and Gaussian kernels additionally satisfy two important properties, boundedness and stationarity, which we now define and discuss.
\begin{definition}
    A kernel $k$ is said to be bounded if there exists a constant $B>0$ such that $k(x, x) \leq B$ for any $x \in \calX$.
\end{definition}
In fact, this is equivalent to $|k(x, x')| \leq B$ for all pairs $x, x' \in \calX$: by $k$ being positive definite, it holds that $k(x, x) - 2 k(x, x') + k(x', x') \geq 0$ and $k(x, x) + 2 k(x, x') + k(x', x') \geq 0$, with equality when $x=x'$; therefore
\begin{align*}
    |k(x, x')| \leq ( k(x, x) + k(x', x') ) / 2 \leq B.
\end{align*}
\begin{definition}
    A kernel $k$ is said to be \emph{stationary}, translation-invariant, or shift-invariant, if it only depends on the difference $x-x'$, i.e., there is a function $k_0$ such that $k(x, x')=k_0(x-x')$. 
\end{definition}
A stationary kernel $k$ is automatically bounded by $B = k_0(0)$; for Mat\'ern and Gaussian kernels, $k_0(0)=\tau^2$. 
Both are important properties in the context of MMD. Boundedness (and measurability) of $k$ is sufficient for $\MMD_k(\bP,\bQ)$ to be well defined and finite for all Borel probability measures $\bP,\bQ$. Further, if $k$ is continuous and stationary on $\calX=\bR^d$, then $\MMD_k$ is a \emph{distance} on the set of Borel probability measures if and only if the support of the Fourier transform of $k_0$ is the entire $\bR^d$; this condition holds for both Mat\'ern and Gaussian kernels~\citep{Sriperumbudur2009}. We return to these points, and their role in our results, in \Cref{sec:kernel_mean_embeddings_and_mmd}.

\begin{table}[!t]
\begin{tabularx}{\textwidth}{@{}p{0.35\textwidth}X@{}}
\toprule
  Polynomial of degree $q$ & $k_{\mathrm{Poly}}(x, x') = \tau^2 (x^\top x' + c)^q$, for $c \geq 0$ and $q \in \bN$.\\
  Mat\'ern of order $\nu$, and smoothness $s=\nu+d/2$ & $k_\nu(x, x') = \frac{\tau^2}{\Gamma(\nu)2^{\nu - 1}} (\frac{\sqrt{2 \nu}}{l} \| x - x' \|_2 )^\nu K_\nu(\frac{\sqrt{2 \nu}}{l} \| x - x' \|_2)$, where $K_\nu$ for $\nu>0$ is the modified Bessel function of the second kind. \\
  Gaussian & $k_\mathrm{Gauss}(x, x') = \tau^2 \exp(-\|x - x'\|^2_2/(2l^2) )$ \\
  Brownian motion & $k_\mathrm{BM}(x, x') = \tau^2 \min(x, x')$ \\
  Fractional Brownian motion & $k_{0,H}(x, x') = \tau^2 \big( x ^{2H} + {x'}^{2H} - \lvert x-x'\rvert^{2H}\big)/2$ \\
\bottomrule
\end{tabularx}
\caption[Example kernels.]{Example kernels on $\calX=\bR^d$. For Brownian motion and fractional Brownian motion, $\calX=[0, \infty)$.}
\label{tab:example_kernels}
\end{table}

\subsection{Reproducing kernel Hilbert spaces}

Every positive definite kernel $k:\calX \times \calX \to \bR$ defines a unique reproducing kernel Hilbert space $\calH_k$ of functions $\calX \to \bR$. This one-to-one mapping is powerful: it lets us work with rich, often infinite-dimensional, function spaces by manipulating only the kernel; for example, properties of $k$, such as smoothness, boundedness, and periodicity, are inherited by functions in the RKHS. Before giving the formal definition of an RKHS, we briefly review the construction of $\calH_k$ from $k$; this construction illuminates the definition and underlies the proof that the map $k \mapsto \calH_k$ is one-to-one. For an in-depth treatment of RKHS theory, we refer the reader to~\citet{Berlinet2004}.

\begin{remark}[Equivalent definitions of an RKHS]
There are multiple equivalent definitions of an RKHS: (1) as a Hilbert space with continuous evaluation functionals; (2) as a closure of the span of $k(x, \cdot)$; (3) as a Hilbert space with a reproducing kernel. We use (3) as the primary definition, review (2), and present the equivalence between (2) and (3) as a theorem~\citep[Moore–Aronszajn theorem]{Berlinet2004}. Lastly, though we do not review it, the equivalence between (1) and (3) is a corollary of the Riesz representation theorem~\citep[Theorem 1]{Berlinet2004}.
\end{remark}
Consider a space $\calH_k^0=\spn{k(x, \cdot) \given x \in \calX}$ consisting of all functions $f:\calX \to \bR$ of the form $f(x)=\sum_{n=1}^N a_n k(x_n, x)$ for any $N \in \bN_{\geq 1}$ and $a_{1:N} \in \bR^N$, $x_{1:N} \in \calX^N$.
Define a function $\langle \cdot, \cdot \rangle_{\calH_k^0}: \calH_k^0 \times \calH_k^0 \to \bR$ as $\langle f, g \rangle_{\calH_k^0}=\sum_{n=1}^N \sum_{m=1}^M a_n b_m k(x_n, y_m)$ for any $f = \sum_{n=1}^N a_n k(x_n, \cdot)$ and $g = \sum_{m=1}^M b_m k(y_m, \cdot)$. It is easy to show $\langle \cdot, \cdot \rangle_{\calH_k^0}$ is an inner product on $\calH_k^0$, and for any $f \in \calH_k^0$ and $x \in \calX$ it holds that
\begin{equation}
\label{eq:pre-reproducing-property}
    \langle f, k(x, \cdot) \rangle_{\calH_k^0} = f(x).
\end{equation}
Under the metric induced by $\langle \cdot, \cdot \rangle_{\calH_k^0}$, the space $\calH_k^0$ need not be complete, meaning there may be a Cauchy sequence $\{f_n\}_{n \in \bN}$ that does not have a limit in $\calH_k^0$. We define $\calH_k$, the \emph{reproducing kernel Hilbert space} induced by $k$, as the completion of $\calH_k^0$. For every $f \in \calH_k$, there will be a Cauchy sequence $\{f_n\}_{n \in \bN} \in \calH_k^0$ that pointwise converges to $f$; this allows us to define
\begin{equation}
    \langle f, g \rangle_{\calH_k} = \lim_{n \to \infty} \langle f_n, g_n \rangle_{\calH_k^0},
\end{equation}
which can be shown to be an inner product on $\calH_k$. The space $\calH_k$ with this inner product is complete, making it a Hilbert space. Finally, by continuity the property in~\eqref{eq:pre-reproducing-property} extends to the entire $\calH_k$,
\begin{equation}
\label{eq:reproducing-property}
    \langle f, k(x, \cdot) \rangle_{\calH_k} = f(x). 
\end{equation}
The property in~\eqref{eq:reproducing-property}, called the \emph{reproducing property}, is the key reason RKHSs are convenient to work with. We are now ready to give a non-constructive definition of an RKHS.

\begin{definition}[Reproducing kernel Hilbert spaces]
\label{def:rkhs}
    Let $\calH_k$ be a Hilbert space of functions $f:\calX \to \bR$, with inner product $\langle \cdot, \cdot \rangle_{\calH_k}$. $\calH_k$ is said to be a reproducing kernel Hilbert space if there is a function $k:\calX \times \calX \to \bR$ such that
    \begin{enumerate}
        \item $k(x, \cdot) \in \calH_k$ for all $x \in \calX$,
        \item $\langle f, k(x, \cdot) \rangle_{\calH_k} = f(x)$ for any $f \in \calH_k$. \hfill \textit{(the reproducing property)}
    \end{enumerate}
    Such $k$ is said to induce $\calH_k$, and is called a reproducing kernel for $\calH_k$.
\end{definition}

Every reproducing kernel is positive definite~\citep[Lemma 2]{Berlinet2004}. Moreover, the correspondence between kernels and RKHSs is one-to-one: every kernel $k$ induces a unique $\calH_k$, and the reproducing kernel of $\calH_k$ is necessarily $k$ by the Moore–Aronszajn theorem~\citep[Theorem 3]{Berlinet2004}. The proof uses the construction of $\calH_k$ from $k$ demonstrated above.

\begin{remark}
    Identifying what types of functions lie in an RKHS induced by a particular kernel is non-trivial, due to a somewhat opaque construction of $\calH_k$ from $k$ covered above. Completion of the space $\calH_k^0$, in particular, should be expected to greatly increase the size of the space, similar to how completing the set of all rationals produces the much larger set of real numbers. Fortunately, some kernels, notably the Mat\'ern family of kernels, have been shown to induce RKHSs that coincide with Sobolev spaces, the better-understood function spaces, with a norm equivalent to the RKHS norm. We cover Sobolev spaces and their connection to the RKHS induced by Mat\'ern kernels in~\Cref{sec:quantifying_smoothness_of_functions}. Further, though we do not cover it in this thesis, Mercer's theorem provides a series representation for functions in the RKHS for continuous kernels on compact domains~\citep[Theorem 4.51]{Steinwart2008-rf} and in certain more general cases~\citep{Steinwart2012-cj}.
\end{remark}

The following section covers a key matter that makes RKHSs useful in the context of this work: kernel mean embeddings of distributions.

\subsection{Kernel mean embeddings and MMD}
\label{sec:kernel_mean_embeddings_and_mmd}

Let $\calP(\calX)$ denote the set of Borel probability measures on a Borel space $\calX$, and $(\calH_k, \langle \cdot, \cdot \rangle_{\calH_k})$ be a reproducing kernel Hilbert space (RKHS) induced by a real-valued kernel $k: \calX \times \calX \to \bR$. Throughout, we will assume $k$ is measurable. 
\begin{definition}[Kernel mean embedding]
\label{def:kme}
Let $\calP_k(\calX) \subseteq \calP(\calX)$ denote the set
\begin{equation*}
    \calP_k(\calX) = \left\{\bP \in \calP(\calX): \int_\calX \sqrt{k(x, x)} \bP(\d x) < \infty \right\}.
\end{equation*}
Then, for any $\bP \in \calP_k(\calX)$, the Bochner integral~\citep{schwabik2005topics}
\begin{equation*}
    \mu_{k,\bP}(\cdot)=\int_\calX k(\cdot, x') \bP(\d x')
\end{equation*}
is called a \emph{kernel mean embedding} (KME) $\mu_{k,\bP} \in \calH_k$ of $\bP$.
\end{definition}
This definition cannot be extended beyond $\bP \in \calP_k(\calX)$: by~\citep[Theorem 1.4.3]{schwabik2005topics}, the Bochner integral defining the KME exists and is finite if and only if $\bP \in \calP_k(\calX)$. Therefore, KMEs are defined for all Borel probability measures, $\bP \in \calP(\calX)$, if and only if $\calP_k(\calX)=\calP(\calX)$, which in turn holds if and only if $k$ is bounded~\citep[Proposition 2]{Sriperumbudur2009}. Informally speaking, when $k$ is unbounded, $\calP_k(\calX)$ only contains $\bP$ that assign sufficiently small mass to regions of $\calX$ in which $\sqrt{k(x, x)}$ is large.

As we will see in the later chapters, applications of MMD rely on KMEs being able to identify distributions: distinct $\bP \neq \bQ$ should map to distinct embeddings $\mu_{k,\bP} \neq \mu_{k,\bQ}$. This property is formalised by \emph{characteristic} kernels.
\begin{definition}[(Mean-)characteristic kernel]
    The kernel $k$ is said to be \emph{characteristic} on $\calX$ if the mapping $\bP \mapsto \mu_{k,\bP}$ is injective for all $\bP \in \calP_k(\calX)$.
\end{definition}
As discussed in~\Cref{sec:properties_of_kernels}, both the Mat\'ern family and the Gaussian kernels have been shown to be characteristic on $\bR^d$~\citep{Sriperumbudur2009,ziegel2022characteristic}. 
In general, beyond translation-invariant kernels on Euclidean spaces, this property is challenging to establish; in~\Cref{sec:kqe} we will introduce alternative kernel embeddings with their own notion of characteristic that is easier to verify, and use the `mean-' prefix above to distinguish the two.

\begin{definition}[Maximum Mean Discrepancy]
The \emph{maximum mean discrepancy} (MMD) is a discrepancy between $\bP$ and $\bQ$ defined as the distance between their kernel mean embeddings in $\calH_k$,
\begin{align}\label{eq:MMD_withembeddings}
    \MMD_k(\bP, \bQ) = \| \mu_{k,\bP} - \mu_{k,\bQ} \|_{\calH_k}.
\end{align}
\end{definition}
Naturally, MMD is a distance on $\calP_k(\calX)$ whenever $k$ is characteristic on $\calX$, i.e., $\MMD_k(\bP, \bQ) = 0$ if and only if $\bP = \bQ$. 
The MMD can be computed exactly in rare cases \citep{pmlr-v271-briol25a}, but typically needs to be estimated. Using the fact that inner product commutes with Bochner integration~\citep[Proposition E.11]{cohn2013measure} and the reproducing property, specifically
\begin{equation*}
\begin{aligned}
     \langle \mu_{k,\bP} , \mu_{k,\bQ} \rangle_{\calH_k} &= \left\langle \int_\calX k(\cdot, x) \bP(\d x) , \int_\calX k(\cdot, y) \bQ(\d y) \right\rangle_{\calH_k} \\
     &= \int_\calX \int_\calX \left\langle  k(\cdot, x)  , \ k(\cdot, y)  \right\rangle_{\calH_k} \bP(\d x) \bQ(\d y) \\
     &= \int_\calX \int_\calX  k(x, y) \bP(\d x) \bQ(\d y),
\end{aligned}
\end{equation*}
we can write
\begin{equation}
\label{eq:MMD2_exact}
\begin{aligned}
     \MMD_k^2(\bP,\bQ) &= \| \mu_{k,\bP} - \mu_{k,\bQ} \|^2_{\calH_k} \\
     &= \langle \mu_{k,\bP} - \mu_{k,\bQ}, \mu_{k,\bP} - \mu_{k,\bQ} \rangle_{\calH_k} \\
     &= \langle \mu_{k,\bP} , \mu_{k,\bP} \rangle_{\calH_k} - 2\langle \mu_{k,\bP} , \mu_{k,\bQ} \rangle_{\calH_k} + \langle \mu_{k,\bQ}, \mu_{k,\bQ} \rangle_{\calH_k} \\
     &=\int_{\calX} \int_{\calX}  k(x,y) \bP(\d x) \bP(\d y) - 2 \int_{\calX} \int_{\calX}  k(x,y) \bP(\d x) \bQ(\d y) \nonumber \\
     &\hspace{5cm}+ \int_{\calX} \int_{\calX}  k(x,y) \bQ(\d x) \bQ(\d y).
\end{aligned}
\end{equation}
This expression is convenient to work with as it can be estimated through approximations of the integrals. Let $x_1, \dots, x_N \sim \bP$, $y_1, \dots, y_M \sim \bQ$ and let $\bP_N = \nicefrac{1}{N} \sum_{n=1}^N \delta_{x_n}$ and $\bQ_M = \nicefrac{1}{M} \sum_{m=1}^M \delta_{y_m}$, where $\delta_x$ is a Dirac measure at $x$. The squared MMD can be approximated through a V-statistic as
\begin{align*}
    \MMD_k^2(\bP_N,\bQ_M) = \frac{1}{N^2} \sum_{n, n' = 1}^N k(x_n, x_{n'}) &- \frac{2}{NM} \sum_{n=1}^N \sum_{m=1}^M k(x_n, y_m) \\
    &+  \frac{1}{M^2} \sum_{m, m' = 1}^M k(y_m, y_{m'}).\nonumber
\end{align*}
This is equivalent to approximating the integral $\mu_{k,\bP}(x)$ with $\nicefrac{1}{N} \sum_{n=1}^N k(x, x_n)$.
Alternatively, we can use an unbiased U-statistic approximation \citep{gretton2012kernel}.
Both of these estimates can be calculated straightforwardly via evaluations of the kernel $k$ at a computational cost~$\bigo( \max(N, M)^2)$, and converge to $\MMD_k^2(\bP, \bQ)$ at the standard rate $\bigo(\min(N, M)^{-\nicefrac{1}{2}})$.

\paragraph{Specifying the kernel.} Whenever there is only one kernel $k$ used within a chapter, we will simplify the notation to omit the subscript $k$, and write
\begin{equation*}
    \calH \coloneqq \calH_k,\qquad \mu_\bP \coloneqq \mu_{k, \bP},\qquad \MMD \coloneqq \MMD_k.
\end{equation*}

\section{MMD-minimising numerical integration}
\label{sec:kernel_and_bayesian_quadrature}

Of the tasks MMD is applied to, numerical integration deserves a special mention. It is the unique setting where the discrepancy objective is the numerical error: in an RKHS, the worst-case quadrature error equals the MMD between the weighted empirical measure and $\bP$. It is central to this thesis, re-appearing in three out of the four core chapters: Chapter~3 extends it to conditional expectations; Chapter~4 uses it to improve kernel mean estimation and thus MMD itself in simulation-based inference; Chapter~5 studies its hyperparameter calibration and its impact on uncertainty quantification. 

Let $\bP$ be a probability measure on a Borel space $(\calX, \calF_\calX)$, $f$ be a measurable real-valued function on $\calX$, and $x_1, \dots, x_N$ be points in $\calX$. In numerical integration, the value of an intractable integral $I = \int_\calX f(x) \bP(\d x)$ is approximated with a weighted sum, or a \emph{quadrature rule}~\citep{davis2007methods},
\begin{equation*}
    \hat I = \sum_{n=1}^N w_n f(x_n).
\end{equation*}
Numerical integration is ubiquitous whenever expectations or definite integrals lack closed forms: in computational statistics and machine learning (for example, marginal likelihoods for model selection), finance and economics, and across the sciences and engineering (such as expected outcomes of experiments or simulations). The problem becomes harder in high dimensions, for rough/low-smoothness integrands $f$, and when $f$ is expensive to evaluate~\citep{novak2008tractability}. 
For general-purpose use, the most common approach is \emph{Monte Carlo} methods~\citep{Robert2004,mcbook}; standard Monte Carlo takes the form
\begin{equation*}
    I_\mathrm{MC} = \frac{1}{N} \sum_{n=1}^N f(x_n), \text{ for } x_1,\dots,x_N \sim \bP,
\end{equation*}
which converges at rate $\mathcal{O}(N^{-1/2})$ under mild conditions. When exact sampling from $\bP$ is not available, approximate samplers are used~\citep{neal2011mcmc,doucet20011tutorial}. This approach ignores the geometry of the sample locations and any structure of $f$, and can therefore require many evaluations. Alternative schemes re-weigh the points, to produce quadrature rules which are equivalent to integrating against a weighted empirical measure,
\begin{equation*}
    \hat I = \int_\calX f(x) \bP^w_N(\d x), \qquad \bP^w_N = \sum_{n=1}^N w_n \delta_{x_n}.
\end{equation*}
Thus, whenever $\bP^w_N$ closely approximates $\bP$, $\hat I$ provides a good estimate of $I$. In kernel quadrature~\citep{sommariva2006numerical,dick2013high,bach2017equivalence,kanagawa2020convergence}, the weights are set so that $\bP^w_N = \sum_{n=1}^N w_n \delta_{x_n}$ is closest to $\bP$ in MMD,
\begin{equation*}
    I_\mathrm{KQ} = \sum_{n=1}^N w^\mathrm{KQ}_n f(x_n), \qquad w^\mathrm{KQ}_{1:N} = \argmin_{w_{1:N}} \MMD_k \left( \sum_{n=1}^N w_n \delta_{x_n}, \bP \right).
\end{equation*}
When the Gram matrix $k(x_{1:N}, x_{1:N})$ is invertible, which is guaranteed whenever the kernel is strictly positive definite and the points $x_{1:N}$ are pairwise distinct, it holds that the weights $w^\mathrm{KQ}_{1:N}$ are unique, and $I_\mathrm{KQ}$ and $\min_{w_{1:N}} \MMD_k \left( \sum_{n=1}^N w_n \delta_{x_n}, \bP \right)$ are the posterior mean and standard deviation of a zero-mean \emph{Gaussian process} conditioned on $(x_{1:N}, f(x_{1:N}))$ \citep[Section 7.2.2]{kanagawa2025gaussian}. This leads naturally to a probabilistic interpretation: Bayesian quadrature (BQ). To define it formally, we first introduce Gaussian process regression.

\subsection{Gaussian process regression.}
\label{sec:gp-regression}

Let $\calX$ be a set. A stochastic process $\{f_\GP(x)\}_{x \in \calX}$ is a Gaussian process with mean $m: \calX \to \bR$ and kernel $k: \calX \times \calX \to \bR$, written $f_\GP \sim \GP(m, k)$, if for any $N \in \bN_{\geq 1}$ and any $x_{1:N} \in \calX^N$,
\begin{equation*}
    f_\GP(x_{1:N}) \sim \calN(m(x_{1:N}), k(x_{1:N}, x_{1:N})),
\end{equation*}
i.e., the $N$-dimensional random variable $f_\GP(x_{1:N})$ follows a multivariate Gaussian distribution with mean $m(x_{1:N})$ and covariance equal to the kernel Gram matrix $k(x_{1:N}, x_{1:N})$.
Consider the task of approximating an unknown function $f: \calX \to \bR$ given $N$ (possibly noisy) observations $y_1, \dots, y_N$ at input points $x_1, \dots, x_N$ in $\calX$, with data-generating process assumed to be
\begin{align*}
    y_n = f(x_n) + \varepsilon_n, \qquad \text{for all } n \in \{1,\ldots,N\},
\end{align*}
where $\varepsilon_1, \dots, \varepsilon_N$ is the observation noise. In Gaussian process (GP) regression (or kriging, \citet{OHagan1978, Stein1999, rassmussen2006gaussian}), the function $f$ is assigned a Gaussian process prior $\GP(m, k)$. When the observations are assumed to be noise-free, i.e., $\varepsilon_1 = \dots = \varepsilon_N = 0$ almost surely, the setting is known as Gaussian process \emph{interpolation}. 
\begin{remark}
\label{rem:gp_zero_mean}
    The GP prior is often taken to be zero-mean, $m(x) =0$ for all $x \in \calX$. This is not restrictive as the data $y_{1:N}$ can be centered, and any additional prior information about the function can be encoded in the kernel rather than the prior mean. In this case, the kernel $k$ determines the properties, such as the smoothness and correlation length, of the GP prior.
\end{remark}
For $\varepsilon_1, \dots, \varepsilon_N$ assumed to be independently sampled from Gaussians, $\varepsilon_n \sim \calN(0, \sigma_{\varepsilon_n}^2)$ with $\sigma_{\varepsilon_n}>0$ for $n \in \{1, \dots, N\}$, the posterior process conditioned on $(x_{1:N}, y_{1:N})$ is again a GP with the posterior mean $m_N$ and covariance $k_N$,
\begin{equation}
\label{eq:posterior-moments}
  \begin{aligned}
    &m_N(x)  = m(x) + k(x, x_{1:N})^\top \left( k(x_{1:N}, x_{1:N}) + \diag(\sigma_{\varepsilon_{1:N}}^2) \right)^{-1} \left(y_{1:N} - m(x_{1:N})\right), \\
    &k_N(x,x') = k(x,x') - k(x, x_{1:N})^\top \left( k(x_{1:N}, x_{1:N}) + \diag(\sigma_{\varepsilon_{1:N}}^2) \right)^{-1} k(x_{1:N},x'),
\end{aligned}
\end{equation}
where $\diag(\sigma_{\varepsilon_{1:N}}^2)$ is the $N \times N$ diagonal matrix with $\sigma_{\varepsilon_1}^2, \dots, \sigma_{\varepsilon_N}^2$ on the diagonal. GP regression is said to be \emph{homoscedastic} when the noise is the same for all observations, $\sigma_{\varepsilon_1}=\dots=\sigma_{\varepsilon_N}$, and \emph{heteroscedastic} otherwise. Homoscedastic regression is the default in the literature: unless otherwise specified, `GP regression' refers to homoscedastic GP regression. 
Although the posterior is closed form, standard GP inference scales as $\bigo(N^3)$ because it relies
on a Cholesky factorisation of $k(x_{1:N}, x_{1:N}) + \diag(\sigma_{\varepsilon_{1:N}}^2)$~\citep[Algorithm 2.1]{rassmussen2006gaussian}. The implications of this for BQ will be explored in~\Cref{sec:bq_background}, and~\Cref{sec:mmdsbi,sec:mmdintegration}.
Provided $k(x_{1:N}, x_{1:N})$ is invertible, setting $\sigma_{\varepsilon_1}=\dots=\sigma_{\varepsilon_N} = 0$ in \eqref{eq:posterior-moments} gives the posterior moments for GP interpolation. A strictly positive definite kernel together with pairwise distinct $x_1, \dots, x_N$ is sufficient for invertibility; when $k$ is simply positive definite, the matrix may still be invertible, but it must be carefully checked. 
While a strictly positive definite $k$, specifically a Gaussian or a Mat\'ern $k$, is the most common scenario in interpolation, implying invertible $k(x_{1:N}, x_{1:N})$, we nevertheless cover what happens when $k(x_{1:N}, x_{1:N})$ is singular for some mutually distinct $x_{1:N}$.

\paragraph{Singular Gram matrices.} When $k(x_{1:N}, x_{1:N})$ is not invertible, $k(x_{1:N}, x_{1:N})$ has rank $r < N$, two scenarios are possible: either there are redundant datapoints, or the data is incompatible with the noiseless interpolation with kernel $k$. Interpolation with a linear kernel in $\bR$ is a simple example: when there is a $c \in \bR$ such that $y_n = c x_n$ for all $n \in \{1, \dots, N\}$, i.e., $(x_1, y_1), \dots, (x_N, y_N)$ lie along the same line, all datapoints but one are redundant and can be discarded; otherwise, when no such $c$ exists, linear interpolation is not possible and regression must be performed instead. 
To see this for a general $k$, suppose $k(x_{1:r}, x_{1:r})$ is full rank, and take $A \in \bR^{N \times r}$ such that $k(x_{1:N}, x_{1:N}) = AA^\top$. Then by definition of Gaussian processes, the finite-dimensional marginal $f(x_{1:N}) \sim \mathcal{N}(0, AA^\top)$ can be written in distribution as $f(x_{1:N}) \stackrel{d}{=} Az$ for $z \sim \mathcal{N}(0, \Id_r)$, where $\Id_r$ is the $r \times r$ identity matrix. 
Since $r < N$, for any draw $z$ exactly $N - r$ components of $f(x_{1:N})$ are fully determined by the remaining $r$. In particular, partitioning $A = \begin{pmatrix} A_1 \ A_2 \end{pmatrix}^\top$ with $A_1 \in \mathbb{R}^{r \times r}$ and $A_2 \in \mathbb{R}^{(N-r) \times r}$, the full-rank assumption on $k(x_{1:r}, x_{1:r}) = A_1 A_1^\top$ implies $A_1$ is invertible, and $f(x_{r+1:N}) = A_2 A_1^{-1} f(x_{1:r})$. Therefore, observations $y_{1:N}$ are compatible with $\GP(0, k)$ interpolation if and only if $y_{r+1:N} = A_2 A_1^{-1} y_{1:r}$, in which case it suffices to interpolate using the $r$ independent points $(x_{1:r}, y_{1:r})$ with the invertible kernel matrix $k(x_{1:r}, x_{1:r})$, and discard the rest. Lastly, it is worth noting that, provided $y_{r+1:N} = A_2 A_1^{-1} y_{1:r}$, discarding datapoints is equivalent to using the Moore-Penrose pseudoinverse $A^{+\top} A^+$ of $k(x_{1:N}, x_{1:N})$, and to GP regression as the Gaussian noise variance tends to zero; specifically, for any $x \in \calX$,
\begin{align*}
    k(x_{1:r}, x)^\top k(x_{1:r}, x_{1:r})^{-1} y_{1:r} 
    &= k(x_{1:N}, x)^\top A^{+\top} A^+ y_{1:N} \\
    &= \lim_{\sigma_\varepsilon \to 0} k(x_{1:N}, x)^\top (k(x_{1:N}, x_{1:N}) + \sigma_\varepsilon^2 \Id_N)^{-1} y_{1:N}. 
\end{align*}
We omit the proof: it is straightforward and of low relevance to the thesis since we will only interpolate with strictly positive definite kernels.

\subsection{Bayesian quadrature.}
\label{sec:bq_background}
In BQ \citep{Diaconis1988,OHagan1991,Rasmussen2003,Briol2019PI}, the integral
\begin{equation*}
    I = \int_\calX f(x) \bP(\d x)
\end{equation*}
is modelled by GP interpolation on the integrand $f$, based on evaluations $(x_{1:N}, f(x_{1:N}))$. For the rest of this thesis, we only consider zero-mean GP priors, $\GP(0, k)$, a common and unrestrictive assumption discussed in \Cref{rem:gp_zero_mean}. Provided the kernel $k$ is square-root $\bP$-integrable,
\begin{equation}
\label{eq:assumptions_on_moments_in_bq}
    \int_\calX \sqrt {k(x, x)} \bP(\d x) < \infty
\end{equation}
samples from the posterior $\GP(m_N, k_N)$ are integrable almost surely, and the integral of the posterior GP is a random variable with univariate Gaussian distribution~\citep[Example 5.3]{kukush2020gaussian}. Further, the mean and variance of said Gaussian, which we denote $\calN(I_\BQ, \sigma^2_\BQ)$, are exactly the integrated posterior mean $m_N$ and variance $k_N$ of the GP,
\begin{equation*}
    I_\BQ = \int_\calX  m_N(x) \bP(\d x), \qquad \sigma^2_\BQ = \int_\calX \int_\calX  k_N(x, x') \bP(\d x) \bP(\d x').
\end{equation*}
After substituting in the exact posterior moments in~\eqref{eq:posterior-moments} for the interpolation setting, $\sigma_{\varepsilon_1} = \dots = \sigma_{\varepsilon_N} = 0$, with a zero prior mean $m \equiv 0$, we get
\begin{equation}
\label{eq:bq-posterior-moments}
\begin{split}
    I_\BQ & = \mu_{k,\bP}(x_{1:N})^\top k(x_{1:N}, x_{1:N})^{-1} f(x_{1:N}), \\
    \sigma^2_\BQ &= \int_\calX \int_\calX k(x, x') \bP(\d x) \bP(\d x') - \mu_{k,\bP}(x_{1:N})^\top k(x_{1:N}, x_{1:N}) ^{-1} \mu_{k,\bP}(x_{1:N}).
\end{split}
\end{equation}
As mentioned earlier and proved in~\citet[Section 7.2.2]{kanagawa2025gaussian}, $I_\BQ = I_\mathrm{KQ}$, and $\sigma^2_\BQ = \MMD_k^2(\bP, \bP^w_N)$ for $w^\top_{1:N}= \mu_{k,\bP}(x_{1:N})^\top k(x_{1:N}, x_{1:N})^{-1}$. From~\eqref{eq:bq-posterior-moments}, it is clear that $I_\BQ$ and $\sigma^2_\BQ$ are available in closed form whenever
\begin{align*}
    \mu_{k,\bP}(x')=\int_\calX k(x, x') \bP(\d x) \text{ for any } x' \in \calX,\quad \int_\calX \int_\calX k(x, x') \bP(\d x) \bP(\d x') 
\end{align*}
are available in closed form. While this is a rather strong requirement that does not hold for all pairs $k, \bP$, a list of well-known pairs can be found in~\citep{pmlr-v271-briol25a,Nishiyama2016,nishiyama2020model}; notably, for both Mat\'ern and Gaussian kernels $k$, closed forms are known when $\bP$ is uniform or Gaussian. Even when none of these pairs are appropriate for the problem at hand, there are still multiple solutions: importance sampling and Stein reproducing kernels. Importance sampling replaces the task with a $\bQ$-integral
\begin{equation*}
    I = \int_\calX f(x) \bP(\d x) = \int_\calX f(x) \frac{p(x)}{q(x)} \bQ(\d x),
\end{equation*}
for some $\bQ$ such that there is a kernel $k$ for which closed forms are available, and densities $p, q$ of $\bP, \bQ$. Stein reproducing kernels apply a Stein operator to build a kernel whose mean embedding under $\bP$ is some constant, which can then be tuned as a hyperparameter~\citep{anastasiou2023stein}.

As discussed in the context of GP interpolation above, BQ requires $k(x_{1:N}, x_{1:N})$ to be invertible. Mutually distinct points and a strictly positive definite kernel, such as the Gaussian or the Mat\'ern family, are sufficient.

The square-root integrability condition in \eqref{eq:assumptions_on_moments_in_bq} is precisely $\bP\in\calP_k(\calX)$, the condition under which the KME $\mu_{k,\bP}$ is defined in~\Cref{def:kme}: this is one of many links between Gaussian processes and kernel methods; see~\citet{kanagawa2025gaussian} for a thorough treatment. While checking \eqref{eq:assumptions_on_moments_in_bq} directly may be feasible for standard BQ, it can be simpler to reason in terms of the entire $\calP_k(\calX)$ when approximating a family of integrals against $\{\bP_\theta\}_{\theta\in\Theta}$, as in~\Cref{sec:mmdintegration}. As discussed in~\Cref{sec:kernel_mean_embeddings_and_mmd}, $\calP_k(\calX)=\calP(\calX)$ if and only if $k$ is bounded.

\paragraph{Efficiency.}
The convergence rate of $I_\BQ$ is well-studied~\citep{Briol2019PI,Kanagawa2019,wynne2021convergence} and is particularly fast for low- to mid-dimensional smooth integrands. 
This has to be contrasted with the computational cost, which is inherited from GP interpolation and is $\bigo(N^3)$. 
For this reason, BQ has primarily been applied to problems where evaluating the integrand is expensive and only a small number of evaluations $N$ is available. Modifications exist for cheaper problems; we review these in~\Cref{app:reducing_the_cost}.

\section{Quantifying smoothness of functions}
\label{sec:quantifying_smoothness_of_functions}

Smoothness of a function can be a powerful tool in analysing further properties of said function, as well as how it interacts with functions of different smoothness. In statistical learning theory, the gap between the smoothness of the function that is describing the true data-generating process and the smoothness of a function class used to approximate said process sometimes allows us to prove asymptotic and finite-sample bounds on the approximation error.

Throughout this work, we employ two ways of describing how smooth a given function $f$ is: H\"older continuity of its derivatives (or $f$ being in a H\"older space,~\citet[Section 5.1]{evans2010partial}), and existence of weak square-integrable derivatives (or $f$ being in a Sobolev space,~\citet[Section 5.1]{adams2003sobolev}). The relationship between H\"older and Sobolev spaces is captured by the Sobolev embedding theorem~\citep[Theorem 4.12]{adams2003sobolev}, which we address later in this section; both classes of spaces are valid approaches to describing smoothness, with some problems proving easier to tackle using one or the other. In this thesis, we make use of H\"older spaces in~\Cref{sec:gpcv}, and Sobolev spaces in~\Cref{sec:mmdsbi,sec:mmdintegration}.

Throughout this section, let $\calX \subseteq \bR^d$ be open. We will use the following standard notation for spaces of $s$ times continuously differentiable $f:\calX \to \bR$.

\begin{definition}[Spaces of continuously differentiable functions]
\label{def:cs_space}
    Let $s \in \bN$. The space $C^s(\calX)$ contains all functions $f: \calX \to \bR$ on an open $\calX \subseteq \bR^d$ whose partial derivatives
    \begin{equation*}
        \partial^\alpha f=\frac{\partial^{|\alpha|} f}{\partial^{\alpha_1} x_1 \dots \partial^{\alpha_d} x_d}, \qquad \qquad |\alpha|\coloneqq \sum_{i=1}^d \alpha_i,
    \end{equation*}
    exist and are continuous on $\calX$ for every multi-index $\alpha \in \bN^d$ with $|\alpha| \leq s$.
\end{definition}

In particular, $C^0(\calX)$ is the space of continuous functions, and $C^\infty(\calX)$ is the space of infinitely continuously differentiable functions.

\subsection{H\"older spaces}

We start with classic function spaces used to quantify function smoothness through fractional continuity of its partial derivatives: H\"older spaces. First, let us define this notion of fractional continuity.

\begin{definition}[H\"older continuity]
\label{def:holder_continuity}
A function $f: \calX \to \bR$ on an open $\calX \subseteq \bR^d$ is $\alpha$-\emph{Hölder continuous} for $0 < \alpha \leq 1$ if there is a constant $L \geq 0$ such that, for all $x, x' \in \calX$,
\begin{equation*}
    |f(x) - f(x')| \leq L \| x - x' \|_2^ \alpha.
\end{equation*}
\end{definition}
Any such constant $L$ is called a \emph{H\"older constant} of $f$. When $\alpha=1$, $f$ is said to be a \emph{Lipschitz} function. For a compact convex $\calX$, every continuously differentiable function, $f \in C^1(\calX)$, is $\alpha$-H\"older continuous for all $0< \alpha \leq 1$~\citep{Folland2001-bd}. Hence, to quantify the smoothness of a continuously differentiable function, it is natural to consider the H\"older continuity of its partial derivatives. For $s$ times continuously differentiable functions, $f \in C^s(\calX)$, this is formalised by the notion of H\"older spaces.

\begin{definition}[H\"older spaces]
  \label{def:hoelder_space}
  Let $s \in \bN$ and $0 < \alpha \leq 1$. The \emph{H\"older space} $C^{s, \alpha}(\calX)$ contains all functions $f \in C^s(\calX)$ on an open $\calX \subseteq \bR^d$ whose $s$-th partial derivatives, $\partial^\beta f$ for $|\beta| =s$, are $\alpha$-H\"older continuous.
\end{definition}

In this notation, $C^{0, \alpha}(\calX)$ is the space of continuous functions that are $\alpha$-H\"older continuous.
For a bounded $\calX$, an $\alpha_1$-H\"older continuous function is $\alpha_2$-H\"older continuous when $\alpha_1 > \alpha_2$; further, as already discussed, when $\calX$ is compact and convex, continuously differentiable functions are $\alpha$-H\"older continuous for any $0< \alpha \leq 1$. This together with the fact that not all Lipschitz functions $C^{0, 1}(\calX)$ are differentiable gives the following strict inclusions on compact convex $\calX$:
\begin{itemize}
  \item $C^{s_1, \alpha_1}(\calX) \subsetneq C^{s_2, \alpha_2}(\calX)$ if (a) $s_1 > s_2$ or (b) $s_1 = s_2$ and $\alpha_1 > \alpha_2$,
  \item $C^{s+1}(\calX) \subsetneq C^{s, 1}(\calX)$.
\end{itemize}

\subsection{Sobolev spaces.}
\label{sec:bg_sobolev_spaces}

Originally studied in the context of partial differential equations, Sobolev spaces $\calW^{s, p}(\calX)$ turn to power-$p$ integrability and a weakened notion of differentiability to quantify smoothness. We start by formalising power-$p$ integrability, with the help of the notion of a seminorm, a function that has all the properties of a norm but may take the value zero on non-zero elements.

\begin{definition}[$\calL^p$ spaces]
\label{def:lp_space}
Let $\calX \subseteq \bR^d$ be open, and $\bP$ be a measure on $\calX$. For $p \in [1, \infty)$, the space $\calL^p(\calX, \bP)$ with a seminorm $\| f \|_{\calL^p(\calX, \bP)}$ contains all functions $f:\calX \to \bR$ that are power-$p$ Lebesgue integrable under $\bP$, 
\begin{equation*}
    \int_\calX |f(x)|^p  \bP(\d x)< \infty, \qquad \qquad \| f \|_{\calL^p(\calX, \bP)} = \left(\int_\calX |f(x)|^p  \bP(\d x) \right)^{1/p}.
\end{equation*}
Further, the space $\calL^\infty(\calX, \bP)$ with a seminorm $\| f \|_{\calL^\infty(\calX, \bP)}$ contains all functions $f:\calX \to \bR$ that are bounded $\bP$-almost everywhere,
\begin{equation*}
    \| f \|_{\calL^\infty(\calX, \bP)} = \inf\left\{M \geq 0: \bP(\{x: |f(x)|>M\})=0\right\}.
\end{equation*} 
\end{definition}
When $\bP=\mu$, the Lebesgue measure, we will simplify the notation to $\calL^p(\calX) \coloneqq \calL^p(\calX, \mu)$. For any $p \in [1, \infty]$, the seminorm $\| \cdot \|_{\calL^p(\calX, \bP)}$ is not a norm in general: for any $f \neq f'$ (equivalently, $f-f' \neq 0$) that coincide $\bP$-almost everywhere, it will hold that
\begin{equation*}
    \| f-f' \|_{\calL^p(\calX, \bP)} = 0.
\end{equation*}
This does become a norm on the space $L^p(\calX, \bP)$ of \emph{equivalence classes} of functions, that group together functions that agree $\bP$-almost everywhere. Nevertheless, it is common~\citep[Chapter 2]{adams2003sobolev} to ignore the distinction between an equivalence class and a function, whenever the application or result in question is not affected by function values on a set of measure zero. This will be the case throughout this thesis, so for clarity, we keep to the function spaces $\calL^p$. Further, whenever some function space $\calF \subseteq \calL^p(\calX, \bP)$ only contains continuous functions, and $\bP$ has full support on $\calX$, each equivalence class collapses to a single continuous function: continuous $f, f'$ can agree $\bP$-almost everywhere if and only if $f \equiv f'$. In other words, the distinction between an equivalence class and a function fully disappears for such $\calF$.

One important example of such $\calF$ is Sobolev spaces $\calW^{s, 2}(\bR^d)$ for $s > d/2$; further, these are RKHS, and a particularly important tool in statistical learning theory and kernel methods. To formally introduce Sobolev spaces, we first define the notion of weak derivatives, an integration by parts-based generalisation of the concept of a derivative to functions that are not differentiable. 

\begin{definition}[Weak derivatives]
\label{def:weak_derivatives}
    Let $\calX \subseteq \bR^d$ be open, $\alpha \in \bN^d$, and $f, g$ be locally integrable, i.e., $f,g \in \calL^1(S)$ for every compact $S \subset \calX$. We say $g$ is an $\alpha$-\emph{weak derivative} of $f$ if
    \begin{equation*}
        \int_\calX f(x) \partial^\alpha \phi(x) \d x = (-1)^{|\alpha|} \int_\calX g(x) \phi(x) \d x
    \end{equation*}
    holds for any $\phi \in C_c^\infty(\calX)$, a smooth function with compact support.
\end{definition}

In other words, a locally integrable function is a weak derivative if it closely resembles the behaviour of the ordinary derivative: for any infinitely continuously differentiable function with a compact support, integration by parts holds as it would for an ordinary derivative. It is clear that the weak derivative does not have to be unique, but two $\alpha$-weak derivatives $g, g'$ of $f$ will agree Lebesgue-almost everywhere; in particular, if an ordinary derivative $\partial^\alpha f$ exists, it is Lebesgue-almost everywhere equal to any weak derivative. 

As such, by $D^\alpha f$ we will refer to any function $g$ that satisfies the definition of a weak derivative of $f$; although, analogously to the $\calL^p$ spaces, it can be formally defined as the equivalence class of functions that agree almost everywhere. Again, the differences on sets of measure zero will have no effect within this thesis, and it will be sufficient to work with an arbitrary representative of said equivalence class.
Finally, the settings in~\Cref{sec:mmdsbi,sec:mmdintegration} require notation for weak derivatives with respect to particular variables, which we introduce in the corresponding chapters.

\begin{definition}[Sobolev spaces]
\label{def:sobolev_space}
    Let $\calX \subseteq \bR^d$ be open, and $s \in \bN$. For $p \in [1, \infty)$, the Sobolev space $\calW^{s, p}(\calX)$ with norm $\|\cdot \|_{\calW^{s, p}(\calX)}$ contains all Lebesgue-measurable functions $f: \calX \to \bR$ whose weak derivatives $D^\alpha f$  for any multi-index $\alpha \in \bN^d$ with $|\alpha|=\sum_{i=1}^d \alpha_i \leq s$ lie in $\calL^p(\calX)$. The $s,p$-Sobolev norm is defined as
    \begin{equation*}
        \|f \|_{\calW^{s, p}(\calX)} = \left(\sum_{\substack{\alpha \in \bN^d \\ |\alpha| \leq s}}  \|D^\alpha f \|^p_{\calL^p(\calX)} \right)^{\nicefrac{1}{p}}.
    \end{equation*}
\end{definition}

The definition implies higher-order Sobolev spaces lie in lower-order ones,
\begin{equation*}
    \calW^{s, 2}(\calX) \subseteq \calW^{s', 2}(\calX) \text{ for any integer } 0<s'\leq s<\infty.
\end{equation*}
The Sobolev embedding theorem~\citep[Theorem 4.12]{adams2003sobolev} establishes functional properties of the inclusion map, as well as the relationships between H\"older and Sobolev spaces. For the purposes of this thesis, the most useful result of this type is $\calW^{s, 2}(\bR^d) \subseteq C^{\lceil s - d/2 \rceil - 1}(\bR^d)$ when $s > d/2$, i.e., the functions in $\calW^{s, 2}(\bR^d)$ are continuous and $\lceil s - d/2 \rceil - 1$ times continuously differentiable. 

The Sobolev spaces $\calW^{s, 2}$ for $s > d/2$ are particularly important in practice and in the context of this thesis: they are reproducing kernel Hilbert spaces, with kernels that consequently encode smoothness of $s$. We elaborate on this in the next section, and refer to~\citet{adams2003sobolev} for an in-depth treatment of general Sobolev spaces $\calW^{s, p}$. 

\subsection{Sobolev kernels.}
\label{sec:bg_sobolev_kernels}

We now use Sobolev spaces to introduce a notion of kernel smoothness, and restate a combination of classic results demonstrating smoothness of Mat\'ern kernels defined in~\Cref{tab:example_kernels}; this makes the methods in~\Cref{sec:mmdsbi,sec:mmdintegration} applicable in practice with Mat\'ern kernels. First, we introduce a notion of equivalence of normed spaces.
\begin{definition}[Norm-equivalence]
    Two normed spaces $(\calF_1, \|\cdot\|_1)$ and $(\calF_2, \|\cdot\|_2)$ are said to be norm-equivalent if $\calF_1=\calF_2$ in the set sense, and the norms $\| \cdot \|_1$, $\| \cdot \|_2$ are equivalent, meaning there are constants $c, C>0$ such that for any $f \in \calF_1=\calF_2$,
    \begin{equation*}
        c \| f \|_1 \leq \| f \|_2 \leq C \| f \|_1.
    \end{equation*}
\end{definition}
It is worth noting that while the term `norm-equivalence' is standard in machine learning, elsewhere it is more common to say instead that $\calF_1=\calF_2$, with equivalent norms. 

\begin{definition}[Sobolev kernel]
\label{def:sobolev_kernel}
    A kernel on an open $\calX \subseteq \bR^d$ is said to be a \emph{Sobolev kernel} of smoothness $s>d/2$ when it induces an RKHS that is norm-equivalent to a Sobolev space $\calW^{s, 2}(\calX)$.
\end{definition}

In fact, the large size of Sobolev spaces implies a stronger property: Sobolev kernels are \emph{strictly} positive definite. For completeness, we provide the proof.

\begin{lemma}
\label{res:sobolev_kernels_are_spd}
    Any Sobolev kernel $k$ is strictly positive definite.
\end{lemma}
\begin{proof}
Suppose that, for some $\alpha_{1:N} \in \bR^N$ and mutually distinct $x_{1:N} \in \calX^N$,
\begin{equation*}
    \alpha_{1:N}^\top k(x_{1:N}, x_{1:N})  \alpha_{1:N} = \sum_{n=1}^N \sum_{n'=1}^N \alpha_n \alpha_{n'} k(x_n, x_{n'}) = 0.
\end{equation*}
We will show this implies $\alpha_1 = \dots = \alpha_N = 0$.
By the reproducing property,
\begin{align*}
    \sum_{n=1}^N \sum_{n'=1}^N \alpha_n \alpha_{n'} k(x_n, x_{n'}) 
    &= \left\langle \sum_{n=1}^N \alpha_n k(x_n, \cdot), \sum_{n=1}^N \alpha_n k(x_n, \cdot) \right \rangle_{\calH_k} \\
    &= \left \| \sum_{n=1}^N \alpha_n k(x_n, \cdot) \right\|_{\calH_k}^2.
\end{align*}
Therefore, $\sum_{n=1}^N \alpha_n k(x_n, \cdot)$ is the zero element in $\calH_k$, and for any $g \in \calH_k$,
\begin{equation}
\label{eq:sob_kern_are_spd}
    \sum_{n=1}^N \alpha_n g(x_n) = \left \langle g, \sum_{n=1}^N \alpha_n k(x_n, \cdot) \right \rangle_{\calH_k} = 0
\end{equation}
holds by the reproducing property. Since $x_1, \ldots, x_N$ are distinct points in the open set $\calX$, there exist pairwise disjoint open sets $U_1, \ldots, U_N \subset \calX$ with $x_n \in U_n$ for each $n$. Then, for any fixed $n \in \{1, \dots, N\}$, by standard mollifier construction~\citep[C.4]{evans2010partial} there exists an infinitely smooth $\varphi_n: \calX \to \bR$ such that $\varphi_n(x_n) = 1$ and $\varphi_n(x) = 0$ whenever $x \notin C_n$, for some compact $C_n \subset U_n$. By definition, all infinitely smooth functions with compact support are Sobolev; then, substituting $g = \varphi_n$ into~\eqref{eq:sob_kern_are_spd} gives $\alpha_n = 0$. Repeating this for every $n \in \{1, \dots, N\}$ concludes the proof.
\end{proof}

This result is important for MMD-minimising integration, which, as discussed in~\Cref{sec:kernel_and_bayesian_quadrature}, requires the Gram matrix $k(x_{1:N}, x_{1:N})$ to be invertible. Provided $x_{1:N}$ are mutually distinct, a strictly positive definite $k$ is sufficient; therefore, MMD-minimising quadrature with Sobolev kernels is always valid.

\paragraph{Mat\'ern kernels are Sobolev.} Due to their popularity in machine learning literature and in practice, Mat\'ern kernels are the running example of Sobolev kernels throughout this thesis. Other Sobolev kernels include the compactly supported Wendland kernels~\citep[Chapter 9]{Wendland2005} favoured in scattered data approximation literature, and the Sobolev generalised hypergeometric kernels~\citep{JMLR:v26:25-0022} that contain Mat\'ern and Wendland kernels as special cases. The precise relationship between Mat\'ern kernels and Sobolev spaces is as follows.

\begin{theorem}
\label{res:matern-is-sobolev-integer-nu}
    Let $\calX=\bR^d$, or $\calX \subset \bR^d$ be open, bounded, and convex. The RKHS of the Mat\'ern kernel $k_\nu: \calX \times \calX \to \bR$ for $\nu = a/2$ for $a \in \bN_{\geq 1}$ is norm-equivalent to the Sobolev space $\calW^{s, 2}(\calX)$ for $s = \nu + d/2$.
\end{theorem}
\begin{proof}
    For $\calX=\bR^d$, this is shown in~\citet[Theorem 6.13 and Corollary 10.13]{Wendland2005}.
    For $\calX \subset \bR^d$ that are open, bounded, and have a Lipschitz boundary, it is an immediate corollary of \citet[Theorem 6.13 and Corollary 10.48]{Wendland2005}.  Since convex sets have a Lipschitz boundary~\citep{stein1970singular}, the result holds.
\end{proof}
The conditions on $\calX$ are important: they ensure the existence of an \emph{extension operator} $\calW^{s, 2}(\calX) \to \calW^{s, 2}(\bR^d)$~\citep[Section 5.17]{adams2003sobolev}, that essentially ensures any function in $\calW^{s, 2}(\calX)$ can be extended into $\calW^{s, 2}(\bR^d)$ without its norm increasing (up to a multiplicative constant). This allows for results in $\bR^d$ to be restricted to $\calX$. Extension operators exist for sufficiently smooth $\calX$; one of the more general cases can be found in~\citet[Theorems 6.1 and 6.7]{devore1993besov}.

The result in~\Cref{res:matern-is-sobolev-integer-nu} restricts the Mat\'ern order $\nu$ to half-integers, $a/2$ for $a \in \bN_{\geq 1}$, because of the integer $s$ in the definition of Sobolev spaces. However, it holds in the more general case, for \emph{Sobolev-Slobodeckij spaces} $\calW^{s, 2}$ that generalise Sobolev spaces for real $s>0$~\citep[Chapter 7]{adams2003sobolev}. We omit the theory of Sobolev-Slobodeckij spaces as it is not required for results in this thesis; however, for completeness, we now cover the extension of~\Cref{res:matern-is-sobolev-integer-nu} to any real $\nu > 0$. 

\begin{theorem}
\label{res:matern-is-sobolev-for-any-nu}
    Let $\calX=\bR^d$, or $\calX \subset \bR^d$ be open, bounded, and convex. The RKHS of the Mat\'ern kernel $k_\nu: \calX \times \calX \to \bR$ for $\nu > 0$ is norm-equivalent to the Sobolev space $\calW^{s, 2}(\calX)$ for $s = \nu + d/2$.
\end{theorem}
\begin{proof}
    For $\calX=\bR^d$, this is shown in~\citet[Theorem 6.13 and Corollary 10.13]{Wendland2005}. An extension operator introduced in~\citet{rychkov1999restrictions} for Besov spaces (of which fractional Sobolev spaces are a special case) allows us to extend the result to open and bounded $\calX \subset \bR^d$ with a Lipschitz boundary. As mentioned in the proof of \Cref{res:matern-is-sobolev-integer-nu}, convex sets have a Lipschitz boundary, completing the proof.
\end{proof}

Lastly, we formally state that the inclusion property of Sobolev spaces stated after~\Cref{def:sobolev_space} extends to fractional orders.

\begin{proposition}[Proposition 2.1 and Corollary 2.3 in~\citet{dinezza2012hitchhikers}]
\label{res:sobolev-nested}
    Let $\calX = \bR^d$ or $\calX \subset \bR^d$ be open, bounded, and convex. Then,
    \begin{equation*}
        \calW^{s, 2}(\calX) \subseteq \calW^{s', 2}(\calX) \text{ for any } 0<s'\leq s<\infty.
    \end{equation*}
\end{proposition}

The aforementioned Sobolev embedding theorem establishes a stronger version of this result: continuous embeddings, rather than mere set inclusion, between Sobolev-Slobodeckij and H\"older spaces. We refer the interested reader to~\citep[Theorem 4.12]{adams2003sobolev}.

\part{Novel Methods Based on the MMD}
\label{sec:part1}

\chapter{Efficient MMD Estimators for Simulation Based Inference}
\label{sec:mmdsbi}
\begin{tcolorbox}
The results in this chapter were published in the following paper:

\begin{itemize}
    \item Bharti, A., Naslidnyk, M., Key, O., Kaski, S., \& Briol, F.-X. (2023). Optimally-Weighted Estimators of the Maximum Mean Discrepancy for Likelihood-Free Inference. International Conference on Machine Learning.
\end{itemize}

All theoretical results in this chapter are due to the author. Experiments were carried out by Dr Ayush Bharti and are included to support the theory numerically.
\end{tcolorbox}

Using \emph{simulator-based models} to study the behaviour of complex systems or phenomena is common across science and engineering, in fields such as population genetics \citep{Beaumont2010}, astronomy \citep{Akeret_2015}, radio propagation \citep{Bharti2022}, and agent-based modelling \citep{Jennings1999}. A sample from such a model is generated through a simple procedure: (1) sample from a tractable distribution (such as a Gaussian or a uniform); (2) apply the \emph{simulator} to each sample. This simplicity, however, comes at the cost of an intractable likelihood, and the need for methods alternative to likelihood-based inference. 
A broad family of \emph{likelihood-free inference} methods has been developed in response; see \citet{Lintusaari2016, Cranmer2020} for surveys.

\begin{figure}
	 \centering
	 \includegraphics[width = 0.85\linewidth]{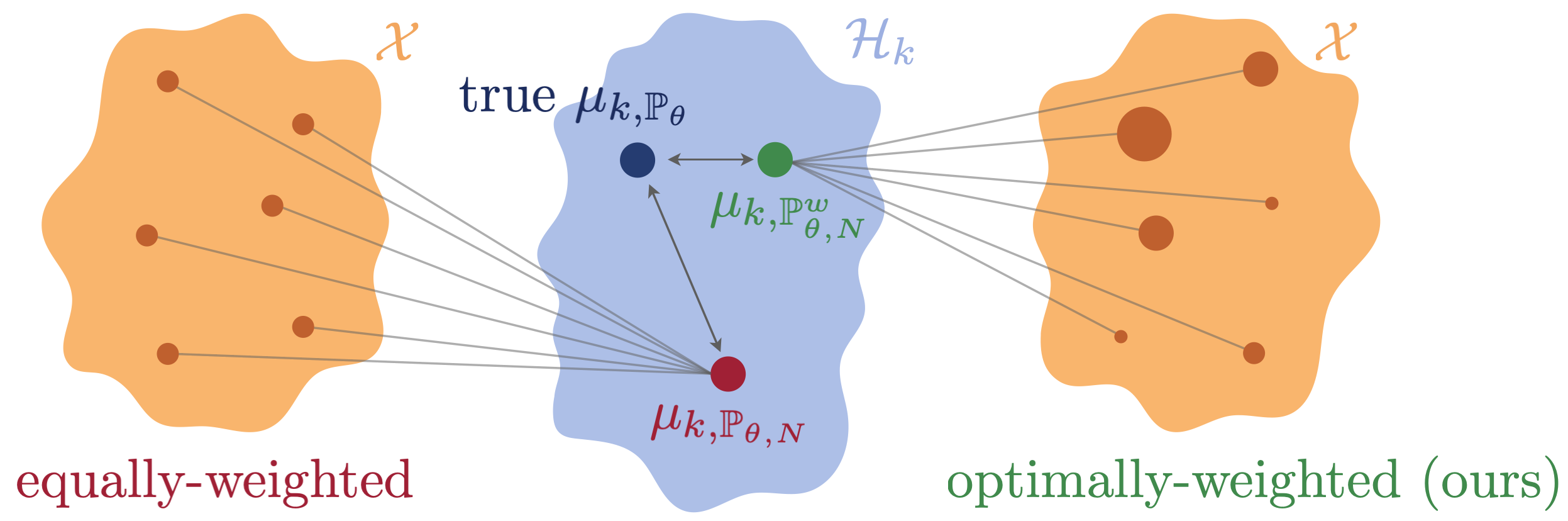}
\caption[Illustration of an optimally-weighted MMD estimator for SBI.]{Estimating the MMD requires approximating the embedding $\mu_{k,\bP_\theta}$ of the model $\bP_\theta$ in the RKHS $\calH_k$.
The classical approach approximates it using $N$ \textcolor{purple}{equally-weighted} i.i.d. samples from $\bP_\theta$, denoted \textcolor{purple}{$\mu_{k,\bP_{\theta,N}}$}.
We show that this estimator can be improved by using \textcolor{ForestGreen}{optimally-weighted} samples, denoted \textcolor{ForestGreen}{$\mu_{k,\bP^w_{\theta,N}}$.}}
\label{fig:sketch}
\end{figure}

A common approach for likelihood-free inference involves measuring some notion of discrepancy between data sampled from the model and data sampled from the real process. Due to its low \emph{sample complexity}, that is, how many samples from the distributions are needed to produce an estimate within some margin of error, the MMD is a common choice. The faster the estimation error goes to zero with the number of samples, the less we need to simulate from the model, and hence, the smaller the computational cost. This is key especially for expensive simulators, which in the most extreme cases can take up to hundreds or thousands of CPU hours per simulation; see \citet{Niederer2019} for an example in cardiac modelling. Other examples include tsunami models based on shallow water equations that require several GPU hours per run \citep{Behrens2015}, runaway electron analysis models for nuclear fusion devices that require 24 CPU hours per run \citep{DREAM}, and models of large-scale wind farms that require 100 CPU hours per run \citep{Kirby2022}.

The MMD has been used in a range of frameworks for likelihood-free inference, including for approximate Bayesian computation (ABC) \citep{Park2015, Mitrovic2016, Kajihara2018, Bharti2022,Legramanti2022}, for minimum distance estimation (MDE) \citep{briol2019statistical, cherief2021, Alquier2021,Niu2023,Key2025}, for generalised Bayesian inference \citep{cherief2020mmd,Pacchiardi2021}, for Bayesian nonparametric learning \citep{Dellaporta2022}, and for training generative adversarial networks \citep{Dziugaite2015,Li2015GMMN,Li2017MMDGAN,Binkowski2018}.

Here, we do not revisit the question of whether the MMD is the best choice of discrepancy for a particular problem. Instead, we assume that the MMD has been chosen, and focus on constructing estimators with strong sample complexity for this distance. The most common estimators for the MMD are U- or V-statistic estimators, and these have sample complexity of $\bigo(N^{-\nicefrac{1}{2}})$, under mild conditions \citep{briol2019statistical}, where $N$ is the number of samples. In recent work, \citet{Niu2023} showed that this can be improved to $\bigo(N^{-1+\varepsilon})$ for any $\varepsilon>0$ through the use of a V-statistic estimator and randomised quasi-Monte Carlo (RQMC) sampling. This significant improvement does come at the cost of restrictive assumptions: the simulator must be written in a form where the inputs are uniform random variables, and must satisfy stringent smoothness conditions which are difficult to verify in practice.

We propose a novel set of \emph{optimally-weighted} estimators with sample complexity of $\bigo(N^{-s_c/r})$ where $r$ is the dimension of the base space and $s_c$ is a parameter depending on the smoothness of the kernel and the simulator. This results in strictly better rates than U- and V-statistic estimators with i.i.d. samples for any $s_c$, and than RQMC when $s_c>r$. Additionally, the optimality of the weights guarantees that even if this condition is not satisfied, the rate is never worse than existing i.i.d. estimators.

\section{Simulation-Based Inference with the MMD}

We consider the classic parameter estimation problem, where we observe $M$ i.i.d. samples $y_1, \dots, y_M \in \calX$ from some data-generating distribution $\bQ \in \calP(\calX)$. Given $y_{1:M}$ and a parametric family of model distributions $ \{ \bP_\theta : \theta \in \Theta\} \subseteq \calP(\calX)$ with parameter space $\Theta$, we are interested in recovering the parameter value $\theta^* \in \Theta$ such that $\bP_{\theta^*}$ is either equal, or in some sense closest, to $\bQ$. 

The challenge in simulation-based inference, or likelihood-free inference, is that the likelihood associated with $\bP_\theta$ is intractable: in other words, it cannot be evaluated pointwise. This prevents the use of classical methods such as maximum likelihood estimation or (exact) Bayesian inference. Instead, we assume that we are able to simulate i.i.d. samples from $\bP_\theta$, and such models are hence called generative models or simulator-based models. Such models are characterised through their generative process, a pair $(G_\theta,\bU)$ consisting of a simple distribution $\bU$ (such as a multivariate Gaussian or uniform distribution) on a space $\calU$ and a map $G_\theta: \calU \rightarrow \calX$ called the generator or simulator. We will call $\bU$ a base measure and $\calU$ the base space, and consider $\calU \subset \bR^r$ and $\calX \subseteq \bR^d$.
To sample $x \sim \bP_\theta$, one can first sample $u \sim \bU$, then apply the generator $x = G_\theta(u)$. To perform parameter estimation for these models, it is common to repeatedly sample simulated data from the model for different parameter values and compare them to $y_{1:M}$ using a discrepancy. We now recall the discrepancy which will be the focus of this chapter.

The MMD has been used within a range of frameworks. 
In a frequentist setting, the MMD was proposed for minimum distance estimation by \citet{briol2019statistical},
\begin{align}
    \hat{\theta}_M = \underset{\theta \in \Theta}{\argmin}~ \MMD_k^2(\bP_\theta,\bQ_M).
    \label{eq:MMD_estimator}
\end{align}
In practice, the minimiser is computed through an optimisation algorithm, which requires evaluations of the squared MMD or of its gradient. Such evaluations are intractable, but any estimator can be used within a stochastic optimisation algorithm. Similar optimisation problems and stochastic approximations also arise when using the MMD for generative adversarial networks \citep{Dziugaite2015,Li2015GMMN} and for nonparametric learning \citep{Dellaporta2022}.

In a Bayesian setting, the MMD has been used to create several pseudo-posteriors by updating a prior distribution $p$ on $\Theta$ using data. For example, the K2-ABC posterior of \citet{Park2015} is a pseudo-posterior of the form
\begin{align}
\label{eq:ABC}
    p_{\mathrm{ABC}}(\theta | y_1, \dots, y_M) \propto  \bE_{x_1, \dots, x_N \sim \bP_\theta} \left[ \mathds{1}_{\{\MMD^2_k(\bP_{\theta, N},\bQ_M) < \varepsilon\}} \right] p(\theta),
\end{align}
where the indicator function $\mathds{1}_{\{A\}}$ is equal to 1 if event $A$ holds.
Here, the MMD is used to determine whether a particular instance of the parametric model is within an $\varepsilon$ distance from the data. The K2-ABC algorithm approximates this pseudo-posterior through sampling of the model $\bP_\theta$, which leads to the use of an estimator of the squared MMD.

Finally, the MMD has also been used for generalised Bayesian inference, where it is used to construct the MMD-Bayes posterior \citep{cherief2020mmd}
\begin{align*}
    p_{\mathrm{GBI}}(\theta | y_1, \dots, y_M) \propto  \exp( - \MMD_k^2(\bP_\theta,\bQ_M)) p(\theta).
\end{align*}
Once again, this pseudo-posterior is intractable, but it can be approximated through pseudo-marginal MCMC, in which case an unbiased estimator is used in place of the squared MMD \citep{Pacchiardi2021}.

\paragraph{Sample complexity of MMD estimators.}  \label{sec:complexity_bg}

As highlighted above, the performance of these likelihood-free inference methods relies heavily on how accurately we can estimate the MMD using samples; that is, how fast our estimator approaches $\MMD_k(\bP_\theta,\bQ)$ as a function of $N$ and $M$, the number of simulated and observed datapoints, respectively. Let $\widehat{\MMD}_k(\bP_{\theta,N},\bQ_M)$ be any estimator of the MMD based on $N$ simulated datapoints. Using the triangle inequality, this error can be decomposed as follows: 
\begin{align}
    |\MMD_k(\bP_\theta,\bQ)-\widehat{\MMD}_k(\bP_{\theta,N},\bQ_M)| 
    & \leq  
    |\MMD_k(\bP_\theta,\bQ)-\MMD_k(\bP_\theta,\bQ_M)| \nonumber \\
    & \; + |\MMD_k(\bP_\theta,\bQ_M)- \widehat{\MMD}_k(\bP_{\theta,N},\bQ_M)| \label{eq:error_decomposition}
\end{align}
where the first term describes the approximation error due to having a finite number of datapoints $M$, and the second term describes the error due to a finite number $N$ of simulator evaluations. To understand the behaviour of the first term, we can use the following sample complexity result for the V-statistic. The proof is a direct application of the triangle inequality together with Lemma 1 in \citep{briol2019statistical}. 
\begin{theorem} Suppose that $\sup_{x,x'} k(x,x') <\infty$ and let $\bQ_M$ consist of $M$ i.i.d. samples from $\bQ \in \calP_k(\calX)$. Then, for any $\bP \in \calP_k(\calX)$, we have with high probability 
\begin{align*}
    \left| \MMD_k(\bP,\bQ) - \MMD_k(\bP,\bQ_M)\right| =  \bigo(M^{-\nicefrac{1}{2}}).
\end{align*}
\end{theorem}
When $\widehat{\MMD}_k(\bP_{\theta, N}, \bQ_M)$ is also a V-statistic approximation, both terms in~\eqref{eq:error_decomposition} can be tackled with this result and the overall error is of size $\bigo(M^{-\nicefrac{1}{2}}+N^{-\nicefrac{1}{2}})$. This shows that we should take $N = \bigo(M)$ to ensure a good enough approximation of the MMD. Though this rate has the advantage of being independent of the dimension of $\calX$, it is relatively slow in $N$. We therefore require a large number of simulated datapoints, which can be computationally expensive.

\citet{Niu2023} recently proposed an alternate approach based on randomised quasi-Monte Carlo (RQMC) \citep{dick2013high} samples within a V-statistic. Using stronger assumptions on $\bU, k$ and $G_\theta$, they are able to obtain an estimator with improved sample complexity. We now state their assumptions and result below.

Below, we use standard smoothness and differentiation notation introduced in~\Cref{sec:quantifying_smoothness_of_functions}. Additionally, for a two-variable function $f:\calX \times \calX \to \bR$, the $\partial^{\alpha,\alpha} f$ is the $\alpha$-partial derivative in each variable. The notation $a_v:b_{-v}$ represents a point $u\in[a,b]^r$ with $u_j=a_j$ for $j\in v$, and $u_j=b_j$ for $j\notin v$.

\begin{oldassumption}
\label{as:oldpoints}
    The base space $\calU = [0,1]^r$, the base measure $\bU$ is uniform on $\calU$, and $u_1, \dots, u_N \in \calU$ form an RQMC point set.
\end{oldassumption}

\begin{oldassumption}
\label{as:oldgenerator}
    The generator $G_\theta:[0,1]^r \rightarrow \calX$ is such that
\begin{enumerate}
  \setlength{\itemsep}{1pt}
  \setlength{\parskip}{0pt}
  \setlength{\parsep}{0pt}
     \item $\partial^{(1,\ldots,1)} G_{\theta,j} \in C([0,1]^r)$ for all $j \in \{1, \dots, d\}$. 
    \item for all $j \in \{1, \dots, d\}$ and  $v \in \{0,1\}^r\setminus(0,\ldots,0)$, there is a $p_j \in [1, \infty]$, $\sum_{j=1}^d p_j^{-1} \leq 1$, such that for $g(\cdot)=\partial^v G_{\theta,j}(\cdot:1_{-v})$ it holds that $\|g\|_{\calL^{p_j}([0,1]^{|v|})} < \infty$.
\end{enumerate}
\end{oldassumption}
\begin{oldassumption}
\label{as:oldkernels}
    For any $x \in \calX$, $k(x, \cdot) \in C^r(\calX)$ and $\forall \alpha \in \bN^d,|\alpha|\leq r,\sup_{x \in \calX}\partial^{\alpha,\alpha}k(x,x)<C_k$ for a constant $C_k>0$ depending only on $k$.
\end{oldassumption}
\begin{theorem}
Under~\Cref{as:oldpoints,as:oldgenerator,as:oldkernels} and $\bQ \in \calP_k(\calX)$,
\begin{align*}
    \left| \MMD_k(\bP_\theta,\bQ) - \MMD_k(\bP_{\theta,N},\bQ)\right| =  \bigo(N^{-1+\varepsilon}).
\end{align*}
\end{theorem}

In this case, the second term in~\eqref{eq:error_decomposition} decreases at a faster rate than the first term and the overall error decreases as $\bigo(M^{-\nicefrac{1}{2}}+N^{-1+\varepsilon})$ for any $\varepsilon>0$. As a result, (ignoring log-terms) we can take $N = \bigo(M^{\nicefrac{1}{2}})$, meaning a much smaller number of simulations are required.
However, the technical conditions required are either very restrictive ($\bU$ must be uniform), or will be difficult to verify in practice (the conditions on $G_\theta$ are not very interpretable and difficult to verify). Hence, the range of cases where RQMC can be applied is limited. Additionally, when both $k$ and $G_\theta$ are smooth, faster rates can be obtained using our optimally-weighted estimator presented in the next section.

\section{Optimally-Weighted Estimators}
\label{sec:methodology}

We now present our estimator, which weights simulated data.
To that end, we denote the empirical measure of the simulated data as $\bP_{\theta,N}^w = \sum_{n=1}^N w_n \delta_{x_n}$ where $x_{1:N} = G_\theta(u_{1:N})$, and $w_{1:N} \in \bR$ are the weights associated with $x_{1:N} \in \calX$. Assuming for a moment that these weights are known, then we have 
\begin{align} \label{eq:optimally_weighted_MMD2}
    \MMD^2_k(\bP_{\theta,N}^w,\bQ_M) = \sum_{n, n' = 1}^N w_n w_{n'} k(x_n, x_{n'}) &- \frac{2}{M} \sum_{n=1}^N \sum_{m=1}^M  w_n k(x_n, y_m) \nonumber \\
    &+ \frac{1}{M^2} \sum_{m, m' = 1}^M k(y_m,y_{m'}).
\end{align}
Clearly, $w_n=1/N$ recovers the V-statistic approximation of $\MMD^2_k$, but here we have additional flexibility in how to select these weights. We will make use of a tight upper bound on the approximation error, proved in \Cref{app:proofs_optimal_weights}. 
\begin{theorem}\label{thm:optimal_weights}
Let $c:\calU \times \calU \rightarrow \bR$ be such that $k(x, \cdot) \circ G_\theta \in \calH_c$ for all $x \in \calX$, and $\bQ \in \calP_k(\calX)$.
Then, $\exists K>0$ independent of $u_{1:N}, w_{1:N}$ but dependent on $c, k, G_\theta$, such that
\begin{align*}
    | \MMD_k(\bP_\theta, \bQ) - \MMD_k(\bP_{\theta,N}^w, \bQ)|  \leq K \times \MMD_c \left(\bU, \sum_{n=1}^N w_n \delta_{u_n}\right).
\end{align*}
Provided $c(u_{1:N},u_{1:N})$ is invertible, weights minimising this upper bound can be computed in closed form as
\begin{align}
\label{eq:optimal_weights}
    w_{1:N}^* & = \argmin_{w_{1:N} \in \bR^N}  \MMD_c\left(\bU,\sum_{n=1}^N w_n \delta_{u_n}\right)  =  c(u_{1:N},u_{1:N})^{-1}  \mu_{c,\bU}(u_{1:N}),
\end{align}
where $\mu_{c,\bU}(u_{1:N})$ is the KME of $\bU$ in the RKHS $\calH_c$ evaluated at $u_{1:N}$.
\end{theorem}
Our \textit{optimally-weighted (OW) estimator} is the weighted estimator in~\eqref{eq:optimally_weighted_MMD2}
with the optimal weights in~\eqref{eq:optimal_weights}. Invertibility of $c(u_{1:N}, u_{1:N})$ holds provided $c$ is strictly positive definite and $u_{1:N}$ are mutually distinct; the former holds for Sobolev (\Cref{sec:bg_sobolev_kernels}) and Gaussian kernels, and the latter holds almost surely when $u_{1:N}$ are samples from a continuous distribution. To calculate the weights $w_{1:N}^*$, we need to evaluate $\mu_{c,\bU}$ pointwise in closed-form. The key insight is that although $\mu_{k,\bP_\theta}$ will usually not be available in closed-form,  the same is not true for $\mu_{c,\bU}$. This is because, unlike $\bP_\theta$, $\bU$ is usually a simple distribution such as a uniform, Gaussian, Gamma or Poisson. Additionally, we have full flexibility in our choice of $c$ so long as $k(x, \cdot) \circ G_\theta \in \calH_c$. Note that both terms in the upper bound in \Cref{thm:optimal_weights} depend on the kernel $c$, meaning that $c$ cannot simply be chosen for computational convenience and must also be chosen such that these quantities are as small as possible. This choice will be explored in further detail through theory in \Cref{sec:theory_sbi}, and experiments in \Cref{sec:experiments_sbi}.

\begin{remark}
    The optimal weights in \Cref{thm:optimal_weights} are equivalent to Bayesian quadrature (BQ) weights. We can therefore think of our estimator as performing BQ to approximate all integrals against $\bP$ in~\eqref{eq:MMD2_exact}. This interpretation is helpful for selecting $c$: the kernel should be chosen so that the corresponding Gaussian process is a good prior for the integrands in~\eqref{eq:MMD2_exact}. This correspondence will also help us derive sample complexity results in the next section.
\end{remark}

Our estimator minimises $\MMD_c \left(\bU, \sum_{n=1}^N w_n \delta_{u_n}\right)$ over the choice of weights, but we also have flexibility over the choice of $u_{1:N}$. Unfortunately, this optimisation cannot be solved in closed-form, and is in fact usually not convex. There is a wide range of methods which have been proposed to do point selection so as to minimise an MMD with equally-weighted points. Kernel thinning \citep{dwivedi2024kernel}, support points \citep{Mak2018} and Stein thinning \citep{Riabiz2020} are methods based on the MMD to subsample points given a large dataset. Kernel herding \citep{Chen2010,Bach2012} and Stein points \citep{Chen2018,Chen2019} are sequential point selection methods which use an MMD as objective. In addition, similar point selection methods have also been proposed for BQ \citep{Gunter2014,Briol2015,Bardenet2019}: these are closest to our OW setting.

\section{Theoretical Guarantees} \label{sec:theory_sbi}

\paragraph{Sample complexity.} The following theorem establishes a sample complexity of $\bigo(N^{-s_c/r})$ for our optimally-weighted estimator, where $s_c$ is a parameter depending on the smoothness of $k$ and $G_\theta$. We achieve a better rate than RQMC under milder conditions, as discussed below.

\begin{assumption}
\label{as:points}
    The base space $\calU \subset \bR^r$ is bounded, open, and convex, the data space $\calX$ is the entire $\bR^d$ or is bounded, open, and convex. 
    The base measure $\bU$ has a density $f_\bU: \calU \to [C_\bU, C_\bU']$ for some $C_\bU$, $C_\bU'>0$, and $\bP_\theta$ has a density bounded above.  
    The points $u_1, \dots, u_N \in \calU$ are mutually distinct, and have a fill distance of asymptotics $h_N = \bigo(N^{-\nicefrac{1}{r}})$, where $h_N = \sup_{u \in \calU} \min_{n \in \{1,\dots, N\}} \|u - u_n\|_2$.
\end{assumption}

Our assumptions on $\calU$ and $\bU$ are milder than those of \Cref{as:oldpoints}, which requires $\bU$ to be uniform. The assumptions on $\calX$ and $\bP_\theta$ are likely to hold for simulators in practice. 
For the point set $u_{1:N}$, we replace the RQMC requirement with a strictly weaker set of conditions: mutual distinctness and fill distance decay. The latter is a standard space-filling condition ensuring coverage of $\calU$. It holds for regular grids, in expectation for independent samples, and for the various quasi-random designs catalogued in \citet{wynne2021convergence}; all of these point sets also consist of mutually distinct points.

\begin{assumption}
\label{as:generator}
    The generator is a map $G_\theta:\calU \rightarrow \calX$ such that for some integer $s > r/2$, any $j \in \{1, \dots, d\}$ and any multi-index $\alpha \in \bN^r$ of size $|\alpha|\leq s$, the partial derivative $\partial^\alpha G_{\theta,j}$ exists and is bounded (in absolute value).
\end{assumption}
Assumption \Cref{as:generator} is more interpretable and easier to check than~\Cref{as:oldgenerator} (specifically part 2) as it just requires knowing how many derivatives $G_\theta$ has.
As stated in~\citet{Niu2023}, a simpler condition that implies~\Cref{as:oldgenerator} needs $G_\theta$ to be smooth up to the order $s \geq r$, which rules out the standard choices of a Mat\'ern $k$ with $\nu \in \{\nicefrac{1}{2}, \nicefrac{3}{2}, \nicefrac{5}{2}\}$ for large enough $r$. In contrast, we only ask that $s > r/2$. Next, we outline kernel smoothness assumptions, using the general notion of Sobolev kernels introduced in \Cref{def:sobolev_kernel}, and allowing for the infinitely smooth Gaussian kernel. Recall that, by~\Cref{res:matern-is-sobolev-integer-nu}, the commonly used Mat\'ern kernels are Sobolev. 
\begin{assumption}
\label{as:kernels_sbi}
    $k$ is a Sobolev kernel on $\calX$ of smoothness $s_k>d/2$, or a Gaussian kernel, and $c$ is a Sobolev kernel on $\calU$ of smoothness $s_c \in (r/2, \min(s_k, s)]$ for $s$ introduced in~\Cref{as:generator}.
\end{assumption}
\Cref{as:kernels_sbi} places fewer restrictions on the choice of $k$ than \Cref{as:oldkernels}. Although both allow for $k$ to be the Gaussian kernel, as a corollary of the Sobolev embedding theorem~\citep[Theorem 4.12]{adams2003sobolev},~\Cref{as:oldkernels} only holds for a Sobolev $k$ if $\lceil s_k - d/2 \rceil \geq r + 1$ (i.e., smooth $k$), while our lower bound on $s_k$ is much less restrictive. The conditions on $c$ are needed to ensure $k(x, \cdot) \circ G_\theta \in \calH_c$. Note that these could be weakened using the work of \citep{kanagawa2020convergence,Teckentrup2020,wynne2021convergence}, but at the expense of more restrictive conditions on $u_{1:N}$ in \Cref{as:points}. Lastly, assuming $c$ is a Sobolev kernel ensures strict positive definiteness by~\Cref{res:sobolev_kernels_are_spd}. The Gram matrix $c(u_{1:N}, u_{1:N})$ is therefore invertible whenever $u_{1:N}$ are mutually distinct, which is the case in practice as there is no need to apply the deterministic $G_\theta$ to the same datapoint $u_n$ twice.
\begin{theorem}\label{thm:rate_of_convergence}
Under \Cref{as:points,as:generator,as:kernels_sbi}, it holds that $k(x, \cdot) \circ G_\theta \in \calH_c$ for all $x \in \calX$, and for any $\bQ \in \calP_k(\calX)$,
\begin{align*}
    \left| \MMD_k(\bP_\theta,\bQ) - \MMD_k(\bP_{\theta,N}^w,\bQ)\right| = \bigo(N^{- s_c / r}). 
\end{align*}
\end{theorem}
Since $s_c>r/2$, this result shows that our method improves the sample complexity over the V-statistic for any $r$, and over RQMC when $s_c>r$.

Although the rate in \Cref{thm:rate_of_convergence} is in terms of $s_c$, the smoothness $s_k$ of $k$ and $s$ of $G_\theta$ enter through the admissible range of $s_c$: \Cref{as:kernels_sbi} requires that the kernel $c$ is not smoother than either $G_\theta$ or $k$. This suggests a practical recipe for selecting kernels. To get the fastest rate in \Cref{thm:rate_of_convergence}, we should pick a kernel $c$ with smoothness $s_c$ close to $\min\{s, s_k\}$: in other words, $c$ that is as smooth as possible, but not smoother than $G_\theta$ or $k$. If the smoothness of $G_\theta$ is unknown, a conservative choice is to take $c$ with smaller smoothness to ensure \Cref{as:kernels_sbi} holds. Finally, when we are free to choose $k$, this suggests picking it at least as smooth as $G_\theta$; or, conservatively, taking a Gaussian $k$ so $s_c$ is only constrained by $s$.

\paragraph{Computational cost.} The total computational cost of our method is the sum of (i) the cost of simulating from the model, which is $\bigo(N  C_{\mathrm{gen}})$, where $C_{\mathrm{gen}}$  is the cost of sampling one datapoint, and (ii) the cost of estimating MMD, which is $\bigo(N^2+NM+M^2)$ for the V-statistic and $\bigo(N^3+NM+M^2)$ for the OW estimator. Our method is hence slightly more expensive when $N$ is large. However, the cost of the simulator is often the computational bottleneck, sometimes taking up to tens or hundreds of CPU hours per run; see \citet{Behrens2015, Kirby2022}. As a result, proposing data-efficient likelihood-free inference methods \citep{Beaumont2009, Gutmann2016, Greenberg2019} is still an active research area. In cases where $\bigo(N C_{\mathrm{gen}}) \gg \bigo(N^3)$, the OW estimator is more efficient than the V-statistic as it requires fewer simulations to estimate the MMD. If the simulator is not costlier than estimating the MMD and assuming a fixed computational budget, then the OW estimator achieves lower error than the V-statistic if $s_c / r > 3/4$ and assumptions \Cref{as:points,as:generator,as:kernels_sbi} hold. This result is straightforwardly derived from \Cref{thm:rate_of_convergence}, as we detail in \Cref{app:cost_error}.

\section{Experiments}
\label{sec:experiments_sbi}

We now illustrate the performance of our OW estimator on various benchmark simulators and on challenging likelihood-free inference tasks. Guided by the assumptions, we take $k$ and $c$ to be either Gaussian kernels, or Mat\'ern kernels. We will denote Mat\'ern orders by $\nu_k$ and $\nu_c$ respectively; as shown in~\Cref{res:matern-is-sobolev-integer-nu}, these are Sobolev kernels of smoothness $s_k = \nu_k+d/2$ and $s_c = \nu_c+r/2$. The lengthscale of kernels $k$ and $c$ is set using the median heuristic \citep{Garreau2018}, unless otherwise stated. Our code is available at \url{https://github.com/bharti-ayush/optimally-weighted_MMD}.

\tocless\subsection{Benchmarking on popular simulators\label{sec:benchmark}}

We begin by comparing the V-statistic with our OW estimator on several popular benchmark simulators with different dimensions for $\calU \subseteq \bR^r$ and $\calX \subseteq \bR^d$. The experiments are conducted for $u_{1:N}$ being i.i.d. as well as RQMC points. We fix $\theta$ for each model (see \Cref{app:trueParam} for exact values) and estimate the MMD$^2$ between $\bP_{\theta,N}$ and $\bP_{\theta,M}$, with $k$ and $c$ both being the Gaussian kernel. We set $M = 10,000$ to be large in order to make $\bP_{\theta,M}$ an accurate approximation of $\bP_\theta$, and $N=2^8$ to facilitate comparison with RQMC, which requires $N$ to be a power of two. 

The results are in \Cref{tab:MMDerror}. For RQMC points, the errors are generally either similar for the two estimators (g-and-k, two moons, and Lotka-Volterra models) or smaller for the OW estimator (bivariate Beta and MA$(2)$), with the OW estimator achieving lower errors in all cases barring the M/G/1 queueing model. This is expected since the M/G/1 model has a discontinuous generator, and our theory, therefore, does not hold. It is also important to note that although RQMC performs very well here even without the optimal weights, the simulators were chosen in order to make this comparison feasible. In many cases, $\bU$ will not be uniform, and therefore the RQMC approach will not be possible to implement, and only the i.i.d. approach is feasible. 

\begin{table*}[t!]
\centering
\caption[Benchmarking OW-MMD on popular simulators.]{Average and standard deviation (in parentheses) of estimated MMD$^2 ~(\times 10^{-3})$ between $\bP_{\theta, N}$ and $\bP_{\theta, M}$ over 100 runs for the V-statistic and our optimally-weighted (OW) estimator. Settings: $M = 10000$, $N = 256$. }
\resizebox{\textwidth}{!}{%
\begin{tabular}{@{}c  lll | ll | ll@{}}
\toprule
\multicolumn{1}{c}{Model} & $r$ & $d$ & References & i.i.d. V-stat & i.i.d. OW (ours) & RQMC V-stat & RQMC OW (ours)\\ \midrule
g-and-k                   & 1                            & 1                                 & \citep{Bharti22a,Niu2023}                                              & 2.25 (1.52)                                                                                 & \textbf{0.086} (0.049)                                                                           & 0.060 (0.037)                                                                                & \textbf{0.059} (0.037)                                                                           \\
Two moons               & 2         & 2           & \citep{Lueckmann2021,Wiqvist2021}  & 2.36 (1.94)                      & \textbf{0.057} (0.054)                  & 0.056 (0.044)                  & \textbf{0.055} (0.044)         \\
Bivariate Beta            & 5                                                                                        & 2                                &    \citep{Nguyen2020,Niu2023}                                            & 2.13 (1.17)                                                                                 & \textbf{0.555} (0.227)                                                                           & 0.222 (0.111)                                                                                & \textbf{0.193} (0.088)                                                                            \\
MA(2)        & 12                                                                                       & 10                                            &    \citep{Marin2011,Nguyen2020}                                & 2.42 (0.796)                                                                                 & \textbf{0.705} (0.107)                                                                            & 0.381 (0.054)                                                                                & \textbf{0.322} (0.052)                                                                            \\
M/G/1 queue            & 10                                                                                       & 5                                &                                       \citep{Pacchiardi2021, Jiang2018}          & 2.52 (1.19)                                                                                & \textbf{1.71} (0.568)                                                                            & \textbf{0.595} (0.134)                                                                                    & 0.646 (0.202)                                                                         \\
Lotka-Volterra            & 600                                                                                      & 2                                &   \citep{briol2019statistical, Wiqvist2021}                                              & 2.13 (1.10)                                                                                & \textbf{2.04} (0.956)                                                                            & 1.44 (0.955)                                                                                           & \textbf{1.42} (0.942)                                                                                       \\ \bottomrule
\end{tabular}
}
\label{tab:MMDerror}
\end{table*}

For the i.i.d. points, the improvement in performance is much more significant. The OW estimator achieves the lowest error for all the models when  $u_{1:N}$ are taken to be i.i.d. uniforms. Its error is reduced by a factor of around $20$ and $40$ for the g-and-k and the two moons models, respectively, compared to the V-statistic. As expected from our sample complexity results, the magnitude of this improvement reduces as $r$ (the dimension of $\calU$) increases. However, the OW estimator still performs slightly better than the V-statistic for the Lotka-Volterra model where $r=600$.

\tocless\subsection{Multivariate g-and-k distribution\label{sec:mvgk}}

We now assess the impact of various practical choices on the performance of our method. To do so, we consider the multivariate extension of the g-and-k distribution introduced in \citep{Drovandi2011} and used as a benchmark in \citep{Li2017Copula, Jiang2018, Nguyen2020}. This flexible parametric family of distributions does not have a closed-form likelihood, but is easy to simulate from.
We define a distribution in this family through $ (G_\theta, \bU)$, where
\begin{align*}
     G_{\theta}(u)  =   \theta_1 \hspace{-0.5ex} +  \theta_2 \hspace{-0.5ex} \left[1 \hspace{-0.5ex} + \hspace{-0.5ex} 0.8 \frac{1 - \exp(-\theta_3 z(u))}{1 + \exp(-\theta_3 z(u))}\right] \hspace{-0.5ex} \left(1 \hspace{-0.5ex} + \hspace{-0.5ex} z(u)^2 \right)^{\theta_4} \hspace{-0.5ex} z(u),
\end{align*}
with $\theta=(\theta_1,\theta_2,\theta_3,\theta_4,\theta_5)$, $z(u) = \Sigma^{\frac{1}{2}}u$ and
$\bU = \calN(0, I_r)$, where $\Sigma \in \bR^{d \times d}$ is a symmetric tridiagonal Toeplitz matrix such that $\Sigma_{ii} = 1$ and $\Sigma_{ij} = \theta_5$.
The parameters $\theta_1$,$\theta_2$,$\theta_3$, and $\theta_4$ govern the location, scale, skewness, and kurtosis respectively, and $r = d$.
An alternative formulation is through  $(\tilde{\bU},\tilde{G}_\theta)$ where $\tilde{\bU} = \text{Unif}(0,1)^r$, and $\tilde{G}_\theta = G_{\theta} \circ \Phi^{-1}$ where $\Phi$ is the cumulative distribution function of a $\calN(0,1)$.

\begin{figure*}
     \centering
     \begin{subfigure}[b]{0.51\textwidth}
         \centering
         \includegraphics[width=\textwidth]{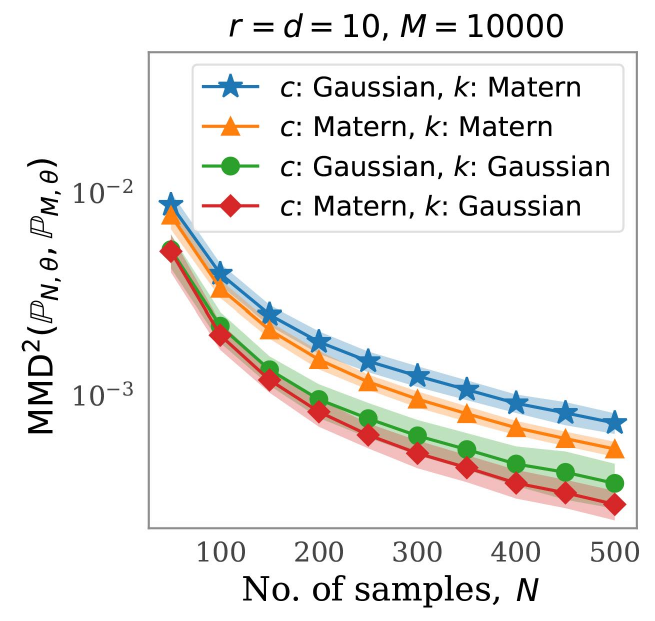}
         \caption{}
         \label{fig:mvgk_a}
     \end{subfigure}
     \begin{subfigure}[b]{0.47\textwidth}
         \centering
         \includegraphics[width=\textwidth]{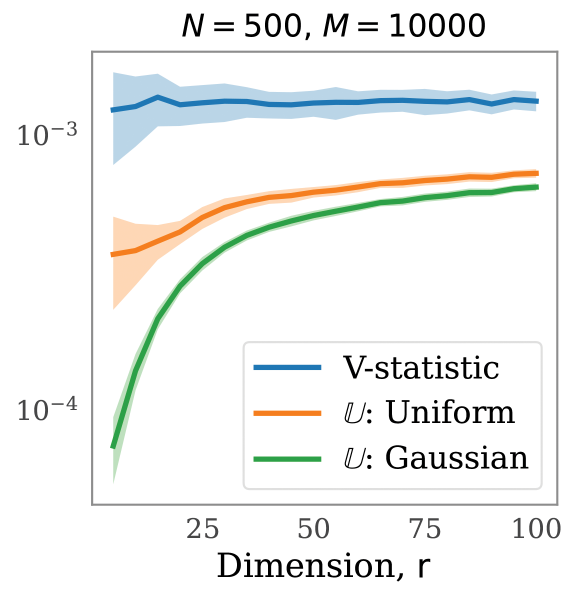}
         \caption{}
         \label{fig:mvgk_b}
     \end{subfigure}
     \vspace*{1cm}
     \hspace*{0.001\textwidth}
     \begin{subfigure}[b]{0.5\textwidth}
         \centering 
         \includegraphics[width=\textwidth]{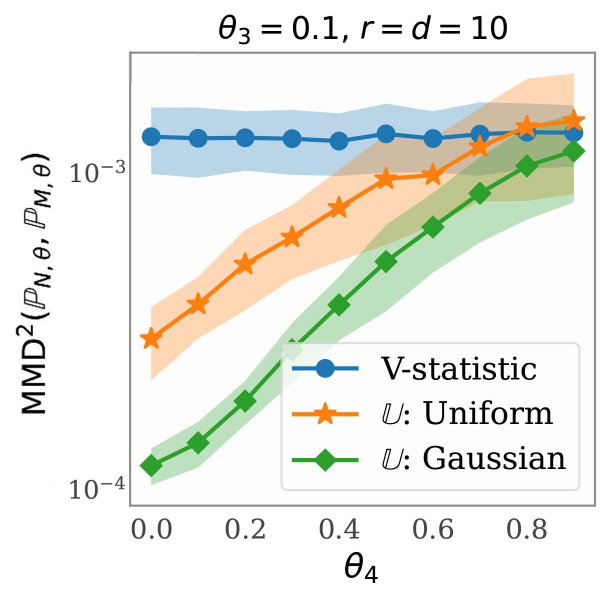}
         \caption{}
         \label{fig:mvgk_c}
     \end{subfigure}
     \begin{subfigure}[b]{0.475\textwidth}
         \centering
         \includegraphics[width=\textwidth]{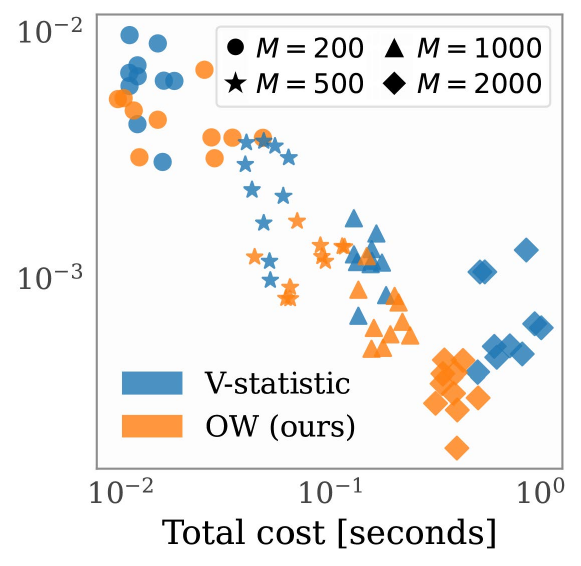}
         \caption{}
         \label{fig:mvgk_d}
     \end{subfigure}
        \caption[Error in estimating MMD$^2$ for the multivariate g-and-k distribution.]{Error in estimating MMD$^2$ for the multivariate g-and-k distribution. (a) Error of our OW estimator for different choices of $k$ and $c$. Increasing the smoothness of $k$ improves the performance. (b) Comparison of V-statistic and OW estimator as a function of dimension. OW performs better for both parameterisations of $\bU$, with the Gaussian giving lowest error. (c) Value of $\theta_4$ also impacts the performance of the OW estimator. (d) Error vs. total computation cost for different $M$. OW performs better than V-statistic for similar cost: $N=M$ for V-statistic, whereas $N = (68, 126, 200, 317)$ for OW.}
        \label{fig:mvgk_all}
\end{figure*}

\paragraph{Varying choice of $k$ and $c$.}
We first investigate the performance of our OW estimator for different combinations of $k$ and $c$, the choices being either the Gaussian or the Mat\'ern kernel. We estimate the squared MMD for each combination as a function of $N$, with $d = 10$ and $M=10,000$. The Lebesgue measure formulation is used when computing the embeddings for both kernels. The Mat\'ern kernels are set to orders $\nu_k = \nu_c = 2.5$, and the parameter value to $\theta_0=(3,1,0.1,0.1,0.1)$. The resulting curves are shown in \Cref{fig:mvgk_a}. Our method performs best when $k$ is the Gaussian kernel, i.e., infinitely smooth. The performance degrades slightly when $k$ is Mat\'ern, while the combination of $c$ as the Gaussian and $k$ as the Mat\'ern kernel is the worst. This is expected from our theory because the composition of $G_\theta$ and $k$ is not smooth, but we approximate it with an infinitely smooth function. Hence, choosing a very smooth $k$ is always beneficial from a computational viewpoint.

\paragraph{Varying dimensions $r$ and $d$.}
We now analyse the impact of the choice of measure, either Gaussian or uniform. \Cref{fig:mvgk_b} shows the OW and V-statistic estimators as  the dimension $r=d$ varies. The parameter values are the same as before, $N = 500$, and the Gaussian kernel is used for both $k$ and $c$. We observe that the OW estimator performs better than the V-statistic even in dimensions as high as 100. In lower dimensions, the Gaussian embedding achieves lower error than the uniform for this model, with their performance converging around $d = 60$. This is likely due to the fact that $\tilde{G}_\theta$ is an easier function to approximate than $G_\theta$, but this is harder to assess a priori for the user and highlights some open questions not yet covered by our theory.

\paragraph{Varying model parameters.}
Building on the previous result, we show that the performance of the OW estimation is also impacted by $\theta$. In \Cref{fig:mvgk_c}, we analyse the performance of the estimators as a function of parameter $\theta_4$. The Gaussian kernel is used for both $k$ and $c$. While the V-statistic is not impacted by the choice of $\theta_4$, the performance of our estimators degrades as $\theta_4$ increases. We expect that this difference in performance is due to the regularity of the generator varying with $\theta$.

\paragraph{Performance vs.\ computational cost.}
Finally, since the OW estimator tends to be more computationally expensive and this simulator is relatively cheap ($\approx$ 1~ms to generate one sample), we also compare estimators for a fixed computational budget. To that end, we vary $M$ and take $N=M$ for the V-statistic and $N = 2M^{2/3}$ for the OW estimator. \Cref{fig:mvgk_d} shows their performance with respect to their total computational cost, including the cost of simulating from the model ($d = r = 5$). We see that the OW estimator achieves lower error on average than the V-statistic. Hence, it is preferable to use the OW estimator even for a computationally cheap simulator like the multivariate g-and-k.

\paragraph{Composite goodness-of-fit test.}
We demonstrate the performance of our method when applied to composite goodness-of-fit testing, using the method proposed by \citet{Key2025} with a test statistic based on the squared MMD.
Given i.i.d. draws from some distribution $\bQ$, the test considers whether  $\bQ$ is an element of some parametric family $\{\bP_{\theta}: \theta \in \Theta\}$ (null hypothesis) or not (alternative hypothesis).
The approach uses a parametric bootstrap \citep{Stute1993} to estimate the distribution of the squared MMD under the null hypothesis, which can then be used to decide whether or not to reject. This requires repeatedly performing two steps: (i) estimating a parameter value through an MMD estimator of the form in~\eqref{eq:MMD_estimator}, and (ii) estimating the squared MMD between $\bQ$ and the model at the estimated parameter value. See \Cref{app:gof} for the full algorithm. This needs to be done up to $B$ times, where $B$ can be hundreds or thousands, which can be a significant challenge computationally. This limits the number of simulated samples $N$ that can be used at each step, and is therefore a prime use case for our OW estimator.

We performed this test with a level of $0.05$ using the V-statistic and OW estimator, with $B=200$. We considered the multivariate g-and-k model with unknown $\theta_1, \theta_2, \theta_3$, and $\theta_5$ but fixed $\theta_4 = 0.1$. We used $N=100$ and $M=500$ and considered two cases: $\bQ$ is a multivariate g-and-k with $\theta_4=0.1$ (null holds) or $\theta_4=0.5$ (alternative holds). When the null hypothesis holds, we should expect the tests to reject the null at a rate close to $0.05$, whereas when the alternative holds, we should reject at a rate close to $1$.
\Cref{tab:composite_test_results} shows that our test based on the OW estimator performs significantly better in that respect than the V-statistic. This is due to the fact that the OW estimator is able to improve both the estimate of the parameter and the estimate of the test statistic, thus improving the overall performance.

\begin{table}[]
    \centering
    \caption[Composite goodness-of-fit test with V-statistic vs. OW-MMD.]{Fraction of repeats for which the null was rejected. An ideal test would have $0.05$ when the null holds, and $1$ otherwise. }
    \resizebox{0.8\columnwidth}{!}{%
    \begin{tabular}{c c c}
        \toprule
        Cases & i.i.d. V-stat & i.i.d. OW (ours)  \\ \midrule
         $\theta_4 =0.1$ (null holds) & 0.040 & 0.047 \\
         $\theta_4 =0.5$ (alternative holds) & 0.040 & 0.413 \\
        \bottomrule
    \end{tabular}
    }
    \label{tab:composite_test_results}

\end{table}

\tocless\subsection{Large-scale offshore wind farm model}

Finally, we consider a low-order wake model \citep{Niayifar2016, Kirby2023} for large-scale offshore wind farms. The model simulates an estimate of the farm-averaged local turbine thrust coefficient \citep{Nishino2016}, which is an indicator of the energy produced. The parameter $\theta$ is the angle (in degrees) at which the wind is blowing. The turbulence intensity is assumed to have zero-mean additive Gaussian noise (meaning $\bU = \calN(0,10^{-3})$), which then goes through the non-linear mapping of the generator. Although this model is an approximation of the state-of-the-art models that can take around 100 CPU hours per run~\citep[e.g.,][]{Kirby2022}, one sample from this model takes $\approx 2$~minutes, which is still computationally prohibitive for likelihood-free inference. 
This example is indicative of the expensive simulators widely used in science and is thus suitable for our method. 

We apply the ABC method of~\eqref{eq:ABC} to estimate $\theta$ with both the OW estimator and the V-statistic. The tolerance threshold $\varepsilon$ is taken in terms of percentile, i.e., the proportion of the data that yields the least MMD distances. We use $1000$ parameter values from the $\text{Unif}(0,30)$ prior on $\Theta$. As the cost of the model far exceeds that of estimating the MMD, we take $N=10$ for both estimators. With a small $N$, setting the lengthscale of $c$ using the median heuristic is difficult, so we fix it to be 1.  The simulated datasets took $\approx 245$~hours to generate, while estimating the MMD took around $0.13$~s and $0.36$~s for the V-statistic and the OW estimator, respectively. 

\begin{figure}[t!]
	 \centering
	 \includegraphics[ width = 0.4 \linewidth]{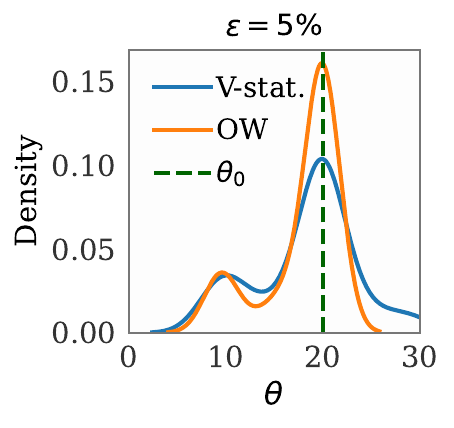}
	 \includegraphics[ width = 0.365 \linewidth]{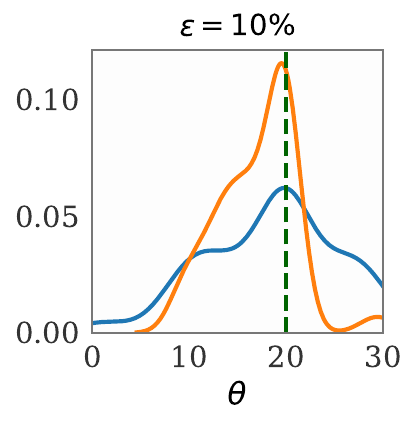}
\caption[ABC posteriors for the wind farm model.]{ABC posteriors for the wind farm model. Our OW estimator yields posterior samples that are more concentrated around the true $\theta_0$ than the V-statistic. Settings: $M=100$, $\theta_0=20$.}
\label{fig:windfarm}
\end{figure}

The resulting posteriors, which are approximations of the ABC posterior obtained if the MMD were computable in closed-form, are shown in \Cref{fig:windfarm}. We observe that the OW estimator's posterior is much more concentrated around the true value than that of the V-statistic for both values of $\varepsilon$. This is because the OW estimator approximates the MMD more accurately than the V-statistic for the same $N$. Hence, our method can achieve similar performance to the V-statistic with much smaller $N$, saving hours of computation time. 

\section{Conclusion}

We proposed an optimally-weighted MMD estimator which has improved sample complexity over the V-statistic when the generator and kernel are smooth, and the dimensionality is small or moderate. Thus, our estimator requires fewer datapoints than alternatives in this setting, making it especially advantageous for computationally expensive simulators, which are widely used in the natural sciences and engineering. However, a number of open questions remain, and we highlight the most relevant below. 

The parameterisation of a simulator through a generator $G_\theta$ and a measure $\bU$ is usually not unique, and it is often unclear which parameterisation is most amenable to our method. One approach would be to choose a parameterisation where the dimension of $\bU$ is small so as to improve the convergence rate. However, our result in \Cref{thm:rate_of_convergence} also contains rate constants which are difficult to get a handle on, and it is therefore difficult to identify which parameterisation is best among those with fixed smoothness and dimensionality.

Finally, our sample complexity result could be extended. One limitation is that we focus on the MMD and not its gradient, meaning that our results are not directly applicable to gradient-based likelihood-free inference, such as the method used for our g-and-k example \citep{briol2019statistical}. A future line of work could also investigate if our ideas translate to other distances used for likelihood-free inference, such as the Wasserstein distance \citep{Bernton2019} and Sinkhorn divergence \citep{Genevay2017,Genevay2019}.

\chapter{MMD-based Estimators for Conditional Expectations}
\label{sec:mmdintegration}
\begin{tcolorbox}
The results in this chapter were published in the following paper:

\begin{itemize}
    \item Chen, Z*., Naslidnyk*, M., Gretton, A., \& Briol, F.-X. (2024). Conditional Bayesian Quadrature. Uncertainty in Artificial Intelligence.
\end{itemize}

with * indicating equal contribution. All theoretical results in this chapter are due to the author. Experiments were carried out by Zonghao Chen and are included to support the theory numerically.
\end{tcolorbox}

In this chapter we study the problem of estimating \emph{parametric} expectations that appear across machine learning, statistics, and science and engineering. Let $f$ be a real-valued function on $\calX \times \Theta$ for sets $\calX$ and $\Theta$, and $\{\bP_\theta\}_{\theta\in\Theta}$ be a family of probability distributions on $\calX$. For each $\theta$, we aim to approximate
\begin{align}
\label{eq:cbq_integral}
    I(\theta) = \bE_{X \sim \bP_\theta}[f(X,\theta)]=\int_\calX f(x, \theta) \bP_\theta(\d x).
\end{align}
An important subclass of parametric expectations are conditional expectations, when $\bP_\theta(\cdot) = \bP(\cdot | \theta)$ for some conditional distribution $\bP$. 
We assume $I$ is sufficiently smooth in $\theta$ so that $I(\theta)$ and $I(\theta')$ are close when $\theta$ and $\theta'$ are close.

Expectations of this type appear in rare-event simulation for tail probabilities \citep{Tang2013}, in moment-generating, characteristic, and cumulative distribution computations \citep{Giles2015,Krumscheid2018}, in computing conditional value-at-risk and option pricing \citep{longstaff2001valuing,alfonsi2022many}, and both Bayesian \citep{Lopes2011,Kallioinen2021} and general global sensitivity analysis \citep{Sobol2001}. They also enter as inner quantities in nested expectations of the form $\bE_{\theta\sim\bQ}[\phi(I(\theta))]$~\citep{Hong2009,Rainforth2018}, which occur in expected information gain for Bayesian experimental design \citep{Chaloner1995} and in expected value of partial perfect information \citep{heath2017review}.

A common workflow is to select $T$ parameter values $\theta_1, \dots, \theta_T$ and, for each $t$, draw $N$ samples from $\bP_{\theta_t}$ to evaluate $f$, for a total budget of $NT$ evaluations. Standard numerical integration methods, such as Monte Carlo or MMD minimisation reviewed in \Cref{sec:kernel_and_bayesian_quadrature}, only approximate $I(\theta_1), \dots, I(\theta_T)$. However, applications frequently require that $I(\theta)$ for a $\theta \notin \{\theta_1,\dots,\theta_T\}$ is estimated, or even that $I(\theta)$ is estimated uniformly over  all $\theta \in \Theta$; this creates a need for a second step that uses $I(\theta_1), \dots, I(\theta_T)$ to estimate $I(\theta)$.

As will be reviewed in detail in \Cref{sec:computing_conditional_expectations}, methods that estimate $I(\theta)$ suffer from the following drawbacks.
\begin{enumerate}
    \item \textbf{High sample complexity.} Accurate estimates typically demand large $N$ and $T$, which is prohibitive when sampling or evaluating $f$ is expensive.
    \item \textbf{Lack of uncertainty quantification.} Obtaining a finite-sample, per-$\theta$ quantification of uncertainty for $I(\theta)$ is often not feasible.
\end{enumerate}

\begin{figure}[t]
\centering
\includegraphics[width=260pt]{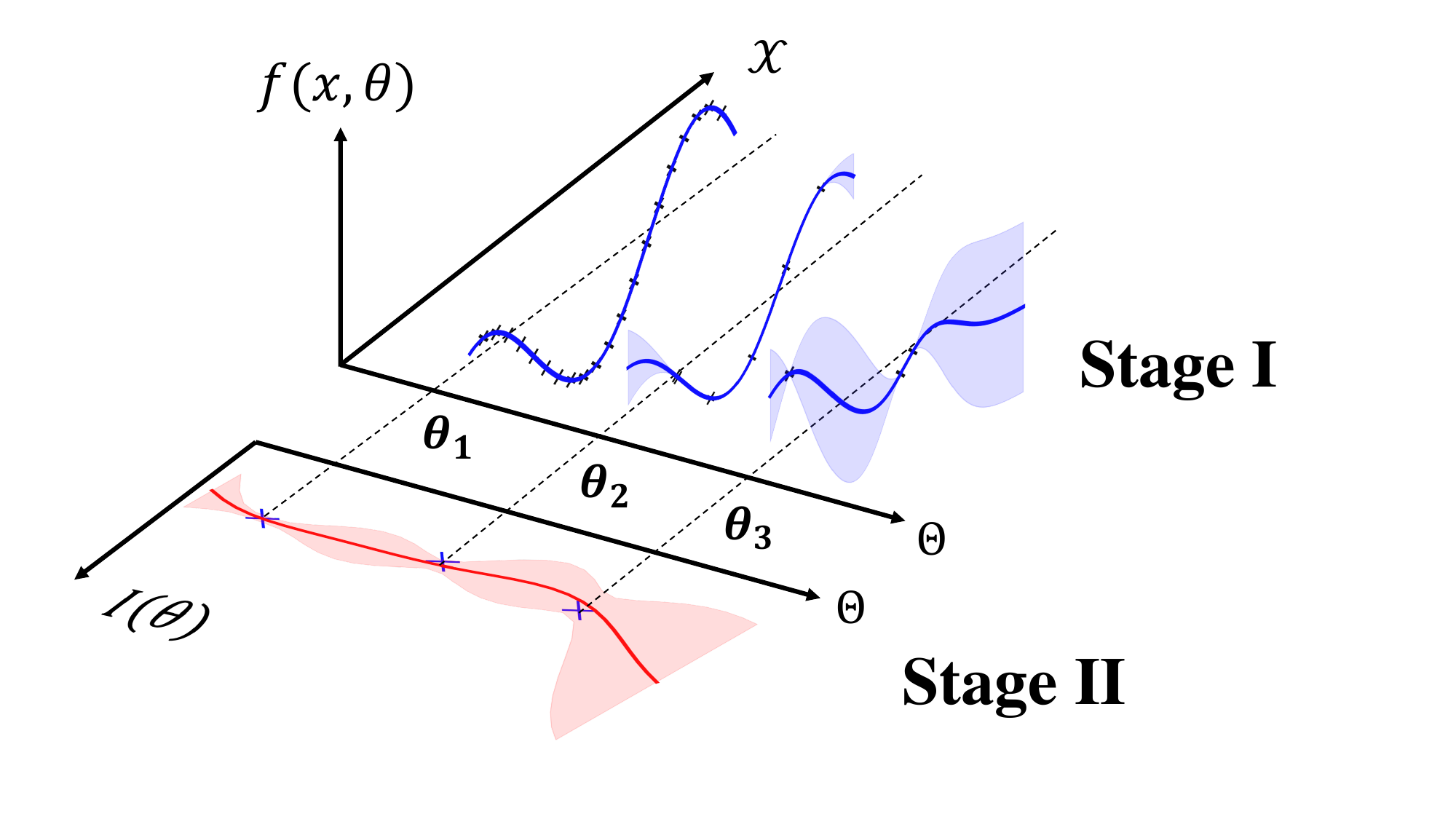}
\caption[Illustration of \textit{conditional Bayesian quadrature} (CBQ).]{Illustration of \textit{conditional Bayesian quadrature} (CBQ). Stage one fits a GP to $f(x,\theta_t)$ for each $\theta_t \in \{\theta_1,\dots,\theta_T\}$ and integrates to obtain BQ estimates \textcolor{blue}{$I_{\BQ}(\theta_1),\dots, I_{\BQ}(\theta_T)$}. Stage two places a GP over $\theta\mapsto I(\theta)$ and fuses these to yield \textcolor{red}{$I_{\mathrm{CBQ}}(\theta)$} with posterior uncertainty shown by shaded regions.}
\label{fig:illustration}
\end{figure}

To tackle these limitations, we propose \emph{conditional Bayesian quadrature} (CBQ), a two-stage probabilistic numerical method that extends Bayesian quadrature \citep{Diaconis1988,OHagan1991,Rasmussen2003,Briol2019PI} to parametric expectations. CBQ places a hierarchical Gaussian process (GP) prior: stage one produces GP posteriors for $f(\cdot,\theta_t)$ and integrates them to obtain BQ estimates $I_{\BQ}(\theta_t)$; stage two places a GP over $\theta\mapsto I(\theta)$ and combines the first-stage integrals to produce a univariate Gaussian posterior for $I(\theta)$ at any $\theta$, with a mean and variance parametrised by $\theta$. We illustrate the approach in~\Cref{fig:illustration}.

CBQ addresses the two limitations above in the following way. First, under mild smoothness assumptions on $f$ and $I$, and when the dimensions of $\calX$ and $\Theta$ are moderate, we show both theoretically and empirically that CBQ converges rapidly and is therefore more sample efficient than the baselines, enabling target accuracies with smaller $N$ and $T$. Second, the GP posterior for $I(\theta)$ delivers finite-sample Bayesian uncertainty quantification.

\section{Computing Conditional Expectations}
\label{sec:computing_conditional_expectations}

Let $\calX \subseteq \bR^d$, $\Theta \subseteq \bR^p$, and $f(\cdot,\theta) \in \calL^2(\bP_\theta)$ for all $\theta \in \Theta$, a minimal integrability assumption which ensures that Monte Carlo estimators satisfy the central limit theorem. We aim to compute the parametric expectation~\eqref{eq:cbq_integral} using evaluations $x_{1:N}^t, f(x_{1:N}^t)$, where $x_1^t, \dots, x_N^t \sim \bP_{\theta_t}$ for all $t \in \{1,\dots,T\}$. Allowing $N_t$ to vary by $t$ is a straightforward extension, but we keep $N$ fixed for any $t$ for notational simplicity. 

Existing methods of computing parametric expectations fall into two categories: sampling-based methods and regression-based methods.
\paragraph{Sampling-based Methods.} 
The \emph{Monte Carlo} (MC) estimator \citep{Robert2004} of $I$ takes the form
$I_{\mathrm{MC}}(\theta_t)\coloneqq\frac{1}{N}\sum_{i=1}^N f(x_i^t,\theta_t)$. As mentioned at the beginning of the chapter, MC cannot estimate $I(\theta)$ for $\theta \notin \{\theta_{1}, \dots,\theta_T \}$; furthermore, for every $t$, it uses only the $N$ samples at $\theta_t$ to construct the estimator, ignoring the evaluations at nearby points $\theta_{t'}$. In the special case when $f$ does not depend on $\theta$, i.e., $f(x, \theta) = f(x)$ for all $\theta \in \Theta$, \emph{importance sampling} (IS, \citet{Glynn1989,Madras1999,Tang2013,Demange-Chryst2022}) offers a more suitable alternative. Provided each $\bP_\theta$ admits a Lebesgue density $p_\theta$ with full support on $\calX$, IS estimates $I(\theta)$ for any $\theta \in \Theta$ using all $NT$ samples as
\begin{equation*}
    I_\mathrm{IS}(\theta) \coloneqq \frac{1}{TN} \sum_{t=1}^T \sum_{n=1}^N \frac{p_{\theta}(x_n^t)}{p_{\theta_t}(x_n^t)} f(x_n^t),
\end{equation*}
a sum that weighs each $f(x_n^t)$ by $p_{\theta}(x_n^t)/p_{\theta_t}(x_n^t)$ to account for the fact that $x_n^t$ was not obtained from $\bP_\theta$, but from an importance distribution $\bP_{\theta_t}$. Unfortunately, this approach does not apply when $f$ depends on $\theta$, and it is usually difficult to identify which importance distributions lead to an accurate estimator for small $N$ and $T$.
 
\paragraph{Regression-based Methods.}
The main regression-based method is least-squares Monte Carlo (LSMC) \citep{longstaff2001valuing}, which proceeds in two stages: (1) compute MC estimators $I_\mathrm{MC}(\theta_1), \dots, I_\mathrm{MC}(\theta_T)$ (2) estimate $I(\theta)$ with polynomial regression using $\theta_{1:T}, I_\mathrm{MC}(\theta_{1:T})$. \citet{Han2009,Hu2020} suggested a more flexible alternative, replacing the second stage with kernel ridge regression; we will refer to this method as kernelised LSMC (KLSMC). Notably, KLSMC coincides with standard estimators for conditional kernel mean embeddings based on vector-valued kernel ridge regression~\citep{Park20:CME}, and is a generalisation of the kernel mean shrinkage estimators of \citet{muandet2016kernelmeanshrinkage,chau2021deconditional}.
In LSMC and KLSMC, the regression method in the second stage affects both the performance and computational cost. LSMC with polynomial order $q$ has cost $\bigo(TN+q^3)$, whereas KLSMC costs $\bigo(TN+T^3)$; however, KLSMC is typically more accurate when $I(\theta)$ is not well approximated by a low-order polynomial.

\paragraph{Other Related Work.} Multi-task and meta-learning approaches \citep{xi2018bayesian,gessner2020active,Sun2021,Sun2023} estimate $I(\theta)$ by treating it as a multi-task objective, and encode task relationships via vector-valued RKHSs or task priors. These methods generally do not model the specific mapping $\theta \mapsto I(\theta)$ and are suboptimal in our setting.
Multilevel Monte Carlo \citep{Giles2015} is also popular in estimating expensive expectations, as it reduces cost by combining simulations at multiple resolutions. However, it is not able to provide estimates at $\theta \notin \{\theta_{1}, \dots,\theta_T \}$ or $I(\theta)$ uniformly over $\theta \in \Theta$.

\section{Conditional Bayesian Quadrature}
\label{sec:cbq}

\emph{Conditional Bayesian quadrature} (CBQ) provides a Bayesian hierarchical model for $I(\theta)$ for any $\theta \in \Theta$; the posterior mean of this model is the \emph{CBQ estimator}. CBQ is a regression-based method that proceeds in two stages: Stage 1 performs Bayesian quadrature (BQ, see~\Cref{sec:bq_background}) on $I(\theta_1), \dots, I(\theta_T)$, and Stage 2 uses the estimates and uncertainty quantification in Stage 1 to perform GP regression (see~\Cref{sec:gp-regression}) on $I(\theta)$. We describe the method now.
\paragraph{Stage 1:} For every $t \in \{1, \dots,T\}$, compute the BQ posterior mean $I_\BQ(\theta_t)$ and variance $\sigma^2_\BQ(\theta_t)$ over $I(\theta_t)$ conditioned on $(x^t_{1:N}, f(x^t_{1:N}, \theta_t))$ as in~\eqref{eq:bq-posterior-moments},
\begin{equation*}
\begin{split}
    I_\BQ(\theta_t) & = \mu_{k_\calX,\bP_{\theta_t}}(x^t_{1:N})^\top k_\calX(x^t_{1:N}, x^t_{1:N})^{-1} f(x^t_{1:N}, \theta_t), \\
    \sigma^2_\BQ(\theta_t) &= \bE_{X, X' \sim \bP_{\theta_t}} [k_\calX(X, X')] - \mu_{k_\calX,\bP_{\theta_t}}(x^t_{1:N})^\top k_\calX(x^t_{1:N}, x^t_{1:N}) ^{-1} \mu_{k_\calX,\bP_{\theta_t}}(x^t_{1:N}).
\end{split}
\end{equation*}
\paragraph{Stage 2:} Perform GP regression over $I(\theta)$ with a zero prior mean, data $(\theta_{1:T}, I_\BQ(\theta_{1:T}))$, and an additive noise model $\varepsilon_t \sim \calN(0,\lambda_\Theta + \sigma^2_\BQ(\theta_t))$, for $t \in \{1, \dots, T\}$. As in~\eqref{eq:posterior-moments}, the posterior mean and covariance 
are given by
\begin{align*}
\begin{aligned}
    I_{\mathrm{CBQ}}(\theta)  &\coloneqq k_\Theta(\theta, \theta_{1:T}) \big(k_\Theta(\theta_{1:T}, \theta_{1:T}) + \diag(\lambda_\Theta+ \sigma^2_\BQ(\theta_{1:T}))\big)^{-1} I_\BQ(\theta_{1:T}), \\
    k_{\mathrm{CBQ}}(\theta,\theta')  &\coloneqq k_{\Theta}(\theta,\theta') - k_\Theta(\theta,\theta_{1:T}) \big( k_{\Theta}(\theta_{1:T}, \theta_{1:T}) \\ &\hspace{2cm} + \diag(\lambda_\Theta+ \sigma^2_\BQ(\theta_{1:T})) \big)^{-1} k_\Theta(\theta_{1:T},\theta').
\end{aligned}
\end{align*}
\Cref{fig:illustration} summarises CBQ. The posterior mean $I_\mathrm{CBQ}(\theta)$ is the CBQ estimator for $I(\theta)$, and the variance $k_{\mathrm{CBQ}}(\theta,\theta)$ quantifies its uncertainty. 
Stage 1 uses the kernel $k_{\mathcal X}$ for Bayesian quadrature over $x$, encoding prior structure in the map $x \mapsto f(x, \theta)$ for every $\theta \in \Theta$. Stage 2 places a zero-mean GP over $\theta$ with kernel $k_\Theta$, encoding prior structure in the map $\theta \mapsto I(\theta)$. The closer the properties of the prior match its corresponding function, the faster the estimator $I_\mathrm{CBQ}$ converges to the true $I$; this is formalised in~\Cref{sec:theory_cbq}. The data $\{(x_{1:N}^t, f(x_{1:N}^t,\theta_t))\}_{t=1}^T$ enters the posterior through the Stage 1 evaluations $I_{\mathrm{BQ}}(\theta_{1:T})$, which are independent since each $I_\BQ(\theta_t)$ is a deterministic function of independent samples $\theta_t, x^t_1, \dots, x^t_N$ across $t = 1, \dots, T$. The CBQ estimator can then be written as a quadrature rule,
\begin{align*}
\begin{aligned}
    I_{\mathrm{CBQ}}(\theta) &= \sum_{t=1}^T\sum_{n=1}^N w_t  v^t_n  f(x^t_n, \theta_t), \\
    w^\top_{1:T} &= k^\top_\Theta(\theta, \theta_{1:T}) \big(k_\Theta(\theta_{1:T}, \theta_{1:T}) + \diag(\lambda_\Theta+ \sigma^2_\BQ(\theta_{1:T}))\big)^{-1}, \\
    {v^t_{1:N}}^\top &= \mu_{k_\calX,\bP_{\theta_t}}(x^t_{1:N})^\top k_\calX(x^t_{1:N}, x^t_{1:N})^{-1}.
\end{aligned}
\end{align*}
The Stage 2 GP regression is heteroscedastic~\citep{Le2005}: we use the uncertainty estimate in Stage 1 to inform the noise model in Stage 2; in particular, when the number of samples $N$ grows, the BQ variance $\sigma^2_\BQ(\theta_t)$ will decrease, thus reducing the noise in Stage 2 as we are more certain about $I(\theta_t)$. The term $\lambda_\Theta$ is a `jitter' or `nugget' term introduced for numerical stability; we explain its role in~\Cref{sec:theory_cbq}. Heteroscedasticity has previously been shown to be common in practice for LSMC \citep{Fabozzi2017}. 

CBQ is closely related to  LSMC and KLSMC as it simply corresponds to different choices for the two stages. 
The main difference is in Stage 1, where we use BQ rather than MC. This is where we expect the greatest gains for our approach due to the fast convergence rate of BQ estimators (this will be confirmed in \Cref{sec:theory_cbq}). For Stage 2, we use heteroscedastic GP regression rather than polynomial or kernel ridge regression. As such, the second stage of KLSMC and CBQ is identical up to a minor difference in the way in which the Gram matrix $k_{\Theta}(\theta_{1:T}, \theta_{1:T})$ is regularised before inversion. 
Finally, one significant advantage of CBQ over LSMC and KLSMC is that it is a fully Bayesian model: we obtain a posterior distribution on $I(\theta)$ for any $\theta \in \Theta$.

The total computational cost of our approach is $\mathcal O(TN^3+T^3)$: $T$ BQ estimators of size $N$ (Stage 1) and a GP of size $T$ (Stage 2). We avoid sparse/variational GP approximations \citep{titsias2009variational} because an additional approximation layer can slow the asymptotic convergence of CBQ. Although CBQ is more expensive than LSMC ($\mathcal O(TN+q^3)$) or KLSMC ($\mathcal O(TN+T^3)$), \Cref{thm:convergence} and \Cref{sec:experiments_cbq} show that its faster convergence in $N$ and $T$ often offsets the higher per-iteration cost, especially when integrand evaluations dominate compute, where a data-efficient method like CBQ is preferable.

Interestingly, CBQ also provides us with a joint Gaussian posterior on the expectation at $\theta^\ast_1, \ldots, \theta^\ast_{T_{\text{Test}}} \in \Theta$ which has mean vector $I_{\mathrm{CBQ}}(\theta^\ast_{1:T_{\text{Test}}})$ and covariance matrix $k_{\mathrm{CBQ}}(\theta^\ast_{1:T_{\text{Test}}},\theta^\ast_{1:T_{\text{Test}}})$. This can be computed at an  $\bigo(T^2 T_{\text{test}})$ cost; as observed, CBQ takes into account covariances between test points in that the integral value will be similar for similar parameter values, whereas standard BQ treats each integral value independently.

A natural alternative would be to place a GP prior directly on $(x,\theta) \mapsto f(x,\theta)$ and condition on all 
$N \times T$ observations. 
The implied distribution on $I(\theta_1), \ldots, I(\theta_T)$ would also be a multivariate Gaussian distribution. 
This approach coincides with the multi-output Bayesian quadrature (MOBQ) approach of \citet{xi2018bayesian} where multiple integrals are considered simultaneously. 
However, the computational cost of MOBQ is $\bigo(N^3 T^3)$, due to fitting a GP on $N T$ observations, and quickly becomes intractable as $N$ or $T$ grow. 
A further comparison of BQ and MOBQ can be found in~\Cref{appendix:cbq_mobq}.
The same holds if $f$ does not depend on $\theta$, in which case the task reduces to the conditional mean process studied in Proposition 3.2 of \citet{chau2021deconditional}, and when $T=1$, we recover standard Bayesian quadrature. 

\paragraph{Hyperparameters.}

CBQ has two sets of hyperparameters: Bayesian quadrature hyperparameters in stage 1, and GP regression hyperparameters in stage 2. We use the following procedure. First, to justify the choice of zero mean priors and stabilise both steps, we standardise the function values by subtracting the empirical mean and dividing by the empirical standard deviation. Then, the kernels $k_\calX$ and $k_\Theta$ are chosen to reflect properties of $x \mapsto f(x, \theta)$ and $\theta \mapsto f(x, \theta)$ (such as smoothness), and to ensure closed-form kernel means are available, as discussed in~\Cref{sec:bq_background}. We will suggest and motivate a specific kernel choice in \Cref{sec:theory_cbq}. Finally, the remaining hyperparameters, the regulariser $\lambda_\Theta$ and the parameters of the kernels, are selected by maximum likelihood, as described in detail in \Cref{appendix:hyperparameter_selection}.

\section{Theoretical Guarantees}\label{sec:theory_cbq}

Our main theoretical result in \Cref{thm:convergence} below guarantees that CBQ is able to recover the true value of $I(\theta)$ when $N$ and $T$ grow. The result of this theorem depends on the smoothness of the problem, and the smoothness of the kernels; we will use Sobolev kernels introduced in~\Cref{def:sobolev_kernel} to quantify the latter. For a multi-index $\alpha = (\alpha_1, \dots, \alpha_p) \in \bN^p$, by $D_\theta^\alpha f$ we denote the $|\alpha|=\sum_{i=1}^p \alpha_i$-th order weak derivative of a function $f$ on $\Theta$. 

\begin{theorem}\label{thm:convergence}
Let $x \mapsto f(x, \theta)$ be a function of smoothness $s_f > d/2$, and $\theta \mapsto f(x, \theta)$ be a function of smoothness $s_I > p/2$ such that $\sup_{\theta \in \Theta} \max_{|\alpha|\leq s_I} \| D_\theta^\alpha f(\cdot, \theta) \|_{\calW^{s_f, 2}(\calX)}<\infty$. Suppose the following assumptions hold.
\begin{enumerate}[itemsep=0.1pt,topsep=0pt,leftmargin=*]
\item [A1] The domains $\calX \subset \bR^d$ and $\Theta\subset \bR^p$ are open, convex, and bounded.
\customlabel{as:domains}{A1}
\item [A2] The parameters and samples satisfy: 
$\theta_1, \dots, \theta_T \sim \bQ$, and $x_{1:N}^t \sim \bP_{\theta_t}$ for all $t \in \{1,\ldots, T\}$. 
\customlabel{as:pars_and_samples}{A2}
\item [A3] $\bQ$ has density $q$ such that $\inf_{\theta \in \Theta} q(\theta)>0$ and $\sup_{\theta \in \Theta} q(\theta) < \infty$, and $\bP_\theta$ has density $p_\theta$ such that $\theta \mapsto p_\theta(x)$ is of smoothness $s_I > p/2$, and for any $\theta \in \Theta$, it holds that $\inf_{\theta \in \Theta, x \in \calX} p_{\theta}(x)>0$ and $\sup_{\theta \in \Theta}\max_{|\alpha|\leq s_I} \|D_\theta^\alpha p_\theta(\cdot)\|_{\calL^\infty(\calX)}<\infty$.
\customlabel{as:densities}{A3}
\item [A4] $k_\calX$ and $k_\Theta$ are Sobolev kernels of smoothness $s_\calX \in (d/2, s_f]$ and $s_\Theta \in (p/2, s_I]$ respectively.
\customlabel{as:kernels_cbq}{A4}
\item [A5] The regulariser satisfies $\lambda_\Theta = \bigo(T^{\nicefrac{1}{2}})$.
\customlabel{as:regulariser}{A5}
\end{enumerate}
Then, we have that for any $\delta \in (0, 1)$ there is an $N_0>0$ such that for any $N \geq N_0$ with probability at least $1-\delta$ it holds that
\begin{align*}
    \left\| I_\mathrm{CBQ} - I \right\|_{\calL^2(\Theta)}
    \leq  C_0(\delta) N^{-\nicefrac{s_\calX}{d} + \varepsilon} + C_1(\delta) T^{-\nicefrac{1}{4}}  ,
\end{align*}
for any arbitrarily small $\varepsilon>0$, and the constants $C_0(\delta)=\bigo(1/\delta)$ and $C_1(\delta)=\bigo(\log(1/\delta))$ are independent of $N, T, \varepsilon$.
\end{theorem}
To prove the result, we represent the CBQ estimator as a \emph{noisy importance-weighted kernel ridge regression} (NIW-KRR) estimator. Then, we extend convergence results for the \emph{noise-free} IW-KRR estimator established in~\citet[Theorem 4]{gogolashvili2023importance} to bound Stage 2 error in terms of the error in Stage 1, which in turn we bound via results on the convergence of GP interpolation from \citet{wynne2021convergence}. See~\Cref{appendix:convergence_rate} for the proof.

We now briefly discuss the assumptions.
\ref{as:domains} ensures the point sets eventually cover the domain. \ref{as:pars_and_samples} requires that $\theta_{1:T}$ and $x_{1:N}^t$ fill $\Theta$ and $\calX$ sufficiently fast in probability as $N$ and $T$ grow. \ref{as:densities} ensures that the pushforward points fill $\calX$. \ref{as:kernels_cbq} ensures the first and second stage GPs have appropriate regularity for the problem. Further, it guarantees that $k_\calX$ is strictly positive definite (by \Cref{res:sobolev_kernels_are_spd}), which, together with the fact that $x_1^t, \dots, x_N^t$ are almost surely mutually distinct (since by \ref{as:pars_and_samples} they are drawn from a continuous distribution), ensures the Gram matrix $k_\calX(x_{1:N}^t, x_{1:N}^t)$ is almost surely invertible. \ref{as:regulariser} requires the nugget $\lambda_\Theta$ to grow with $T$, which is natural in a bounded domain where the conditioning of the Gram matrix deteriorates as $T \to \infty$. We also implicitly assume that kernel hyperparameters are known.

Several of these assumptions admit straightforward generalisations. \ref{as:domains} extends to any bounded domain with Lipschitz boundary satisfying an interior cone condition \citep{kanagawa2020convergence, wynne2021convergence}. The point set assumptions in \ref{as:pars_and_samples} could be replaced by active learning designs or grids following existing BQ convergence theory \citep{Kanagawa2019, kanagawa2020convergence, wynne2021convergence}. The smoothness range in \ref{as:kernels_cbq} can be significantly widened using the approach of \citet{kanagawa2020convergence}, and the known-hyperparameters assumption can be relaxed to estimation in bounded sets \citep{Teckentrup2020}.

We are now ready to discuss the implications of the theorem.
Firstly, the result is expressed in probability to account for randomness in $\theta_{1:T}$ and $x_{1:N}^t$, and provides a rate of $\bigo(T^{-1/4}+ N^{- s_\calX/d + \varepsilon})$. We can see that growing $N$ will only help up to some extent (as the first term approaches zero fast), but that growing $T$ is essential to ensure convergence. This is intuitive since we cannot expect to approximate $I(\theta)$ uniformly simply by increasing $N$ at some fixed points in $\Theta$. 
Despite this, we will see in \Cref{sec:experiments_cbq} that increasing $N$ will be essential to improving performance in practice. The rate in $N$ will typically be very fast for smooth targets, but is significantly slowed down for large $d$, demonstrating that our method is mostly suitable for low- to mid-dimensional problems, a common feature shared by Bayesian quadrature-based algorithms~\citep{Briol2019PI, frazier2018bayesian}.
There have been some attempts to scale BQ/CBQ to high dimensions: for example,~\citet[Section 5.4]{Briol2019PI} decompose the integrand into a sum of low-dimensional functions; however, this is only possible in limited settings when the integrand has certain forms of sparsity.

Although the bound is dominated by a term $\bigo(T^{-1/4})$ in $T$, the proof can be extended to provide a more general result with rate up to $\bigo(T^{-1/3})$ under an additional `source condition' which requires stronger regularity from $f$; this is further discussed in~\Cref{appendix:convergence_rate}. Compared to baselines, we note that we cannot expect a similar result for IS since IS does not apply when $f$ depends on $\theta$. 
For LSMC, we also cannot guarantee consistency of the algorithm when $I(\theta)$ is not a polynomial (unless $q \rightarrow \infty$; see \citet{stentoft2004convergence}). 
Although we are not aware of any such result, we expect KLSMC to have the same rate in $T$ as CBQ, and for CBQ to be significantly faster than KLSMC in $N$. This is due to the second stage of KLSMC being essentially the same as that for CBQ, and KLSMC using MC rather than BQ in the first stage: by~\citet{novak1988deterministic}, the convergence rate of BQ, $N^{-s_\calX/d}$, is faster than that of MC, $N^{-1/2}$, in the case where the function $x \to f(x, \theta)$ is of smoothness at least $s_\calX > d/2$.

Lastly, while the rate in $N$ matches the minimax rate for nonparametric regression \citep{Stone1982}, the rate in $T$ does not. This highlights that optimal convergence rates for heteroscedastic GP regression remain unresolved; we leave this substantial question to future work.

\begin{figure*}[t]
    \centering
    \begin{minipage}{\textwidth}
    \centering
    \includegraphics[width=240pt]{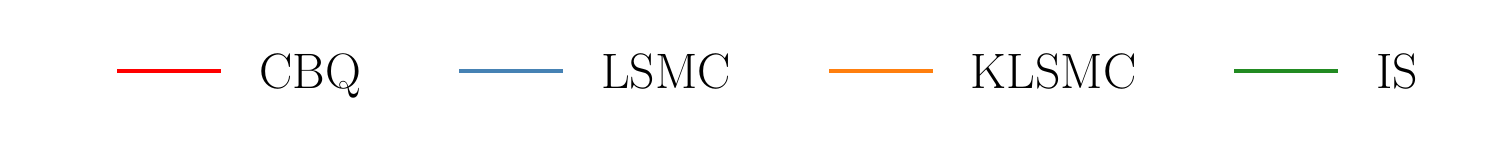}
    \end{minipage}
    \centering
    \begin{subfigure}{0.33\textwidth}
        \centering
        \hspace{-10pt}
        \includegraphics[width=1.0\linewidth]{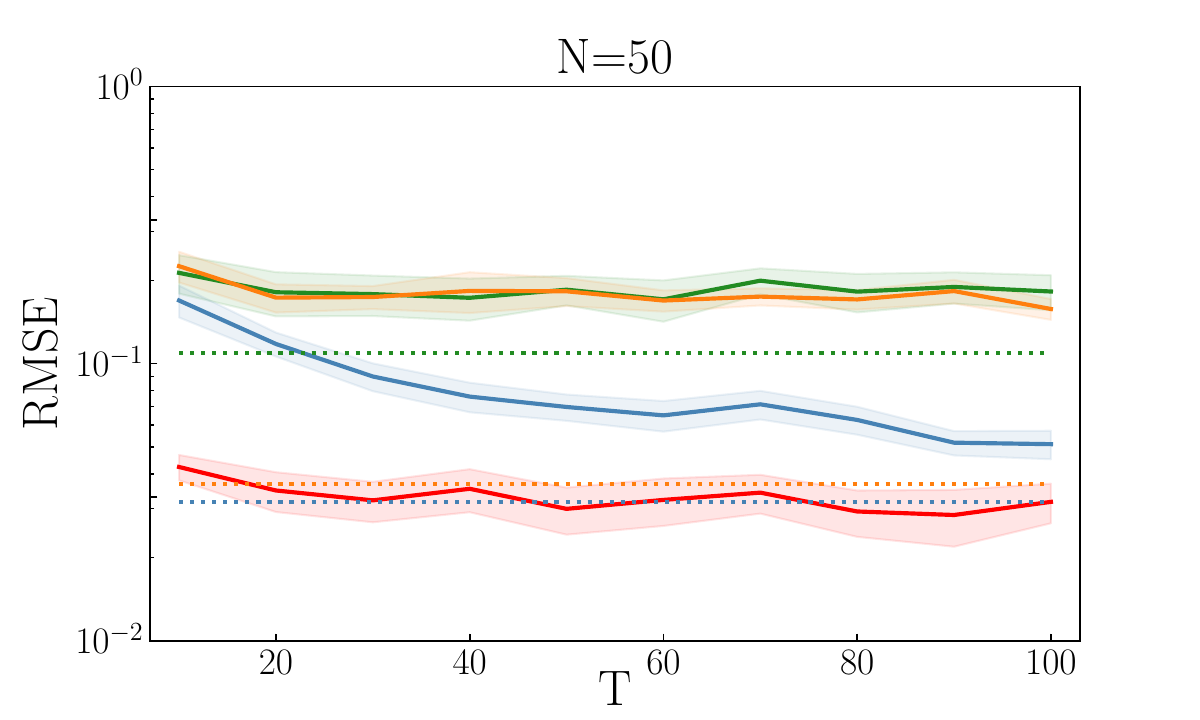}
        \label{fig:bayes_sensitivity_1}
    \end{subfigure}%
    \hfill 
    \begin{subfigure}{0.33\textwidth}
        \centering
        \hspace{-10pt}
        \includegraphics[width=1.0\linewidth]{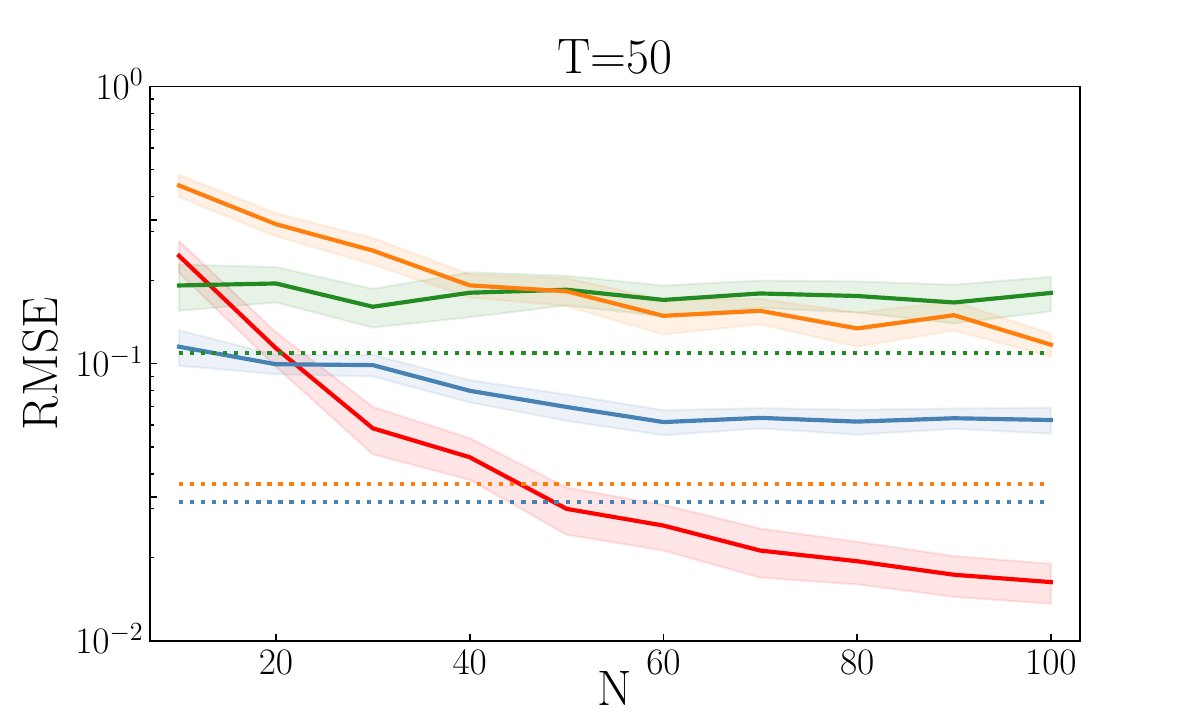}
        \label{fig:bayes_sensitivity_2}
    \end{subfigure}%
    \hfill
    \begin{subfigure}{0.33\textwidth}
        \centering
        \hspace{-10pt}
        \includegraphics[width=1.0\linewidth]{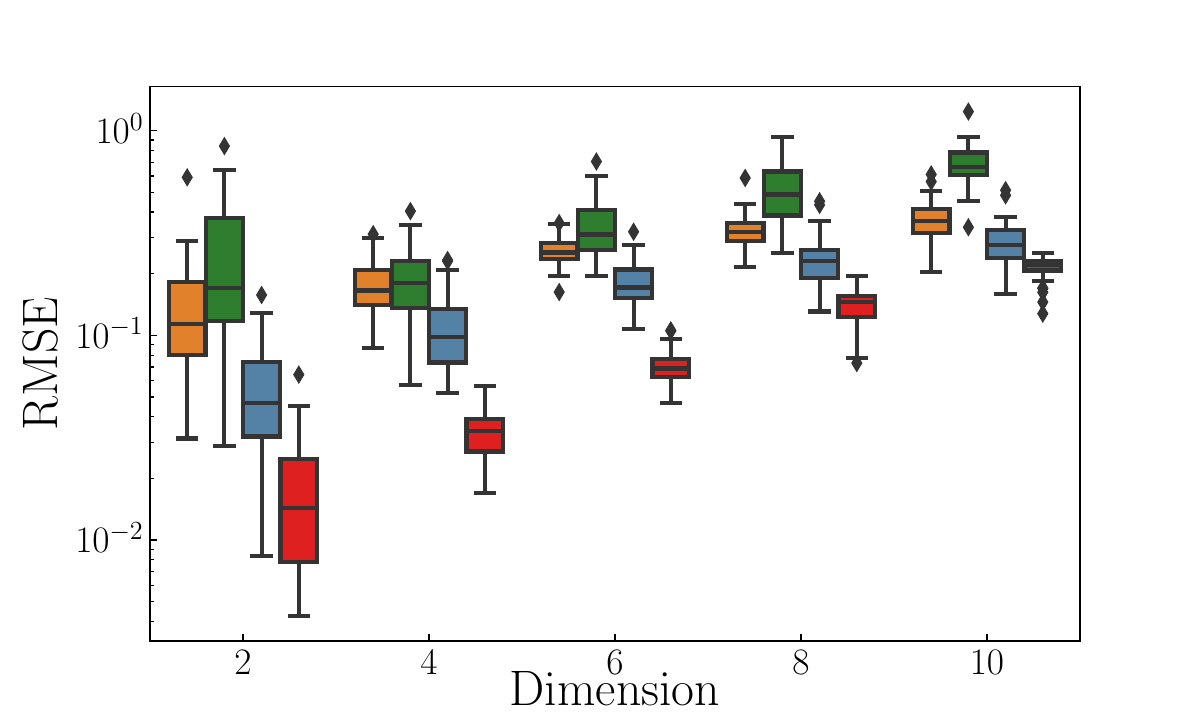}
        \label{fig:bayes_sensitivity_3}
    \end{subfigure}
    \caption[Bayesian sensitivity analysis for linear models.]{\emph{Bayesian sensitivity analysis for linear models.} \textbf{Left:} RMSE of all methods when $d=2$ and $N=50$. \textbf{Middle:} RMSE of all methods when $d=2$ and $T=50$. \textbf{Right:} RMSE of all methods when $N=T=100$.}
    \label{fig:bayes_sensitivity}
\end{figure*}

\section{Experiments}\label{sec:experiments_cbq}

We will now evaluate the empirical performance of CBQ against baselines including IS, LSMC and KLSMC. 
For the first three experiments, we focus on the case where $f$ does not depend on $\theta$ (i.e,. $f(x, \theta) = f(x)$), and for the fourth experiment, we focus on the case where $f$ depends on both $x$ and $\theta$. 
All methods use $\theta_1, \dots, \theta_T \sim \bQ$ ($\bQ$ is specified individually for each experiment) and $x_{1:N}^t \sim \bP_{\theta_t}$ to ensure a fair comparison, and we therefore use $\bP_{\theta_1}, \ldots, \bP_{\theta_T}$ as our importance distributions in IS. 
For experiments on nested expectations, we use standard Monte Carlo for the outer expectation and use CBQ along with all baseline methods to compute the conditional expectation for the inner expectation.

Detailed descriptions of hyperparameter selection for CBQ and all baseline methods can be found in \Cref{appendix:hyperparameter_selection}. 
Detailed experimental settings can be found in \Cref{appendix:bayes_sensitivity} to \Cref{appendix:decision} along with detailed checklists on whether the assumptions of \Cref{thm:convergence} can be satisfied in each experiment. 

\paragraph{Synthetic Experiment: Bayesian Sensitivity Analysis for Linear Models.}
The prior and likelihood in a Bayesian analysis often depend on hyperparameters, and determining the sensitivity of the posterior to these is critical for assessing robustness~\citep{oakley2004probabilistic,Kallioinen2021}. One way to do this is to study how posterior expectations of interest depend on these hyperparameters, a task usually requiring the computation of conditional expectations. We consider this problem in the context of Bayesian linear regression with a zero-mean Gaussian prior with covariance $\theta \Id_d$ and $\theta \in (1, 3)^d$. Using a Gaussian likelihood, we can obtain a conjugate Gaussian posterior $\bP_{\theta}$ on the regression weights. We can then analyse sensitivity by computing the conditional expectation $I(\theta)$ of some quantity of interest $f$. For example, if  $f(x)=x^\top x$, then $I(\theta)$ is the second moment of the posterior, whereas if $f(x) = x^\top y^\ast$ for some new observation $y^\ast$, then $I(\theta)$ is the predictive mean. In these simple settings, $I(\theta)$ can be computed analytically, making this a good synthetic example for benchmarking.

\begin{figure}[h]
    \centering
    \includegraphics[width=0.66\textwidth]{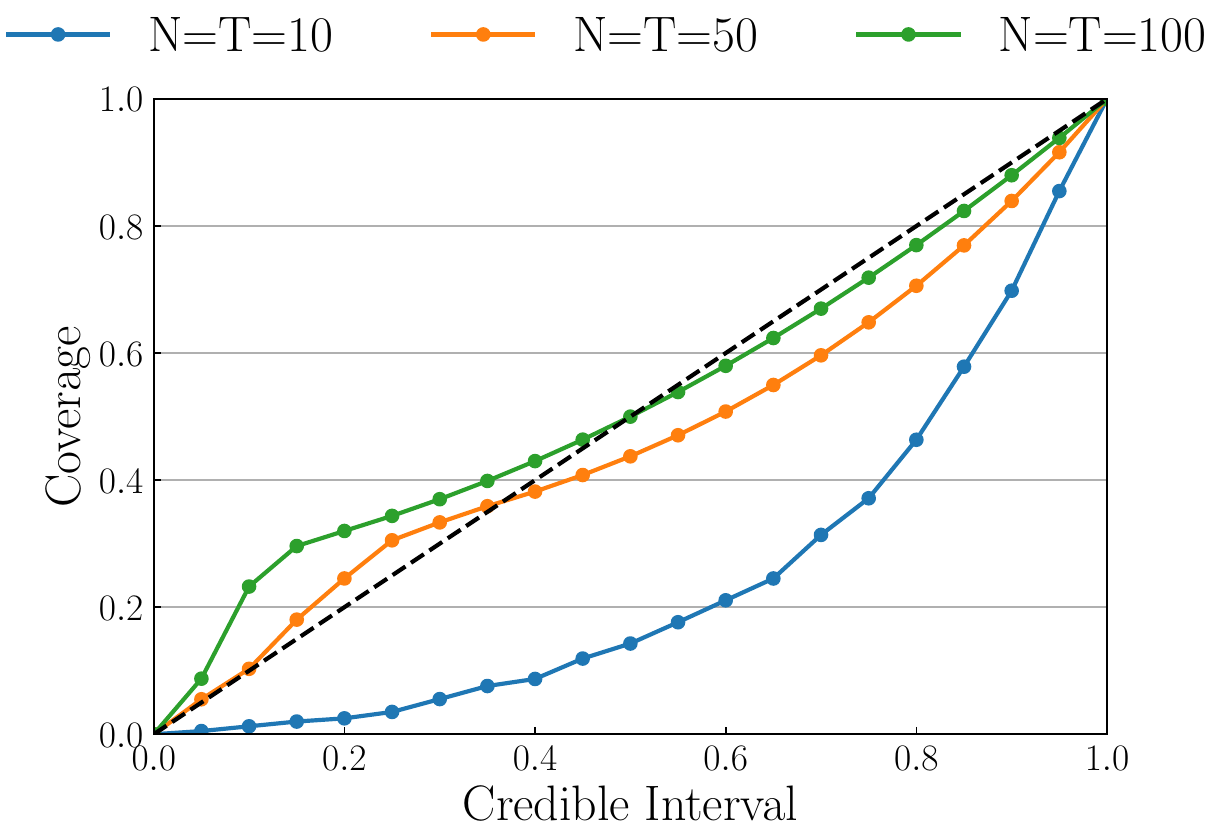}
    \caption[Bayesian linear model sensitivity analysis in $d=2$.]{\emph{Bayesian linear model sensitivity analysis in $d=2$}.}
    \label{fig:calib_bayes}
\end{figure}

Our results in \Cref{fig:bayes_sensitivity} are for the second moment, whilst the results for the predictive mean are in \Cref{appendix:bayes_sensitivity}.  
We measure performance in terms of root mean squared error (RMSE) and use $ \bQ = \operatorname{Unif}(1, 3)^d$.  For CBQ, $k_\calX$ is chosen to be a Gaussian kernel so that the kernel mean embedding $\mu$ has a closed form, and $k_\Theta$ is a Mat\'ern-3/2 kernel.
\Cref{fig:bayes_sensitivity} shows the performance of CBQ against baselines with varying $N$, $T$ and $d$. LSMC performs well for this problem, and this can be explained by the fact that $I(\theta)$ is a polynomial in $\theta$.
Despite this, the left and middle plots show that CBQ consistently outperforms all competitors. Specifically, its rate of convergence is initially much faster in $N$ than in $T$, which confirms the intuition from \Cref{thm:convergence}. The dotted lines also give the performance of baselines under a very large number of samples $N=T=1000$, and we see that CBQ is either comparable or better than these even when it has access only to much smaller $N$ and $T$. In the rightmost panel, we see that the baselines gradually catch up with CBQ as $d$ grows, which is again expected since the rate in \Cref{thm:convergence} is $O(N^{- s_{\calX}/d+\varepsilon})$ in $N$.

Our last plot is in \Cref{fig:calib_bayes} and studies the calibration of the CBQ posterior. The coverage is the $\%$ of times a credible interval contains $I(\theta)$ under repetitions of the experiment. The black diagonal line represents perfect calibration, whilst any curve lying above or below the black line indicates underconfidence or overconfidence, respectively. 
We observe that when $N$ and $T$ are as small as $10$, CBQ is overconfident. When $N$ and $T$ increase, CBQ becomes underconfident, i.e., the posterior variance is more inflated than needed from a frequentist viewpoint.
Calibration plots for the rest of the experiments can be found in \Cref{appendix:experiments_cbq} and demonstrate similar results. 
It is generally preferable to be underconfident than overconfident, and CBQ does a good job most of the time. 
We expect that overconfidence in small $N$ and $T$ can be explained by the poor performance of empirical Bayes, and therefore caution users not to rely too heavily on the reported uncertainty in this regime.


\begin{figure*}[t]
    \begin{minipage}{0.99\textwidth}
    \centering
    \includegraphics[width=\textwidth]{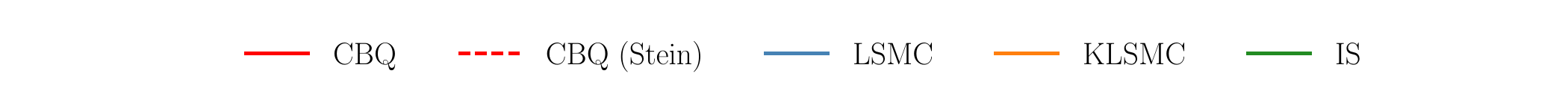}
    \end{minipage}
    
    \centering
    
    \begin{subfigure}{0.33\textwidth}
        \centering
        \includegraphics[width=1.0\linewidth]{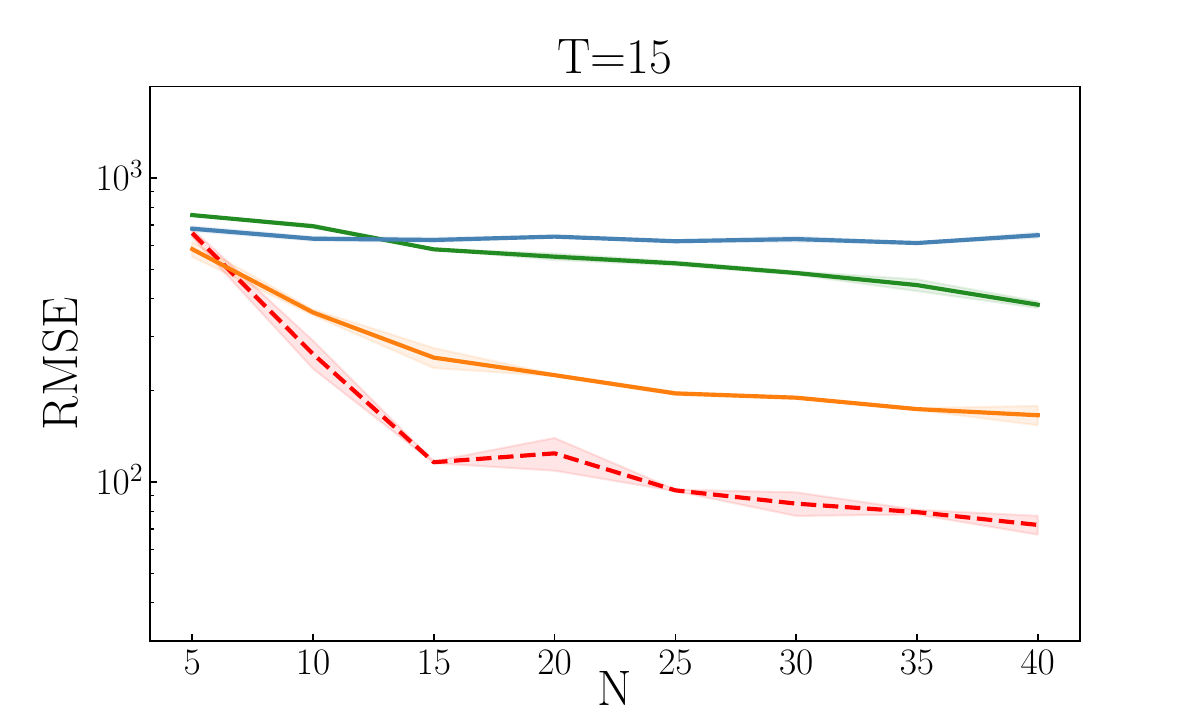}
        \label{fig:sir}
    \end{subfigure}%
    \hfill 
    \begin{subfigure}{0.33\textwidth}
        \centering
        \includegraphics[width=1.0\linewidth]{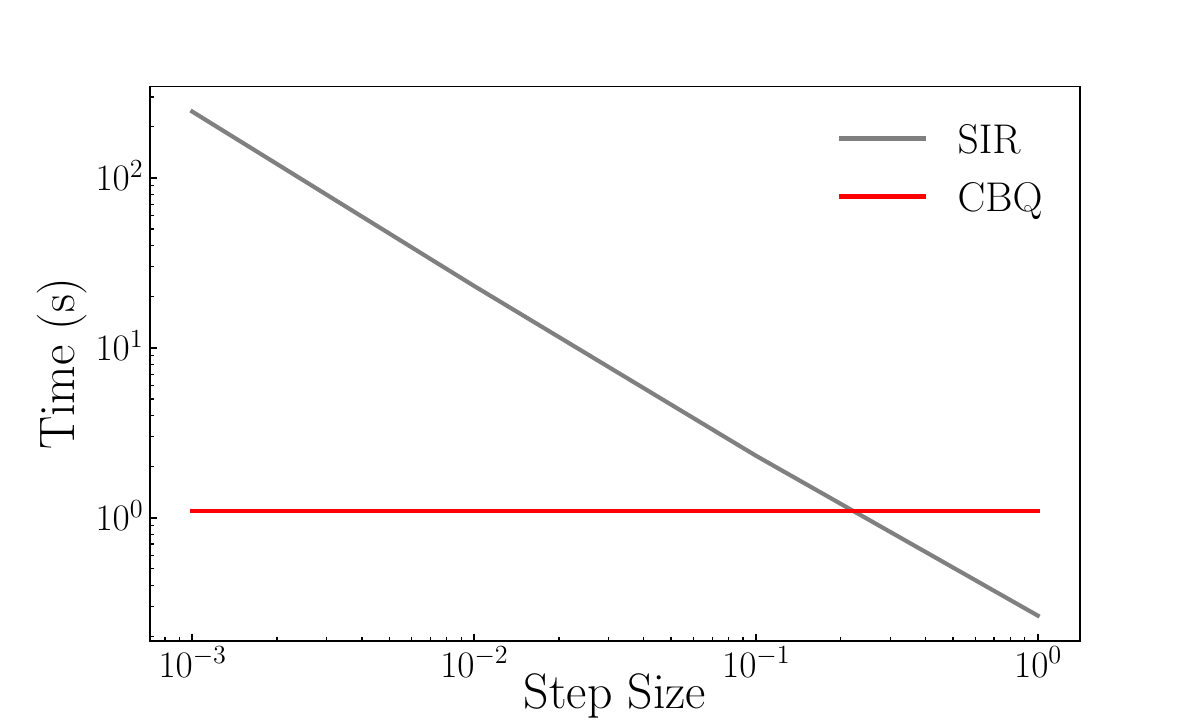}
        \label{fig:sir_time}
    \end{subfigure}
        \hfill
    \begin{subfigure}{0.33\textwidth}
        \centering
        \includegraphics[width=1.0\linewidth]{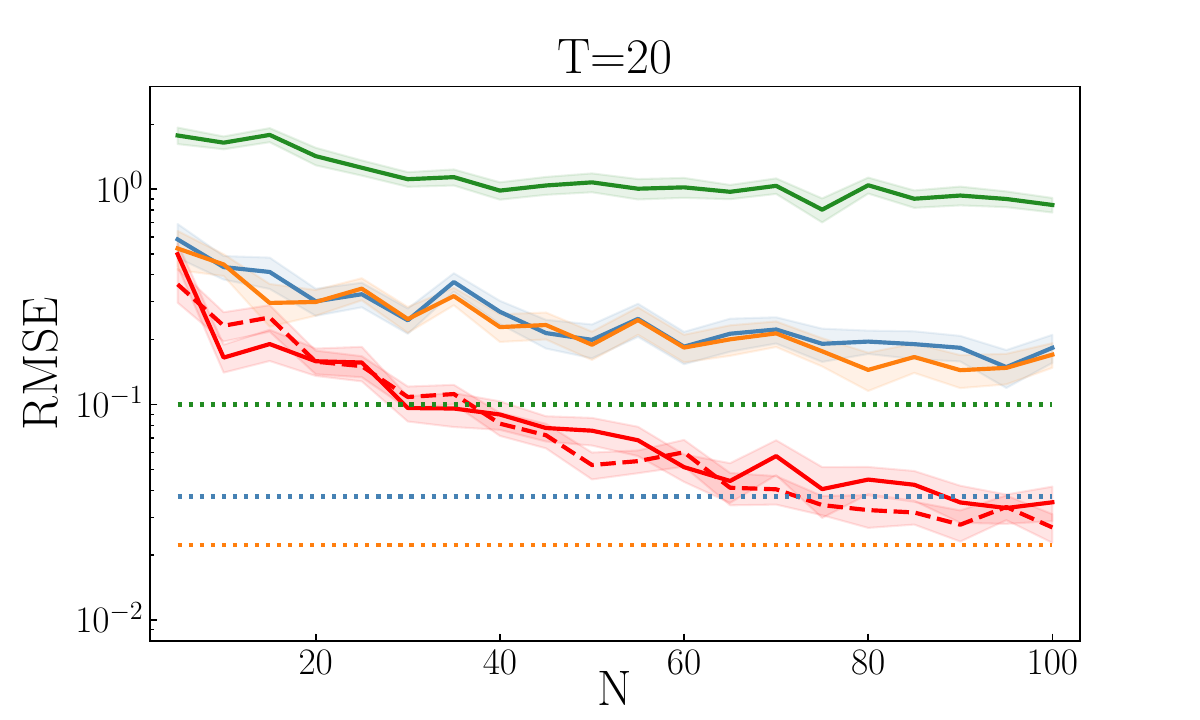}
        \label{fig:finance}
    \end{subfigure}
    \caption[Bayesian sensitivity analysis for SIR Model $\&$ Option pricing in mathematical finance.]{\emph{Bayesian sensitivity analysis for SIR Model $\&$ Option pricing in mathematical finance.} \textbf{Left:} RMSE of all methods for the SIR example with $T=15$. \textbf{Middle:} The computational cost (in wall clock time) for CBQ ($T=15, N=40$) and for obtaining one single numerical solution from SIR under different discretisation step sizes. In practice, the process of obtaining samples from SIR equations is repeated $NT$ times.
    \textbf{Right:} RMSE of all methods for the finance example with $T=20$.
    }
    \label{fig:finance_sir}
\end{figure*}

\paragraph{Bayesian Sensitivity Analysis for the Susceptible-Infectious-Recovered (SIR) Model.}
The SIR model is commonly used to simulate the dynamics of infectious diseases through a population~\citep{kermack1927sir}. In this model, the dynamics are governed by a system of differential equations parametrised by a positive infection rate and a recovery rate (see \Cref{appendix:sir}). 
The accuracy of the numerical solution to this system typically hinges on the step size.
While smaller step sizes yield more accurate solutions, they are also associated with a much higher computational cost. 
For example, using a step size of $0.1$ days for simulating a $150$-day period would require a computation time of $3$ seconds for generating a single sample, which is more costly than running CBQ on $N=40, T=15$ samples.
The cost would become even larger as the step size gets smaller, as depicted in the middle panel of \Cref{fig:finance_sir}. Consequently, when performing Bayesian sensitivity for SIR, there is a clear necessity for more data-efficient algorithms such as CBQ.

We perform a sensitivity analysis for the parameter $\theta$ of our $\operatorname{Gamma}(\theta, 10)$ prior on the infection rate $x$. The parameter $\theta$ represents the initial belief of the infection rate deduced from the study of the virus in the laboratory at the beginning of the outbreak.
We are interested in the expected peak number of infected individuals: $f(x)= \max_r N^r_I(x)$, where $N^r_I(x)$ is the solution to the SIR equations and represents the number of infections at day $r$. 
It is important to study the sensitivity of $I(\theta)$ to the shape parameter $\theta$. 
The total population is set to be $10^6$ and 
$\bQ = \operatorname{Unif}\left(2,9\right)$ and $\bP_{\theta_t} = \operatorname{Gamma}(\theta_t, 10)$. 
We use a Monte Carlo estimator with $5000$ samples as the pseudo ground truth and evaluate the RMSE across all methods. 
For CBQ, we employ a Stein kernel for $k_\calX$, with the Mat\'ern-3/2 as the base kernel, and $k_\Theta$ is selected to be a Mat\'ern-3/2 kernel.

We can see in the left panel of \Cref{fig:finance_sir} that CBQ clearly outperforms baselines including IS, LSMC and KLSMC in terms of RMSE.
Although the CBQ estimator exhibits a higher computational cost compared to baselines, we have demonstrated in the middle panel of \Cref{fig:finance_sir} that, due to the increased computational expense of obtaining samples with a smaller step size, using CBQ is ultimately more efficient overall within the same period of time.


\paragraph{Option Pricing in Mathematical Finance.}
Financial institutions are often interested in computing the expected loss of their portfolios if a shock were to occur in the economy, which itself requires the computation of conditional expectations; it is, in fact, in this context that LSMC and KLSMC were first proposed. This is typically a challenging computational problem since simulating from the stock of interest often requires the numerical solution of stochastic differential equations over a long time horizon (see \citet{achdou2005computational}), making data-efficient methods such as CBQ particularly desirable.

Our next experiment is representative of this class of problems, but has been chosen to have a closed-form expected loss and to be amenable to cheap simulation of the stock to enable extensive benchmarking. We consider a butterfly call option whose price $S(\tau)$ at time $\tau \in [0,\infty)$ follows the Black-Scholes formula; see \Cref{appendix:black_scholes} for full details. The payoff at time $\tau$ can be expressed as
$\psi(S({\tau}))=\max (S(\tau)-K_1, 0) + \max (S(\tau)-K_2, 0) - 2\max (S(\tau) - (K_1 + K_2) / 2, 0)$ for two fixed constants $K_1, K_2 \geq 0$.
We follow the set-up in \citet{alfonsi2021multilevel, alfonsi2022many} assuming that a shock occurs at time $\eta$ when the price is $S(\eta)=\theta \in (0,\infty)$, and this shock multiplies the price by $1 + s$ for some $s \geq 0$. As a result, the expected loss of the option is $L = \bE_{\theta \sim \bQ}
[ \max ( I(\theta), 0)]$, where $I(\theta) =  \int_{0}^\infty f(x) \bP_\theta(\d x)$, $x=S(\zeta)$ is the price at the time $\zeta$ at which the option matures, $f(x) = \psi(x)-\psi((1+s)x)$, and $\bP_\theta$ and $\bQ$ are two log-normal distributions induced from the Black-Scholes model. 

Results are presented in the rightmost panel of \Cref{fig:finance_sir}. We take $K_1 = 50, K_2 = 150, \eta=1, s = 0.2$ and $\zeta=2$. 
For CBQ, $k_\Theta$ is selected to be a Mat\'ern-3/2 kernel and $k_\calX$ is either a Stein kernel with Mat\'ern-3/2 as base kernel or a logarithmic Gaussian kernel (see \Cref{appendix:black_scholes}), in which case $k_{\calX}$ is too smooth to satisfy the assumption of our theorem. 

As expected, CBQ exhibits much faster convergence in $N$ than IS, LSMC or KLSMC, and outperforms these baselines even when they are given a substantial sample size of $N=T=1000$ (see dotted lines). We can also see that CBQ with the log-Gaussian kernel or with Stein kernel have similar performance, despite the log-Gaussian kernel not satisfying the smoothness assumptions of our theory. 

\begin{figure}[h]
  \centering
  \hspace{-20pt}
  \includegraphics[width=0.8\linewidth]{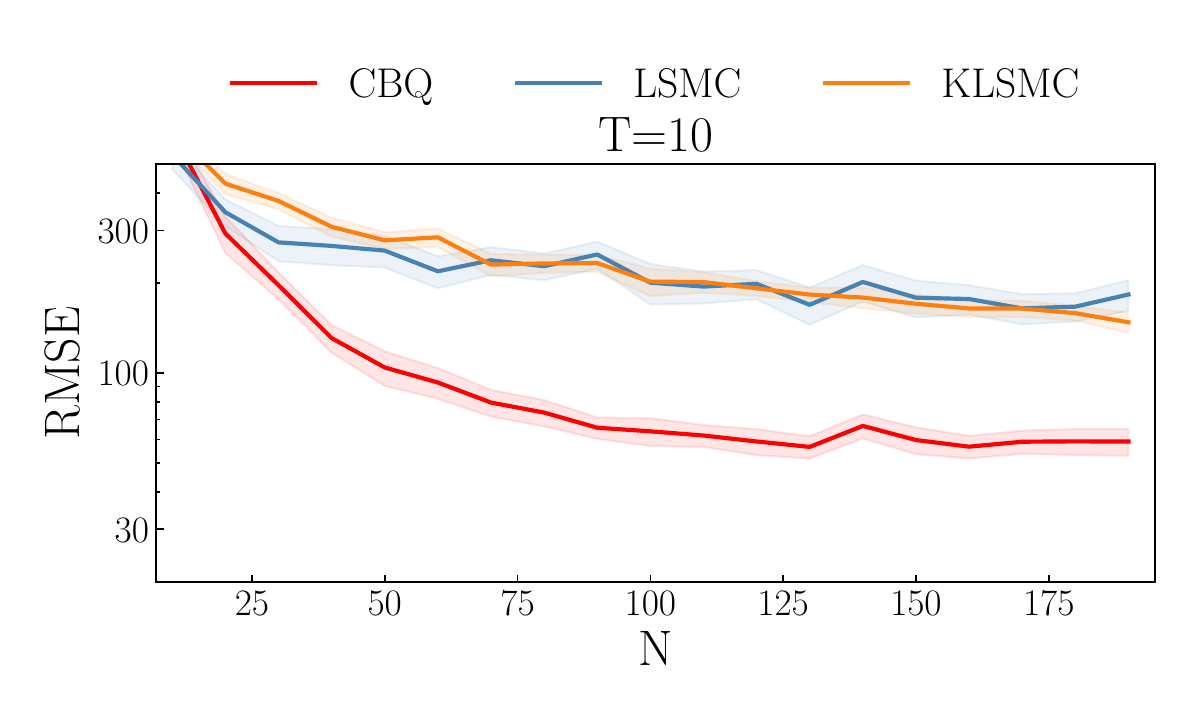}
  \caption[Uncertainty decision making in health economics.]{\emph{Uncertainty decision making in health economics.} We study RMSE for different estimators of EVPPI.}
  \label{fig:decision}
\end{figure}

\paragraph{Uncertainty Decision Making in Health Economics.} 
In the medical world, it is important to trade off the costs and benefits of conducting additional experiments on patients.
One important measure in this context is the expected value of partial perfect information (EVPPI), which quantifies the expected gain from conducting experiments to obtain precise knowledge of some unknown variables \citep{brennan2007calculating}. 
The EVPPI can be expressed as $\bE_{\theta \sim \bQ}[\max_c I_c(\theta) ] - \max_c \bE_{\theta \sim \bQ}[I_c(\theta)]$ where $f_c$ represents a measure of patient outcome (such as quality-adjusted life-years) under treatment $c$ among a set of potential treatments $\calC$, $\theta$ denotes the additional variables we could measure, and $I_c(\theta) = \int_{\calX} f_c(x, \theta) \bP_\theta(\d x)$ denotes the expected patient outcome given our measurement of $\theta$. We highlight that for these applications, $N$ and $T$ are often small due to the very high monetary cost and complexity of collecting patient data in the real world.

We study the potential use of CBQ for this problem using the synthetic problem of \citet{Giles2019}, where $\bP_{\theta}$ and $\bQ$ are Gaussians (see \Cref{appendix:decision}). We compute EVPPI with $f_1(x, \theta)=10^4 (\theta_1 x_5 x_6 + x_7 x_8 x_{9})-(x_1 + x_2 x_3 x_4)$ and $f_2(x, \theta) = 10^4 (\theta_2 x_{13} x_{14} + x_{15} x_{16} x_{17})-(x_{10} + x_{11} x_{12} x_4)$. 
The exact practical meanings of $x$ and $\theta$ can be found in \Cref{appendix:decision}.
We draw $10^6$ samples from the joint distribution to generate a pseudo ground truth, and evaluate the RMSE across different methods.
Note that IS is no longer applicable here because $f$ depends on both $x$ and $\theta$, so we only compare against KLSMC and LSMC. 
For CBQ, $k_\calX$ is a Mat\'ern-3/2 kernel and $k_\Theta$ is also a Mat\'ern-3/2 kernel. 
In \Cref{fig:decision}, we can see that CBQ consistently outperforms baselines with much smaller RMSE. The results are also consistent with different values of $T$; see \Cref{appendix:decision}.

\section{Conclusions}
We propose CBQ, a novel algorithm which is tailored for the computation of conditional expectations in the setting where obtaining samples or evaluating functions is costly. 
We show both theoretically and empirically that CBQ exhibits a fast convergence rate and provides the additional benefit of Bayesian uncertainty quantification. 
Looking forward, we believe further gains in accuracy could be obtained by developing active learning schemes to select $N$, $T$, and the location of $\theta_{1:T}$ and $x_{1:N}^t$ for all $t$ in an adaptive manner. 
Additionally, CBQ could be extended for nested expectation problems by using a second level of BQ based on the output of the heteroscedastic GP in stage 2, potentially leading to a further increase in accuracy. 

\chapter{Calibration for MMD-minimising Integration}
\label{sec:gpcv}
\begin{tcolorbox}
This chapter is an extended version of the following paper:

\begin{itemize}
    \item Naslidnyk, M., Kanagawa, M., Karvonen, T., \& Mahsereci, M. (2025). Comparing scale parameter estimators for Gaussian process interpolation with the Brownian motion prior: Leave-one-out cross validation and maximum likelihood. SIAM/ASA J. Uncertain. Quantif., 13(2), 679–717.
\end{itemize}

Specifically, while said paper considers uncertainty quantification for Gaussian process interpolation, this chapter extends this to Bayesian quadrature. Results in~\Cref{sec:limit-behaviour-for-sigma} and \Cref{res:uq-theorem-exp} apply to both settings and appear in the paper.

The theoretical results were obtained by me, with the exception of \Cref{res:uq-theorem-exp}, which was primarily by Dr Toni Karvonen. Experiments were completed in collaboration with Dr Maren Mahsereci.
\end{tcolorbox}

In~\Cref{sec:kernel_and_bayesian_quadrature} and the previous two chapters, we discussed that Bayesian quadrature (BQ), the probabilistic interpretation of MMD-minimising quadrature, performs numerical integration via Gaussian process (GP) interpolation on the integrand. The GP posterior mean defines the quadrature rule; the induced posterior variance estimates its uncertainty.
The uncertainty estimate is then used as a proxy for true quadrature error: in stopping rules, adaptive learning schemes, and reporting error bars. These procedures are only valid when the uncertainty is \emph{well-calibrated}, meaning it is commensurate with the true squared predictive error. Overconfident posteriors cause premature stopping and under-coverage; over-cautious ones waste samples and mask improvements. Therefore, uncertainty calibration is key.

Somewhat counter-intuitively, pointwise GP calibration alone is not enough to ensure uncertainty calibration in BQ. Even when the posterior standard deviation $\sqrt{k_N(x,x)}$ accurately estimates the prediction error $|f(x)-m_N(x)|$ for every input $x$, two effects may contribute to a mismatch between the BQ posterior standard deviation $\sigma_\BQ$ and BQ prediction error $|I - I_{\BQ}|$.
First, the cross-covariances $k_N(x, x')$ for $x \neq x'$, ignored in pointwise calibration but present in $\sigma_\BQ$, can accumulate to a substantial positive or negative contribution.
Second, $|I - I_{\BQ}|$ integrating the signed residuals $f(x)-m_N(x)$ may lead to error cancellation absent from $|f(x)-m_N(x)|$. Together, these effects mean it is important to consider BQ calibration specifically.

As we will show, for BQ uncertainty estimates to be well-calibrated, the kernel of the GP prior must be carefully selected. We will theoretically compare two methods for choosing the kernel: cross-validation and maximum likelihood estimation. Focusing on the amplitude parameter estimation of a Brownian motion kernel in the noiseless setting, we prove that, for both BQ and the underlying GP, cross-validation can yield asymptotically well-calibrated credible intervals for a broader class of ground-truth functions than maximum likelihood estimation, suggesting an advantage of the former over the latter.  Finally, motivated by the findings, we propose \emph{interior cross validation}, a procedure that adapts to an even broader class of ground-truth functions.

\section{Uncertainty Quantification via Kernel Scaling} 
\label{sec:gpcv-introduction}

Uncertainty quantification is a key property of BQ, crucial for applications involving decision-making, safety-critical systems, and scientific discovery. Recall from \Cref{sec:bq_background} that in BQ, the unknown integral
\begin{equation*}
    I = \int_\calX f(x) \bP(\d x)
\end{equation*}
is estimated using GP interpolation on the integrand $f$. Specifically, a {\em prior distribution} for $f$ is defined as a GP, by specifying its {\em kernel} and mean function.
Given $N$ observations of $f$, the {\em posterior distribution} of $f$ is another GP with mean function $m_N$ and kernel (or covariance function) $k_N$. The posterior GP induces a Gaussian posterior distribution $\calN(I_\BQ, \sigma^2_\BQ)$ over the integral, where
\begin{equation}
\label{eq:bq_mean_and_var_are_integrals}
    I_\BQ = \int_\calX m_N(x) \bP(\d x), \qquad \sigma^2_\BQ = \int_\calX \int_\calX k_N(x, x') \bP(\d x) \bP(\d x').
\end{equation}
The value of the integral $I$ can then be predicted as the posterior mean $I_\BQ$, and its uncertainty is quantified using the posterior standard deviation $\sigma_\BQ$.
Specifically, a {\em credible interval} of $I$ can be constructed as the interval $[I_\BQ - \alpha \sigma_\BQ,  I_\BQ + \alpha \sigma_\BQ ]$ for a constant $\alpha > 0$ (for example, $\alpha \approx 1.96$ leads to the 95\% credible interval).

For BQ uncertainty estimates to be reliable, the posterior standard deviation $\sigma_\BQ$ should, ideally, decay at the {\em same} rate as the prediction error $|I_\BQ - I|$ decreases, with the increase of sample size $N$. Otherwise, the uncertainty estimates are either asymptotically {\em overconfident} or {\em underconfident}.  For example, if  $\sigma_\BQ$ goes to $0$ faster than the error $|I_\BQ - I|$, then the credible interval $[I_\BQ - \alpha \sigma_\BQ, I_\BQ + \alpha \sigma_\BQ  ] $ will {\em not} contain the true value $I$ as $N$ increases for {\em any} fixed constant $\alpha > 0$ (asymptotically overconfident). If  $\sigma_\BQ$ goes to $0$ slower than the error $|I_\BQ - I|$, then the credible interval $[I_\BQ - \alpha \sigma_\BQ, I_\BQ + \alpha \sigma_\BQ  ] $ will get larger than the error $|I_\BQ - I|$ as $N$ increases (asymptotically underconfident). Neither case is desirable in practice, as BQ credible intervals will not be accurate estimates of prediction errors.

\begin{figure}[H]
    \centering
    \includegraphics[width=\linewidth]{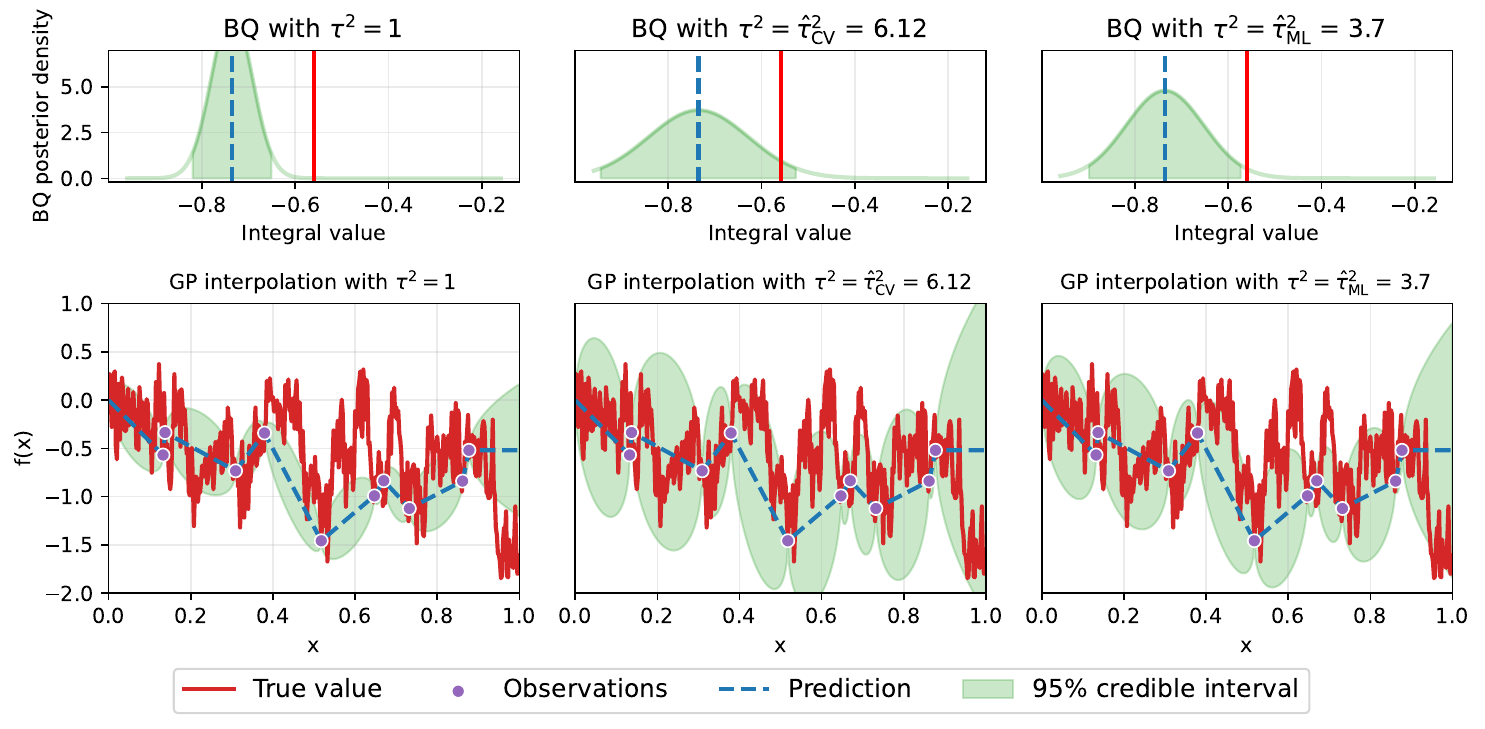}
    \caption[BQ of a fractional Brownian motion integrand.]{\emph{Top row:} BQ of a fractional Brownian motion (fBm) integrand $f(x)$ with Hurst parameter $H=0.2$ (smoothness $s+\alpha=0.2$), using the Brownian motion kernel~\eqref{eq:bm-kernel-intro} with amplitudes $\tau^2=1$ (left), $\tau^2=\tausqcvest=6.12$ given by the LOO-CV estimator (middle), and $\tau^2=\tausqmlest=3.7$ given by the ML estimator (right). Vertical lines mark the \textcolor{red}{true value of the integral $\int_{0}^{T} f(x) \d x$} (solid line) and the \textcolor{blue}{BQ posterior mean $I_{\mathrm{BQ}}=\int_{0}^{T} m_N(x) \d x$} (dashed line); the shaded bands show the \textcolor{ForestGreen}{95\% BQ credible interval $[ I_\mathrm{BQ} - 1.96\tau \sigma_\BQ,  I_\mathrm{BQ} + 1.96\tau \sigma_\BQ ]$}. \emph{Bottom row:} The GP interpolation underlying the top row: an fBm path that is the \textcolor{red}{true integrand $f(x)$} (solid line), the \textcolor{RoyalPurple}{training data $x_{1:N}, f(x_{1:N})$}, and the \textcolor{blue}{GP posterior mean $m_N(x)$} (dashed line); the shaded bands show the \textcolor{ForestGreen}{95\% GP credible interval $[ m_N(x) - 1.96\tau \sqrt{k_N(x, x)},  m_N(x) + 1.96\tau \sqrt{k_N(x, x)} ]$}.}
    \label{fig:FBM-H02}
\end{figure}

\begin{figure}[H]
    \centering
    \includegraphics[width=\linewidth]{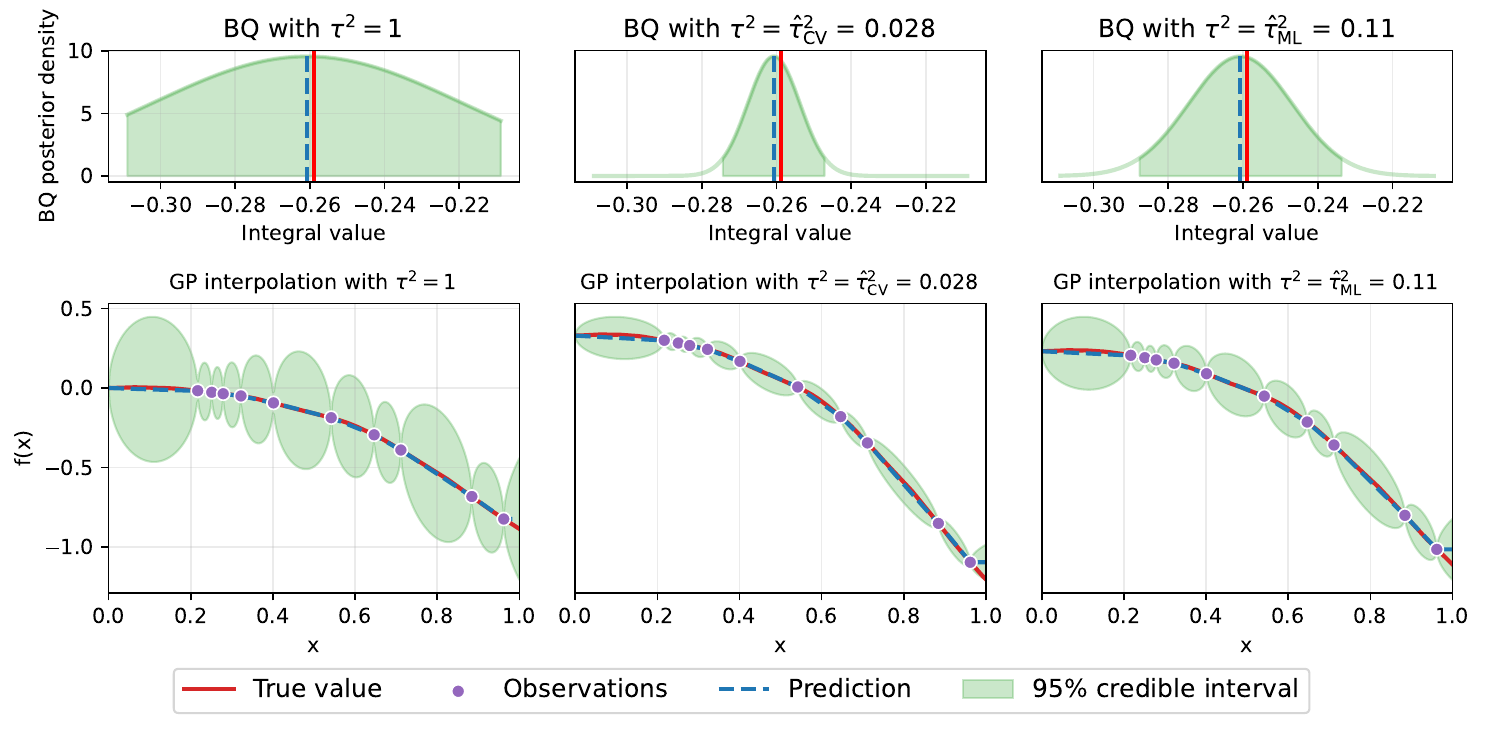}
    \caption[BQ of an integrated fractional Brownian motion integrand.]{BQ of an integrated fractional Brownian motion (fBm) integrand $f(x)$ with Hurst parameter $H=0.5$ (smoothness $s+\alpha=1.5$), using the Brownian motion kernel~\eqref{eq:bm-kernel-intro} with amplitudes $\tau^2=1$ (left), $\tau^2=\tausqcvest=0.028$ given by the LOO-CV estimator (middle), and $\tau^2=\tausqmlest=0.11$ given by the ML estimator (right). For the explanation of the figures, see the caption of \Cref{fig:FBM-H02}.}
    \label{fig:IFBM-H05}
\end{figure}

Unfortunately, in general, the posterior standard deviation $\sigma_\BQ$ does {\em not} decay at the same rate as the prediction error $| I - I_\BQ  |$, because $\sigma_\BQ$ does {\em not} depend on the true integrand $f$; see~\eqref{eq:bq-posterior-moments}. Exceptionally, if the function $f$ is a sample path of the GP prior (the well-specified case), uncertainty estimates can be well-calibrated. However, in general, $f$ is not exactly a sample path of the GP prior (the misspecified case), and the posterior standard deviation $\sigma_\BQ$ does not scale with the prediction error $| I - I_\BQ  |$. \Cref{fig:FBM-H02,fig:IFBM-H05}  (the left panels) show examples where the true function $f$ is not a sample of the GP prior and where the GP uncertainty estimates are not well-calibrated.

\tocless\subsection{Amplitude Parameter Estimation}

To obtain sensible uncertainty estimates, it is therefore
necessary to adapt the posterior standard deviation $\sigma_\BQ$ to the function $f$. One simple way to achieve this is to introduce the {\em amplitude parameter} $\tau^2 > 0$ and parameterise the kernel as
\begin{equation} \label{eq:scale-kernel}
k_\tau(x,x') \coloneqq \tau^2 k(x,x'),
\end{equation}
where $k$ is the original kernel. GP interpolation with this kernel $k_\tau$ yields the posterior mean function $m_N$, which is not influenced by $\tau^2$, and the posterior covariance function $\tau^2 k_N$, which is scaled by $\tau^2$. Consequently, BQ mean $I_\BQ$ is not influenced by $\tau^2$, and the BQ posterior is scaled as $\tau^2 \sigma^2_\BQ$. If $\tau^2$ is estimated from observed data of $f$, the estimate $\hat{\tau}^2$ depends on $f$, and so does the resulting posterior standard deviation $\hat{\tau} \sigma_\BQ$.

One approach to amplitude parameter estimation, in both GP interpolation and BQ, is the method of {\em maximum likelihood (ML)}, which optimises $\tau^2$ to maximise the marginal likelihood of the GP \citep[Section 5.4]{rassmussen2006gaussian}. The ML approach is popular for general hyperparameter optimisation in GP regression. Another less common way in the GP literature is {\em cross-validation (CV)}, which optimises $\tau^2$ to maximise the average predictive likelihood with held-out data~\citep{sundararajan2001predictive}. For either approach, the optimised amplitude parameter can be obtained analytically in computational complexity $\bigo(N^3)$.
\Cref{fig:FBM-H02,fig:IFBM-H05} (middle and right panels) demonstrate that both approaches yield uncertainty estimates better calibrated than the original estimates without the amplitude parameter.

Do these amplitude parameter estimators lead to asymptotically well-calibrated BQ uncertainty estimates? To answer this question, it is necessary to understand their convergence properties as the sample size $N$ increases.
Most existing theoretical works focus on the well-specified case where there is a `true' amplitude parameter $\tau_0^2$ such that the unknown $f$ is a GP with the kernel $\tau_0^2 k$. In this case, both the ML and CV estimators have been shown to be consistent in estimating the true $\tau_0^2$ \citep[e.g.,][]{Ying1991, Zhang2004,Bachoc2017, bachoc2020asymptotic}.
However, in general, no `true' amplitude parameter $\tau^2_0$ exists such that the integrand $f$ is a GP with the covariance $\tau_0^2 k$. In such misspecified cases, not much is known about the convergence properties of either estimator.
\citet{Karvonen2020} analyse the ML estimator for the amplitude parameter, assuming that $f$ is a deterministic function. They derive upper bounds (and lower bounds in some cases) for the ML estimator; see \citet{Wang2021} for closely related work. To our knowledge, no theoretical work exists for the CV estimator for the amplitude parameter in the misspecified case. \citet{Bachoc2013} and \citet{petit2021gaussian} empirically compare the ML and CV estimators under different model misspecification settings. We will review other related works in \Cref{sec:related-work}.

\tocless\subsection{Contributions}

This work studies the convergence properties of the ML and CV estimators $\tausqmlest$ and $\tausqcvest$ of the amplitude parameter $\tau^2$ to understand whether they lead to asymptotically well-calibrated uncertainty estimates. In particular, we provide the first theoretical analysis of the CV estimator $\tausqcvest$ when the GP prior is misspecified, and also establish novel results for the ML estimator $\tausqmlest$.

To facilitate the analysis, we focus on the following simplified setting. For a constant $T > 0$, let $[0, T] \subset \bR$ be the input domain. Let $k$ in~\eqref{eq:scale-kernel} be the Brownian motion kernel
\begin{equation}
\label{eq:bm-kernel-intro}
    k(x, x') = \min(x, x') \quad \text{ for } \quad x, x' \in [0, T].
\end{equation}
With this choice, a sample path of the GP prior, roughly speaking, has smoothness of 1/2; we will formalise this in later sections.

We assume that the integrand $f$ has the smoothness $s + \alpha$, where $s \in \bN$ and $0 < \alpha \leq 1$.
The GP prior has well-specified smoothness if $s = 0$ and $\alpha = 1/2$. Other settings of $s$ and $\alpha$ represent misspecified cases. If $s = 0$ and $\alpha <1/2$, the true integrand $f$ is rougher than the GP prior (\Cref{fig:FBM-H02}); if $s = 0$ and $\alpha > 1/2$ or $s \geq 1$, the integrand $f$ is smoother than the GP prior. We focus on the noise-free setting where the function values $f(x_1), \dots, f(x_N)$ are observed at input points $x_1, \dots, x_N \in [0,T]$.

Our main results are new upper and lower bounds for the asymptotic rates of the CV estimator $\tausqcvest$ and the ML estimator $\tausqmlest$ as $N \to \infty$ (\Cref{sec:limit-behaviour-for-sigma}). The results suggest that the CV estimator can yield asymptotically well-calibrated BQ uncertainty estimates for a broader class of integrands $f$ than the ML estimator; thus, the former has an advantage over the latter (\Cref{sec:discussion}).   More specifically, asymptotically well-calibrated uncertainty estimates may be obtained with the CV estimator for the range $0 < s + \alpha \leq 3/2$ of smoothness of the true function, while this range becomes $0 < s + \alpha \leq 1$ with the ML estimator and is narrower. This finding is consistent with the example in~\Cref{fig:IFBM-H05}, where the true function has smoothness $s + \alpha = 3/2$ and is thus smoother than the GP prior. The uncertainty estimates of the CV estimator appear to be well-calibrated, while those of the ML estimator are unnecessarily wide, failing to adapt to the smoothness. Motivated by these insights, we propose a method called \emph{interior cross-validation}, and show it accommodates an even wider range of smoothness of the true function than the CV estimator.

The rest of the chapter is structured as follows. After reviewing related works in
\Cref{sec:related-work}, we introduce the necessary background on the ML and CV approaches to amplitude parameter estimation in \Cref{sec:kernel-parameter-estimation}. We describe the setting of the theoretical analysis in \Cref{sec:setting},   present our main results in  \Cref{sec:limit-behaviour-for-sigma}, and discuss their consequences for uncertainty quantification in  \Cref{sec:discussion}. We report simulation experiments in \Cref{sec:experiments}, conclude in \Cref{sec:conclusion_gpcv}, and present proofs in \Cref{sec:proofs}.

\tocless\subsection{Related work\label{sec:related-work}}

We review related theoretical works on hyperparameter selection in GP interpolation. There is a lack of work specifically in the BQ setting, with the exception of~\citep{Karvonen2020} who showed that BQ with the scale parameter estimated by maximum likelihood can become `slowly' overconfident at worst in the Sobolev setting. Nevertheless, the GP results are relevant to our setting. We categorise them into two groups based on how the true unknown function $f$ is modelled: random and deterministic.

\paragraph{Random setting.}

One group of works models the ground truth $f$ as a random function, specifically as a GP. Most of these works model $f$ as a GP with a Mat\'ern-type kernel and analyse the ML estimator. Under the assumption that the GP prior is correctly specified,  asymptotic properties of the ML estimator for the amplitude parameter\footnote{In these works, $\tau^2$ is often referred to as the `scale' parameter; we adopt the term `amplitude' to avoid confusion with the lengthscale parameter.} and other parameters have been studied \citep{Stein1999, Ying1991, Ying1993, LohKam2000, Zhang2004, Loh2005, Du2009, Anderes2010, WangLoh2011, Kaufman2013,Bevilacqua2019}.
Recently \citet{LohSunWen2021} and \citet{LohSun2023} have constructed consistent estimators of various parameters for many commonly used kernels, including Mat\'erns. \citet{Chen2021} and \citet{Petit2023} consider a periodic version of Mat\'ern GPs,  and show the consistency of the ML estimator for its smoothness parameter. To our knowledge, the only existing theoretical result for ML estimation of the amplitude parameter in the misspecified random setting considers oversmoothing~\citep[Theorem 4.2]{karvonen2021estimation}. Oversmoothing refers to the situation where the chosen kernel is smoother than the true function. In~\Cref{sec:random-setting} (\Cref{res:holder-spaces-exp-ml}), we provide a result for the undersmoothing case, which occurs when the chosen kernel is less smooth than the true function.

In contrast, few theoretical works exist for the CV estimator.  \citet{Bachoc2017} study the leave-one-out (LOO) CV estimator for the Mat\'ern-1/2 model (or the Laplace kernel) with one-dimensional inputs, in which case the GP prior is an Ornstein--Uhlenbeck (OU) process. Assuming the well-specified case where the true function is also an OU process, they prove the consistency and asymptotic normality of the CV estimator for the microergodic parameter in the fixed-domain asymptotic setting. \citet{bachoc2018asymptotic} and \citet{bachoc2020asymptotic} discuss another CV estimator that uses the mean square prediction error as the scoring criterion of CV (thus different from the one discussed here) in the increasing-domain asymptotics.   \citet{Bachoc2013} and \citet{petit2021gaussian}  perform empirical comparisons of the ML and CV estimators under different model misspecification settings. Thus, to our knowledge, no theoretical result exists for the CV estimator of the amplitude parameter in the random misspecified setting, which we provide in \Cref{sec:random-setting} (\Cref{res:holder-spaces-exp}).

\paragraph{Deterministic setting.}
Another line of research assumes that the ground truth $f$ is a fixed function belonging to a specific function space \citep{Stein1993}. \citet{XuStein2017} assumed that the ground truth $f$ is a monomial on $[0,1]$ and proved some asymptotic results for the ML estimator when the kernel $k$ is Gaussian.
As mentioned earlier, \citet{Karvonen2020} proved asymptotic upper (and, in certain cases, also lower) bounds on the ML estimator $\tausqmlest$ of the amplitude parameter $\tau^2$; see \citet{Wang2021} for a closely related work. \citet{Karvonen2023} has studied the ML and LOO-CV estimators for the smoothness parameter in the Mat\'ern model; see also  \citet{Petit2023}. \citet{BenSalem2019} and \citet{KarvonenOates2023} proved non-asymptotic results on the lengthscale parameter in the Mat\'ern and related models. Thus, there has been no work for the CV estimator of the amplitude parameter $\tau^2$ in the deterministic setting, which we provide in \Cref{sec:results-deterministic} (\Cref{res:holder-spaces}); we also prove a corresponding result for the ML estimator (\Cref{res:holder-spaces-ml}).

\section{Kernel parameter estimation}
\label{sec:kernel-parameter-estimation}

The selection of the kernel $k$ is typically performed by defining a parametric family of kernels $\{k_\theta \}_{\theta \in \Theta}$ and selecting the parameter $\theta$ based on an appropriate criterion.  Here $\Theta$ is a parameter set, and $k_\theta: \calX \times \calX \to \bR$ for each $\theta \in \Theta$ is a kernel.

\paragraph{Maximum likelihood (ML) estimation.}
The ML estimator maximises the log-likelihood of the GP $f$ with kernel $k_\theta$ under the data $(x_{1:N}, f(x_{1:N}))$,
\begin{align*}
  \log p(f(x_{1:N}) \given x_{1:N}, \theta) &= -\frac{1}{2} \bigg( f(x_{1:N})^\top k_\theta(x_{1:N}, x_{1:N})^{-1} f(x_{1:N}) \\
  &\qquad\qquad+ \log \det k_\theta(x_{1:N}, x_{1:N}) + N \log (2\pi) \bigg),
\end{align*}
where $\det k_\theta(x_{1:N}, x_{1:N})$ is the determinant of the Gram matrix $k_\theta(x_{1:N}, x_{1:N})$ \citep[e.g.,][Section~5.4.1]{rassmussen2006gaussian}.
With the additive terms that do not depend on $\theta$ removed from $\log p(f(x_{1:N}) \given x_{1:N}, \theta)$, this is equivalent to minimising the loss function
\begin{equation} \label{eq:ML-loss-function}
    L_\mathrm{ML}(\theta) \coloneqq f(x_{1:N})^\top k_\theta (x_{1:N}, x_{1:N})^{-1} f(x_{1:N}) + \log \det k_\theta (x_{1:N}, x_{1:N}).
\end{equation}
In general, $L_\mathrm{ML}(\theta)$ may not have a unique minimiser, so that any ML estimator satisfies
\begin{equation*}
     \hat{\theta}_\mathrm{ML} \in \argmin_{\theta \in \Theta} L_\mathrm{ML}(\theta).
\end{equation*}

\paragraph{Leave-one-out cross-validation (LOO-CV).}
The LOO-CV estimator \citep[Section~5.4.2]{rassmussen2006gaussian}, which we may simply call the CV estimator, is an alternative to the ML estimator. It maximises the average log-predictive likelihood
\begin{equation} \label{eq:cv-estimator-any-kernel}
  \sum_{n=1}^N \log p( f(x_n) \given x_n, x_{\setminus n}, f(x_{\setminus n}), \theta)
\end{equation}
with held-out data $(x_n, f(x_n))$, where $n = 1, \dots, N$, based on the data $(x_{\setminus n}, f(x_{\setminus n}))$, where $x_{\setminus n}$ denotes the input points with $x_n$ removed,
\begin{equation*}
  x_{\setminus n} =
  \begin{bmatrix} x_1 & \dots & x_{n-1} & x_{n+1} & \dots& x_N \end{bmatrix}^\top \in \calX^{N - 1}.
\end{equation*}
Let $m_{\theta, \setminus  n}$ and $ k_{\theta, \setminus  n}$ denote the posterior mean and covariance functions of GP regression with the kernel $k_\theta$ and the data $(x_{\setminus n}, f(x_{\setminus n}))$.
Because each $p( f(x_n) \given x_n, x_{\setminus n}, f(x_{\setminus n}), \theta)$ is the Gaussian density of $f(x_n)$ with mean $m_{\theta, \setminus  n}(x_n)$ and variance $k_{\theta, \setminus  n}(x_n,x_n)$, removing additive terms that do not depend on $\theta$ and reversing the sign in~\eqref{eq:cv-estimator-any-kernel} gives the CV objective function,
\begin{equation} \label{eq:CV-loss-function}
    L_\mathrm{CV}(\theta) = \sum_{n=1}^N \frac{\left[f(x_n) - m_{\theta, \setminus n}(x_n)\right]^2}{ k_{\theta, \setminus  n}(x_n, x_n)}  + \log k_{\theta, \setminus  n} (x_n, x_n).
\end{equation}%
The CV estimator is then defined as its minimiser
\begin{equation*}
  \hat{\theta}_\mathrm{CV} \in \argmin_{\theta \in \Theta} L_\mathrm{CV}(\theta).
\end{equation*}
As for the ML estimator, the CV objective function and its first-order gradients can be computed in closed form in $\bigo(N^3)$~\citep{sundararajan2001predictive}.

\paragraph{Amplitude parameter estimation.}
As explained in \Cref{sec:gpcv-introduction}, we consider the family of kernels $k_\tau (x,x') \coloneqq \tau^2 k(x,x')$ parameterised with the amplitude parameter $\tau^2>0$, and study the estimation of $\tau^2$ using the CV and ML estimators, denoted as $\tausqcvest$ and  $\tausqmlest$, respectively. In this case, both $\tausqcvest$ and  $\tausqmlest$ can be derived in closed form by differentiating~\eqref{eq:CV-loss-function} and~\eqref{eq:ML-loss-function}.

Let $m_{n-1}$ and $k_{n-1}$ be the posterior mean and covariance functions of GP regression using the kernel $k$ and the first $n-1$ training observations $(x_1, f(x_1)), \dots, (x_{n-1}, f(x_{n-1}))$. Let $m_0(\cdot) \coloneqq 0$ and $k_0(x,x) \coloneqq k(x,x)$. Then the ML estimator is given by
\begin{equation}
\label{eq:sigma-ml}
  \tausqmlest = \frac{f(x_{1:N})^\top k(x_{1:N}, x_{1:N})^{-1} f(x_{1:N})}{N} = \frac{1}{N} \sum_{n=1}^N \frac{[ f(x_n) - m_{n-1}(x_n) ]^2}{k_{n-1}(x_n, x_n)}.
\end{equation}
This expression of the ML estimator is relatively well-known; see, for example,~\citet[Section~4.2.2]{XuStein2017} or~\citet[Proposition~7.5]{KarvonenOates2023}.
On the other hand, the CV estimator $\tausqcvest$ is given by
\begin{align}
\label{eq:sigma-cv}
    \tausqcvest = \frac{1}{N} \sum_{n=1}^N  \frac{\left[f(x_n) - m_{\setminus  n}(x_n)\right]^2}{ k_{\setminus  n}(x_n, x_n)},
\end{align}%
where  $m_{\setminus n}$ and $k_{\setminus n}$ are the posterior mean and covariance functions of GP interpolation using the kernel $k$ and data $(x_{\setminus n}, f(x_{\setminus n}))$ with $(x_n, f(x_n))$ removed,
\begin{align*}
    m_{\setminus n}(x) &= k(x_{\setminus n}, x)^\top k(x_{\setminus n}, x_{\setminus n})^{-1} f(x_{\setminus n}), \\
    k_{\setminus n}(x, x') &= k(x, x') -  k(x_{\setminus n}, x)^\top k(x_{\setminus n}, x_{\setminus n})^{-1} k(x_{\setminus n}, x').
\end{align*}%

Notice the similarity between the two expressions~\eqref{eq:sigma-ml} and~\eqref{eq:sigma-cv}. The difference is that the ML estimator uses $k_{n-1}$ and $m_{n-1}$, which are based on the first $n-1$ training observations, while the CV estimator uses  $k_{\setminus n}$ and $m_{\setminus n}$ obtained with $N-1$ observations, for each $n = 1,\dots, N$. Therefore, the CV estimator uses all the datapoints more evenly than the ML estimator. This difference may be the source of the difference in their asymptotic properties established later.

\begin{remark}
  \label{remark:ml-cv}
  As suggested by the similarity between~\eqref{eq:sigma-ml} and~\eqref{eq:sigma-cv}, there is a deeper connection between ML and CV estimators in general. For instance, \citet[Proposition~2]{FongHolmes2020} have shown that the Bayesian marginal likelihood equals the average of leave-$p$-out CV scores. Another notable example is the work in~\citet{ginsbourger2021fast}, where the authors showed that, when corrected for the covariance of residuals, the CV estimator of the amplitude parameter reverts to MLE.
\end{remark}

\section{Setting}
\label{sec:setting}

This section describes the settings and tools for our theoretical analysis: the Brownian motion kernel in \Cref{sec:bm-kernel-setting}; sequences of partitions in \Cref{sec:sequence-of-partitions}; fractional Brownian motion in \Cref{sec:fbm}; and functions of finite quadratic variation in \Cref{sec:quadratic-variation}.

\tocless\subsection{Brownian motion kernel\label{sec:bm-kernel-setting}}

As explained at the beginning of the chapter, for the kernel $k$ we focus on the Brownian motion kernel on the domain $\Omega = [0, T]$ for some $T > 0$,
\begin{equation*}
  k(x, x') = \min(x, x').
\end{equation*}
The resulting kernel $k_\tau(x,x') = \tau^2 k(x,x')$ induces a Brownian motion prior for GP interpolation.
We assume the input points $x_1, \dots, x_N$ are ordered,
\begin{equation*}
    0 < x_1 < x_2 < \dots < x_N \leq T.
\end{equation*}
The positivity ensures that the Gram matrix $k(x_{1:N}, x_{1:N})$ is non-singular; the proof is given in~\Cref{sec:explicit-post-mean-cov}.

As is well-known~\citep[see, for instance,][Example~1]{Diaconis1988} and can be seen in \Cref{fig:FBM-H02,fig:IFBM-H05}, the posterior mean function $m_N$ in~\eqref{eq:posterior-moments} using the Brownian motion kernel becomes the {\em piecewise linear interpolant} of the observations $( x_{1:N}, f(x_{1:N}) )$. See~\eqref{eq:explicit-post-mean} and~\eqref{eq:explicit-post-cov} in \Cref{sec:explicit-post-mean-cov} for the proof and explicit expressions of the posterior mean and covariance functions.

\tocless\subsection{Sequences of partitions\label{sec:sequence-of-partitions}}

For our asymptotic analysis, we assume that the input points $x_1, \dots, x_N \in [0,T]$ cover the domain $[0,T]$ more densely as the sample size $N$ increases. To make the dependence on the size $N$ explicit, we write $\prt_N \coloneqq x_{N,1:N} \subset [0,T]$ as a point set of size $N$, and assume that they are ordered as
\begin{equation*}
    0 \eqqcolon x_{N,0} < x_{N, 1} < x_{N, 2} < \dots < x_{N, N} = T
\end{equation*}%
Then $\prt_N$ defines a partition of $[0,T]$ into $N$ subintervals $[x_{N,n}, x_{N, n+1}]$.
When there is no risk of confusion, we may write $x_n$ instead of $x_{N,n}$ for simplicity.
Note that we do {\em not} require the nesting $\prt_N \subset \prt_{N+1}$ of partitions.

We define the {\em mesh size} of partition $\prt_N$ as the longest subinterval,
$$
\|\prt_N\| \coloneqq \max_{n \in \{0, 1,\ldots,N-1\}} (x_{N, n+1} - x_{N, n} )
$$
The decay rate of the mesh size $\|\prt_N\|$ quantifies how quickly the points in $\prt_N$ cover the interval $[0,T]$. In particular, the decay rate $\|\prt_N \| = \bigo(N^{-1})$ implies that the length of every subinterval is asymptotically upper bounded by $1/N$.
At the same time, if each subinterval is asymptotically lower bounded by $1/N$, we call the sequence of partitions $(\prt_N)_{N \in \bN_{\geq 1}}$ {\em quasi-uniform}, more formally defined in~\citet[Definition 4.6]{Wendland2005} as follows.

\begin{definition}
For each $N \in \bN_{\geq 1}$, let $\prt_N \coloneqq (x_{N,n})_{n=1}^N \subset [0,T]$. Define $\Delta x_{N, n} \coloneqq x_{N, n+1} - x_{N, n}$.
Then the sequence of partitions $(\prt_N)_{N \in \bN_{\geq 1}}$ is called {\em quasi-uniform} if there exists a constant  $1 \leq C_\mathrm{qu} < \infty$ such that
\begin{equation*}
\sup_{N \in \bN_{\geq 1}} \frac{\max_n \Delta x_{N, n}}{\min_n \Delta x_{N, n}} = C_\mathrm{qu}.
\end{equation*}
\end{definition}

Quasi-uniformity, as defined here, requires that the ratio of the longest subinterval, $\max_{n} \Delta x_{N, n}$, to the shortest one, $\min_{n} \Delta x_{N, n}$, is upper-bounded by $C_{\mathrm qu}$ for all $N \in \bN_{\geq 1}$. Since $\min_n \Delta x_{N, n} \leq T N^{-1}$ and $\max_n \Delta x_{N, n} \geq T N^{-1}$ for any partition of $[0, T]$, quasi-uniformity implies that all subintervals are asymptotically upper and lower bounded by $1/N$, as we have, for all $N \in \bN_{\geq 1}$ and $n_0 \in \{0, \dots, N-1\}$,
\begin{equation}
  \label{eq:quasi-uniformity-2}
 \frac{T N^{-1} }{C_\mathrm{qu}}  \leq \min_n \Delta x_{N, n} \leq \Delta x_{N, n_0} \leq \max_n \Delta x_{N, n} \leq T C_\mathrm{qu} N^{-1}.
\end{equation}
Therefore, quasi-uniform sequences of partitions are \emph{space-filling designs} that cover the space `almost' uniformly. Trivially, equally-spaced points (or uniform grids) satisfy the quasi-uniformity with $C_{\mathrm qu} = 1$.~\citet{wenzel2021novel} showed that points chosen sequentially to minimise GP posterior variance for a Sobolev kernel are quasi-uniform. We refer to~\citet[p.\@~6]{wynne2021convergence} for further examples and a discussion on quasi-uniformity.

\tocless\subsection{Fractional Brownian motion\label{sec:fbm}} 

\Cref{sec:random-setting} considers the random setting where $f$ is a {\em fractional (or integrated fractional) Brownian motion} (see \citet[e.g.,][Chapter IX]{mandelbrot1982fractal}). 
Examples of these processes can be seen in Figures \Cref{fig:FBM-H02,fig:IFBM-H05,fig:cv,fig:cv-vs-ml}.

A fractional Brownian motion on $[0,T]$ with Hurst parameter $0 < H  < 1$ is a Gaussian process whose kernel is given by
\begin{equation}
\label{eq:fbm-def}
    k_{0,H}(x,x') =  \big( \, \lvert x \rvert^{2H} + \lvert x' \rvert^{2H} - \lvert x-x'\rvert^{2H}\big) / 2.
\end{equation}
Note that if $H = 1/2$, this is the Brownian motion kernel: $k_{0,1/2}(x,x') = \min (x,x')$.
The Hurst parameter $H$ quantifies the smoothness of the fractional Brownian motion.
If $f_{\mathrm FBM} \sim \GP(0, k_{0,H})$ for $H \in (0, 1)$, then $f_{\mathrm FBM} \in C^{0, H-\varepsilon} ( [0, T] )$ almost surely for arbitrarily small $\varepsilon > 0$ \citep[e.g.,][Proposition~1.6]{Nourdin2012}.\footnote{That $f_{\mathrm FBM} \notin C^{0, H} ( [0, T] )$ almost surely for $f_{\mathrm FBM} \sim \GP(0, k_{0,H})$ with $H \in (0, 1)$ is a straightforward corollary of, for example, Theorem~3.2 in~\citet{wang_almost-sure_2007}.}

An integrated fractional Brownian motion with Hurst parameter $H$ is defined via the integration of a fractional Brownian motion with the same Hurst parameter: if $f_{\mathrm FBM} \sim \GP(0, k_{0, H})$, then
\begin{equation*}
  f_\text{iFBM}(x) = \int_0^x f_\text{FBM}(z) \, \mathrm{d} z, \quad x \in [0,T]
\end{equation*}
is an integrated Brownian motion with Hurst parameter $H$. It is a zero-mean GP with the kernel
\begin{equation} \label{eq:iFBM-kernel-explicit}
  \begin{split}
  k_{1,H}(x, x') ={}& \int_0^x \int_0^{x'} \big( \lvert z \rvert^{2H} + \lvert z' \rvert^{2H} - \lvert z-z'\rvert^{2H}\big) / 2 \, \mathrm{d} z \, \mathrm{d} z' \\
  ={}& \frac{1}{2(2H+1)} \bigg( x' x^{2H+1} + x (x')^{2H+1} \\
  &\hspace{1cm} - \frac{1}{2(H+1)} \big[ x^{2H+2} + (x')^{2H+2} - |x - x'|^{2H+2} \big] \bigg).
  \end{split}
\end{equation}
Because differentiating an integrated fractional Brownian motion $f_{\mathrm iFBM} \sim \GP(0, k_{1, H})$ yields a fractional Brownian motion $f_{\mathrm FBM} \sim \GP(0, k_{0,H})$, a sample path of the former satisfies $f_{\mathrm iFBM} \in C^{1,H-\varepsilon}([0, T])$ almost surely for arbitrarily small $\varepsilon > 0$; therefore the smoothness of $f_{\mathrm iFBM}$ is $1 + H$.

\tocless\subsection{Functions of finite quadratic variation\label{sec:quadratic-variation}}

Some of our asymptotic results use the notion of functions of {\em finite quadratic variation}, defined below.
\begin{definition}
For each $N \in \bN_{\geq 1}$, let $\prt_N \coloneqq x_{N, 1:N} \subset [0,T]$, and suppose that $\|\prt_N\| \to 0$ as $N \to \infty$.
Then a function $f : [0, T] \to \bR$ is defined to have {\em finite quadratic variation} with respect to $\prt \coloneqq (\prt_N)_{N \in \bN_{\geq 1}}$, if the limit
\begin{equation}
\label{eq:quadratic-variation}
    V^2(f) \coloneqq \lim_{N \to \infty } \sum_{n=0}^{N-1} \big[ f(x_{N, n+1}) - f(x_{N, n}) \big]^2
\end{equation}
exists and is finite.
We write $V^2(f, \prt)$ when it is necessary to indicate the sequence of partitions.
\end{definition}

Quadratic variation is defined for a specific sequence of partitions $(\prt_N)_{N \in \bN_{\geq 1}}$ and may take different values for different sequences of partitions~\citep[Remark 1.36]{morters2010brownian}.
For conditions that guarantee the invariance of quadratic variation on the sequence of partitions, see, for instance,~\citet{ContBas2023}.
Note also that the notion of quadratic variation differs from that of $p$-variation for $p=2$, which is defined as the supremum over all possible sequences of partitions whose mesh sizes tend to zero.

If $f \in C^{0,\alpha}([0, T])$ with $\alpha > 1/2$ and $\|\prt_N\| = \bigo(N^{-1})$ as $N \to \infty$, then we have $V^2(f) = 0$, because in this case
\begin{align*}
  \sum_{n=0}^{N-1} \big[ f(x_{N, n+1}) - f(x_{N, n}) \big]^2 \leq   N L^2 \max_n (\Delta x_{N, n})^{2 \alpha} &= \bigo (N^{1 - 2\alpha}) \to 0
\end{align*}
as $N \to \infty$.
Therefore, given the inclusion properties of H\"older spaces given in \Cref{sec:quantifying_smoothness_of_functions}, we arrive at the following standard proposition.

\begin{proposition}
\label{res:qv_of_smooth_functions}
Suppose that the partitions $(\prt_N)_{N \in \bN_{\geq 1}}$ are such that $\| \prt_N \| = \bigo (N^{-1})$. If $f \in C^{s, \alpha}([0, T])$ for $s+\alpha>1/2$, then $V^2(f) = 0$.
\end{proposition}
If the mesh size tends to zero faster than $1/\log N$, in that $\| \prt_N \|=o(1/\log N)$, then the quadratic variation of almost every sample path of the Brownian motion on the interval $[0, T]$ equals~$T$~\citep{10.2307/2959347}.
This is of course true for partitions which have the faster decay $\| \prt_N \|=\bigo(N^{-1})$.

\section{Theoretical Analysis}
\label{sec:limit-behaviour-for-sigma}

This section presents our main results on the asymptotic properties of the CV and ML estimators, $\tausqcvest$ and $\tausqmlest$, for the amplitude parameter. \Cref{sec:results-deterministic} considers the deterministic setting where the integrand $f$ is fixed and assumed to belong to a H\"older space. \Cref{sec:random-setting} studies the random setting where $f$ is an (integrated) fractional Brownian motion. In~\Cref{sec:icv-estimators}, we use the insights obtained in the proofs for the deterministic and random settings to propose a \emph{interior cross-validation} (ICV) estimator, and show its asymptotic properties are an improvement on those of CV and ML estimators.

\tocless\subsection{Deterministic setting\label{sec:results-deterministic}} 

We present our main results for the deterministic case where the integrand $f$ is fixed and assumed to be in a H\"older space $C^{s, \alpha}([0, T])$.  \Cref{res:holder-spaces} below provides asymptotic upper bounds on the CV estimator $\tausqcvest$ for different values of the smoothness parameters $s$ and $\alpha$ of the H\"older space.

\begin{theorem}[Rate of CV decay in H\"older spaces]
\label{res:holder-spaces}
Suppose that $f$ is an element of $C^{s, \alpha}([0, T])$, with $s \geq 0$ and $0 < \alpha \leq 1$, such that  $f(0)=0$, and the interval partitions $(\prt_N)_{N \in \bN_{\geq 1}}$ have bounded mesh sizes $\|\prt_N \|=\bigo(N^{-1})$ as $N \to \infty$. Then
\begin{equation}
\label{eq:main-result}
\tausqcvest = \bigo\big( N^{1 - \min\{2(s + \alpha), 3\}} \big)
= \begin{cases}
    \bigo\left(N^{1 - 2 \alpha}\right) &\text{ if } \quad s = 0, \\
    \bigo\left(N^{-1 - 2\alpha}\right) &\text{ if } \quad  s = 1 \text{ and } \alpha < 1/2, \\
    \bigo\left(N^{- 2}\right) &\text{ if } \quad s = 1 \text{ and } \alpha \geq 1/2, \\
    \bigo\left(N^{- 2}\right) &\text{ if } \quad s \geq 2.
    \end{cases}
\end{equation}
\end{theorem}
\begin{proof}
See \Cref{sec:proofs-deterministic}.
\end{proof}

\Cref{res:holder-spaces-ml} below is a corresponding result for the ML estimator $\tausqmlest$. Note that a similar result has been obtained by \citet[Proposition~4.5]{Karvonen2020}, where the function $f$ is assumed to belong to a Sobolev space and the kernel is a Mat\'ern kernel. \Cref{res:holder-spaces-ml} is a version of this result where $f$ is in a H\"older space and the kernel is the Brownian motion kernel; we provide it for completeness and ease of comparison.

\begin{theorem}[Rate of ML decay in H\"older spaces]
\label{res:holder-spaces-ml}
Suppose that $f$ is a non-zero element of $C^{s, \alpha}([0, T])$, with $s \geq 0$ and $0 < \alpha \leq 1$, such that  $f(0)=0$, and the interval partitions $(\prt_N)_{N \in \bN_{\geq 1}}$ have bounded mesh sizes $\|\prt_N \|=\bigo(N^{-1})$ as $N \to \infty$. Then
\begin{equation}
\label{eq:main-result-ml}
\tausqmlest = \bigo\big( N^{1 - \min\{2(s + \alpha), 2\}} \big)
= \begin{cases}
    \bigo\left(N^{1 - 2 \alpha}\right) &\text{ if } \quad s = 0, \\
    \Theta\left(N^{- 1}\right) &\text{ if } \quad s \geq 1.
    \end{cases}
\end{equation}
\end{theorem}
\begin{proof}
  See \Cref{sec:proofs-deterministic}. The proof is similar to that of \Cref{res:holder-spaces}.
\end{proof}

\begin{figure}
    \centering
    \includegraphics[width=0.8\textwidth]{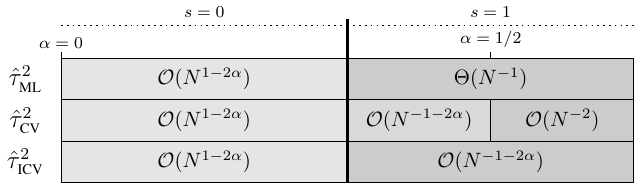}
    \caption[Illustrating rates in \Cref{res:holder-spaces,res:holder-spaces-ml,res:holder-spaces-icv}.]{Rates of decay for the ML, CV and ICV estimators from \Cref{res:holder-spaces,res:holder-spaces-ml,res:holder-spaces-icv}.
    Observe that the CV estimator's range of adaptation to the smoothness $s+\alpha$ is wider than the ML estimator's, and the ICV estimator's range of adaptation is wider than that for both the CV and ML estimators.}
    \label{fig:rate-regimes-intro-deterministic}
\end{figure}

\Cref{fig:rate-regimes-intro-deterministic} summarises the rates of \Cref{res:holder-spaces,res:holder-spaces-ml}. When $s + \alpha \leq 1$ (or $s = 0$ and $\alpha \leq 1$), the rates of  $\tausqcvest$ and $\tausqmlest$ are $\bigo(N^{1-2\alpha})$, so both of them may decay (or grow, for $s+\alpha<1/2$) adaptively to the smoothness $s+\alpha$ of the integrand $f$. However, when $s + \alpha > 1$, the situation is different: the decay rate of $\tausqmlest$ is always $\Theta(N^{-1})$ and thus insensitive to $\alpha$, while that of $\tausqcvest$ is $\bigo(N^{-1 - 2\alpha})$ for $s=1$ and $\alpha \in (0, 1/2]$. Therefore, the CV estimator may be adaptive to a broader range of the smoothness $0 < s + \alpha \leq 3/2$ of the integrand $f$ than the ML estimator (whose range of adaptation is $0 < s + \alpha \leq 1$).

Note that \Cref{res:holder-spaces,res:holder-spaces-ml} provide asymptotic upper bounds (except for the case $s \geq 1$ of \Cref{res:holder-spaces-ml}) and may not be tight if the integrand $f$ is smoother than `typical' functions in $C^{s,\alpha}([0,T])$.\footnote{For example, if $f(x) = | x - 1/2 |$ with $T = 1$, we have $f \in C^{0,1} ([0,T])$, as $f$ is Lipschitz continuous in this case. However, $f$ is almost everywhere infinitely differentiable except at one point $x = 1/2$, so it is, in this sense, much smoother than `typical' functions in $C^{0,1} ([0, T])$.} In \Cref{sec:random-setting}, we show that the bounds are indeed tight in expectation if $f$ is a fractional (or integrated fractional) Brownian motion with smoothness $s + \alpha$.

In the deterministic setting, a potential approach for obtaining a matching lower bound could use the rate of decay of the Fourier coefficients as a notion of smoothness, instead of the H\"older smoothness condition on the function $f$. Certain self-similarity conditions based on the decay rate and behaviour of Fourier coefficients are routinely used to study coverage of Bayesian credible sets~\citep[e.g.,][]{Szabo2015, HadjiSzabo2021} as they define classes of functions that cannot `deceive' parameter estimators. Motivated by this, we attempted to adapt the argument in~\citet[Section~4.2]{sniekers2015adaptive} and \citet[Section~10]{sniekers2020adaptive} to derive a matching lower bound under a self-similarity assumption on the Fourier coefficients. However, the bounds obtained through this approach proved sub-optimal in our setting. A different technique may therefore be required.

\begin{remark}
The proof of \Cref{res:holder-spaces-ml} shows that for $s = 1$ we have $\tausqmlest = \Theta(N^{-1})$ whenever $\|\prt_N \| \to 0$ as $N \to \infty$.
More precisely, it establishes that
  \begin{equation*}
    N \tausqmlest \to \| f' \|_{\calL^2([0, T])} \coloneqq\int_0^T f'(x)^2 \, \mathrm{d} x \quad \text{ as } \quad N \to \infty.
  \end{equation*}
Note that the $\calL^2([0, T])$ norm of $f'$ in the right hand side equals the norm of $f$ in the reproducing kernel Hilbert space of the Brownian motion kernel~\citep[e.g.,][Section~10]{VaartZanten2008}
Therefore, this fact is consistent with a more general statement in \citet[Proposition~3.1]{Karvonen2020}.

\end{remark}

In addition to the above results, \Cref{res:fqv-estimator} below shows the limit of the CV estimator $\tausqcvest$ if the integrand $f$ is of finite quadratic variation.

\begin{theorem}
\label{res:fqv-estimator}
For each $N \in \bN_{\geq 1}$,  let $\prt_N \subset [0,T]$ be the equally-spaced partition of size $N$.
Suppose that $f: [0,T] \to \bR$ has finite quadratic variation $V^2(f)$ with respect to $(\prt_N)_{N \in \bN_{\geq 1}}$, $f(0) = 0$, and $f$ is continuous on the boundary, i.e., $\lim_{x \to 0^+} f(x) = f(0)$ and $\lim_{x \to T^-} f(x) = f(T)$.
Moreover, suppose that the quadratic variation $V^2(f)$ remains the same for all sequences of quasi-uniform partitions with constant $C_\mathrm{qu}=2$.\footnote{In \Cref{sec:discussion-thm-fqv-estimator}, we discuss the relaxation of this requirement.}
Then
\begin{equation} \label{eq:sigma-cv-limit-qvar}
  \lim_{N \to \infty} \tausqcvest = \frac{V^2(f)}{T}.
\end{equation}
\end{theorem}
\begin{proof}
See \Cref{sec:proofs-deterministic}.
\end{proof}

For the ML estimator $\tausqmlest$, it is straightforward to obtain a similar result by using~\eqref{eq:quasi-uniformity-2} and~\eqref{eq:sigma-ml-w-deltas} in \Cref{sec:explicit-post-mean-cov}: Under the same conditions as \Cref{res:fqv-estimator}, we have
\begin{equation} \label{eq:ml-quad-var}
  \lim_{N \to \infty} \tausqmlest =\frac{V^2(f)}{T}.
\end{equation}

\Cref{res:fqv-estimator} and~\eqref{eq:ml-quad-var} are consistent with \Cref{res:holder-spaces,res:holder-spaces-ml}, which assume $f \in C^{s, \alpha}([0,T])$ with $s + \alpha > 1/2$ and imply $\tausqcvest \to 0$ and $\tausqmlest \to 0$ as $N \to \infty$. As summarised in~\Cref{res:qv_of_smooth_functions}, we have $V^2(f) = 0$ for $f \in C^{s, \alpha}([0,T])$ with $s + \alpha > 1/2$, so \Cref{res:fqv-estimator} and~\eqref{eq:ml-quad-var} imply that $\tausqcvest \to 0$ and $\tausqmlest \to 0$  as $N \to \infty$.

When $f$ is a Brownian motion, in which case the Brownian motion prior is well-specified, the smoothness of $f$ is $s + \alpha = 1/2$, and the quadratic variation $V^2(f)$ becomes a positive constant~\citep{10.2307/2959347}. \Cref{prop:bm-almost-sure} in the next subsection shows that this fact,  \Cref{res:fqv-estimator}, and~\eqref{eq:ml-quad-var} lead to the consistency of the ML and CV estimators in the well-specified setting.

\tocless\subsection{Random setting\label{sec:random-setting}}

In \Cref{sec:results-deterministic}, we obtained asymptotic upper bounds on the CV and ML amplitude estimators when the true function $f$ is a fixed function in a H\"older space. This section shows that these asymptotic bounds are tight in expectation when $f$ is a fractional (or integrated fractional) Brownian motion.

\begin{figure}
    \centering
    \includegraphics[width=0.8\textwidth]{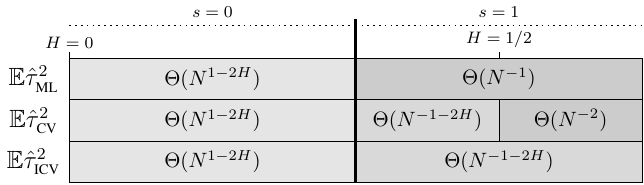}
    \caption[Illustrating rates in \Cref{res:holder-spaces-exp,res:holder-spaces-exp-ml,res:holder-spaces-icv-exp}.]{Expected decay rates for the ML, CV and ICV estimators from \Cref{res:holder-spaces-exp,res:holder-spaces-exp-ml,res:holder-spaces-icv-exp}.
    Observe that the CV estimator's range of adaptation to the smoothness $s+H$ is wider than the ML estimator's, and the ICV estimator's range of adaptation is wider than that for both the CV and ML estimators.
    }
    \label{fig:rate-regimes-intro}
\end{figure}

That is, we consider the asymptotics of the expectations $\bE[\tausqcvest]$ and $\bE[\tausqmlest]$ under the assumption that $f \sim \GP(0, k_{s, H})$, where $k_{s, H}$ is the kernel of a fractional Brownian motion~\eqref{eq:fbm-def} for $s = 0$ or that of an integrated fractional Brownian motion~\eqref{eq:iFBM-kernel-explicit} for $s = 1$, with $0 < H <1$ being the Hurst parameter. Recall that $f \sim \GP(0, k_{s, H})$ belongs to the H\"older space $C^{s, H - \varepsilon}([0,T])$ almost surely for arbitrarily small $\varepsilon > 0$, so its smoothness is $s + H$.  \Cref{fig:rate-regimes-intro} summarises the obtained upper and lower rates, corroborating the upper rates in \Cref{fig:rate-regimes-intro-deterministic}.

\Cref{res:holder-spaces-exp,res:holder-spaces-exp-ml} below establish the asymptotic upper and lower bounds for the CV and ML estimators, respectively.

\begin{theorem}[Expected CV rate for fractional Brownian motion]
\label{res:holder-spaces-exp}
    Suppose that $(\prt_N)_{N \in \bN_{\geq 1}}$ are quasi-uniform and $f \sim \GP(0, k_{s,H})$ with $s \in \{0, 1\}$ and $0 < H < 1$.
    Then
    \begin{equation*}
      \bE[\tausqcvest] = \Theta ( N^{1 - \min\{2(s + H),3\}} ) =
      \begin{cases}
        \Theta\left(N^{1 - 2  H}\right) &\text{ if } \quad s = 0 \text{ and } H \in (0, 1), \\
        \Theta\left(N^{-1 - 2H}\right) &\text{ if } \quad  s = 1 \text{ and } H < 1/2, \\
        \Theta\left(N^{- 2}\right) &\text{ if } \quad s = 1 \text{ and } H \geq 1/2. \\
      \end{cases}
    \end{equation*}
\end{theorem}
\begin{proof}
  See \Cref{sec:proofs-random}.
\end{proof}

\begin{theorem}[Expected ML rate for fractional Brownian motion]
\label{res:holder-spaces-exp-ml}
      Suppose that $(\prt_N)_{N \in \bN_{\geq 1}}$ are quasi-uniform and $f \sim \GP(0, k_{s,H})$ with $s \in \{0, 1\}$ and $0 < H < 1$.
    Then
    \begin{equation*}
      \bE[\tausqmlest] = \Theta ( N^{1 - \min\{2(s + H),2\}} ) =
      \begin{cases}
        \Theta\left(N^{1 - 2  H}\right) &\text{ if } \quad s = 0 \text{ and } H \in (0, 1), \\
        \Theta\left(N^{-1}\right) &\text{ if } \quad  s = 1 \text{ and } H \in (0, 1).
      \end{cases}
    \end{equation*}
\end{theorem}
\begin{proof}
  See \Cref{sec:proofs-random}. The proof is similar to that of \Cref{res:holder-spaces-exp}.
\end{proof}

\Cref{res:holder-spaces-exp,res:holder-spaces-exp-ml} show that the CV estimator is adaptive to the unknown smoothness $s + H$ of the integrand $f$ for a broader range $0< s+H \leq 3/2$ than the ML estimator, whose range of adaptation is $0 < s+H \leq 1$. These results imply that the CV estimator can be asymptotically well-calibrated for a broader range of unknown smoothness than the ML estimator, as discussed in \Cref{sec:discussion}.

When the smoothness of $f$ is less than $1/2$, i.e., when $s + H < 1/2$, the Brownian motion prior, whose smoothness is $1/2$, is smoother than $f$. In this case, the expected rates of $\tausqcvest$ and $\tausqmlest$ are $ \Theta\left(N^{1 - 2  H}\right)$ and increase as $N$ increases. The increase of $\tausqcvest$ and $\tausqmlest$ can be interpreted as compensating the overconfidence of the posterior standard deviation $\sigma_\BQ$, which decays too fast to be asymptotically well-calibrated. This interpretation agrees with the illustration in \Cref{fig:FBM-H02}.

On the other hand, when $s+ H > 1/2$, the integrand $f$ is smoother than the Brownian motion prior. In this case,  $ \tausqcvest$ and $\tausqmlest$ decrease as $N$ increases, compensating for the under-confidence of the posterior standard deviation $\sigma_\BQ$. See \Cref{fig:IFBM-H05} for an illustration.

When $s + H = 1/2$, this is the well-specified case in that the smoothness of $f$ matches the Brownian motion prior. In this case,  \Cref{res:holder-spaces-exp,res:holder-spaces-exp-ml} yield $\bE[\tausqcvest] = \Theta(1)$ and $\bE[\tausqmlest] = \Theta(1)$, i.e., when the CV and ML estimators converge, they converge to a positive constant.
The following proposition, which follows from \Cref{res:fqv-estimator} and~\eqref{eq:ml-quad-var}, shows that this limiting constant is the true value of the amplitude parameter $\tau_0^2$ in the well-specified setting $f \sim \GP(0, \tau_0^2 k)$, recovering similar results in the literature \citep[e.g.,][Theorem 2]{Bachoc2017}.

\begin{proposition} \label{prop:bm-almost-sure}
  Suppose that $f \sim \GP(0, \tau_0^2 k)$ for $\tau_0 > 0$ and that partitions  $(\prt_N)_{N \in \bN_{\geq 1}}$ are equally-spaced. Then
  \begin{equation*}
    \lim_{N \to \infty} \tausqcvest = \lim_{N \to \infty} \tausqmlest = \tau_0^2 \quad \text{ almost surely}.
  \end{equation*}
\end{proposition}
\begin{proof}
  Since the quadratic variation of almost all sample paths of the unscaled (i.e., $\tau_0 = 1$) Brownian motion on $[0, T]$ equals $T$~\citep{10.2307/2959347}, the claim follows from~\eqref{eq:sigma-cv-limit-qvar} and~\eqref{eq:ml-quad-var}.
\end{proof}

In~\Cref{sec:discussion}, we discuss the implications of the obtained asymptotic rates of $\tausqcvest$ and $\tausqmlest$ on the reliability of the resulting BQ uncertainty estimates. Before turning to that discussion, we propose a modification to the cross-validation procedure, motivated by the results in \Cref{res:holder-spaces} and \Cref{res:holder-spaces-exp}, that may have better asymptotic properties than the CV estimator.

\tocless\subsection{Interior cross-validation estimators\label{sec:icv-estimators}}

The proofs of~\Cref{res:holder-spaces,res:holder-spaces-exp} show that when $s=1$ and $\alpha \in (1/2, 1]$, the bound on $\tausqcvest$ is dominated by the bound on what we call the boundary terms.
These are the terms corresponding to $n = 1$ and $n = N$ in~\eqref{eq:sigma-cv}; see also~\eqref{eq:boundary-terms}.
That the boundary terms dominate is unsurprising since prediction at boundary points is a more challenging task than prediction at the interior. Motivated by this observation, we propose an alternative estimation method called \emph{interior cross validation} (ICV) that maximises
\begin{equation*} 
\label{eq:icv-estimator-any-kernel}
  \sum_{n=2}^{N-1} \log p( f(x_n) \given x_n, x_{\setminus n}, f(x_{\setminus n}), \theta) .
\end{equation*}
The corresponding amplitude parameter estimator is
\begin{equation}
\label{eq:icv-estimator}
    \tausqicvest = \frac{1}{N} \sum_{n=2}^{N-1}  \frac{\left[f(x_n) - m_{\setminus  n}(x_n)\right]^2}{ k_{\setminus  n}(x_n, x_n)}.
\end{equation}
With the boundary points removed, the estimator's range of adaptation to the smoothness of the true function is greater than that for the CV estimator, as illustrated in~\Cref{fig:rate-regimes-intro-deterministic} for the deterministic setting and~\Cref{fig:rate-regimes-intro} for the random setting. We present formal results for the deterministic and the random settings in the following theorems. 

\begin{theorem}[Rate of ICV decay in H\"older spaces]
\label{res:holder-spaces-icv}
Suppose that $f$ is an element of $C^{s, \alpha}([0, T])$, with $s \geq 0$ and $0 < \alpha \leq 1$, such that $f(0)=0$, and the interval partitions $(\prt_N)_{N \in \bN_{\geq 1}}$ have bounded mesh sizes $\|\prt_N \|=\bigo(N^{-1})$ as $N \to \infty$. Then
\begin{equation*}
\tausqicvest = \bigo\big( N^{1 - \min\{2(s + \alpha), 4\}} \big)
= \begin{cases}
    \bigo\left(N^{1 - 2 \alpha}\right) &\text{ if } \quad s = 0, \\
    \bigo\left(N^{-1 - 2\alpha}\right) &\text{ if } \quad  s = 1, \\
    \bigo\left(N^{- 3}\right) &\text{ if } \quad s \geq 2.
    \end{cases}
\end{equation*}
\end{theorem}
\begin{proof}
See \Cref{sec:proofs-icv}.
\end{proof}

\begin{theorem}[Expected ICV rate for fractional Brownian motion]
\label{res:holder-spaces-icv-exp}
    Suppose that $(\prt_N)_{N \in \bN_{\geq 1}}$ are quasi-uniform and $f \sim \GP(0, k_{s,H})$ with $s \in \{0, 1\}$ and $0 < H < 1$.
    Then
    \begin{equation*}
      \bE \tausqicvest = \Theta ( N^{1 - \min\{2(s + H),4\}} ) =
      \begin{cases}
        \Theta\left(N^{1 - 2  H}\right) &\text{ if } \quad s = 0, \\
        \Theta\left(N^{-1 - 2H}\right) &\text{ if } \quad  s = 1.
      \end{cases}
    \end{equation*}
\end{theorem}
\begin{proof}
  See \Cref{sec:proofs-icv}.
\end{proof}

This idea can be taken further. For the Brownian motion kernel, an estimator that does not attempt to predict on points `close enough' to the boundary,
\begin{equation*}
    \tausqicvest[N_0] = \frac{1}{N} \sum_{n=N_0}^{N-N_0}  \frac{\left[f(x_n) - m_{\setminus  n}(x_n)\right]^2}{ k_{\setminus  n}(x_n, x_n)}
\end{equation*}
for some fixed $N_0$, has the same range of adaptation as $\tausqicvest= \tausqicvest[1]$, the estimator that only ignores the points on the boundary. However, for smoother kernels like  integrated fractional Brownian motion (iFBM) and the Mat\'ern family, $\tausqicvest[N_0]$ may exhibit adaptation beyond the level $s=2$. The number of boundary points $N_0$ to remove would likely depend on the smoothness of the kernel. We conjecture that this could be understood through an analogy with finite differences: the leave-one-out residuals at interior points behave like centered difference stencils, whose width, and thus sensitivity to boundary effects, increases with the smoothness of the kernel. Investigating model-dependent cross-validation estimators that discard a proportion of boundary points would be an interesting direction for future work.

\section{Consequences for credible intervals}
\label{sec:discussion}

This section discusses whether the estimated amplitude parameter, given by the CV or ML estimator, leads to asymptotically well-calibrated credible intervals.
 With the kernel $\hat{\tau}^2 k(x,x')$, where $\hat{\tau}^2 = \tausqcvest$ or   $\hat{\tau}^2 = \tausqmlest$,  an $\alpha$-credible interval is given by
 \begin{equation} \label{eq:CI-discus}
     [I_\BQ - \alpha \hat{\tau} \sigma_\BQ,\quad I_\BQ + \alpha \hat{\tau} \sigma_\BQ]
 \end{equation}
where $\alpha > 0$ is a constant (e.g., $\alpha \approx 1.96$ leads to the 95\% credible interval).

As discussed in \Cref{sec:gpcv-introduction}, this credible interval~\eqref{eq:CI-discus} is asymptotically well-calibrated, if it shrinks to $0$ at the same speed as the decay of the error $|I_\BQ - I|$ as $N$ increases, i.e., the ratio
\begin{equation} \label{eq:ratio-discus}
    \frac{|I - I_\BQ | }{ \hat{\tau} \sigma_\BQ}
\end{equation}
should neither diverge to infinity nor converge to $0$.
If this ratio diverges to infinity, the credible interval~\eqref{eq:CI-discus} is asymptotically overconfident, in that~\eqref{eq:CI-discus} shrinks to $0$ faster than the actual error $|I - I_\BQ |$. If the ratio converges to $0$, the credible interval is asymptotically underconfident, as it increasingly overestimates the actual error. Therefore, the ratio~\eqref{eq:ratio-discus} should ideally converge to a positive constant for the credible interval~\eqref{eq:CI-discus} to be reliable.

For ease of analysis,  we focus on the random setting in~\Cref{sec:random-setting} where $f$ is a fractional (or integrated fractional) Brownian motion and where we obtained asymptotic upper and lower bounds for $\bE[\tausqcvest]$ and $\bE[\tausqmlest]$.
We study how the expectation of the posterior variance $\bE[\hat{\tau}^2] \sigma^2_\BQ$ scales with the expected squared error $\bE [ (I - I_\BQ)^2  ]$.
Specifically, we analyse their ratio for $\hat{\tau}^2 = \tausqcvest$ and $\hat{\tau}^2 = \tausqmlest$,
\begin{equation} \label{eq:ratios-expectation-bq}
  R_\mathrm{CV}^\BQ (N) \coloneqq \frac{\bE [ I - I_\BQ  ]^2}{\bE[\tausqcvest] \sigma^2_\BQ } \quad \text{ and } \quad R_\mathrm{ML}^\BQ (N) \coloneqq \frac{\bE [ I - I_\BQ  ]^2}{\bE[\tausqmlest] \sigma^2_\BQ }.
\end{equation}
The ratio diverging to infinity (resp.~converging to $0$) as $N \to \infty$ suggests that the credible interval~\eqref{eq:CI-discus} is asymptotically overconfident (resp.~underconfident) for a non-zero probability of the samples of $f$. Thus ideally, the ratio should converge to a positive constant.

\Cref{res:uq-theorem-exp-bq} establishes the asymptotic rates of the ratios in~\eqref{eq:ratios-expectation-bq}. To facilitate the analysis, we strengthen the requirement on $(\prt_N)_{N \in \bN_{\geq 1}}$, from quasi-uniformity to uniformity (i.e., quasi-uniformity with $C_{\mathrm{qu}}=1$), and ensure the integrating measure $\bP$ is such that an integral against $\bP$ can be lower- and upper bounded with an integral against the Lebesgue measure.

\begin{theorem}
  \label{res:uq-theorem-exp-bq}
  Suppose that $(\prt_N)_{N \in \bN_{\geq 1}}$ are uniform and $f \sim \GP(0, k_{s, H})$ for $s \in \{0,1 \}$ and $0 < H < 1$, and the integrating measure $\bP$ has a density $f_\bP: [0, T] \to [c_0, C_0]$ for some $c_0, C_0>0$. Then,
  \begin{equation*}
    R_\mathrm{CV}^\BQ (N)
    =
      \begin{cases}
        \Theta(1) &\text{ if } \quad s = 0 \text{ and } H \in (0, 1), \\
        \Theta(1) &\text{ if } \quad s = 1 \text{ and } H \in (0, 1),
      \end{cases}
  \end{equation*}
  and
  \begin{equation*}
    R_\mathrm{ML}^\BQ (N)
    =
      \begin{cases}
        \Theta(1) &\text{ if } \quad s = 0 \text{ and } H \in (0, 1), \\
        \Theta\left(N^{-2H}\right) &\text{ if } \quad s = 1 \text{ and } H < 1/2 \\
        \Theta\left(N^{-1}\right) &\text{ if } \quad s = 1 \text{ and } H \geq 1/2.
      \end{cases}
  \end{equation*}
\end{theorem}
\begin{proof}
  See \Cref{sec:proofs-discussion}.
\end{proof}

We have the following observations from \Cref{res:uq-theorem-exp-bq}, which suggest an advantage of the CV estimator over the ML estimator for BQ uncertainty quantification.
\begin{itemize}
    \item The ratio for the CV estimator neither diverges to infinity nor decays to $0$ across the entire range $0 < s+H < 2$, which is broader than that of the ML estimator, $0 < s+H < 1$. This observation suggests that the CV estimator can yield asymptotically well-calibrated credible intervals for a broader range of the unknown smoothness $s + H$ of the function $f$ than the ML estimator.

    \item In the range $1 < s+H < 2$, the ML estimator may yield asymptotically underconfident credible intervals, at a $N^{-1}$ gap at worst.
\end{itemize}

Further, we may analogously assess the impact of $\bE[\tausqcvest]$ and $\bE[\tausqmlest]$ in~\Cref{sec:random-setting} on uncertainty quantification in the underlying GP interpolation. Define error-variance ratios analogous to~\eqref{eq:ratios-expectation-bq} for pointwise GP uncertainty quantification,
\begin{equation} \label{eq:ratios-expectation-gp}
  R_\mathrm{CV}^\GP (x, N) \coloneqq \frac{\bE [ f(x) - m_N(x)  ]^2}{\bE[\tausqcvest] k_N(x, x)} \quad \text{ and } \quad R_\mathrm{ML}^\GP (x, N) \coloneqq \frac{\bE [ f(x) - m_N(x)  ]^2}{\bE[\tausqmlest] k_N(x, x) }.
\end{equation}
Then, the following holds.

\begin{theorem}
  \label{res:uq-theorem-exp}
  Suppose that $(\prt_N)_{N \in \bN_{\geq 1}}$ are quasi-uniform and $f \sim \GP(0, k_{s, H})$ for $s \in \{0,1 \}$ and $0 < H < 1$. Then,
  \begin{equation*}
    \sup_{x \in [0, T]} R_\mathrm{CV}^\GP (x, N)
    =
      \begin{cases}
        \Theta(1) &\text{ if } \quad s = 0 \text{ and } H \in (0, 1), \\
        \Theta(1) &\text{ if } \quad s = 1 \text{ and } H \in (0, 1/2), \\
        \Theta\left(N^{1-2H}\right) &\text{ if } \quad s = 1 \text{ and } H \in (1/2, 1), \\
      \end{cases}
  \end{equation*}
  and
  \begin{equation*}
    \sup_{x \in [0, T]} R_\mathrm{ML}^\GP (x, N)
    =
      \begin{cases}
        \Theta(1) &\text{ if } \quad s = 0 \text{ and } H \in (0, 1), \\
        \Theta\left(N^{-2H}\right) &\text{ if } \quad s = 1 \text{ and } H \in (0, 1).
      \end{cases}
  \end{equation*}
\end{theorem}
\begin{proof}
  See \Cref{sec:proofs-discussion}.
\end{proof}

The difference in rates between \Cref{res:uq-theorem-exp} and \Cref{res:uq-theorem-exp-bq} illustrates the point made at the beginning of this chapter: in general, pointwise GP calibration need not match BQ calibration. In our case, BQ appears asymptotically more confident than GP for $3/2 < s+H < 2$ for both CV and ML estimators; notably, the CV estimator credible intervals flip from underconfidence (for GP interpolation) to being well-calibrated (for BQ).
Moreover, for the interior CV estimator introduced in~\eqref{eq:icv-estimator}, it immediately follows from the proof in~\Cref{sec:proofs-discussion} that
\begin{equation*}
\sup_{x \in [0, T]} R_\mathrm{ICV}^\GP (x, N)
=
    \begin{cases}
    \Theta(1) &\text{ if } \quad s = 0 \text{ and } H \in (0, 1), \\
    \Theta(1) &\text{ if } \quad s = 1 \text{ and } H \in (0, 1), 
    \end{cases}
\end{equation*}
but
\begin{equation*}
R_\mathrm{ICV}^\BQ (N)
=
    \begin{cases}
    \Theta(1) &\text{ if } \quad s = 0 \text{ and } H \in (0, 1), \\
    \Theta(1) &\text{ if } \quad s = 1 \text{ and } H \in (0, 1/2), \\
    \Theta(N^{1-2H}) &\text{ if } \quad s = 1 \text{ and } H \in (1/2, 1), 
    \end{cases}
\end{equation*}
which implies that, in terms of utility of the ML, CV, and ICV estimators for asymptotic calibration, (1) the ICV estimator is most optimal for pointwise GP calibration, (2) the CV estimator is most optimal for BQ calibration.

\section{Experiments}
\label{sec:experiments}

This section describes numerical experiments to substantiate the theoretical results in \Cref{sec:limit-behaviour-for-sigma}.
We define test functions in \Cref{sec:test-functions}, show empirical asymptotic results for the CV estimator in \Cref{sec:experiment-cv}, and report comparisons between the CV and ML estimators in \Cref{sec:comparison-CV-ML}.

For a continuous function $f$, define $s[f] \in \bN$ and $\alpha[f] \in (0, 1]$ as
\begin{equation} \label{eq:smoothness-1037}
\begin{split}
        s[f] &\coloneqq \sup\{ s \in \bN: f \in C^{s}([0, T]) \}, \\
        \alpha[f] &\coloneqq \sup\{ \alpha \in (0, 1] : f \in C^{s[f],\alpha}([0, T]) \}.
\end{split}
\end{equation}
Then, for arbitrarily small $\varepsilon_1,\varepsilon_2 > 0$, we have
\begin{equation*}
  f \in C^{\max ( s[f]-\varepsilon_1, 0),\alpha[f]-\varepsilon_2}([0, T]) \quad \text{ and } \quad f \notin C^{s[f]+\varepsilon_1,\alpha[f]+\varepsilon_2}([0, T]).
\end{equation*}
In this sense, $s[f]$ and $\alpha[f]$ characterise the smoothness of $f$.

\tocless\subsection{Test functions\label{sec:test-functions}}

We generate test functions $f: [0,1] \to \bR$ as sample paths of stochastic processes with varying degrees of smoothness, as defined below.
The left columns of \Cref{fig:cv,fig:cv-vs-ml} show samples of these functions.

\begin{itemize}
    \item
To generate nowhere differentiable test functions, we use the Brownian motion (BM), the Ornstein--Uhlenbeck process (OU), and the fractional Brownian motion (FBM\footnote{We use \url{https://github.com/crflynn/fbm} to sample from FBM.}), which are zero-mean GPs with kernels
\begin{align*}
    k_\mathrm{BM}(x,x') & = \min(x, x'),\quad   k_\mathrm{OU}(x,x')  = \big(e^{- \lambda \lvert x-x' \rvert} - e^{-\lambda (x+x')}\big) / 4,  \\
    k_\mathrm{FBM}(x,x') & =  \big( \, \lvert x \rvert^{2H} + \lvert x' \rvert^{2H} - \lvert x-x'\rvert^{2H}\big) / 2,
\end{align*}
where $\lambda > 0$ and $0<H<1$ is the Hurst parameter (recall that the FBM $=$ BM if $H = 1/2$). We set $\lambda = 0.2$ in the experiments below.
Almost all samples $f$ from these processes satisfy $s[f] = 0$.
For BM and OU we have $\alpha[f] = 1/2$ and for FBM $\alpha[f] = H$ (see \Cref{sec:fbm}).
It is well-known that the OU process with the kernel $k_\mathrm{OU}$ above satisfies the stochastic differential equation
\begin{equation} \label{eq:OU-SDE}
  \mathrm{d} f(t) = -\lambda f(t) \mathrm{d} t + \sqrt{\frac{\lambda}{2}} \, \mathrm{d} B(t),
\end{equation}
where $B$ is the standard Brownian motion whose kernel is $k_{\mathrm BM}$.

\item
To generate differentiable test functions, we use once (iFBM) and twice (iiFBM) integrated fractional Brownian motions
\begin{equation*}
    f_\mathrm{iFBM}(x) =\int_0^x f_\mathrm{FBM}(z) \, \mathrm{d} z \quad \text{ and } \quad f_\mathrm{iiFBM}(x) =\int_0^x f_\mathrm{iFBM}(z) \, \mathrm{d} z,
\end{equation*}
where $f_\mathrm{FBM} \sim \GP(0, k_{\mathrm FBM})$.
See~\eqref{eq:iFBM-kernel-explicit} for the iFBM kernel.
With $H$ the Hurst parameter of the original FBM, almost all samples $f$ from the above processes satisfy $s[f] = 1$ and $\alpha[f] = H$ (iFBM) or $s[f] = 2$ and $\alpha[f] = H$ (iiFBM).

\item
We also consider a piecewise infinitely differentiable function $f(x) = \sin 10x + [x>x_0]$, where $x_0$ is randomly sampled from the uniform distribution on $[0,1]$ and $[x > x_0]$ is $1$ if $x > x_0$ and $0$ otherwise. This function is of finite quadratic variation with $V^2(f) = 1$.
\end{itemize}

Denote $\hat{\tau}^2 = \lim_{N \to \infty} \tausqcvest$. 
 For the above test functions, with equally-spaced partitions, we expect the following asymptotic behaviour for the CV estimator from
\Cref{res:holder-spaces,res:fqv-estimator,res:holder-spaces-exp}, \Cref{prop:bm-almost-sure}, the definition of quadratic variation, and~\eqref{eq:OU-SDE},
\begin{align*}
  \text{BM ($s[f]=0$, $\alpha[f]=1/2$):} &\quad\quad \tausqcvest = \bigo(1) \quad &&\hspace{-1cm}\text{ and } \quad \hat{\tau}^2 = 1, \\
  \text{OU ($s[f]=0$, $\alpha[f]=1/2$):} &\quad\quad \tausqcvest = \bigo(1) \quad &&\hspace{-1cm}\text{ and } \quad \hat{\tau}^2 = \lambda/2, \\
  \text{FBM  ($s[f]=0$, $\alpha[f]=H$):} &\quad\quad \tausqcvest = \bigo(N^{1 - 2H}) \quad &&\hspace{-1cm}\text{ and } \quad \hat{\tau}^2 = 0, \\
  \text{iFBM  ($s[f]=1$, $\alpha[f]=H$):} &\quad\quad \tausqcvest = \bigo(N^{-1 - 2H}) \quad &&\hspace{-1cm}\text{ and } \quad \hat{\tau}^2 = 0, \\
  \text{iiFBM  ($s[f]=2$, $\alpha[f]=H$):} &\quad\quad \tausqcvest = \bigo(N^{-2}) \quad &&\hspace{-1cm}\text{ and } \quad \hat{\tau}^2 = 0, \\
  \sin 10x + [x > x_0]: &\quad\quad \tausqcvest = \bigo(1) \quad &&\hspace{-1cm}\text{ and } \quad \hat{\tau}^2 = 1.
\end{align*}
Note that the above rate for the iFBM holds for $0 < H \leq 1/2$.
The chosen functions allow us to cover a range of $\alpha[f]$ and $s[f]$ relevant to the varying rate of convergence in~\Cref{res:holder-spaces,res:holder-spaces-exp}, as well as a range of $V^2(f)$ relevant to the limit in~\Cref{res:fqv-estimator}, $\lim_{N \to \infty} \tausqcvest = V^2(f) / T$.

\tocless\subsection{Asymptotics of the CV estimator\label{sec:experiment-cv}}

\Cref{fig:cv} shows the asymptotics of $\tausqcvest$, where each row corresponds to one stochastic process generating test functions $f$; the rows are displayed in increasing order of smoothness as quantified by $s[f] + \alpha[f]$.
The estimates are obtained for equally-spaced partitions of sizes $N=10,10^2,\dots,10^5$.
In each row, the left panel plots a single sample of generated test functions $f$. The middle panel shows the mean and credible intervals (of two standard deviations) of $\tausqcvest$ for 100 samples of $f$ for each sample size $N$. The right panel describes the convergence rate of $\tausqcvest$ to its limit point $\hat{\tau}^2 = \lim_{N \to \infty} \tausqcvest$ on the log scale.

We have the following observations.
\begin{itemize}
    \item The first two rows (the BM and OU) and the last (the piecewise infinitely differentiable function) confirm \Cref{res:fqv-estimator}, which states the convergence $\tausqcvest \to V^2(f) / T$ as $N \to \infty$. While \Cref{res:fqv-estimator} does not provide convergence rates, the rates in the first two rows appear to be $N^{-1/2}$. In the last row the rate is $N^{-2}$.
    \item The remaining rows show that the observed rates of $\tausqcvest$ to $0$ are in complete agreement with the rates predicted by \Cref{res:holder-spaces,res:holder-spaces-exp}.
 In particular, the rates are adaptive to the smoothness $s[f] + \alpha [f]$ of the function if $s[f] + \alpha[f] \leq 3/2$, as predicted.
\end{itemize}

\tocless\subsection{Comparison of CV and ML estimators\label{sec:comparison-CV-ML}}

\begin{figure}
    \centering
    \includegraphics[width=\textwidth]{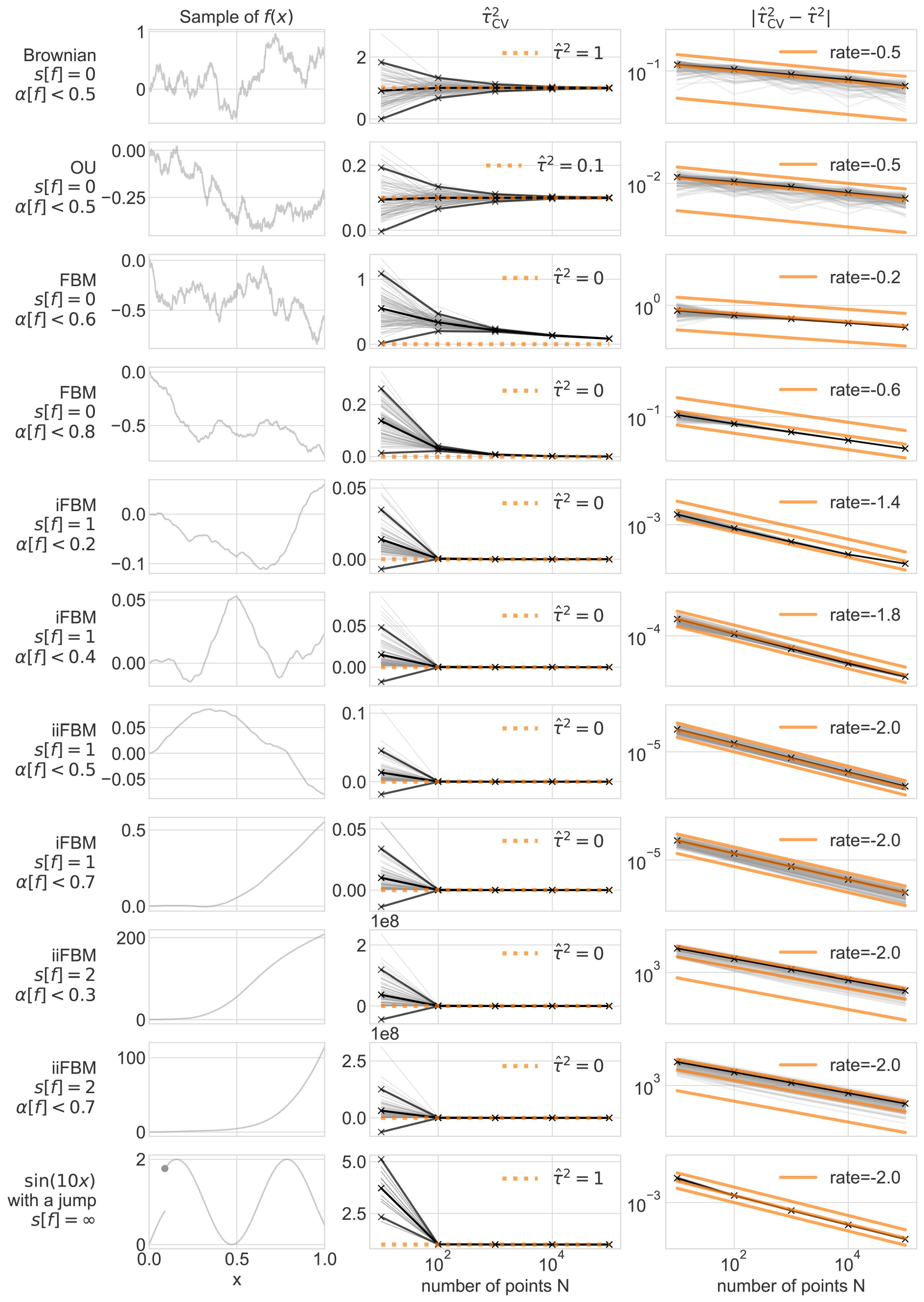}
    \caption[Asymptotics of CV estimators for functions of varying smoothness.]{Asymptotics of CV estimators for functions of varying smoothness as quantified by $s[f]$ and $\alpha[f]$ in~\eqref{eq:smoothness-1037}. Runs on individual 100 samples from $f$ are in grey, means and confidence intervals (of two standard deviations) are in black.}
    \label{fig:cv}
\end{figure}

\begin{figure}
    \centering
    \includegraphics[width=\textwidth]{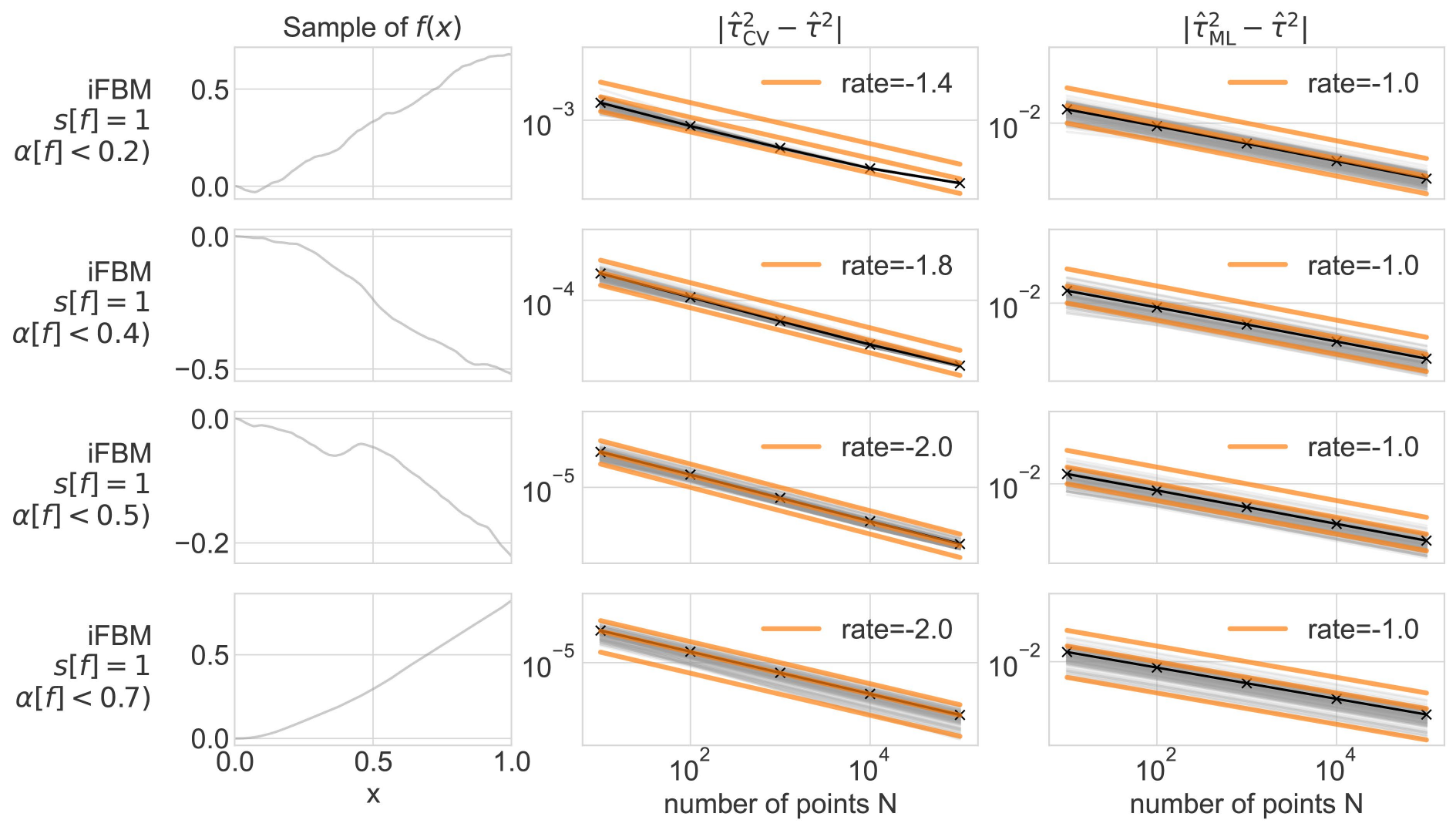}
    \caption[Asymptotics of CV estimator compared to asymptotics of ML estimators.]{Asymptotics of CV estimator compared to asymptotics of ML estimators, for once differentiable functions.}
    \label{fig:cv-vs-ml}
\end{figure}

\Cref{fig:cv-vs-ml} shows the decay rates of $\tausqcvest$ and $\tausqmlest$ to $0$ for test functions $f$ with $s[f] = 1$, under the same setting as for \Cref{fig:cv}. In this case, \Cref{res:holder-spaces-ml,res:holder-spaces-exp-ml} predict that  $\tausqmlest$ decays at the rate $\Theta(N^{-1})$  regardless of the smoothness; this is confirmed in the right column. In contrast, the middle column shows again that $\tausqcvest$ decays with a rate that adapts to  $s[f]$ and $\alpha[f]$ as long as $s[f] + \alpha[f] \leq 3/2$, as predicted by \Cref{res:holder-spaces,res:holder-spaces-exp}. These results empirically support our theoretical finding that the CV estimator is adaptive to the unknown smoothness $s[f] + \alpha[f]$ of a function $f$ for a broader range of smoothness than the ML estimator.

Additionally, in~\Cref{sec:cv-vs-ml-sobolev}, we compare the asymptotics of the CV and ML estimators when the underlying kernel is a Mat\'ern kernel, and the Sobolev smoothness of the true functions differs from that of the kernel. Similarly to the results presented in this section, we observe that the CV estimator exhibits a larger range of adaptation than the ML estimator.


\section{Conclusion}
\label{sec:conclusion_gpcv}

We have analysed the asymptotics of the CV and ML estimators for the kernel amplitude parameter in Bayesian quadrature with the Brownian motion kernel. As a novel contribution, our analysis covers the misspecified case where the smoothness of the integrand $f$ is different from that of the samples from the GP prior. Our main results in \Cref{res:holder-spaces,res:holder-spaces-ml,res:holder-spaces-exp,res:holder-spaces-exp-ml} indicate that both CV and ML estimators can adapt to the unknown smoothness of $f$, but the range of smoothness for which this adaptation happens is broader for the CV estimator. Accordingly, the CV estimator can make BQ uncertainty estimates asymptotically well-calibrated for a wider range of smoothness than the ML estimator, as indicated in \Cref{res:uq-theorem-exp-bq}. In this sense, the CV estimator has an advantage over the ML estimator. The experiments provide supporting evidence for the theoretical results.

The natural next steps are to (1) supplement the asymptotic upper bounds in \Cref{res:holder-spaces,res:holder-spaces-ml} of the deterministic setting with matching lower bounds; and (2) extend the analyses of both the deterministic and random settings to more generic finitely smooth kernels, higher dimensions, and a noisy setting.

The matching lower bounds, if obtained, would enable the analysis of the ratio between the prediction error $|I - I_\BQ |$ and the posterior standard deviation $\hat{\tau} \sigma_\BQ$ in the deterministic setting, corresponding to the one in \Cref{sec:discussion} for the random setting.  Such an analysis would need additional assumptions on the integrand $f$, such as the homogeneity of the smoothness of $f$ across the input space.  It also requires a sharp characterisation of the error $| I - I_\BQ  |$, which could use super convergence results in \citet[Section~11.5]{Wendland2005} and \citet{Schaback2018}.
Most natural kernel classes for extension are Mat\'erns and other kernels whose RKHS are norm-equivalent to Sobolev spaces; we conduct initial empirical analysis in~\Cref{sec:cv-vs-ml-sobolev} and observe results consistent with the main results in this chapter. To this end, it would be possible to adapt the techniques used in \citet{Karvonen2020} for analysing the ML estimator to the CV estimator. In any case, much more advanced techniques
than those used here would be needed. A potentially more straightforward extension could be one to multiple times integrated Brownian motion kernels for which Gaussian process interpolation corresponds to spline interpolation~\citep[Chapter~1]{Wahba1990}.
In particular, finding an analytic expression for the mean and variance of a cubic spline kernel given in, for example, Equation~(6.28) of \citet{rassmussen2006gaussian} can be reduced to the problem of inverting a tridiagonal matrix targeted in~\citet{mallik2001inverse} and~\citet{kilicc2008explicit}.

\part{Kernel-Based Distances Beyond the MMD}
\label{sec:part2}
\chapter{Kernel Quantile Embeddings}
\label{sec:kqe}
\begin{tcolorbox}
The results in this chapter were published in the following paper:

\begin{itemize}
    \item Naslidnyk, M., Chau, S. L., Briol, F.-X., \& Muandet, K. (2025). Kernel Quantile Embeddings and Associated Probability Metrics. International Conference on Machine Learning.
\end{itemize}

All theoretical results were obtained by me. The code for experiments and the experimental framework were implemented by me, with contributions from Dr Siu Lun Chau, who executed half of the benchmarking runs.
\end{tcolorbox}

In this chapter, we address the second challenge in~\Cref{sec:challenges_and_contributions}: investigating alternative kernel-based discrepancies. As covered in~\Cref{sec:kernel_mean_embeddings_and_mmd}, maximum mean discrepancy relies on kernel mean embeddings to represent distributions as mean functions in an RKHS. However, the question of whether alternative kernel-based embeddings, particularly nonlinear counterparts, could exhibit desirable properties has long remained underexplored, in part due to the associated computational challenges.  Recently, this gap has begun to be addressed, with works investigating kernelised medians~\citep{nienkotter2022kernel}, cumulants~\citep{Bonnier2023}, and variances~\citep{makigusa2024two}. Inspired by generalised quantiles, we introduce alternative embeddings based on the concept of quantiles in an RKHS, which we term \emph{kernel quantile embeddings (KQEs)}. Similar to the construction of KMEs, KQEs are obtained by considering the directional quantiles of a feature map obtained from a reproducing kernel. KQEs also lead naturally to a family of distances which we call \emph{kernel quantile discrepancies (KQDs)}. This approach is motivated by the statistics and econometrics literature~\citep{Kosorok1999,dominicy2013method,ranger2020minimum,stolfi2022sparse}, where matching quantiles has been shown to be effective in constructing statistical estimators and hypothesis tests.

We identify several desirable properties of KQEs.
Firstly, from a theoretical point of view, we show in \Cref{res:cramer-wold} and \Cref{res:if_meanchar_then_quantchar} that KQEs can represent distributions on any space for which we can define a kernel, and that the conditions to make a kernel \emph{quantile-characteristic}, that is, for KQEs to be a one-to-one representation of a probability distribution, are weaker than for the classical notion of characteristic, which we now call \emph{mean-characteristic}. We then show in \Cref{res:consistency_KQE} that KQEs can be estimated at a rate of $\bigo(N^{-\nicefrac{1}{2}})$ in the number of samples $N$; the same rate as that of the empirical estimator of KMEs~\citep{tolstikhin2017minimax}. As a result, KQDs are probability metrics under much weaker conditions than the MMD (see \Cref{res:KQE_characterise_dists}), while maintaining comparable computational guarantees, including a finite-sample consistency with rate $\bigo(N^{-\nicefrac{1}{2}})$ (up to log terms) for their empirical estimators (see \Cref{res:consistency_KQD}).

Secondly, we establish several connections between KQDs, Wasserstein distances \citep{Kantorovich1942,villani2009optimal}, and generalisations or approximations thereof. In particular, special cases of our KQDs recover existing sliced Wasserstein (SW) distances~\citep{bonneel2015sliced,wang_two-sample_2022,wang2025statistical} and can interpolate between the Wasserstein distance and MMD, similar to Sinkhorn divergences \citep{cuturi2013sinkhorn,Genevay2019}. These results are presented in Connections \ref{res:connections_slicedwasserstein}, \ref{res:connections_maxslicedwasserstein}, and \ref{res:connections_sinkhorn}.

Finally, we consider a specific instance of KQDs based on Gaussian averaging over kernelised quantile directions, which we name the \emph{Gaussian expected kernel quantile discrepancy (e-KQD)}. Beyond the desirable theoretical properties described above, we show that the Gaussian e-KQD also has attractive computational properties. In particular, we show that it has a natural estimator which only requires sampling from a Gaussian measure on the RKHS, and which can be computed with complexity $\bigo(N\log^2(N) )$. It is studied empirically in \Cref{sec:experiments_kqe} with experiments on two-sample hypothesis testing, where we show that it is competitive with the MMD: it often outperforms estimators of the MMD of the same asymptotic complexity, and in some cases even outperforms MMD at higher computational costs.

We begin by reviewing existing definitions of quantiles, and Wasserstein and sliced Wasserstein distances.

\section{Preliminaries: Quantiles and Wasserstein distances}
\label{sec:background_quantiles}

\textbf{Univariate quantiles.} Let $\calX \subseteq \bR$. For $\alpha \in [0,1]$, the \emph{$\alpha$-quantile of $\bP \in \calP(\calX)$} is defined as $\rho^{\alpha}_\bP = \inf \{y \in \calX : \text{Pr}_{Y\sim \bP}[Y \leq y] \geq \alpha\}$.
When $\bP$ has a continuous and strictly monotonic cumulative distribution function $F_\bP$, quantiles can also be defined through the inverse of that function $\rho^{\alpha}_\bP \coloneqq F^{-1}_\bP(\alpha)$. Notable special cases include $\alpha = 0.5$, corresponding to the median, and $\alpha = 0.25,0.75$, corresponding to lower and upper quartiles respectively. Importantly, $\bP$ is fully characterised by its quantiles $\{\rho^{\alpha}_\bP\}_{\alpha \in [0,1]}$.

From a computational viewpoint, univariate quantiles can be straightforwardly estimated using order statistics. Suppose $y_1, \dots, y_N \sim \bP$, and denote by $[y_{1:N}]_n$ the $n$-th order statistic of $y_{1:N}$ (i.e., the $n$-th smallest value in the vector $[y_1 \dots y_N]^\top$). The $\alpha$-quantile of $\bP$, denoted $\rho^\alpha_\bP$, can be estimated using $[y_{1:N}]_{\lceil \alpha N \rceil}$ where $\lceil \cdot \rceil$ denotes the ceiling function. This estimator is known to converge at a rate of $\bigo(N^{-\nicefrac{1}{2}})$~\citep[Section 2.3.2]{serfling2009approximation}.

\begin{figure}
    \centering
    \resizebox{\columnwidth}{!}{
    \includegraphics[height=1.1in]{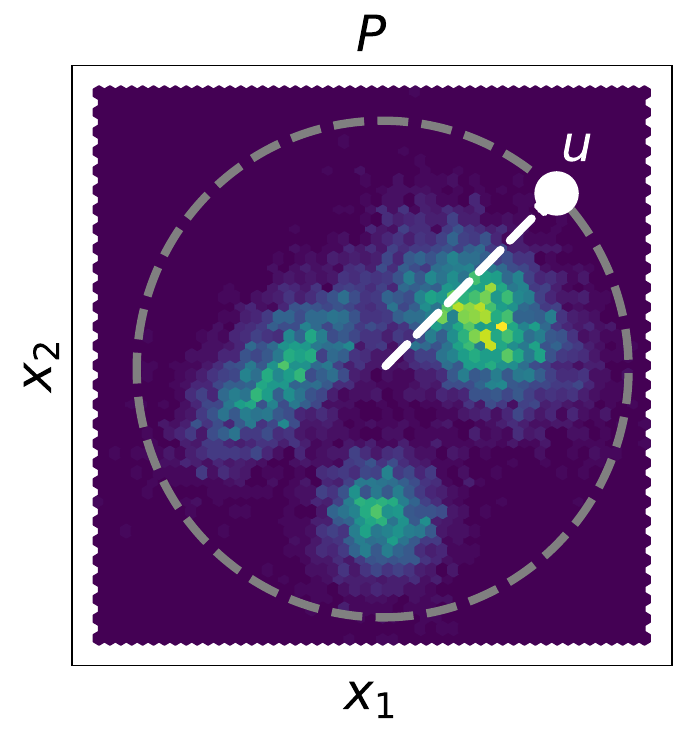}
    \includegraphics[height=1.1in]{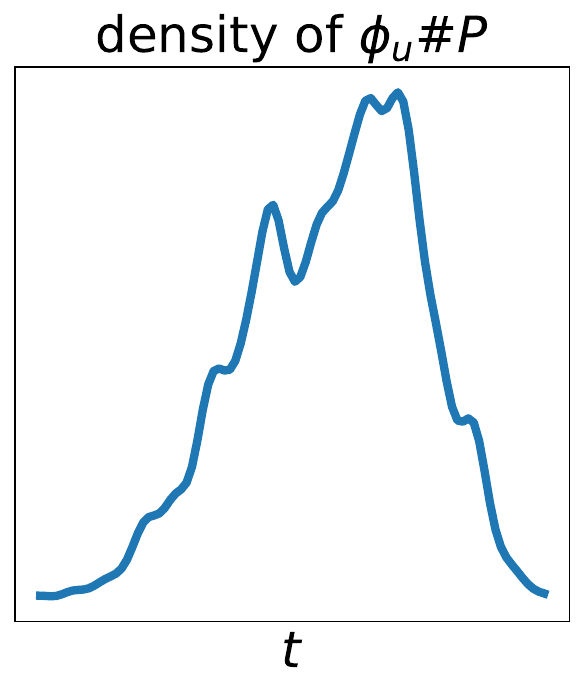}
    \includegraphics[height=1.1in]{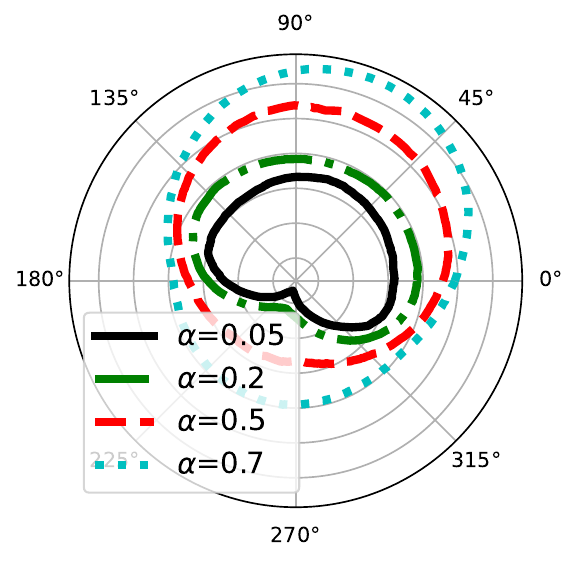}}
    \caption[Illustration of bivariate quantiles.]{\textit{Illustration of bivariate quantiles.} \textit{Left:} Bivariate distribution $\bP$. \textit{Center:} Density of the projection of $\bP$ onto direction $u$ on the unit circle, with $\phi_u(x)=\langle u, x \rangle$. \textit{Right:} Different quantiles for all possible directions $u$.}
    \label{fig:2d-quantiles}
\end{figure}

\textbf{Multivariate quantiles.} Suppose now that $\calX \subseteq \bR^d$ for $d > 1$. The previous definition of quantiles depends on the existence of an ordering in $\calX$, and its natural generalisation to $d>1$ is therefore not unique \citep{Serfling2002}. In this chapter, we will focus on the notion of \textit{$\alpha$-directional quantile of $\bP$ along some direction $u$ in the unit sphere $S^{d-1}$}~\citep{Kong2012},
\begin{equation*}
    \rho^{\alpha,u}_\bP \coloneqq \rho^{\alpha}_{\phi_u \#\bP} u,\qquad \phi_u(y)= \langle u, y \rangle.
\end{equation*}
Here, $\phi_u: \calX \to \bR$ is the projection map onto $u$, and $\rho^{\alpha}_{\phi_u\#\bP}$ is the standard one-dimensional $\alpha$-quantile of $\phi_u \# \bP$, the law of $\phi_u(X)$ for $X \sim \bP$. We note that this quantile is now a $d$-dimensional vector rather than a scalar.
The $\alpha$-directional quantiles for $d=2$ are illustrated in \Cref{fig:2d-quantiles}, in which the probability measure $\bP$ is projected onto some line; see the left and middle plots. Once again, we can use quantiles to characterise $\bP$, although we must now consider all $\alpha$-quantiles over a sufficiently rich family of projections $\{\rho_\bP^{\alpha, u} : \alpha \in [0,1], u \in S^{d-1}\}$; see Theorem 5 of \citet{Kong2012} for sufficient regularity conditions.

Although these multivariate quantiles
satisfy scale equivariance and rotation equivariance, they do not satisfy location equivariance.
To remedy this issue,~\citet{fraiman2012quantiles} introduced a related notion, the \emph{centered} $\alpha$-directional quantile,
\begin{equation}
\label{eq:direct-q}
    \tilde \rho^{\alpha, u}_\bP \coloneqq \left(\rho^\alpha_{\phi_u \# \bP} - \phi_u(\bE_{X\sim \bP}[X])\right)u + \bE_{X \sim \bP} [X],
\end{equation}
Further details are provided in \Cref{appendix:centered_quantiles}.

\paragraph{Wasserstein distances.}

Let $c: \calX \times \calX \to \bR$ be a metric on $\calX$, and $\Gamma(\bP, \bQ) \subseteq \calP(\calX \times \calX)$ denote the space of joint distributions on $\calX \times \calX$ with first and second marginals $\bP$ and $\bQ$, respectively. The \textit{$p$-Wasserstein distance}~\citep{Kantorovich1942,villani2009optimal} quantifies the cost of optimally transporting one distribution to another under `cost' $c: \calX \times \calX \rightarrow \bR$. It is a probability metric under mild conditions~\citep[Section 6]{villani2009optimal}, and is defined as
\begin{align*}
    W_p(\bP, \bQ) = \left(\inf_{\pi \in \Gamma(\bP, \bQ)} \bE_{(X,Y)\sim \pi}\left[c(X,Y)^p \right]\right)^{\nicefrac{1}{p}}.
\end{align*}
When $\calX \subseteq \bR^d$, the metric $c$ is typically taken to be the Euclidean distance $c(x,y)=\|x-y\|_2$.
The Wasserstein distance can then be estimated by solving an optimal transport problem using empirical measures constructed through samples of $\bP$ and $\bQ$, an approach that suffers from a high computational cost of $\bigo(N^3)$ and, when $\bP, \bQ$ have at least $2p$ moments, slow convergence of $\bigo(N^{-1 / \max(d, 2p)})$ when $\calX \subseteq \bR^d$ for $d>1$ \citep{Fournier2015}.

However, when $d=1$, $W_p$ can be computed at a lower cost of $\bigo(N \log N)$ with convergence of $\bigo(N^{-\nicefrac{1}{2p}})$ when $\bP, \bQ$ have at least $2p$ moments.
This motivated the introduction of the \textit{sliced Wasserstein} (SW) distance~\citep{bonneel2015sliced}. Recall that $\phi_u(x) = u^\top x$. The SW distance projects high-dimensional distributions $\bP, \bQ$ onto elements on the unit sphere $u \in S^{d-1}$ sampled uniformly, computes the Wasserstein distance between the projected distributions, now in $\bR$, and averages over the projections:
\begin{equation*}
    \text{SW}_p(\bP, \bQ) =\left( \bE_{u \sim \bU(S^{d-1})}\left[W_p^p(\phi_u\#\bP, \phi_u\#\bQ)\right] \right)^{\nicefrac{1}{p}}.
\end{equation*}
A further refinement, the \textit{max-sliced Wasserstein (max-SW)} distance~\citep{Deshpande2018}, aims to identify the optimal projection that maximises the 1D Wasserstein distance,
\begin{equation*}
    \text{max-SW}_p(\bP, \bQ) =\left( \sup_{u \in S^{d-1}} W_p^p(\phi_u\#\bP, \phi_u\#\bQ) \right)^{\nicefrac{1}{p}}.
\end{equation*}
Both slicing distances reduce the computational complexity to $\bigo(LN \log N)$ and the convergence rate to $\bigo(L^{-\nicefrac{1}{2}} + N^{-\nicefrac{1}{2p}} )$, where $L$ is either the number of projections, or the number of iterations of the optimiser. A further extension is the \textit{generalised sliced Wasserstein} (GSW,~\citet{Kolouri2019}), which replaces the linear projection $\phi_u$ with a non-linear mapping. While the conditions for GSW to be a probability metric are highly non-trivial to verify, the authors showed that they hold for polynomials of odd degree.

Another approximation of the Wasserstein distance involves the introduction of an entropic regularisation term~\citep{cuturi2013sinkhorn}, which reduces the cost to $\bigo(N^2)$ and can be estimated with sample complexity $\bigo(N^{-\nicefrac{1}{2}})$ \citep{Genevay2019}. The solution to this regularised problem, with self-cost terms subtracted, is referred to as the \emph{Sinkhorn divergence}. Interestingly,~\citet{ramdas2017wasserstein,Feydy2019} demonstrated that by varying the strength of the regularisation, the Sinkhorn divergence interpolates between the Wasserstein distance and the MMD with a kernel corresponding to the energy distance.

\section{Kernel Quantile Embeddings and Discrepancies}
\label{sec:KQE_KQD}

We introduce directional quantiles in the RKHS and the corresponding discrepancies. Unlike in~\Cref{sec:background_quantiles}, the measures and their quantiles now live in different spaces: the measures are on $\calX$, and the quantiles are in the RKHS $\calH$ induced by a kernel on $\calX$.
This leads to greater flexibility: the approach works for any space on which a kernel can be defined. Throughout, we assume the kernel $k$ is measurable.

\subsection{Kernel Quantile Embeddings}

Let $S_\calH = \{u \in \calH : \|u\|_\calH = 1\}$ be the unit sphere of an RKHS $\calH$ induced by the kernel $k$. For $\bP \in \calP(\calX)$, we define its \textit{$\alpha$-quantile along RKHS direction $u \in S_\calH$} as a function $\rho_\bP^{\alpha, u}:\calX \rightarrow \bR$ in $ \calH$ with
\begin{equation}
\label{eq:quemb_for_kernels}
    \rho_\bP^{\alpha,u}(x) \coloneqq \rho^\alpha_{u \# \bP} u(x).
\end{equation}
By the reproducing property, it holds that $\rho^\alpha_{u \# \bP} u(x) = \rho^\alpha_{\phi_u \# [\psi \# \bP]} u(x)$, where $\psi(x) = k(x, \cdot)$ is the canonical feature map $\calX \to \calH$, and $\phi_u(h) = \langle u, h\rangle_{\calH}$ is the $\calH \to \bR$ equivalent of the projection operator onto $u$ defined in~\Cref{sec:background_quantiles}. Thus, when $\dim(\calH)<\infty$, the RKHS quantiles of $\bP$ on $\calX$ are exactly the multivariate quantiles of the measure of $k(X, \cdot)$, $X \sim \bP$, on $\calH$.
In other words, KQEs can be thought of as two-step embeddings: we first embed $X \sim \bP \in \calP(\calX)$ as an RKHS element and then compute its directional quantiles to obtain the KQEs.

\textbf{Centered vs uncentered quantiles.}
Just as done for multivariate quantiles in~\eqref{eq:direct-q}, a centered version of RKHS quantiles can be defined as
\begin{equation*}
    \tilde \rho_\bP^{\alpha,u}(x) \coloneqq \left(\rho^\alpha_{u \# \bP} - \langle u, \mu_\bP \rangle_\calH \right)u(x) + \mu_\bP(x), 
\end{equation*}
where $\mu_\bP$ is the KME of $\bP$. This coincides with~\eqref{eq:direct-q} for the measure being
the law of $k(X, \cdot)$ with $X \sim \bP$. The impact of centering is examined in detail in~\Cref{appendix:centered_quantiles}, but two key observations are relevant here: (1) omitting centering eliminates the computational overhead of calculating means; (2)
the only equivariance violated for the uncentered directional quantile is location equivariance: shifting $k(X, \cdot)$ by $h$ shifts the quantile by $\langle h, u \rangle_\calH u$, rather than by $h$ itself.
However, when KQEs are used to compare two distributions, the additional term $\langle h, u \rangle_\calH u$ cancels out as it does not depend on the measure. For these reasons, we primarily work with the uncentered RKHS quantiles.

We now consider the properties of the set $\{\rho_\bP^{\alpha,u} : \alpha \in [0, 1], u \in S_\calH\}$, for a distribution $\bP$. It may be of independent interest to study kernel quantiles not as a set, but as a map $[0, 1] \times S_\calH \to \calH$; this is left for future work. 

\textbf{Quantile-characteristic kernels.} The kernel $k$ is said to be \emph{quantile-characteristic} if the mapping $\bP \mapsto \{\rho_\bP^{\alpha,u} : \alpha \in [0, 1], u \in S_\calH\}$ is injective for $\bP \in \calP(\calX)$. In $\bR^d$, the Cram\'er-Wold theorem~\citep{cramer1936some} states that the set of all one-dimensional projections (or, equivalently, all quantiles of all one-dimensional projections) determines the measure. One may therefore recognise our next theorem as an RKHS-specific extension of the Cram\'er-Wold theorem. Earlier Hilbert space extensions required higher-dimensional projections and imposed restrictive moment assumptions~\citep{cuesta2007sharp}. Being concerned with the RKHS case specifically allows us to prove the result under mild assumptions, as stated below.
\begin{assumption}
\label{as:input_space}
    \hspace{-0.1cm}$\calX$ is Hausdorff, separable, and $\sigma$-compact.
\end{assumption}
Being Hausdorff ensures points in $\calX$ can be separated, and separability says $\calX$ has a countable dense subset. $\sigma$-compactness means $\calX$ is a union of countably many compact sets. These are mild conditions, notably satisfied by Polish spaces, including discrete topological spaces with at most countably many elements and topological manifolds. 

It is possible to drop the $\sigma$-compactness and separability. When $\calX$ is Hausdorff and completely regular, quantile-characteristic properties still hold on Radon probability measures, the "non-pathological" Borel probability measures. We discuss this in~\Cref{sec:proof_projections_determine_distribution} and refer to~\citet{Willard1970} for a review of general topological properties.
\begin{assumption}
\label{as:kernel}
    The kernel $k$ is continuous, and separating on $\calX$: for any $x \neq y \in \calX$, it holds that $k(x, \cdot) \neq k(y, \cdot)$.
\end{assumption}
This is a mild condition: most commonly used kernels, such as the Mat\'ern, Gaussian, and Laplacian kernels, are separating. The constant kernel $k(x, x')=c$ is an example of a non-separating kernel. Trivially, a non-separating kernel for which $k(x, \cdot) = k(y, \cdot)$ will not distinguish between Dirac measures $\delta_x$ and $\delta_y$.

The proof of the following result uses~\emph{characteristic functionals}, an extension of characteristic functions to measures on spaces beyond $\bR^d$. Unlike moments, these are defined for any probability measure, which is the key to the generality of KQEs. Further discussion and proof are in~\Cref{sec:proof_projections_determine_distribution}.
\begin{theorem}[\textbf{Cram\'er-Wold Theorem in RKHS}]
\label{res:cramer-wold}
    Under~\Cref{as:input_space,as:kernel}, the kernel $k$ is quantile-characteristic, i.e., the mapping $\bP \mapsto \{\rho_\bP^{\alpha,u} : \alpha \in [0, 1], u \in S_\calH\}$ is injective.
\end{theorem}

The mildness of the assumptions in \Cref{res:cramer-wold} naturally raises the question: \emph{is being quantile-characteristic a less restrictive condition than being mean-characteristic?}
This indeed holds, as shown in the result below.
\begin{theorem}
\label{res:if_meanchar_then_quantchar}
    Every mean-characteristic kernel $k$ is also quantile-characteristic. The converse does not hold.
\end{theorem}
This result, proven in \Cref{appendix:proof_if_meanchar_then_quantchar}, has a powerful implication.
For any discrepancy $D(\bP, \bQ)$ that aggregates the KQEs injectively (i.e., $D(\bP, \bQ)=0 \iff \rho_\bP^{\alpha,u}=\rho_\bQ^{\alpha,u}$ for all $\alpha, u$), it holds that $\MMD(\bP, \bQ) > 0 \Rightarrow D(\bP, \bQ) > 0$, but $D(\bP, \bQ) > 0 \not \Rightarrow \MMD(\bP, \bQ) > 0$. This means $D$ can tell apart every pair of measures MMD can, and sometimes more (see the proof for examples). This is intuitive: MMD is an injective aggregation of means ($\MMD(\bP, \bQ)=0 \iff \bE_\bP[u]=\bE_\bQ[u]$ for all $u$), and the set of all quantiles captures all the information in the mean, but not vice versa.
Before introducing a specific family of quantile discrepancies, we discuss sample versions of KQEs.
\begin{figure}
    \centering
    \resizebox{\columnwidth}{!}{
    \includegraphics[height=1.1in]{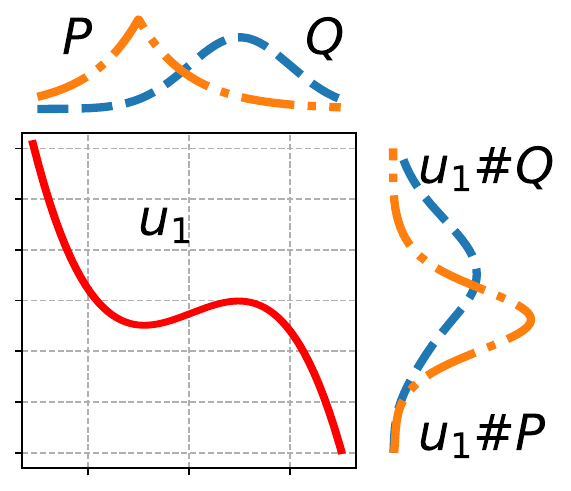}
    \includegraphics[height=1.1in]{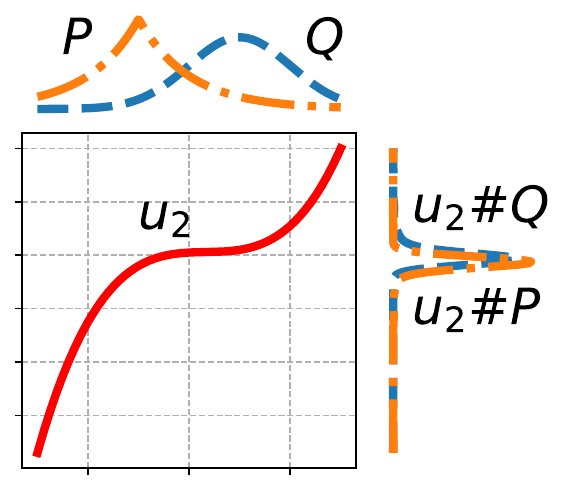}
    \includegraphics[height=1.1in]{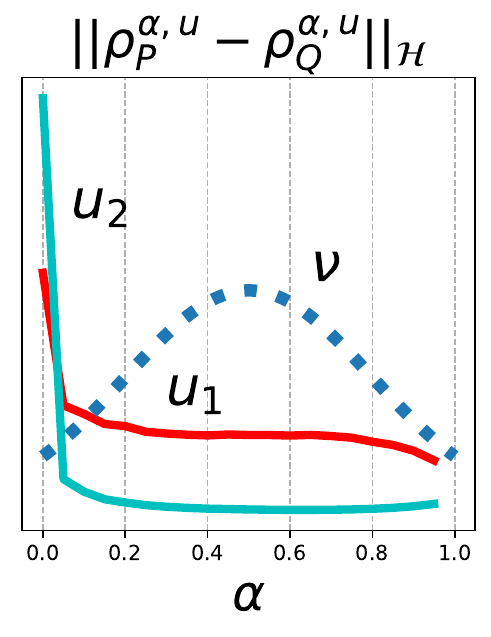}}
    \caption[Illustration of the impact of the slicing direction on KQEs.]{\textit{Illustration of the impact of the slicing direction on KQEs.} Suppose $X \sim \bP$, the KQEs $\rho_\bP^{\alpha, u}(x) \coloneqq \rho_{u \#\bP}^{\alpha} u(x)$ are obtained by considering the $\alpha$-th quantile of $u(X)$. Clearly, these quantiles might vary significantly depending on the slicing direction used.
    }
    \label{fig:kernelised-quantiles}
\end{figure}

\textbf{Estimating KQEs.}
For fixed $\alpha \in [0, 1]$ and $u \in S_\calH$, estimating the directional quantile $\rho_\bP^{\alpha,u}$ with samples $x_1, \dots, x_N \sim \bP$ boils down to estimating the $\bR$-quantile $\rho_{u \# \bP}^{\alpha}$ using samples $u(x_{1:N})$. We employ the classic, model-free approach to estimate a quantile by using the order statistic estimator,
\begin{equation}
\label{eq:dq_estimator}
    \rho^{\alpha,u}_{\bP_N}(x) \coloneqq \rho_{u \# \bP_N}^{\alpha}u(x) = [ u(x_{1:N}) ]_{\lceil \alpha N \rceil}u(x),
\end{equation}
where $\bP_N = \nicefrac{1}{N}\sum_{n=1}^N \delta_{x_n}$. In other words,~\eqref{eq:dq_estimator} uses the $\alpha$-quantile of the set $u(x_{1:N})$, i.e., the $\lceil \alpha N \rceil$-th smallest element of $u(x_{1:N})$. We now state an RKHS version of a classic result on convergence of quantile estimators; the proof is provided in~\Cref{appendix:proof_consistency_KQE}.

\begin{theorem}[\textbf{Finite-Sample Consistency for Empirical KQEs}]\label{res:consistency_KQE}
    Suppose the PDF of $u\#\bP$ is bounded away from zero, $f_{u\#\bP}(x) \geq c_u > 0$, and $x_1, \dots, x_N \sim \bP$. Then, with probability at least $1-\delta$, and $C(\delta, u) = \bigo(\sqrt {\log(2/\delta)})$,
    \begin{equation*}
        \| \rho^{\alpha, u}_{\bP_N} - \rho^{\alpha, u}_\bP \|_\calH \leq C(\delta, u) N^{-\nicefrac{1}{2}}.
    \end{equation*}
\end{theorem}
We do not need to assume~\Cref{as:input_space,as:kernel} to prove consistency; this was only needed to establish that $k$ is quantile-characteristic, and we may still have a consistent estimator when the kernel is not quantile-characteristic. The condition $f_{u\#\bP}(x) \geq c_u > 0$ lets us avoid making any assumptions on $\calX$, other than the existence of a kernel $k$ on $\calX$.

\subsection{Kernel Quantile Discrepancies}

We quantify the difference between $\bP, \bQ \in \calP(\calX)$ in unit-norm direction $u$ as a $\nu$-weighted expectation of power-$p$ distance (in the RKHS) between KQEs,
\begin{align*}
    \tau_p(\bP, \bQ;\nu, u) = \Big(\int_0^1 \big\| \rho_\bP^{\alpha,u} - \rho_\bQ^{\alpha,u} \big\|_\calH^p \nu(\d \alpha) \Big)^{\nicefrac{1}{p}}.
\end{align*}
\Cref{fig:kernelised-quantiles} illustrates how $u\#\bP$ and $u\#\bQ$ vary depending on direction $u$, and the impact it has on $\tau_p$. The weighting measure $\nu$ on $[0, 1]$ assigns importance to each $\alpha$-quantile. For example, the Lebesgue measure $\nu \equiv \mu$ treats all quantiles as equally important, whereas a partially-supported measure would allow us to ignore certain quantiles.

Based on $\tau_p(\bP, \bQ; \nu, u)$, we introduce a novel family of \textit{Kernel Quantile Discrepancies (KQDs)} that aggregate the directional differences $\tau_p(\bP, \bQ; \nu, u)$ over $u \in S_\calH$: the $\calL^p$-type distance \textit{expected KQD ($\ekqd$)} that uses the average as the aggregate function, and the $\calL^\infty$-type distance \textit{supremum KQD ($\supkqd$)} that aggregates with the supremum:
\begin{equation}
\label{eq:general_distances}
\begin{split}
    \ekqd_p(\bP, \bQ; \nu, \gamma) &= \left(\bE_{u \sim \gamma} \left[\tau^p_p\left(\bP, \bQ; \nu, u\right)\right]\right)^{\nicefrac{1}{p}},\\
    \supkqd_p(\bP, \bQ; \nu) &= \big(\sup_{u \in S_\calH} \tau^p_p\left(\bP, \bQ; \nu,u \right)\big)^{\nicefrac{1}{p}},
\end{split}
\end{equation}
where $\gamma$ is a measure on the unit sphere $S_\calH$ of the RKHS.

Next, we demonstrate that under mild conditions $\ekqd$ and $\supkqd$ are indeed distances, and establish connections with existing methods. 
\begin{theorem}[\textbf{KQDs as Probability Metrics}]\label{res:KQE_characterise_dists}
    Under~\Cref{as:input_space},~\Cref{as:kernel}, and if $\nu$ has full support on $[0, 1]$, $\supkqd_p$ is a distance. Further, if $\gamma$ has full support on $S_\calH$, $\ekqd_p$ is a distance.
\end{theorem}
The proof is in \Cref{appendix:proof_quantile_characteristic}. As discussed in \Cref{sec:KQE_KQD}, \Cref{as:input_space,as:kernel} are minor. The assumptions on the support of $\nu$ and $\gamma$ ensure that no quantile level in $[0,1]$ and no parts of $S_\calH$ are missed entirely. This is satisfied, for example, for the uniform $\nu$ (that considers all quantiles to be equally important), and when $\calH$ is separable, for any centered Gaussian $\gamma=\calN(0, S)$ with a non-degenerate $S$ by~\citep[Corollary 5.3]{kukush2020gaussian}. For example, an $\calH \mapsto \calH$ covariance operator $S[f](x) = \int_\calX k(x, y) f(y) \beta(\d y)$ is non-degenerate and well-defined provided (1) $\beta$ on $\calX$ has full support, and (2) $\int_\calX \sqrt {k(x, x)} \beta(\d x) < \infty$. This choice of $\gamma$ also happens to be computationally convenient, as discussed in~\Cref{sec:estimator}.

In contrast, while conditions under which MMD is a distance are well-understood for continuous bounded translation-invariant kernels on Euclidean spaces~\citep{sriperumbudur2011universality}, they are challenging to establish beyond this setting. For instance, it is known that commonly used graph kernels are not characteristic~\citep{kriege2020survey}.

When $\nu$ is chosen as the Lebesgue measure $\mu$, an important connection emerges between $\ekqd$, $\supkqd$, and sliced Wasserstein distances. This connection is formalised in the next result, with a proof provided in \Cref{appendix:proof_connections_slicedwasserstein}.

\begin{connection}[\textbf{SW}]
\label{res:connections_slicedwasserstein}
Suppose $\bP, \bQ$ have $p$-finite moments. Then, $\ekqd_p(\bP, \bQ; \nu, \gamma)$ for $\nu \equiv \mu$ corresponds to a kernel expected sliced $p$-Wasserstein distance, which has not been introduced in the literature. For $\calX \subseteq \bR^d$, linear $k(x, y)=x^\top y$, and uniform $\gamma$, this recovers the expected sliced $p$-Wasserstein distance~\citep{bonneel2015sliced}.
\end{connection}
\begin{connection}[\textbf{Max-SW}]
\label{res:connections_maxslicedwasserstein}
Suppose $\bP, \bQ$ have $p$-finite moments. Then, $\supkqd_p(\bP, \bQ; \nu)$ for $\nu \equiv \mu$ is the kernel max-sliced $p$-Wasserstein distance~\citep{wang_two-sample_2022}. For $\calX \subseteq \bR^d$, linear $k(x, y)=x^\top y$, and uniform $\gamma$, it recovers the max-sliced $p$-Wasserstein~\citep{Deshpande2018}.
\end{connection}

For $d=1$, we recover the standard Wasserstein. When $k$ is non-linear but induces a finite-dimensional RKHS, $\ekqd$ is connected to the generalised sliced Wasserstein distances of~\citet{kolouri_generalized_2022}; we explore this in~\Cref{appendix:proof_connections_slicedwasserstein}.
Lastly, we establish a connection to Sinkhorn divergence.

\begin{connection}[\textbf{Sinkhorn}]
\label{res:connections_sinkhorn}
    Sinkhorn divergence~\citep{cuturi2013sinkhorn}, like $\ekqd$ and $\supkqd$, combines the strengths of kernel embeddings and Wasserstein distances. Furthermore, for $p=2$ and $\nu \equiv \mu$, the centered version of $\ekqd$ and $\supkqd$ developed in~\Cref{appendix:centered_quantiles} can be represented as a sum of MMD and kernelised expected or max-sliced Wasserstein distances, thus positioning these measures as mid-point interpolants between MMD and SW distances.
\end{connection}
It is important to note that the MMD term within the Sinkhorn divergence is restricted to a specific kernel tied to the energy distance; in contrast, $\ekqd$ and $\supkqd$ offer much greater flexibility in the choice of kernel. Moreover, as will be shown empirically in~\Cref{sec:experiments_kqe}, the computational complexity of $\ekqd$ for a particular choice of $\gamma$ can be made significantly lower than that of Sinkhorn divergences, which have a cost of $\bigo(N^2)$.

\paragraph{Estimating $\ekqd$.} We propose a Monte Carlo estimator for $\ekqd$, and refer to~\citet{wang_two-sample_2022} for an optimisation-based, $\bigo(N^3 \log(N))$ estimator for $\supkqd$.
Let $x_1, \dots, x_N \sim \bP$, $y_1, \dots, y_N \sim \bQ$ and let $u_1, \dots, u_L \in S_\calH$ to be $L$ unit-norm functions sampled from $\gamma$, and $f_\nu$ to be the density of $\nu$. Denote $\bP_N = \nicefrac{1}{N}\sum_{n=1}^N \delta_{x_n}, \bQ_N = \nicefrac{1}{N} \sum_{n=1}^N \delta_{y_n}$. Then, similar to the order statistic estimator of the quantiles in~\eqref{eq:dq_estimator}, $\ekqd^p_p(\bP_N, \bQ_N; \nu, \gamma_L)$ is the estimator of $\ekqd^p_p(\bP, \bQ; \nu, \gamma)$, where
\begin{align}
\label{eq:estimator_ekqd}
    &\ekqd^p_p(\bP_N, \bQ_N; \nu, \gamma_L) = \frac{1}{LN} \sum_{l=1}^L \sum_{n=1}^N \left|\big[u_l(x_{1:N})\big]_n - \big[u_l(y_{1:N})\big]_n \right|^p f_\nu\left( \frac{n}{N} \right) 
\end{align}
Here, $[u_l(x_{1:N})]_n$ is the $n$-th order statistic, i.e., the $n$-th smallest element of $u_l(x_{1:N})=[u_l(x_1), \dots, u_l(x_N)]^\top$. For $p=1$, we get the following result, proven in~\Cref{appendix:proof_consistency_KQD}.
\begin{theorem}[\textbf{Finite-Sample Consistency for Empirical KQDs}]
\label{res:consistency_KQD}
    Let $\nu$ have a density, $\bP, \bQ$ be measures on $\calX$ such that
    $\bE_{X \sim \bP} \sqrt{k(X, X)}<\infty$ and $\bE_{X \sim \bQ} \sqrt{k(X, X)}<\infty$, and $x_1, \dots, x_N \sim \bP, y_1, \dots, y_N \sim \bQ$. Then, with probability at least $1-\delta$, and $C(\delta) = \bigo(\sqrt{\log(1/\delta)})$ that depends only on $\delta, k, \nu$,
    \begin{align*}
        | \ekqd_1(\bP_N, \bQ_N;\nu, \gamma_L) - \ekqd_1(\bP, \bQ;\nu, \gamma) |  \leq C(\delta)(L^{-\nicefrac{1}{2}} + N^{-\nicefrac{1}{2}}).
    \end{align*}
\end{theorem}
The rate does not depend on $\mathrm{dim}(\calX)$. This is a major advantage of projection/slicing-based discrepancies~\citep{Nadjahi2020}, which comes at the cost of dependence on the number of projections $L$. Setting $L=N/\log N$ recovers the MMD rate (up to log-terms), at matching complexity (see~\Cref{sec:estimator}). Here, we do not need $\ekqd$ to be a distance: indeed, we did not assume~\Cref{as:input_space,as:kernel}.
The condition of square root integrability of $k(X, X)$ under $\bP, \bQ$ is immediately satisfied when $k$ is bounded, and can in fact be further weakened to $\bE_{X \sim \bP} \bE_{Y \sim \bQ} \sqrt{k(X, X) - 2 k(X, Y) + k(Y, Y)} < \infty$.
Requiring that $\nu$ has a density is mild and necessary to reduce the problem to CDF convergence, which, by the classic Dvoretzky-Kiefer-Wolfowitz inequality of~\citet{dvoretzky1956asymptotic}, has rate $N^{-1/2}$ under no assumptions on the underlying distributions. The strength of this inequality allows us to assume nothing more of $\calX$ than the fact that it is possible to define a kernel on it.

Further, for any integer $p>1$, the $N^{-1/2}$ rate still holds, if and only if it holds that for $J_p(R) \coloneqq \left(F_{u\#R}(t) (1 - F_{u\#R}(t))\right)^{p/2} / f^{p-1}_{u\#R}(t)$, both $J_p(\bP)$ and $J_p(\bQ)$ are integrable over $u \sim \gamma$ and Lebesgue measure on $u(\calX)$.
In turn, this may be reduced to a problem of controlling $d-1$ volumes of level sets of $u$. We discuss this extension further in~\Cref{res:consistency_KQD_p} in~\Cref{appendix:proof_consistency_KQD}.

\section{Gaussian Kernel Quantile Discrepancy}\label{sec:estimator}

\begin{algorithm}[tb]
    \caption{Gaussian $\ekqd$}
    \label{alg:gaussian_ekqd}
\begin{algorithmic}
    \STATE {\bfseries Input:} Data $x_1, \dots, x_N \sim \bP, y_1, \dots, y_N \sim \bQ$, samples from the reference measure $z_1, \dots, z_M \sim \xi$, kernel $k$, density $f_\nu$, number of projections $L$, power $p$.
    \STATE Initialise $\ekqd^p \leftarrow 0$ and $\tau_{p,l}^p \leftarrow 0$ for $l \in \{1, \dots, L\}$.
    \FOR{$l=1$ {\bfseries to} $L$}
    \STATE Sample $\lambda_1, \dots, \lambda_M \sim \calN(0, \Id_M)$
    \STATE Compute $f_l(x_{1:N}) \leftarrow \lambda_{1:M}^\top k(z_{1:M}, x_{1:N})/\sqrt{M},$
    \STATE \hspace{1.3cm} $f_l(y_{1:N}) \leftarrow \lambda_{1:M}^\top k(z_{1:M}, y_{1:N}) /\sqrt{M}$
    \STATE Compute $\|f_l\|_\calH \leftarrow \sqrt{\lambda_{1:M}^\top k(z_{1:M}, z_{1:M}) \lambda_{1:M}/M}$
    \STATE Compute $u_l(x_{1:N}) \leftarrow f_l(x_{1:N})/\|f_l\|_\calH$, \\ \hspace{1.3cm} $u_l(y_{1:N}) \leftarrow f_l(y_{1:N})/\|f_l\|_\calH$
    \STATE Sort $u_l(x_{1:N})$ and $u_l(y_{1:N})$
    \FOR{$n=1$ {\bfseries to} $N$}
    \STATE $\tau_{p,l}^p \gets \tau_{p, l}^p + \big|[u_l(x_{1:N})]_n - [u_l(y_{1:N})]_n\big|^p f_\nu( n/N )$
    \ENDFOR
    \STATE $\ekqd^p \gets \ekqd^p + \tau_{p,l}^p/L$
    \ENDFOR
    \STATE Return $\ekqd^p$
\end{algorithmic}
\end{algorithm}

We now conduct a further empirical study of the squared kernel distance $\ekqd^2_p$. Unlike its supremum-based counterpart $\supkqd$, $\ekqd$ can be approximated simply by drawing samples from $\gamma$ on $S_\calH$, avoiding the challenges associated with optimising for the supremum. Although a uniform $\gamma$ is a natural choice, no such measure exists when $\dim(\calH)$ is infinite~\citep[Section 1.3]{kukush2020gaussian}. Instead, we follow a well-established strategy from the inverse problems literature~\citep{stuart2010inverse} and take $\gamma$ to be the projection onto $S_\calH$ of a Gaussian measure on $\calH$. Using established techniques for sampling Gaussian measures, we build an efficient estimator for $\ekqd_p(\bP, \bQ;\nu,\gamma)$. Gaussian measures on Hilbert spaces are a natural extension of the familiar Gaussian measures on $\bR^d$: a measure $\calN(0, C)$ on $\calH$
is said to be a \emph{centered Gaussian measure} with covariance operator $C:\calH \to \calH$ if, for every $f \in \calH$, the pushforward of $\calN(0, C)$ under the $\calH \to \bR$ projection map $\phi_f(\cdot)= \langle f, \cdot \rangle_\calH$ is the Gaussian measure $\calN(0, \langle C[f], f \rangle_\calH)$ on $\bR$. For further details on Gaussian measures in Hilbert spaces, we refer to \citet{kukush2020gaussian}.

Let $\gamma'$ be a centered Gaussian measure on $\calH$ whose covariance function $C: \calH \to \calH$ is an integral operator with some reference measure $\xi$ on $\calX$,
\begin{align*}
    \gamma' = \calN(0, C),\quad C[f](x) = \int_\calX k(x, y) f(y) \xi(\d y),
\end{align*}
and let $\gamma$ be the pushforward of $\gamma'$ by the projection $\calH \to S_\calH$ that maps any $f \in \calH$ to $f/\|f\|_\calH \in S_\calH$. By the change of variables formula for pushforward measures~\citep[Theorem 3.6.1]{bogachev2007measure}, it holds that
\begin{align*}
    \ekqd_p^p(\bP, \bQ; \nu, \gamma)
    &= \bE_{u \sim \gamma}\left[ \tau_p^p\left(\bP, \bQ; \nu, u \right) \right] = \bE_{f \sim \gamma'} \left[ \tau_p^p\left(\bP, \bQ; \nu, \nicefrac{f}{\|f\|_\calH}\right)\right].
\end{align*}
This equality reduces sampling from $\gamma$ to sampling from a centered Gaussian measure with an integral operator covariance function. The next proposition reduces sampling from (a finite-sample approximation of) $\gamma$ to sampling from the standard Gaussian on the real line; proof is in~\Cref{sec:proof_sampling_from_gm}.
\begin{proposition}[\textbf{Sampling from a Gaussian measure}]
\label{res:sampling_from_gm}
    Let $z_1, \dots, z_M \sim \xi$, and $\gamma'_M$ to be the estimate of $\gamma'$ based on the Monte Carlo estimate $C_M$ of the covariance operator $C$,
    \begin{align*}
        \gamma'_M = \calN(0, C_M),\quad C_M[g](x) = \frac{1}{M}\sum_{m=1}^M k(x, z_m) g(z_m).
    \end{align*}
    Let $f(x) = M^{-1/2}\sum_{m=1}^M \lambda_m k(x, z_m)$ with $\lambda_1, \dots, \lambda_M \sim \calN(0, 1)$. Then, $f \sim \gamma'_M$.
\end{proposition}
\Cref{alg:gaussian_ekqd} brings together the $\ekqd$ estimator in~\eqref{eq:estimator_ekqd}, and the procedure for sampling from the Gaussian measure in~\Cref{res:sampling_from_gm}. The $\nu$ choice is left up to the user; the uniform $\nu$ remains a default choice. We proceed to analyse the cost.
This estimator has complexity $ \bigo(L \max(NM, M^2, N \log N))$:
$\bigo(L)$ for iterating over directions $l \in \{1, \dots, L\}$;
$\bigo(NM)$ for computing $f_l(x_{1:N})$ and $f_l(y_{1:N})$;
$\bigo(M^2)$ for computing $\| f_l \|_\calH$; and
$\bigo(N \log N)$ for sorting $u_l(x_{1:N})$ and $u_l(y_{1:N})$. For $L \coloneqq \log N$ and $M \coloneqq \log N$, the complexity therefore reduces to $ \bigo(N \log^2 N)$; i.e., near-linear (up to log-terms).

\section{Experiments}
\label{sec:experiments_kqe}

\begin{figure*}[t]
    \centering
    \includegraphics[width=\linewidth]{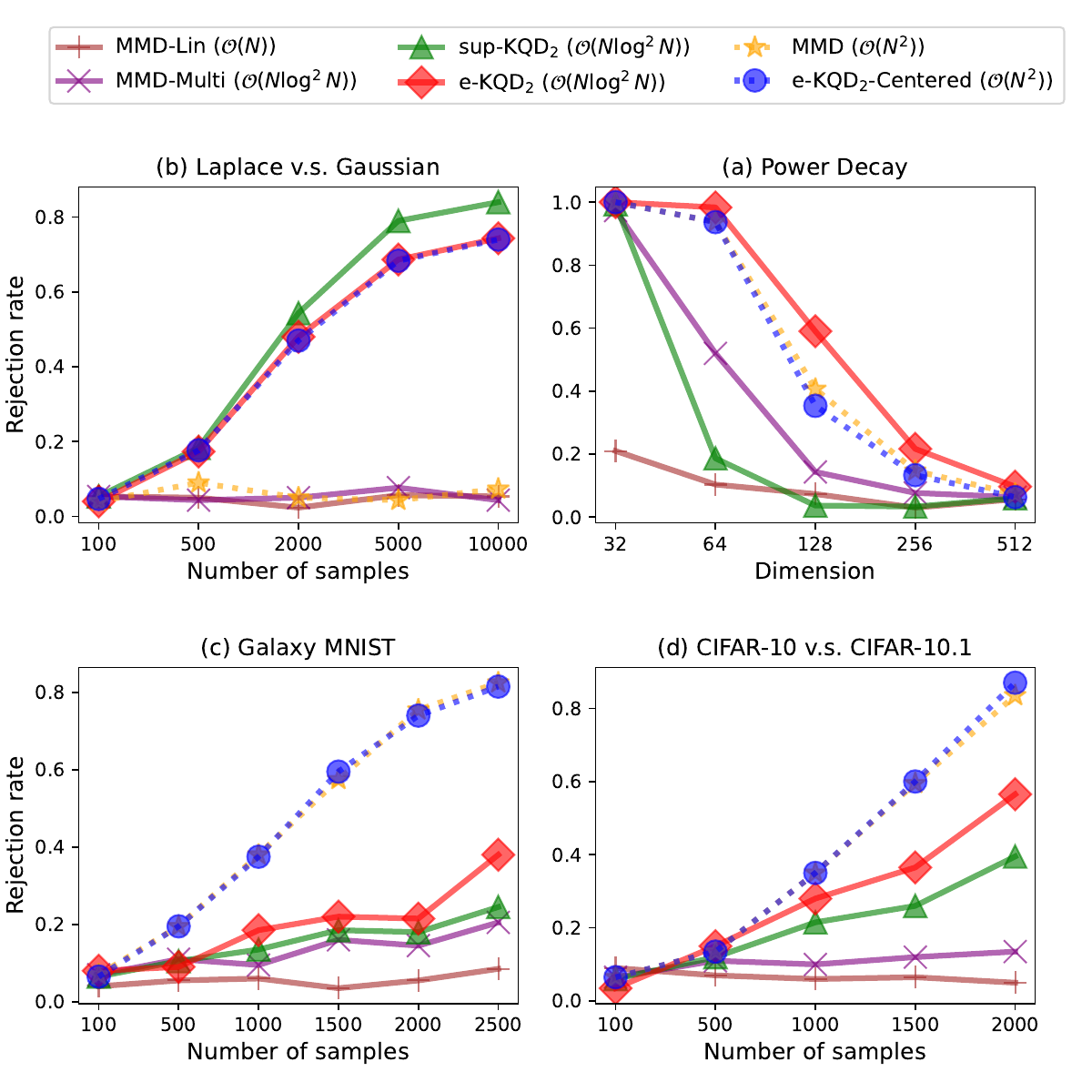}
    \caption[Experimental results comparing KQDs with baseline approaches.]{Experimental results comparing our proposed methods with baseline approaches. Methods represented by dotted lines exhibit quadratic complexity for a \textbf{single} computation of the test statistic, while the remaining methods achieve near-linear or linear computational efficiency. A higher rejection rate indicates better performance in distinguishing between distributions. \textbf{Overall, quadratic-time quantile-based estimators perform comparably to quadratic-time MMD estimators, while near-linear time quantile-based estimators often outperform their MMD-based counterparts.}}
    \label{fig:main_experiments}
\end{figure*}

We empirically demonstrate the effectiveness of KQDs for nonparametric two-sample hypothesis testing, which aims to determine whether two arbitrary probability distributions, $\bP$ and $\bQ$, differ statistically based on their respective i.i.d. samples. Two-sample testing is widely adopted in scientific discovery fields, such as model verification~\citep{gao2024model}, out-of-domain detection~\citep{magesh2023principled}, and comparing epistemic uncertainties~\citep{chau2025credal}. Specifically, we test the null hypothesis $H_0: \bP = \bQ$ against the alternative $H_1: \bP \neq \bQ$. In such tests, (estimators of) probability metrics are commonly used as test statistics, including the Kolmogorov-Smirnov distances~\citep{Kolmogorov60:Foundations}, Wasserstein distance~\citep{wang_two-sample_2022}, energy distances~\citep{SzeRiz05,sejdinovic2013equivalence}, and most relevant to our work, the MMD~\citep{gretton2006kernel,gretton2009fast, gretton2012kernel}. For an excellent overview of kernel-based two-sample testing, we refer readers to \citet{schrab2025unified}.

Experiments are repeated to calculate the rejection rate, which is the proportion of tests where the null hypothesis is rejected.
A high rejection rate indicates better performance at distinguishing between distributions.
It is equally important to ensure proper control of Type I error, defined as the rejection rate when the null hypothesis $H_0$ is true. Specifically, the Type I error rate should not exceed the specified level. Without controlling for Type I error, an inflated rejection rate might not reflect the estimator’s ability to detect genuine differences but instead indicate the test rejects more often than it should. We consider a significance level $\alpha$ of 0.05 throughout and report on Type I control in~\Cref{appendix:experiments}.

To determine the rejection threshold for each test statistic, we employ a permutation-based approach: for each trial, we pool the two sets of samples, randomly reassign labels $300$ times to simulate draws under $H_0$, compute the test statistic on each permuted split, and take the 95th percentile of this empirical null distribution as our threshold. This fully nonparametric thresholding ensures Type I error control without additional distributional assumptions~\citep{lehmann1986testing}.

Our experiments aim to demonstrate that, within a comparable computational budget, statistics computed using quantile-characteristic kernels can deliver results competitive with those of MMD tests based on mean-characteristic kernels. Additionally, we seek to explore the inherent trade-offs of the proposed methods. We focus on the nonparametric two-sample testing problem, as it represents one of the most successful applications of the mean-embedding-based MMD and its variants. The code is available at \url{https://github.com/MashaNaslidnyk/kqe}.

\tocless\subsection{Benchmarking}
We consider the following distances as test statistics in our experiments. Detailed descriptions of these estimators are provided in \Cref{appendix_sec:estimators}. For KQDs, we take the reference measure $\xi$ (cf. Proposition~\ref{res:sampling_from_gm}) to be $\nicefrac{1}{2} \bP_N + \nicefrac{1}{2} \bQ_N$, where $\bP_N$ corresponds to the empirical distribution $\nicefrac{1}{N}\sum_{n=1}^N \delta_{x_n}$, analogously for $\bQ_N$. Such $\xi$ is a general choice that is appropriate in the absence of additional information about the space $\calX$. We take power $p=2$ for all KQD-based discrepancies in our experiments; identical experiments for $p=1$ lead to the same conclusions and are presented for completeness in~\Cref{sec:exp_for_p1}. Other than in the second experiment, we use the Gaussian kernel $k(x,x') = \exp(-\|x-x'\|^2/2 l^2)$ with $l$ the lengthscale chosen using the median heuristic method, i.e., $l = \operatorname{Median}(\{\|x_n - x_{n'}\|_2 : 1 \leq n < n' \leq N)$~\citep{gretton2012kernel}. For the reader's convenience, we present all methods on the same plot, regardless of their computational complexity. However, it is important to note that directly comparing test power across methods with varying sampling complexities may be unfair and misleading.
\begin{itemize}[leftmargin=*]
\setlength\itemsep{0em}
    \item \textbf{$\ekqd$ (ours).} For $\ekqd$, we set the number of projections to $L = \log N$ and the number of samples drawn from the Gaussian reference to $M=\log N$. Consequently, the overall computational complexity is $ \bigo(N \log^2(N))$.
    \item \textbf{$\ekqd$-centered (ours).} The centered version of $\ekqd$, as discussed in~\Cref{appendix:centered_quantiles}, can be expressed as the sum of an $\ekqd$ term and the classical MMD. While the $\ekqd$ component follows the same sampling configuration as above, the MMD computation is the dominant factor in complexity, leading to an overall cost of $\bigo(N^2)$.
    \item \textbf{$\supkqd$ (ours).} $\supkqd$ adopts the same sampling configuration as $\ekqd$ (thus cost $\bigo(N\log^2(N))$). Instead of averaging over projections, it selects the maximum across all projections. This approach serves as a fast approximation of the kernel max-sliced Wasserstein distance of \citet{wang_two-sample_2022}, where a Riemannian block coordinate descent method is used to optimise an entropic regularised objective at a computational cost of $\bigo(N^3 \log(N))$. In contrast, our approach identifies the largest directional quantile difference across the sampled projections. While we do not claim that this provides an accurate estimate of the true distance, this approach allows for controlled complexity and facilitates comparisons between averaging or taking the supremum.

    \item \textbf{MMD.} The MMD is included as a benchmark to be compared with $\ekqd$-centered and has complexity $\bigo(N^2)$. The MMD is estimated using the U-statistic formulation.

    \item \textbf{MMD-Multi.} A fast MMD approximation based on incomplete U-statistic introduced in \citet{Schrab2022} is included to benchmark against our $\ekqd$ distance. Configurations of MMD-Multi are chosen to match the complexity of $\ekqd$ for a fair comparison.
    \item \textbf{MMD-Lin.} MMD-Linear from \citet[Lemma 14.]{gretton2012kernel} estimates the MMD with complexity $\bigo(N)$.
\end{itemize}

\tocless\subsection{Experimental results\label{sec:experimental_results}}

We conduct four experiments: two using synthetic data, allowing full control over the simulation environment, and two based on high-dimensional image data to showcase the practicality and competitiveness of our proposed methods. Additional experiments are reported in~\Cref{appendix:experiments}, specifically: studying the impact of changing the measures $\nu$ and $\xi$, comparing with sliced Wasserstein distances, and comparing with MMD based on other KME approximations.

\textbf{1. Power-decay experiment.} This experiment investigates the effect of the curse of dimensionality on our tests, following the setup of Experiment A in \citet{wang_two-sample_2022}. Prior work by \citet{ramdas2015decreasing} has shown that MMD-based methods are particularly vulnerable to the curse of dimensionality. Here, we assess whether our quantile-based test statistic exhibits similar limitations.

We fix $N=200$ and take $\bP$ to be an isotropic Gaussian distribution of dimension $d$. Similarly, we take $\bQ$ to be a $d$-dimensional Gaussian distribution with a diagonal covariance matrix $\Sigma = \diag(\{4, 4, 4, 1, \dots, 1\})$. As we increase the dimension $d \in \{32, 64, 128, 256, 512\}$, the testing problem becomes increasingly challenging. \Cref{fig:main_experiments}a presents the results. We observe that $\ekqd$ exhibits the slowest decline in test power among all methods, irrespective of their computational complexity. Notably, it maintains its performance significantly better than its $\bigo(N \log^2(N))$ benchmark, MMD-Multi. These results suggest that quantile-based discrepancies exhibit greater robustness to high-dimensional data.

\textbf{2. Laplace vs. Gaussian.} This experiment aims to illustrate~\Cref{res:if_meanchar_then_quantchar} by demonstrating that while a kernel may not be mean-characteristic, meaning it cannot distinguish between two distributions using standard KMEs and MMDs, it can still be quantile-characteristic. In such cases, the distributions can still be effectively distinguished using our KQEs and KQDs. To demonstrate this, we take $\bP$ to be a standard Gaussian in $d=1$, and $\bQ$ to be a Laplace distribution with matching first and second moments. We vary $N \in \{100, 500, 2000, 5000, 10000\}$ and select a polynomial kernel of degree $3$, i.e., $k(x,x') = (\langle x,x' \rangle + 1)^3$, for all our methods. This ensures that $k$ cannot distinguish between the two distributions due to their matching first and second moments, which leads to their KMEs being identical.

\Cref{fig:main_experiments}b shows that our KQDs, irrespective of their computational complexity, exhibit increasing test power as the sample size grows. In contrast, MMD-based methods fail entirely to detect any differences between $\bP$ and $\bQ$. Notably, although $\ekqd$-centered can be expressed as the sum of an MMD term and an $\ekqd$ term, the underperformance of the MMD component in this scenario is effectively compensated by the $\ekqd$ term, enabling successful testing.

\textbf{3. Galaxy MNIST.} We examine performance on real-world data through galaxy images~\citep{walmsley2022galaxy} in dimension $d=3\times64\times64=12288$, following the setting from \citet{biggs2024mmd}. These images consist of four classes. $\bP$ corresponds to images sampled uniformly from the first three classes, while $\bQ$ consists of samples from the same classes with probability $0.85$ and from the fourth class with probability $0.15$. A Gaussian kernel with lengthscale chosen using the median heuristic method is chosen for all estimators. Sample sizes are chosen from $N\in\{100, 500, 1000, 1500, 2000, 2500\}$.

\Cref{fig:main_experiments}c presents the results.
$\ekqd$-centered and MMD exhibit nearly identical performance, suggesting that the MMD term is dominating in the $\ekqd$-centered estimator.
Among the near-linear time test statistics, $\ekqd$ and $\supkqd$ show a slight advantage over MMD-Multi in distinguishing between the distributions of Galaxy images.

\textbf{4. CIFAR-10 vs. CIFAR-10.1.} We conclude with an experiment on telling apart the CIFAR-10~\citep{Krizhevsky12:ImageNet} and CIFAR-10.1~\citep{recht2019imagenet} test sets, following again \citet{liu2020learning} and \citet{biggs2024mmd}. The dimension is $d = 3 \times 32 \times 32 = 3072$. This is a challenging task, as CIFAR-10.1 was designed to provide new samples from the CIFAR-10 distribution, making it an alternative test set for models trained on CIFAR-10. We conduct the test by drawing $N$ samples from CIFAR-10, and $N$ samples from CIFAR-10.1, with $N \in \{100, 500, 1000, 1500, 2000\}$.

\Cref{fig:main_experiments}d presents the results. Consistent with previous observations, test statistics with quadratic computational complexity exhibit nearly identical performance. However, our quantile discrepancy estimators with near-linear complexity significantly outperform the fast MMD estimators (MMD-Multi) of the same complexity, highlighting the practical advantages of our methods in real-world testing scenarios where computational efficiency is a critical consideration.

\begin{figure}[!t]
    \centering
    \includegraphics[width=0.85\linewidth]{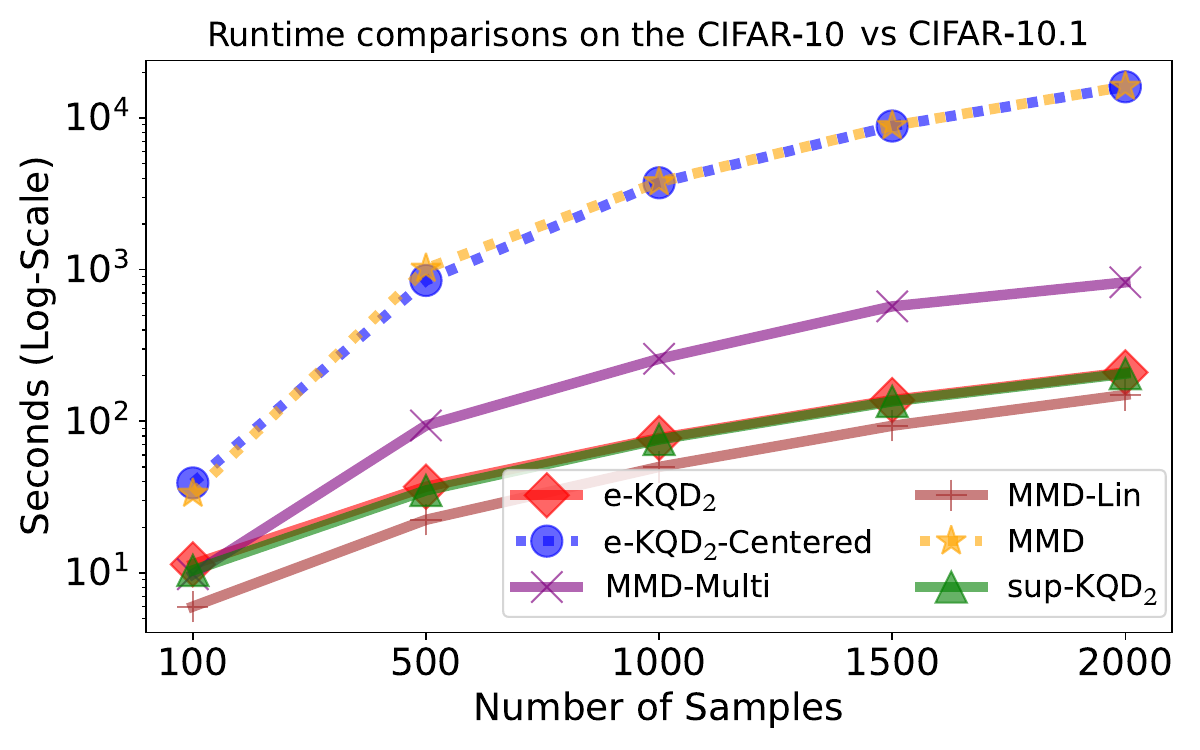}
    \caption[Runtime comparison of KQDs against baseline methods.]{Comparing the time (in seconds) required to complete the CIFAR-10 vs. CIFAR-10.1 experiment, plotted on a logarithmic scale. A shorter time indicates a faster algorithm. These results align with our complexity analysis.}
    \label{fig:run_time}
\end{figure}

An empirical runtime comparison of all methods is presented in \Cref{fig:run_time}, which shows the time (in seconds) required to complete this experiment. The empirical results align with our complexity analysis: the near-linear estimators exhibit comparable performance, while the quadratic estimators are significantly slower. The proposed near-linear KQD estimator is suitable for larger-scale datasets.

\section{Conclusion}

This work explores representations of distributions in an RKHS beyond the mean, using functional quantiles to capture richer distributional characteristics. We introduce kernel quantile embeddings (KQEs) and their associated kernel quantile discrepancies (KQDs), and establish that the conditions required for KQD to define a distance are strictly more general than those needed for MMD to be a distance. Additionally, we propose an efficient estimator for the expected KQD based on Gaussian measures, and demonstrate its effectiveness compared to MMD and its fast approximations through extensive experiments in two-sample testing.
Our findings demonstrate the potential of KQEs as a powerful alternative to traditional mean-based representations.

Several promising avenues remain. Firstly, future work could explore more sophisticated methods for improving the empirical estimates of KQEs.
The study of optimal kernel selection to maximise test power when using KQD for hypothesis testing, analogous to existing work on MMDs~\citep{jitkrittum2020testing,liu2020learning,schrab2023mmd} could also be explored.
Secondly, considering the demonstrated potential of functional quantiles for representing marginal distributions, it is natural to ask whether they could provide a powerful alternative to conditional mean embeddings (CMEs) \citep{Song10:KCOND,Park20:CME}, the Hilbert space representations of conditional distributions. These directions could extend the applicability of KQEs to tasks where KMEs are currently used, such as (conditional) independence testing, causal inference, reinforcement learning, learning on distributions, generative modelling, robust parameter estimation, and Bayesian representations of distributions via kernel mean embeddings, as explored in \citet{flaxman2016bayesian,chau2021deconditional,chau2021bayesimp}, among others.

\part{Discussion and Future Work}
\label{sec:conclusion}
This thesis advanced kernel-based discrepancy methods for statistical computation along two axes. First, we built on the success of the Maximum Mean Discrepancy (MMD), an efficient and expressive kernel-based discrepancy, and showed how to extend and adapt it to specific statistical inference and computation tasks in order to broaden its applicability. Second, we considered alternative RKHS embeddings for richer distributional comparisons, and proposed new kernel-based discrepancies that retain computational tractability while relaxing assumptions needed to be a metric.

The results of this thesis suggest two broader lessons. First, exploit problem structure and prior knowledge: task-specific estimators and uncertainty quantification can significantly improve the performance of kernel-based methods. Second, closed functional forms do not guarantee practicality: although MMD is a difference of kernel mean functions, estimating said functions can be expensive; by contrast, our kernel quantile discrepancies lack a closed form but compensate for it with efficient estimators. Specifically, these are supported in each chapter as follows.

\paragraph{Optimally-Weighted MMD Estimator for Simulation-Based Inference.} In Chapter 5, we introduced an optimally-weighted MMD estimator for simulation-based (also known as likelihood-free) inference that exploits the simulator’s smoothness. By replacing the usual V- or U-statistic with Bayesian quadrature to estimate the MMD, we achieved faster convergence, thus reducing the number of simulations required.
\paragraph{MMD-minimising Quadrature for Conditional Expectations.} In Chapter 4, we extended Bayesian quadrature, an MMD-minimising numerical integration method, to efficiently compute conditional, or parametric, expectations. The proposed two-stage method enables information sharing across parameters and provides accurate estimates with a quantification of uncertainty for each conditional integral while reducing the sample requirements in challenging integration tasks.
\paragraph{Calibration for MMD-minimising Quadrature.} In Chapter 6, we examined the trustworthiness of uncertainty calibration in Bayesian quadrature, for the case of the Brownian motion kernel. We showed that selecting the kernel amplitude by cross-validation rather than maximum likelihood produces posterior variances that better match true integration error. Further, we contrasted this with uncertainty calibration for Gaussian processes, and showed that, despite Bayesian quadrature making explicit use of Gaussian processes, integration and prediction benefit from \emph{different} calibration strategies.
\paragraph{Kernel Quantile Discrepancies.} In Chapter 7, we went beyond kernel mean embeddings and MMD by introducing kernel-based discrepancies that compare quantiles rather than means. We proved these discrepancies are valid metrics under substantially weaker conditions than MMD and, at fixed sample budgets, they match or outperform MMD. Specific variants recover optimal-transport metrics: they can be viewed as kernelised sliced Wasserstein distances.

Each chapter has already outlined future work specific to each line of work. Here, we focus on broader directions of future work in kernel-based discrepancies, based on the two broader lessons stated above.

\paragraph{Task-specific kernel selection.} 
Kernel choice is central to practical applications. The kernel that best calibrates credible intervals for a \emph{target linear functional} $L$ on the RKHS $\calH_k$ can vary with $L$.~\Cref{sec:gpcv} showed this: the kernel that best calibrates Bayesian quadrature (with $L$ the integration operator) can differ from the kernel that best calibrates pointwise predictions (with $L$ the point evaluation). The approach can be extended to other $L$, such as a gradient map, and solutions to differential equations.

Beyond linear functionals, kernel selection remains an open problem for most tasks in~\Cref{sec:introduction}. In hypothesis testing, kernels are often chosen to maximise a proxy for test power~\citep{gretton2012optimal}; however, this is sometimes deemed data-intensive, with alternative approaches aggregating kernels for greater total power~\citep{schrab2023mmd}. In simulation-based inference, the kernel choice problem parallels choosing summary statistics in classical approximate Bayesian computation~\citep{nunes2010optimal}, with some advances made in that area~\citep{gonzalez2019akl,hsu2019bayesian}. In parameter estimation, kernel choice has been studied through a robustness lens~\citep{cherief2020mmd,cherief2021}. Open problems include characterising efficiency–robustness trade-offs and adapting kernels to latent low-dimensional structure~\citep{oates2022minimum}.

\paragraph{Optimal representations in the finite-sample regime.} Even when the KME fully characterises the distribution, its practical value in the finite-sample regime is limited by estimation error. 
By contrast, our proposed kernel quantile discrepancy is defined through an infinite class of RKHS test functions; nevertheless, experiments suggest $\log N$ functions are often sufficient in practice. 
This observation prompts a question more general to kernel selection: should multiple RKHS functions be used to represent the distribution, to increase discriminative power of a kernel-based discrepancy?

There is evidence towards a positive answer. In hypothesis testing, aggregating multiple kernels has been shown to increase test power by covering complementary alternatives~\citep{schrab2023mmd,biggs2024mmd}. Relatedly, in both testing and generative modelling~\citep{Li2017MMDGAN}, non-characteristic kernels can outperform characteristic ones at finite $N$ in high-dimensional regimes: intuitively speaking, a characteristic kernel may `focus' on many weakly informative features, giving lower test power than a non-characteristic kernel ignoring unimportant features. This perspective suggests that optimal finite-sample representation may be low-dimensional, provided the dimensions are carefully selected.

A possible direction is to proceed via spectral decomposition, by selecting eigenfunctions of the kernel integral operator by criteria that reflect finite-sample utility, and then restricting the discrepancy to the top subset. This idea appears in earlier work on test construction from spectral components and, more recently, in spectral feature selection for kernel discrepancies~\citep{eric2007testing,hagrass2024spectral}; further,~\citep{ozier2024extending} used it to lend interpretability to their method. Overall, there is an argument for viewing kernel discrepancies through a finite-sample lens: instead of insisting on full (asymptotic) characterisation via a single, rich kernel and a corresponding representer, learn a set of RKHS functions and a discrepancy that concentrates discriminative power where the data can actually support it. 
    
\paragraph{Discrepancies for dependent and structured data.} 
    Lastly, many theoretical results in this thesis, \Cref{sec:part1} in particular, rely on a set of standard assumptions: Euclidean (often bounded) domains, i.i.d. samples, densities bounded away from zero, and smoothness of the true function available a priori. Relaxing these assumptions would broaden applicability; however, it is non-trivial and would likely require substantial changes to the methods themselves, in line with the idea of adapting estimators to the task. 
    
    These extensions are practically relevant. There is rich literature on kernels applied in non-Euclidean domains, such as graphs~\citep{kriege2020survey}, arbitrary-length sequences~\citep{hue2010large}, and manifolds~\citep{Jayasumana2016}; in these settings, applying our methods and using Sobolev smoothness assumptions to derive convergence rates may still be feasible under a general notion of a Sobolev space~\citep{hajlasz1996sobolev}. Independence of the data fails in routine settings of time series, spatial, and spatio-temporal data. In this scenario, one option is to assume the dependence is weak and aim to provide guarantees for our methods under this assumption, as is done in~\citep{cherief2021}; or adapt the method, as is done in~\citep{chwialkowski2014wild}. Lastly, when the smoothness of the target function $f$ is unknown or is challenging to estimate (for example, when $f$ is a complex simulator), kernel smoothness must be chosen based on the data, to avoid slow convergence. This is a well-studied problem~\citep{Karvonen2023,Szabo2015}; suggesting the feasibility of incorporating kernel smoothness learning into our methods.

Taken together, these directions aim for a general-purpose toolkit of kernel-based discrepancies that are expressive and task-aware.

\addcontentsline{toc}{chapter}{Bibliography}

\bibliography{references}

@article{bach2022information,
	author    = {Bach, Francis},
	journal   = {IEEE Transactions on Information Theory},
	number    = {2},
	pages     = {752--775},
	publisher = {IEEE},
	title     = {Information theory with kernel methods},
	volume    = {69},
	year      = {2022},
	doi       = {10.1109/TIT.2022.3211077}
}

@book{Berlinet2004,
	author    = {Berlinet, A. and Thomas-Agnan, C.},
	publisher = {Springer Science+Business Media},
	title     = {Reproducing Kernel {H}ilbert Spaces in Probability and Statistics},
	year      = {2004},
	doi       = {10.1007/978-1-4419-9096-9}
}

@InProceedings{Bharti2023,
	title     = {Optimally-weighted Estimators of the Maximum Mean Discrepancy for Likelihood-Free Inference},
	author    = {Bharti, Ayush and Naslidnyk, Masha and Key, Oscar and Kaski, Samuel and Briol, Francois-Xavier},
	booktitle = {Proceedings of the 40th International Conference on Machine Learning},
	pages     = {2289--2312},
	year      = {2023},
	series    = {PMLR 202}
}

@inproceedings{biggs2024mmd,
	author    = {Biggs, Felix and Schrab, Antonin and Gretton, Arthur},
	booktitle = {Advances in Neural Information Processing Systems},
	pages     = {75151--75188},
	title     = {{MMD-FUSE}: learning and combining kernels for two-sample testing without data splitting},
	volume    = {36},
	year      = {2023}
}

@book{bobkov2019one,
	author    = {Bobkov, Sergey and Ledoux, Michel},
	publisher = {American Mathematical Society},
	title     = {One-dimensional empirical measures, order statistics, and {K}antorovich transport distances},
	volume    = {261},
	year      = {2019},
	doi       = {10.1090/memo/1259}
}

@article{Bodenham2023,
	author    = {Bodenham, D. A. and Kawahara, Y.},
	journal   = {Statistics and Computing},
	number    = {5},
	pages     = {1--14},
	publisher = {Springer US},
	title     = {eu{MMD}: efficiently computing the {MMD} two-sample test statistic for univariate data},
	volume    = {33},
	year      = {2023},
	doi       = {10.1007/s11222-023-10271-x}
}

@book{bogachev2007measure,
	author    = {Bogachev, Vladimir Igorevich},
	publisher = {Springer},
	title     = {Measure theory},
	volume    = {1},
	year      = {2007},
	doi       = {10.1007/978-3-540-34514-5}
}

@article{bonneel2015sliced,
	author    = {Bonneel, Nicolas and Rabin, Julien and Peyr{\'e}, Gabriel and Pfister, Hanspeter},
	journal   = {Journal of Mathematical Imaging and Vision},
	pages     = {22--45},
	publisher = {Springer},
	title     = {Sliced and {R}adon {W}asserstein barycenters of measures},
	volume    = {51},
	year      = {2015},
	doi       = {10.1007/s10851-014-0506-3}
}

@inproceedings{Bonnier2023,
	author    = {Bonnier, P. and Oberhauser, H. and Szab{\'{o}}, Z.},
	booktitle = {Advances in Neural Information Processing Systems},
	title     = {Kernelized cumulants: beyond kernel mean embeddings},
	volume    = {36},
	year      = {2023},
	pages     = {11049--11074},
}

@article{briol2019statistical,
	author    = {Briol, Francois-Xavier and Barp, Alessandro and Duncan, Andrew B and Girolami, Mark},
	journal   = {arXiv:1906.05944},
	title     = {Statistical inference for generative models with maximum mean discrepancy},
	year      = {2019}
}

@InProceedings{chatalic2022nystrom,
	title     = {{N}ystr{\"o}m kernel mean embeddings},
	author    = {Chatalic, Antoine and Schreuder, Nicolas and Rosasco, Lorenzo and Rudi, Alessandro},
	booktitle = {Proceedings of the 39th International Conference on Machine Learning},
	pages     = {3006--3024},
	year      = {2022},
	series    = {PMLR 162}
}

@inproceedings{chau2021bayesimp,
	author    = {Chau, Siu Lun and Ton, Jean-Francois and Gonz{\'a}lez, Javier and Teh, Yee and Sejdinovic, Dino},
	booktitle = {Advances in Neural Information Processing Systems},
	pages     = {3466--3477},
	title     = {{B}ayes{I}mp: uncertainty quantification for causal data fusion},
	volume    = {34},
	year      = {2021}
}

@inproceedings{chau2021deconditional,
	author    = {Chau, Siu Lun and Bouabid, Shahine and Sejdinovic, Dino},
	booktitle = {Advances in Neural Information Processing Systems},
	pages     = {17813--17825},
	title     = {Deconditional downscaling with {G}aussian processes},
	volume    = {34},
	year      = {2021}
}

@inproceedings{chau2022rkhs,
	author    = {Chau, Siu Lun and Hu, Robert and Gonzalez, Javier and Sejdinovic, Dino},
	booktitle = {Advances in Neural Information Processing Systems},
	pages     = {13050--13063},
	title     = {{RKHS-SHAP}: {S}hapley values for kernel methods},
	volume    = {35},
	year      = {2022}
}

@inproceedings{chau2023explaining,
	author    = {Chau, Siu Lun and Muandet, Krikamol and Sejdinovic, Dino},
	booktitle = {Advances in Neural Information Processing Systems},
	pages     = {50769--50795},
	title     = {Explaining the uncertain: stochastic {S}hapley values for {G}aussian process models},
	volume    = {36},
	year      = {2023}
}

@InProceedings{chau2025credal,
	title     = {Credal Two-Sample Tests of Epistemic Uncertainty},
	author    = {Chau, Siu Lun and Schrab, Antonin and Gretton, Arthur and Sejdinovic, Dino and Muandet, Krikamol},
	booktitle = {Proceedings of The 28th International Conference on Artificial Intelligence and Statistics},
	pages     = {127--135},
	year      = {2025},
	series    = {PMLR 258}
}

@inproceedings{chwialkowski2015fast,
	author    = {Chwialkowski, Kacper P and Ramdas, Aaditya and Sejdinovic, Dino and Gretton, Arthur},
	booktitle = {Advances in Neural Information Processing Systems},
	title     = {Fast two-sample testing with analytic representations of probability measures},
	volume    = {28},
	year      = {2015},
	pages     = {1981--1989}
}

@article{cramer1936some,
	author    = {Cram{\'e}r, Harald and Wold, Herman},
	journal   = {Journal of the London Mathematical Society},
	number    = {4},
	pages     = {290--294},
	publisher = {Wiley Online Library},
	title     = {Some theorems on distribution functions},
	volume    = {1},
	year      = {1936},
	doi       = {10.1112/jlms/s1-11.4.290}
}

@article{cuesta2007sharp,
	author    = {Cuesta-Albertos, Juan Antonio and Fraiman, Ricardo and Ransford, Thomas},
	journal   = {Journal of Theoretical Probability},
	number    = {2},
	pages     = {201--209},
	publisher = {Springer},
	title     = {A sharp form of the {C}ram{\'e}r--{W}old theorem},
	volume    = {20},
	year      = {2007},
	doi       = {10.1007/s10959-007-0060-7}
}

@inproceedings{cuturi2013sinkhorn,
	author    = {Cuturi, Marco},
	booktitle = {Advances in Neural Information Processing Systems},
	title     = {{S}inkhorn distances: lightspeed computation of optimal transport},
	volume    = {26},
	year      = {2013},
	pages     = {2292--2300}
}

@inproceedings{Deshpande2018,
	author    = {Deshpande, I. and Zhang, Z. and Schwing, A.},
	booktitle = {IEEE Conference on Computer Vision and Pattern Recognition},
	pages     = {3483--3491},
	title     = {Generative modeling using the sliced {W}asserstein distance},
	year      = {2018},
	doi       = {10.1109/cvpr.2018.00367}
}

@article{dominicy2013method,
	author    = {Dominicy, Yves and Veredas, David},
	journal   = {Journal of Econometrics},
	number    = {2},
	pages     = {235--247},
	publisher = {Elsevier},
	title     = {The method of simulated quantiles},
	volume    = {172},
	year      = {2013},
	doi       = {10.2139/ssrn.1561185 }
}

@article{dvoretzky1956asymptotic,
	author    = {Dvoretzky, Aryeh and Kiefer, Jack and Wolfowitz, Jacob},
	journal   = {The Annals of Mathematical Statistics},
	pages     = {642--669},
	publisher = {JSTOR},
	title     = {Asymptotic minimax character of the sample distribution function and of the classical multinomial estimator},
	year      = {1956},
	doi       = {10.1007/978-1-4613-8505-9_17}
}

@InProceedings{Feydy2019,
	title     = {Interpolating between optimal transport and {MMD} using {S}inkhorn divergences},
	author    = {Feydy, Jean and S\'{e}journ\'{e}, Thibault and Vialard, Fran\c{c}ois-Xavier and Amari, Shun-ichi and Trouve, Alain and Peyr\'{e}, Gabriel},
	booktitle = {Proceedings of the Twenty-Second International Conference on Artificial Intelligence and Statistics},
	pages     = {2681--2690},
	year      = {2019},
	series    = {PMLR 89}
}

@inproceedings{flaxman2016bayesian,
	author    = {Flaxman, Seth and Sejdinovic, Dino and Cunningham, John P. and Filippi, Sarah},
	title     = {{B}ayesian learning of kernel embeddings},
	year      = {2016},
	booktitle = {Proceedings of the 32nd Conference on Uncertainty in Artificial Intelligence},
	pages     = {182--191}
}

@article{Fournier2015,
	author    = {Fournier, Nicolas and Guillin, Arnaud},
	journal   = {Probability Theory and Related Fields},
	number    = {3-4},
	pages     = {707--738},
	title     = {On the rate of convergence in {W}asserstein distance of the empirical measure},
	volume    = {162},
	year      = {2015},
	doi       = {10.1007/s00440-014-0583-7}
}

@article{fraiman2012quantiles,
	author    = {Fraiman, Ricardo and Pateiro-L{\'o}pez, Beatriz},
	journal   = {Journal of Multivariate Analysis},
	pages     = {1--14},
	publisher = {Elsevier},
	title     = {Quantiles for finite and infinite dimensional data},
	volume    = {108},
	year      = {2012},
	doi       = {10.1016/j.jmva.2012.01.016}
}

@inproceedings{gao2024model,
	author    = {Irena Gao and Percy Liang and Carlos Guestrin},
	title     = {Model Equality Testing: which Model Is This {API} Serving?},
	booktitle = {International Conference on Learning Representations},
	year      = {2025}
}

@InProceedings{Genevay2019,
	title     = {Sample Complexity of {S}inkhorn Divergences},
	author    = {Genevay, Aude and Chizat, L\'{e}na\"{i}c and Bach, Francis and Cuturi, Marco and Peyr\'{e}, Gabriel},
	booktitle = {Proceedings of the Twenty-Second International Conference on Artificial Intelligence and Statistics},
	pages     = {1574--1583},
	year      = {2019},
	series    = {PMLR 89}
}

@inproceedings{gretton2006kernel,
	author    = {Gretton, Arthur and Borgwardt, Karsten and Rasch, Malte and Sch{\"o}lkopf, Bernhard and Smola, Alex},
	booktitle = {Advances in Neural Information Processing Systems},
	pages     = {513--520},
	title     = {A kernel method for the two-sample-problem},
	volume    = {19},
	year      = {2006}
}

@inproceedings{gretton2009fast,
	author    = {Gretton, Arthur and Fukumizu, Kenji and Harchaoui, Zaid and Sriperumbudur, Bharath K},
	booktitle = {Advances in Neural Information Processing Systems},
	title     = {A fast, consistent kernel two-sample test},
	volume    = {22},
	year      = {2009},
	pages     = {673--681},
}

@article{gretton2012kernel,
	author    = {Arthur Gretton and Karsten M. Borgwardt and Malte J. Rasch and Bernhard Sch{{\"o}}lkopf and Alexander Smola},
	title     = {A Kernel Two-Sample Test},
	journal   = {Journal of Machine Learning Research},
	year      = {2012},
	volume    = {13},
	number    = {25},
	pages     = {723--773}
}

@book{Hirsch1976,
	author    = {Hirsch, Morris W.},
	publisher = {Springer-Verlag},
	series    = {Graduate Texts in Mathematics},
	title     = {Differential Topology},
	volume    = {33},
	year      = {1976},
	doi       = {10.1007/978-1-4684-9449-5}
}

@article{huber1964robust,
	author    = {Huber, Peter J},
	journal   = {Ann. Math. Statist.},
	number    = {4},
	pages     = {73--101},
	title     = {Robust Estimation of a Location Parameter},
	volume    = {35},
	year      = {1964},
	doi       = {10.1007/978-1-4612-4380-9_35}
}

@InProceedings{jitkrittum2020testing,
	title     = {Testing goodness of fit of conditional density models with kernels},
	author    = {Jitkrittum, Wittawat and Kanagawa, Heishiro and Sch\"{o}lkopf, Bernhard},
	booktitle = {Proceedings of the 36th Conference on Uncertainty in Artificial Intelligence (UAI)},
	pages     = {221--230},
	year      = {2020},
	series    = {PMLR 124}
}

@article{Kantorovich1942,
	author    = {Kantorovich, L. V.},
	journal   = {Dokl. Akad. Nauk SSSR},
	number    = {7-8},
	pages     = {227--229},
	title     = {On the translocation of masses},
	volume    = {37},
	year      = {1942},
	doi       = {10.1287/mnsc.5.1.1}
}

@book{Kolmogorov60:Foundations,
	author    = {Kolmogorov, Andrey N.},
	edition   = {2},
	publisher = {Chelsea Pub Co},
	title     = {Foundations of the Theory of Probability},
	year      = {1960},
	doi       = {10.2307/2332488}
}

@inproceedings{Kolouri2019,
	author    = {Kolouri, S. and Nadjahi, K. and Simsekli, U. and Badeau, R. and Rohde, G. K.},
	booktitle = {Advances in Neural Information Processing Systems},
	pages     = {261--272},
	title     = {Generalized sliced {W}asserstein distances},
	year      = {2019}
}

@inproceedings{kolouri_generalized_2022,
	author    = {Kolouri, Soheil and Nadjahi, Kimia and Shahrampour, Shahin and Şimşekli, Umut},
	booktitle = {{IEEE} {International} {Conference} on {Acoustics}, {Speech} and {Signal} {Processing}},
	doi       = {10.1109/ICASSP43922.2022.9746016},
	pages     = {4513--4517},
	title     = {Generalized sliced probability metrics},
	year      = {2022}
}

@article{Kong2012,
	author    = {Kong, L. and Mizera, I.},
	journal   = {Statistica Sinica},
	number    = {4},
	pages     = {1589--1610},
	title     = {Quantile tomography: using quantiles with multivariate data},
	volume    = {22},
	year      = {2012},
	doi       = {10.5705/ss.2010.224}
}

@article{Kosorok1999,
	author    = {Kosorok, M. R.},
	journal   = {Biometrika},
	number    = {4},
	pages     = {909--921},
	title     = {Two-sample quantile tests under general conditions},
	volume    = {86},
	year      = {1999},
	doi       = {10.1093/biomet/86.4.909}
}

@article{kriege2020survey,
	author    = {Kriege, Nils M and Johansson, Fredrik D and Morris, Christopher},
	journal   = {Applied Network Science},
	pages     = {1--42},
	publisher = {Springer},
	title     = {A survey on graph kernels},
	year      = {2020},
	doi       = {10.1007/s41109-019-0195-3}
}

@inproceedings{Krizhevsky12:ImageNet,
	author    = {Krizhevsky, Alex and Sutskever, Ilya and Hinton, Geoffrey E},
	booktitle = {Advances in Neural Information Processing Systems},
	title     = {{ImageNet} Classification with Deep Convolutional Neural Networks},
	volume    = {25},
	year      = {2012},
	pages     = {1097--1105}
}

@book{kukush2020gaussian,
	author    = {Kukush, Alexander},
	publisher = {John Wiley \& Sons},
	title     = {{G}aussian measures in {H}ilbert space: construction and properties},
	year      = {2020},
	doi       = {10.1002/9781119476825}
}

@book{lehmann1986testing,
	author    = {Lehmann, Erich Leo and Romano, Joseph P and Casella, George},
	publisher = {Springer},
	title     = {Testing statistical hypotheses},
	volume    = {3},
	year      = {1986},
	doi       = {10.2307/2533531}
}

@InProceedings{lerasle2019monk,
	title     = {{MONK} -- outlier-robust mean embedding estimation by median-of-means},
	author    = {Lerasle, Matthieu and Szabo, Zoltan and Mathieu, Timoth{\'e}e and Lecue, Guillaume},
	booktitle = {Proceedings of the 36th International Conference on Machine Learning},
	pages     = {3782--3793},
	year      = {2019},
	series    = {PMLR 97}
}

@article{Li2007,
	author    = {Li, Y. and Liu, Y. and Zhu, J.},
	journal   = {Journal of the American Statistical Association},
	number    = {477},
	pages     = {255--268},
	title     = {Quantile regression in reproducing kernel {H}ilbert spaces},
	volume    = {102},
	year      = {2007},
	doi       = {10.1198/016214506000000979}
}

@InProceedings{liu2020learning,
	title     = {Learning Deep Kernels for Non-Parametric Two-Sample Tests},
	author    = {Liu, Feng and Xu, Wenkai and Lu, Jie and Zhang, Guangquan and Gretton, Arthur and Sutherland, Danica J.},
	booktitle = {Proceedings of the 37th International Conference on Machine Learning},
	pages     = {6316--6326},
	year      = {2020},
	series    = {PMLR 119}
}

@article{magesh2023principled,
	author    = {Akshayaa Magesh and Venugopal V. Veeravalli and Anirban Roy and Susmit Jha},
	title     = {Principled Out-of-Distribution Detection via Multiple Testing},
	journal   = {Journal of Machine Learning Research},
	year      = {2023},
	volume    = {24},
	number    = {378},
	pages     = {1--35},
}

@article{makigusa2024two,
	author    = {Makigusa, Natsumi},
	journal   = {Communications in Statistics-Theory and Methods},
	number    = {15},
	pages     = {5421--5438},
	publisher = {Taylor \& Francis},
	title     = {Two-sample test based on maximum variance discrepancy},
	volume    = {53},
	year      = {2024},
	doi       = {10.1080/03610926.2023.2220851}
}

@article{minsker_geometric_2015,
	author    = {Minsker, Stanislav},
	doi       = {10.3150/14-BEJ645},
	journal   = {Bernoulli},
	number    = {4},
	title     = {Geometric median and robust estimation in {B}anach spaces},
	volume    = {21},
	year      = {2015}
}

@inproceedings{Muandet12:SMM,
	author    = {Muandet, Krikamol and Fukumizu, Kenji and Francesco Dinuzzo and Sch\"{o}lkopf, Bernhard},
	booktitle = {Advances in Neural Information Processing Systems},
	pages     = {10--18},
	title     = {Learning from Distributions via Support Measure Machines},
	year      = {2012}
}

@article{muandet2021counterfactual,
	author    = {Krikamol Muandet and Motonobu Kanagawa and Sorawit Saengkyongam and Sanparith Marukatat},
	title     = {Counterfactual Mean Embeddings},
	journal   = {Journal of Machine Learning Research},
	year      = {2021},
	volume    = {22},
	number    = {162},
	pages     = {1--71},
}

@article{muller1997integral,
	author    = {M{\"u}ller, Alfred},
	journal   = {Advances in applied probability},
	number    = {2},
	pages     = {429--443},
	publisher = {Cambridge University Press},
	title     = {Integral probability metrics and their generating classes of functions},
	volume    = {29},
	year      = {1997},
	doi       = {10.2307/1428011}
}

@inproceedings{Nadjahi2020,
	author    = {Nadjahi, K. and Durmus, A. and Chizat, L. and Kolouri, S. and Shahrampour, S. and Şimşekli, U.},
	booktitle = {Advances in Neural Information Processing Systems},
	title     = {Statistical and topological properties of sliced probability divergences},
	year      = {2020},
	pages     = {20802--20812}
}

@article{nienkotter2022kernel,
	author    = {Nienk{\"o}tter, Andreas and Jiang, Xiaoyi},
	journal   = {IEEE Transactions on Pattern Analysis and Machine Intelligence},
	number    = {5},
	pages     = {5872--5888},
	publisher = {IEEE},
	title     = {Kernel-based generalized median computation for consensus learning},
	doi       = {10.1109/TPAMI.2022.3202565},
	volume    = {45},
	year      = {2022}
}

@article{Niu2023,
	author    = {Niu, Z. and Meier, J. and Briol, F-X},
	journal   = {Electronic Journal of Statistics},
	number    = {1},
	pages     = {1411--1456},
	title     = {Discrepancy-based inference for intractable generative models using quasi-{M}onte {C}arlo},
	volume    = {17},
	year      = {2023},
	doi       = {10.1214/23-ejs2131}
}

@inproceedings{Park20:CME,
	author    = {Park, Junhyung and Muandet, Krikamol},
	booktitle = {Advances in Neural Information Processing Systems},
	pages     = {21247--21259},
	title     = {A Measure-Theoretic Approach to Kernel Conditional Mean Embeddings},
	year      = {2020}
}

@inproceedings{rabin2011wasserstein,
	author    = {Rabin, Julien and Peyr{\'e}, Gabriel and Delon, Julie and Bernot, Marc},
	booktitle = {Scale Space and Variational Methods in Computer Vision},
	pages     = {435--446},
	title     = {{W}asserstein barycenter and its application to texture mixing},
	year      = {2012},
	publisher = {Springer Berlin Heidelberg},
	doi       = {10.1007/978-3-642-24785-9_37}
}

@inproceedings{ramdas2015decreasing,
	author    = {Ramdas, Aaditya and Reddi, Sashank Jakkam and P{\'o}czos, Barnab{\'a}s and Singh, Aarti and Wasserman, Larry},
	booktitle = {AAAI Conference on Artificial Intelligence},
	title     = {On the decreasing power of kernel and distance based nonparametric hypothesis tests in high dimensions},
	volume    = {29},
	year      = {2015},
	doi       = {10.1609/aaai.v29i1.9692},
	pages     = {3571--3577}
}

@article{ramdas2017wasserstein,
	author    = {Ramdas, Aaditya and Garc{\'\i}a Trillos, Nicol{\'a}s and Cuturi, Marco},
	journal   = {Entropy},
	number    = {2},
	title     = {On {W}asserstein two-sample testing and related families of nonparametric tests},
	volume    = {19},
	year      = {2017},
	doi       = {10.3390/e19020047}
}

@article{ranger2020minimum,
	author    = {Ranger, Jochen and Kuhn, J{\"o}rg-Tobias and Szardenings, Carsten},
	journal   = {Multivariate Behavioral Research},
	number    = {6},
	pages     = {941--957},
	publisher = {Taylor \& Francis},
	title     = {Minimum distance estimation of multidimensional diffusion-based item response theory models},
	volume    = {55},
	year      = {2020},
	doi       = {10.1080/00273171.2019.1704676}
}

@InProceedings{recht2019imagenet,
	title     = {Do {I}mage{N}et Classifiers Generalize to {I}mage{N}et?},
	author    = {Recht, Benjamin and Roelofs, Rebecca and Schmidt, Ludwig and Shankar, Vaishaal},
	booktitle = {Proceedings of the 36th International Conference on Machine Learning},
	pages     = {5389--5400},
	year      = {2019},
	series    = {PMLR 97},
}

@inproceedings{Schrab2022,
	author    = {Schrab, A. and Kim, I. and Guedj, B. and Gretton, A.},
	booktitle = {Advances in Neural Information Processing Systems},
	pages     = {18793--18807},
	title     = {Efficient aggregated kernel tests using incomplete {U}-statistics},
	year      = {2022}
}

@article{schrab2023mmd,
	author    = {Antonin Schrab and Ilmun Kim and MÃ©lisande Albert and BÃ©atrice Laurent and Benjamin Guedj and Arthur Gretton},
	title     = {{MMD} Aggregated Two-Sample Test},
	journal   = {Journal of Machine Learning Research},
	year      = {2023},
	volume    = {24},
	number    = {194},
	pages     = {1--81},
}

@article{schrab2025unified,
	author    = {Schrab, Antonin},
	journal   = {arXiv:2503.07084},
	title     = {A Unified View of Optimal Kernel Hypothesis Testing},
	year      = {2025}
}

@article{sejdinovic2013equivalence,
	author    = {Sejdinovic, Dino and Sriperumbudur, Bharath and Gretton, Arthur and Fukumizu, Kenji},
	journal   = {The Annals of Statistics},
	pages     = {2263--2291},
	title     = {Equivalence of distance-based and {RKHS}-based statistics in hypothesis testing},
	year      = {2013},
	doi       = {10.1214/13-aos1140}
}

@article{Sejdinovic2024,
	author    = {Sejdinovic, D.},
	journal   = {arXiv:2410.22754},
	title     = {An overview of causal inference using kernel embeddings},
	year      = {2024}
}

@article{Serfling2002,
	author    = {Serfling, R.},
	journal   = {Statistica Neerlandica},
	number    = {2},
	pages     = {214--232},
	title     = {Quantile functions for multivariate analysis: approaches and applications},
	volume    = {56},
	year      = {2002},
	doi       = {10.1111/1467-9574.00195}
}

@book{serfling2009approximation,
	author    = {Serfling, Robert J},
	publisher = {John Wiley \& Sons},
	title     = {Approximation theorems of mathematical statistics},
	year      = {2009},
	doi       = {10.1002/9780470316481}
}

@article{Sheather1990,
	author    = {Sheather, S. J. and Marron, J. S.},
	journal   = {Journal of the American Statistical Association},
	number    = {410},
	pages     = {410--416},
	title     = {Kernel quantile estimators},
	volume    = {85},
	year      = {1990},
	doi       = {10.1080/01621459.1990.10476214}
}

@inproceedings{smola07hilbert,
	author    = {Smola, Alexander J. and Gretton, Arthur and Song, Le and Bernhard Sch\"{o}lkopf},
	booktitle = {Algorithmic Learning Theory},
	pages     = {13--31},
	title     = {A {H}ilbert space embedding for distributions},
	year      = {2007},
	publisher = {Springer Berlin Heidelberg},
	doi       = {10.1007/978-3-540-75488-6_5}
}

@inproceedings{Song10:KCOND,
	author    = {Le Song and Jonathan Huang and Alex Smola and Kenji Fukumizu},
	booktitle = {Proceedings of the 26th International Conference on Machine Learning},
	title     = {{H}ilbert Space Embeddings of Conditional Distributions with Applications to Dynamical Systems},
	year      = {2009},
	doi       = {10.1145/1553374.1553497}, 
	pages     = {961--968}
}

@article{Sriperumbudur2009,
	author    = {Bharath K. Sriperumbudur and Arthur Gretton and Kenji Fukumizu and Bernhard Sch{{\"o}}lkopf and Gert R.G. Lanckriet},
	title     = {{H}ilbert space embeddings and metrics on probability measures},
	journal   = {Journal of Machine Learning Research},
	year      = {2010},
	volume    = {11},
	number    = {50},
	pages     = {1517--1561},
}

@article{sriperumbudur2011universality,
	author    = {Bharath K. Sriperumbudur and Kenji Fukumizu and Gert R.G. Lanckriet},
	title     = {Universality, Characteristic Kernels and {RKHS} Embedding of Measures},
	journal   = {Journal of Machine Learning Research},
	year      = {2011},
	volume    = {12},
	number    = {70},
	pages     = {2389--2410},
}

@article{stolfi2022sparse,
	author    = {Stolfi, Paola and Bernardi, Mauro and Petrella, Lea},
	journal   = {Econometrics and Statistics},
	publisher = {Elsevier},
	title     = {Sparse simulation-based estimator built on quantiles},
	year      = {2022},
	doi       = {10.1016/j.ecosta.2022.01.006}
}

@article{stuart2010inverse,
	author    = {Stuart, Andrew M},
	journal   = {Acta numerica},
	pages     = {451--559},
	publisher = {Cambridge University Press},
	title     = {Inverse problems: a {B}ayesian perspective},
	volume    = {19},
	year      = {2010},
	doi       = {10.1017/s0962492910000061}
}

@article{szabo2016learning,
	author    = {Zolt{{\'a}}n Szab{{\'o}} and Bharath K. Sriperumbudur and Barnab{{\'a}}s P{{\'o}}czos and Arthur Gretton},
	title     = {Learning Theory for Distribution Regression},
	journal   = {Journal of Machine Learning Research},
	year      = {2016},
	volume    = {17},
	number    = {152},
	pages     = {1--40},
}

@article{SzeRiz05,
	author    = {G\'abor J. Sz\'ekely and Maria L. Rizzo},
	journal   = {Journal of Multivariate Analysis},
	number    = {1},
	pages     = {58--80},
	title     = {A new test for multivariate normality},
	volume    = {93},
	year      = {2005},
	doi       = {10.1016/j.jmva.2003.12.002}
}

@article{tolstikhin2017minimax,
	author    = {Ilya Tolstikhin and Bharath K. Sriperumbudur and Krikamol Muandet},
	title     = {Minimax Estimation of Kernel Mean Embeddings},
	journal   = {Journal of Machine Learning Research},
	year      = {2017},
	volume    = {18},
	number    = {86},
	pages     = {1--47},
}

@book{vakhania1987probability,
	author    = {Vakhania, Nicholas and Tarieladze, Vazha and Chobanyan, S},
	publisher = {Springer Science \& Business Media},
	title     = {Probability distributions on {B}anach spaces},
	volume    = {14},
	year      = {1987},
	doi       = {10.1007/978-94-009-3873-1}
}

@book{villani2009optimal,
	author    = {Villani, C{\'e}dric},
	publisher = {Springer},
	title     = {Optimal transport: old and new},
	volume    = {338},
	year      = {2009},
	doi       = {10.1007/978-3-540-71050-9}
}

@article{walmsley2022galaxy,
	author    = {Walmsley, Mike and Lintott, Chris and G{\'e}ron, Tobias and Kruk, Sandor and Krawczyk, Coleman and Willett, Kyle W and Bamford, Steven and Kelvin, Lee S and Fortson, Lucy and Gal, Yarin and Keel, William and Masters, Karen L and Mehta, Vihang and Simmons, Brooke D and Smethurst, Rebecca and Smith, Lewis and Baeten, Elisabeth M and Macmillan, Christine},
	journal   = {Monthly Notices of the Royal Astronomical Society},
	number    = {3},
	pages     = {3966--3988},
	publisher = {Oxford University Press},
	title     = {Galaxy {Z}oo {DECaLS}: detailed visual morphology measurements from volunteers and deep learning for 314 000 galaxies},
	volume    = {509},
	year      = {2022},
	doi       = {10.1093/mnras/staf025}
}

@InProceedings{wang2025statistical,
	title     = {Statistical and Computational Guarantees of Kernel Max-Sliced {W}asserstein Distances},
	author    = {Wang, Jie and Boedihardjo, March and Xie, Yao},
	booktitle = {Proceedings of the 42nd International Conference on Machine Learning},
	pages     = {62373--62400},
	year      = {2025},
	series    = {PMLR 267},
}

@inproceedings{wang_two-sample_2022,
	author    = {Wang, Jie and Gao, Rui and Xie, Yao},
	booktitle = {Proceedings of the 25th {International} {Conference} on {Artificial} {Intelligence} and {Statistics}},
	pages     = {8022--8055},
	title     = {Two-Sample Test with Kernel Projected {W}asserstein Distance},
	year      = {2022}
}

@article{ziegel2022characteristic,
	author    = {Ziegel, Johanna and Ginsbourger, David and D{\"{u}}mbgen, Lutz},
	journal   = {Bernoulli},
	number    = {2},
	pages     = {1441--1457},
	title     = {Characteristic kernels on {H}ilbert spaces, {B}anach spaces, and on sets of measures},
	volume    = {30},
	year      = {2024},
	doi       = {10.3150/23-bej1639}
}

@book{Willard1970,
	title     = {General Topology},
	author    = {Willard, Stephen},
	publisher = {Addison Wesley Longman Publishing},
	series    = {Addison-Wesley Series in Mathematics},
	year      = {1970},
    isbn      = {\isbnhref{9780486434797}}
}

@book{Wendland2005,
	author    = {Wendland, H.},
	title     = {Scattered Data Approximation},
	publisher = {Cambridge University Press},
	year      = {2005},
	series    = {Cambridge Monographs on Applied and Computational Mathematics},
	number    = {17},
	doi       = {10.1017/cbo9780511617539}
}

@article{sundararajan2001predictive,
	title     = {Predictive approaches for choosing hyperparameters in {G}aussian processes},
	author    = {Sundararajan, Sellamanickam and Keerthi, S. Sathiya},
	journal   = {Neural Computation},
	volume    = {13},
	number    = {5},
	pages     = {1103--1118},
	year      = {2001},
	doi       = {10.1162/08997660151134343 }
}

@book{rassmussen2006gaussian,
	title     = {{G}aussian processes for machine learning},
	author    = {Rasmussen, Carl Edward and Williams, Christopher K},
	volume    = {2},
	year      = {2006},
	publisher = {MIT press Cambridge, MA},
	doi       = {10.7551/mitpress/3206.001.0001}
}

@article{Anderes2010,
	author    = {Anderes, E.},
	journal   = {Annals of Statistics},
	number    = {2},
	pages     = {870--893},
	title     = {On the consistent separation of scale and variance for {G}aussian random fields},
	volume    = {38},
	year      = {2010},
	doi       = {10.1214/09-aos725}
}

@book{mandelbrot1982fractal,
	title     = {The Fractal Geometry of Nature},
	author    = {Mandelbrot, Benoit B},
	year      = {1982},
	publisher = {WH Freeman New York},
	doi       = {10.21236/ada273271}
}

@article{petit2021gaussian,
	title     = {Parameter Selection in {G}aussian Process Interpolation: an Empirical Study of Selection Criteria},
	volume    = {11},
	doi       = {10.1137/21m1444710},
	number    = {4},
	journal   = {SIAM/ASA Journal on Uncertainty Quantification},
	publisher = {Society for Industrial & Applied Mathematics (SIAM)},
	author    = {Petit,  Sébastien J. and Bect,  Julien and Feliot,  Paul and Vazquez,  Emmanuel},
	year      = {2023},
	pages     = {1308--1328}
}

@article{Bachoc2013,
	author    = {Bachoc, F.},
	title     = {Cross validation and Maximum Likelihood estimations of hyperparameters of {G}aussian processes with model misspecification},
	journal   = {Computational Statistics \& Data Analysis},
	volume    = {66},
	pages     = {55--69},
	year      = {2013},
    doi = {10.1016/j.csda.2013.03.016}
}

@article{OHagan1978,
	author    = {O'Hagan, A.},
	title     = {Curve fitting and optimal design for prediction},
	journal   = {Journal of the Royal Statistical Society: Series B},
	volume    = {40},
	number    = {1},
	pages     = {1--42},
	year      = {1978},
    doi = {10.1111/j.2517-6161.1978.tb01643.x}
}

@article{wynne2021convergence,
	author    = {George Wynne and FranÃ§ois-Xavier Briol and Mark Girolami},
	title     = {Convergence guarantees for {G}aussian process means with misspecified likelihoods and smoothness},
	journal   = {Journal of Machine Learning Research},
	year      = {2021},
	volume    = {22},
	number    = {123},
	pages     = {1--40},
}

@article{Bachoc2017,
	author    = {Bachoc, F. and Lagnoux, A. and Nguyen, T. M. N.},
	title     = {Cross-validation estimation of covariance parameters under fixed-domain asymptotics},
	journal   = {Journal of Multivariate Analysis},
	volume    = {160},
	pages     = {42--67},
	year      = {2017},
    doi = {10.1016/j.jmva.2017.06.003}
}

@article{bachoc2018asymptotic,
	title     = {Asymptotic analysis of covariance parameter estimation for {G}aussian processes in the misspecified case},
	author    = {Bachoc, Fran{\c{c}}ois},
	journal   = {Bernoulli},
	volume    = {24},
	number    = {2},
	pages     = {1531--1575},
	year      = {2018},
	publisher = {Bernoulli Society for Mathematical Statistics and Probability},
    doi = {10.3150/16-BEJ906}
}

@article{bachoc2020asymptotic,
	title     = {Asymptotic properties of the maximum likelihood and cross validation estimators for transformed {G}aussian processes},
	author    = {Bachoc, Fran{\c{c}}ois and Betancourt, Jos{\'e} and Furrer, Reinhard and Klein, Thierry},
	journal   = {Electronic Journal of Statistics},
	volume    = {14},
	number    = {1},
	pages     = {1962--2008},
	year      = {2020},
	publisher = {Institute of Mathematical Statistics and Bernoulli Society},
    doi = {10.1214/20-EJS1712}
}

@inbook{VaartZanten2008,
	author    = {van der Vaart, A. W. and van Zanten, J. H.},
	title     = {Reproducing Kernel {H}ilbert spaces of {G}aussian Priors},
	booktitle = {Pushing the Limits of Contemporary Statistics: Contributions in Honor of
	Jayanta K.\ Ghosh},
	volume    = {3},
	publisher = {Institute of Mathematical Statistics},
	pages     = {200--222},
	year      = {2008},
    doi = {10.1214/074921708000000156}
}

@article{Schaback2018,
	author    = {Schaback, R.},
	title     = {Superconvergence of kernel-based interpolation},
	journal   = {Journal of Approximation Theory},
	volume    = {235},
	pages     = {1--19},
	year      = {2018},
    doi = {10.1016/j.jat.2018.05.002}
}

@incollection{Diaconis1988,
	author    = {Diaconis, Persi},
	title     = {{B}ayesian numerical analysis},
	booktitle = {Statistical Decision Theory and Related Topics IV},
	publisher = {Springer-Verlag New York},
	volume    = {1},
	pages     = {163--175},
	year      = {1988},
    doi = {10.1007/978-1-4613-8768-8_20}
}

@article{Ying1991,
	author    = {Ying, Zhiliang},
	title     = {Asymptotic properties of a maximum likelihood estimator with data from a {G}aussian process},
	journal   = {Journal of Multivariate Analysis},
	volume    = {36},
	number    = {2},
	pages     = {280--296},
	year      = {1991},
    doi = {10.1016/0047-259X(91)90062-7}
}

@article{Ying1993,
	author    = {Ying, Z.},
	title     = {Maximum likelihood estimation of parameters under a spatial sampling scheme},
	journal   = {The Annals of Statistics},
	volume    = {21},
	number    = {3},
	pages     = {1567--1590},
	year      = {1993},
    doi = {10.1214/aos/1176349272}
}

@book{Stein1999,
	author    = {Stein, Michael L.},
	title     = {Interpolation of Spatial Data: {S}ome Theory for Kriging},
	publisher = {Springer},
	year      = {1999},
	series    = {Springer Series in Statistics},
}

@article{LohKam2000,
	author    = {Loh, W.-L. and Kam, T.-K.},
	title     = {Estimating structured correlation matrices in smooth {G}aussian random field models},
	journal   = {Annals of Statistics},
	volume    = {28},
	number    = {3},
	pages     = {880--904},
	year      = {2000},
    doi = {10.1214/aos/1015952003}
}

@article{Stein1993,
	author    = {Stein, M. L.},
	title     = {Spline smoothing with an estimated order parameter},
	journal   = {The Annals of Statistics},
	volume    = {21},
	number    = {3},
	pages     = {1522--1544},
	year      = {1993},
    doi = {10.1214/aos/1176349270}
}

@article{Zhang2004,
	author    = {Zhang, H.},
	title     = {Inconsistent estimation and asymptotically equal interpolations in model-based geostatistics},
	journal   = {Journal of the American Statistical Association},
	volume    = {99},
	number    = {465},
	pages     = {250--261},
	year      = {2004},
    doi = {10.1198/016214504000000241}
}

@article{Loh2005,
	author    = {Loh, W.-L.},
	title     = {Fixed-domain asymptotics for a subclass of {M}atérn-type {G}aussian random fields},
	journal   = {The Annals of Statistics},
	volume    = {33},
	number    = {5},
	pages     = {2344--2394},
	year      = {2005},
    doi = {10.1214/009053605000000516}
}

@article{Du2009,
	author    = {Du, J. and Zhang, H. and Mandrekar, V. S.},
	title     = {Fixed-domain asymptotic properties of tapered maximum likelihood estimators},
	journal   = {The Annals of Statistics},
	volume    = {37},
	number    = {6A},
	pages     = {3330--3361},
	year      = {2009},
    doi = {10.1214/08-aos676}
}

@article{WangLoh2011,
	author    = {Wang, D. and Loh, W.-L.},
	title     = {On fixed-domain asymptotics and covariance tapering in {G}aussian random field models},
	journal   = {Electronic Journal of Statistics},
	volume    = {5},
	pages     = {238--269},
	year      = {2011},
    doi = {10.1214/11-ejs607}
}

@book{Nourdin2012,
	author    = {Nourdin, I.},
	title     = {Selected aspects of fractional {B}rownian motion},
	publisher = {Springer},
	year      = {2012},
	series    = {Bocconi \& Springer Series},
	number    = {4},
    doi = {10.1007/978-88-470-2823-4}
}

@article{Kaufman2013,
	author    = {Kaufman, C. G. and Shaby, B. A.},
	title     = {The role of the range parameter for estimation and prediction in geostatistics},
	journal   = {Biometrika},
	volume    = {100},
	number    = {2},
	pages     = {473--484},
	year      = {2013},
    doi = {10.1093/biomet/ass079}
}

@article{Szabo2015,
	author    = {Szab\'{o}, B. and van der Vaart, A. W. and van Zanten, J. H.},
	title     = {Frequentist coverage of adaptive nonparametric {B}ayesian credible sets},
	journal   = {The Annals of Statistics},
	volume    = {43},
	number    = {4},
	pages     = {1391--1428},
	year      = {2015},
    doi = {10.1214/14-aos1270}
}

@article{XuStein2017,
	author    = {Xu, W. and Stein, M. L.},
	title     = {Maximum Likelihood Estimation for a Smooth {G}aussian Random Field Model},
	journal   = {SIAM/ASA Journal on Uncertainty Quantification},
	volume    = {5},
	number    = {1},
	pages     = {138--175},
	year      = {2017},
    doi = {10.1137/15m105358x}
}

@article{BenSalem2019,
	author    = {{Ben Salem}, M. and Bachoc, F. and Roustant, O. and Gamboa, F. and Tomaso, L.},
	title     = {{G}aussian process-based dimension reduction for goal-oriented sequential design},
	journal   = {SIAM/ASA Journal on Uncertainty Quantification},
	volume    = {7},
	number    = {4},
	pages     = {1369--1397},
	year      = {2019},
    doi = {10.1137/18M1167930}
}

@article{Karvonen2020,
	title     = {Maximum likelihood estimation and uncertainty quantification for {G}aussian process approximation of deterministic functions},
	author    = {Karvonen, Toni and Wynne, George and Tronarp, Filip and Oates, Chris and Sarkka, Simo},
	journal   = {SIAM/ASA Journal on Uncertainty Quantification},
	volume    = {8},
	number    = {3},
	pages     = {926--958},
	year      = {2020},
	publisher = {SIAM},
    doi = {10.1137/20m1315968}
}

@article{Bevilacqua2019,
	author    = {Bevilacqua, M. and Faouzi, T. and Furrer, R. and Porcu, E.},
	title     = {Estimation and prediction using generalized {W}endland covariance functions under fixed domain asymptotics},
	journal   = {The Annals of Statistics},
	volume    = {47},
	number    = {2},
	pages     = {828--856},
	year      = {2019},
    doi = {10.1214/17-AOS1652}
}

@article{FongHolmes2020,
	author    = {Fong, E. and Holmes, C. C.},
	title     = {On the marginal likelihood and cross-validation},
	journal   = {Biometrika},
	volume    = {107},
	number    = {2},
	pages     = {489--496},
	year      = {2020},
    doi = {10.1093/biomet/asz077}
}

@article{HadjiSzabo2021,
	author    = {Hadji, A. and Szab\'{o}, B.},
	title     = {Can we trust {B}ayesian uncertainty quantification from {G}aussian process priors with squared exponential covariance kernel?},
	journal   = {SIAM/ASA Journal on Uncertainty Quantification},
	volume    = {9},
	number    = {1},
	pages     = {185--230},
	year      = {2021},
    doi = {10.1137/19m1253010}
}

@article{Wang2021,
	author    = {Wang, W.},
	title     = {On the inference of applying {G}aussian process modeling to a deterministic function},
	journal   = {Electronic Journal of Statistics},
	volume    = {15},
	number    = {2},
	pages     = {5014--5066},
	year      = {2021},
    doi = {10.1214/21-ejs1912}
}

@article{Chen2021,
	author    = {Chen, Y. and Owhadi, H. and Stuart, A. M.},
	title     = {Consistency of empirical {B}ayes and kernel flow for hierarchical parameter estimation},
	journal   = {Mathematics of Computation},
	volume    = {90},
	number    = {332},
	pages     = {2527--2578},
	year      = {2021},
    doi = {10.1090/mcom/3649}
}

@article{LohSunWen2021,
	author    = {Loh, W.-L. and Sun, S. and Wen, J.},
	title     = {On fixed-domain asymptotics, parameter estimation and isotropic {G}aussian random fields with {M}at{\'e}rn covariance functions},
	journal   = {The Annals of Statistics},
	volume    = {49},
	number    = {6},
	pages     = {3127--3152},
	year      = {2021},
    doi = {10.1214/21-aos2077}
}

@Article{Karvonen2023,
	author    = {Karvonen, Toni},
	title     = {Asymptotic bounds for smoothness parameter estimates in {G}aussian process interpolation},
	journal   = {SIAM/ASA Journal on Uncertainty Quantification},
	volume    = {11},
	number    = {4},
	pages     = {1225--1257},
	year      = {2023},
    doi = {10.1137/22M149288X}
}

@article{Petit2023,
	title     = {An asymptotic study of the joint maximum likelihood estimation of the regularity and the amplitude parameters of a periodized {Mat\'ern} model},
	volume    = {19},
	doi       = {10.1214/25-ejs2380},
	number    = {1},
	journal   = {Electronic Journal of Statistics},
	publisher = {Institute of Mathematical Statistics},
	author    = {Petit,  Sébastien J.},
	year      = {2025},
}

@article{KarvonenOates2023,
	author    = {Toni Karvonen and Chris J. Oates},
	title     = {Maximum likelihood estimation in {G}aussian process regression is ill-posed},
	journal   = {Journal of Machine Learning Research},
	year      = {2023},
	volume    = {24},
	number    = {120},
	pages     = {1--47},
}

@article{ContBas2023,
	author    = {Cont, R. and Bas, P.},
	title     = {Quadratic variation and quadratic roughness},
	journal   = {Bernoulli},
	volume    = {29},
	number    = {1},
	pages     = {496--522},
	year      = {2023},
    doi = {10.3150/22-BEJ1466}
}

@article{LohSun2023,
	author    = {Loh, W.-L. and Sun, S.},
	title     = {Estimating the parameters of some common {G}aussian random fields with nugget under fixed-domain asymptotics},
	journal   = {Bernoulli},
	volume    = {29},
	number    = {3},
	pages     = {2519--2534},
	year      = {2023},
    doi = {10.3150/22-bej1551}
}

@article{10.2307/2959347,
	author    = {R. M. Dudley},
	journal   = {The Annals of Probability},
	number    = {1},
	pages     = {66--103},
	publisher = {Institute of Mathematical Statistics},
	title     = {Sample Functions of the {G}aussian Process},
	volume    = {1},
	year      = {1973},
    doi = {10.1214/aop/1176997026}
}

@article{sniekers2020adaptive,
	title     = {Adaptive {B}ayesian credible bands in regression with a {G}aussian process prior},
	author    = {Sniekers, Suzanne and van der Vaart, Aad},
	journal   = {Sankhya A},
	volume    = {82},
	number    = {2},
	pages     = {386--425},
	year      = {2020},
	publisher = {Springer},
    doi = {10.1007/s13171-019-00185-0}
}

@article{sniekers2015adaptive,
	author    = {Suzanne Sniekers and Aad van der Vaart},
	title     = {Adaptive {B}ayesian credible sets in regression with a {G}aussian process prior},
	volume    = {9},
	journal   = {Electronic Journal of Statistics},
	number    = {2},
	publisher = {Institute of Mathematical Statistics and Bernoulli Society},
	pages     = {2475--2527},
	year      = {2015},
	doi       = {10.1214/15-EJS1078},
}

@article{wang_almost-sure_2007,
	title     = {Almost-sure path properties of fractional {Brownian} sheet},
	volume    = {43},
	doi       = {10.1016/j.anihpb.2006.09.005},
	number    = {5},
	journal   = {Annales de l'Institut Henri Poincare (B) Probability and Statistics},
	author    = {Wang, Wensheng},
	year      = {2007},
	pages     = {619--631},
}

@article{wenzel2021novel,
	title     = {A novel class of stabilized greedy kernel approximation algorithms: convergence, stability and uniform point distribution},
	author    = {Wenzel, Tizian and Santin, Gabriele and Haasdonk, Bernard},
	journal   = {Journal of Approximation Theory},
	volume    = {262},
	year      = {2021},
	publisher = {Elsevier},
    doi = {10.1016/j.jat.2020.105508}
}

@article{karvonen2021estimation,
	title     = {Estimation of the scale parameter for a misspecified {G}aussian process model},
	author    = {Karvonen, Toni},
	journal   = {arXiv:2110.02810},
	year      = {2021}
}

@article{kilicc2008explicit,
	title     = {Explicit formula for the inverse of a tridiagonal matrix by backward continued fractions},
	author    = {K{\i}l{\i}{\c{c}}, Emrah},
	journal   = {Applied Mathematics and Computation},
	volume    = {197},
	number    = {1},
	pages     = {345--357},
	year      = {2008},
	publisher = {Elsevier},
    doi = {10.1016/j.amc.2007.07.046}
}

@article{mallik2001inverse,
	title     = {The inverse of a tridiagonal matrix},
	author    = {Mallik, Ranjan K},
	journal   = {Linear Algebra and its Applications},
	volume    = {325},
	number    = {1--3},
	pages     = {109--139},
	year      = {2001},
	publisher = {Elsevier},
    doi = {10.1016/s0024-3795(00)00262-7}
}

@book{Wahba1990,
	author    = {Wahba, G.},
	title     = {Spline Models for Observational Data},
	publisher = {Society for Industrial and Applied Mathematics},
	series    = {CBMS-NSF Regional Conference Series in Applied Mathematics},
	number    = {59},
	year      = {1990},
    doi = {10.1137/1.9781611970128}
}

@article{ginsbourger2021fast,
	title     = {Fast Calculation of {G}aussian Process Multiple-Fold Cross-Validation Residuals and their Covariances},
	volume    = {34},
	doi       = {10.1080/10618600.2024.2353633},
	number    = {1},
	journal   = {Journal of Computational and Graphical Statistics},
	publisher = {Informa UK Limited},
	author    = {Ginsbourger,  David and Sch\"{a}rer,  Cédric},
	year      = {2024},
	pages     = {1--14}
}

@book{morters2010brownian,
	title     = {Brownian motion},
	author    = {M{\"o}rters, Peter and Peres, Yuval},
	volume    = {30},
	year      = {2010},
	publisher = {Cambridge University Press},
    doi = {10.1017/CBO9780511750489}
}

@article{mahalanobis1936,
	title     = {On the generalized distance in statistics},
	author    = {Mahalanobis, Prasanta Chandra},
	journal   = {Proceedings of the National Institute of
	Sciences of India},
	volume    = {2},
	pages     = {49--55},
	year      = {1936},
}

@article{csiszar1963informationstheoretische,
	title     = {{Eine informationstheoretische Ungleichung und ihre Anwendung auf den Beweis der Ergodizit{\"a}t von Markoffschen Ketten}},
	author    = {Csisz{\'a}r, Imre},
	journal   = {A Magyar Tudom{\'a}nyos Akad{\'e}mia Matematikai Kutat{\'o} Int{\'e}zet{\'e}nek K{\"o}zlem{\'e}nyei},
	volume    = {8},
	number    = {1-2},
	pages     = {85--108},
	year      = {1963},
	publisher = {Akad{\'e}miai Kiad{\'o}}
}

@article{bhattacharyya1943,
	title     = {On a measure of divergence between two statistical populations defined by their probability distributions},
	author    = {Bhattacharyya, Anil Kumar},
	journal   = {Bulletin of the Calcutta Mathematical Society},
	year      = {1943},
	volume    = {35},
	pages     = {99--109}
}

@article{Stute1993,
	author    = {Stute, W. and Manteiga, W. G. and Quindimil, M. P.},
	doi       = {10.1007/BF02613687},
	journal   = {Metrika},
	number    = {1},
	pages     = {243--256},
	title     = {Bootstrap based goodness-of-fit tests},
	volume    = {40},
	year      = {1993}
}

@Article{Drovandi2011,
	author    = {Christopher C. Drovandi and Anthony N. Pettitt},
	journal   = {Computational Statistics {\&} Data Analysis},
	title     = {Likelihood-free {B}ayesian estimation of multivariate quantile distributions},
	year      = {2011},
	number    = {9},
	pages     = {2541--2556},
	volume    = {55},
    doi = {10.1016/j.csda.2011.03.019}
}

@Article{Beaumont2009,
	author    = {M. A. Beaumont and J.-M. Cornuet and J.-M. Marin and C. P. Robert},
	title     = {Adaptive approximate {B}ayesian computation},
	journal   = {Biometrika},
	year      = {2009},
	volume    = {96},
	number    = {4},
	pages     = {983--990},
    doi = {10.1093/biomet/asp052}
}

@Article{Beaumont2010,
	author    = {Mark A. Beaumont},
	title     = {Approximate {B}ayesian Computation in Evolution and Ecology},
	journal   = {Annual Review of Ecology, Evolution, and Systematics},
	year      = {2010},
	volume    = {41},
	number    = {1},
	pages     = {379--406},
	publisher = {Annual Reviews},
    doi = {10.1146/annurev-ecolsys-102209-144621}
}

@Article{Lintusaari2016,
	author    = {Jarno Lintusaari and Michael U. Gutmann and Ritabrata Dutta and Samuel Kaski and Jukka Corander},
	journal   = {Systematic Biology},
	title     = {Fundamentals and Recent Developments in Approximate {B}ayesian Computation},
	year      = {2017},
	volume    = {66},
	pages     = {66--82},
    doi = {10.1093/sysbio/syw077}
}

@article{Akeret_2015,
	year      = 2015,
	volume    = {2015},
	number    = {08},
	pages     = {043--043},
	author    = {Joël Akeret and Alexandre Refregier and Adam Amara and Sebastian Seehars and Caspar Hasner},
	title     = {Approximate {B}ayesian computation for forward modeling in cosmology},
	journal   = {Journal of Cosmology and Astroparticle Physics},
    doi = {10.1088/1475-7516/2015/08/043}
}

@article{Riabiz2020,
	title     = {Optimal Thinning of {MCMC} Output},
	volume    = {84},
	doi       = {10.1111/rssb.12503},
	number    = {4},
	journal   = {Journal of the Royal Statistical Society: Series B},
	publisher = {Oxford University Press (OUP)},
	author    = {Riabiz,  Marina and Chen,  Wilson Ye and Cockayne,  Jon and Swietach,  Pawel and Niederer,  Steven A. and Mackey,  Lester and Oates,  Chris. J.},
	year      = {2022},
	pages     = {1059--1081}
}

@article{Mak2018,
	author    = {Mak, S. and Joseph, V. Roshan},
	journal   = {Annals of Statistics},
	number    = {6A},
	pages     = {2562--2592},
	title     = {Support points},
	volume    = {46},
	year      = {2018},
    doi = {10.1214/17-aos1629}
}

@inproceedings{Bach2012,
	author    = {Bach, F. and Lacoste-Julien, S. and Obozinski, G.},
	booktitle = {Proceedings of the 29th International Conference on Machine Learning},
	pages     = {1355--1362},
	title     = {On the equivalence between herding and conditional gradient algorithms},
	year      = {2012}
}

@article{dick2013high,
	author    = {Dick, Josef and Kuo, Frances Y. and Sloan, Ian H.},
	journal   = {Acta Numerica},
	number    = {April 2013},
	pages     = {133--288},
	title     = {High-dimensional integration: the quasi-{M}onte {C}arlo way},
	volume    = {22},
	year      = {2013},
    doi = {10.1017/s0962492913000044}
}

@InProceedings{Chen2019,
	title     = {Stein Point {M}arkov Chain {M}onte {C}arlo},
	author    = {Chen, Wilson Ye and Barp, Alessandro and Briol, Francois-Xavier and Gorham, Jackson and Girolami, Mark and Mackey, Lester and Oates, Chris},
	booktitle = {Proceedings of the 36th International Conference on Machine Learning},
	pages     = {1011--1021},
	year      = {2019},
	series    = {PMLR 97},
}

@InProceedings{Chen2018,
	title     = {Stein Points},
	author    = {Chen, Wilson Ye and Mackey, Lester and Gorham, Jackson and Briol, Francois-Xavier and Oates, Chris},
	booktitle = {Proceedings of the 35th International Conference on Machine Learning},
	pages     = {844--853},
	year      = {2018},
	series    = {PMLR 80},
}

@inproceedings{Chen2010,
	author    = {Chen, Yutian and Welling, Max and Smola, Alex},
	title     = {Super-samples from kernel herding},
	year      = {2010},
	publisher = {AUAI Press},
	booktitle = {Proceedings of the 26th Conference on Uncertainty in Artificial Intelligence},
	pages     = {109--116},
}

@article{OHagan1991,
	author    = {O'Hagan, A},
	journal   = {Journal of Statistical Planning and Inference},
	pages     = {245--260},
	title     = {{B}ayes-{H}ermite quadrature},
	volume    = {29},
	year      = {1991},
    doi = {10.1016/0378-3758(91)90002-v}
}

@article{Briol2019PI,
	author    = {Briol, F-X. and Oates, C. J. and Girolami, M. and Osborne, M. A. and Sejdinovic, D.},
	journal   = {Statistical Science},
	number    = {1},
	pages     = {1--22},
	title     = {Probabilistic integration: a role in statistical computation? (with discussion)},
	volume    = {34},
	year      = {2019},
    doi = {10.1214/18-STS660}
}

@Article{Bharti2022,
	author    = {Ayush Bharti and Francois-Xavier Briol and Troels Pedersen},
	journal   = {{IEEE} Transactions on Antennas and Propagation},
	title     = {A General Method for Calibrating Stochastic Radio Channel Models With Kernels},
	year      = {2022},
	number    = {6},
	pages     = {3986--4001},
	volume    = {70},
	doi       = {10.1109/tap.2021.3083761},
	publisher = {Institute of Electrical and Electronics Engineers ({IEEE})},
}

@Article{Marin2011,
	author    = {Jean-Michel Marin and Pierre Pudlo and Christian P. Robert and Robin J. Ryder},
	title     = {Approximate {B}ayesian computational methods},
	journal   = {Statistics and Computing},
	year      = {2011},
	number    = {6},
	pages     = {1167--1180},
	volume    = {22},
    doi = {10.1007/s11222-011-9288-2}
}

@Article{Cranmer2020,
	author    = {Kyle Cranmer and Johann Brehmer and Gilles Louppe},
	journal   = {Proceedings of the National Academy of Sciences},
	title     = {The frontier of simulation-based inference},
	year      = {2020},
	number    = {48},
	pages     = {30055--30062},
	volume    = {117},
	publisher = {Proceedings of the National Academy of Sciences},
    doi = {10.1073/pnas.1912789117}
}

@InProceedings{Park2015,
	title     = {{K2-ABC}: approximate {B}ayesian computation with kernel embeddings},
	author    = {Park, Mijung and Jitkrittum, Wittawat and Sejdinovic, Dino},
	booktitle = {Proceedings of the 19th International Conference on Artificial Intelligence and Statistics},
	pages     = {398--407},
	year      = {2016},
	series    = {PMLR 51},
}

@article{Bernton2019,
	author    = {Bernton, E. and Jacob, P. E. and Gerber, M. and Robert, C. P.},
	journal   = {Journal of the Royal Statistical Society: Series B},
	number    = {2},
	pages     = {235--269},
	title     = {Approximate {B}ayesian computation with the {W}asserstein distance},
	volume    = {81},
	year      = {2019},
    doi = {10.1111/rssb.12312}
}

@InProceedings{Jiang2018,
	title     = {Approximate {B}ayesian Computation with {K}ullback-{L}eibler Divergence as Data Discrepancy},
	author    = {Jiang, Bai},
	booktitle = {Proceedings of the 21st International Conference on Artificial Intelligence and Statistics},
	pages     = {1711--1721},
	year      = {2018},
	series    = {PMLR 84},
}

@article{Key2025,
	author    = {Oscar Key and Arthur Gretton and Fran{\c{c}}ois-Xavier Briol and Tamara Fernandez},
	title     = {Composite Goodness-of-fit Tests with Kernels},
	journal   = {Journal of Machine Learning Research},
	year      = {2025},
	volume    = {26},
	number    = {51},
	pages     = {1--60},
}

@InProceedings{Kajihara2018,
	title     = {Kernel Recursive {ABC}: Point Estimation with Intractable Likelihood},
	author    = {Kajihara, Takafumi and Kanagawa, Motonobu and Yamazaki, Keisuke and Fukumizu, Kenji},
	booktitle = {Proceedings of the 35th International Conference on Machine Learning},
	pages     = {2400--2409},
	year      = {2018},
	series    = {PMLR 80},
}

@Article{peyre2019computational,
	author    = {Gabriel Peyr{\'{e}} and Marco Cuturi},
	journal   = {Foundations and Trends{\textregistered} in Machine Learning},
	title     = {Computational Optimal Transport: with Applications to Data Science},
	year      = {2019},
	number    = {5-6},
	pages     = {355--607},
	volume    = {11},
	doi       = {10.1561/2200000073},
	publisher = {Now Publishers},
}

@InProceedings{Genevay2017,
	title     = {Learning Generative Models with {S}inkhorn Divergences},
	author    = {Genevay, Aude and Peyre, Gabriel and Cuturi, Marco},
	booktitle = {Proceedings of the Twenty-First International Conference on Artificial Intelligence and Statistics},
	pages     = {1608--1617},
	year      = {2018},
	series    = {PMLR 84},
}

@InProceedings{Mitrovic2016,
	title     = {{DR-ABC}: approximate {B}ayesian Computation with Kernel-Based Distribution Regression},
	author    = {Mitrovic, Jovana and Sejdinovic, Dino and Teh, Yee-Whye},
	booktitle = {Proceedings of The 33rd International Conference on Machine Learning},
	pages     = {1482--1491},
	year      = {2016},
	series    = {PMLR 48},
}

@InProceedings{Li2015GMMN,
	title     = {Generative Moment Matching Networks},
	author    = {Li, Yujia and Swersky, Kevin and Zemel, Rich},
	booktitle = {Proceedings of the 32nd International Conference on Machine Learning},
	pages     = {1718--1727},
	year      = {2015},
	series    = {PMLR 37},
}

@inproceedings{Binkowski2018,
	author    = {Bi{\'{n}}kowski, M. and Sutherland, D. J. and Arbel, M. and Gretton, A.},
	booktitle = {International Conference on Learning Representations},
	title     = {Demystifying {MMD} {GAN}s},
	year      = {2018}
}

@inproceedings{Gunter2014,
	author    = {Gunter, T. and Garnett, R. and Osborne, M. and Hennig, P. and Roberts, S.},
	booktitle = {Advances in Neural Information Processing Systems},
	pages     = {2789--2797},
	title     = {Sampling for inference in probabilistic models with fast {B}ayesian quadrature},
	year      = {2014}
}

@inproceedings{Bardenet2019,
	author    = {Belhadji, A. and Bardenet, R. and Chainais, P.},
	booktitle = {Advances in Neural Information Processing Systems},
	pages     = {12927--12937},
	title     = {Kernel quadrature with {DPPs}},
	year      = {2019}
}

@inproceedings{Briol2015,
	author    = {Briol, F-X. and Oates, C. J. and Girolami, M. and Osborne, M. A.},
	booktitle = {Advances in Neural Information Processing Systems},
	pages     = {1162--1170},
	title     = {Frank-{W}olfe {B}ayesian quadrature: probabilistic integration with theoretical guarantees},
	year      = {2015}
}

@article{Legramanti2022,
	title     = {Concentration of discrepancy-based approximate {B}ayesian computation via {R}ademacher complexity},
	volume    = {53},
	doi       = {10.1214/24-aos2453},
	number    = {1},
	journal   = {The Annals of Statistics},
	publisher = {Institute of Mathematical Statistics},
	author    = {Legramanti,  Sirio and Durante,  Daniele and Alquier,  Pierre},
	year      = {2025},
}

@inproceedings{cherief2020mmd,
	title     = {{MMD}-{B}ayes: robust {B}ayesian estimation via maximum mean discrepancy},
	author    = {Ch\'erief-Abdellatif, Badr-Eddine and Alquier, Pierre},
	booktitle = {Symposium on Advances in Approximate {B}ayesian Inference},
	pages     = {1--21},
	year      = {2020},
	OPTorganization= {PMLR}
}

@article{Alquier2021,
	title     = {Universal robust regression via maximum mean discrepancy},
	volume    = {111},
	doi       = {10.1093/biomet/asad031},
	number    = {1},
	journal   = {Biometrika},
	publisher = {Oxford University Press (OUP)},
	author    = {Alquier,  P and Gerber,  M},
	year      = {2023},
	pages     = {71--92}
}

@article{cherief2021,
	title     = {Finite sample properties of parametric {MMD} estimation: robustness to misspecification and dependence},
	volume    = {28},
	doi       = {10.3150/21-bej1338},
	number    = {1},
	journal   = {Bernoulli},
	publisher = {Bernoulli Society for Mathematical Statistics and Probability},
	author    = {Ch\'erief-Abdellatif, Badr-Eddine and Alquier,  Pierre},
	year      = {2022},
}

@InProceedings{Dellaporta2022,
	title     = {Robust {B}ayesian Inference for Simulator-based Models via the {MMD} Posterior Bootstrap}, 
	author    = {Dellaporta, Charita and Knoblauch, Jeremias and Damoulas, Theodoros and Briol, Francois-Xavier},
	booktitle = {Proceedings of The 25th International Conference on Artificial Intelligence and Statistics},
	pages     = {943--970},
	year      = {2022},
	series    = {PMLR 151},
}

@article{Nguyen2020,
	author    = {Nguyen, H. D. and Arbel, J. and Lu, H. and Forbes, F.},
	journal   = {IEEE Access},
	pages     = {131683--131698},
	title     = {Approximate {B}ayesian Computation Via the Energy Statistic},
	volume    = {8},
	year      = {2020},
    doi = {10.1109/access.2020.3009878}
}

@inproceedings{Li2017MMDGAN,
	author    = {Li, C.-L. and Chang, W.-C. and Cheng, Y. and Yang, Y. and P{\'{o}}czos, B.},
	booktitle = {Advances in Neural Information Processing Systems},
	pages     = {2203--2213},
	title     = {{MMD} {GAN}: towards deeper understanding of moment matching network},
	year      = {2017}
}

@inproceedings{Dziugaite2015,
	author    = {Dziugaite, Gintare Karolina and Roy, Daniel M. and Ghahramani, Zoubin},
	title     = {Training generative neural networks via maximum mean discrepancy optimization},
year = {2015},
booktitle = {Proceedings of the 31st Conference on Uncertainty in Artificial Intelligence},
	pages     = {258--267}
}

@article{Niederer2019,
	author    = {Niederer, S. A. and Lumens, J. and Trayanova, N. A.},
	doi       = {10.1038/s41569-018-0104-y},
	journal   = {Nature Reviews Cardiology},
	number    = {2},
	pages     = {100--111},
	pmid = {30361497},
	publisher = {Springer US},
	title     = {Computational models in cardiology},
	volume    = {16},
	year      = {2019}
}

@inproceedings{Jennings1999,
	title     = {Agent-Based Computing: promise and Perils},
	author    = {Nicholas R. Jennings},
	booktitle = {Proceedings of the 16th International Joint Conference on Artificial Intelligence},
	year      = {1999},
	pages     = {1429--1436},
}

@article{Pacchiardi2021,
	title     = {Generalized {B}ayesian likelihood-free inference},
	volume    = {18},
	doi       = {10.1214/24-ejs2283},
	number    = {2},
	journal   = {Electronic Journal of Statistics},
	publisher = {Institute of Mathematical Statistics},
	author    = {Pacchiardi,  Lorenzo and Khoo,  Sherman and Dutta,  Ritabrata},
	year      = {2024},
}

@Article{Wiqvist2021,
	author    = {Samuel Wiqvist and Jes Frellsen and Umberto Picchini},
	title     = {Sequential Neural Posterior and Likelihood Approximation},
	year      = {2021},
	journal   = {arXiv:2102.06522},
}

@InProceedings{Lueckmann2021,
	title     = { Benchmarking Simulation-Based Inference },
	author    = {Lueckmann, Jan-Matthis and Boelts, Jan and Greenberg, David and Goncalves, Pedro and Macke, Jakob},
	booktitle = {Proceedings of The 24th International Conference on Artificial Intelligence and Statistics},
	pages     = {343--351},
	year      = {2021},
	series    = {PMLR 130},
}

@Article{Li2017Copula,
	author    = {J. Li and D.J. Nott and Y. Fan and S.A. Sisson},
	journal   = {Computational Statistics {\&} Data Analysis},
	title     = {Extending approximate {B}ayesian computation methods to high dimensions via a {G}aussian copula model},
	year      = {2017},
	pages     = {77--89},
	volume    = {106},
	doi       = {10.1016/j.csda.2016.07.005},
	publisher = {Elsevier {BV}},
}

@book{adams2003sobolev,
	title     = {Sobolev spaces},
	author    = {Adams, Robert A and Fournier, John JF},
	year      = {2003},
	publisher = {Elsevier},
    isbn = {\isbnhref{9780080541297}}
}

@InProceedings{Bharti22a,
	title     = {Approximate {B}ayesian Computation with Domain Expert in the Loop},
	author    = {Bharti, Ayush and Filstroff, Louis and Kaski, Samuel},
	booktitle = {International Conference on Machine Learning},
	pages     = {1893--1905},
	year      = {2022}
}

@article{Kirby2022, 
	title     = {Two-scale interaction of wake and blockage effects in large wind farms}, 
	volume    = {953}, 
	doi       = {10.1017/jfm.2022.979}, 
	journal   = {Journal of Fluid Mechanics}, 
	author    = {Kirby, Andrew and Nishino, Takafumi and Dunstan, Thomas D.}, 
	year      = {2022},
}

@article{Niayifar2016,
	author    = {Niayifar, A. and Port{\'{e}}-Agel, F.},
	journal   = {Energies},
	number    = {9},
	pages     = {1--13},
	title     = {Analytical modelling of wind farms: a new approach for power prediction},
	volume    = {9},
	year      = {2016},
    doi = {10.3390/en9090741}
}

@article{Kirby2023,
	title     = {Data‐driven modelling of turbine wake interactions and flow resistance in large wind farms},
	volume    = {26},
	doi       = {10.1002/we.2851},
	number    = {9},
	journal   = {Wind Energy},
	publisher = {Wiley},
	author    = {Kirby,  Andrew and Briol,  Fran\c{c}ois‐Xavier and Dunstan,  Thomas D. and Nishino,  Takafumi},
	year      = {2023},
	pages     = {968--984}
}

@article{Nishino2016,
	author    = {Nishino, T.},
	journal   = {Journal of Physics: Conference Series},
	number    = {3},
	title     = {Two-scale momentum theory for very large wind farms},
	volume    = {753},
	year      = {2016},
    doi = {10.1088/1742-6596/753/3/032054}
}

@article{Gutmann2016,
	author    = {Michael U. Gutmann and Jukka Corander},
	title     = {{B}ayesian Optimization for Likelihood-Free Inference of Simulator-Based Statistical Models},
	journal   = {Journal of Machine Learning Research},
	year      = {2016},
	volume    = {17},
	number    = {125},
	pages     = {1--47},
}

@InProceedings{Greenberg2019,
	title     = {Automatic Posterior Transformation for Likelihood-Free Inference},
	author    = {Greenberg, David and Nonnenmacher, Marcel and Macke, Jakob},
	booktitle = {Proceedings of the 36th International Conference on Machine Learning},
	pages     = {2404--2414},
	year      = {2019},
	series    = {PMLR 97},
}

@Article{Behrens2015,
	author    = {J. Behrens and F. Dias},
	journal   = {Philosophical Transactions of the Royal Society A: Mathematical, Physical and Engineering Sciences},
	title     = {New computational methods in tsunami science},
	year      = {2015},
	number    = {2053},
	volume    = {373},
	doi       = {10.1098/rsta.2014.0382},
	publisher = {The Royal Society},
}

@article{devore1993besov,
	title     = {Besov spaces on domains in $\mathbb{R}^d$},
	author    = {DeVore, Ronald A and Sharpley, Robert C},
	journal   = {Transactions of the American Mathematical Society},
	volume    = {335},
	number    = {2},
	pages     = {843--864},
	year      = {1993},
    doi = {10.2307/2154408}
}

@book{stein1970singular,
	title     = {Singular integrals and differentiability properties of functions},
	author    = {Stein, Elias M},
	volume    = {2},
	year      = {1970},
	publisher = {Princeton University Press},
    doi = {10.1515/9781400883882}
}

@article{DREAM,
	title     = {{DREAM}: a fluid-kinetic framework for tokamak disruption runaway electron simulations},
	journal   = {Computer Physics Communications},
	volume    = {268},
	year      = {2021},
	doi       = {10.1016/j.cpc.2021.108098},
	author    = {Mathias Hoppe and Ola Embreus and Tünde Fülöp}
}

@article{Teckentrup2020,
	author    = {Teckentrup, Aretha L.},
	journal   = {SIAM-ASA Journal on Uncertainty Quantification},
	number    = {4},
	pages     = {1310--1337},
	title     = {Convergence of {G}aussian process regression with estimated hyperparameters and applications in {B}ayesian inverse problems},
	volume    = {8},
	year      = {2020},
    doi = {10.1137/19m1284816}
}

@article{Constantine1996,
	author    = {Constantine, G. M. and Savits, Thomas H.},
	year      = {1996},
	title     = {A multivariate {F}a\'a di {B}runo formula with applications},
	pages     = {503--520},
	volume    = {348},
	number    = {2},
	journal   = {Transactions of the American Mathematical Society},
    doi = {10.1090/s0002-9947-96-01501-2}
}

@book{evans2018measure,
	title     = {Measure theory and fine properties of functions},
	author    = {Evans, Lawrence C and Garzepy, Ronald F},
	year      = {2018},
	publisher = {Routledge},
    doi = {10.1201/9781003583004}
}

@ARTICLE{SciPy,
	author    = {Virtanen, Pauli and Gommers, Ralf and Oliphant, Travis E. and
	Haberland, Matt and Reddy, Tyler and Cournapeau, David and
	Burovski, Evgeni and Peterson, Pearu and Weckesser, Warren and
	Bright, Jonathan and {van der Walt}, St{\'e}fan J. and
	Brett, Matthew and Wilson, Joshua and Millman, K. Jarrod and
	Mayorov, Nikolay and Nelson, Andrew R. J. and Jones, Eric and
	Kern, Robert and Larson, Eric and Carey, C J and
	Polat, {\.I}lhan and Feng, Yu and Moore, Eric W. and
	{VanderPlas}, Jake and Laxalde, Denis and Perktold, Josef and
	Cimrman, Robert and Henriksen, Ian and Quintero, E. A. and
	Harris, Charles R. and Archibald, Anne M. and
	Ribeiro, Ant{\^o}nio H. and Pedregosa, Fabian and
	{van Mulbregt}, Paul and {SciPy 1.0 Contributors}},
	title     = {{SciPy} 1.0: fundamental Algorithms for Scientific
	Computing in {P}ython},
	journal   = {Nature Methods},
	year      = {2020},
	volume    = {17},
	pages     = {261--272},
	doi       = {10.1038/s41592-019-0686-2},
}

@article{aronszajn1950theory,
	title     = {Theory of reproducing kernels},
	author    = {Aronszajn, Nachman},
	journal   = {Transactions of the American Mathematical Society},
	volume    = {68},
	number    = {3},
	pages     = {337--404},
	year      = {1950},
    doi = {10.21236/ada296533}
}

@article{mahalanobis1930,
	author    = {Mahalanobis, Prasanta Chandra},
	title     = {On tests and measures of group divergence},
	journal   = {J. Asiat. Soc. Bengal},
	year      = {1930},
	volume    = {26},
	pages     = {541--588},
}

@book{dudley2002,
	title     = {Real Analysis and Probability},
	doi       = {10.1017/cbo9780511755347},
	publisher = {Cambridge University Press},
	author    = {Dudley,  R. M.},
	year      = {2002},
}

@inproceedings{rubenstein2019practical,
	title     = {Practical and consistent estimation of f-divergences},
	author    = {Rubenstein, Paul and Bousquet, Olivier and Djolonga, Josip and Riquelme, Carlos and Tolstikhin, Ilya O},
	booktitle = {Advances in Neural Information Processing Systems},
	volume    = {32},
	year      = {2019},
	pages     = {4070--4080}
}

@InProceedings{mcallester2020formal,
	title     = {Formal limitations on the measurement of mutual information},
	author    = {McAllester, David and Stratos, Karl},
	booktitle = {Proceedings of the 23rd International Conference on Artificial Intelligence and Statistics},
	pages     = {875--884},
	year      = {2020},
	series    = {PMLR 108},
}

@article{nguyen2010estimating,
	title     = {Estimating divergence functionals and the likelihood ratio by convex risk minimization},
	author    = {Nguyen, XuanLong and Wainwright, Martin J and Jordan, Michael I},
	journal   = {IEEE Transactions on Information Theory},
	volume    = {56},
	number    = {11},
	pages     = {5847--5861},
	year      = {2010},
	publisher = {IEEE},
    doi = {10.1109/tit.2010.2068870}
}

@article{sommariva2006numerical,
	title     = {Numerical cubature on scattered data by radial basis functions},
	author    = {Sommariva, Alvise and Vianello, Marco},
	journal   = {Computing},
	volume    = {76},
	pages     = {295--310},
	year      = {2006},
	publisher = {Springer},
    doi = {10.1007/s00607-005-0142-2}
}

@inproceedings{Rasmussen2003,
	author    = {Rasmussen, Carl and Ghahramani, Zoubin},
	booktitle = {Advances in Neural Information Processing Systems},
	pages     = {489--496},
	title     = {{B}ayesian {M}onte {C}arlo},
	year      = {2002}
}

@article{Lin2025,
	title     = {{BASIL}: fast broadband line-rich spectral-cube fitting and image visualization via {B}ayesian quadrature},
	volume    = {700},
	doi       = {10.1051/0004-6361/202452828},
	journal   = {Astronomy and Astrophysics},
	publisher = {EDP Sciences},
	author    = {Lin,  Y. and Adachi,  M. and Spezzano,  S. and Edenhofer,  G. and Eberle,  V. and Osborne,  M. A. and Caselli,  P.},
	year      = {2025}
}

@inproceedings{dlr201186,
	year      = {2023},
	booktitle = {OBMS 2023},
	title     = {Automating The Selection Of Battery Models With {B}ayesian Quadrature And {B}ayesian Optimization},
	author    = {Kuhn, Yannick and Horstmann, Birger and Latz, Arnulf},
	}

@article{Osborne2012,
	title     = {Real-time information processing of environmental sensor network data using {B}ayesian {G}aussian processes},
	volume    = {9},
	doi       = {10.1145/2379799.2379800},
	number    = {1},
	journal   = {ACM Transactions on Sensor Networks},
	publisher = {Association for Computing Machinery (ACM)},
	author    = {Osborne,  Michael A. and Roberts,  Stephen J. and Rogers,  Alex and Jennings,  Nicholas R.},
	year      = {2012},
	pages     = {1--32}
}

@article{COUSIN2024102483,
	title     = {Optimal design of experiments for computing the fatigue life of an offshore wind turbine based on stepwise uncertainty reduction},
	journal   = {Structural Safety},
	volume    = {110},
	year      = {2024},
	doi       = {10.1016/j.strusafe.2024.102483},
	author    = {Alexis Cousin and Nicolas Delépine and Martin Guiton and Miguel {Munoz Zuniga} and Timothée Perdrizet},
}

@book{Hennig2022,
	author    = {Hennig, Philipp and Osborne, Michael A. and Kersting, Hans},
	publisher = {Cambridge University Press},
	title     = {Probabilistic Numerics: computation as Machine Learning},
	year      = {2022},
    doi = {doi.org/10.1017/9781316681411.006}
}

@book{ritter2000,
	author    = {Klaus Ritter},
	title     = {Average-Case Analysis of Numerical Problems},
	doi       = {10.1007/bfb0103934},
	journal   = {Lecture Notes in Mathematics},
	publisher = {Springer Berlin Heidelberg},
	year      = {2000}
}

@article{Sriperumbudur2012,
	title     = {On the empirical estimation of integral probability metrics},
	volume    = {6},
	doi       = {10.1214/12-ejs722},
	journal   = {Electronic Journal of Statistics},
	publisher = {Institute of Mathematical Statistics},
	author    = {Sriperumbudur,  Bharath K. and Fukumizu,  Kenji and Gretton,  Arthur and Sch\"{o}lkopf,  Bernhard and Lanckriet,  Gert R. G.},
	year      = {2012},
}

@article{simon2023metrizing,
	title     = {Metrizing weak convergence with maximum mean discrepancies},
	author    = {Simon-Gabriel, Carl-Johann and Barp, Alessandro and Sch{\"o}lkopf, Bernhard and Mackey, Lester},
	journal   = {Journal of Machine Learning Research},
	volume    = {24},
	number    = {184},
	pages     = {1--20},
	year      = {2023},
}

@phdthesis{Tang2013,
	author    = {Tang, Xiaojin},
	school    = {Boston University},
	title     = {Importance sampling for efficient parametric simulation},
	year      = {2013}
}

@article{Giles2015,
	title     = {Multilevel {M}onte {C}arlo approximation of distribution functions and densities},
	author    = {Giles, Michael B and Nagapetyan, Tigran and Ritter, Klaus},
	journal   = {SIAM/ASA journal on Uncertainty Quantification},
	volume    = {3},
	number    = {1},
	pages     = {267--295},
	year      = {2015},
	publisher = {SIAM},
    doi = {10.1137/140960086}
}

@article{Krumscheid2018,
	title     = {Multilevel {M}onte {C}arlo approximation of functions},
	author    = {Krumscheid, Sebastian and Nobile, Fabio},
	journal   = {SIAM/ASA Journal on Uncertainty Quantification},
	volume    = {6},
	number    = {3},
	pages     = {1256--1293},
	year      = {2018},
	publisher = {SIAM},
    doi = {Krumscheid2018}
}

@article{longstaff2001valuing,
	title     = {Valuing {A}merican options by simulation: a simple least-squares approach},
	author    = {Longstaff, Francis A and Schwartz, Eduardo S},
	journal   = {The Review of Financial Studies},
	volume    = {14},
	number    = {1},
	pages     = {113--147},
	year      = {2001},
	publisher = {Oxford University Press},
    doi = {10.1093/rfs/14.1.113}
}

@article{alfonsi2022many,
	title     = {How Many Inner Simulations to Compute Conditional Expectations with Least-square {M}onte {C}arlo?},
	volume    = {25},
	doi       = {10.1007/s11009-023-10038-x},
	number    = {3},
	journal   = {Methodology and Computing in Applied Probability},
	publisher = {Springer Science and Business Media LLC},
	author    = {Alfonsi,  Aurélien and Lapeyre,  Bernard and Lelong,  Jér\^ome},
	year      = {2023},
}

@article{Lopes2011,
	author    = {Lopes, Hedibert F. and Tobias, Justin L.},
	journal   = {Annual Review of Economics},
	pages     = {107--131},
	title     = {Confronting prior convictions: on issues of prior sensitivity and likelihood robustness in {B}ayesian analysis},
	volume    = {3},
	year      = {2011},
    doi = {10.1146/annurev-economics-111809-125134}
}

@article{Kallioinen2021,
	title     = {Detecting and diagnosing prior and likelihood sensitivity with power-scaling},
	volume    = {34},
	doi       = {10.1007/s11222-023-10366-5},
	number    = {1},
	journal   = {Statistics and Computing},
	publisher = {Springer Science and Business Media LLC},
	author    = {Kallioinen,  Noa and Paananen,  Topi and B\"{u}rkner,  Paul-Christian and Vehtari,  Aki},
	year      = {2023},
}

@article{Sobol2001,
	title     = {Global sensitivity indices for nonlinear mathematical models and their {M}onte {C}arlo estimates},
	author    = {Sobol, Ilya M},
	journal   = {Mathematics and Computers in Simulation},
	volume    = {55},
	number    = {1-3},
	pages     = {271--280},
	year      = {2001},
	publisher = {Elsevier},
    doi = {10.1016/s0378-4754(00)00270-6}
}

@inproceedings{Hong2009,
	title     = {Estimating the mean of a non-linear function of conditional expectation},
	author    = {Hong, L Jeff and Juneja, Sandeep},
	booktitle = {Winter Simulation Conference},
	pages     = {1223--1236},
	year      = {2009},
    doi = {10.1109/wsc.2009.5429428}
}

@article{heath2017review,
	title     = {A review of methods for analysis of the expected value of information},
	author    = {Heath, Anna and Manolopoulou, Ioanna and Baio, Gianluca},
	journal   = {Medical Decision Making},
	volume    = {37},
	number    = {7},
	pages     = {747--758},
	year      = {2017},
	publisher = {Sage Publications Sage CA: Los Angeles, CA},
    doi = {10.1177/0272989x17697692}
}

@article{Glynn1989,
	author    = {Glynn, Peter and Igelhart, Donald},
	journal   = {Management Science},
	number    = {1367-1392},
	title     = {Importance sampling for stochastic simulations},
	volume    = {35},
	year      = {1989},
    doi = {10.1287/mnsc.35.11.1367}
}

@article{Madras1999,
	title     = {Importance sampling for families of distributions},
	author    = {Madras, Neal and Piccioni, Mauro},
	journal   = {The Annals of Applied Probability},
	volume    = {9},
	number    = {4},
	pages     = {1202--1225},
	year      = {1999},
	publisher = {Institute of Mathematical Statistics},
    doi = {10.1214/aoap/1029962870}
}

@article{Demange-Chryst2022,
	title     = {Efficient estimation of multiple expectations with the same sample by adaptive importance sampling and control variates},
	volume    = {33},
	doi       = {10.1007/s11222-023-10270-y},
	number    = {5},
	journal   = {Statistics and Computing},
	publisher = {Springer Science and Business Media LLC},
	author    = {Demange-Chryst,  Julien and Bachoc,  Fran\c{c}ois and Morio,  Jér\^ome},
	year      = {2023},
}

@book{Robert2004,
	title     = {{M}onte {C}arlo statistical methods},
	author    = {Robert, Christian P and Casella, George and Casella, George},
	volume    = {2},
	year      = {1999},
	publisher = {Springer},
    doi = {10.1007/978-1-4757-4145-2}
}

@article{Han2009,
	author    = {Han, G. S. and Kim, B. H. and Lee, J.},
	doi       = {10.1016/j.eswa.2008.05.004},
	journal   = {Expert Systems with Applications},
	number    = {3},
	pages     = {4431--4436},
	publisher = {Elsevier Ltd},
	title     = {Kernel-based {M}onte {C}arlo simulation for {A}merican option pricing},
	volume    = {36},
	year      = {2009}
}

@article{Hu2020,
	author    = {Hu, W. and Zastawniak, T.},
	journal   = {Quantitative Finance},
	number    = {5},
	pages     = {851--865},
	publisher = {Routledge},
	title     = {Pricing high-dimensional {A}merican options by kernel ridge regression},
	volume    = {20},
	year      = {2020},
    doi = {10.1080/14697688.2020.1713393}
}

@article{muandet2016kernelmeanshrinkage,
	title     = {Kernel mean shrinkage estimators},
	author    = {Muandet, Krikamol and Sriperumbudur, Bharath and Fukumizu, Kenji and Gretton, Arthur and Sch{\"o}lkopf, Bernhard},
	journal   = {Journal of Machine Learning Research},
	volume    = {17},
	year      = {2016},
}

@InProceedings{xi2018bayesian,
	title     = {{B}ayesian quadrature for multiple related integrals},
	author    = {Xi, Xiaoyue and Briol, Francois-Xavier and Girolami, Mark},
	booktitle = {Proceedings of the 35th International Conference on Machine Learning},
	pages     = {5373--5382},
	year      = {2018},
	series    = {PMLR 80},
}

@InProceedings{gessner2020active,
	title     = {Active multi-information source {B}ayesian quadrature},
	author    = {Gessner, Alexandra and Gonzalez, Javier and Mahsereci, Maren},
	booktitle = {Proceedings of The 35th Uncertainty in Artificial Intelligence Conference},
	pages     = {712--721},
	year      = {2020},
	series    = {PMLR 115},
}

@inproceedings{Sun2021,
	author    = {Sun, Zhuo and Barp, Alessandro and Briol, Fran{\c{c}}ois-Xavier},
	booktitle = {International Conference on Machine Learning},
	pages     = {32819--32846},
	title     = {Vector-valued control variates},
	year      = {2023}
}

@inproceedings{Sun2023,
	author    = {Sun, Zhuo and Oates, Chris J and Briol, Fran{\c{c}}ois-Xavier},
	booktitle = {Conference on Uncertainty in Artificial Intelligence},
	pages     = {2047--2057},
	title     = {Meta-learning control variates: variance reduction with limited data},
	year      = {2023}
}

@article{Ababou1994,
	title     = {On the condition number of covariance matrices in kriging, estimation, and simulation of random fields},
	author    = {Ababou, Rachid and Bagtzoglou, Amvrossios C and Wood, Eric F},
	journal   = {Mathematical Geology},
	volume    = {26},
	pages     = {99--133},
	year      = {1994},
	publisher = {Springer},
	doi       = {10.1007/BF02065878}
}

@article{Andrianakis2012,
	title     = {The effect of the nugget on {G}aussian process emulators of computer models},
	author    = {Andrianakis, Ioannis and Challenor, Peter G},
	journal   = {Computational Statistics \& Data Analysis},
	volume    = {56},
	number    = {12},
	pages     = {4215--4228},
	year      = {2012},
	publisher = {Elsevier},
    doi = {10.1016/j.csda.2012.04.020 }
}

@inproceedings{li2022multilevel,
	title     = {Multilevel {B}ayesian Quadrature},
	author    = {Li, Kaiyu and Giles, Daniel and Karvonen, Toni and Guillas, Serge and Briol, Fran{\c{c}}ois-Xavier},
	booktitle = {International Conference on Artificial Intelligence and Statistics},
	pages     = {1845--1868},
	year      = {2022}
}

@inproceedings{Kanagawa2019,
	title     = {Convergence guarantees for adaptive {B}ayesian quadrature methods},
	author    = {Kanagawa, Motonobu and Hennig, Philipp},
	booktitle = {Advances in Neural Information Processing Systems},
	volume    = {32},
	year      = {2019},
	pages     = {6237--6248}
}

@article{Marques2013,
	title     = {A spherical {G}aussian framework for {B}ayesian {M}onte {C}arlo rendering of glossy surfaces},
	author    = {Marques, Ricardo and Bouville, Christian and Ribardi{\`e}re, Micka{\"e}l and Santos, Lu{\'\i}s Paulo and Bouatouch, Kadi},
	journal   = {IEEE Transactions on Visualization and Computer Graphics},
	volume    = {19},
	number    = {10},
	pages     = {1619--1632},
	year      = {2013},
	publisher = {IEEE},
    doi = {10.1109/tvcg.2013.79 }
}

@inproceedings{Oates2017heart,
	title     = {Probabilistic models for integration error in the assessment of functional cardiac models},
	author    = {Oates, Chris J. and Niederer, Steven and Lee, Angela and Briol, Fran{\c{c}}ois-Xavier and Girolami, Mark},
	booktitle = {Advances in Neural Information Processing Systems},
	volume    = {30},
	year      = {2017},
	pages     = {110--118}
}

@article{Karvonen2019,
	title     = {Symmetry exploits for {B}ayesian cubature methods},
	volume    = {29},
	doi       = {10.1007/s11222-019-09896-8},
	number    = {6},
	journal   = {Statistics and Computing},
	publisher = {Springer Science and Business Media LLC},
	author    = {Karvonen,  Toni and S\"{a}rkk\"{a},  Simo and Oates,  Chris. J.},
	year      = {2019},
	pages     = {1231--1248}
}

@article{Karvonen2017symmetric,
	title     = {Fully symmetric kernel quadrature},
	author    = {Karvonen, Toni and S{\"a}rkk{\"a}, Simo},
	journal   = {SIAM Journal on Scientific Computing},
	volume    = {40},
	number    = {2},
	pages     = {697--720},
	year      = {2018},
	publisher = {SIAM},
    doi = {10.1137/17m1121779}
}

@article{Jagadeeswaran2018,
	title     = {Fast automatic {B}ayesian cubature using lattice sampling},
	author    = {Jagadeeswaran, Rathinavel and Hickernell, Fred J},
	journal   = {Statistics and Computing},
	volume    = {29},
	number    = {6},
	pages     = {1215--1229},
	year      = {2019},
	publisher = {Springer},
    doi = {10.1007/s11222-019-09895-9}
}

@InProceedings{Hayakawa2023,
	title     = {Sampling-based {N}ystr\"om approximation and kernel quadrature},
	author    = {Hayakawa, Satoshi and Oberhauser, Harald and Lyons, Terry},
	booktitle = {Proceedings of the 40th International Conference on Machine Learning},
	pages     = {12678--12699},
	year      = {2023},
	series    = {PMLR 202},
}

@inproceedings{Adachi2022,
	author    = {Adachi, M. and Hayakawa, S. and Oberhauser, H. and Jorgensen, M. and Osborne, M.A.},
	booktitle = {Advances in Neural Information Processing Systems},
	title     = {Fast {B}ayesian inference with batch {B}ayesian quadrature via kernel recombination},
	year      = {2022},
	pages     = {16533--16547}
}

@inproceedings{Zhu2020,
	author    = {Zhu, Harrison and Liu, Xing and Kang, Ruya and Shen, Zhichao and Flaxman, Seth and Briol, Fran\c{c}ois-Xavier},
	booktitle = {Advances in Neural Information Processing Systems},
	pages     = {5837--5849},
	title     = {{B}ayesian probabilistic numerical integration with tree-based models},
	year      = {2020}
}

@inproceedings{Ott2023,
	author    = {Ott, Katharina and Tiemann, Michael and Hennig, Philipp and Briol, Fran\c{c}ois-Xavier},
	booktitle = {Proceedings of the 39th Conference on Uncertainty in Artificial Intelligence},
	pages     = {1606--1617},
	title     = {{B}ayesian numerical integration with neural networks},
	year      = {2023},
    series = {PMLR 216}
}

@inproceedings{Le2005,
	title     = {Heteroscedastic {G}aussian process regression},
	author    = {Le, Quoc V and Smola, Alex J and Canu, St{\'e}phane},
	booktitle = {Proceedings of the 22nd International Conference on Machine Learning},
	pages     = {489--496},
	year      = {2005},
    doi = {10.1145/1102351.1102413}
}

@article{Fabozzi2017,
	author    = {Fabozzi, F. J. and Paletta, T. and Tunaru, R.},
	journal   = {European Journal of Operational Research},
	number    = {2},
	pages     = {698--706},
	publisher = {Elsevier B.V.},
	title     = {An improved least squares {M}onte {C}arlo valuation method based on heteroscedasticity},
	volume    = {263},
	year      = {2017},
    doi = {10.1016/j.ejor.2017.05.048}
}

@InProceedings{titsias2009variational,
	title     = {Variational learning of inducing variables in sparse {G}aussian processes},
	author    = {Titsias, Michalis},
	booktitle = {Proceedings of the 12th International Conference on Artificial Intelligence and Statistics},
	pages     = {567--574},
	year      = {2009},
	series    = {PMLR 5},
}

@article{kanagawa2020convergence,
	title     = {Convergence analysis of deterministic kernel-based quadrature rules in misspecified settings},
	author    = {Kanagawa, Motonobu and Sriperumbudur, Bharath K and Fukumizu, Kenji},
	journal   = {Foundations of Computational Mathematics},
	volume    = {20},
	pages     = {155--194},
	year      = {2020},
	publisher = {Springer},
    doi = {10.1007/s10208-018-09407-7}
}

@article{gogolashvili2023importance,
	title     = {When is Importance Weighting Correction Needed for Covariate Shift Adaptation?},
	author    = {Gogolashvili, Davit and Zecchin, Matteo and Kanagawa, Motonobu and Kountouris, Marios and Filippone, Maurizio},
	journal   = {arXiv:2303.04020},
	year      = {2023}
}

@article{Garreau2018,
	author    = {Garreau,  Damien and Jitkrittum,  Wittawat and Kanagawa,  Motonobu},
	title     = {Large sample analysis of the median heuristic},
	journal   = {arXiv:1707.07269},
	year      = {2017},
}

@incollection{frazier2018bayesian,
	title     = {{B}ayesian optimization},
	author    = {Frazier, Peter I},
	booktitle = {Recent advances in optimization and modeling of contemporary problems},
	pages     = {255--278},
	year      = {2018},
	publisher = {Informs},
    doi = {10.1287/educ.2018.0188}
}

@article{Stone1982,
	author    = {Stone, C. J.},
	journal   = {The Annals of Statistics},
	number    = {4},
	pages     = {1040--1053},
	title     = {Optimal global rates of convergence for nonparametric regression},
	volume    = {10},
	year      = {1982},
    doi = {10.1214/aos/1176345969}
}

@article{stentoft2004convergence,
	title     = {Convergence of the least squares {M}onte {C}arlo approach to {A}merican option valuation},
	author    = {Stentoft, Lars},
	journal   = {Management Science},
	volume    = {50},
	number    = {9},
	pages     = {1193--1203},
	year      = {2004},
	publisher = {INFORMS},
    doi = {10.1287/mnsc.1030.0155}
}

@book{novak1988deterministic,
	title     = {Deterministic and stochastic error bounds in numerical analysis},
	author    = {Novak, Erich},
	volume    = {1349},
	year      = {1988},
	publisher = {Springer},
    doi = {10.1007/bfb0079792}
}

@InProceedings{Rainforth2018,
  title = 	 {On Nesting {M}onte {C}arlo Estimators},
  author =       {Rainforth, Tom and Cornish, Rob and Yang, Hongseok and Warrington, Andrew and Wood, Frank},
  booktitle = 	 {Proceedings of the 35th International Conference on Machine Learning},
  pages = 	 {4267--4276},
  year = 	 {2018},
  series = 	 {PMLR 80}
}

@article{Chaloner1995,
	title     = {{B}ayesian experimental design: a review},
	author    = {Chaloner, Kathryn and Verdinelli, Isabella},
	journal   = {Statistical Science},
	pages     = {273--304},
	year      = {1995},
	publisher = {JSTOR},
    doi = {10.1214/ss/1177009939}
}

@article{Nishiyama2016,
	author    = {Nishiyama, Yu and Fukumizu, Kenji},
	journal   = {Journal of Machine Learning Research},
	number    = {180},
	pages     = {1--28},
	title     = {Characteristic kernels and infinitely divisible distributions},
	volume    = {17},
	year      = {2016},
}

@article{nishiyama2020model,
	title     = {Model-based kernel sum rule: kernel {B}ayesian inference with probabilistic models},
	author    = {Nishiyama, Yu and Kanagawa, Motonobu and Gretton, Arthur and Fukumizu, Kenji},
	journal   = {Machine Learning},
	volume    = {109},
	number    = {5},
	pages     = {939--972},
	year      = {2020},
	publisher = {Springer},
    doi = {10.1007/s10994-019-05852-9}
}

@article{anastasiou2023stein,
	title     = {{S}tein's method meets computational statistics: a review of some recent developments},
	author    = {Anastasiou, Andreas and Barp, Alessandro and Briol, Fran{\c{c}}ois-Xavier and Ebner, Bruno and Gaunt, Robert E and Ghaderinezhad, Fatemeh and Gorham, Jackson and Gretton, Arthur and Ley, Christophe and Liu, Qiang and and Mackey, Lester and Oates, Chris J and Reinert, Gesine and Swan, Yvik},
	journal   = {Statistical Science},
	volume    = {38},
	number    = {1},
	pages     = {120--139},
	year      = {2023},
	publisher = {Institute of Mathematical Statistics},
    doi = {10.1214/22-sts863}
}

@article{kanagawa2025gaussian,
	title     = {{G}aussian Processes and Reproducing Kernels: connections and Equivalences},
	author    = {Kanagawa, Motonobu and Hennig, Philipp and Sejdinovic, Dino and Sriperumbudur, Bharath K},
	journal   = {arXiv:2506.17366},
	year      = {2025}
}

@article{bach2017equivalence,
	title     = {On the equivalence between kernel quadrature rules and random feature expansions},
	author    = {Bach, Francis},
	journal   = {Journal of Machine Learning Research},
	volume    = {18},
	number    = {21},
	pages     = {1--38},
	year      = {2017},
}

@inproceedings{gretton2012optimal,
	title     = {Optimal kernel choice for large-scale two-sample tests},
	author    = {Gretton, Arthur and Sejdinovic, Dino and Strathmann, Heiko and Balakrishnan, Sivaraman and Pontil, Massimiliano and Fukumizu, Kenji and Sriperumbudur, Bharath K},
	booktitle = {Advances in Neural Information Processing Systems},
	volume    = {25},
	year      = {2012},
	pages     = {1205--1213},
}

@article{oakley2004probabilistic,
	title     = {Probabilistic sensitivity analysis of complex models: a {B}ayesian approach},
	author    = {Oakley, Jeremy E and O'Hagan, Anthony},
	journal   = {Journal of the Royal Statistical Society: Series B},
	volume    = {66},
	number    = {3},
	pages     = {751--769},
	year      = {2004},
	publisher = {Wiley Online Library},
    doi = {10.1111/j.1467-9868.2004.05304.x}
}

@article{kermack1927sir,
	title     = {A contribution to the mathematical theory of epidemics},
	author    = {Kermack, William Ogilvy and McKendrick, Anderson G},
	journal   = {Journal of the Royal Statistical Society: Series A},
	volume    = {115},
	number    = {772},
	pages     = {700--721},
	year      = {1927},
	publisher = {The Royal Society London},
    doi = {10.1098/rspa.1927.0118}
}

@book{achdou2005computational,
	title     = {Computational methods for option pricing},
	author    = {Achdou, Yves and Pironneau, Olivier},
	year      = {2005},
	publisher = {SIAM},
    doi = {10.1137/1.9780898717495}
}

@article{alfonsi2021multilevel,
	title     = {Multilevel {M}onte {C}arlo for computing the {SCR} with the standard formula and other stress tests},
	author    = {Alfonsi, Aur{\'e}lien and Cherchali, Adel and Acevedo, Jose Arturo Infante},
	journal   = {Insurance: Mathematics and Economics},
	volume    = {100},
	pages     = {234--260},
	year      = {2021},
	publisher = {Elsevier},
    doi = {10.1016/j.insmatheco.2021.05.005}
}

@article{brennan2007calculating,
	title     = {Calculating partial expected value of perfect information via {M}onte {C}arlo sampling algorithms},
	author    = {Brennan, Alan and Kharroubi, Samer and O'Hagan, Anthony and Chilcott, Jim},
	journal   = {Medical Decision Making},
	volume    = {27},
	number    = {4},
	pages     = {448--470},
	year      = {2007},
	publisher = {Sage Publications Sage CA: Los Angeles, CA},
    doi = {10.1177/0272989x07302555}
}

@article{Giles2019,
	title     = {Decision-making under uncertainty: using {MLMC} for efficient estimation of {EVPPI}},
	author    = {Giles, Michael B and Goda, Takashi},
	journal   = {Statistics and Computing},
	volume    = {29},
	pages     = {739--751},
	year      = {2019},
	publisher = {Springer},
    doi = {10.1007/s11222-018-9835-1}
}

@article{oates2019convergence,
	title     = {Convergence rates for a class of estimators based on {S}tein's method},
	author    = {Oates, Chris J. and Cockayne, Jon and Briol, Fran{\c{c}}ois-Xavier and Girolami, Mark},
	year      = {2019},
	number    = {2},
	pages     = {1141--1159},
	volume    = {25},
	journal   = {Bernoulli},
    doi = {10.3150/17-bej1016}
}

@article{caponnetto2007optimal,
	title     = {Optimal rates for the regularized least-squares algorithm},
	author    = {Caponnetto, Andrea and De Vito, Ernesto},
	journal   = {Foundations of Computational Mathematics},
	volume    = {7},
	pages     = {331--368},
	year      = {2007},
	publisher = {Springer},
    doi = {10.1007/s10208-006-0196-8}
}

@book{fremlin2000measure,
	title     = {Measure theory},
	author    = {Fremlin, David Heaver},
	volume    = {4},
	year      = {2000},
	publisher = {Torres Fremlin},
    isbn      = {\isbnhref{9780953812943}}
}

@misc{jax2018github,
	author    = {James Bradbury and Roy Frostig and Peter Hawkins and Matthew James Johnson and Chris Leary and Dougal Maclaurin and George Necula and Adam Paszke and Jake Vander{P}las and Skye Wanderman-{M}ilne and Qiao Zhang},
	title     = {{JAX}: composable transformations of {P}ython+{N}um{P}y programs},
	version = {0.3.13},
	year      = {2018},
}

@Book{bishop:2006:PRML,
	author    = {Christopher M. Bishop},
	title     = {Pattern Recognition and Machine Learning},
	publisher = {Springer},
	year      = {2006},
    doi = {10.1007/978-0-387-45528-0}
}

@article{ming2021linked,
	title     = {Linked {G}aussian Process Emulation for Systems of Computer Models using {M}at{\'e}rn Kernels and Adaptive Design},
	volume    = {9},
	doi       = {10.1137/20m1323771},
	number    = {4},
	journal   = {SIAM/ASA Journal on Uncertainty Quantification},
	publisher = {Society for Industrial & Applied Mathematics (SIAM)},
	author    = {Ming,  Deyu and Guillas,  Serge},
	year      = {2021},
	pages     = {1615--1642}
}

@book{schwabik2005topics,
	title     = {Topics in {B}anach space integration},
	author    = {Schwabik, Stefan and Ye, Guoju},
	volume    = {10},
	year      = {2005},
	publisher = {World Scientific},
    doi = {10.1142/9789812703286}
}

@article{salicru1994applications,
	title     = {On the applications of divergence type measures in testing statistical hypotheses},
	author    = {Salicr{\'u}, M and Morales, D and Men{\'e}ndez, ML and Pardo, L},
	journal   = {Journal of Multivariate Analysis},
	volume    = {51},
	number    = {2},
	pages     = {372--391},
	year      = {1994},
	publisher = {Elsevier},
    doi = {10.1006/jmva.1994.1068}
}

@inproceedings{sejdinovic2012hypothesis, 
	author    = {Sejdinovic, Dino and Gretton, Arthur and Sriperumbudur, Bharath and Fukumizu, Kenji}, 
	title     = {Hypothesis testing using pairwise distances and associated kernels}, 
	year      = {2012}, 
	booktitle = {Proceedings of the 29th International Conference on Machine Learning}, 
	pages     = {787--794}, 
}

@book{basu2011statistical,
	title     = {Statistical inference: the minimum distance approach},
	author    = {Basu, Ayanendranath and Shioya, Hiroyuki and Park, Chanseok},
	year      = {2011},
	publisher = {CRC press},
    doi = {10.1201/b10956}
}

@book{davis2007methods,
	title     = {Methods of numerical integration},
	author    = {Davis, Philip J and Rabinowitz, Philip},
	year      = {2007},
	publisher = {Courier Corporation},
    doi = {10.1016/c2013-0-10566-1}
}

@inproceedings{karvonen2017classical,
	title     = {Classical quadrature rules via {G}aussian processes},
	author    = {Karvonen, Toni and S{\"a}rkk{\"a}, Simo},
	booktitle = {2017 IEEE 27th International Workshop on Machine Learning for Signal Processing (MLSP)},
	pages     = {1--6},
	year      = {2017},
    doi = {10.1109/mlsp.2017.8168195}
}

@InProceedings{acuna2021f,
	title     = {f-{D}omain adversarial learning: theory and algorithms},
	author    = {Acuna, David and Zhang, Guojun and Law, Marc T. and Fidler, Sanja},
	booktitle = {Proceedings of the 38th International Conference on Machine Learning},
	pages     = {66--75},
	year      = {2021},
	series    = {PMLR 139},
}

@article{gretton2009covariate,
	title     = {Covariate shift by kernel mean matching},
	author    = {Gretton, Arthur and Smola, Alex and Huang, Jiayuan and Schmittfull, Marcel and Borgwardt, Karsten and Sch{\"o}lkopf, Bernhard},
	journal   = {Dataset shift in machine learning},
	volume    = {3},
	number    = {4},
	year      = {2009},
    doi = {10.7551/mitpress/9780262170055.003.0008}
}

@article{dwivedi2024kernel,
	author    = {Raaz Dwivedi and Lester Mackey},
	title     = {Kernel Thinning},
	journal   = {Journal of Machine Learning Research},
	year      = {2024},
	volume    = {25},
	number    = {152},
	pages     = {1--77},
}

@article{borgwardt2006integrating,
	title     = {Integrating structured biological data by kernel maximum mean discrepancy},
	author    = {Borgwardt, Karsten M and Gretton, Arthur and Rasch, Malte J and Kriegel, Hans-Peter and Sch{\"o}lkopf, Bernhard and Smola, Alex J},
	journal   = {Bioinformatics},
	volume    = {22},
	number    = {14},
	pages     = {49--57},
	year      = {2006},
	publisher = {Oxford University Press},
    doi = {10.1093/bioinformatics/btl242}
}

@article{kullback1951information,
	title     = {On information and sufficiency},
	author    = {Kullback, Solomon and Leibler, Richard A},
	journal   = {The Annals of Mathematical Statistics},
	volume    = {22},
	number    = {1},
	pages     = {79--86},
	year      = {1951},
	publisher = {JSTOR},
    doi = {10.1214/aoms/1177729694}
}

@InProceedings{pmlr-v271-briol25a,
	title     = {A Dictionary of Closed-Form Kernel Mean Embeddings},
	author    = {Briol, Francois-Xavier and Karvonen, Toni and Gessner, Alexandra and Mahsereci, Maren},
	booktitle = {International Conference on Probabilistic Numerics},
	pages     = {84--94},
	year      = {2025},
	series    = {PMLR 271},
	}

@book{novak2008tractability,
	title     = {Tractability of Multivariate Problems: linear information},
	author    = {Novak, E. and Wo{\'z}niakowski, H.},
	series    = {EMS tracts in mathematics},
	year      = {2008},
	publisher = {European Mathematical Society},
    doi = {10.4171/026}
}

@book{mcbook,
	author    = {Art B. Owen},
	year      = 2013,
	title     = {{M}onte {C}arlo theory, methods and examples}
}

@article{neal2011mcmc,
	title     = {{MCMC} using {H}amiltonian dynamics},
	author    = {Neal, Radford M},
	journal   = {Handbook of {M}arkov chain {M}onte {C}arlo},
	volume    = {2},
	number    = {11},
	year      = {2011},
	publisher = {Chapman and Hall/CRC},
    doi = {10.1201/b10905-6 }
}

@incollection{doucet20011tutorial,
	author    = {Arnaud Doucet and Adam M. Johansen},
	title     = {A Tutorial on Particle Filtering and Smoothing: fifteen Years Later},
	booktitle = {The Oxford Handbook of Nonlinear Filtering},
	publisher = {Oxford University Press},
	year      = {2011}
}

@book{vanDerVaart1998,
	author    = {Aad W. van der Vaart},
	title     = {Asymptotic Statistics},
	year      = {1998},
	publisher = {Cambridge University Press},
	series    = {Cambridge Series in Statistical and Probabilistic Mathematics},
    doi = {10.1017/cbo9780511802256}
}

\appendix

\chapter{Efficient MMD Estimators for Simulation Based Inference: Supplementary Materials}
\label{sec:app_sbi}
In \Cref{app:proofs}, we present the proofs and derivations of all the theoretical results in the chapter, while \Cref{app:expDetails} contains additional details regarding our experiments.

\section{Proofs of Theoretical Results}\label{app:proofs}

In this section, we prove~\Cref{thm:optimal_weights,thm:rate_of_convergence} and intermediate results required, and expand on the technical background. 

\subsection{Proof of \Cref{thm:optimal_weights}}\label{app:proofs_optimal_weights}
\begin{proof}
Let $\bP_{\theta, N}^w = \sum_{n=1}^N w_n \delta_{x_n} = \sum_{n=1}^N w_n \delta_{G_\theta(u_n)}$. Using the fact that the MMD is a metric, we can use the reverse triangle inequality to get
\begin{align*}
 &\left| \MMD_k(\bP_\theta,\bQ) - \MMD_k(\bP_{\theta, N}^w,\bQ)\right|
 \leq  \MMD_k(\bP_\theta,\bP_{\theta, N}^w). 
\end{align*}
Define a kernel $c_\theta$ on $\calU$ as $c_\theta(u, u') = k(G_\theta(u), G_\theta(u'))$. As $\bP_\theta$ is a pushforward of $\bU$ under $G_\theta$, it holds that,
\begin{align*}
 \MMD^2_k(\bP_\theta,\bP_{\theta, N}^w) &= \int_{\calX} \int_{\calX}  k(x,x') \bP_\theta (\d x)  \bP_\theta (\d x') - 2 \sum_{n=1}^N w_n \int_{\calX}  k(x_n,x) \bP_\theta (\d x) \\
 &\hspace{5.55cm}+ \sum_{n,n'=1}^N w_n w_{n'} k(x_n, x_{n'})\\
 &= \int_{\calU} \int_{\calU}  k(G_\theta(u),G_\theta(u')) \bU (\d u)  \bU (\d u') \\
 &\hspace{4cm} - 2 \sum_{n=1}^N w_n \int_{\calU}  k(G_\theta(u_n), G_\theta(u)) \bU (\d u) \\
 &\hspace{4cm} + \sum_{n,n'=1}^N w_n w_{n'} k(G_\theta(u_n), G_\theta(u_{n'}))\\
 & = \MMD^2_{c_\theta}(\bU,\sum_{n=1}^N w_n \delta_{u_n}). 
\end{align*}
Since $c_\theta(u, \cdot) \in \calH_c$ for all $u \in \calU$ (by the assumption that $k(x, \cdot) \circ G_\theta \in \calH_c$ for all $x \in \calX$), it holds that $\calH_{c_\theta} \subseteq \calH_c$. If $\calH_{c_\theta} = \calH_c$, we have $\MMD_k(\bP_\theta,\bP_{\theta, N}^w)=\MMD_c(\bU,\sum_{n=1}^N w_n \delta_{u_n})$, and the result holds for $K=1$.

Suppose $\calH_{c_\theta} \subset \calH_c$. Then, by~\citet[Theorem I.13.IV]{aronszajn1950theory}, for any $f \in \calH_{c_\theta}$ there is a constant $K$ independent of $f$ such that $\|f\|_{\calH_c} \leq K \|f\|_{\calH_{c_\theta}}$. Together with the fact that  $\MMD_{c_\theta}$ is an integral-probability metric with underlying function class being the unit-ball in $\calH_{c_\theta}$, this gives
\begin{align*}
    \MMD_{c_\theta}(\bU,\sum_{n=1}^N w_n \delta_{u_n}) &= \sup_{\|f\|_{\calH_{c_\theta}}\leq 1} \left| \int_{\calU} f(u) \bU(\d u) - \sum_{n=1}^N w_n  f(u_n) \right| \\
    &=K \times \sup_{\|f\|_{\calH_{c_\theta}}\leq 1/K} \left| \int_{\calU} f(u) \bU(\d u) - \sum_{n=1}^N w_n  f(u_n) \right| \\
    &\leq K \times\sup_{\substack{f \in \calH_{c_\theta}\\\|f\|_{\calH_c}\leq 1}} \left| \int_{\calU} f(u) \bU(\d u) - \sum_{n=1}^N w_n  f(u_n) \right| \\
    &\leq K \times \sup_{\|f\|_{\calH_c}\leq 1} \left| \int_{\calU} f(u) \bU(\d u) - \sum_{n=1}^N w_n  f(u_n) \right| \\
    &= K \times \MMD_c\left(\bU,\sum_{n=1}^N w_n \delta_{u_n}\right),
\end{align*}
where the second equality is simply a reparameterisation from $f$ to $Kf$, and the inequalities use the fact that supremum of a set is not greater than supremum of its superset, and
\begin{align*}
    \{f \in \calH_{c_\theta} \ |\ K \|f\|_{\calH_{c_\theta}} \leq 1 \} \subseteq \{f \in \calH_{c_\theta}\ |\ \|f\|_{\calH_c} \leq 1 \} \subseteq \{f \in \calH_c\ |\ \|f\|_{\calH_c} \leq 1 \}.
\end{align*}
To prove the result about the exact form of $w$, we note that 
\begin{align*}
 \underset{ w \in \bR^N}{\argmin } ~ \MMD_c\left(\bU,\sum_{n=1}^N w_n \delta_{u_n}\right)= \underset{ w \in \bR^N}{\argmin }~ \MMD^2_c\left(\bU,\sum_{n=1}^N w_n \delta_{u_n}\right),
\end{align*}
and 
\begin{align*}
\MMD^2_c\left(\bU,\sum_{n=1}^N w_n \delta_{u_n}\right)  = \int_{\calU} \int_{\calU}  c(u, u') &\bU(\d u) \bU(\d  u')  \\
&- 2 \sum_{n=1}^N w_n \int_{\calU}  c(u_n,u) \bU(\d u) \\
&+ \sum_{n,n'=1}^N w_n w_{n'} c(u_n, u_{n'}).
\end{align*}
The latter is a quadratic form in $w$, meaning it can be minimised in closed-form over $w$ and the optimal weights are given by $w^*$. This completes the proof of the second part of the theorem.

\end{proof}

\subsection{Chain rule in Sobolev spaces}

The proof of~\Cref{thm:rate_of_convergence}, specifically the result $k(x, \cdot) \circ G_\theta \in \calH_c$ for Sobolev $k$ and $c$, will use a specific form of a chain rule for Sobolev spaces introduced in \Cref{sec:bg_sobolev_spaces}.

For general $c$ and $k$, $k(x, \cdot) \circ G_\theta \in \calH_c$ is non-trivial to check. Here, we introduce sufficient conditions on $c$, $k$, and $G_\theta$ that are easily interpretable and correspond to common practical settings. Specifically, we consider Sobolev $c$ and $k$, and $G_\theta$ of a certain degree of smoothness, which reduces the problem to a form of a chain rule for Sobolev spaces.\footnote{Though various forms of the chain rule for Sobolev spaces exist in the literature (for example,~\citet[Section 4.2.2]{evans2018measure}), they tend to either consider $F \circ f$, where $f$ is in the Sobolev space (rather than $F$), or place overly strong assumptions on $f$.} The rest of the section proceeds as follows: first, we introduce the background definitions and results, then show that the required form of the chain rule holds for first order derivatives (\Cref{lemma:chain_rule_w1p}), and finally extend the result to higher order derivatives (\Cref{thm:k_circ_G_is_Sobolev}).

For convenience, we denote by $D_{x_i} f$ the first-order weak derivative of $f$ in $i$-th dimension $x_i$, and by $D^{\alpha_i}_{x_i} f$ the  $\alpha_i$-th order weak derivative of $f$ in $i$-th dimension $x_i$. For a multi-index $\alpha = (\alpha_1, \dots, \alpha_d) \in \bN^d$,
\begin{equation*}
    D^\alpha f=D^{\alpha_1}_{x_1} \dots D^{\alpha_d}_{x_d} f, \qquad \qquad |\alpha|\coloneqq \sum_{i=1}^d \alpha_i.
\end{equation*}

We start by recalling an important result characterising Sobolev functions as limit points of sequences of $C^\infty(\calX)$ functions.
Since it is a necessary and sufficient condition, we will use this result both to operate on a function in a Sobolev space using the "friendlier" smooth functions, and to prove that a function of interest lies in a Sobolev space by finding a sequence of smooth functions that approximates it accordingly.

\begin{theorem}[Theorem 3.17, \citet{adams2003sobolev}]
\label{thm:sobolev_function_approximation}
For an open set $\calX \subseteq \bR^d$, a function $f: \calX \to \bR$ lies in the Sobolev space $\calW^{1, 2}(\calX)$ and has weak derivatives $D_{x_j}[f]$, $j \in \{1, \dots, d\}$ 
if and only if there exists a sequence of functions $f_n \in C^\infty(\calX) \cap \calW^{1, 2}(\calX)$ such that for $j \in \{1, \dots, d\}$
\begin{align}
\label{eq:sobolev_function_approximation_funcs}
    \| f - f_n \|_{\calL^2(\calX)} \to 0,&\quad n \to \infty,  \\
    \left\| D_{x_j}[f] - \frac{\partial f_n}{\partial x_j} \right\|_{\calL^2(\calX)} \to 0,&\quad n \to \infty, \label{eq:sobolev_function_approximation_funcs2}
\end{align}
where $\frac{\partial f_n}{\partial x_j}$ is the ordinary derivative of $f_n$ with respect to $x_j$.
\end{theorem}

Note that the functions $f_n$ converge to $f$ in the Sobolev $\calW^{1, 2}(\calX)$ norm, $\| f - f_n \|_{\calW^{1, 2}(\calX)} = (\|f -f_n\|_{\calL^2(\calX)} + \sum_{j=1}^d \|D_{x_j} f- \partial f_n / \partial x_j \|_{\calL^2(\calX)})^{1/2} \to 0$ as $n \to \infty$ , if and only if \eqref{eq:sobolev_function_approximation_funcs} and \eqref{eq:sobolev_function_approximation_funcs2} hold.

\paragraph{Chain rule for $\calW^{1, 2}$.} We now prove that chain rule holds for $\varphi \circ G_\theta$ for $\varphi$ in a Sobolev space $W^{1, 2}(\calX)$. For clarity, we will explicitly state the assumptions on $G_\theta$ in the main text. Recall that a measure $\bP_\theta$ on $\calX \subseteq \bR^d$ is said to be a pushforward of a measure $\bU$ on $\calU \subseteq \bR^r$ under $G_\theta: \calU \to \calX$ if for any $\calX$-measurable $f: \calX \to \bR$ it holds that $\int_\calX f(x) \bP_\theta(\d x)= \int_\calU \left[f \circ G_\theta\right](u) \bU(\d u)$. 

\begin{lemma}[Chain rule for $\calW^{1, 2}$]
\label{lemma:chain_rule_w1p}

Suppose
\begin{itemize}
  \item $\varphi \in \calW^{1, 2}(\calX)$. 
  \item $\calU \subset \bR^{r}$ is bounded, $\calX \subset \bR^d$ is open, and $\calX = G_\theta(\calU)$ for some $G_\theta = (G_{\theta, 1}, \dots, G_{\theta, d})^\top$.
  The partial derivative $\partial G_{\theta, j}/\partial u_i$ exists and $|\partial G_{\theta, j}/\partial u_i| \leq C_G$ for some $C_G$ for all $i \in \{1, \dots, r\}$ and $j \in \{1, \dots, d\}$.
  \item $\bU$ is a probability distribution on $\calU$ that has a density $f_\bU: \calU \to [C_\bU, \infty)$ for $C_\bU>0$. 
  \item $\bP_\theta$ is a pushforward of $\bU$ under $G_\theta$, and has a density $f_{\bP_\theta}$ such that $f_{\bP_\theta}(x) \leq C_{\bP_\theta}$ for all $x \in \calX$ for some $C_{\bP_\theta}$.
\end{itemize}
Then $\varphi \circ G_\theta \in \calW^{1, 2}(\calU)$, and for $i \in \{1, \dots, r\}$, its weak derivative $D_{u_i}[\varphi \circ G_\theta]$ is equal to $\sum_{j=1}^d [D_{x_j} \varphi \circ G_\theta] \frac{\partial {G_{\theta, j}}}{\partial u_i}$.
\end{lemma}

\begin{proof}

Since $\calX$ is open, by \Cref{thm:sobolev_function_approximation} there is a sequence $\varphi_n \in C^\infty(\calX) \cap \calW^{1, 2}(\calX)$ such that
\begin{align*}
    \| \varphi - \varphi_n \|_{\calL^2(\calX)} \to 0,&\quad n \to \infty, \\
    \left\| D_{x_j}\varphi - \frac{\partial \varphi_n}{\partial x_j} \right\|_{\calL^2(\calX)} \to 0,&\quad n \to \infty,
\end{align*}
The proof proceeds as follows: we show that the sequence $\varphi_n \circ G_\theta$ approximates $\varphi \circ G_\theta$, and $\frac{\partial \left[\varphi_n \circ G_\theta\right]}{\partial u_i}$ approximates the sum in the statement of the lemma, $\sum_{j=1}^d [D_{x_j} \varphi \circ G_\theta] \frac{\partial {G_{\theta, j}}}{\partial u_i}$, in $\calL^2(\calU)$--norm. Then, by the sufficient condition in \Cref{thm:sobolev_function_approximation}, $\varphi \circ G_\theta$ lies in $\calW^{1, 2}(\calU)$, and its weak derivative in $u_i$ is $\sum_{j=1}^d [D_{x_j} \varphi \circ G_\theta](u) \frac{\partial {G_{\theta, j}}}{\partial u_i}(u)$, for any $i \in \{1, \dots, r\}$.

Since $\bP_\theta$ has a density,
for any $\calX$-measurable $f$ it holds that
\begin{align*}
    \int_\calX f(x) f_{\bP_\theta}(x) \d x = \int_\calU \left[f \circ G_\theta\right](u) f_\bU(u) \d u.
\end{align*}
Together with density bounds, this gives $\|\varphi \circ G_\theta - \varphi_n \circ G_\theta\|_{\calL^2(\calU)} \to 0$ as
\begin{align*}
   \int_\calU &\left(\varphi \circ G_\theta(u) - \varphi_n \circ G_\theta(u) \right)^2 \d u \\
   &\qquad\leq C_\bU^{-1} \int_\calU \left(\varphi \circ G_\theta(u) - \varphi_n \circ G_\theta(u) \right)^2 f_\bU(u) \d u \\
   &\qquad= C_\bU^{-1} \int_\calX \left(\varphi(x)  - \varphi_n(x)\right)^2 f_{\bP_\theta}(x) \d x \\
   &\qquad\leq C_\bU^{-1} C_{\bP_\theta} \int_\calX \left(\varphi(x)  - \varphi_n(x)\right)^2 \d x.
\end{align*}
In the same fashion, $\|D_{x_j}\varphi \circ G_\theta - \frac{\partial \varphi_n}{\partial x_j} \circ G_\theta\|_{\calL^2(\calU)} \to 0$ since
\begin{align*}
   \int_\calU \big(D_{x_j}\varphi \circ G_\theta(u) &- \frac{\partial \varphi_n}{\partial x_j} \circ G_\theta (u)\big)^2 \d u \\
   & \leq C_\bU^{-1} \int_\calU \big(D_{x_j}\varphi \circ G_\theta (u) - \frac{\partial \varphi_n}{\partial x_j} \circ G_\theta (u) \big)^2 f_\bU(u) \d u \\
   &= C_\bU^{-1} \int_\calX \big(D_{x_j}\varphi(x)  - \frac{\partial \varphi_n}{\partial x_j} (x)\big)^2 f_{\bP_\theta}(x) \d x \nonumber  \\
   &\leq C_\bU^{-1} C_{\bP_\theta} \int_\calX \big(D_{x_j}\varphi (x)  - \frac{\partial \varphi_n}{\partial x_j} (x)\big)^2 \d x.
\end{align*}
Since $\varphi$ and $G_\theta$ are both differentiable, the ordinary chain rules applies to $\varphi_n \circ G_\theta$,
\begin{align*}
    \frac{\partial [\varphi_n \circ G_\theta]}{\partial u_i}= \sum_{j=1}^d \left[\frac{\partial \varphi_n }{\partial x_j} \circ G_\theta \right] \frac{\partial {G_{\theta, j}}}{\partial u_i},
\end{align*}
and for any $i \in \{1, \dots, r\}$ the convergence of derivatives 
 $\|[D_{x_j} \varphi \circ G_\theta] \frac{\partial {G_{\theta, j}}}{\partial u_i} - \frac{\partial [\varphi_n \circ G_\theta]}{\partial u_i} \|_{\calL^2(\calU)} \to 0$ follows since
\begin{align}
    \int_\calU \Big( \sum_{j=1}^d \big[D_{x_j} \varphi \circ G_\theta\big] \frac{\partial {G_{\theta, j}}}{\partial u_i} &- \frac{\partial [\varphi_n \circ G_\theta]}{\partial u_i} \Big)^2  \d u \\
    &= \int_\calU \Big( \sum_{j=1}^d \big[D_{x_j}\varphi \circ G_\theta - \frac{\partial \varphi_n}{\partial x_j} \circ G_\theta\big]  \frac{\partial {G_{\theta, j}}}{\partial u_i} \Big)^2 \d u \nonumber \\
    &\leq d \sum_{j=1}^d \int_\calU \Big(  \big[D_{x_j}\varphi \circ G_\theta - \frac{\partial \varphi_n}{\partial x_j} \circ G_\theta\big]  \frac{\partial {G_{\theta, j}}}{\partial u_i} \Big)^2 \d u \nonumber \\
    &\leq d C_G^2 \sum_{j=1}^d \int_\calU \big(  D_{x_j}\varphi \circ G_\theta - \frac{\partial \varphi_n}{\partial x_j} \circ G_\theta  \big)^2 \d u \nonumber
\end{align}
where
the first inequality is using the inequality $(\sum_{i=1}^d a_i)^2 \leq d \sum_{i=1}^d a_i^2$.
This completes the proof.

\end{proof}

\paragraph{Chain rule for $\calW^{s, 2}$.} To extend~\Cref{lemma:chain_rule_w1p} to Sobolev spaces of order higher than 1, we need the following version of the weak derivative product rule, for a product of a function $f$ in $\calW^{1, 2}$ and a bounded differentiable function $g$ with bounded derivatives. Other versions of the product rule, for different regularity assumptions on $g$, exist in the literature (for example,~\citet{adams2003sobolev}); we will require this specific form.
\begin{lemma}[Product rule]
\label{lemma:product_rule}
    Suppose $\calX \subseteq \bR^d$ is open, $f \in \calW^{1, 2}(\calX)$, $g(x)$ is differentiable on $\calX$, and $|g(x)| \leq L$, $|[\partial g / \partial x_i ] (x)| \leq L$ for all $x \in \calX$ for some constant $L$. Then $fg \in \calW^{1, 2}(\calX)$ and for any $i \in \{1, \dots, d\}$,
    \begin{align*}
        D_{x_i} [fg] = [ D_{x_i} f ] g + f \big[\partial g / \partial x_i\big]
    \end{align*}
\end{lemma}
\begin{proof}
    By the criterion in~\Cref{thm:sobolev_function_approximation}, there is a sequence of smooth functions $f_n$ approximating $f$, meaning 
    \begin{align*}
        \int_\calX (f(x) - f_n(x))^2 \d x \to 0 \text{ as } n \to \infty,\\
        \int_\calX \big(D_{x_i} f(x) - [\partial f_n/ \partial x_i](x)\big)^2 \d x \to 0 \text{ as } n \to \infty.
    \end{align*}
     We will show that $f_n g$ approximates $fg$ with weak derivatives taking the form $[ D_{x_i} f ] g + f [\partial g / \partial x_i]$; by the aforementioned criterion, it will follow that $fg \in \calW^{1, 2}(\calX)$.

    First, we establish convergence of functions. As $n \to \infty$,
    \begin{align*}
        \| fg - f_n g \|^2_{\calL^2(\calX)} &= \int_\calX \left(f(x) g(x) - f_n (x) g(x)\right)^2 \d x \\
        &\leq L^2 \int_\calX \left(f(x) - f_n (x) \right)^2 \d x \to 0.
    \end{align*}
    By the ordinary chain rule, $\partial [f_n g] / \partial x_i = [\partial f_n / \partial x_i]g + f [\partial g / \partial x_i]$.
    Then, applying triangle inequality for norms and the fact that $(a+b)^2 \leq 2a^2 + 2b^2$ for any $a$, $b$, we get that for $n \to \infty$,
    \begin{align*}
        &\left\| \frac{\partial f_n }{ \partial x_i}g + f_n \frac{\partial g}{\partial x_i} - \left[ D_{x_i} f \right] g - f \frac{\partial g }{ \partial x_i} \right\|^2_{\calL^2(\calX)} \\
        &\hspace{2cm}\leq 2\left\| \frac{\partial f_n}{ \partial x_i}g - \left[ D_{x_i} f \right] g \right\|^2_{\calL^2(\calX)} + 2\left\| f_n \frac{\partial g}{ \partial x_i} - f \frac{\partial g}{ \partial x_i} \right\|^2_{\calL^2(\calX)} \\
        &\hspace{2cm}\leq 2 L^2 \left\| \frac{\partial f_n}{ \partial x_i} - \left[ D_{x_i} f \right] \right\|^2_{\calL^2(\calX)} + 2 L^2 \left\| f_n - f \right\|_{\calL^2(\calX)} \to 0.
    \end{align*}
    This completes the proof.
\end{proof}

We are now ready to extend the chain rule from order 1, proven in~\Cref{lemma:chain_rule_w1p}, to arbitrary order $s$.

\begin{theorem}[Chain rule for $\calW^{s, 2}$]
\label{thm:k_circ_G_is_Sobolev}
Suppose
\begin{itemize}
  \item $\varphi \in \calW^{s_\varphi, 2}(\calX)$.
  \item $\calU \subset \bR^r$ is bounded, $\calX \subset \bR^d$ is open, and $\calX = G_\theta(\calU)$ for some $G_\theta = (G_{\theta, 1}, \dots, G_{\theta, d})^\top$. For some $s_G$ and any $|\alpha|\leq s_G$, $j \in \{1, \dots, r\}$, the derivative $\partial^\alpha G_{\theta,j}$ exists and is in $\calL^\infty(\calU)$.
  \item $\bU$ is a probability distribution on $\calU$ that has a density $f_\bU: \calU \to [C_\bU, \infty)$ for $C_\bU>0$.
  \item $\bP_\theta$ is a pushforward of $\bU$ under $G_\theta$ with a density bounded above.
\end{itemize}
Then $\varphi \circ G_\theta \in \calW^{s_0, 2}(\calU)$ for $s_0=\min\{s_\varphi, s_G\}$, and for any $s \leq s_0$ and $|\alpha_0| = s$, the derivative takes an $\alpha_0$-specific $(\kappa, \beta, \alpha, \eta)$--form
\begin{align}
\label{eq:general_faa_di_bruno}
    D^{\alpha_0}[\varphi \circ G_\theta] = \sum_{i=1}^I \sum_{j=1 }^{d^{\kappa_i}} \left[ D^{\beta_{ij}} \varphi \circ G_\theta \right] \prod_{l=1}^{\kappa_i} \partial^{\alpha_{ijl}} G_{\theta, \eta_{ijl} },
\end{align}
where $I \in \bN_{\geq 1}$, and for any $i \in \{1, \dots, I\}$, $s \geq \kappa_i \in \bN$; $\beta_{ij} \in \bN^d$ is a multi-index of size $\kappa_i$ for $j \in \{1, \dots, d^{\kappa_i}\}$; $\alpha_{ijl} \in \bN^r$ is of size $|\alpha_{ijl}| \leq s$, and $\eta_{ijl} \in \{1, \dots, d\}$ for $l \in \{1, \dots, \kappa_i\}$.
\end{theorem}

By saying the $(\kappa, \beta, \alpha, \eta)$ form is $\alpha_0$-specific, we mean that the values of $I, (\kappa, \beta, \alpha, \eta)$ depend on $\alpha_0$, and may be different for $\alpha'_0 \neq \alpha_0$; we do not index $I, (\kappa, \beta, \alpha, \eta)$ by $\alpha_0$ for the sake of readability.

Before proving this result, let us point out that the $(\kappa, \beta, \alpha, \eta)$--form introduced in the theorem can be seen as a form of Faà di Bruno's formula, which generalises the chain rule to higher derivatives~\citep[Theorem 1]{Constantine1996}. 
However, since our ultimate goal is to show $\varphi \circ G_\theta \in \calW^{s_0, 2}(\calU)$, and the expression for the derivative is simply a means for proving that, an unspecified $(\kappa, \beta, \alpha, \eta)$--form suffices. It is simpler to prove the general $(\kappa, \beta, \alpha, \eta)$ case without using explicit Faà di Bruno forms.

\begin{proof}[Proof of~\Cref{thm:k_circ_G_is_Sobolev}]
Note that $\varphi \circ G_\theta \in \calW^{s_0, 2}(\calU)$ if and only if $\varphi \circ G_\theta \in \calW^{s, 2}(\calU)$ for $s \leq s_0$. We use this to construct a proof by induction: we show the statement holds for $s=1$, and that $\varphi \circ G_\theta \in \calW^{s, 2}(\calU)$ implies $\varphi \circ G_\theta \in \calW^{s+1, 2}(\calU)$ if $s+1\leq s_0$ (and the weak derivatives take a $(\kappa, \beta, \alpha, \eta)$-form stated in~\eqref{eq:general_faa_di_bruno}).

\paragraph{Case $s=1$:} $\varphi \circ G_\theta$ is in $\calW^{1, 2}(\calU)$. 
    
Suppose $\alpha_0=e[m]$ for some unit vector $e[m] = (0, \dots, 0, 1, 0, \dots, 0)$ where the $1$ is the $m$-th element. Then, as proven in~\Cref{lemma:chain_rule_w1p}, $D^{e[m]}[\varphi \circ G_\theta] = D_{u_m}[\varphi \circ G_\theta]$ is equal to $\sum_{j=1}^d [D_{x_j} \varphi \circ G_\theta] [\partial {G_{\theta, j}} / \partial u_m]=\sum_{j=1}^d [D^{e[j]} \varphi \circ G_\theta] \partial^{e[m]} {G_{\theta, j}} $, so the statement holds for $I=1$, $\kappa_1 = 1$, $\beta_{1j} = e[j]$, $\alpha_{1j1} = e[m]$, $\eta_{1j1} = j$.

\paragraph{Case $s$ implies $s+1$:} If $s+1 \leq s_0$ and $\varphi \circ G_\theta$ is in $\calW^{s, 2}(\calU)$, and for every $|\alpha_0|=s$~\eqref{eq:general_faa_di_bruno} holds for some $\alpha_0$-specific $(\kappa, \beta, \alpha, \eta)$, then $\varphi \circ G_\theta$ is in $\calW^{s+1, 2}(\calU)$, and for any $|\tilde{\alpha}_0|=s+1$ there is a  $(\tilde{\kappa}, \tilde{\beta}, \tilde{\alpha}, \tilde{\eta})$--form, $|\tilde{\kappa}|=\tilde I$,
\begin{align}
\label{eq:induction_proof_eq5}
    D^{\tilde{\alpha}_0}[\varphi \circ G_\theta] = \sum_{i=1}^{\tilde{I}} \sum_{j=1 }^{d^{\tilde{\kappa}_i}} \left[ D^{\tilde{\beta}_{ij}} \varphi \circ G_\theta \right] \prod_{l=1}^{\tilde{\kappa}_i} \partial^{\tilde{\alpha}_{ijl}} G_{\theta, \tilde{\eta}_{ijl} }.
\end{align}
\par By induction assumption, $\varphi \circ G_\theta$ is in $\calW^{s, 2}(\calU)$, so it is in $\calW^{s+1, 2}(\calU)$ if and only if $D^{\alpha_0} \left[ \varphi \circ G_\theta \right]$ is in $\calW^{1, 2}(\calU)$ for any $\alpha_0$ of size $s$. The latter can be shown by studying the $(\kappa, \beta, \alpha, \eta)$--form that $D^{\alpha_0}[\varphi \circ G_\theta]$ takes by~\eqref{eq:general_faa_di_bruno}, for some $\alpha_0$--specific $(\kappa, \beta, \alpha, \eta)$. Since $s_\varphi \geq s_0 \geq s+1$ (the last inequality holds by the induction assumption), it holds that $\calW^{s_\varphi, 2}(\calX) \subseteq \calW^{s_0, 2}(\calX) \subseteq \calW^{s+1, 2}(\calX)$. 
Then $\varphi \in \calW^{s+1, 2}(\calX)$, and since $|\beta_{ij}|=\kappa_i \leq s$ by definition of $\beta_{ij}$, we have $D^{\beta_{ij}} \varphi \in \calW^{1, 2}(\calX)$ for all $i, j$. Then by~\Cref{lemma:chain_rule_w1p}, its composition with $G_\theta$ is in $\calW^{1, 2}(\calU)$, i.e., $D^{\beta_{ij}}\varphi \circ G_\theta \in \calW^{1, 2}(\calU)$. Consequently, $D^{\alpha_0}\left[\varphi \circ G_\theta\right]$ as per~\eqref{eq:general_faa_di_bruno} is a sum over the product of functions in $\calW^{1, 2}(\calU)$, and bounded functions with bounded derivatives; by~\Cref{lemma:product_rule}, such product is in $\calW^{1, 2}(\calU)$, and it follows that $D^{\alpha_0}\left[\varphi \circ G_\theta\right] \in \calW^{1, 2}(\calU)$ as well.

Finally, we show that for any fixed $\tilde{\alpha}_0$ such that $|\tilde{\alpha}_0|=s+1$ there are $\tilde{I}, \tilde{\kappa}, \tilde{\beta}, \tilde{\alpha}, \tilde{\eta}$ for which~\eqref{eq:induction_proof_eq5} holds; this will conclude the induction step.
Suppose $\alpha_0$ of size $s$, $|\alpha_0| = s$, is such that $\tilde{\alpha}_0 = \alpha_0 + e[m]$ for some $\alpha_0$ (that is unrelated to $\alpha_0$ in the previous part of the proof) and a unit vector $e[m]$ (such pair of $m$ and $\alpha_0$ must exist as $|\tilde{\alpha}_0| = s + 1$). For this $\alpha_0$, in a slight abuse of notation, we shall say that $\kappa, \beta, \alpha, \eta$ are such that $D^{\alpha_0}[\varphi \circ G_\theta]$ takes a $(\kappa, \beta, \alpha, \eta)$ form. Then, by the sum rule for weak derivatives and the product rule of~\Cref{lemma:product_rule}, $D^{\tilde{\alpha}_0}\left[\varphi \circ G_\theta\right] = D_{u_m}\left[D^{\alpha_0}\left[\varphi \circ G_\theta\right]\right]$ takes the form
\begin{align}
        D^{\tilde{\alpha}_0}\left[\varphi \circ G_\theta\right] &= D_{u_m}\left[D^{\alpha_0}\left[\varphi \circ G_\theta\right]\right] \\
        & = \sum_{i=1}^I \sum_{j=1 }^{d^{\kappa_i}}  D_{u_m}\left[D^{\beta_{ij}} \varphi \circ G_\theta\right] \prod_{l=1}^{\kappa_i} \partial^{\alpha_{ijl}} G_{\theta, \eta_{ijl} } \nonumber \\
        &\qquad + \sum_{i=1}^I \sum_{j=1 }^{d^{\kappa_i}} \left[ D^{\beta_{ij}} \varphi \circ G_\theta \right] \partial^{e[m]} \Big[\prod_{l=1}^{\kappa_i} \partial^{\alpha_{ijl}} G_{\theta, \eta_{ijl} }\Big].
\label{eq:induction_proof_eq}
\end{align}   
By the product rule for regular derivatives,
\begin{equation*}
\begin{aligned}   
    \partial^{e[m]} \Big[\prod_{l=1}^{\kappa_i} \partial^{\alpha_{ijl}} G_{\theta, \eta_{ijl} }\Big] = \sum_{l_0=1}^{\kappa_i} \partial^{\alpha_{ijl_0} + e[m]} G_{\theta, \eta_{ijl_0}} \prod_{\substack{l \in \{1, \dots, \kappa_i \} \\ l \neq l_0}} \partial^{\alpha_{ijl}} G_{\theta, \eta_{ijl} }.
\end{aligned}   
\end{equation*}
Since $D^{\beta_{ij}} \varphi \in \calW^{1, 2}(\calX)$, the statement in~\Cref{lemma:chain_rule_w1p} applies to its composition with $G_\theta$, i.e.,
\begin{align*}
    D_{u_m}\left[D^{\beta_{ij}} \varphi \circ G_\theta\right] 
    &= \sum_{j_0=1}^d \left[D_{x_{j_0}} \left[D^{\beta_{ij}} \varphi\right] \circ G_\theta\right] \frac{\partial {G_{\theta, j_0}}}{\partial u_m} \\
    &= \sum_{j_0=1}^d \left[D^{\beta_{ij} + e[j_0]} \varphi \circ G_\theta\right] \frac{\partial {G_{\theta, j_0}}}{\partial u_m},
\end{align*}
where, recall, $e[j_0]$ is a $d$-dimensional unit vector with $1$ as the $j_0$-th element. Substituting these into~\eqref{eq:induction_proof_eq}, we get
\begin{align}
\begin{split}
\label{eq:induction_proof_eq2}
        D^{\tilde{\alpha}_0}[\varphi \circ G_\theta] 
        &= \sum_{i=1}^I \sum_{j=1 }^{d^{\kappa_i}}  \sum_{j_0=1}^d \left[D^{\beta_{ij} + e[j_0]} \varphi \circ G_\theta\right] \frac{\partial {G_{\theta, j_0}}}{\partial u_m} \prod_{l=1}^{\kappa_i} \partial^{\alpha_{ijl}} G_{\theta, \eta_{ijl} }  \\
        &\quad +\sum_{i=1}^I \sum_{l_0=1}^{\kappa_i} \sum_{j=1 }^{d^{\kappa_i}} \left[ D^{\beta_{ij}} \varphi \circ G_\theta \right] \partial^{\alpha_{ijl_0} + e[m]} G_{\theta, \eta_{ijl_0}} \prod_{\substack{l \in \{1, \dots, \kappa_i \} \\ l \neq l_0}} \partial^{\alpha_{ijl}} G_{\theta, \eta_{ijl} }
\end{split}
\end{align}  
Now all that is left to do is find $\tilde{I}, \tilde{\kappa}, \tilde{\beta}, \tilde{\alpha}, \tilde{\eta}$ for which this will take the $(\tilde{\kappa}, \tilde{\beta}, \tilde{\alpha}, \tilde{\eta})$--form similar to~\eqref{eq:general_faa_di_bruno}. One can already see this should be possible, due to the flexibility in the definition of $(\tilde{\kappa}, \tilde{\beta}, \tilde{\alpha}, \tilde{\eta})$--forms; for completeness, we give the exact values now.

Define $\kappa_0 = 0$. Take $\tilde{I} = I+\sum_{i=1}^I \kappa_i$, and
\begin{flalign*}
    \tilde{\kappa}_i = 
    \begin{cases}
        \kappa_i+1, & i \in \{1, \dots, I\}, \\
        \kappa_p, & i \in (I+\sum_{h=0}^{p-1} \kappa_h, I+\sum_{h=0}^{p} \kappa_h] \text{ for } p \in \{1, \dots, I \},
    \end{cases}&&
\end{flalign*}
\begin{flalign*}
    \tilde{\beta}_{ij} = 
    \begin{cases}
        \beta_{i\lfloor j/d \rfloor} + e[j \mod d], & i \in \{1, \dots, I\},\ j \in \{1, \dots, d^{\kappa_i+1}\}, \\
        \beta_{pj}, & i \in (I+\sum_{h=0}^{p-1} \kappa_h, I+\sum_{h=0}^{p} \kappa_h],\\ 
         & j \in \{1, \dots, d^{\kappa_p}\} \text{ for } p \in \{1, \dots, I \},
    \end{cases}&&
\end{flalign*}
\begin{flalign*}
    \tilde{\alpha}_{ijl} = 
    \begin{cases}
         \alpha_{i\lfloor j/d \rfloor l}, & i \in \{1, \dots, I\},\ j \in \{1, \dots, d^{\kappa_i+1}\},\ l \in \{1, \dots, \kappa_i \}, \\
         e[m], & i \in \{1, \dots, I\},\ j \in \{1, \dots, d^{\kappa_i+1}\},\ l = \kappa_i+1, \\
         \alpha_{pjl}, & i \in (I+\sum_{h=0}^{p-1} \kappa_h, I+\sum_{h=0}^{p} \kappa_h],\ j \in \{1, \dots, d^{\kappa_p}\},\\ 
         & l \in \{1, \dots, \kappa_p \} \setminus \{i - I-\sum_{h=0}^{p-1} \kappa_h\}, \\
         \alpha_{pjl} + e[m], & i \in (I+\sum_{h=0}^{p-1} \kappa_h, I+\sum_{h=0}^{p} \kappa_h],\ j \in \{1, \dots, d^{\kappa_p}\},
         \\ 
         & l = i - I-\sum_{h=0}^{p-1} \kappa_h \text{ for } p \in \{1, \dots, I \},
    \end{cases}&&
\end{flalign*}
\begin{flalign*}
    \tilde{\eta}_{ijl} = 
    \begin{cases}
        \eta_{i\lfloor j/d \rfloor l}, & i \in \{1, \dots, I\},\ j \in \{1, \dots, d^{\kappa_i+1}\},\ l \in \{1, \dots, \kappa_i \}, \\
        j \mod d , & i \in \{1, \dots, I\},\ j \in \{1, \dots, d^{\kappa_i+1}\},\ l = \kappa_i+1, \\
        \eta_{pjl}, & i \in (I+\sum_{h=0}^{p-1} \kappa_h, I+\sum_{h=0}^{p} \kappa_h],\ j \in \{1, \dots, d^{\kappa_p}\},\\ 
         & l \in \{1, \dots, \kappa_p \}  \text{ for } p \in \{1, \dots, I \},
    \end{cases}&&
\end{flalign*}
where $j \mod d$ is the remainder of dividing $j$ by $d$. Then,~\eqref{eq:induction_proof_eq2} becomes
\begin{align*}
    D^{\tilde{\alpha}_0}[\varphi \circ G_\theta] = \sum_{i=1}^{\tilde{I}} \sum_{j=1 }^{d^{\tilde{\kappa}_i}} \left[ D^{\tilde{\beta}_{ij}} \varphi \circ G_\theta \right] \prod_{l=1}^{\tilde{\kappa}_i} \partial^{\tilde{\alpha}_{ijl}} G_{\theta, \tilde{\eta}_{ijl} }.
\end{align*}
This completes the proof of the induction step, and the theorem.
\end{proof}

\subsection{Proof of~\Cref{thm:rate_of_convergence}}

Before proving the main theorem, we introduce an auxiliary lemma, which is a straightforward corollary of~\citet[Theorem 9]{wynne2021convergence}.

\begin{lemma}[Corollary of Theorem 9 in~\citet{wynne2021convergence}]
\label{lemma:bq_convergence}
    Suppose for any $N \geq N_0 \in \bN_{\geq 1}$,
    \begin{itemize}
        \item $\bU$ is a measure on a convex, open, and bounded $\calU \subset \bR^r$ that has a density $f_\bU: \calU \to [0, C_\bU']$ for some $C_\bU'>0$.
        \item $u_{1:N}$ are such that the fill distance $h_N =\bigo(N^{-1/r})$.
        \item $w_{1:N}$ are the optimal weights obtained based on the kernel $c_{\beta_N}$ and measure $\bU$, parameterised by $\beta_N \in B$ for some parameter space $B$,
        \item for any $\beta \in B$, $c_{\beta}$ is a Sobolev kernel of smoothness $s_c$; $s_c$ is independent of $\beta$.
    \end{itemize}
    Then, for some $C_0$ independent of $N$ and $f$, and any $f \in \calH_c$ with $\|f\|_{\calH_c} = 1$,
    \begin{align*}
        \left| \int_\calU f (u)\bU(\d u) - \sum_{n=1}^N w_n f(u_n) \right| \leq C_0 N^{-s_c/r}.
    \end{align*}
\end{lemma}
\begin{proof}
    The expression on the left hand side of~\citet[Theorem 9]{wynne2021convergence} is $|\int_\calU f (u)\bU(\d u) - \sum_{n=1}^N w_n f(u_n)|$; the notation from their paper to this result maps as $\theta \to \beta$, $p \to f_\bU$, $\calX \to \calU$, $x \to u$, $\Theta \to B$, and the prior mean $\mu(\beta)=0$ for any $\beta \in B$. First, we show the assumptions in the Theorem hold.
    
    {Assumption 1 (Assumptions on the Domain):} An open, bounded, and convex $\calU$ satisfies the assumption, as discussed in~\citet{wynne2021convergence}.
    
    {Assumption 2 (Assumptions on the Kernel Parameters):} The smoothness of $c_\beta$ was assumed to be $s_c$ regardless of the value of $\beta \in B$, meaning $\tau(\beta)=\tau_c^-=\tau_c^+=s_c>r/2$. Lastly, the norm equivalence constants of \citet[Equation 3]{wynne2021convergence} are the same for all $\beta$, since the respective RKHS and Sobolev spaces are the same, so the set of extreme values $B_m^*$ is finite and does not depend on $m$; we denote $B_c^*=B_m^*$, to highlight that $B_c^*$ only depends on the choice of kernel family $c$ and not $m$.
    
    {Assumption 3 (Assumptions on the Kernel Smoothness Range):} As discussed in Assumption 2, $\tau(\beta)=s_c$ for any $\beta \in B$, so the set in the statement of Assumption 3 has only one element.
    
    {Assumption 4 (Assumptions on the Target Function and Mean Function):} The target function $f$ is in $\calH_c$, i.e., $\tau_f = \tau_c^- = \tau_c^+ = s_c$. The mean function $\mu(\beta)$ was taken to be zero, so has zero norm.

    Lastly, take $h_0$ such that $h_1 \leq h_0$; as we assumed $h_N = \bigo(N^{-1/r})$, it holds that $h_0 \leq h_N$ for all $N \geq 1$. Therefore, all the assumptions are satisfied and~\citet[Theorem 9]{wynne2021convergence} applies; moreover, the bounding expression is $C_0 N^{-s_c/r}$ for some $C_0$ independent of $N$ and $f$ since
    \begin{itemize}
        \item $h_N = \bigo(N^{-1/r})$, and as $\tau_f=\tau_c^-=\tau_c^+=s_c$ as discussed in the verification of assumptions, $h_N^{\max(\tau_f, \tau_c^-)} = \bigo(N^{-s_c/r})$,
        \item the rest of the multipliers do not depend on $N$ and $f$: $C$ depends only on $\calU$, $r$, $\tau_f = s_c$, and $B^*$; $\|f_\bU \|_{\calL^2(\calU)}$ is a constant and finite since $f_\bU$ is bounded above; $\tau_f-\tau_c^+=0$ so rising to its power produces $1$; the norm $\|f\|_{\calH_c} = 1$; for any $N \geq N_0$, $\mu(\beta_N) = 0$.
    \end{itemize}
    This completes the proof.
\end{proof}

Now we are ready to prove the main theorem.
\begin{proof}[Proof of \Cref{thm:rate_of_convergence}]
By~\Cref{thm:k_circ_G_is_Sobolev}, $k(x, \cdot) \circ G_\theta \in \calW^{\min(s_k, s), 2}(\calU)$, for a $G_\theta$ that satisfies~\Cref{as:generator}, and a Sobolev $k$ of smoothness $s_k$. By~\Cref{as:kernels_sbi}, $s_c \leq \min(s_k, s)$, and therefore $k(x, \cdot) \circ G_\theta \in \calH_c$ by~\Cref{res:sobolev-nested}. Then, we can use~\Cref{thm:optimal_weights} and state
    \begin{align*}
        | \MMD_k(\bP_\theta,\bQ_M) - \MMD_k(\bP_{\theta, N},\bQ_M)| \leq K \times \MMD_c \left(\bU, \sum_{n=1}^N w_n \delta_{u_n}\right).
    \end{align*}
    By the reproducing property, it holds that \begin{align*}
        \MMD_c\left(\bU, \sum_{n=1}^N w_n \delta_{u_n}\right) = \sup_{\substack{f \in \calH_c \\ \|f\|_{\calH_c}=1}} \left( \int_\calU f (u)\bU( \d u) - \sum_{n=1}^N w_n f(u_n) \right).
    \end{align*}
    The expression under the supremum is bounded by~\Cref{lemma:bq_convergence} with $C_0 N^{-s_c/r}$, for $C_0$ independent of $N$ and $f$. Therefore, $\MMD_c(\bU, \sum_{n=1}^N w_n \delta_{u_n}) \leq C_0 N^{-s_c/r}$, and the result holds.
\end{proof}

Note that while the result was formulated for the special case of convex spaces, it applies more generally to any open, connected, bounded $\calX \subset\bR^d$, $\calU \subset \bR^r$ with Lipschitz boundaries, with no changes to the proof. The applicability to $\calX=\bR^d$ remains unchanged; $\calU$, however, must remain bounded for~\Cref{thm:k_circ_G_is_Sobolev} to hold.

\subsection{Computational and sample complexity}\label{app:cost_error}

We derive the condition under which the OW estimator achieves better sample complexity than the V-statistic for the same order of computational cost, see Table~\ref{tab:cost_error} for the rates.

Suppose the cost for both V-statistic and OW is $\bigo(\tilde{N})$. Then, the sample complexity for the V-statistic can be written in terms of $\tilde N$ as $\bigo(\tilde{N}^{-1/4})$. Similarly, for the OW estimator, the sample complexity in terms of $\tilde N$ is $\bigo(\tilde{N}^{-s_c/3r})$. 
The more accurate estimator is therefore the one whose error rate goes to zero quicker. Therefore, the OW estimator is more accurate than the V-statistic if
\begin{align*}
   \nicefrac{ s_c }{r} > \nicefrac{3}{4}.
\end{align*}
For the common choice of Mat\'ern-5/2, $s_c=5/2 + r/2$, which implies $r< 10$.

\begin{table}[]
\centering
\caption[Computational and sample complexity of the V-statistic vs. OW-MMD.]{Computational and sample complexity rates of the V-statistic and the OW estimator with respect to $N$.}
\begin{tabular}{@{}lll@{}}
\toprule
            & \multicolumn{1}{c}{\textbf{Cost}} & \multicolumn{1}{c}{\textbf{Error}} \\ \midrule
V-statistic & $\bigo(N^2)$                & $\bigo(N^{-\nicefrac{1}{2}})$            \\
OW          & $\bigo(N^3)$                & $\bigo(N^{- \nicefrac{s_c}{r} })$    \\ \bottomrule
\end{tabular}
\label{tab:cost_error}
\end{table}

\section{Experimental details}
\label{app:expDetails}

True parameter values of the benchmark simulators in \Cref{sec:benchmark} is given in \Cref{app:trueParam}. \Cref{app:compositetest} provides details regarding the experiments in \Cref{sec:mvgk}. Finally, the link to the source code of the wind farm simulator is in \Cref{app:windfarm}.

\subsection{Benchmark Simulators}\label{app:trueParam}
We now provide further details on the benchmark simulators. For drawing i.i.d. or RQMC points, we use the implementation from \texttt{SciPy} \citet{SciPy}. Below, we report the parameter value $\theta$ used to generate the results in Table~\ref{tab:MMDerror} for each model. We refer the reader to the respective reference in Table~\ref{tab:MMDerror} for a description of the model and their parameters.

\noindent\textbf{g-and-k distribution}: $(A, B, g, k) = (3,1,0.1,0.1)$

\noindent\textbf{Two moons}: $(\theta_1, \theta_2) = (0,0)$

\noindent\textbf{Bivariate Beta}: $(\theta_1, \theta_2, \theta_3, \theta_4, \theta_5) = (1,1,1,1,1)$

\noindent\textbf{Moving average (MA) 2}: $(\theta_1, \theta_2) = (0.6,0.2)$

\noindent\textbf{M/G/1 queue}: $(\theta_1, \theta_2, \theta_3)= (1,5,0.2)$

\noindent\textbf{Lotka-Volterra}: $(\theta_{11}, \theta_{12}, \theta_{13})= (5,0.025,6)$

\subsection{Composite goodness-of-fit test}
\label{app:compositetest}

\begin{figure}[h]
  \setlength{\algomargin}{0.2em}
  \begin{minipage}[t]{\textwidth}
    \begin{algorithm2e}[H]
      \KwIn{$\bP_\theta$, $\bQ_M$, $\alpha$, $B$}
      $\hat{\theta}_{M} = \underset{\theta}{\argmin} \MMD_k^2(\bP_\theta, \bQ_M)$ \;
      
      \For{$b \in \{1, \ldots, B\}$}{
          $\bQ_{M,(b)} = \frac{1}{M} \sum_{m=1}^M \delta_{y_m^{(b)}}$, $y_1^{(b)}, \dots, y_M^{(b)} \sim \bP_{\hat{\theta}_M}$\;
          
          $\hat{\theta}_{(b)}^M = \underset{\theta \in \Theta}{\argmin} \MMD_k^2(\bP_\theta, \bQ_{M,(b)})$\;
          
          $\Delta_{(b)} = \MMD_k^2(\bP_{\hat{\theta}_{(b)}^M}, \bQ_{M,(b)})$\;
      }
      $c_\alpha = \mathrm{quantile}(\{\Delta_{(1)}, \ldots, \Delta_{(B)}\}, 1 - \alpha)$\;
      \BlankLine
      $\bP_{\hat{\theta}_M, N} = \frac{1}{N} \sum_{n=1}^N \delta_{x_n}$, where $x_1, \dots, x_N \sim \bP_{\hat{\theta}_M}$\;
      \eIf{$\MMD_k^2(\bP_{\hat{\theta}_M, N}, \bQ_M) > c_\alpha$}{\Return \text{reject}\;}{\Return do not reject\;}
    
      \caption{Composite goodness-of-fit test}
      \label{alg:composite_test}
    \end{algorithm2e}
  \end{minipage}
  \hfill
  \begin{minipage}[t]{\textwidth}
    \begin{algorithm2e}[H]
      \KwIn{$\bP_\theta$, $\bQ_M$, $N$, $I$, $R$, $S$, $\eta$, $\Theta^\text{init}$}

      \SetKwProg{Fn}{Function}{ is}{end}
      \Fn{loss($\theta$)}{
        $\bP_{\theta, N} = \frac{1}{N} \sum_{n=1}^N \delta_{x_n}$, where $x_1, \dots, x_N \sim \bP_\theta$ \;
        \Return $\MMD_k^2(\bP_{\theta, N}, \bQ_M)$\;
      }
      
      $\theta^\text{trial}_{(1)},\ldots,\theta^\text{trial}_{(I)} \sim \Theta^\text{init}$ \;
      Select $\theta_{(1)}^\text{init}, \ldots, \theta_{(R)}^\text{init} \in \{\theta^\text{trial}_{(i)}\}_{i=1}^I$ that yield the smallest $\mathrm{loss}(\theta^\text{init}_{(i)})$\;
      $\hat{\theta}^\text{opt}_{(1)},\ldots,\hat{\theta}^\text{opt}_{(R)} = $
      \For{$i \in \{1, \ldots, R\}$}{
          $\hat{\theta}^\text{opt}_{(i)} = $ adam\_optimizer(loss, $S$, $\eta$, $\theta^\text{init}_{(i)}$)
      }
      \Return $\theta^* \in \{\hat{\theta}^\text{opt}_\text{(i)}\}_{i=1}^R$ such that $\forall i,\ \mathrm{loss}(\theta^*) \leq \mathrm{loss}(\hat{\theta}_{(i)}^\text{opt})$\;
    
      \caption{Random-restart optimiser}
      \label{alg:optimisation}
    \end{algorithm2e}
  \end{minipage}
\end{figure}

\Cref{alg:composite_test} shows the details of the composite goodness-of-fit test using the parametric bootstrap. The algorithm is written for the V-statistic estimator, but each instance of the squared MMD can be replaced with our OW estimator.
In practice, to compute ${\argmin_\theta} \MMD_k^2(\bP_\theta, \bQ_M)$ we use gradient-based optimisation, as described in \Cref{alg:optimisation}.
The definitions of the hyperparameters of these two algorithms, and the values that we use, are given in \Cref{tab:hyper_def_app}.

\begin{table}[t]
\begin{center}
\centering
\caption[Hyperparameters in the composite goodness-of-fit test.]{Definitions of the hyperparameters.}
\label{tab:hyper_def_app}
\begin{tabular}{c c l}
    \toprule
    hyperparameter & value & \\
    \midrule
    $\alpha$ & 0.05 & level of the test  \\
    $B$ & 200 & number of bootstrap samples \\
    $N$ & 100 & number of samples from the simulator \\
    $M$ & 500 & number of observations in the data \\
    $I$ & 50 & number of initial parameters sampled \\
    $R$ & 10 & number of initial parameters to optimise \\
    $S$ & 200 & number of gradient steps \\
    $\eta$ & 0.04 & step size \\
    \bottomrule
\end{tabular}
\end{center}
\end{table}

$\Theta^\text{init}$ is the distribution from which the initial parameters are sampled, and is a uniform distribution with the following ranges: $\theta_1: (0.001, 5)$, $\theta_2: (0.001, 5)$, $\theta_3: (0.001, 1)$, $\theta_5: (0.001, 1)$.
To compute the fraction of times that the null hypothesis is rejected (\Cref{tab:composite_test_results}) we repeat the experiment $150$ times.

\label{app:gof}

\subsection{Large scale wind farm model}\label{app:windfarm}

The low-order wake model is described in \citet{Kirby2023} and the code is available at \url{https://github.com/AndrewKirby2/ctstar_statistical_model/blob/main/low_order_wake_model.py}.

\chapter{MMD-based Estimators for Conditional Expectations: Supplementary Materials}
\label{sec:app_cbq}
\section{Proofs of Theoretical Results}
\label{appendix:convergence_rate}

To validate our methodology, we established a rate at which the CBQ estimator converges to the true value of the conditional expectation $I$ in the $\calL^2(\Theta)$ norm, $\|I_\mathrm{CBQ} - I\|_{\calL^2(\Theta)}=\int_\Theta (I_\mathrm{CBQ}(\theta) - I(\theta))^2 \d \theta$ in~\Cref{thm:convergence_generalised}. The more specific version of this result was presented in the main text in~\Cref{thm:convergence}. In this section, we prove a more general version of~\Cref{thm:convergence} (as well as several intermediate results), and expand on the technical background required.

For the duration of the appendix, we will denote by $M$ the total number of points in $\Theta$ instead of $T$, to avoid notation clashes with the integral operator $T$. Additionally, we will be explicit on the dependency of the BQ mean $I_\BQ$ and variance $\sigma^2_\BQ$ at the point $\theta$ on the samples $x_{1:N}^\theta \sim \bP_\theta$, meaning
\begin{align*}
    I_\BQ(\theta; x_{1:N}^\theta) &= \mu^\top_\theta(x_{1:N}^\theta) \left(k_\calX(x_{1:N}^\theta, x_{1:N}^\theta) + \lambda_\calX \Id_N \right)^{-1} f(x_{1:N}^\theta, \theta) \\
    \sigma^2_\BQ(\theta; x_{1:N}^\theta) &= \bE_{X, X' \sim \bP_\theta} [k_\calX(X, X')] -\mu^\top_\theta(x_{1:N}^\theta) \\
    &\hspace{2cm}\left(k_\calX(x_{1:N}^\theta, x_{1:N}^\theta) + \lambda_\calX \Id_N \right)^{-1} \mu_\theta(x_{1:N}^\theta).
\end{align*}
Here, we added a `nugget' or `jitter' term, $\lambda_\calX \Id_N$ for a small $\lambda_\calX$, to the Gram matrix $k_\calX(x_{1:N},x_{1:N})$ to ensure it can be numerically inverted~\citep{Ababou1994,Andrianakis2012}. This is done to increase the generality of our results; the main text takes $\lambda_\calX=0$ to simplify presentation. 
Finally, we will shorten $x_{1:N}^{\theta_t}$ to $x_{1:N}^t$ to avoid bulky notation. The rest of the section is structured as follows. In~\Cref{sec:technical_assumptions} we present technical assumptions, and state in~\Cref{thm:convergence_generalised} the main convergence result the proof of which is deferred until the necessary Stage 1 and 2 results are proven. In~\Cref{sec:stage1}, we provide the necessary Stage 1 bounds that will be used in the proof of the main result. In~\Cref{sec:stage2}, we provide the necessary auxiliary results and the bound for Stage 2 in terms of Stage 1 errors. Finally, in~\Cref{sec:proof_of_main_theorem} we combine the bounds from both stages to prove~\Cref{thm:convergence_generalised}, the more general version of~\Cref{thm:convergence}.

\subsection{Main Result}
\label{sec:technical_assumptions}
Before presenting our findings, we list and justify the assumptions made. Throughout, we use Sobolev spaces defined in \Cref{sec:bg_sobolev_spaces} to quantify a function's smoothness.

We write $\theta = \begin{bmatrix} \theta_{(1)} & \dots & \theta_{(p)}\end{bmatrix}$ for any $\theta \in \Theta \subseteq \bR^p$.  For a multi-index $\alpha = (\alpha_1, \dots, \alpha_d) \in \bN^d$, by $D_\theta^\alpha g$ we denote the $|\alpha|=\sum_{i=1}^d \alpha_i$-th order weak derivative $D_\theta^\alpha g=D^{\alpha_1}_{\theta_{(1)}} \dots D^{\alpha_p}_{\theta_{(p)}} g$ for a function $g$ on $\Theta \subseteq \bR^p$.
Further, we assume the kernels $k_\Theta, k_\calX$ are Sobolev kernels; Mat\'ern kernels are important examples of Sobolev kernels.

The following is a more general form of the assumptions in~\Cref{thm:convergence}: specifically, we allow for the case when $\theta_{1:T}$ came from a distribution that does not necessarily have a density, and do not assume $\lambda_\calX=0$.
\begin{enumerate}[itemsep=0.1pt,topsep=0pt,leftmargin=*]
\item [B0] 
\begin{enumerate}
    \item[(a)] $f(x, \theta)$ lies in the Sobolev space $\calW^{s_f, 2}(\calX)$ for any $\theta \in \Theta$. 
    \customlabel{as:app_true_f_smoothness}{B0.(a)} 
    \item[(b)] $f(x, \theta)$ lies in the Sobolev space $\calW^{s_I, 2}(\Theta)$ for any $x \in \calX$.
    \customlabel{as:app_true_I_smoothness}{B0.(b)} 
    \item[(c)] $M_f = \sup_{\theta \in \Theta} \max_{|\alpha| \leq s_I} \| D_\theta^\alpha f(\cdot, \theta) \|_{\calW^{s_I, 2}(\calX)} < \infty$.
    \customlabel{as:app_true_I_norm_bounds}{B0.(c)} 
\end{enumerate}
\item [B1] 
\begin{enumerate}
    \item[(a)] $\calX \subset \bR^d$ is open, convex, and bounded.  
    \customlabel{as:app_domains_x}{B1.(a)} 
    \item[(b)] $\Theta\subset \bR^p$ is open, convex, and bounded. 
    \customlabel{as:app_domains_theta}{B1.(b)} 
\end{enumerate}
\item [B2]
\begin{enumerate}
    \item[(a)] $\theta_t$ were sampled i.i.d. from some $\bQ$, and $\bQ$ is equivalent to the uniform distribution on $\Theta$, meaning $\bQ(A)=0$ for a set $A \subset \Theta$ if and only if $\operatorname{Unif}(A)=0$. 
    \customlabel{as:app_theta_samples}{B2.(a)} 
    \item[(b)] $x_{1:N}^t \sim \bP_{\theta_t}$ for all $t \in \{1, \cdots, T\}$.  
    \customlabel{as:app_x_samples}{B2.(b)}
\end{enumerate}
\item [B3] $\bP_\theta$ has a density $p_\theta$ for any $\theta \in \Theta$, and the densities are such that 
    \begin{enumerate}
    \item[(a)] $\inf_{\theta \in \Theta, x \in \calX} p_\theta(x)=\eta>0$ and $\sup_{\theta \in \Theta}\|p_\theta\|_{\calL^2(\calX)}=\eta_0 < \infty$.
    \customlabel{as:app_densities1}{B3.(a)}
    \item[(b)] $p_\theta(x)$ lies in the Sobolev space $\calW^{s_I, 2}(\Theta)$ for any $x \in \calX$.
    \customlabel{as:app_densities2}{B3.(b)}
    \item[(c)] 
    $M_p = \sup_{\substack{\theta \in \Theta\\x \in \calX}}\max_{|\alpha|\leq s_I} |D_\theta^\alpha p_\theta(x)| < \infty$. 
    \customlabel{as:app_densities_linf}{B3.(c)}
    \end{enumerate}
\item [B4] 
\begin{enumerate}
    \item[(a)] $k_\calX$ is a Sobolev kernel of smoothness $s_\calX \in (d/2, s_f]$. 
    \customlabel{as:app_kernel_x}{B4.(a)}
    \item[(b)] $k_\Theta$ is a Sobolev kernel of smoothness $s_\Theta \in (p/2, s_I]$.
    \customlabel{as:app_kernel_theta}{B4.(b)}
    \item[(c)] $\kappa = \sup_{\theta \in \Theta}k_\Theta(\theta, \theta) < \infty$.
    \customlabel{as:app_kernel_theta_bounded}{B4.(c)}
\end{enumerate}
\item [B5]
\begin{enumerate}
    \item[(a)] $\lambda_\Theta = cM^{1/2}$, for $c>(4/C_6) \kappa \log(4/\delta)$ for some $C_6\leq1$. 
    \customlabel{as:app_regulariser_theta}{B5.(a)}
    \item[(b)] $\lambda_\calX \geq 0$. 
    \customlabel{as:app_regulariser_x}{B5.(b)}
\end{enumerate}
\end{enumerate}

Assumption B0 corresponds to conditions specified in the text of~\Cref{thm:convergence} prefacing the list of assumptions. Assumption~\ref{as:app_true_I_smoothness} implies $I(\theta) \in \calW^{s_I, 2}(\Theta)$: $f(x, \theta) p_\theta(x) \in \calW^{s_I, 2}(\Theta)$ by the product rule for weak derivatives (see, for instance,~\citet[Section 4.2.2]{evans2018measure}), and the integral lies in $\calW^{s_I, 2}(\Theta)$ by $\calW^{s_I, 2}(\Theta)$ being a complete space. 
Assumption~\ref{as:app_true_I_norm_bounds} ensures the $\calX$-Sobolev norm of any weak derivative of $\theta \to f(\cdot, \theta)$ is uniformly bounded across all $\theta$; this will be satisfied unless $f$ is so irregular that said Sobolev norms can get arbitrarily close to infinity.
Assumption~\ref{as:app_densities_linf}, similarly, ensures that any weak derivative of $\theta \to p_\theta(x)$ is bounded across all $\theta$ and $x$.
It is worth pointing out that assumption~\ref{as:app_kernel_theta_bounded}, boundedness of the kernel, follows from assumption~\ref{as:app_kernel_theta}; however, we keep it separate as some results will only require that the kernel is bounded, not necessarily that it is Sobolev.

Crucially, in the proofs in the next section we will see that the assumptions imply that the setting of the model in Stage 1 satisfies the assumptions of~\cite[Theorem 4]{wynne2021convergence}, and the setting of the model in Stage 2 satisfies the assumptions necessary to establish convergence of a noisy importance-weighted kernel ridge regression estimator, the two key results we will use to prove the convergence rate of the estimator.

We now state the main convergence result, which is a version of~\Cref{thm:convergence} for $\lambda_\calX \geq 0$. The proof of both this result and the more specific~\Cref{thm:convergence} are postponed until~\Cref{sec:proof_of_main_theorem}, as they rely on intermediary results.

\begin{theorem}[Generalised~\Cref{thm:convergence}]
\label{thm:convergence_generalised}
    Suppose all technical assumptions in~\Cref{sec:technical_assumptions} hold. Then for any $\delta \in (0, 1)$ there is an $N_0>0$ such that for any $N \geq N_0$, with probability at least $1-\delta$ it holds that
    \begin{align*}
        \| I_\mathrm{CBQ} - I \|_{\calL^2(\Theta, \bQ)} &\leq \left(1 + c^{-1} M^{-\frac{1}{2}}\left(\lambda_\calX + C_2 N^{-1 + 2\varepsilon} \left( N^{-\frac{s_\calX}{d}+\frac{1}{2} + \varepsilon} + C_3 \lambda_\calX \right)^2\right) \right) \\
        & \times \left( C_7(\delta) N^{-\frac{1}{2} + \varepsilon} \left( N^{-\frac{s_\calX}{d} + \frac{1}{2} + \varepsilon} + C_5 \lambda_\calX \right) + C_8(\delta) M^{-\frac{1}{4}} \| I\|_{\calH_\Theta} \right)
    \end{align*}
    for any arbitrarily small $\varepsilon>0$, constants $C_2, C_3, C_5$, $C_7(\delta) = \bigo(1/\delta)$ and $C_8(\delta) = \bigo(\log(1/\delta))$ independent of $N, M, \varepsilon$.
\end{theorem}

\subsection{Stage 1 bounds}
\label{sec:stage1}

Recall that we use the shorthand $x_{1:N}^t$ for $x_{1:N}^{\theta_t}$. In this section, we bound the BQ variance $\sigma^2_\BQ(\theta; x_{1:N}^\theta)$ in expectation in~\Cref{res:bound_on_bq_var}, and the difference between $I_\BQ(\theta; x_{1:N}^\theta)$ and $I$ in the norm of the RKHS $\calH_\Theta$ induced by the kernel $k_\Theta$ in~\Cref{res:bound_on_bq_error}. Later in~\Cref{sec:stage2}, the error of the estimator $I_\mathrm{CBQ}$ will be bounded in terms of these quantities.

\begin{theorem}
\label{res:bound_on_bq_var}
    Suppose Assumptions~\ref{as:app_true_f_smoothness},~\ref{as:app_domains_x},~\ref{as:app_densities1},~\ref{as:app_densities2},~\ref{as:app_kernel_x}, and~\ref{as:app_regulariser_x} hold. Then there is a $N_0>0$ such that for all $N \geq N_0$ it holds that
    \begin{align*}
        \bE_{y_1^\theta, \dots, y_N^\theta \sim \bP_\theta} \sigma^2_\BQ(\theta; y_{1:N}^\theta) \leq  \lambda_\calX + C_2 N^{-1 + 2\varepsilon} \left( N^{-\frac{s_\calX}{d}+\frac{1}{2} + \varepsilon} + C_3 \lambda_\calX \right)^2
    \end{align*}
    for any $\theta \in \Theta$, any arbitrarily small $\varepsilon>0$, and $C_2, C_3$ independent of $\theta,N,\varepsilon, \lambda_\calX$.
\end{theorem}

The term $N_0$ quantifies how likely the points $y_{1:N}^\theta$ are to `fill out' the space $\calX$, for any $\theta$. Intuitively speaking, $N_0$ is smallest when for all $\theta$, the $\bP_\theta$ is uniform. 

\begin{proof}
Recall
\begin{align*}
    I_\BQ(\theta; y_{1:N}^\theta) & = \mu_\theta(y_{1:N}^\theta)^\top \left(k_{\calX}(y_{1:N}^\theta, y_{1:N}^\theta)+ \lambda_\calX \Id_N\right)^{-1} f(y_{1:N}^\theta, \theta),\\
    \sigma^2_\BQ(\theta; y_{1:N}^\theta) &= \bE_{X,X'\sim \bP_\theta}[k_{\calX}(X,X')] \\ &\hspace{2cm}-\mu_\theta(y_{1:N}^\theta)^\top \left(k_{\calX}(y_{1:N}^\theta, y_{1:N}^\theta)+ \lambda_\calX \Id_N\right)^{-1} \mu_\theta(y_{1:N}^\theta).
\end{align*}
We seek to bound $\sigma^2_\BQ(\theta; y_{1:N}^\theta)$. \citet[Proposition 3.8]{kanagawa2025gaussian} pointed out that the Gaussian noise posterior is the worst-case error in the $\calH_{\calX}^{\lambda_\calX}$, the RKHS induced by the kernel $k_\calX^{\lambda_\calX}(x, x') = k_\calX(x, x') + \lambda_\calX \delta(x, x')$ (where $\delta(x, x') = 1$ if $x=x'$, and $0$ otherwise). 
Through straightforward algebraic manipulations and using the reproducing property, one can show that for the vector $w_\theta = \left[\int_\calX k_\calX(x, y_{1:N}^\theta) \bP_\theta(\d x) \right]^\top \left(k_{\calX}(y_{1:N}^\theta, y_{1:N}^\theta)+ \lambda_\calX \Id_N\right)^{-1} \in \bR^N$, 
\begin{align}
\label{eq:variance_bound_proof_1}
    \sigma^2_\BQ(\theta; y_{1:N}^\theta) - \lambda_\calX =\sup_{\|f\|_{\calH_{\calX}^{\lambda_\calX}} \leq 1} \left| w_\theta f(y_{1:N}^\theta) - \int_\calX f(x) \bP_\theta(\d x)\right|^2.
\end{align}
Since $\calH_\calX^{\lambda_\calX}$ is induced by the sum of kernels, $k_\calX^{\lambda_\calX}(x, x') = k_\calX(x, x') + \lambda_\calX \delta(x, x')$, it holds that $\calH_\calX \subseteq \calH_\calX^{\lambda_\calX}$, and $\| f  \|_{\calH_\calX^{\lambda_\calX}} \leq \| f  \|_{\calH_\calX}$~\citep[Theorem I.13.IV]{aronszajn1950theory}. Therefore, the class of functions $f$ for which $\| f  \|_{\calH_\calX} \leq 1$ is larger than that for which $\| f  \|_{\calH_\calX^{\lambda_\calX}} \leq 1$, and
\begin{align}
\label{eq:variance_bound_proof_2}   \sup_{\|f\|_{\calH_{\calX}^{\lambda_\calX}} \leq 1} \left| w_\theta f(y_{1:N}^\theta) - \int_\calX f(x) \bP_\theta(\d x)\right| \leq \sup_{\|f\|_{\calH_{\calX}} \leq 1} \left| w_\theta f(y_{1:N}^\theta) - \int_\calX f(x) \bP_\theta(\d x)\right|.
\end{align}
Next, note that for $\hat{f}_\theta(x) = k_\calX(x, y_{1:N}^\theta)^\top \left(k_{\calX}(y_{1:N}^\theta, y_{1:N}^\theta)+ \lambda_\calX \Id_N\right)^{-1} f(y_{1:N}^\theta)$,
\begin{align}
\begin{split}
\label{eq:variance_bound_proof_3}
   \left| w_\theta f(y_{1:N}^\theta) - \int_\calX f(x) \bP_\theta(\d x)\right| &= \left| \int_\calX \left(\hat{f}_\theta(x) - f(x) \right) \bP_\theta(\d x)\right| \\
   &\leq  \int_\calX \left|\hat{f}_\theta(x) - f(x) \right| \bP_\theta(\d x)  \\
    & \leq \|\hat{f}_\theta - f\|_{\calL^2(\calX)} \|p_\theta \|_{\calL^2(\calX)},
\end{split}
\end{align}
where the last inequality is an application of H\"older's inequality. By Assumption~\ref{as:app_densities1}$, \|p_\theta \|_{\calL^2(\calX)}$ is bounded above by $\eta_0$. In order to apply~\citep[Theorem 4]{wynne2021convergence} to bound $\|\hat{f}_\theta - f\|_{\calL^2(\calX)}$, we show the assumptions of that Theorem hold.
    
{Assumption 1 (Assumptions on the Domain):} An open, bounded, and convex $\calX$ satisfies the assumption, as discussed in~\citet{wynne2021convergence}.
    
{Assumption 2 (Assumptions on the Kernel Parameters) and Assumption 3 (Assumptions on the Kernel Smoothness Range):} Our setting is more specific than the one~\citep[Theorem 4]{wynne2021convergence}: the kernel $k_\calX$ is Mat\'ern, and therefore all smoothness constants mentioned in Assumptions 2 and 3 have the same value, $s_\calX$.

{Assumption 4 (Assumptions on the Target Function and Mean Function):} The target function $f$ was assumed to have higher smoothness than $k_\calX$ in~\ref{as:app_true_f_smoothness}, and~\ref{as:app_kernel_x}; the mean function was taken to be zero.

{Assumption 5 (Additional Assumptions on Kernel Parameters):} By \ref{as:app_kernel_x} and~\ref{as:app_true_f_smoothness} the smoothness of the true function $s_f \geq s_\calX >d/2$, which verifies both statements in the Assumption since all smoothness constants of the kernel are equal to $s_\calX$.

Therefore~\citep[Theorem 4]{wynne2021convergence} holds, and for $\calW^{0,2}(\calX)=\calL^2(\calX)$
\begin{align*}
    \|\hat{f}_\theta - f\|_{\calL^2(\calX)} \leq K_3 \| f \|_{\calH_\calX} h_{y_{1:N}^\theta}^{\frac{d}{2}} \left( h_{y_{1:N}^\theta}^{s_\calX-\frac{d}{2}} + \lambda_\calX \right),
\end{align*}
for any $N$ for which the fill distance $h_{y_{1:N}^\theta} \leq h_0$ for some $h_0$, and $K_3$ and $h_0$ that depend on $\calX, s_f, s_\calX$.\footnote{Note that the result in~\citep[Theorem 4]{wynne2021convergence} features $\| f \|_{\calW^{s_\calX,2}(\calX)}$, not $\| f \|_{\calH_\calX}$. The bound in terms $\| f \|_{\calH_\calX}$ holds since $\calH_\calX$ was assumed to be a Sobolev RKHS.}

For $y_{1:N}^\theta \sim \bP_\theta$, we can guarantee that $h_{y_{1:N}^\theta} \leq h_0$ in expectation using~\citep[Lemma 2]{oates2019convergence}, which says that provided the density $\inf_{x} p_\theta(x)>0$, there is a $C_\theta$ such that $\bE h_{y_{1:N}^\theta} \leq C_\theta N^{-1/d + \varepsilon}$ for an arbitrarily small $\varepsilon>0$, for $C_\theta$ that depends on $\theta$ through $\inf_{x} p_\theta(x)$. The smaller $\inf_{x} p_\theta(x)$, the larger $C_\theta$. Since we assumed $\inf_{x, \theta} p_\theta(x)=\eta>0$ there is a $K_4$ such that $C_\theta \leq K_4$ for any $\theta$. Therefore, we may take $N_0$ to be the smallest $N$ for which $\bE h_{y_{1:N}^\theta} \leq K_4 N^{-1/d + \varepsilon}$ holds, and have for all $N \geq N_0$
\begin{equation}
\label{eq:wynne_bound_on_exp}
    \bE_{y_1^\theta, \dots, y_N^\theta \sim \bP_\theta}\|\hat{f}_\theta - f\|_{\calL^2(\calX)} \leq K_3 K_4^{\frac{d}{2}} \| f \|_{\calH_\calX} N^{-\frac{1}{2} + \varepsilon} \left( K_4^{s_\calX-\frac{d}{2}} N^{-\frac{s_\calX}{d} + \frac{1}{2} + \varepsilon} + \lambda_\calX \right).
\end{equation}
Putting together~\cref{eq:variance_bound_proof_1,eq:variance_bound_proof_2,eq:variance_bound_proof_3,eq:wynne_bound_on_exp} and Assumption~\ref{as:app_densities1}, we get the result,
\begin{align*}
    \bE_{y_1^\theta, \dots, y_N^\theta \sim \bP_\theta} &\sigma^2_\BQ(\theta; y_{1:N}^\theta) - \lambda_\calX \\
    &=
    \sup_{\|f\|_{\calH_{\calX}^{\lambda_\calX}} \leq 1} \bE_{y_1^\theta, \dots, y_N^\theta \sim \bP_\theta} \left| w_\theta f(y_{1:N}^\theta) - \int_\calX f(x) \bP_\theta(\d x)\right|^2 \\
    &\leq 
    \sup_{\|f\|_{\calH_{\calX}} \leq 1} \bE_{y_1^\theta, \dots, y_N^\theta \sim \bP_\theta} \left| w_\theta f(y_{1:N}^\theta) - \int_\calX f(x) \bP_\theta(\d x)\right|^2 \\
    &\leq 
    \sup_{\|f\|_{\calH_{\calX}} \leq 1} \bE_{y_1^\theta, \dots, y_N^\theta \sim \bP_\theta} \|\hat{f}_\theta - f\|^2_{\calL^2(\calX)} \|p_\theta \|^2_{\calL^2(\calX)} \\
    &\leq 
    \eta^2_0 K^2_3 K_4^d N^{-1 + 2\varepsilon} \left( K_4^{s_\calX-\frac{d}{2}} N^{-\frac{s_\calX}{d} + \frac{1}{2} + \varepsilon} + \lambda_\calX \right)^2 \\
    &\eqqcolon C_2 N^{-1 + 2\varepsilon} \left( N^{-\frac{s_\calX}{d}+\frac{1}{2} + \varepsilon} + C_3 \lambda_\calX \right)^2.
\end{align*}
\end{proof}
Before bounding the error $\|I_\BQ - I\|_{\calH_\Theta}$, we give the following general auxiliary result for an arbitrary Sobolev space of function over some open $\Omega \subseteq \bR^d$.

\begin{proposition}
\label{res:sobolev_product_bound}
    Suppose $f, g$ lie in a Sobolev space $\calW^{s,2}(\Omega)$ for some of smoothness $s$, and for all $|\alpha|\leq s$ the weak derivative $D^\alpha g$ is bounded. Take $M = \max_{|\alpha|\leq s} \|D^\alpha g\|_{\calL^\infty(\Omega)}$. Then, there is a constant $K$ such that
    \begin{equation*}
        \|fg\|_{\calW^{s,2}(\Omega)} \leq KM \|f\|_{\calW^{s,2}(\Omega)}.
    \end{equation*}
\end{proposition}
\begin{proof}
    Recall that the norm in a Sobolev space is defined as
    \begin{equation}
    \label{eq:sobolev_norm_of_product_defn}
        \|fg\|^2_{\calW^{s,2}(\Omega)} = \sum_{|\alpha|\leq s} \|D^\alpha[fg]\|^2_{\calL^2(\Omega)}.
    \end{equation}
    Fix some $\alpha$ such that $|\alpha|\leq s$. By the product rule to weak derivatives (see, for instance,~\citet[Section 4.2.2]{evans2018measure}), it holds that
    \begin{equation*}
        D^{\alpha} [fg] = \sum_{\substack{|\alpha'|\leq|\alpha|}} \sum_{\substack{|\alpha''|\leq|\alpha|}} C_{\alpha',\alpha'',\alpha} D^{\alpha'} [f] D^{\alpha''} [g],
    \end{equation*}
    for all $\alpha', \alpha''$ being multi-indices of the same dimension as $\alpha$, and some real constants $C_{\alpha',\alpha'',\alpha}>0$ that only depend on $\alpha$ and not $f$ or $g$. Then
    \begin{align*}
        \|D^{\alpha}[fg]\|^2_{\calL^2(\Omega)} 
        &= \left\|\sum_{\substack{|\alpha'|\leq|\alpha|}} \sum_{\substack{|\alpha''|\leq|\alpha|}} C_{\alpha',\alpha'',\alpha} D^{\alpha'} [f] D^{\alpha''} [g]\right\|^2_{\calL^2(\Omega)} \\
        &\stackrel{(A)}{\leq} 
        \left(\sum_{\substack{|\alpha'|\leq|\alpha|}} \sum_{\substack{|\alpha''|\leq|\alpha|}} C_{\alpha',\alpha'',\alpha} \| D^{\alpha'} [f] D^{\alpha''} [g]\|_{\calL^2(\Omega)} \right)^2 \\
        &\stackrel{(B)}{\leq} 
        2 \begin{pmatrix} d\\|\alpha| \end{pmatrix} \sum_{\substack{|\alpha'|\leq|\alpha|}} \sum_{\substack{|\alpha''|\leq|\alpha|}} C_{\alpha',\alpha'',\alpha} 
        \| D^{\alpha'} [f] D^{\alpha''} [g]\|^2_{\calL^2(\Omega)} \\
        &\stackrel{(C)}{\leq} 
        2M^2 \begin{pmatrix} d\\|\alpha| \end{pmatrix} \sum_{\substack{|\alpha'|\leq|\alpha|}} \sum_{\substack{|\alpha''|\leq|\alpha|}} C_{\alpha',\alpha'',\alpha} 
        \| D^{\alpha'} [f] \|^2_{\calL^2(\Omega)} \\
        &\leq 2M^2 \begin{pmatrix} d\\|\alpha| \end{pmatrix} \sum_{\substack{|\alpha'|\leq|\alpha|}} \sum_{\substack{|\alpha''|\leq|\alpha|}} C_{\alpha',\alpha'',\alpha} 
        \| f \|^2_{\calW^{s,2}(\Omega)},
    \end{align*}
    where $(A)$ holds by triangle inequality, $(B)$ holds as, by Cauchy-Schwarz, $(\sum_{i=1}^n a_i)^2 \leq n \sum_{i=1}^n a_i^2$ for any real $a_i$, and as the number of multi-indices in $\bN^d$ of size at most $\alpha$ is `$d$ choose $|\alpha|$', and $(C)$ by the definition $M = \max_{|\alpha|\leq s} \|D^\alpha g\|_{\calL^\infty}(\Omega)$. Substituting this into~\eqref{eq:sobolev_norm_of_product_defn}, we get that for $K = 2\sum_{|\alpha| \leq s} \begin{pmatrix} d\\|\alpha| \end{pmatrix} \sum_{\substack{|\alpha'|\leq|\alpha|}} \sum_{\substack{|\alpha''|\leq|\alpha|}} C_{\alpha',\alpha'',\alpha}$,
    \begin{equation*}
        \|fg\|^2_{\calW^{s,2}(\Omega)} \leq K^2 M^2 \| f \|^2_{\calW^{s,2}(\Omega)}.
    \end{equation*}
\end{proof}
With the Sobolev norm bound in place, we are ready to give the bound on $\|I_\BQ - I\|_{\calH_\Theta}$.
\begin{theorem}
\label{res:bound_on_bq_error}
    Suppose Assumptions~\ref{as:app_true_f_smoothness},~\ref{as:app_true_I_norm_bounds},~\ref{as:app_domains_x},~\ref{as:app_x_samples}, \ref{as:app_densities1}, \ref{as:app_densities2}, \ref{as:app_densities_linf}, \ref{as:app_kernel_x}, \ref{as:app_kernel_theta} and~\ref{as:app_regulariser_x} hold. Then there is a $N_0>0$ such that for all $N \geq N_0$ with probability at least $1-\delta/2$ it holds that
    \begin{equation*}
        \|I_\BQ - I\|_{\calH_\Theta} \leq \frac{2}{\delta} C_4 N^{-\frac{1}{2} + \varepsilon} \left( N^{-\frac{s_\calX}{d} + \frac{1}{2} + \varepsilon} + C_5 \lambda_\calX \right).
    \end{equation*}
    for any arbitrarily small $\varepsilon>0$, and $C_4, C_5$ independent of $N,\varepsilon, \lambda_\calX$.
\end{theorem}
\begin{proof}
    Recall that, as $\calH_\Theta$ is a Sobolev RKHS (meaning $k_\Theta$ is a Sobolev kernel) of smoothness $s_\Theta$, it holds that $C'_1 \|g\|_{\calW^{s_\Theta, 2}(\Theta)} \leq \|g\|_{\calH_\Theta} \leq C'_2 \|g\|_{\calW^{s_\Theta, 2}(\Theta)}$ for some constants $C'_1,C'_2>0$ and any $g \in \calH_\Theta$.
    Take $\hat{f}(x, \theta) = k_\calX(x, x_{1:N}^\theta)^\top \left(k_{\calX}(x_{1:N}^\theta, x_{1:N}^\theta)+ \lambda_\calX \Id_N\right)^{-1} f(x_{1:N}^\theta, \theta)$. Then,
    \begin{align*}
        \|I_\BQ &- I\|^2_{\calH_\Theta} \\
        &= 
        \langle I_\BQ - I, I_\BQ - I\rangle_{\calH_\Theta}  \\
        &= 
        \left\langle \int_\calX \left( \hat f(x, \theta) - f(x, \theta) \right) p_\theta(x)  \mathrm{d} x, \int_\calX \left( \hat f(x', \theta) - f(x', \theta) \right) p_\theta(x')  \mathrm{d} x' \right\rangle_{\calH_\Theta}  \\
        &\leq \int_\calX \int_\calX \left\langle   \left( \hat f(x, \theta) - f(x, \theta) \right) p_\theta(x) , \left( \hat f(x', \theta) - f(x', \theta) \right) p_\theta(x')  \right\rangle_{\calH_\Theta} \mathrm{d} x \mathrm{d} x' \\
        &\stackrel{(A)}{\leq} 
        \left(\int_\calX \left\| \left( \hat f(x, \theta) - f(x, \theta) \right) p_\theta(x)  \right\|_{\calH_\Theta} \mathrm{d} x \right)^2 \\
        &\stackrel{(B)}{\leq} 
        {C'_2}^2 K^2 M_p^2 \left(\int_\calX \left\| \hat f(x, \theta) - f(x, \theta) \right\|_{\calW^{s_\Theta, 2}(\Theta)} \mathrm{d} x \right)^2,
    \end{align*}
    where $(A)$ holds by the Cauchy-Schwarz, $(B)$ by~\Cref{res:sobolev_product_bound} and $\calH_\Theta$ being a Sobolev RKHS. 
    As for the remaining term,
    \begin{align*}
        \int_\calX &\left\| \hat f(x, \theta) - f(x, \theta) \right\|^2_{\calW^{s_\Theta, 2}(\Theta)} \d x \\
        &= 
        \sum_{|\alpha|\leq s_\Theta} \int_\calX \int_\Theta \left( D_\theta^\alpha \hat f(x, \theta) - D_\theta^\alpha f(x, \theta) \right)^2 \d \theta \d x \\
        &= 
        \sum_{|\alpha|\leq s_\Theta} \int_\Theta \int_\calX \left( D_\theta^\alpha \hat f(x, \theta) - D_\theta^\alpha f(x, \theta) \right)^2 \d x \d \theta \\
        &= 
        \sum_{|\alpha|\leq s_\Theta} \int_\Theta \left\| D_\theta^\alpha \hat f(x, \theta) - D_\theta^\alpha f(x, \theta) \right\|^2_{\calL^2(\calX)} \d \theta
    \end{align*}
    Since $D_\theta^\alpha \hat{f}(x, \theta) = k_\calX(x, x_{1:N}^\theta)^\top \left(k_{\calX}(x_{1:N}^\theta, x_{1:N}^\theta)+ \lambda_\calX \Id_N\right)^{-1} D_\theta^\alpha f(x_{1:N}^\theta, \theta)$, and the $\calX$-smoothness of $D_\theta^\alpha f$ is the same as that of $f$, we may use~\citet[Theorem 4]{wynne2021convergence} to bound $\|D_\theta^\alpha \hat f(x, \theta) - D_\theta^\alpha f(x, \theta)\|_{\calL^2(\calX)}$ identically to the proof of~\Cref{res:bound_on_bq_var}. Then, we have that
    \begin{align*}
        \bE_{x_1^\theta, \dots, x_N^\theta \sim \bP_\theta}&\|D_\theta^\alpha \hat f(x, \theta) - D_\theta^\alpha f(x, \theta)\|_{\calL^2(\calX)} \\
        &\leq 
        K_3 K_4^{\frac{d}{2}} \| D_\theta^\alpha f \|_{\calH_\calX} N^{-\frac{1}{2} + \varepsilon} \left( K_4^{s_\calX-\frac{d}{2}} N^{-\frac{s_\calX}{d} + \frac{1}{2} + \varepsilon} + \lambda_\calX \right) \\
        &\stackrel{(A)}{\leq} 
        K_3 K_4^{\frac{d}{2}} C'_2 M_f N^{-\frac{1}{2} + \varepsilon} \left( K_4^{s_\calX-\frac{d}{2}} N^{-\frac{s_\calX}{d} + \frac{1}{2} + \varepsilon} + \lambda_\calX \right),
    \end{align*}
    where $(A)$ holds by Assumption~\ref{as:app_true_I_norm_bounds}, and $k_\calX$ being a Sobolev kernel and $C'_2$ being a norm equivalence constant. 
    By Markov's inequality, for any $\delta/2 \in (0,1)$ it holds with probability at least $1-\delta/2$ that 
    \begin{align*}
        \|D_\theta^\alpha \hat f(x, \theta) &- D_\theta^\alpha f(x, \theta)\|_{\calL^2(\calX)} \\
        &\leq \frac{2}{\delta}K_3 K_4^{\frac{d}{2}} C'_2 M_f N^{-\frac{1}{2} + \varepsilon} \left( K_4^{s_\calX-\frac{d}{2}} N^{-\frac{s_\calX}{d} + \frac{1}{2} + \varepsilon} + \lambda_\calX \right)
    \end{align*}
    Lastly, the number of $\alpha$ such that $|\alpha| \leq s_\Theta$ is the combination `$p$ choose $s_\Theta$'. Then,
    \begin{align*}
        \|I_\BQ &- I\|^2_{\calH_\Theta}  \\
        &\leq  
        {C'_2}^2 K^2 M_p^2 \begin{pmatrix} p \\ s_\Theta \end{pmatrix} \left( \frac{2}{\delta}K_3 K_4^{\frac{d}{2}} C'_2 M_f N^{-\frac{1}{2} + \varepsilon} \left( K_4^{s_\calX-\frac{d}{2}} N^{-\frac{s_\calX}{d} + \frac{1}{2} + \varepsilon} + \lambda_\calX \right) \right)^2 \\
        &\eqqcolon
        \frac{4}{\delta^2} C_4^2 N^{-1 + 2\varepsilon} \left( N^{-\frac{s_\calX}{d} + \frac{1}{2} + \varepsilon} + C_5 \lambda_\calX \right) ^2.
    \end{align*}
\end{proof}

\subsection{Stage 2 bounds}
\label{sec:stage2}

In this section, we establish convergence of the estimator $I_\mathrm{CBQ}$ to the true function $I$ in the norm $\calL^2(\Theta, \bQ)$, first in terms of the error $\|I_\BQ(\cdot; x^\theta_{1:N}) - I(\cdot)\|_{\calH_\Theta}$ in~\Cref{res:convergence_of_iwkrr}, and additionally in the variance $\sigma^2_\BQ(\theta; x^\theta_{1:N})$ in~\Cref{res:unweighted_bound_stage_2}. To do so, we represent the CBQ estimator as
\begin{equation}
\label{eq:cbq_noisy_weights}
\begin{split}
    I_\mathrm{CBQ}(\theta) &= k_\Theta(\theta, \theta_{1:M})^\top \left( k_\Theta(\theta_{1:M}, \theta_{1:M}) + \diag \left[ \frac{M \lambda}{w(\theta_{1:M}) + \varepsilon(\theta_{1:M}; x^{1:M}_{1:N})} \right] \right)^{-1} \\
    &\hspace{3cm} \times I_\BQ(\theta_{1:M}; x^{1:M}_{1:N}).
\end{split}
\end{equation}
for vector notation $\varepsilon(\theta_{1:M}; x^{1:M}_{1:N}) = [\varepsilon(\theta_1; x^1_{1:N}), \ldots,  \varepsilon(\theta_M; x^M_{1:N}) ]^\top \in \bR^M$, and $\lambda$, the weight $w: \Theta \to \bR$ and the noise term $\varepsilon: \Theta \to \bR$ given by
\begin{equation}
\begin{split}
\label{eq:weights}
    &\lambda = \lambda_\Theta M^{-1} \\
    &w(\theta) = \bE_{y_1^\theta, \dots, y_N^\theta \sim \bP_\theta}\frac{\lambda_\Theta}{\lambda_\Theta + \sigma^2_\BQ(\theta; y^\theta_{1:N})},\\
    &\varepsilon(\theta; x^\theta_{1:N}) = \frac{\lambda_\Theta}{\lambda_\Theta + \sigma^2_\BQ(\theta; x^\theta_{1:N})} - \bE_{y_1^\theta, \dots, y_N^\theta \sim \bP_\theta} \frac{\lambda_\Theta}{\lambda_\Theta + \sigma^2_\BQ(\theta; y^\theta_{1:N})}.
\end{split}
\end{equation}
The equality to the CBQ estimator given in the main text can be easily seen, as the term under the $\diag$ is
\begin{equation*}
    \frac{M \lambda}{w(\theta_{1:M}) + \varepsilon(\theta_{1:M}; x^{1:M}_{1:N})} = \frac{M \lambda_\Theta M^{-1}}{\frac{\lambda_\Theta}{\lambda_\Theta + \sigma^2_\BQ(\theta; x^\theta_{1:N})}} = \lambda_\Theta + \sigma^2_\BQ(\theta; x^\theta_{1:N}).
\end{equation*}
If the noise term in~\eqref{eq:cbq_noisy_weights} were absent (meaning, equal to zero), the estimator would become the \emph{importance-weighted kernel ridge regression} (IW-KRR) estimator. The convergence of the IW-KRR estimator was studied in~\citet[Theorem 4]{gogolashvili2023importance}. In this section, we extend their results to the case of noisy weights ($\varepsilon \not\equiv 0$), which are additionally correlated with the noise in $I_\BQ(\theta_i; x^i_{1:N})$ (through the shared datapoints $x^i_{1:N}$). 

Note that, while we only provide results specific for $I_\mathrm{CBQ}$, the proof can be extended with minor modifications to the more general case of arbitrary noisy IW-KRR with weights that satisfy conditions in~\citet{gogolashvili2023importance}, and zero-mean weight noise.

The convergence results for the \emph{noisy importance-weighted kernel ridge regression} estimator in~\Cref{sec:convergence_of_niwkrr} will rely on a representation of $I_\mathrm{CBQ}$ in terms of a sample-level version of a certain weighted integral operator. Then, we bound the gap between $I_\mathrm{CBQ}$ and $I$ in terms of (1) the gap between the sample-level version of said operator, and the population-level version, and (2) the gap between $I_\BQ$ and $I$. Next, we define said operator, and additional notation used in the proofs.

\subsubsection{Notation}
\label{sec:notation}
We will be working on positive, bounded, self-adjoint $\calH_\Theta \to \calH_\Theta$ operators 
\begin{equation}
\begin{split}
    T[g](\theta) &= \int_\Theta k_\Theta(\theta, \theta') g(\theta') w(\theta') \bQ(\d \theta') \\
    \hat{T}[g](\theta) &= \frac{1}{M} k_\Theta(\theta, \theta_{1:M}) \diag\left[w(\theta_{1:M}) + \varepsilon(\theta_{1:M}; x^{1:M}_{1:N}) \right] g(\theta_{1:M}).
\end{split}
\end{equation}
for the weight function $w$ and noise term $\varepsilon$ as defined in~\eqref{eq:weights}.
We will denote $\mathrm{HS}$ to be the Hilbert space of Hilbert-Schmidt operators $\calH_\Theta \to \calH_\Theta$, $\|\cdot\|_\text{HS}$ to be the Hilbert-Schmidt norm, and $\|\cdot\|_\mathrm{op}$ to be the operator norm. As is customary, we will write $T+\lambda$ to mean the operator $T+\lambda \Id_{\calH_\Theta}$, where $\Id_{\calH_\Theta}$ is the identity operator $\calH_\Theta \to \calH_\Theta$.

\subsubsection{Auxiliary results}
\label{sec:auxilliary_results_stage_2}

The results given in this section are key to proving the main Stage 2 result,~\Cref{res:convergence_of_iwkrr}. The following result bounds the Hilbert-Schmidt norm on the `gap' between the population-level $T$ and the sample-level $\hat T$, when their difference is `sandwiched' between $(T+\lambda)^{-1/2}$. With some manipulation, this term will appear in the proof of~\Cref{res:convergence_of_iwkrr}.
\begin{lemma}[Modified Lemma 18 in~\citet{gogolashvili2023importance}]
\label{res:s1}
Suppose Assumptions~\ref{as:app_x_samples},~\ref{as:app_kernel_theta_bounded} hold, and the operators $T, \hat{T}$ be as defined in \Cref{sec:notation}. Then, with probability greater than $1 - \delta/2$,
    \begin{equation*}
        S_1 \coloneqq \|(T+\lambda)^{-1/2} (T-\hat{T}) (T+\lambda)^{-1/2} \|_{\mathrm{HS}} \leq \frac{4\kappa}{\lambda \sqrt M} \log(4/\delta).
    \end{equation*}
Additionally, if $\lambda \sqrt{M} > (4/C_6) \kappa \log(4/\delta)$ for some $C_6\leq1$, it holds that $S_1 < C_6 \leq 1$.
\end{lemma}
The fact that $S_1$ is strictly less than $1$ will be important in the proof of the main Stage 2 result,~\Cref{res:convergence_of_iwkrr}, as it will allow us to apply Neumann series expansion to $\|(\Id - (T+\lambda)^{-1/2} (T-\hat{T}) (T+\lambda)^{-1/2})^{-1}\|_\mathrm{op}$.

\begin{proof}
    Denote a feature function $\varphi_\theta (\cdot) \coloneqq k_\Theta(\theta, \cdot)$. Let $\xi, \xi_1, \dots, \xi_M$ be random variables in $\mathrm{HS}$ defined as
    \begin{align*}
        \xi &= (T+\lambda)^{-1/2} (w(\theta)+\varepsilon(\theta; x^\theta_{1:N})) \varphi_\theta \langle \varphi_\theta, \cdot \rangle_{\calH_\Theta} (T+\lambda)^{-1/2} \\
        \xi_i &= (T+\lambda)^{-1/2} (w(\theta_i) + \varepsilon(\theta_i; x^i_{1:N})) \varphi_{\theta_i} \langle \varphi_{\theta_i}, \cdot \rangle_{\calH_\Theta} (T+\lambda)^{-1/2}
    \end{align*}
    
    First, note that as $(\theta, x_{1:N}), (\theta_1, x^{1}_{1:N}),\dots,(\theta_M, x^M_{1:N})$ are i.i.d., it follows that $\xi, \xi_1, \dots, \xi_M$ are i.i.d. random variables in $\mathrm{HS}$. Further, as $\bE_{x_1^\theta, \dots, x_N^\theta \sim \bP_\theta} \varepsilon(\theta; x^\theta_{1:N}) =0$, it holds that
    \begin{align*}
        \bE_{\theta \sim \bQ} \bE_{x_1^\theta, \dots, x_N^\theta \sim \bP_\theta} \xi 
        &= \bE_{\theta \sim \bQ} \left[(T+\lambda)^{-1/2} w(\theta) \varphi_\theta \langle \varphi_\theta, \cdot \rangle_{\calH_\Theta} (T+\lambda)^{-1/2}\right] \\
        &=(T+\lambda)^{-1/2} T (T+\lambda)^{-1/2},
    \end{align*}
    where the identity $\bE_{\theta \sim \bQ} \left[ w(\theta) \varphi_\theta \langle \varphi_\theta, \cdot \rangle_{\calH_\Theta} \right]g(\theta') = \bE_{\theta \sim \bQ}  w(\theta) k_\Theta(\theta, \theta') g(\theta)$, which holds for any $g \in \calH_\Theta$, is used to get the last equality. Therefore 
    \begin{align*}
        S_1 = \left\| \frac{1}{M} \sum_{i=1}^M \xi_i - \bE_{\theta \sim \bQ} \bE_{x_1^\theta, \dots, x_N^\theta \sim \bP_\theta} \xi \right\|_{\mathrm{HS}}.
    \end{align*} 
    Then, by the Bernstein inequality for Hilbert space-valued random variables~\citep[Proposition 2]{caponnetto2007optimal}, the claimed bound on $S_1$ holds if there exist $L>0, \sigma>0$ such that
    \begin{equation*}
        \bE_{\theta \sim \bQ} \bE_{x_1^\theta, \dots, x_N^\theta \sim \bP_\theta} \left[\| \xi - \bE_{\theta \sim \bQ} \bE_{x_1^\theta, \dots, x_N^\theta \sim \bP_\theta} \xi  \|^m_{\mathrm{HS}}\right] \leq \frac{1}{2} m! \sigma^2 L^{m-2}
    \end{equation*}
    holds for all integer $m \geq 2$. We will show the condition holds.
    For convenience, denote $\bE_\xi f(\xi) \coloneqq \bE_{\theta \sim \bQ} \bE_{x_1^\theta, \dots, x_N^\theta \sim \bP_\theta} f(\xi)$. 
    First, suppose $\xi'$ is an independent copy of $\xi$. Identically to the proof of~\citet[Lemma 18]{gogolashvili2023importance}, it holds that
    \begin{align*}
        \bE_\xi\left[\| \xi - \bE_\xi \xi  \|^m_{\mathrm{HS}}\right] \stackrel{(A)}{\leq} \bE_\xi \bE_{\xi'} \left[\| \xi - \xi'  \|^m_{\mathrm{HS}}\right] &\stackrel{(B)}{\leq} 2^{m-1} \bE_\xi \bE_{\xi'}\left[\| \xi \|^m_{\mathrm{HS}} +  \|\xi'  \|^m_{\mathrm{HS}}\right] \\
        &= 2^m \bE_\xi \| \xi \|^m_{\mathrm{HS}},
    \end{align*}
    where $(A)$ holds by Jensen inequality, and $(B)$ uses $|a+b|^m \leq 2^{m-1}(|a|^m + |b|^m)$. Next, observe that
    \begin{align*}
        \bE_\xi \| \xi \|^m_{\mathrm{HS}} 
        &= 
        \bE_{\theta \sim \bQ} \bE_{x_1^\theta, \dots, x_N^\theta \sim \bP_\theta} \|(T+\lambda)^{-1/2} (w(\theta)\\
        &\hspace{4cm}+\varepsilon(\theta; x^\theta_{1:N})) \varphi_\theta \langle \varphi_\theta, \cdot \rangle_{\calH_\Theta} (T+\lambda)^{-1/2} \|^m_{\mathrm{HS}} \\
        &\stackrel{(A)}{=} 
        \bE_{\theta \sim \bQ} \Big[ \bE_{x_1^\theta, \dots, x_N^\theta \sim \bP_\theta} \left[ (w(\theta)+\varepsilon(\theta; x^\theta_{1:N}))^m \right]  \\
        &\hspace{4cm} \times \|(T+\lambda)^{-1/2}  \varphi_\theta \langle \varphi_\theta, \cdot \rangle_{\calH_\Theta} (T+\lambda)^{-1/2} \|^m_{\mathrm{HS}}  \Big] \\
        & \stackrel{(B)}{\leq}
        \bE_{\theta \sim \bQ} \left[  \|(T+\lambda)^{-1/2}  \varphi_\theta \langle \varphi_\theta, \cdot \rangle_{\calH_\Theta} (T+\lambda)^{-1/2} \|^m_{\mathrm{HS}}  \right] \\
        & \stackrel{(C)}{\leq} 
        \kappa^m \lambda^{-m} \\
        &= 
        \frac{1}{2}m! \sigma^2 L^{m-2},
    \end{align*}
    where $L=\sigma=\kappa \lambda^{-1}$, $(A)$ holds by linearity of norms as $w(\theta)+\varepsilon(\theta; x^\theta_{1:N}) \in \bR$, $(B)$ holds since $\sigma^2_\BQ(\theta; x^\theta_{1:N}) \geq 0$, so
    \begin{equation*}
        \left(w(\theta) + \varepsilon(\theta; x^\theta_{1:N})\right)^m = \left(\frac{\lambda_\Theta}{\lambda_\Theta + \sigma^2_\BQ(\theta; x^\theta_{1:N})}\right)^m \leq 1.
    \end{equation*}
    To show $(C)$ holds, take $\{e_j\}_{j=1}^\infty$ to be some orthonormal basis of $\calH_\Theta$. Then,
    \begin{align*}
        \|(T+\lambda)^{-1/2}  \varphi_\theta &\langle \varphi_\theta, \cdot \rangle_{\calH_\Theta} (T+\lambda)^{-1/2} \|^2_{\mathrm{HS}} \\
        &= 
        \sum_{j=1}^\infty \|(T+\lambda)^{-1/2}  \varphi_\theta \langle \varphi_\theta, \cdot \rangle_{\calH_\Theta} (T+\lambda)^{-1/2} e_j \|^2_{\calH_\Theta} \\
        &= 
        \sum_{j=1}^\infty \|(T+\lambda)^{-1/2}  \varphi_\theta \langle \varphi_\theta, (T+\lambda)^{-1/2} e_j \rangle_{\calH_\Theta}  \|^2_{\calH_\Theta} \\
        &\leq 
        \|(T+\lambda)^{-1/2}  \varphi_\theta  \|^2_{\calH_\Theta} \sum_{j=1}^\infty \langle \varphi_\theta, (T+\lambda)^{-1/2} e_j \rangle^2_{\calH_\Theta}  \\
        &= 
        \|(T+\lambda)^{-1/2}  \varphi_\theta  \|^2_{\calH_\Theta} \sum_{j=1}^\infty \langle (T+\lambda)^{-1/2} \varphi_\theta,  e_j \rangle^2_{\calH_\Theta}  \\
        &\stackrel{(A)}{\leq}
        \|(T+\lambda)^{-1/2}  \varphi_\theta  \|^2_{\calH_\Theta} \|(T+\lambda)^{-1/2}  \varphi_\theta  \|^2_{\calH_\Theta}  \\
        &= 
        \langle (T+\lambda)^{-1}  \varphi_\theta  , \varphi_\theta  \rangle^2_{\calH_\Theta}  \\
        &\leq \kappa^2 \lambda^{-2},
    \end{align*}
    where $(A)$ holds by Bessel's inequality. Then by the Bernstein inequality in~\citet[Proposition 2]{caponnetto2007optimal}, it holds that
    \begin{equation*}
        S_1 \leq \frac{2\kappa}{\lambda \sqrt M} \left(\frac{1}{\sqrt M} + 1\right) \log(4/\delta) \leq \frac{4\kappa}{\lambda \sqrt M} \log(4/\delta),
    \end{equation*}
    with probability at least $1 - \delta/2$. As $\lambda \sqrt{M} > (16/3) \kappa \log(4/\delta)$, $S_1  < 3/4$.
\end{proof}

Next, we bound another relevant term that also quantifies the `gap' between $T$ and $\hat T$. Unlike $S_1$, we will not require it to be upper bounded by $1$, as it will only appear in~\Cref{res:convergence_of_iwkrr} as a bounding term to the error.

\begin{lemma}
\label{res:s2}
Suppose Assumptions~\ref{as:app_x_samples},~\ref{as:app_kernel_theta_bounded} hold, and the operators $T, \hat{T}$ be as defined in \Cref{sec:notation}. Then, with probability greater than $1 - \delta/2$,
    \begin{equation*}
        S_2 \coloneqq \|(T+\lambda)^{-1/2} (T-\hat{T})\|_{\mathrm{HS}} \leq \frac{4\kappa}{ \sqrt {\lambda M}} \log(4/\delta).
    \end{equation*}
Additionally, if $\lambda \sqrt{M} > (4/C_6) \kappa \log(4/\delta)$, it holds that $S_2 < C_6 \sqrt \lambda$.
\end{lemma}
\begin{proof}
    The proof is identical to that of~\Cref{res:s1}.
\end{proof}

The last auxiliary result we need is a simple bound on the following operator norm.
\begin{lemma}
\label{res:t_t_plus_lambda_inv_bounded}
Let $T: \calH_\Theta \to \calH_\Theta$ be a positive operator. Then,
    \begin{equation*}
        \| T  (T+\lambda)^{-1} \|_\mathrm{op} \leq 1.
    \end{equation*}
\end{lemma}
\begin{proof}
    Since $T$ is positive, for any $f \in \calH_\Theta$ it holds that $\|Tf\|_{\calH_\Theta} \leq \|(T+\lambda) f\|_{\calH_\Theta}$. Therefore, by taking $f=(T+\lambda)^{-1} g$, we get that
    \begin{align*}
        \|T (T+\lambda)^{-1} \|_\mathrm{op} 
        &= \sup_{\substack{g \in \calH_\Theta \\ \|g\|_{\calH_\Theta}=1}} \|T (T+\lambda)^{-1} g \|_{\calH_\Theta} \\
        &\leq \sup_{\substack{g \in \calH \\ \|g\|_{\calH_\Theta}=1}} \|(T+\lambda)  (T+\lambda)^{-1} g\|_{\calH_\Theta}  = 1.
    \end{align*}
\end{proof}

\subsubsection{Convergence of the noisy IW-KRR estimator}
\label{sec:convergence_of_niwkrr}

With the auxiliary results in place, we now extend~\citet[Theorem 4]{gogolashvili2023importance} to the case of noisy weights. We start by establishing convergence in $\calL^2(\Theta, \bQ_w)$, where $\bQ_w$ is the measure defined as $\bQ_w(A) = \int_A w(\theta)\bQ(\mathrm{d} \theta)$ that must be finite and positive. By~\citet[Proposition 232D]{fremlin2000measure}, for $\bQ_w(A)$ to be a finite positive measure, it is sufficient for $w(\theta)$ to be continuous and bounded. By their definition in~\eqref{eq:weights}, 
\begin{equation*}
    w(\theta) = \bE_{y_1^\theta, \dots, y_N^\theta \sim \bP_\theta}\frac{\lambda_\Theta}{\lambda_\Theta + \sigma^2_\BQ(\theta; y^\theta_{1:N})}
\end{equation*}
the weights are bounded by $1$, and are continuous in $\theta$ if $p_\theta$ is continuous in $\theta$ (as the dependence of $\sigma^2_\BQ(\theta; y^\theta_{1:N})$ on $\theta$ for a fixed $y^\theta_{1:N}$ is, again, only through $p_\theta$ appearing under integrals and in polynomials). The continuity of $p_\theta$ holds as, by~\ref{as:app_true_I_smoothness},~\ref{as:app_kernel_theta}, $p_\theta$ lies in a Sobolev space of smoothness over $p/2$, and therefore by Sobolev embedding theorem~\citep[Theorem 4.12]{adams2003sobolev} $p_\theta$ is continuous in $\theta$.

\begin{theorem}
\label{res:convergence_of_iwkrr}
Suppose Assumptions~\ref{as:app_true_I_smoothness},~\ref{as:app_domains_theta},~\ref{as:app_theta_samples},~\ref{as:app_x_samples}, and~\ref{as:app_kernel_theta_bounded} hold, and $\lambda \sqrt{M} > (4/C_6) \kappa \log(4/\delta)$ for some $C_6\leq1$. Then,
\begin{align*}
    \| I_\mathrm{CBQ} - I \|_{\calL^2(\Theta, \bQ_w)} &\leq (1-C_6)^{-1} \left( C_6 \sqrt \lambda + 1\right) \|I_\BQ-I \|_{\calH_\Theta} \\
    &\hspace{2cm}+ \left(\frac{8 (1-C_6)^{-1} \kappa}{ \sqrt {\lambda M}} \log(4/\delta)  + \sqrt \lambda \right) \| I\|_{\calH_\Theta}.
\end{align*}
\end{theorem}
\begin{proof}
    First, note that $I_\mathrm{CBQ}(\theta) = (\hat{T} + \lambda)^{-1}\hat{T}[I_\BQ]$, 
    which can be checked easily by seeing that $(\hat{T} + \lambda)I_\mathrm{CBQ}(\theta) = \hat{T}[I_\BQ]$ for the \emph{weighted} operator $\hat T$ as defined in~\Cref{sec:notation} and $I_\mathrm{CBQ}$ as defined in~\eqref{eq:cbq_noisy_weights}.
    Then, for $I_{\lambda} = (T+\lambda)^{-1} T[I]$, by triangle inequality the error is bounded as
    \begin{align}
    \label{eq:i_cbq_minus_i_triangle}
        \| I_\mathrm{CBQ} - I \|_{\calL^2(\Theta, \bQ_w)} \leq \| I_\mathrm{CBQ} - I_\lambda \|_{\calL^2(\Theta, \bQ_w)} + \| I_\lambda - I \|_{\calL^2(\Theta, \bQ_w)}
    \end{align}
    The second term, $\| I_\lambda - I \|^2_{\calL^2(\Theta, \bQ_w)}$, can be bounded in terms of $\lambda$ as
    \begin{align}
        \| I_\lambda - I \|_{\calL^2(\Theta, \bQ_w)} 
        &= 
        \| \lambda (T+\lambda)^{-1}[I] \|_{\calL^2(\Theta, \bQ_w)} \nonumber \\
        &= 
        \| \lambda T^{1/2} (T+\lambda)^{-1}[I] \|_{\calH_\Theta} \nonumber \\
        &\stackrel{(A)}{\leq} 
        \lambda \|  T (T+\lambda)^{-1} \|^{1/2}_\mathrm{op} \|(T+\lambda)^{-1/2}  \|_\mathrm{op} \| I \|_{\calH_\Theta} \nonumber \\
        &\leq 
        \sqrt{\lambda} \| I \|_{\calH_\Theta} , \label{eq:i_lambda_minus_i_bound}
    \end{align}
    where $(A)$ holds by~\Cref{res:t_t_plus_lambda_inv_bounded} and $T$ being a positive operator.
    Next, the $\calL^2(\Theta, \bQ_w)$ norm between $I_\mathrm{CBQ} - I_{\lambda}$ can be bounded as
    \begin{align*}
        \| I_\mathrm{CBQ} - &I_{\lambda} \|_{\calL^2(\Theta, \bQ_w)} \\
        &= \| T^{1/2} (I_\mathrm{CBQ} - I_{\lambda}) \|_{\calH_\Theta} \\
        &= 
        \| T^{1/2} ((\hat{T}+\lambda)^{-1} \hat{T}[I_\BQ] - (T+\lambda)^{-1} T[I] ) \|_{\calH_\Theta} \\
        &\stackrel{(A)}{\leq}
        \| T  (T+\lambda)^{-1} \|_\mathrm{op}^{1/2} \| ( \Id - (T+\lambda)^{-1/2} (T - \hat{T}) (T+\lambda)^{-1/2} )^{-1} \|_\mathrm{op} \\
        &\qquad \times \Big(\| (T+\lambda)^{-1/2} (\hat{T}[I_\BQ]-T[I]) \|_{\calH_\Theta} \\
        &\qquad\qquad\qquad+ \| (T+\lambda)^{-1/2} (T-\hat{T}) (T+\lambda)^{-1} T[I] \|_{\calH_\Theta} \Big) \\
        &\stackrel{(B)}{\leq} 
        \| T  (T+\lambda)^{-1} \|_\mathrm{op}^{1/2} \| ( \Id - (T+\lambda)^{-1/2} (T - \hat{T}) (T+\lambda)^{-1/2} )^{-1} \|_\mathrm{op} \\
        &\qquad \times \Big( \| (T+\lambda)^{-1/2} \hat{T}[I_\BQ-I] \|_{\calH_\Theta} \\
        &\qquad\qquad\qquad+ \| (T+\lambda)^{-1/2} (T - \hat T)[I]\|_{\calH_\Theta} \\
        &\qquad\qquad\qquad+ \| (T+\lambda)^{-1/2} (T-\hat{T}) (T+\lambda)^{-1} T[I] \|_{\calH_\Theta} \Big) \\
        &\eqqcolon 
        U_0 \times U_1 \times (U_2 + U_3 + U_4),
    \end{align*}
    where $\|\cdot\|_\mathrm{op}$ denotes the operator norm, $(A)$ holds by~\citet[Lemma 17]{gogolashvili2023importance}, and $(B)$ is an application of triangle inequality,
    \begin{align*}
        \| (T+\lambda)^{-1/2} (\hat{T}[I_\BQ]-T[I]) \|_{\calH_\Theta} &\leq \| (T+\lambda)^{-1/2} \hat{T}[I_\BQ-I] \|_{\calH_\Theta} \\
        &\qquad\qquad+ \| (T+\lambda)^{-1/2} (T - \hat T)[I] \|_{\calH_\Theta}.
    \end{align*}
    We will bound the terms $U_0,U_1, U_2, U_3, U_4$, and the result will follow. First, we have that $U_0 = \| T  (T+\lambda)^{-1} \|_\mathrm{op}^{1/2} \leq 1$ by \Cref{res:t_t_plus_lambda_inv_bounded}. To upper bound $U_1=\| ( \Id - (T+\lambda)^{-1/2} (T - \hat{T}) (T+\lambda)^{-1/2} )^{-1} \|_\mathrm{op}$ we may expand it as Neumann series, if $\| (T+\lambda)^{-1/2} (T - \hat{T}) (T+\lambda)^{-1/2} \|_\mathrm{op} \eqqcolon \| B_\lambda \|_\mathrm{op} < 1$. This holds as
    \begin{align*}
        \| B_\lambda \|_\mathrm{op} \stackrel{(A)}{\leq} \| B_\lambda \|_{\mathrm{HS}} \stackrel{(B)}{ < } C_6\leq 1,
    \end{align*}
    where $(A)$ holds as the operator norm is bounded by the Hilbert-Schmidt norm, and $(B)$ by~\Cref{res:s1}. 
    Therefore,
    \begin{align*}
        &\left\| ( \Id - (T+\lambda)^{-1/2} (T - \hat{T}) (T+\lambda)^{-1/2} )^{-1} \right\|_\mathrm{op} 
        \\
        &\hspace{2cm}\stackrel{(A)}{=} \left\| \sum_{i=0}^\infty \left((T+\lambda)^{-1/2} (T - \hat{T}) (T+\lambda)^{-1/2} \right)^i \right\|_\mathrm{op} \\
        &\hspace{2cm}\stackrel{(B)}{\leq} \sum_{i=0}^\infty \left\| (T+\lambda)^{-1/2} (T - \hat{T}) (T+\lambda)^{-1/2}  \right\|^i_\mathrm{op}\\
        &\hspace{2cm}\stackrel{(C)}{\leq} \sum_{i=0}^\infty \left\| (T+\lambda)^{-1/2} (T - \hat{T}) (T+\lambda)^{-1/2}  \right\|^i_{\mathrm{HS}}\\
        &\hspace{2cm}\stackrel{(D)}{=} \left(1 - \left\| (T+\lambda)^{-1/2} (T - \hat{T}) (T+\lambda)^{-1/2}  \right\|_{\mathrm{HS}}\right)^{-1}\\
        &\hspace{2cm}\stackrel{(E)}{\leq} (1 - C_6)^{-1},
    \end{align*}
    where $(A)$ holds by the Neumann series expansion, $(B)$ by the triangle inequality, and the fact that operator norm is sub-multiplicative for bounded operators, $(C)$ since the operator norm is bounded by the Hilbert-Schmidt norm, $(D)$ by the geometric series, and $(E)$ by~\Cref{res:s1}.

    To bound $U_2 = \| (T+\lambda)^{-1/2} \hat{T}[I_\BQ-I] \|_{\calH_\Theta}$, observe that
    \begin{align*}
        U_2 &= \| (T+\lambda)^{-1/2} \hat{T}[I_\BQ-I] \|_{\calH_\Theta} \\
        &\leq \| (T+\lambda)^{-1/2} \hat{T} \|_\mathrm{op} \|I_\BQ-I \|_{\calH_\Theta} \\
        &\leq
        \left( \| (T+\lambda)^{-1/2} (T - \hat T) \|_\mathrm{op} + \| (T+\lambda)^{-1/2} T \|_\mathrm{op} \right) \|I_\BQ-I \|_{\calH_\Theta} \\
        &\stackrel{(A)}{\leq}
        \left( S_2 + 1\right) \|I_\BQ-I \|_{\calH_\Theta},
    \end{align*}
    where $(A)$ holds by~\Cref{res:s2,res:t_t_plus_lambda_inv_bounded}.

    Both $U_3$ and $U_4$ are upper bounded by the $S_2$ term in~\Cref{res:s2}, as 
    \begin{align*}
        U_3 &=\| (T+\lambda)^{-1/2} (T - \hat T)[I]\|_{\calH_\Theta} \leq \| (T+\lambda)^{-1/2} (T - \hat T)\|_\mathrm{op} \| I \|_{\calH_\Theta} \\
        &\hspace{5.4cm}= S_2 \| I \|_{\calH_\Theta}, \\
        U_4 &=\| (T+\lambda)^{-1/2} (T-\hat{T}) (T+\lambda)^{-1} T[I] \|_{\calH_\Theta} \\
        &\leq
        \| (T+\lambda)^{-1/2} (T - \hat T)\|_\mathrm{op} \| (T+\lambda)^{-1} T \|_\mathrm{op} \| I \|_{\calH_\Theta} \\
        &\stackrel{(A)}{\leq} 
        S_2 \| I \|_{\calH_\Theta},
    \end{align*}
    where $(A)$ holds by~\Cref{res:t_t_plus_lambda_inv_bounded}. 
    Putting the upper bounds on $U_0, U_1, U_2, U_3, U_4$ together, we get
    \begin{align*}
        \| I_\mathrm{CBQ} - I_{\lambda} \|_{\calL^2(\Theta, \bQ_w)} &\leq U_0 \times U_1 \times (U_2 + U_3 + U_4) \\
        &\leq 4 \left(( S_2 + 1) \|I_\BQ-I \|_{\calH_\Theta} + 2S_2 \| I\|_{\calH_\Theta} \right).
    \end{align*}
    By applying the union bound, we get that that with probability at least $1-\delta$,
    \begin{align*}
        \| I_\mathrm{CBQ} &- I_{\lambda} \|_{\calL^2(\Theta, \bQ_w)}  \\
        &\leq 
        U_0 \times U_1 \times (U_2 + U_3 + U_4) \\
        &\leq 
        (1-C_6)^{-1} \left(( S_2 + 1) \|I_\BQ-I \|_{\calH_\Theta} + 2S_2 \| I\|_{\calH_\Theta} \right) \\
        &\stackrel{(A)}{\leq}
        (1-C_6)^{-1} \left(\left( C_6 \sqrt \lambda + 1\right) \|I_\BQ-I \|_{\calH_\Theta} + \frac{8\kappa}{ \sqrt {\lambda M}} \log(4/\delta) \| I\|_{\calH_\Theta} \right),
    \end{align*}
    where $(A)$ holds by~\Cref{res:s2}. Inserting this and the bound in~\eqref{eq:i_lambda_minus_i_bound} into~\eqref{eq:i_cbq_minus_i_triangle} gives
    \begin{align*}
        \| I_\mathrm{CBQ} - I \|_{\calL^2(\Theta, \bQ_w)} &\leq (1-C_6)^{-1} \left( C_6 \sqrt \lambda + 1\right) \|I_\BQ-I \|_{\calH_\Theta} \\
        &\hspace{2cm}+ \left(\frac{8 (1-C_6)^{-1} \kappa}{ \sqrt {\lambda M}} \log(4/\delta)  + \sqrt \lambda \right) \| I\|_{\calH_\Theta}.
    \end{align*}
\end{proof}
Finally, we use the $\calL^2(\Theta, \bQ_w)$ bound in~\Cref{res:convergence_of_iwkrr} to establish a bound in $\calL^2(\Theta, \bQ)$ in terms of the BQ variance.
\begin{corollary}
\label{res:unweighted_bound_stage_2}
    Suppose Assumptions~\ref{as:app_true_I_smoothness},~\ref{as:app_domains_theta},~\ref{as:app_theta_samples},~\ref{as:app_x_samples}, and~\ref{as:app_kernel_theta_bounded}, and $\lambda \sqrt{M} > (4/C_6) \kappa \log(4/\delta)$ for some $C_6\leq1$. Then
    \begin{align*}
        \| I_\mathrm{CBQ} - I \|_{\calL^2(\Theta, \bQ)} &\leq \left(1 + \frac{1}{c \sqrt{M}} \sup_{\theta \in \Theta} \bE_{y_1^\theta, \dots, y_N^\theta \sim \bP_\theta}\sigma^2_\BQ(\theta; y^\theta_{1:N})\right) \\
        &\qquad \times \Bigg( (1-C_6)^{-1} \left( C_6 \sqrt \lambda + 1\right) \|I_\BQ-I \|_{\calH_\Theta} \\
        &\qquad \qquad + \left(\frac{8 (1-C_6)^{-1} \kappa}{ \sqrt {\lambda M}} \log(4/\delta)  + \sqrt \lambda \right) \| I\|_{\calH_\Theta}\Bigg)
    \end{align*}
\end{corollary}
\begin{proof}
    Observe that for any $g \in \calL^2(\Theta, \bQ)$, it holds that 
    \begin{equation*}
        \|g\|^2_{\calL^2(\Theta, \bQ_w)} \geq \left(\inf_{\theta \in \Theta} w(\theta) \right) \times \|g\|^2_{\calL^2(\Theta, \bQ)}.
    \end{equation*}
    Then, since
    \begin{align*}
        w(\theta) = \bE_{y_1^\theta, \dots, y_N^\theta \sim \bP_\theta}\frac{\lambda_\Theta}{\lambda_\Theta + \sigma^2_\BQ(\theta; y^\theta_{1:N})} &\geq \frac{\lambda_\Theta}{\lambda_\Theta + \bE_{y_1^\theta, \dots, y_N^\theta \sim \bP_\theta} \sigma^2_\BQ(\theta; y^\theta_{1:N})} \\
        &= \frac{1}{1 + \lambda_\Theta^{-1} \bE_{y_1^\theta, \dots, y_N^\theta \sim \bP_\theta} \sigma^2_\BQ(\theta; y^\theta_{1:N})},
    \end{align*}
    the bound in~\Cref{res:convergence_of_iwkrr}, the definition of $\lambda$ in~\eqref{eq:weights}, and Assumption~\ref{as:app_regulariser_theta} give the desired statement.
\end{proof}

\subsection{Proof of \Cref{thm:convergence}}
\label{sec:proof_of_main_theorem}
We are now ready to prove our main convergence result, which is a version of~\Cref{thm:convergence} for $\lambda_\calX \geq 0$. We start by restating it for the convenience of the reader.

\begin{proof}[Restatement of~\Cref{thm:convergence_generalised}]
    Suppose all technical assumptions in~\Cref{sec:technical_assumptions} hold. Then for any $\delta \in (0, 1)$ there is an $N_0>0$ such that for any $N \geq N_0$, with probability at least $1-\delta$ it holds that
    \begin{align*}
        \| I_\mathrm{CBQ} &- I \|_{\calL^2(\Theta, \bQ)} \\
        &\leq \left(1 + c^{-1} M^{-\frac{1}{2}}\left(\lambda_\calX + C_2 N^{-1 + 2\varepsilon} \left( N^{-\frac{s_\calX}{d}+\frac{1}{2} + \varepsilon} + C_3 \lambda_\calX \right)^2\right) \right) \\
        &\qquad \times \left( C_7(\delta) N^{-\frac{1}{2} + \varepsilon} \left( N^{-\frac{s_\calX}{d} + \frac{1}{2} + \varepsilon} + C_5 \lambda_\calX \right) + C_8(\delta) M^{-\frac{1}{4}} \| I\|_{\calH_\Theta} \right)
    \end{align*}
    for any arbitrarily small $\varepsilon>0$, constants $C_2, C_3, C_5$, $C_7(\delta) = \bigo(1/\delta)$ and $C_8(\delta) = \bigo(\log(1/\delta))$ independent of $N, M, \varepsilon$.
\end{proof}

\begin{proof}[Proof of~\Cref{thm:convergence_generalised}]
    By inserting~\Cref{res:bound_on_bq_error,res:bound_on_bq_var} into~\Cref{res:unweighted_bound_stage_2} and applying the union bound, we get that the result holds with probability at least $1-\delta$ and
    \begin{align*}
        C_7(\delta) &= (1-C_6)^{-1} \left( C_6 c^{\frac{1}{2}} + 1\right) C_4 (2/\delta), \\
        C_8(\delta) &= \left( 8c^{-\frac{1}{2}} (1-C_6)^{-1} \kappa \log(4/\delta)  + c^{\frac{1}{2}} \right).
    \end{align*}
\end{proof}

As discussed in the main text, convergence is fastest when the regulariser $\lambda_\calX$ is set to $0$; $\lambda_\calX>0$ ensures greater stability at the cost of a lower speed of convergence. For clarity we show how~\Cref{thm:convergence} in the main text follows from the more general~\Cref{thm:convergence_generalised} by setting $\lambda_\calX=0$. 
    
\begin{proof}[Proof of~\Cref{thm:convergence}]         
    In~\Cref{thm:convergence_generalised}, take $\lambda_\calX=0$. Then
    \begin{align*}
        \| I_\mathrm{CBQ} - I \|_{\calL^2(\Theta, \bQ)} &\leq \left(1 + c^{-1} M^{-\frac{1}{2}} C_2 N^{-\frac{2s_\calX}{d} + \varepsilon} \right) \\
        &\hspace{2cm}\times \left( C_7(\delta) N^{-\frac{s_\calX}{d} + \varepsilon} + C_8(\delta) M^{-\frac{1}{4}} \| I\|_{\calH_\Theta} \right).
    \end{align*}
    As $\bQ$ was assumed equivalent to the uniform distribution in Assumption~\ref{as:app_theta_samples}, the error in uniform measure is bounded by the error in $\bQ$. Therefore, the result holds for
    \begin{align*}
        C_0(\delta)&=\left(1 + c^{-1} C_2 \right) C_7(\delta)= \bigo(1/\delta), \\
        C_1(\delta)&=\left(1 + c^{-1} C_2 \right) \|I\|_{\calH_\Theta} C_8(\delta) = \bigo(\log(1/\delta)).
    \end{align*}
\end{proof}

\section{Hyperparameter Selection}
\label{appendix:hyperparameter_selection}

In this section, we discuss selection of hyperparameters for CBQ, as well as the baselines methods in~\Cref{sec:experiments_cbq}.

\paragraph{Conditional Bayesian quadrature.} The hyperparameter selection for CBQ boils down to the choice of GP interpolation hyperparameters at stage 1 and the choice of GP regression hyperparameters at stage 2. To simplify this choice, we renormalise all our function values before performing GP regression and interpolation. This is done by first subtracting the empirical mean and then dividing by the empirical standard deviation. 
All of our experiments then use prior mean functions $m_\Theta$ and $m_\calX$ which are zero functions, a reasonable choice given the function was renormalised using the empirical mean. This choice is made for simplicity, and we might expect further improvements in accuracy if more information is available.

The choice of covariance functions $k_\calX$ and $k_\Theta$ is made on a case-by-case basis in order to both encode properties we expect the target functions to have, but also to ensure that the corresponding kernel mean is available in closed-form (as per the previous section). Once this is done, we typically still need to make a choice of hyperparameters for both kernel: lengthscales $l_\calX$, $l_\Theta$ and amplitudes $A_\calX, A_\Theta$. 
We also need to select the regularizer $\lambda_\calX, \lambda_\Theta$. 
$\lambda_\calX$ is fixed to be $0$ as suggested by \Cref{thm:convergence}.
The rest of the hyperparameters are selected through empirical Bayes, which consists of maximising the log-marginal likelihood.
For stage 1, the log-marginal likelihood can be written as~\citep{rassmussen2006gaussian}
\begin{align*}
& L(l_\calX, A_\calX) =  -\frac{1}{2} \log \det( k_{\calX}(x_{1:N},x_{1:N}; l_\calX, A_\calX) ) - \frac{N}{2} \log(2 \pi) \\
& \quad -\frac{1}{2}(f(x_{1:N})-m_{\calX}(x_{1:N}))^\top \left(k_{\calX}(x_{1:N},x_{1:N};l_\calX, A_\calX) + \lambda_\calX \Id_N \right)^{-1} \\
&\hspace{7cm}\times(f(x_{1:N})-m_{\calX}(x_{1:N})),
\end{align*}
where $\det$ denotes the determinant of the matrix, and $l_\calX, A_\calX$ are explicitly included in $k$ to emphasise the hyperparameters used to compute the Gram matrix.
The optimisation is implemented through a grid search over $\left[1.0, 10.0, 100.0, 1000.0 \right]$ for the amplitude $A_\calX$ and a grid search over $\left[0.1, 0.3, 1.0, 3.0, 10.0 \right]$ for the lengthscale $l_\calX$. 

If $k_\calX$ is a Stein reproducing kernel, we have an extra hyperparameter $c_\calX$. 
In this case, we use stochastic gradient descent on the log-marginal likelihood to find the optimal value for $c_\calX, l_\calX, A_\calX$, which is implemented with \texttt{JAX} autodiff library~\citep{jax2018github}. 
The reason we are using gradient-based optimisation instead of grid search for the Stein kernel is that the Stein kernel requires an accurate estimate of $c_\calX$ to work well. 
In order to return accurate results, grid search would require a finer grid, which is very expensive, while gradient-based methods would require good initialisation to avoid getting stuck in local minima. Fortunately, since $c_\calX$ indicates the mean of functions in the RKHS, we know that $c_\calX = 0$ is a good initialisation point since we have subtracted the empirical mean when normalising.

Additionally, it is important to note that we could technically use $T$ different kernels $k_\calX^1, \cdots, k_\calX^T$ for each integral in stage 1. However, the hyperparameters of each kernel $k_\calX^t$ would need to be selected using empirical Bayes under the observations $x_{1:N}^t$, which means we would need to repeat the above optimisation $T$ times. In practice, when performing initial experiments, we observed that the estimated hyperparameters were very similar. Our strategy is therefore to select the hyperparameters of $k_\calX^1$ and subsequently reuse them across all $T$ integrals in stage 1. This is done for computational reasons, and we expect CBQ to show better performances if hyperparameters are optimised separately.

For the kernel $k_\Theta$, we also select the hyperparameters by maximising the log-marginal likelihood, 
\begin{align*}
   & L(l_\Theta, A_\Theta) =  -\frac{1}{2} \log | k_{\Theta}(\theta_{1:T},\theta_{1:T}; l_\Theta, A_\Theta)| - \frac{T}{2} \log(2 \pi)
     \\
& \hspace{3cm}-\frac{1}{2} (I_\BQ (\theta_{1:T})- m_{\Theta}(\theta_{1:T}))^\top \\
&\hspace{4cm}\times\left(k_{\Theta}(\theta_{1:T}, \theta_{1:T};l_\Theta, A_\Theta) + \left( \lambda_\Theta + \sigma_{\BQ}^2(\theta_{1:T}) \right) \Id_T \right)^{-1} \\
&\hspace{4cm}\times(I_\BQ (\theta_{1:T})- m_{\Theta}(\theta_{1:T})).
\end{align*}
Similar to above, we also do a grid search over $\left[1.0, 10.0, 100.0, 1000.0 \right]$ for amplitude $A_\Theta$, a grid search over $\left[0.1, 0.3, 1.0, 3.0, 10.0 \right]$ for lengthscale $l_\Theta$ and a grid search over $\left[0.01, 0.1, 1.0 \right]$ for $\lambda_\Theta$, so we select the value that gives the largest log-marginal likelihood. 

\paragraph{Least-squares Monte Carlo.} LSMC implements Monte Carlo in the first stage and polynomial regression in the second stage. In the second stage, the hyperparameters include the regularisation coefficient $\lambda_\Theta$ and the order of the polynomial $p \in \{1,2,3,4\}$.
These hyperparameters are also selected with grid search to give the lowest RMSE on a separate held-out validation set.

\paragraph{Kernel least-squares Monte Carlo.}
KLSMC implements Monte Carlo in the first stage and kernel ridge regression in the second stage. In the second stage, the hyperparameters are analogous to the hyperparameters in the second stage of CBQ, namely $A_\Theta, l_\Theta, \lambda_\Theta$.
These hyperparameters are selected with grid search to give the lowest RMSE on a separate held-out validation set.

\paragraph{Importance sampling.} For IS, there are no hyperparameters to select.

\section{Experimental details}
\label{appendix:experiments_cbq}
We now provide a detailed description of all experiments in the main text.

\subsection{Synthetic Experiment: Bayesian Sensitivity Analysis for Linear Models}
\label{appendix:bayes_sensitivity}

In this synthetic experiment, we do sensitivity analysis on the hyperparameters in Bayesian linear regression. 
The observational data for the linear regression are $Y \in \bR^{m \times d}, Z \in \bR^{m}$ with $m$ being the number of observations and $d$ being the dimension.
We use $x$ to denote the regression weight; this is unusual but is done so as to keep the notation consistent with the main text.
By placing a $\calN(x ; 0, \theta \Id_d)$ prior 
on the regression weights $x \in \bR^{d}$ with $\theta \in \left( 1, 3\right)^d$, and assuming independent $\calN(0, \eta)$ observation noise for some known $\eta>0$, we can obtain (via conjugacy) a multivariate Gaussian posterior $\bP_\theta$ whose mean and variance have a closed form expression~\citep{bishop:2006:PRML}.
\begin{align*}
    \bP_\theta = \calN(\tilde{m}, \tilde{\Sigma}), \quad \tilde{\Sigma}^{-1} = {\frac{1}\theta \Id_d} + \eta Y^\top Y, \quad \tilde{m} = \eta \tilde{\Sigma} Y^\top Z.
\end{align*}
We can then analyse sensitivity by computing the conditional expectation $I(\theta)=\int_\calX f(x)\bP_\theta(\d x)$ of some quantity of interest $f$.
For example, if  $f(x)=x^\top x$, then $I(\theta)$ is the second moment of the posterior and the results are already reported in the main text.
If $f(x) = x^\top y^\ast$ for some new observation $y^\ast$, then $I(\theta)$ is the predictive mean. 
In these simple settings, $I(\theta)$ can be
computed analytically, making this a good synthetic example for benchmarking.
We sample parameter values $\theta_{1:T}$ from a uniform distribution $ \bQ = \operatorname{Unif}(\Theta)$ where $\Theta = (1, 3)^d$, and for each such parameter $\theta_t$, we obtain $N$ observations $x_{1:N}^t$ from $\bP_{\theta_t}$.  
In total, we have $N \times T$ samples.

For conditional Bayesian quadrature (CBQ), we need to carefully choose two kernels $k_\Theta$ and $k_\calX$. Firstly, we choose the kernel $k_\calX$ to be an isotropic Gaussian kernel: $k(x, x') = A_\calX \exp(-\frac{1}{2 l_\calX^2} (x - x')^\top(x - x'))$ for the purpose that the Gaussian kernel mean embedding has a closed form under the Gaussian posterior $\bP_\theta$,
\begin{align}\label{appeq:E14}
    \mu_\theta(x) = A_\calX {\left| \Id_d + l_\calX^{-2} \tilde{\Sigma} \right|}^{-1/2} \exp \left(-\frac{1}{2} (x - \tilde{m})^\top (\tilde{\Sigma} + l_\calX^2 \Id_d)^{-1} (x - \tilde{m})\right)
\end{align}
In addition, the integral of the kernel mean embedding $\mu_\theta$ (known as the initial error) also has a closed form
$\int_{\calX} \mu_\theta(x) \bP_\theta(\d x) = A_\calX l_\calX / \sqrt{| l_\calX^2 \Id_d + 2 \tilde{\Sigma}|}$.

This leaves us with a choice for $k_\Theta$. 
In this synthetic setting, we know that $I(\theta)$ is infinitely times differentiable, but we opt for Mat\'ern-3/2 kernel $k_\Theta(\theta, \theta') = A_\Theta (1+\sqrt{3} |\theta - \theta'|/l_\Theta) \exp (-\sqrt{3} |\theta - \theta'|/l_\Theta)$ to encode a more conservative prior information on the smoothness of $I(\theta)$.

\paragraph{Verifying assumptions.} 
We would like to check whether the assumptions made in \Cref{thm:convergence} hold in this experiment.
\begin{itemize}
    \item A1: 
    Although $\calX=\bR$ is not a compact domain, $\bP_\theta$ is a Gaussian distribution, so the probability mass outside a large compact subset of $\calX$ decays exponentially. $\Theta = \left( 1, 3 \right)^d$ is a compact domain. A1 is therefore approximately satisfied.
    \item A2: A2 is satisfied due to the sampling mechanism of $\theta_{1:T}$ and $\{x_{1:N}^t\}_{t=1}^T$.
    \item A3: $\bQ$ is a uniform distribution so its density $q$ is constant and hence upper bounded and strictly positive. $\bP_\theta$ is a Gaussian distribution, so its density $p_\theta$ is strictly positive on a compact and large domain with finite second moment. A3 is approximately satisfied.
    \item A4: Both $f(x)$ and $I(\theta)$ are infinitely times differentiable, so $s_I=s_f = \infty$. 
    Although $k_\calX$ is Gaussian kernel which does not satisfy the assumption of \Cref{thm:convergence}, we observed similar performance when $k_\calX$ is Mat\'ern-3/2 kernel so $s_\calX = \frac{3}{2} + \frac{d}{2}$, and $k_\Theta$ is Mat\'ern-3/2 kernel so $s_\Theta = \frac{3}{2} + \frac{d}{2}$, where $d$ is the dimension. A4 is satisfied.
    \item A5: $\lambda_\calX$ is picked to be $0$ and $\lambda_\Theta$ is found via grid search among $\{0.01, 0.1, 1.0\}$. A5 is satisfied.
\end{itemize}

\subsection{Bayesian Sensitivity Analysis for Susceptible-Infectious-Recovered (SIR) Model}
\label{appendix:sir}
The SIR model is commonly used to simulate the dynamics of infectious diseases through a population~\citep{kermack1927sir}. 
It divides the population into three sections.
Susceptible (S) represents people who are not infected but can be infected after getting in contact with an infectious individual.
Infectious (I) represents people who are currently infected and can infect susceptible individuals.
Recovered (R) represents individuals who have been infected and then removed from the disease, either by recovering or dying. The dynamics are governed by a system of ordinary differential equations (ODE) as below.
\begin{align*}
    \begin{aligned}
\frac{\mathrm{d} S}{\mathrm{~d} r} &= -x S I, \quad
\frac{\mathrm{d} I}{\mathrm{~d} r} &= x S I-\gamma I, \quad
\frac{\mathrm{d} R}{\mathrm{~d} r} &= \gamma I
\end{aligned}
\end{align*}
with $x$ being the infection rate, $\gamma$ being the recovery rate and $r$ is the time. The solution to the SIR model would be a vector of $\left(N_I^r, N_S^r, N_R^r \right)$ representing the number of infectious, susceptible and recovered at day $r$.

In this experiment, we assume that the recovery rate $\gamma$ is fixed and we place a Gamma prior distribution on $x$; i.e., $\bP_\theta = \operatorname{Gamma}(\theta, \xi)$ where $\theta$ represents the initial belief of the infection rate deduced from the study of the virus in the laboratory at the beginning of the outbreak, and $\xi$ represents the amount of uncertainty on the initial belief. 
We fix the parameter $\xi=10$, the total population is set to be $10^6$ and the recovery rate $\gamma = 0.05$. 
The target of interest is the expected peak number of infected individuals under the prior distribution on $x$: 
\begin{align*}
    I(\theta) = \bE_{x}\left[\max_r N_I^r(x) \mid \theta \right] = \int_{\calX} \max_r N^r_I(x) \bP_\theta(\d x)
\end{align*}
with the integrand $f(x) = \max_r N_I^r(x)$. We are interested in the sensitivity analysis of the shape parameter $\theta$ to the final estimate of the expected peak number of infected individuals.
The initial belief of the infection rate $\theta_{1:T}$ are sampled from the uniform distribution $\bQ = \operatorname{Unif}\left(2,9\right)$ and then $N$ number of $x^t_{1:N}$ are sampled from $\bP_{\theta_t} = \operatorname{Gamma}(\theta_t, \xi)$. 
In this setting, sampling $x$ is very expensive as it necessarily involves solving the system of SIR ODEs, which can be very slow as the discretisation step gets finer.
In the middle panel of \Cref{fig:finance_sir}, we have shown that obtaining one sample from SIR ODEs under discretisation time step $\tau = 0.1$ takes around $3.0$s, whereas running the whole CBQ algorithm takes $1.0$s, not to mention that sampling from SIR ODEs needs to be repeated $N \times T$ times. Therefore, using CBQ is ultimately more efficient overall within the same period of time.

For CBQ, we need to carefully choose two kernels $k_\Theta$ and $k_\calX$.
First we choose $k_\calX$, we use Mat\'ern-3/2 as the base kernel and then apply a Langevin Stein operator to both arguments of the base kernel to obtain $k_\calX$. 
The reason we use a Langevin Stein kernel is that Stein kernel gives an RKHS which is a subset on the Sobolev space with one order less smoothness than the base kernel, and since the smoothness of the integrand $f(x) = \max_r N_I^r(x)$ is unknown, using a Stein kernel enforces weaker prior information than Mat\'ern-3/2.
Furthermore, the kernel mean embedding of a Stein kernel $\mu(x)$ is a constant $c$ by construction. 
The initial error is also a constant $c$ by construction.
Then we choose $k_\Theta$. Since $I(\theta)$ represents the peak number of infections so $I(\theta)$ is expected to be smooth and continuous, and hence we choose $k_\Theta$ as Mat\'ern-3/2 kernel. 
All hyperparameters in $k_\calX$ and $k_\Theta$ are selected according to \Cref{appendix:hyperparameter_selection}.
We use a MC estimator with $5000$ samples as the pseudo ground truth and evaluate the RMSE across all methods.

\paragraph{Verifying assumptions.} 
We would like to check whether the assumptions made in \Cref{thm:convergence} hold in this experiment.
\begin{itemize}
    \item A1: Although $\calX=\bR^+$ is not a compact domain, $\bP_\theta$ is a Gamma distribution so the probability mass outside a large compact subset of $\calX$ around the origin decays exponentially. $\Theta = \left(2, 9 \right)^d$ is a compact domain. A1 is approximately satisfied.
    \item A2: A2 is satisfied due to the sampling mechanism of $\theta_{1:T}$ and $\{x_{1:N}^t\}_{t=1}^T$.
    \item A3: $\bQ$ is a uniform distribution so its density $q$ is constant and hence upper bounded and strictly positive. $\bP_\theta$ is a Gamma distribution so its density $p_\theta$ is strictly positive within a large compact subset of $\calX$ and has finite second moment. A3 is approximately satisfied.
    \item A4: $f(x) = \max_r N_I^r(x)$ is the maximum number of infections so $f(x)$ is not necessarily smooth. $I(\theta)$ represents the peak number of infections with varying initial estimate of the infection rate, so $I(\theta)$ is smooth and continuous with $s_I \leq 1$. 
    $k_\calX$ is Stein kernel with Mat\'ern-3/2 kernel as the base, so the corresponding RKHS will have functions which are rough (i.e., of smoothness $1/2$) but is only a subset of a Sobolev space. In addition, $k_\Theta$ is Mat\'ern-3/2 kernel so $s_\Theta = \frac{3}{2} + \frac{1}{2} = 2$. It is therefore unclear if A4 is satisfied.
    \item A5: $\lambda_\calX$ is picked to be $0$ and $\lambda_\Theta$ is found via grid search among $\{0.01, 0.1, 1.0\}$. A5 is satisfied.
\end{itemize}


\subsection{Option Pricing in Mathematical Finance}
\label{appendix:black_scholes}
In this experiment, we consider specifically an asset whose price $S({\tau})$ at time $\tau$ follows the Black-Scholes formula $S(\tau) = S_0 \exp \left(\sigma W(\tau) - \sigma^2 \tau/2 \right)$ for $\tau \geq 0$, where $\sigma$ is the underlying volatility, $S_0$ is the initial price and $W$ is the standard Brownian motion.
The financial derivative we are interested in is a butterfly call option whose payoff at time $\tau$ can be expressed as $\psi(S({\tau}))=\max (S(\tau)-K_1, 0) + \max (S(\tau)-K_2, 0) - 2\max (S(\tau) - (K_1+K_2)/2, 0)$.

In addition to the expected payoff, insurance companies are interested in computing the expected loss of their portfolios if a shock would occur in the economy.
We follow the setting in \citet{alfonsi2021multilevel, alfonsi2022many} assuming that a shock occur at time $\eta$, at which time the option price is $S(\eta)=\theta$, and this shock multiplies the option price by $1 + s$. The option price at maturity time $\zeta$ is denoted as $S(\zeta) = x$. The expected loss caused by the shock can be expressed as 
\begin{align*}
    L = \bE [\max (I(\theta), 0 )], \text{   } I(\theta) = \int_0^\infty \psi(x)-\psi \left((1 + s) x \right) \bP_\theta(\d x)
\end{align*}
So the integrand is $f(x) = \psi(x)-\psi((1+s)x)$.

Following the setting in \citet{alfonsi2021multilevel, alfonsi2022many}, we consider the initial price $S_0 = 100$, the volatility $\sigma = 0.3$, the strikes $K_1 = 50, K_2 = 150$, the option maturity $\zeta=2$ and the shock happens at $\eta=1$ with strength $s = 0.2$. 
The option price at which the shock occurs are $\theta_{1:T}$ sampled from the log normal distribution deduced from the Black-Scholes formula $\theta_1, \dots, \theta_T \sim \bQ = \operatorname{Lognormal}( \log S_0 - \frac{\sigma^2}{2} \eta, \sigma^2 \eta)$. 
Then $x^t_{1:N}$ are sampled from another log normal distribution also deduced from the Black-Scholes formula $x^t_1, \dots, x^t_N \sim \bP_{\theta_t} = \operatorname{Lognormal}( \log \theta_t - \frac{\sigma^2}{2} (\zeta - \eta), \sigma^2 (\zeta - \eta))$. 

For CBQ, we need to carefully choose two kernels $k_\calX$ and $k_\Theta$. First we choose the kernel $k_\calX$ to be a log-Gaussian kernel for the purpose that the log-Gaussian kernel mean embedding has a closed form under log-normal distribution $\bP_\theta = \operatorname{Lognormal}(\bar{m}, \bar{\sigma}^2)$ with $\bar{m} = \log \theta - \frac{\sigma^2}{2}(\zeta - \eta)$ and  $\bar{\sigma}^2 = \sigma^2 (\zeta - \eta)$. 
The log Gaussian kernel is defined as $k_\calX(x, x') = A_\calX \exp(-\frac{1}{2 l_\calX^2} (\log x - \log x')^2)$
and the kernel mean embedding has the form
\begin{align*}
    \mu_\theta(x) = \frac{A_\calX}{\sqrt{1 + \frac{\bar{\sigma}^2}{l_\calX^2}}} \left. \exp \left(-\frac{\bar{m}^2 + (\log x)^2 }{2(\bar{\sigma}^2 + l_\calX^2)}\right) x^{\frac{\bar{m}}{\bar{\sigma}^2 + l_\calX^2}}  \right.
\end{align*}
The initial error, which is the integral of kernel mean $\mu_\theta(x)$ does not have a closed form expression, so we use the empirical average as an approximation. Then, we choose the kernel $k_\Theta$ to be a Mat\'ern-3/2 kernel.

For this experiment, we also implement CBQ with Langevin Stein reproducing kernel. We use Mat\'ern-3/2 as the base kernel and then apply the Langevin Stein operator to both arguments of the base kernel to obtain $k_\calX$. 
The reason we use a Stein kernel is that Stein kernels have an RKHS whose functions have one order less smoothness than the base kernel, and since the integrand has very low smoothness (due to the maximum function), we do not want to use an overly smooth kernel. 
The kernel mean embedding of a Stein kernel is a
constant $c$ by construction.
The kernel $k_\Theta$ is selected as Mat\'ern-3/2 kernel.
All hyperparameters in $k_\calX$ and $k_\Theta$ for CBQ and hyperparameters for baseline methods are selected according to \Cref{appendix:hyperparameter_selection}.

\paragraph{Verifying assumptions.} 
We would like to check whether the assumptions made in \Cref{thm:convergence} hold in this experiment.
\begin{itemize}
    \item A1: Although $\calX=\bR^+$ is not a compact domain, $\bP_\theta$ is a lognormal distribution so the probability mass outside a large compact subset of $\calX$ decays super exponentially. A similar argument can be made for $\Theta$ as well. A1 is therefore approximately satisfied.
    \item A2: A2 is satisfied due to the sampling mechanism of $\theta_{1:T}$ and $\{x_{1:N}^t\}_{t=1}^T$.
    \item A3: $\bQ$ is a lognormal distribution, so its density $q$ is upper bounded and strictly positive within a large compact subset of $\Theta$. $\bP_\theta$ is also a lognormal distribution so its density $p_\theta$ is strictly positive within a large compact subset of $\calX$ and has finite second moment. A3 is approximately satisfied.
    \item A4: $f(x)$ is a combination of piecewise linear functions so $s_f = 1$ and $I(\theta)$ is infinitely times differentiable so $s_f = \infty$. 
    When $k_\calX$ is a Stein kernel with Mat\'ern-3/2 kernel as the base, the functions in the corresponding RKHS have smoothness $1/2$, whereas when $k_\calX$ is the log Gaussian kernel, the functions are infinitely differentiable. Neither of these choices satisfies the assumption, although Stein kernel contains many (but not necessarily all) functions of smoothness $1/2$. $k_\Theta$ is Mat\'ern-3/2 kernel so $s_\Theta = \frac{3}{2} + \frac{1}{2} = 2$. It is therefore unclear if A4 is satisfied.
    \item A5: $\lambda_\calX$ is picked to be $0$ and $\lambda_\Theta$ is found via grid search among $\{0.01, 0.1, 1.0\}$. A5 is satisfied.
\end{itemize}

\subsection{Uncertainty Decision Making in Health Economics}
\label{appendix:decision}
In the medical world, it is important to compare the cost and the relative advantages of conducting extra medical experiments. 
The expected value of partial perfect information (EVPPI) quantifies the expected gain from conducting extra experiments to obtain precise knowledge of some unknown variables \citep{brennan2007calculating},
\begin{align*}
    \text{EVPPI} = \bE \Bigl[\max_c I_c(\theta) \Bigr] - \max_c \bE \Bigl[I_c(\theta) \Bigr], \text{   } I_c(\theta) = \int_{\calX} f_c(x, \theta) \bP_\theta(\d x)
\end{align*}
where $c \in \calC$ is a set of potential treatments and $f_c$ measures the potential outcome of treatment $c$. Our method is applicable for estimating the conditional expectation $I_c(\theta)$ of the first term. 

We adopt the same experimental setup as delineated in \citet{Giles2019}, wherein $x$ and $\theta$ have a joint 19-dimensional Gaussian distribution, meaning that $\bP_\theta$ is a Gaussian distribution. 
The specific meanings of all $x$ and $\theta$ are outlined in \Cref{tab:mytable}.
All these variables are independent except that $\theta_1, \theta_2, x_6, x_{14}$ are pairwise correlated with a correlation coefficient $0.6$.
The observations $\theta_{1:T}$ are sampled from the marginal Gaussian distribution $\bQ$ and then $N$ observations of $x^t_{1:N}$ are sampled from $\bP_{\theta_t}$.

We are interested in a binary decision-making problem ($\calC = \{1, 2\}$) with $f_1(x, \theta)=10^4 (\theta_1 x_5 x_6 + x_7 x_8 x_{9})-(x_1 + x_2 x_3 x_4)$ and $f_2(x, \theta) = 10^4 (\theta_2 x_{13} x_{14} + x_{15} x_{16} x_{17})-(x_{10} + x_{11} x_{12} x_4)$. 
In computing EVPPI, we estimate $I_c(\theta)$ with CBQ and baselines, and then use standard MC for the rest of the expectations.
We draw $10^6$ samples from the joint distribution to generate a pseudo ground truth, and evaluate the RMSE across different methods. 
Note that IS is no longer applicable here because $f_c$ now depends on both $x$ and $\theta$, so we only comparing CBQ against KLSMC and LSMC.

For CBQ, we need to carefully choose two kernels. First, we take $k_\calX$ to be a Mat\'ern-3/2 to ensure that the kernel mean embedding under a Gaussian distribution $\bP_\theta = \calN(\tilde{\mu}, \tilde{\Sigma})$ has a closed form. 
Specifically, we initially sample $u$ from $\calN(0, \Id_d)$, then calculate $x = \tilde{m} + L^\top u$ where $L$ is the lower triangular matrix derived from the Cholesky decomposition of the covariance matrix $\tilde{\Sigma}$. 
The integral now becomes
\begin{align}\label{appeq:transform}
    I_c(\theta) = \int_{\bR^d} f(x)\calN(x; \tilde{m},\tilde{\Sigma}) \d x = \int_{\bR^d} f(\tilde{m} + L^\top u) \calN(u; 0, \Id_d) du
\end{align}
The closed-form expression of kernel mean embedding for a Mat\'ern-3/2 kernel and an isotropic Gaussian measure can be found in~\citet[Appendix S.3]{ming2021linked}.
Then we pick $k_\Theta$. 
We know there is a high chance that $I_c(\theta)$ is infinitely times differentiable, but we opt for Mat\'ern-3/2 kernel to encode a more conservative prior information on the smoothness of $I_c(\theta)$ because we do not have a closed form of it.
All hyperparameters in $k_\calX$ and $k_\Theta$ are selected according to \Cref{appendix:hyperparameter_selection}.

\begin{table}[t]
\centering
\begin{tabular}{
>{\centering\arraybackslash}p{1.5cm}
>{\centering\arraybackslash}p{1cm}
>{\centering\arraybackslash}p{1cm}
>{\centering\arraybackslash}p{6cm}}
\toprule
Variables & Mean & Std & Meaning \\
\midrule
$X_1$ & 1000 & 1.0 & Cost of treatment \\
$X_2$ & 0.1 & 0.02 & Probability of admissions \\
$X_3$ & 5.2 & 1.0 & Days of hospital \\
$X_4$ & 400 & 200 & Cost per day \\
$X_5$ & 0.3 & 0.1 & Utility change if response \\
$X_6$ & 3.0 & 0.5 & Duration of response \\
$X_7$ & 0.25 & 0.1 & Probability of side effects \\
$X_8$ & -0.1 & 0.02 & Change in utility if side effect \\
$X_{9}$ & 0.5 & 0.2 & Duration of side effects \\
$X_{10}$ & 1500 & 1.0 & Cost of treatment \\
$X_{11}$ & 0.08 & 0.02 & Probability of admissions \\
$X_{12}$ & 6.1 & 1.0 & Days of hospital \\
$X_{13}$ & 0.3 & 0.05 & Utility change if response \\
$X_{14}$ & 3.0 & 1.0 & Duration of response \\
$X_{15}$ & 0.2 & 0.05 & Probability of side effects \\
$X_{16}$ & -0.1 & 0.02 & Change in utility if side effect \\
$X_{17}$ & 0.5 & 0.2 & Duration of side effects \\
$\theta_1$ & 0.7 & 0.1 & Probability of responding \\
$\theta_2$ & 0.8 & 0.1 & Probability of responding \\
\bottomrule

\end{tabular}
\caption{Variables in the health economics experiment.}
\label{tab:mytable}
\end{table}

\paragraph{Verifying assumptions.} 
We would like to check whether the assumptions made in \Cref{thm:convergence} hold in this experiment.
\begin{itemize}
    \item A1: 
    Although $\calX=\bR$ is not a compact domain, $\bP_\theta$ is a Gaussian distribution, so the probability mass outside a large compact subset of $\calX$ decays exponentially. Similarly, $\Theta = \bR$ is not a compact domain, but $\bQ$ is a Gaussian distribution so the probability mass outside a large compact subset of $\Theta$ decays exponentially. A1 is approximately satisfied.
    \item A2: A2 is satisfied due to the sampling mechanism of $\theta_{1:T}$ and $\{x_{1:N}^t\}_{t=1}^T$.
    \item A3: $\bQ$ is also a Gaussian distribution so its density $q$ is upper bounded and strictly positive on a compact and large domain. $\bP_\theta$ is a Gaussian distribution so its density $p_\theta$ is strictly positive on a compact and large domain with finite second moment. A3 is approximately satisfied.
    \item A4: Both the integrand $f$ and the conditional expectation $I_c(\theta)$ are infinitely times differentiable, so $s_f = s_I = \infty$. On the other hand, due to the choice of Mat\'ern-3/2 kernels, $s_{\Theta}=3/2+1/2=2$ and $s_{\calX}=3/2+9/2=6$. A4 is therefore satisfied.
    \item A5: $\lambda_\calX$ is picked to be $0$ and $\lambda_\Theta$ is found via grid search among $\{0.01, 0.1, 1.0\}$. A5 is satisfied.
\end{itemize}

\tocless
\section{Comparison of Conditional Bayesian Quadrature and Multi-Output Bayesian Quadrature\label{appendix:cbq_mobq}}

In \Cref{sec:cbq} in the main text, we mentioned a comparison of CBQ and multi-output Bayesian quadrature (MOBQ,~\citet{xi2018bayesian}) in terms of their computational complexity. 
For $T$ parameter values $\theta_1, \cdots, \theta_T$ and $N$ samples from each probability distribution $\bP_{\theta_1}, \ldots, \bP_{\theta_T}$, the computational cost is $\bigo(TN^3 + T^3)$ for CBQ and $\bigo(N^3T^3)$ for MOBQ. 
We now give a more thorough comparison of CBQ and MOBQ in this section. 

When the integrand $f$ only depends on $x$ (Bayesian sensitivity analysis for linear models, option pricing in mathematical finance), MOBQ only requires one kernel $k_\calX$. 
\begin{align*}
    I_{\mathrm{MOBQ}}(\theta^\ast) &= \left(\int_\calX k_\calX(x, x_{1:NT}) \bP_{\theta^\ast}(\d x) \right) \\
    & \hspace{2cm} \times \left(k_\calX(x_{1:NT}, x_{1:NT}) + \lambda_\calX \Id_{NT} \right)^{-1} f(x_{1:NT}),
\end{align*}
where $x_{1:NT} \in \bR^{NT}$ is a concatenation of $x_{1:N}^1, \cdots, x_{1:N}^T$.
When the integrand $f$ depends on both $x$ and $\theta$ (uncertainty decision making in health economics), MOBQ requires two kernels $k_\calX$ and $k_\Theta$.
\begin{align*}
\begin{aligned}
    I_{\mathrm{MOBQ}}(\theta^\ast) &= \left( \int_\calX k_\calX(x, x_{1:NT})    \odot k_\Theta(\theta^\ast, \theta_{1:NT}) \bP_{\theta^\ast}(\d x)  \right) \\ &
    \Big(k_\calX(x_{1:NT}, x_{1:NT}) \odot k_\Theta(\theta_{1:NT}, \theta_{1:NT})  + \lambda_\calX \Id_{NT} \Big)^{-1} f(x_{1:NT})
\end{aligned}
\end{align*}
where $\odot$ is the element-wise product, and $\theta_{1:NT} = \left[\theta_1, \cdots, \theta_1, \cdots, \theta_T, \cdots, \theta_T \right] \in \bR^{NT}$.
From the above two equations, we can see that the computation cost of $\bigo(N^3T^3)$ mainly comes from the inversion of an $NT \times NT$ kernel matrix. 
It is crucial to note that the MOBQ computational cost is significantly higher for the Stein reproducing kernel during hyperparameter selection (an approach analogous to the `vector-valued control variates' of \citet{Sun2021}), as evaluating the log marginal likelihood at every iteration would require the inversion of an $NT \times NT$ matrix.
\section{Reducing the cost of Bayesian Quadrature}
\label{app:reducing_the_cost}
Due to the computational cost of $\bigo(N^3)$ of inverting the Gram matrix, Bayesian Quadrature is best suited when the cost of evaluating the integrand $f$ is the computational bottleneck. For cheaper problems,\citet{Jagadeeswaran2018,Karvonen2017symmetric,Karvonen2019} propose BQ methods where the computational cost is much lower, but these are applicable only with specific point sets $x_{1:N}$ and distributions $\bP$. \citet{Hayakawa2023} also studies Nystr\"om-type of approximations, whilst \citet{Adachi2022} studies parallelisation techniques. Finally, several alternatives with linear cost in $N$ have also been proposed using tree-based \citep{Zhu2020} or neural-network \citep{Ott2023} models, but these tend to require approximate inference methods such as Laplace approximations or Markov chain Monte Carlo.

\chapter{Calibration for MMD-minimising Integration: Supplementary Materials}
\label{sec:app_gpcv}
\section{Proofs of Theoretical Results}
\label{sec:proofs}

This section provides the proofs of the main results and other lengthy computations.
For $x_0 = 0$ and $x_1, \dots, x_N \in [0,T]$, we will use the following notation whenever it can improve the readability or highlight a point:
\begin{align}
   \Delta x_n &\coloneqq x_{n+1} - x_n, \quad n  = 0, 1, \dots, N-1,  \nonumber \\
   f_n &\coloneqq f(x_n), \quad n = 0, 1, \dots, N. \label{eq:notation-x-f}
\end{align}

\subsection{Explicit expressions for the CV and ML estimators} \label{sec:explicit-post-mean-cov}
Let us define $x_0 = 0$ and use the convention $f(x_0) = 0$.
By a direct computation it is straightforward to verify that the inverse of the Gram matrix of the Brownian motion kernel $k(x, x') = \min(x, x')$ over the points $x_1 < x_2 < \dots < x_N$ is the band matrix
\begin{align*}
  k(x_{1:N}, x_{1:N})^{-1} &=
  \begin{bmatrix}
      x_1 & x_1 & x_1 & \dots & x_1 & x_1 \\
      x_1 & x_2 & x_2 & \dots & x_2 & x_2 \\
      x_1 & x_2 & x_3 & \dots & x_3 & x_3 \\
      \vdots & \vdots & \vdots  & \ddots  & \vdots & \vdots \\
      x_1 & x_2 & x_3 & \dots & x_{N-1} & x_{N-1} \\
      x_1 & x_2 & x_3 & \dots & x_{N-1} & x_N
    \end{bmatrix}^{-1} \\
    &= 
    \begin{bmatrix}
        b_1    & c_1    & 0          & \dots  & 0       & 0      \\
        c_1    & b_2    & c_2        & \dots  & 0       & 0      \\
        0      & c_2    & b_3        & \dots  & 0       & 0      \\
        \vdots & \vdots & \vdots & \ddots & \vdots  & \vdots \\
        0      & 0          & 0      & \dots  & b_{N-1} & c_{N-1}\\
        0      & 0          & 0      & \dots  & c_{N-1} & b_N    \\
    \end{bmatrix} ,
\end{align*}
where
\begin{align*}
    b_i &= \frac{x_{i+1} - x_{i-1}}{(x_{i-1} - x_i) (x_i - x_{i+1})} \text{ for } i \in \{1,\dots,N-1\}, \qquad b_N &= - \frac{1}{x_{N-1} - x_{N}}, \\
    c_i &= \frac{1}{(x_i - x_{i+1})}  \text{ for } i \in \{1,\dots,N-1\}. \\
\end{align*}
It follows that the posterior mean and covariance functions in~\eqref{eq:posterior-moments} can be expressed as
\begin{equation} \label{eq:explicit-post-mean}
  m_N(x) =
  \begin{dcases}
    \frac{ (x_n - x) f(x_{n-1}) + (x - x_{n-1}) f(x_{n}) }{x_n - x_{n-1}} &\hspace{-0.2cm}\text{ if } x \in [x_{n-1}, x_n] \\
    &\quad\text{for some } 1 \leq n \leq N, \\
    f(x_N) &\hspace{-0.2cm}\text{ if } x \in [x_N, T]
  \end{dcases}
\end{equation}
and
\begin{equation} \label{eq:explicit-post-cov}
  k_N(x, x') =
  \begin{dcases}
    \frac{ ( x_n - x') ( x - x_{n-1})  }{x_n - x_{n-1}} &\text{ if } x_{n-1} \leq x \leq x' \leq x_n \\
    &\quad\text{for some } 1 \leq n \leq N, \\
    x - x_N &\text{ if } x_N \leq x \leq x' \leq T, \\
    0 &\text{ otherwise}.
  \end{dcases}
\end{equation}
We omit the case $x' \leq x$ for $k_N(x,x')$ as this case is  obtained by the symmetry $k_N(x,x') = k_N(x', x)$.

Using these expressions, we have, for each $1 \leq n < N$:
\begin{equation*}
  m_{\setminus n}(x_n) = \frac{ (x_n - x_{n + 1}) f(x_{n-1}) + (x_{n - 1} - x_{n}) f(x_{n+1}) }{x_{n-1} - x_{n + 1}}
\end{equation*}
and
\begin{equation*}
  k_{\setminus n} (x_n) = k_{\setminus n}(x_n, x_n) = \frac{ ( x_n - x_{n + 1}) ( x_n - x_{n - 1}) }{x_{n-1} - x_{n + 1}}
\end{equation*}
For $n = N$, we have $m_{\setminus N}(x_N) = f(x_{N-1})$ and $k_{\setminus N}(x_N) = x_N - x_{N-1}$.
Inserting these expressions in~\eqref{eq:sigma-cv} and using the notation~\eqref{eq:notation-x-f}, the CV estimator can be written as
\begin{equation}
\label{eq:sigma-cv-w-deltas}
\begin{split}
    \tausqcvest &= \frac{1}{N} \Bigg[ \frac{(x_2 f_1 -  x_1 f_2  )^2}{ x_1 x_2 \Delta x_1 } + \sum_{n=2}^{N-1}  \frac{( \Delta x_{n-1} [f_{n+1} - f_n] - \Delta x_n [f_n - f_{n-1}] )^2}{ (\Delta x_n + \Delta x_{n-1}) \Delta x_n \Delta x_{n-1} } \\
    &\hspace{1.2cm}+ \frac{(f_N - f_{N-1})^2}{ \Delta x_{N-1} } \Bigg].
\end{split}
\end{equation}

For the ML estimator~\eqref{eq:sigma-ml}, we obtain the explicit expression
\begin{equation}
  \label{eq:sigma-ml-w-deltas}
  \tausqmlest = \frac{1}{N} \sum_{n=1}^N \frac{[ f(x_n) - f(x_{n-1}) ]^2}{ \Delta x_{n-1} }
\end{equation}
by observing that $m_{n-1}(x_n) = f(x_n)$ and $k_{n-1}(x_n) = x_n - x_{n-1}$. 

\begin{remark}
    The leave-$p$-out estimator $\hat{\tau}_{\mathrm{CV}(p)}^2$ can be expressed in a form similar (albeit more complicated) to~\eqref{eq:sigma-cv-w-deltas}. We derive this expression in~\Cref{sec:explicit_expression_for_leave_p_out}. This suggests that the analysis in~\Cref{sec:limit-behaviour-for-sigma} could potentially be generalised to apply to the leave-$p$-out estimators, a possibility that we leave open for future research to explore.
\end{remark}

\subsection{Proofs for \Cref{sec:results-deterministic}} \label{sec:proofs-deterministic}

\begin{proof}[Proof of \Cref{res:holder-spaces}]
The estimator $\tausqcvest$ in~\eqref{eq:sigma-cv-w-deltas} may be written as
\begin{equation} \label{eq:sigma-cv-boundary-interior}
  \tausqcvest = B_{1, N} + I_N + B_{2,N}
\end{equation}
in terms of the boundary terms
\begin{equation} \label{eq:boundary-terms}
  B_{1,N} = \frac{1}{N} \cdot \frac{(x_2 f_1 -  x_1 f_2  )^2}{ x_1 x_2 \Delta x_1 } \quad \text{ and } \quad B_{2,N} = \frac{1}{N} \cdot \frac{(f_N - f_{N-1})^2}{ \Delta x_{N-1} }
\end{equation}
and the interior term 
\begin{equation} \label{eq:interior-term}
    I_N = \frac{1}{N} \sum_{n=2}^{N-1}  \frac{( \Delta x_{n-1} [f_{n+1} - f_n] - \Delta x_n [f_n - f_{n-1}] )^2}{ (\Delta x_n + \Delta x_{n-1}) \Delta x_n \Delta x_{n-1} } .
\end{equation}
The claimed rate in~\eqref{eq:main-result} is $\bigo( N^{-2} )$ if $s \geq 2$ or $s = 1$ and $\alpha \geq 1/2$.
By the inclusion properties of H\"older spaces given in \Cref{sec:quantifying_smoothness_of_functions}, it is therefore sufficient to consider the cases (a) $s = 0$ and (b) $s=1$ and $\alpha \in (0, 1/2]$.

Suppose first that $s = 0$.
Let $L$ be a H\"older constant of a function $f \in C^{0, \alpha}([0,T])$.
Using the H\"older condition, the bounding assumption on $\Delta x_n$, and $f_0 = f(0) = 0$, the boundary terms can be bounded as
\begin{align}
  B_{1,N} &= \frac{1}{N} \cdot \frac{ (x_1 (f_1 - f_2) + \Delta x_1 (f_1 - f_0) )^2}{x_1 x_2 \Delta x_1} \\
  &\leq \frac{1}{N} \cdot \frac{2(x_1^2 (f_1 - f_2)^2 + \Delta x_1^2 (f_1 - f_0)^2)}{ x_1 x_2 \Delta x_1 } \nonumber \\
    &\leq \frac{1}{N} \cdot \frac{ 2L^2 (x_1^2 \Delta x_1^{2\alpha} + x_1^{2\alpha} \Delta x_1^2  ) }{x_1 x_2 \Delta x_1 } \nonumber \\
    & = \bigo(N^{-1} \Delta x_1^{2\alpha - 1}) \nonumber \\
    & = \bigo(N^{-2 \alpha}) \label{eq:bound-B1N}
\end{align}
and
\begin{equation}
  B_{2,N} = \frac{1}{N} \cdot \frac{(f_N - f_{N-1})^2}{ \Delta x_{N-1} } \leq \frac{1}{N} L^2 \Delta x_{N-1}^{2\alpha - 1} = \bigo(N^{- 2\alpha}). \label{eq:bound-B2N}
\end{equation}
Similarly, the interior term is bounded as
\begin{align*}
  I_N &\leq
  \frac{2}{N}\sum_{n=2}^{N-1}  \frac{\Delta x_{n-1}^2 (f_{n+1} - f_n)^2 + \Delta x_n^2 ( f_n - f_{n-1})^2}{ (\Delta x_n + \Delta x_{n-1}) \Delta x_n \Delta x_{n-1} } \\
  &\leq \frac{2L^2}{N} \sum_{n=2}^{N-1}  \frac{\Delta x_{n-1}^2 \Delta x_n^{2\alpha} + \Delta x_n^2 \Delta x_{n-1}^{2\alpha} }{ (\Delta x_n + \Delta x_{n-1}) \Delta x_n \Delta x_{n-1} }
  \\
    & = \frac{2L^2}{N} \sum_{n=2}^{N-1}  \frac{ \Delta x_{n-1} \Delta x_n^{2\alpha - 1} + \Delta x_n \Delta x_{n-1}^{2\alpha - 1}  }{ \Delta x_n + \Delta x_{n-1}} \\
    &= \frac{2L^2}{N} \sum_{n=2}^{N-1}  \bigg( \frac{ \Delta x_{n-1}}{ \Delta x_n + \Delta x_{n-1}} \Delta x_n^{2\alpha - 1} + \frac{\Delta x_n }{ \Delta x_n + \Delta x_{n-1}} \Delta x_{n-1}^{2\alpha - 1} \bigg) \\
    & \leq \frac{2L^2}{N} \sum_{n=2}^{N-1} \big( \Delta x_n^{2\alpha - 1} + \Delta x_{n-1}^{2\alpha - 1} \big) \\
    &= \bigo (N^{1 - 2\alpha}).
\end{align*}
Inserting the above bounds in~\eqref{eq:sigma-cv-boundary-interior} yields $\tausqcvest = \bigo(N^{-2\alpha} + N^{1 - 2\alpha}) = \bigo(N^{1-2\alpha})$, which is the claimed rate when $s=0$.

Suppose then that $s = 1$ and $\alpha \in (0, 1/2]$, so that the first derivative $f'$ of $f \in C^{1, \alpha}([0, T])$ is $\alpha$-H\"older and hence continuous.
Because a continuously differentiable function is Lipschitz, we may set $\alpha = 1$ in the estimates~\eqref{eq:bound-B1N} and~\eqref{eq:bound-B2N} for the boundary terms $B_{1,N}$ and $B_{2,N}$ in the preceding case.
This shows these terms are $\bigo(N^{-2})$.
Because $f$ is differentiable, we may use the mean value theorem to write the interior term as
\begin{align*}
    I_N &= \frac{1}{N} \sum_{n=2}^{N-1}  \frac{\Delta x_{n-1} \Delta x_n}{\Delta x_{n-1} + \Delta x_n} \bigg( \frac{f_{n+1} - f_n}{\Delta x_n} - \frac{f_n - f_{n-1}}{\Delta x_{n-1}} \bigg)^2 \\
    &= \frac{1}{N} \sum_{n=2}^{N-1}  \frac{\Delta x_{n-1} \Delta x_n}{\Delta x_{n-1} + \Delta x_n} \big[ f'(\tilde{x}_n) - f'(\tilde{x}_{n-1}) \big]^2,
\end{align*}
where $\tilde{x}_n \in (x_{n}, x_{n+1})$.
Let $L$ be a H\"older constant of $f'$.
Then the H\"older continuity of $f'$ and the assumption that $\|\prt_N \|=\bigo(N^{-1})$ yield
\begin{align*}
    I_N &\leq \frac{L^2}{N} \sum_{n=2}^{N-1}  \frac{\Delta x_{n-1} \Delta x_n}{\Delta x_{n-1} + \Delta x_n} \lvert \tilde{x}_n - \tilde{x}_{n-1} \rvert^{2\alpha} \\
    &\leq \frac{L^2}{N} \sum_{n=2}^{N-1}  \frac{\Delta x_{n-1} \Delta x_n}{\Delta x_{n-1} + \Delta x_n} (\Delta x_{n-1} + \Delta x_n)^{2\alpha} \\
    &\leq \frac{L^2}{N} \sum_{n=2}^{N-1} \Delta x_n (\Delta x_{n-1} + \Delta x_n)^{2\alpha} \\
    &= \bigo(N^{-2\alpha - 1}).
\end{align*}
Using the above bounds in~\eqref{eq:sigma-cv-boundary-interior} yields $\tausqcvest = \bigo(N^{-2} + N^{-2\alpha-1}) = \bigo(N^{-2\alpha-1})$, which is the claimed rate when $s=1$.
\end{proof}

\begin{proof}[Proof of \Cref{res:holder-spaces-ml}]

  From~\eqref{eq:sigma-ml-w-deltas} we have
  \begin{equation*}
    \tausqmlest = \frac{1}{N} \sum_{n=1}^N \frac{(f_n - f_{n-1})^2}{\Delta x_{n-1}}.
  \end{equation*}
  Suppose first that $s = 0$.
  As in the proof of \Cref{res:holder-spaces}, we get
  \begin{equation} \label{eq:holder-spaces-ml-derivation}
    \tausqmlest = \frac{1}{N} \sum_{n=1}^N \frac{(f_n - f_{n-1})^2}{\Delta x_{n-1}} \leq \frac{L^2}{N} \sum_{n=1}^N \Delta x_{n-1}^{2\alpha - 1} = \bigo\big( N^{1-2\alpha} \big)
  \end{equation}
  when $\|\prt_N \|=\bigo(N^{-1})$.
  Suppose then that $s = 1$.
  By the mean value theorem there are $\xi_n \in (x_{n-1}, x_n)$ such that
  \begin{align*}
    \tausqmlest = \frac{1}{N} \sum_{n=1}^N \frac{(f_n - f_{n-1})^2}{\Delta x_{n-1}} &= \frac{1}{N} \sum_{n=1}^N \Delta x_{n-1} \bigg( \frac{f_n - f_{n-1}}{\Delta x_{n-1}} \bigg)^2 \\
    &= \frac{1}{N} \sum_{n=1}^N \Delta x_{n-1} f'(\xi_n)^2.
  \end{align*}
  Since $f'$ is continuous on $[0, T]$ and hence Riemann integrable, we obtain the asymptotic equivalence
  \begin{equation*}
    N \tausqmlest \to \int_0^T f'(x)^2 \, \mathrm{d} x \quad \text{ as } \quad N \to \infty
  \end{equation*}
  when $\|\prt_N \| \to 0$ as $N \to \infty$.
  The integral is positive because $f$ has been assumed non-constant.
\end{proof}

\begin{proof}[Proof of \Cref{res:fqv-estimator}]
For equally-spaced partitions, $\Delta x_n = x_1 = T/N$ for all $n \in \{0, \dots, N-1\}$, the estimator $\tausqcvest$ in~\eqref{eq:sigma-cv-w-deltas} takes the form
\begin{equation*}
    \tausqcvest = \frac{1}{T} \Bigg[ \frac{(x_2 f_1 -  x_1 f_2  )^2}{ x_1 x_2} + \frac{1}{2}\sum_{n=2}^{N-1} (  (f_{n+1} - f_n) - (f_n - f_{n-1}) )^2 + (f_N - f_{N-1})^2\Bigg].
\end{equation*}
Recall from the proof of \Cref{res:holder-spaces} the decomposition
\begin{equation*}
  \tausqcvest = B_{1, N} + I_N + B_{2,N}
\end{equation*}
in terms of the boundary terms $B_{1,N}$ and $B_{2,N}$ in~\eqref{eq:boundary-terms} and the interior term $I_N$ in~\eqref{eq:interior-term}.
Because $f$ is assumed continuous on the boundary and equispaced partitions are quasi-uniform, both $B_{1,N}$ and $B_{2,N}$ tend to zero as $N \to \infty$.
We may therefore focus on the interior term, which decomposes as
\begin{equation*}
\begin{split}
    I_N = {}& \frac{1}{2}\sum_{n=2}^{N-1} \left(  (f_{n+1} - f_n) - (f_n - f_{n-1}) \right)^2\\
    ={}& \sum_{n=2}^{N-1} (f_{n+1} - f_n)^2
    + ( f_n - f_{n-1} )^2 - \frac{1}{2} ( f_{n+1} - f_{n-1} )^2
\end{split}
\end{equation*}
The sums $\sum_{n=2}^{N-1}  (f_{n+1} - f_n)^2$ and $\sum_{n=2}^{N-1}  ( f_n - f_{n-1} )^2$ tend to $V^2(f)$ by definition. To establish the claimed bound we are therefore left to prove that
\begin{equation}
\label{eq:requirement-double-prt-to-v2f}
    \sum_{n=2}^{N-1} (f_{n+1} - f_{n-1} )^2 \to 2V^2(f) \qquad \text{as} \qquad N \to \infty.
\end{equation}
We may write the sum as
\begin{align*}
  \sum_{n=2}^{N-1} (f_{n+1} - f_{n-1} )^2 &= \sum_{n=1}^{\lfloor \frac{N-1}{2} \rfloor} (f_{2n+1} - f_{2n-1} )^2 + \sum_{n=1}^{\lfloor \frac{N-2}{2} \rfloor} (f_{2n+2} - f_{2n} )^2.
\end{align*}
Consider a sub-partition of $\prt_N$ that consists of odd-index points $x_1, x_3, \dots x_{2\lfloor \frac{N-1}{2} \rfloor+1} $ of $\prt_N$. The sequence of these sub-partitions is quasi-uniform with constant $2$. The assumption that the quadratic variation is $V^2(f)$ for all partitions with quasi-uniformity constant $2$ implies that
\begin{align*}
  \lim_{N \to \infty} \sum_{n=1}^{\lfloor \frac{N-1}{2} \rfloor} (f_{2n+1} - f_{2n-1} )^2 = V^2(f).
\end{align*}
The same will hold for sub-partitions formed of even-index points of $\prt_N$, giving
\begin{align*}
  \lim_{N \to \infty} \sum_{n=1}^{\lfloor \frac{N-2}{2} \rfloor} (f_{2n+2} - f_{2n} )^2 = V^2(f).
\end{align*}
Thus,~\eqref{eq:requirement-double-prt-to-v2f} holds. This completes the proof.
\end{proof}

\subsection{Proofs for \Cref{sec:random-setting}} \label{sec:proofs-random}

\begin{proof}[Proof of \Cref{res:holder-spaces-exp}]
    Recall the explicit expression of $\tausqcvest$ in~\eqref{eq:sigma-cv-w-deltas}:
    \begin{equation}
    \begin{split} \label{eq:expnasion-1056}
        \tausqcvest &= \frac{1}{N} \Bigg[ \frac{(x_2 f_1 -  x_1 f_2  )^2}{ x_1 x_2 \Delta x_1 } + \sum_{n=2}^{N-1}  \frac{( \Delta x_{n-1} [f_{n+1} - f_n] - \Delta x_n [f_n - f_{n-1}] )^2}{ (\Delta x_n + \Delta x_{n-1}) \Delta x_n \Delta x_{n-1} } \\
        &\hspace{1.2cm}+ \frac{(f_N - f_{N-1})^2}{ \Delta x_{N-1} } \Bigg].
    \end{split}
    \end{equation}
    We consider the cases $s = 0$ and $s = 1$ separately. Recall that $f \sim \GP(0, k_{s, H})$ implies that $\bE [f(x)f(x')] = k_{s,H}(x, x')$.

    Suppose first that $s = 0$, in which case $f \sim \GP(0, k_{0, H})$ for the fractional Brownian motion kernel $k_{0,H}$ in~\eqref{eq:fbm-def}.
    In this case the expected values of squared terms in the expression for $\tausqcvest$ are $\bE [x_2 f_1 -  x_1 f_2  ]^2 = x_1 x_2 \Delta x_1  (x_1^{2H - 1} + \Delta x_1^{2H-1}- (x_1+\Delta x_1)^{2H - 1} )$,
    \begin{align*}
      \bE \big[ &\Delta x_{n-1} (f_{n+1} - f_n) - \Delta x_n (f_n - f_{n-1}) \big]^2  \\
      &= \big( \Delta x_n^{2H-1} +  \Delta x_{n-1}^{2H-1} - (\Delta x_{n-1} +\Delta x_n)^{2H-1} \big) \Delta x_{n-1}  \Delta x_n (\Delta x_n + \Delta x_{n-1} ),
    \end{align*}
    and $\bE [f_N - f_{N-1}]^2 = \Delta x_{N-1}^{2H}$.
    Substituting these in the expectation of $\tausqcvest$ and using the fact that $\Delta x_n=\Theta(N^{-1})$ for all $n$ by quasi-uniformity we get
    \begin{equation*}
    \begin{split}
        \bE \tausqcvest &= \frac{1}{N} \Bigg[ (x_1^{2H - 1} + \Delta x_1^{2H-1}- (x_1+\Delta x_1)^{2H - 1} ) \\
        &\hspace{1.2cm} + \sum_{n=2}^{N-1}  \left(\Delta x_{n-1}^{2H-1}  + \Delta x_n^{2H-1} - (\Delta x_{n-1} + \Delta x_n)^{2H-1} \right) + \Delta x_{N-1}^{2H - 1}\Bigg] \\
        &= \frac{1}{N} \Bigg[ \Delta x_1^{2H-1}\left(\left(\frac{x_1}{\Delta x_1}\right)^{2H - 1} + 1 - \left(\frac{x_1}{\Delta x_1^{2H-1}} + 1\right)^{2H - 1} \right) \\
        &\hspace{1.2cm} + \Delta x_n^{2H-1} \sum_{n=2}^{N-1}  \left(\left(\frac{\Delta x_{n-1}}{\Delta x_n}\right)^{2H-1}  + 1 - \left(\frac{\Delta x_{n-1}}{\Delta x_n} + 1\right)^{2H-1} \right) \\
        &\hspace{10cm}+ \Delta x_{N-1}^{2H - 1}\Bigg] \\
        &\eqqcolon{}\frac{1}{N} \Bigg[ \Delta x_1^{2H-1} c_1 + \Delta x_n^{2H-1} \sum_{n=2}^{N-1} c_n + \Delta x_{N-1}^{2H - 1}\Bigg] .
    \end{split}
    \end{equation*}
    Notice that the function $x \mapsto x^{2H-1} + 1 - (x+1)^{2H-1}$ is positive for $x>0$ and $H \in (0, 1)$, and increasing for $H \in (1/2, 1)$ and non-increasing for $H \in (0, 1/2]$. By quasi-uniformity we have $C_\textup{qu}^{-1} \leq \Delta x_{n-1}/\Delta x_n \leq C_\textup{qu}$, and can bound $c_n$ for any $n$ and $N$ as
    \begin{equation*}
    \begin{split}
        0< C_\textup{qu}^{2H-1} + 1 - (C_\textup{qu}+1)^{2H-1} \leq c_n \leq C_\textup{qu}^{1-2H} + 1 - (C_\textup{qu}^{-1}+1)^{2H-1}
    \end{split}
    \end{equation*}
    if $H \in (0, 1/2]$, and
    \begin{equation*}
    \begin{split}
        0< C_\textup{qu}^{1-2H} + 1 - (C_\textup{qu}^{-1}+1)^{2H-1} \leq c_n \leq C_\textup{qu}^{2H-1} + 1 - (C_\textup{qu}+1)^{2H-1}
    \end{split}
    \end{equation*}
    if $H \in (1/2, 1)$. Finally, by quasi-uniformity $\Delta x_n=\Theta(N^{-1})$, and $\bE \tausqcvest=\Theta(N^{-2H}) + \Theta(N^{1-2H})+\Theta(N^{-2H})=\Theta(N^{1-2H})$.

    Suppose then that $s = 1$, in which case $f \sim \GP(0, k_{1, H})$ for the integrated fractional Brownian motion kernel $k_{1,H}$ in~\eqref{eq:iFBM-kernel-explicit}.
    It is straightforward (though, in the case of the second expectation, somewhat tedious) to compute that the expected values of squared terms in the expression~\eqref{eq:expnasion-1056} for $\tausqcvest$ are
    \begin{equation*}
        \bE [x_2 f_1 -  x_1 f_2  ]^2 = \frac{x_1 x_2 \Delta x_1 }{2(H+1)(2H+1)} \big( x_2^{2H+1} - x_1^{2H+1} - \Delta x_1^{2H+1} \big)
    \end{equation*}
    and
    \begin{equation} \label{eq:s=1-exp-2}
      \begin{split}
        \bE \big[ \Delta &x_{n-1} (f_{n+1} - f_n) - \Delta x_n (f_n - f_{n-1}) \big]^2  \\
        ={}& \frac{\Delta x_n \Delta x_{n-1} (\Delta x_{n} + \Delta x_{n-1}) }{2(H+1)(2H+1)} \big[ (\Delta x_{n} + \Delta x_{n-1})^{2H+1} - \Delta x_n^{2H+1} - \Delta x_{n-1}^{2H+1} \big]
      \end{split}
    \end{equation}
    and
    \begin{equation*}
      \begin{split}
        \bE [f_N - f_{N-1}]^2 = \frac{\Delta x_{N-1}}{2H+1} \bigg( x_N^{2H+1} - x_{N-1}^{2H+1} - \frac{1}{2(H+1)} \Delta x_{N-1}^{2H+1} \bigg).
      \end{split}
    \end{equation*}
    Therefore, by~\eqref{eq:expnasion-1056}, $\bE \tausqcvest$ is equal to
    \begin{equation*}
      \begin{split}
        & \frac{\big( x_2^{2H+1} - x_1^{2H+1} - \Delta x_1^{2H+1} \big)}{2(H+1)(2H+1)N} \\
        &+ \frac{1}{2(H+1)(2H+1)N}\sum_{n=2}^{N-1} \big[ (\Delta x_{n} + \Delta x_{n-1})^{2H+1} - \Delta x_n^{2H+1} - \Delta x_{n-1}^{2H+1} \big] \\
        &+ \frac{1}{(2H+1)N} \bigg( x_N^{2H+1} - x_{N-1}^{2H+1} - \frac{1}{2(H+1)} \Delta x_{N-1}^{2H+1} \bigg) \\
        \eqqcolon{}& \frac{1}{2(H+1)(2H+1)} B_{1,N} + \frac{1}{2(H+1)(2H+1)} I_N + \frac{1}{(2H+1)} B_{2,N}.
      \end{split}
    \end{equation*}
    By quasi-uniformity, $B_{1,N} \leq N^{-1} x_2^{2H+1} = \bigo(N^{-2-2H})$.
    Consider then the interior term
    \begin{equation} \label{eq:interior-term-exp-cv-proof}
      \begin{split}
      I_N &= \frac{1}{N} \sum_{n=2}^{N-1} \Delta x_n^{2H+1} \Bigg[ \bigg(1 + \frac{\Delta x_{n-1}}{\Delta x_n} \bigg)^{2H+1} - \bigg(1 + \bigg(\frac{\Delta x_{n-1}}{\Delta x_n} \bigg)^{2H+1} \bigg) \Bigg] \\
      &\eqqcolon \frac{1}{N} \sum_{n=2}^{N-1} \Delta x_n^{2H+1} c'_n.
      \end{split}
    \end{equation}
    Because the function $x \mapsto (1 + x)^{2H+1} - (1 + x^{2H+1})$ is positive and increasing for $x > 0$ if $H \in (0, 1)$ and $C_\textup{qu}^{-1} \leq \Delta x_{n-1} / \Delta x_n \leq  C_\textup{qu}$ by quasi-uniformity, we have
    \begin{equation*}
      0 < (1 + C_\textup{qu}^{-1})^{2H+1} - (1 + C_\textup{qu}^{-(2H+1)}) \leq c'_n \leq \bigg(1 + \frac{\Delta x_{n-1}}{\Delta x_n} \bigg)^{2H+1} \leq (1 + C_\textup{qu})^{2H+1}
    \end{equation*}
    for every $n$.
    Because $N^{-1} \sum_{n=2}^{N-1} \Delta x_n^{2H+1} = \Theta(N^{-1-2H})$ by quasi-uniformity, we conclude from~\eqref{eq:interior-term-exp-cv-proof} that $I_N = \Theta(N^{-1-2H})$.
    For the last term $B_{2, N}$, recall that we have set $x_N = T$.
    Thus
    \begin{equation*}
      \begin{split}
        B_{2,N} &= \frac{1}{N} \bigg( T^{2H+1} - (T - \Delta x_{N-1})^{2H+1} - \frac{1}{2(H+1)} \Delta x_{N-1}^{2H+1} \bigg).
      \end{split}
    \end{equation*}
    By the generalised binomial theorem,
    \begin{equation*}
      T^{2H+1} - (T - \Delta x_{N-1})^{2H+1} = (2H+1) T^{2H} \Delta x_{N-1} + \bigo( \Delta x_{N-1}^2 )
    \end{equation*}
    as $\Delta x_{N-1} \to 0$.
    It follows that under quasi-uniformity we have $B_{2,N} = \Theta(N^{-2})$ for every $H \in (0, 1)$.
    Putting these bounds for $B_{1,N}$, $I_N$ and $B_{2,N}$ together we conclude that
    \begin{equation*}
        \begin{split}
        \bE \tausqcvest &= \frac{1}{2(H+1)(2H+1)} B_{1,N} + \frac{1}{2(H+1)(2H+1)} I_N + \frac{1}{(2H+1)} B_{2,N} \\
        &= \bigo(N^{-2-2H}) + \Theta(N^{-1-2H}) + \Theta(N^{-2}),
        \end{split}
    \end{equation*}
    which gives $\bE \tausqcvest = \Theta(N^{-1-2H})$ if $H \in (0, 1/2]$ and $\bE \tausqcvest = \Theta(N^{-2})$ if $H \in [1/2, 1)$.
\end{proof}

Observe that in the proof of \Cref{res:holder-spaces-exp} it is the boundary term $B_{2,N}$ that determines the rate when there is sufficient smoothness, in that $s = 1$ and $H \in [1/2, 1)$.
Similar phenomenon occurs in the proof of \Cref{res:holder-spaces}.
The smoother a process is, the more correlation there is between its values at far-away points.
Because the Brownian motion (as well as fractional and integrated Brownian motions) has a zero boundary condition at $x = 0$ but no boundary condition at $x = T$ and no information is available at points beyond $T$, the importance of $B_{2,N}$ is caused by the fact that around $T$, the least information about the process is available.

\begin{proof}[Proof of \Cref{res:holder-spaces-exp-ml}]
  From~\eqref{eq:sigma-ml-w-deltas} we get
  \begin{equation*}
    \bE \tausqmlest = \frac{1}{N} \sum_{n=1}^N \frac{\bE [f_n - f_{n-1}]^2}{\Delta x_{n-1}}.
  \end{equation*}
  We may then proceed as in the proof of \Cref{res:holder-spaces-exp} and use quasi-uniformity to show that
  \begin{equation*}
    \bE \tausqmlest = \frac{1}{N} \sum_{n=1}^N \frac{\bE [f_n - f_{n-1}]^2}{\Delta x_{n-1}} = \frac{1}{N} \sum_{n=1}^N \frac{ \Delta x_{n-1}^{2H}}{\Delta x_{n-1}} = \frac{1}{N} \sum_{n=1}^N \Delta x_{n-1}^{2H-1} = \Theta(N^{1-2H})
  \end{equation*}
  when $s = 0$ and
  \begin{equation*}
    \begin{split} 
      \bE \tausqmlest &= \frac{1}{N} \sum_{n=1}^N \frac{\bE [f_n - f_{n-1}]^2}{\Delta x_{n-1}} \\
      &= \frac{1}{(2H+1)N} \sum_{n=1}^N \bigg( x_n^{2H+1} - x_{n-1}^{2H+1} - \frac{1}{2(H+1)} \Delta x_{n-1}^{2H+1} \bigg) \\
      &= \frac{1}{(2H+1)N} \sum_{n=1}^N \bigg( (2H+1) x_n^{2H} \Delta x_{n-1} + \bigo(\Delta x_{n-1}^2) \\
      &\hspace{7.16cm}- \frac{1}{2(H+1)} \Delta x_{n-1}^{2H+1} \bigg) \\
      &= \Theta(N^{-1})
      \end{split}
  \end{equation*}
  when $s = 1$.
\end{proof}
\subsection{Proofs for~\Cref{sec:icv-estimators}} \label{sec:proofs-icv}

For the Brownian motion kernel, the ICV estimator defined in~\eqref{eq:icv-estimator} takes the explicit form
\begin{equation*}
    \tausqicvest = \frac{1}{N} \sum_{n=2}^{N-1}  \frac{( \Delta x_{n-1} [f_{n+1} - f_n] - \Delta x_n [f_n - f_{n-1}] )^2}{ (\Delta x_n + \Delta x_{n-1}) \Delta x_n \Delta x_{n-1} }.
\end{equation*}
We analyse this estimator below.

\begin{proof}[Proof of \Cref{res:holder-spaces-icv}]

The proof of~\Cref{res:holder-spaces} shows that when $s=1$ and $\alpha \in (1/2, 1]$, the bound is dominated by the bound on the boundary terms, $B_{1,N}=\bigo(N^{-2})$ and $B_{2,N}=\bigo(N^{-2})$, since
\begin{equation*}
    \tausqcvest = B_{1,N} + I_N + B_{2,N} = \bigo(N^{-2}) + \bigo(N^{-1-2\alpha}) + \bigo(N^{-2}) = \bigo(N^{-2}).
\end{equation*}
As $\tausqicvest = I_N$, it follows that $\tausqicvest = \bigo(N^{-1-2\alpha})$ when $s=1$.
\end{proof}

\begin{proof}[Proof of \Cref{res:holder-spaces-icv-exp}]

The proof of~\Cref{res:holder-spaces-exp} shows that when $s=1$ and $H \in [1/2, 1)$, the bound is dominated by the bound on the right boundary terms, $B_{2,N}=\Theta(N^{-2})$, since
\begin{equation*}
\begin{split}
    \bE \tausqcvest &= \frac{1}{2(H+1)(2H+1)} B_{1,N} + \frac{1}{2(H+1)(2H+1)} I_N + \frac{1}{(2H+1)} B_{2,N} \\
        &= \bigo(N^{-2-2H}) + \Theta(N^{-1-2H}) + \Theta(N^{-2})
\end{split}
\end{equation*}
As $\bE \tausqicvest = I_N/(2(H+1)(2H+1))$, it follows that $\tausqicvest = \Theta(N^{-1-2H})$ when $s=1$.
\end{proof}

\subsection{Proofs for \Cref{sec:discussion}}
\label{sec:proofs-discussion}

\begin{proof}[Proof of \Cref{res:uq-theorem-exp-bq}]
  Since we assumed $\bP$ has a density that maps $[0, T]$ to $[c_0, C_0]$, for any measurable $g(x)$ it holds that
  \begin{equation*}
      c_0 \int_0^T g(x) \d x \leq \int_0^T g(x) \bP(\d x) \leq C_0 \int_0^T g(x) \d x,
  \end{equation*}
  and
  \begin{equation}
  \label{eq:bq_credible_intervals_proof_5}
      \frac{c_0^2}{C_0^2} R_0 \leq \frac{\bE \Bigg[  \int_0^T (f(x) - m_N(x)) \bP(\d x) \Bigg]^2}{\int_0^T\int_0^T k_N(x, x') \bP(\d x) \bP(\d x')} \leq \frac{C_0^2}{c_0^2} R_0
  \end{equation}
  for
  \begin{equation*}
      R_0 = \frac{\bE \Bigg[  \int_0^T (f(x) - m_N(x)) \d x \Bigg]^2}{\int_0^T\int_0^T k_N(x, x') \d x \d x'}.
  \end{equation*}
 Therefore, bounding $R_0$, the fraction with the Lebesgue measure integrals, is sufficient to bound the fraction with measure $\bP$. We only provide the proof for the case $s = 1$; the case $s = 0$ is simpler and analogous. 
 
 For $k_N$ in~\eqref{eq:explicit-post-cov} and the uniform grid $x_n = nT/N$ for $n \in \{0, \dots, N\}$, 
 \begin{equation}
 \label{eq:bq_credible_intervals_proof_4}
      \int_0^T\int_0^T k_N(x, x') \d x \d x' =  \frac{T^3}{12 N^2}
  \end{equation}
 therefore, once we show 
  \begin{equation*}
      I \coloneqq \bE \Bigg[  \int_0^T (f(x) - m_N(x)) \d x \Bigg]^2 = \Theta(N^{-\max(2H+3, 4)}) 
  \end{equation*}
  the statement will follow immediately from the asymptotics for $\bE \tausqcvest$ and $\bE \tausqmlest$ in \Cref{res:holder-spaces-exp} and \Cref{res:holder-spaces-exp-ml}. 

  Let $h \coloneqq T/N$ be the distance between points on the uniform grid. From the expression for $m_N$ in \Cref{sec:explicit-post-mean-cov} for the uniform case of $x_n=nh$ for $n \in \{0, \dots, N\}$, we get
  \begin{align}
  \label{eq:bq_credible_intervals_proof_1}
      I &= \bE \Bigg[ \sum_{n=1}^N \int_{(n-1)h}^{nh} \Big(f(x) - \frac{nh - x}{h} f((n-1)h) - \frac{x - (n-1)h}{h} f(nh)\Big) \d x \Bigg]^2  \nonumber \\
      &= \frac{1}{h^2}\bE \Bigg[ \sum_{n=1}^N \int_{(n-1)h}^{nh} \Big((x - (n-1)h)(f(nh) - f(x)) \nonumber\\
      &\hspace{4cm}- (nh - x)(f(x) - f((n-1)h))\Big) \d x \Bigg]^2 
  \end{align}
Since $f$ is integrated fractional Brownian motion, for $G(t)$ being fractional Brownian motion it holds that
\begin{align*}
   f(x) = \int_0^x G(t) \d t = \int_0^T G(t) \mathds{1}_{t \leq x} \d t,
\end{align*}
and for $(n-1)h \leq x \leq nh$,
\begin{align*}
   f(x) - f((n-1)h) = \int_{(n-1)h}^{nh} G(t)\mathds{1}_{t \leq x} \d t,\quad f(nh) - f(x) = \int_{(n-1)h}^{nh} G(t)\mathds{1}_{t \geq x} \d t
\end{align*}
Substituting these into~\eqref{eq:bq_credible_intervals_proof_1} and exchanging the order of integration by Fubini's theorem, we get the convenient form
\begin{align*}
  I = \bE \Bigg[ \sum_{n=1}^N \int_{(n-1)h}^{nh} \psi_n(t) G(t) \d t \Bigg]^2, \qquad \psi_n(t) \coloneqq ( nh - h/2 - t)
\end{align*}
Then, since $\bE [G(t)G(s)] = k_\mathrm{FBM}(t,s) = (t^{2H} + s^{2H} - |t - s|^{2H})/2$,
\begin{align*}
   2I &= \sum_{n=1}^N \sum_{m=1}^N \int_{(n-1)h}^{nh} \int_{(m-1)h}^{mh}  \psi_n(t)\psi_m(s) (t^{2H} + s^{2H} - |t - s|^{2H}) \d s \d t \\
   &\overset{(A)}{=}\sum_{n=1}^N \sum_{m=1}^N \int_{-h/2}^{h/2} \int_{-h/2}^{h/2}  ts (t^{2H} + s^{2H} - |t - s + nh - mh|^{2H}) \d s \d t \\
   &\overset{(B)}{=} - \sum_{n=1}^N \sum_{m=1}^N \int_{-h/2}^{h/2} \int_{-h/2}^{h/2}  ts |t - s + nh - mh|^{2H} \d s \d t \\
   &\overset{(C)}{=} - \sum_{n=1}^N \int_{-h/2}^{h/2} \int_{-h/2}^{h/2}  ts |t - s|^{2H} \d s \d t \\
   &\hspace{0.5cm}- 2\sum_{n=1}^{N-1} \sum_{m=n+1}^N \int_{-h/2}^{h/2} \int_{-h/2}^{h/2}  ts (t - s + nh - mh)^{2H} \d s \d t \\
   & \eqqcolon S_1 + S_2. 
\end{align*}
Here, equality $(A)$ is obtained through change of variables, $t \rightarrow t + nh - h/2$ and $s \rightarrow s + mh - h/2$; equality $(B)$ holds due to the antisymmetric function $f(x)=x$ integrating to zero on a symmetric domain $[-h/2, h/2]$; equality $(C)$ holds due to the full expression being symmetric with respect to $m$ and $n$. 

Next, we simplify $S_1$ and $S_2$. Substituting $h=T/N$, $S_1$ may be computed exactly, as
\begin{align}
\label{eq:bq_credible_intervals_proof_2}
    S_1 = - \sum_{n=1}^N \int_{-h/2}^{h/2} \int_{-h/2}^{h/2}  ts |t - s|^{2H} \d s \d t = \frac{ H T^{2H + 4}}{(2H+1)(2H+2)(2H+4)} N^{-2H-3}.
\end{align}
$S_2$ requires more care. First, notice that 
\begin{align*}
    S_2 &= - 2\sum_{n=1}^{N-1} \sum_{m=n+1}^N \int_{-h/2}^{h/2} \int_{-h/2}^{h/2}  ts (t - s + nh - mh)^{2H} \d s \d t \\
        &=- 2\sum_{n=1}^{N-1} \sum_{d=1}^{N-n} \int_{-h/2}^{h/2} \int_{-h/2}^{h/2}  ts (t - s + dh)^{2H} \d s \d t \\
        &=- 2\sum_{d=1}^{N-1} (N-n) \int_{-h/2}^{h/2} \int_{-h/2}^{h/2}  ts (t - s + dh)^{2H} \d s \d t \\
        &\eqqcolon \sum_{d=1}^{N-1} (N-d) I_1(d)
\end{align*}
Performing integration by parts on the inner and outer integrals in $I_1(d)$, we get
\begin{align*}
    I_1(d) &= 
    \frac{ h^2 ( dh - h)^{2H+2} + 2 h^2 (dh)^{2H+2}+ h^2 (dh + h)^{2H+2}}{2(2H+1)(2H+2)} \\
    &\qquad+ \frac{2h ( dh - h)^{2H+3} - 2h(dh + h)^{2H+3}}{(2H+1)(2H+2)(2H+3)} \\
    &\qquad+ \frac{ 2( dh - h)^{2H+4} - 4 ( dh)^{2H+4} +  2( dh + h)^{2H+4}}{(2H+1)(2H+2)(2H+3)(2H+4)}.
\end{align*}
Then, using telescoping sums of the form $\sum_{d=1}^{N-1} (a_{d+1} - a_d) = a_N - a_1$,
\begin{align*}
    \sum_{d=1}^{N-1} N I_1(d) &= 
     h^{2H+4}\frac{- (N-1)^{2H+3}  - (N-1)^{2H+2} - N + N^{2H+3}}{2(2H+1)(2H+2)}  \\
    &\qquad+2 h^{2H+4} \frac{ - ( N-1)^{2H+4} - ( N-1)^{2H+3} + N - N^{2H+4}}{(2H+1)(2H+2)(2H+3)} \\
    &\qquad+ 2 h^{2H+4}\frac{ - ( N-1)^{2H+5} - ( N-1)^{2H+4}  - N +  N^{2H+5}}{(2H+1)(2H+2)(2H+3)(2H+4)}\\
     &\qquad+2 h^{2H+4} N \sum_{d=1}^{N-1}\frac{ d^{2H+2}}{(2H+1)(2H+2)}
\end{align*}
and
\begin{align*}
    \sum_{d=1}^{N-1} dI_1(d) &= 
    h^{2H+4}\frac{ - ( N-1)^{2H+2} + 1 - N^{2H+2}}{2(2H+1)(2H+2)} \\
    &\qquad+h^{2H+4}\frac{- (N-1)^{2H+3} - 1 + N^{2H+3} }{2(2H+1)(2H+2)} \\
    &\qquad+ 2 h^{2H+4}\frac{- (N-1)^{2H+4} + 1 - N^{2H+4}}{(2H+1)(2H+2)(2H+3)} \\
    &\qquad+ 2 h^{2H+4}\frac{ - (N-1)^{2H+3} - 1 + N^{2H+3}}{(2H+1)(2H+2)(2H+3)}  \\
    &\qquad+ 2 h^{2H+4} \frac{- (N-1)^{2H+5} - 1 +  N^{2H+5}}{(2H+1)(2H+2)(2H+3)(2H+4)} \\
    &\qquad+ 2h^{2H+4} \frac{ - (N-1)^{2H+4}  + 1 - N^{2H+4}}{(2H+1)(2H+2)(2H+3)(2H+4)} \\
    &\qquad+2 h^{2H+4}\sum_{d=1}^{N-1}\frac{ (2H+5)d^{2H+3} }{(2H+1)(2H+2)(2H+3)}.
\end{align*}
Subtracting $\sum_{d=1}^{N-1} dI_1(d)$ from $\sum_{d=1}^{N-1} N I_1(d)$, grouping matching powers, and substituting $h=T/N$, we get 
\begin{align*}
    S_2 &=\sum_{d=1}^{N-1} (N - d) I_1(d) \\
    &= N^{-2H-4} C(T, H) \Big[ 2 N^{2H+4} -4(H+2) N^{2H+3} \\
    &\hspace{2cm}+(H+2)(2H+3) N^{2H+2} - H (2H+3) N \\
    &\hspace{2cm}+4(H+2)\Big((2H+3) N \sum_{d=1}^{N-1} d^{2H+2}
    -(2H+5)\sum_{d=1}^{N-1} d^{2H+3}\Big) \Big]
\end{align*}
for
\begin{align*}
    C(T, H) &\coloneqq \frac{T^{2H+4}}{(2H+1)(2H+2)(2H+3)(2H+4)}.
\end{align*}
Summing up with the expression for $S_1$ in~\eqref{eq:bq_credible_intervals_proof_2}, we get
\begin{align}
\label{eq:bq_credible_intervals_proof_3}
    2I = S_1 + S_2  = N^{-2H-4} C(T, H) S_3
\end{align}
for
\begin{align*}
    S_3 &= 2 N^{2H+4} -4(H+2) N^{2H+3} +(H+2)(2H+3) N^{2H+2} \\
    &\qquad+4(H+2)\Big((2H+3) N \sum_{d=1}^{N-1} d^{2H+2}
    -(2H+5)\sum_{d=1}^{N-1} d^{2H+3}\Big)
\end{align*}
We expand the remaining $d$-sums via the Euler-Maclaurin formula for $f(x) = x^p$. Denote by $B_n$ the Bernoulli numbers, and by $p^{\underline{n}} = p(p-1)\cdots(p-n+1)$ the falling factorial. Then, for $m \geq 1$, the Euler–Maclaurin formula for $n^p$ states that
\begin{align*}
    \sum_{n=1}^N n^p = \frac{N^{p+1} - 1}{p+1} + \frac{N^p + 1}{2} + \sum_{k=1}^{m} \frac{B_{2k}}{(2k)!} p^{\underline{2k-1}} ( N^{p+1-2k} - 1) + R_{2m}
\end{align*}
where the remainder takes the exact form
\begin{align*}
    R_{2m,p} = - \frac{p^{\underline{2m}}}{(2m)!} \int_1^N  x^{p-2m} B_{2m}(x - \lfloor x \rfloor) \d x,
\end{align*}
and is bounded as
\begin{align*}
    |R_{2m,p}| \leq \frac{2 \zeta(2m)}{ (2 \pi)^{2m}} |p^{\underline{2m}}| \frac{N^{p-2m+1} - 1}{p - 2m + 1}.
\end{align*}
Expanded until $m=3$, the lowest order that keeps the remainders of sufficiently low order to make asymptotics clear, this takes the form
\begin{align*}
    \sum_{n=1}^N n^p = \frac{N^{p+1} - 1}{p+1} &+ \frac{N^p + 1}{2} + \frac{p}{12} ( N^{p-1} - 1)
    - \frac{p(p-1)(p-2)}{720} ( N^{p-3} - 1) \\
    &+ \frac{1}{6 \times 7!} p(p-1)(p-2)(p-3)(p-4) ( N^{p-5} - 1) + R_{6,p}
\end{align*} 
for
\begin{align*}
    R_{6,p} &= - \frac{p(p-1)(p-2)(p-3)(p-4)(p-5)}{6!} \int_1^N  x^{p-6} B_6(x - \lfloor x \rfloor) \d x, \\
    |R_{6,p}| &\leq \frac{1}{945 \times 2^5} |p(p-1)(p-2)(p-3)(p-4)(N^{p-5} - 1)|.
\end{align*}
Substituting this form for $p=2H+2$ and $p=2H+3$, we get
\begin{align*}
    S_3 &= A(H) N^{2H} \\
    &+ B(H) N \\
    &+ C(H) \\
    &- \frac{1}{3 \times 6!} (2H+4)(2H+3)(2H+2)(2H+1)2H(2H-1) N^{2H-2}  \\
    & + 2(2H+3)(2H+4) N R_{6,2H+2} - 2(2H+4)(2H+5) R_{6,2H+3}
\end{align*}
for
\begin{align*}
    A(H)&= \frac{(2H+1)(2H+2)(2H+3)(2H+4)}{72} \\
    B(H)&= -\frac{H(H-2)(H-1)(H+2)(2H-5)(4H^2 + 28 H + 69)}{945} \\
    C(H)&= \frac{(H-2)(2H-3)(2H-1)(2H+1)(2H+5)(H^2 + 8H + 21)}{1890}.
\end{align*}
As $A(H)>0$ and $B(H)>0$ for $0<H<1$, it holds that
\begin{align*}
    I = \bigo(N^{\max\{2H, 1\}}).
\end{align*}
For large enough $N$ to subsume the $N^b$ terms for $b \leq 0$, $I$ is lower bounded by
\begin{align*}
    I &= A(H) N^{2H} \\
    &+ B(H) N \\
    & - \frac{1}{945 \times 2^5} |(2H+2)(2H+1)2H(2H-1)(2H-2)| N,
\end{align*}
where the last $N$ term comes from the lower bound on $R_{6,2H+2}$. Since for $0<H<1$,
\begin{align*}
    B(H)  - \frac{1}{945 \times 2^5} |(2H+2)(2H+1)2H(2H-1)(2H-2)| \geq 0
\end{align*}
it holds that
\begin{align*}
    S_3 = \Theta(N^{\max\{2H, 1\}}).
\end{align*}
Therefore, by~\eqref{eq:bq_credible_intervals_proof_3},
\begin{align}
    I = N^{-2H-4} \Theta(N^{\max\{2H, 1\}})  \Theta(N^{\max\{-4, -2H-3\}}).
\end{align}
By the $N^{-2}$ asymptotics of the variance shown in~\eqref{eq:bq_credible_intervals_proof_4} and the $\bP$-bounds in~\eqref{eq:bq_credible_intervals_proof_5},
\begin{align*}
      \frac{\bE \Bigg[  \int_0^T (f(x) - m_N(x)) \bP(\d x) \Bigg]^2}{\int_0^T\int_0^T k_N(x, x') \bP(\d x) \bP(\d x')} &= \Theta(N^{\max\{-2, -1-2H\}})\\&=
      \begin{cases}
        \Theta\left(N^{-1-2H}\right) &\text{ if } \quad  s = 1 \text{ and } H < 1/2, \\
        \Theta\left(N^{-2}\right) &\text{ if } \quad s = 1 \text{ and } H \geq 1/2. \\
      \end{cases}
  \end{align*}
and the result follows from \Cref{res:holder-spaces-exp} and \Cref{res:holder-spaces-exp-ml}.
\end{proof}

\begin{proof}[Proof of \Cref{res:uq-theorem-exp}]
  We only provide the proof for the case $s = 1$ and leave the simpler case $s = 0$ to the reader.
  Let $x \in (x_{n-1}, x_n)$.
  From the expression for $m_N$ in \Cref{sec:explicit-post-mean-cov}, we get
  \begin{equation*}
    \begin{split}
      \bE [ f(x) - m_N(x) ]^2 &= \bE \Bigg[ f(x) - \frac{(x_n - x) f(x_{n-1}) + (x - x_{n-1}) f(x_n)}{\Delta x_{n-1}} \Bigg]^2 \\
      &= \frac{1}{\Delta x_{n-1}^2} \bE \big[ (x - x_{n-1})(f(x_n) \\
      &\hspace{2cm}- f(x)) - (x_n - x)(f(x) - f(x_{n-1})) \big]^2.
      \end{split}
  \end{equation*}
  Then, we can use~\eqref{eq:s=1-exp-2} with $x_n$ instead of $x_{n+1}$ and $x$ instead of $x_n$ to get
  \begin{align*}
      \bE [ f(x) - m_N(x) ]^2 &= \frac{(x_n - x)(x - x_{n-1})}{C_H \Delta x_{n-1}} \\
      &\hspace{2cm} \times \big[ \Delta x_{n-1}^{2H+1} - (x_n - x)^{2H+1} - (x - x_{n-1})^{2H+1} \big],
  \end{align*}
  where $C_H = 2(H+1)(2H+1)$.
  The expression for $k_N$ in \Cref{sec:explicit-post-mean-cov} gives
  \begin{equation*}
    \frac{\bE [ f(x) - m_N(x) ]^2}{k_N(x)} = \frac{1}{C_H} \big[ \Delta x_{n-1}^{2H+1} - (x_n - x)^{2H+1} - (x - x_{n-1})^{2H+1} \big].
  \end{equation*}

  By removing the negative terms and using the quasi-uniformity~\eqref{eq:quasi-uniformity-2}, we obtain
  \begin{equation*}
      \sup_{x \in [0, T]} \frac{\bE [ f(x) - m_N(x) ]^2}{k_N(x)} \leq \frac{ (T C_\textup{qu})^{2H+1} }{C_H} N^{-1-2H},
  \end{equation*}
  To see that this bound is tight, observe that for the midpoint $x = (x_n + x_{n-1}) / 2$ we have $x_n - x = x - x_{n-1} = \Delta x_{n-1} / 2$ and
  \begin{equation*}
    \frac{ \bE [ f(x) - m_N(x) ]^2 }{ k_N(x) } = \frac{1}{ C_H} \bigg(1 - \frac{1}{2^{2H}} \bigg) \Delta x_{n-1}^{2H+1} \geq \frac{T^{2H + 1}}{C_H C_\textup{qu}^{2H + 1}} \bigg(1 - \frac{1}{2^{2H}} \bigg) N^{-1-2H}
  \end{equation*}
  by the quasi-uniformity.
  Therefore
  \begin{equation*}
    \sup_{x \in [0, T]} \frac{\bE [ f(x) - m_N(x) ]^2}{k_N(x)} = \Theta(N^{-1-2H})
  \end{equation*}
  when $s = 1$.
  It can be similarly shown that
  \begin{equation*}
    \sup_{x \in [0, T]} \frac{\bE [ f(x) - m_N(x) ]^2}{k_N(x)} = \Theta(N^{1-2H})
  \end{equation*}
  when $s = 0$.
  The claims then follow from the rates for $\bE \tausqcvest$ and $\bE \tausqmlest$ in \Cref{res:holder-spaces-exp,res:holder-spaces-exp-ml}.
\end{proof}

\section{Further discussion on~\Cref{res:fqv-estimator} }
\label{sec:discussion-thm-fqv-estimator}
The requirement of having the same $V^2(f)$ for all sequences of partitions quasi-uniform with constant $2$ can be relaxed somewhat: trivially, it is sufficient that the quadratic variation is $V^2(f)$ specifically with respect to even-points and odd-points sequences of sub-partitions used in the proof in \Cref{sec:proofs-deterministic}. Furthermore, we may even have different quadratic variations with respect to said sequences. Then the results becomes
\begin{equation*} \label{eq:liminf-limsup-v2f-generalisation}
   \lim_{N \to \infty} \hat{\tau}^2_\mathrm{CV} = \frac{\nu}{T} \qquad \text{for} \qquad \nu = \frac{ V_0^2(f) + V_1^2(f)}{2},
\end{equation*}
where $V_0^2(f)$ and $V_1^2(f)$ are quadratic variations with respect to the even- and odd-points sub-partitions, respectively, meaning that
\begin{align*}
  V^2(f) &= \lim_{N \to \infty} \sum_{n=1}^{N-1} (f_{n+1} - f_{n} )^2, \\
  V_0^2(f) &= \lim_{N \to \infty} \sum_{n=1}^{\lfloor \frac{N-2}{2} \rfloor} (f_{2n+2} - f_{2n} )^2, \\
  V_1^2(f) &= \lim_{N \to \infty} \sum_{n=1}^{\lfloor \frac{N-1}{2} \rfloor} (f_{2n+1} - f_{2n-1} )^2.
\end{align*}

\section{Explicit expression for the leave-$p$-out estimator}
\label{sec:explicit_expression_for_leave_p_out}

Using the expressions for posterior mean and covariance functions in~\eqref{eq:explicit-post-mean} and~\eqref{eq:explicit-post-cov}, we may derive an explicit expression for the leave-$p$-out cross-validation (LPO-CV) estimator of the amplitude parameter, given by
\begin{equation*}
  \hat{\tau}_{\mathrm{CV}(p)}^2 = \frac{1}{C(N,p)} \sum_{i=1}^{C(N, p)} \frac{1}{p} \sum_{n=1}^p \frac{[f(x_{p,i,n}) - m_{\setminus \{p, i\}}(x_{p,i,n})]^2}{ k_{\setminus \{p, i\}}(x_{p,i, n})}.
\end{equation*}
The expression is less straightforward than that for $p=1$. Denote by $x_{\lfloor p,i,n \rfloor}$ the largest point in the set $\bx_{\setminus \{p,i\}} = \bx \setminus \{x_{p,i,1}, \ldots, x_{p,i,p}\}$ that does not exceed $x_{p,i,n}$, and by $x_{\lceil p,i,n \rceil}$ the smallest point in the set $\bx_{\setminus \{p,i\}}$ that exceeds $x_{p,i,n}$. Through somewhat cumbersome arithmetic derivations it can be shown that the estimator takes the form
\begin{align*}
  \hat{\tau}_{\mathrm{CV}(p)}^2 = \frac{1}{C(N,p)} \sum_{i=1}^{C(N, p)} \bigg[  B_{p,i,1} + \sum_{n=2}^{p-1} I_{p,i,n} + B_{p,i,p} \bigg]
\end{align*}
where, for $\Delta x^-_{p,i,n} = (x_{p,i,n} - x_{\lfloor p,i,n \rfloor})$ and $\Delta x^+_{p,i,n} = (x_{ \lceil p,i,n \rceil } - x_{p,i,n})$, the inner term is
\begin{equation*}
    I_{p,i,n} = \frac{\Delta x^-_{p,i,n}(f_{\lceil p,i,n \rceil} - f_{p,i,n }) - \Delta x^+_{p,i,n} (f_{p,i,n} - f_{\lfloor p,i,n \rfloor})}{ (\Delta x^+_{p,i,n} + \Delta x^-_{p,i,n}) \Delta x^+_{p,i,n} \Delta x^-_{p,i,n}},
\end{equation*}
and the boundary terms $B_{p,i,1}$ and $B_{p,i,p}$ depend on whether the $i$-th set contains $x_1$ or $x_N$, respectively. Specifically,
\begin{align*}
    B_{p,i,1} &= \begin{cases}
    \frac{(x_{ \lceil p,i,1 \rceil} f_{p,i,1} -  x_{p,i,1} f_{ \lceil p,i,1 \rceil} )^2}{ x_{ p,i,1 } x_{ \lceil p,i,1 \rceil} \Delta x^+_{p,i,1} } & \text{if the $i$-th set contains $x_1$,}\\
    I_{p,i,1} & \text{otherwise,}
    \end{cases} \\
    B_{p,i,p} &= \begin{cases}
    \frac{(f_{p,i,p} - f_{ \lfloor p,i,p \rfloor})^2}{ \Delta x^-_{p,i,p} } & \text{if the $i$-th set contains $x_N$,}\\
    I_{p,i,p} & \text{otherwise.}
    \end{cases}
\end{align*}
Though more cumbersome, it may be feasible to conduct convergence analysis similar to that in~\Cref{sec:limit-behaviour-for-sigma} for $\hat{\tau}_{\mathrm{CV}(p)}^2$. We leave this up to future work.

\section{Comparison of CV and ML estimators for Mat\'ern kernels} 
\label{sec:cv-vs-ml-sobolev}

\begin{figure}
    \centering
    \includegraphics[width=0.8\textwidth]{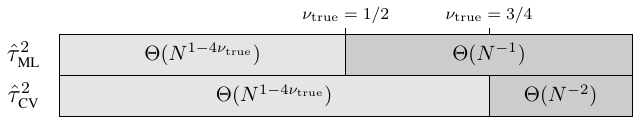}
    \caption[Rates of decay for the ML and CV estimators for the Mat\'ern kernel]{Rates of decay for the ML and CV estimators for the Mat\'ern kernel of order $1$, and a true function that is a linear combination of Mat\'ern kernels of order $\nu_\mathrm{true}$. The ML rate is given in~\citet[Equation 5.2]{Karvonen2020}. The CV rate is empirically observed in~\Cref{fig:cv-vs-ml-sobolev}.
    Observe that the CV estimator's range of adaptation to the smoothness $\nu_\mathrm{true}$ is wider than the ML estimator's.
    }
    \label{fig:sobolev-fig}
\end{figure}

\begin{figure}
    \centering
    \includegraphics[width=\textwidth]{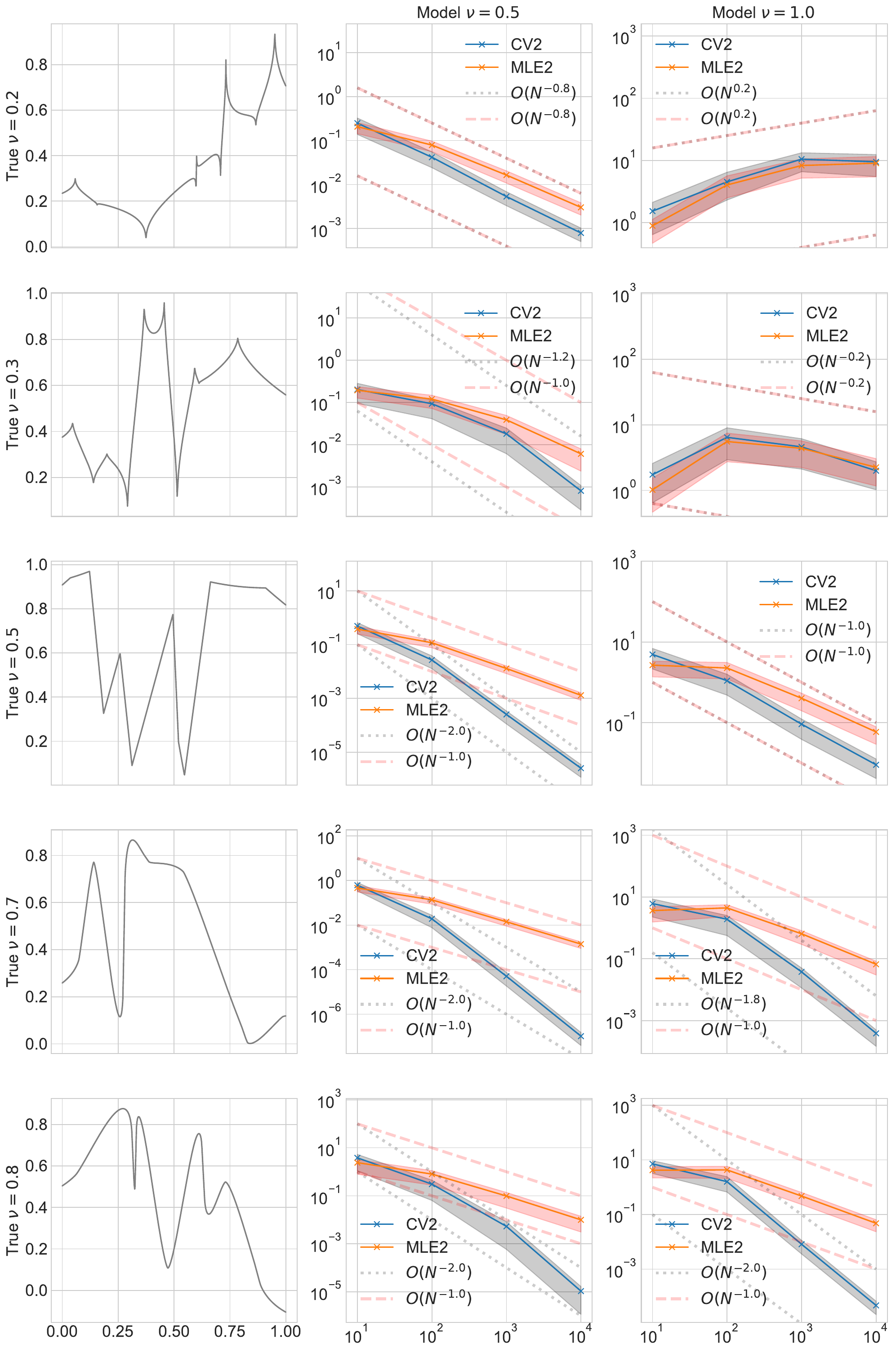}
    \caption[Asymptotics of CV estimator compared to asymptotics of the ML estimator, for the Mat\'ern kernel]{Asymptotics of CV estimator compared to asymptotics of the ML estimator, for the Mat\'ern kernel of order $\nu_\mathrm{model}$, and a true function that is a finite linear combination of Mat\'ern kernels of order $\nu_\mathrm{true}$.}
    \label{fig:cv-vs-ml-sobolev}
\end{figure}

A natural next step is extending the analysis to Sobolev kernels introduced in \Cref{def:sobolev_kernel}, such as the commonly used Mat\'ern kernels. The ML estimator for Mat\'ern kernels was analysed in~\citet{Karvonen2020}.
Their experiments in Section~5.1 suggest that, for $x_+ \coloneqq \max(x, 0)$,
\begin{equation}
\label{eq:karvonen2020mle}
    \tausqmlest = \Theta(N^{2(\nu_\mathrm{model} - 2 \nu_\mathrm{true})_+ -1})
\end{equation}
when $k_{\nu_\mathrm{model}}$ is a Mat\'ern kernel of order $\nu_\mathrm{model}$ and $f$ is a finite linear combination of the form $f = \sum_{i=1}^m \alpha_i k_{\nu_\mathrm{true}}(\cdot, x_i)$ for some $m \in \bN_{\geq 1}$, $\alpha_i \in \bR$, $x_i \in [0, 1]$, and the Mat\'ern kernel $k_{\nu_\mathrm{true}}$ of order $\nu_\mathrm{true}$.
Empirically, we compare this to the rate of the CV estimator in~\Cref{fig:cv-vs-ml-sobolev}. The test functions $f$ are posterior means of a GP with the $k_{\nu_\mathrm{true}}$ kernel conditioned on points $\{(x_1, y_1), \dots, (x_{10}, y_{10})\}$, where each $x_i$ and $y_i$ is sampled i.i.d. from the uniform distribution on $[0,1]$. Since such $f$ are of the form $f = \sum_{i=1}^{10} \alpha_i k_{\nu_\mathrm{true}}(\cdot, x_i)$, we expect the MLE rate in~\eqref{eq:karvonen2020mle} to apply; we use experimental data and the results in~\Cref{res:holder-spaces,res:holder-spaces-exp-ml} to hypothesise what the rate in each individual example is. 
Similarly to the observations for the Brownian motion kernel, we see that the CV estimator adapts to the smoothness of the true function over a larger range of smoothness compared to the ML estimator. For instance, for $\nu_\mathrm{model}=1$, the experimental results suggest that the dependence of rate on $\nu_\mathrm{true}$ is as illustrated in~\Cref{fig:sobolev-fig}. While the CV and the ML estimators adapt to the function smoothness when $\nu_\mathrm{true} \leq 1/2$, for $\nu_\mathrm{true} \in [1/2, 3/4]$ only the CV estimator continues adapting to the smoothness. This implies the CV estimator is less likely to become asymptotically overconfident in the event of undersmoothing.

\chapter{Kernel Quantile Embeddings: Supplementary Materials}
\label{sec:app_kme}
\section{Probability Metrics and Their Estimators }\label{appendix_sec:estimators}

\subsection{Sub-$N^2$ Estimators of the MMD}
A linear estimator of the MMD was proposed in~\citet[Lemma 14]{gretton2012kernel}. This estimator has computational complexity $\bigo(N)$ and a convergence rate of $\bigo(N^{-\nicefrac{1}{2}})$, and we will refer to it as \emph{MMD-Lin}. The estimator is given by
\begin{align*}
    \MMD_\text{lin}^2(\bP, \bQ) \coloneqq \frac{1}{\lfloor N/2\rfloor} \sum_{n=1}^{\lfloor N/2\rfloor} &k(x_{2n-1},x_{2n})+k(y_{2n-1},y_{2n})\\
    &-k(x_{2n-1},y_{2n})-k(x_{2n},y_{2n-1})
\end{align*}
\textit{MMD-Multi} estimators of the MMD are due to \citet{Schrab2022}, and take the following form
\begin{align*}
    \MMD_\text{Multi}^2(\bP, \bQ) \coloneqq \frac{2}{R(2N-R-1)}\sum_{r=1}^R\sum_{n=1}^{N-r} &k(x_n, x_{n+r}) + k(y_n, y_{n+r}) \\
    &- k(x_n, y_{n+r}) - k(x_{n+r}, y_n)
\end{align*}
where $R$ is the number of subdiagonals considered. \textit{MMD-Multi} estimators have computational complexity $\bigo(RN)$. In our experiments, to match the complexity with $\ekqd$, we set $R = \log^2(N)$.

Several estimators with faster convergence rates exist \citep{Niu2023,Bharti2023}, but these have computational cost ranging from $\bigo(N^2)$ to $\bigo(N^{3})$ and require more regularity conditions on $k, \bP$ and $\bQ$, and we therefore omit them from our benchmark. \citet{Bodenham2023} also introduced an estimator with computational complexity of $\bigo(N \log N)$ (and convergence $\bigo(N^{-\nicefrac{1}{2}})$) using slices/projections to $d=1$. However, their approach is restrictive in that it can only be used for the Laplace kernel, and we therefore also do not compare to it.

\subsection{Wasserstein Distance}

The $p$-Wasserstein distance \citep{Kantorovich1942,villani2009optimal} is defined as
\begin{align*}
    W_p(\bP, \bQ) \coloneqq \left(\inf_{\pi \in \Gamma(\bP, \bQ)} \bE_{(X,Y)\sim \pi}\left[c(X,Y)^p \right]\right)^{\nicefrac{1}{p}}.
\end{align*}
Given samples $x_1, \dots, x_N \sim \bP$ and $y_1, \dots, y_N \sim \bQ$, this distance can be approximated using a plug-in $W_p(\nicefrac{1}{N}\sum_{n=1}^N \delta_{x_n},\nicefrac{1}{N} \sum_{n=1}^N \delta_{y_n})$, which can be computed in closed-form at a cost of $\bigo(N^3)$, but converges to $W_p(\bP, \bQ)$ with a convergence rate $\bigo(N^{-\nicefrac{1}{d}})$.

When $\calX \subseteq \bR^d$ and $p=1$, we obtain the $1$-Wasserstein distance which, similarly to the MMD, can be written as an integral probability metric \citep{muller1997integral},
\begin{align*}
    W_1(\bP, \bQ) \coloneqq \sup_{\|f\|_\text{Lip} \leq 1} \left| \bE_{X\sim \bP}[f(X)] - \bE_{X \sim \bQ} [f(X)]\right|
\end{align*}
where $\|f\|_{\text{Lip}} = \sup_{x,y \in \calX, x \neq y} |f(x)-f(y)|/\|x-y\|$ denotes the Lipschitz norm.

When $\bP, \bQ$ are distributions on a one dimensional space $\calX \subseteq \bR$ that have $p$-finite moments, the $p$-Wasserstein distance can be expressed in terms of distance between quantiles of $\bP$ and $\bQ$ (see for instance~\citet[Remark 2.30]{peyre2019computational})
\begin{align}
\label{eq:wasserstein_as_quantiles}
    W_p(\bP, \bQ) = \left( \int_0^1 |\rho^{\alpha}_\bP - \rho^{\alpha}_\bQ|^p \d \alpha \right)^{\nicefrac{1}{p}}
\end{align}
A natural estimator for the Wasserstein distance is therefore based on approximating these one-dimensional quantiles using order statistics. Given $x_1, \dots, x_N \sim \bP$ and $y_1, \dots, y_N \sim \bQ$, denote by $\bP_N = \nicefrac{1}{N}\sum_{n=1}^N \delta_{x_n}$ and $\bQ_N = \nicefrac{1}{N} \sum_{n=1}^N \delta_{y_n}$ the corresponding empirical approximations to $\bP$ and $\bQ$. Then, the $\nicefrac{n}{N}$-th quantile of $\bP_N$ and $\bQ_N$ are exactly the $n$-th order statistics $[x_{1:N}]_n$ and $[y_{1:N}]_n$, i.e., the $n$-th smallest element of $x_{1:N}$ and $y_{1:N}$ respectively. Then $W_p(\bP_N, \bQ_N)$ takes the exact form
\begin{align}
\label{eq:wasserstein_as_order_statistics}
    W_p(\bP_N, \bQ_N) = \left(\sum_{n=1}^N \left| [x_{1:N}]_n - [y_{1:N}]_n\right|^p \right)^{\nicefrac{1}{p}},
\end{align}
and is an estimator of $W_p(\bP, \bQ)$. This estimator costs $\bigo(N \log N)$ to compute (due to the cost of sorting $N$ datapoints), and a convergence rate of $\bigo(N^{-\nicefrac{1}{2}})$ for $p=1$, and minimax convergence rate $\bigo(N^{-\nicefrac{1}{2p}})$ for integer $p>1$ when $\bP, \bQ$ have at least $2p$ finite moments. In some cases, the $p>1$ rate can be improved upon to match the $\bigo(N^{-\nicefrac{1}{2}})$ rate of $p=1$: we refer to~\citet{bobkov2019one} for a thorough overview.

\subsection{Sliced Wasserstein}

The \emph{sliced Wasserstein} (SW) distances \citep{rabin2011wasserstein,bonneel2015sliced} between two distributions $\bP, \bQ$ on $\bR^d$ use one-dimensional projections to reduce computational cost. \paragraph{Expected SW.} For an integer $p \ge 1$, expected SW is defined as
\begin{align*}
    \text{SW}_p(\bP, \bQ)
    &\coloneqq \left(\bE_{u \sim \bU(S^{d-1})} [W_p^p(\phi_u\# \bP, \phi_u\# \bQ)] \right)^{\nicefrac{1}{p}},
\end{align*}
where $\bU(S^{d-1})$ is the uniform distribution on the unit sphere $S^{d-1}$, the measures $\phi_u\# \bP$, $\phi_u\# \bQ$ are pushforwards under the projection operator $\phi_u(x) = \langle u, x \rangle$, and $W_p$ is the one-dimensional $p$-Wasserstein distance as in~\eqref{eq:wasserstein_as_quantiles}. Given $x_1, \dots, x_N \sim \bP$ and $y_1, \dots, y_N \sim \bQ$, the integral over the sphere is approximated by Monte Carlo sampling of $L$ directions $u_1, \dots u_L$, which together with the estimator in~\eqref{eq:wasserstein_as_order_statistics} gives
\begin{align*}
    \widehat{\text{SW}}_p^p(\bP, \bQ) &= \frac{1}{L}\sum_{l=1}^L W_p^p(\phi_{u_l}\# \bP_N, \phi_{u_l}\# \bQ_N)  \\
    &= \frac{1}{LN}\sum_{l=1}^L \sum_{n=1}^N \left(\big[\langle u_l, x_{1:N}\rangle\big]_n - \big[\langle u_l, y_{1:N}\rangle\big]_n \right)^p
\end{align*}
Here, $[\langle u_l, x_{1:N}\rangle]_n$ is the $n$-th order statistic, i.e., the $n$-th smallest element of $\langle u_l, x_{1:N}\rangle=[\langle u_l, x_1\rangle, \dots, \langle u_l, x_N\rangle]^\top$. This estimator can be computed in $\bigo(LN\log N)$ time (the cost of sorting $N$ samples, for $L$ directions) and was shown to converge at rate $\bigo(L^{-\nicefrac12} + N^{-\nicefrac12})$ for $p=1$ \citep{Nadjahi2020}.

\paragraph{Max SW.}
The \emph{max-sliced Wasserstein} (max-SW) distance \citep{Deshpande2018} replaces the average over projections in expected SW with a supremum over directions,
\begin{align*}
    \text{max-SW}_p(\bP, \bQ)
    &\coloneqq \left(\sup_{u \in S^{d-1}} W^p_p(\phi_u \# \bP, \phi_u \# \bQ) \right)^{\nicefrac{1}{p}},
\end{align*}
where, $\phi_u(x) = \langle u, x \rangle$ is again the projection operator, and $W_p$ is the one-dimensional $p$-Wasserstein distance of~\eqref{eq:wasserstein_as_quantiles}. Max-SW emphasises the direction of greatest dissimilarity between the two measures.

Given $x_1, \dots, x_N \sim \bP$ and $y_1, \dots, y_N \sim \bQ$, max-SW is estimated as $W_p(\phi_{u^*} \# \bP, \phi_{u^*} \# \bQ)$, for $u^*$ the projection that maximises $W^p_p(\phi_u \# \bP_N, \phi_u \# \bQ_N)$ as given in~\eqref{eq:wasserstein_as_order_statistics}. In~\citep{Deshpande2018}, $u^*$ was approximated by optimising a heuristic, rather than the actual $W^p_p(\phi_u \# \bP_N, \phi_u \# \bQ_N)$. Then,~\citet{kolouri_generalized_2022} approached the actual problem of
\begin{align*}
    u^*
    &= \argmax_{\|u\|=1} W^p_p(\phi_u \# \bP_N, \phi_u \# \bQ_N).
\end{align*}
by running projected gradient descent on $S^{d-1}$, where each gradient step requires computing the derivative of the 1D Wasserstein distance w.r.t. $u$. Concretely, they initialise $u_1$ randomly and iterate
\begin{align}
\label{eq:max_sw_optimisation}
    u_{t+1} = \mathrm{Proj}_{S^{d-1}}\Big(\mathrm{Optim}\big( \nabla_u W^p_p(\phi_{u_t} \# \bP, \phi_{u_t} \# \bQ), u_{1:t}\big)\Big),
\end{align}
where $\mathrm{Proj}_{S^{d-1}}(x) = x/\|x\|$ is the operator projecting onto the unit sphere, and $\mathrm{Optim}$ is an optimiser of choice, such as Adam. Each evaluation of $W_p$ and its gradient in one dimension costs $\bigo(N\log N)$, so the overall complexity is $\bigo(T N \log N)$ for $T$ gradient steps. It is important to point out that the optimisation may be noisy, with the value objective getting worse after some iterations. Indeed, if $z_{t+1}$ is the solution to $\mathrm{Optim}\big( \nabla_u W^p_p(\phi_{u_t} \# \bP, \phi_{u_t} \# \bQ), u_{1:t}\big)$, it is an improvement over $u_t$, i.e., $W^p_p(\phi_{u_t}\#\bP, \phi_{u_t}\#\bQ) \leq W^p_p(\phi_{z_{t+1}}\#\bP, \phi_{z_{t+1}}\#\bQ)$. Written out explicitly,
\begin{align*}
    \sum_{n=1}^N \left| [\langle u_t, x_{1:N} \rangle]_n - [\langle u_t, y_{1:N} \rangle]_n\right|^p \leq \sum_{n=1}^N \left| [\langle z_{t+1}, x_{1:N} \rangle]_n - [\langle z_{t+1}, y_{1:N} \rangle]_n\right|^p,
\end{align*}
Then, $u_{t+1} = \mathrm{Proj}_{S^{d-1}}(z_{t+1}) = z_{t+1} / \|z_{t+1} \|$, and it may happen that $W^p_p(\phi_{u_t}\#\bP, \phi_{u_t}\#\bQ) > W^p_p(\phi_{u_{t+1}}\#\bP, \phi_{u_{t+1}}\#\bQ)$. The desired 
\begin{equation*}
    W^p_p(\phi_{u_t}\#\bP, \phi_{u_t}\#\bQ) \leq W^p_p(\phi_{u_{t+1}}\#\bP, \phi_{u_{t+1}}\#\bQ)
\end{equation*}
is guaranteed when $\|z_{t+1}\|^p \leq 1$, which may not happen.

\subsection{Generalised sliced Wasserstein}

The \emph{generalised (max-)sliced Wasserstein} (GSW and max-GSW) distances \citep{kolouri_generalized_2022} extend SW and max-SW by using a family of nonlinear feature maps $\{f_\theta : \bR^d \to \bR\}_{\theta \in \Theta}$ instead of linear projections. Formally,
\begin{align*}
    \text{GSW}_p(\bP, \bQ)
    &\coloneqq \left( \bE_{\theta \sim \mu} W_p^p(f_\theta \# \bP, f_\theta \# \bQ ) \right)^{\nicefrac{1}{p}} \\
    \text{max-GSW}_p(\bP, \bQ)
    &\coloneqq\left( \sup_{\theta \in \Theta} W_p^p(f_\theta \# \bP, f_\theta \# \bQ ) \right)^{\nicefrac{1}{p}},
\end{align*}
where $f_\theta \# \bP$ denotes the pushforward of $\bP$ by $f_\theta$ and $\mu$ is a probability measure over the parameter space $\Theta$. For $f_\theta(x)=\langle \theta,x\rangle$ and $\Theta=S^{d-1}$ with uniform $\mu$, GSW reduce to the standard SW distances. For expected GSW, sampling $\theta_1, \dots, \theta_L \sim \mu$ yields an estimator with the same $\bigo(LN \log N)$ computational complexity as expected SW~\citep{kolouri_generalized_2022}. For max-GSW, the projected gradient descent approach of~\eqref{eq:max_sw_optimisation} applies, at the same complexity of $\bigo(T N \log N)$ as for max-SW.

Statistical and topological properties of GSW depend completely on the choice of the family $\{f_\theta:\theta \in\Theta\}$.~\citet{kolouri_generalized_2022} consider the specific case of polynomial $f_\theta$, and show GSW is then a metric on probability distributions on $\bR^d$.

\subsection{Kernel sliced Wasserstein}

A special case of the GSW arises when the feature maps $f_\theta$ are drawn from a reproducing kernel Hilbert space (RKHS). Let $k:\bR^d\times\bR^d\to\bR$ be a positive definite kernel that induces the RKHS $\calH$ with unit sphere $S_\calH$. Then, the \emph{kernel sliced Wasserstein} (KSW) can be introduced as
\begin{align*}
    \text{e-KSW}_p(\bP, \bQ)
    &\coloneqq \left( \bE_{u \sim \gamma} W_p^p(u \# \bP, u \# \bQ ) \right)^{\nicefrac{1}{p}}, \\
    \text{max-KSW}_p(\bP, \bQ)
    &\coloneqq\left( \sup_{u \in S_\calH} W_p^p(u \# \bP, u \# \bQ ) \right)^{\nicefrac{1}{p}},
\end{align*}
where $\gamma$ is some probability measure on $S_\calH$. The expected KSW is a new construct, while max-KSW was introduced in~\citet{wang_two-sample_2022}, and studied further in~\citet{wang2025statistical}; in both papers $k$ was assumed to be universal. Finding the optimal $u^*$ for max-KSW was shown to be NP-hard in~\citet{wang2025statistical}; they propose an estimator at cost $\bigo(T^{3/2} N^2)$. Though still more expensive than computing the V-statistic estimator of MMD, this is an improvement over $\bigo(TN^3)$ in the original work of~\citet{wang_two-sample_2022}.

As pointed out in the main text, the choice of a uniform $\gamma$ in e-KSW, while seemingly natural, may not be feasible as there is no uniform or Lebesgue measure in infinite dimensional spaces. In the chapter, we propose a practical choice of $\gamma$ that facilitates an efficient estimator, and study computational cost. Further, we establish statistical and topological properties that apply to both expected and max-KSW, and do not assume a universal kernel.

\subsection{Sinkhorn Divergence}

The entropic regularisation of optimal transport leads to the \emph{Sinkhorn divergence}~\citep{cuturi2013sinkhorn, Genevay2019}. For distributions $\bP, \bQ$ and regularisation parameter $\varepsilon>0$, the entropic OT cost is defined as
\begin{align*}
    W_{p,\varepsilon}(\bP, \bQ)
    &\coloneqq \left(\inf_{\pi\in\Gamma(\bP, \bQ)} \bE_{(X,Y)\sim\pi}[\|X-Y\|^p]
    +\varepsilon \mathrm{KL}(\pi \| \bP \otimes \bQ )\right)^{\nicefrac{1}{p}}.
\end{align*}
The Sinkhorn divergence then corrects for the entropic bias,
\begin{align}
\label{eq:sinkhorn}
    \mathrm{S}_{p,\varepsilon}(\bP, \bQ)
    &\coloneqq W_{p,\varepsilon}(\bP, \bQ)
    - W_{p,\varepsilon}(\bP, \bP)/2
    - W_{p,\varepsilon}(\bQ, \bQ)/2.
\end{align}
This quantity interpolates between MMD-like behavior for large $\varepsilon$ and true Wasserstein for $\varepsilon\to0$, and can be computed efficiently via Sinkhorn iterations at cost $\bigo(N^2)$ per iteration~\citep{cuturi2013sinkhorn}.

\subsection{Kernel covariance embeddings}

Kernel covariance (operator) embeddings (KCE,~\citet{makigusa2024two}) represent the distribution $\bP$ as the second-order moment of the function $k(X, \cdot)$, for $X \sim \bP$, as an alternative to the first-order moment (the kernel mean embedding). Due to being moments of the same distribution, the two share key positives and drawbacks: KCE for kernel $k$ exists if and only if KME for $k^2$ exists, and the kernel $k$ is covariance characteristic if and only if $k^2$ is mean-characteristic~\citep{bach2022information}. The divergence proposed in~\citet{makigusa2024two} is the distance between the KCE, and is estimated at $\bigo(N^3)$ due to the need to compute full eigendecomposition of the KCE in order to compute the norm. In contrast, our proposed kernel quantile embeddings (KQE) embed quantiles, and therefore the relation to the KCE comes down to matching quantiles (which always exist, and come with an efficient estimator), compared to matching the second moment in the infinite-dimensional RKHS (which may not exist, and requires eigenvalue decomposition).

\subsection{Kernel median embeddings}

The median embedding~\citep{nienkotter2022kernel} of $\bP$ is the geometric median of $k(X, \cdot)$, $X \sim \bP$ in the RKHS, i.e., the RKHS element which, on average, is $L^1$-closest to the point $k(X, \cdot)$. Explicitly put, it is the function $\mathrm{med}_\bP \in \calH$ defined through
\begin{equation*}
    \mathrm{med}_\bP = \argmin_{f \in \calH} \int_\calH \| f(\cdot) - k(x, \cdot) \|_\calH \bP(\d x).
\end{equation*}
The median exists for any separable Hilbert space~\citep{minsker_geometric_2015}. However, even for an empirical $\bP_N = \nicefrac{1}{N}\sum_{n=1}^N \delta_{x_n}$, there is no closed-form solution to this $L^1$-problem, and the median is typically approximated using iterative algorithms like Weiszfeld’s algorithm. The estimator proposed in~\citet{nienkotter2022kernel} has a computational complexity of $\bigo(N^2)$. The property of being median-characteristic, as far as the authors are aware, has not been explored, and no theoretical guarantees are available.

The connection to 1D-projected quantiles as done in KQE, even specifically the 1D-projected median, is also unclear. Expanding the understanding of geometric median embeddings is an area for future research.

\subsection{Other Related Work}

Kernel methods have also been studied in the context of quantile estimation and regression~\citep{Sheather1990,Li2007}. These methods, however, focus on using either kernel density estimation or kernel ridge regression to estimate univariate quantiles. In contrast, our focus lies in exploring directional quantiles in the RKHS, and using them to estimate distances between distributions. We introduce this idea in the following section.

\section{Connection between Centered and Uncentered Quantiles}
\label{appendix:centered_quantiles}

\begin{proposition}[Centered $\ekqd_2$]
    The $\ekqd_2$ and $\supkqd_2$ correspondence derived based on centered directional quantiles, now expressed as $\widetilde{\ekqd_2^2}(\bP, \bQ; \mu, \gamma)$ and $\widetilde{\supkqd_2^2}(\bP, \bQ; \mu, \gamma)$ can be expressed as follows,
\begin{align*}
\widetilde{\ekqd_2^2}(\bP, \bQ; \mu, \gamma)
&= \ekqd_2^2(\bP, \bQ; \mu, \gamma) + \MMD^2(\bP, \bQ)\\
    &\hspace{2.77cm} - \bE_{u\sim\gamma}[(\bE_{X\sim \bP}[u(X)] - \bE_{Y\sim \bQ}[u(Y)])^2] , \\
&\leq \ekqd^2_2(\bP, \bQ; \mu, \gamma) + \MMD^2(\bP, \bQ)\\
\widetilde{\supkqd_2^2}(\bP, \bQ; \mu, \gamma)
    &= \supkqd_2^2(\bP, \bQ; \mu, \gamma) + \MMD^2(\bP, \bQ)  \\
    &\hspace{2.77cm} - \sup_{u\in S_\calH} \left[\left(\bE_{X\sim \bP}[u(X)] - \bE_{Y\sim \bQ}[u(Y)]\right)^2 \right]\\
    &\leq \supkqd_2(\bP, \bQ; \mu, \gamma) + \MMD^2(\bP, \bQ)
\end{align*}
\end{proposition}
\begin{proof}
    Let $\bP, \bQ \in \calP(\calX)$ be measures on some instance space $\calX$. Further, define $\psi: x\mapsto k(x,\cdot)$, and write $\bP_\psi = \psi \# \bP$ and $\bQ_\psi = \psi \# \bQ$. Now $\bP_\psi$ and $\bQ_\psi$ are measures on the RKHS $\calH_k$. Recall the definition of centered directional quantiles in \Cref{sec:background_quantiles},
    \begin{align*}
        {\tilde \rho_{\bP_\psi}}^{\alpha, u} = \left(\rho^\alpha_{\phi_u \# \bP_\psi} - \phi_u(\bE_{Y\sim \bP_\psi}[Y])\right)u + \bE_{Y\sim \bP_\psi}[Y]
    \end{align*}
    Now since we are working in the RKHS $\calH_k$, the expectation term $\bE_{Y\sim \bP_\psi}[Y]$ corresponds to the kernel mean embedding $\mu_\bP \coloneqq \bE_\bP[k(X, \cdot)]$, thus we can rewrite the above expression as,
    \begin{align*}
        {\tilde \rho_{\bP_\psi}}^{\alpha, u} = \left(\rho_{\phi_u \# \bP_\psi} - \langle u, \mu_\bP\rangle\right)u + \mu_\bP
    \end{align*}
    $\tilde{\rho}^{\alpha, u}_{\bQ_\psi}$ can be defined analogously. Now consider integrating the difference between the two centered directional quantiles along all quantile levels, leading to
    \begin{align}
        \tilde{\tau}_2(\bP, \bQ, \mu, u) = \left(\int_0^1 \|{\tilde \rho_{\bP_\psi}}^{\alpha, u} - {\tilde \rho_{\bQ_\psi}}^{\alpha, u}\|_{\calH_k}^2 \mu(\d \alpha)\right)^{\nicefrac{1}{2}} \label{eq:centered-quad-u}
    \end{align}
    We now proceed to show $\tilde{\tau}_2^2(\bP, \bQ, \mu, u)$ ,where $\mu$ is the Lebesgue measure, can be expressed as a sum between an uncentered $\ekqd_2$ term with the MMD. Starting with expanding the RKHS norm inside the integrand,
    \begin{align}
        \|{\tilde \rho_{\bP_\psi}}^{\alpha, u} - {\tilde \rho_{\bQ_\psi}}^{\alpha, u}\|_{\calH_k}^2 &= \|\underbrace{(\rho^{\alpha}_{\phi_u\# \bP_\psi} - \rho^{\alpha}_{\phi_u\# \bQ_\psi} - \langle u, \mu_\bP - \mu_\bQ \rangle)}_{=:A \in \bR} u + \mu_\bP - \mu_\bQ\|_{\calH_k}^2 \nonumber \\
        &= \|Au + (\mu_\bP-\mu_\bQ)\|_{\calH_k}^2 \nonumber \\
        &= 2\langle Au, \mu_\bP-\mu_\bQ\rangle + \|Au\|_{\calH_k}^2 + \|\mu_\bP-\mu_\bQ\|_{\calH_k}^2 \nonumber \\
        &= 2A \langle u, \mu_\bP-\mu_\bQ\rangle + A^2 + \MMD^2(\bP, \bQ) \label{eq:2-norm-diff-centered-quantiles}
    \end{align}
    Plugging the expression from~\eqref{eq:2-norm-diff-centered-quantiles} into~\eqref{eq:centered-quad-u}, we get the following,
    \begin{align}
        \tilde{\tau}_2^2(\bP, \bQ, \mu, u) &= \int_0^1 (2A\langle u, \mu_\bP - \mu_\bQ\rangle + A^2)\mu(\d \alpha) + \MMD^2(\bP, \bQ) \nonumber \\
        &= 2\langle u, \mu_\bP - \mu_\bQ\rangle\int_0^1 A \mu(\d \alpha) + \int_0^1 A^2 \mu(\d \alpha) + \MMD^2(\bP, \bQ) \label{eq:centered-tau}
    \end{align}
    For the first term on the right hand side, notice that,
    \begin{align}
        \int_0^1 A \mu(\d \alpha) = \int_0^1 (\rho^{\alpha}_{\phi_u\# \bP_\psi} - \rho^{\alpha}_{\phi_u\# \bQ_\psi} - \langle u, \mu_\bP - \mu_\bQ \rangle) \mu(\d \alpha) \label{eq:integrating-quantile}
    \end{align}
    Recall standard results from probability theory that integrating the quantile function between $0$ to $1$ with the Lebesgue measure returns you the expectation, specifically, that is,
    \begin{align*}
        \int_0^1 \rho^\alpha_{\phi_u\# \bP_\psi} \mu(\d \alpha) = \bE_{X \sim \bP}[u(X)] = \langle u, \mu_\bP\rangle.
    \end{align*}
    Using this fact, the terms in~\eqref{eq:integrating-quantile} cancels out, leaving $\int_0^1 A\mu(\d \alpha) = 0$. Therefore, continuing from~\eqref{eq:centered-tau}, we have,
    \begin{align*}
        \tilde{\tau}_2^2(\bP, \bQ, \mu, u) &= \int_0^1 A^2 \mu(\d \alpha) + \MMD^2(\bP, \bQ) \\
        &= \int_0^1 (\rho^{\alpha}_{\phi_u\# \bP_\psi} - \rho^{\alpha}_{\phi_u\# \bQ_\psi} - \langle u, \mu_\bP - \mu_\bQ \rangle)^2 \mu(\d \alpha) + \MMD^2(\bP, \bQ) \\
        &= \int_0^1 \|(\rho_{s_{\mu_\bP,u}\# (\phi_u \# \bP_\psi)}^{\alpha} - \rho_{s_{\mu_\bQ,u}\# (\phi_u \# \bQ_\psi)}^{\alpha})u\|^2 \mu(\d \alpha) + \operatorname{MMD^2}(\bP, \bQ)
    \end{align*}
    where $s_{\mu_\bP,u}: \bR \to \bR$ is a shifting function defined as $s_{\mu_\bP,u}(r) = r - \langle u, \mu_{\bP} \rangle$ for $r\in\bR$. Alternatively, after expanding the terms in $A^2$, we can express $\tilde{\tau}_2^2(\bP, \bQ, \mu, u)$ as,
    \begin{align*}
        \tilde{\tau}_2^2(\bP, \bQ, \mu, u) &= \int_0^1 (\rho_{\phi_u\# \bP_\psi} - \rho_{\phi_u\#\bQ_\psi})^2 \mu(\d \alpha) - (\bE[u(X) - u(Y)])^2 \\
        &\hspace{5.7cm}+ \MMD^2(\bP, \bQ) \\
        &= \tau_2^2(\bP, \bQ, \mu, u) + \MMD^2(\bP, \bQ) - (\bE[u(X) - u(Y)])^2
    \end{align*}
As a result, for $\gamma$ a measure on the unit sphere of $\calH_k$, the centered version of $\ekqd_2$ and $\supkqd_2$, now expressed as $\widetilde{\ekqd_2}$ and $\widetilde{\supkqd_2}$, are given by,
\begin{align*}
    \widetilde{\ekqd_2^2}&(\bP, \bQ; \mu, \gamma) \\
    &= \bE_{u\sim \gamma}\left[\tilde{\tau}_2^2(\bP, \bQ; \mu, u)\right] \\
    &= \ekqd_2(\bP, \bQ; \mu, \gamma)^2 + \MMD^2(\bP, \bQ) \\
    &\hspace{3cm}- \bE_{u\sim\gamma}[(\bE_{X\sim \bP}[u(X)] - \bE_{Y\sim \bQ}[u(Y)])^2], \\
    &\leq \ekqd_2(\bP, \bQ; \mu, \gamma)^2 + \MMD^2(\bP, \bQ) \\
    \widetilde{\supkqd_2^2}&(\bP, \bQ; \mu, \gamma) \\
    &= \sup_{u\in S_\calH} \tilde{\tau}_2^2(\bP, \bQ; \mu, u) \\
    &= \sup_{u\in S_\calH} \left(\tau_2^2(\bP, \bQ, \mu, u) - (\bE[u(X)] - \bE[u(Y)])^2\right) + \MMD^2(\bP, \bQ) \\
    &\leq \sup_{u\in S_\calH}\tau_2^2(\bP, \bQ; \mu, u) - \sup_{u\in S_\calH}(\bE[u(X)] - \bE[u(Y)])^2 + \MMD^2(\bP, \bQ) \\
    &\leq \supkqd_2(\bP, \bQ; \mu, \gamma)^2 + \MMD^2(\bP, \bQ).
\end{align*}

\end{proof}

When $\nu\equiv \mu$ and the connections to sliced Wasserstein distances explored in~\Cref{res:connections_slicedwasserstein} and~\Cref{res:connections_maxslicedwasserstein} emerge, the mean-shifting property of Wasserstein distances allows us to express centered KQD as a sum of MMD and uncentered KQD, a curious interpretation of centering.

\section{Proofs of Theoretical Results}\label{appendix:proofs}

This section now provides the proof of all theoretical results in the main text.
\subsection{Proof of~\Cref{res:cramer-wold}}
\label{sec:proof_projections_determine_distribution}

The main result in this section,~\Cref{res:projections_determine_distribution}, shows that the set of $\bR$ measures $\{u \#\bP : u \in S_\calH \}$ fully determines the distribution $\bP$. Since quantiles determine the distribution,~\Cref{res:cramer-wold} follows immediately.

Being concerned with the RKHS case specifically allows us to prove the result under mild conditions by using~\emph{characteristic functionals}, an extension of characteristic functions to measures on spaces beyond $\bR^d$. Characteristic functionals describe Borel probability measures as operators acting on some function space $\calF: \calX \to \bR$.
\begin{definition}[\citet{vakhania1987probability}, Section IV.2.1]
    The~\emph{characteristic functional} $\varphi_\bP: \calF \to \bC$ of a Borel probability measure $\bP$ on $\calX$ is defined as
    \begin{equation*}
        \varphi_\bP(f) = \int_\calX e^{if(x)} \bP (\d x).
    \end{equation*}
\end{definition}

Theorem 2.2(a) in~\citet[Chapter 4]{vakhania1987probability} establishes that a $\bP$-characteristic functional on $\calF$ uniquely determines the distribution $\bP$, on the smallest $\sigma$-algebra under which all functions $f \in \calF$ are measurable. Therefore, when $\calF$ is such that this $\sigma$-algebra coincides with the Borel $\sigma$-algebra, the distribution is fully determined by $\bP$-characteristic functional on $\calF$. We show that, indeed, this holds in our setting, for $\calF=\calH$.
\begin{lemma}
    Suppose~\Cref{as:input_space,as:kernel} holds. Then, the Borel $\sigma$-algebra $\calB(\calX)$ is the smallest $\sigma$-algebra on $\calX$ under which all functions $f \in \calH$ are measurable.
\end{lemma}
\begin{proof}
    Denote by $\hat C(\calX, \calH)$ the smallest $\sigma$-algebra on $\calX$ under which all functions $f \in \calH$ are measurable, and recall that the Borel $\sigma$-algebra is the $\sigma$-algebra that contains all closed sets. Therefore, we need to show that $\hat C(\calX, \calH)$ contains every closed set in $\calX$. We split the proof into two parts: (1) show that $\calH$ contains a countable separating subspace, and (2) show that this implies that every closed set lies in $\hat C (\calX, \calH)$.
    \paragraph{$\calH$ contains a countable separating subspace.} Recall that a function space $\calF$ on $\calX$ is said to be separating when for any $x_1 \neq x_2 \in \calX$, there is a function $f \in \calF$ such that $f(x_1) \neq f(x_2)$. Since $k$ is separating, $\calH$ is separating. Since $\calH$ is separable, it contains a countable dense subspace $\calH_0 \subseteq \calH$. By $\calH_0$ being dense in $\calH$, it must also be separating.
    \paragraph{Every closed set lies in $\hat C (\calX, \calH)$.} By~\citet[Section I.1, Exercise 9]{vakhania1987probability}, all compact sets in $\calX$ lie in $\hat C(\calX, \calH_0)$, by $\calH_0$ being countable, continuous, separating space of real-valued functions. By definition, $\hat C(\calX, \calH_0) \subseteq \hat C(\calX, \calH)$, and so $\hat C(\calX, \calH)$ contains all compact sets. We now show this means every closed set must also lie in $\hat C(\calX, \calH)$.
    
    By $\calX$ being $\sigma$-compact, there is a family of compact sets $\{\calX_i\}_{i=1}^\infty$ such that $\calX = \cup_{i=1}^\infty \calX_i$. Take any closed $K \subseteq X$; then, $K = \cup_{i=1}^\infty (\calX_i \cap K)$. Since $\calX_i \cap K$ is compact as the intersection of a compact set and a closed set, and $\sigma$-algebras are closed under countable unions, $K$ must lie in $\hat C(\calX, \calH)$. As this holds for every closed $K$, we conclude $\calB(\calX)=\hat C(\calX, \calH)$.
\end{proof}

We now restate the RKHS-specific version of the Vakhania result here for completeness.

\begin{theorem}[Theorem 2.2(a) in \citet{vakhania1987probability} for RKHS]
\label{res:char_fnal_is_char}
    Suppose~\Cref{as:input_space,as:kernel} holds, and for Borel probability measures $\bP, \bQ$ on $\calX$, it holds that $\varphi_\bP(f) = \varphi_\bQ(f)$ for every $f \in \calH$. Then, $\bP = \bQ$.
\end{theorem} 

We are now ready to prove the distribution of projections uniquely determines the distribution.

\begin{proposition}
\label{res:projections_determine_distribution}
Under~\Cref{as:input_space,as:kernel}, it holds that
\begin{equation*}
    u\#\bP = u\#\bQ \ \text{ for all } u \in S_\calH \iff \bP = \bQ.
\end{equation*}
\end{proposition}
\begin{proof}
    The main idea of the proof is to show that equality of $u\#\bP$ and $u\#\bQ$ implies equality of characteristic functionals, $\varphi_\bP(f)=\varphi_\bQ(f)$ for all $f \in \calH$ such that $f(x)=tu(x)$ for some $t \in \bR$ and $u$ in the unit sphere. Since such $f$ form the entire $\calH$, the result immediately follows.

    First, recall that $u\#\bP = u\#\bQ$ for all $u$ if and only if their characteristic functions coincide. Then,
    \begin{equation}
    \label{eq:equality_of_char_functions}
        \int_\bR e^{itz} u\#\bP (\d z) = \int_\bR e^{itz} u\#\bQ (\d z) \quad \forall u \in S_\calH,\forall t \in \bR.
    \end{equation}
    Notice that the measure $u\#\bP$ is a pushforward of $\bP$ under the map $x \to u(x)$. Then, for any measurable $g$ it holds that
    \begin{equation}
    \label{eq:pushforward_integration}
        \int_\calX g(u(x)) \bP(\d x) = \int_\bR g(z) u\#\bP(\d z) \quad \forall u \in S_\calH.
    \end{equation}
    Take $g(z) = e^{itz}$, for some $t \in \bR$. Then, for all $u$ it holds that $\int_\bR e^{itz} u\#\bP(\d z)= \int_\bR e^{itz} u\#\bQ(\d z)$, and consequently by~\eqref{eq:equality_of_char_functions} we have that
    \begin{equation}
    \label{eq:equality_of_functional_over_sphere}
        \int_\calX e^{itu(x)} \bP(\d x) = \int_\calX e^{itu(x)} \bQ(\d x)\quad \forall u \in S_\calH,\forall t \in \bR.
    \end{equation}
    Finally, let us pick an $f \in \calH$ and show that $\varphi_\bP(f) = \varphi_\bQ(f)$. Define $u = f/\|f\|$, and $t=\|f\|$; then,
    \begin{equation*}
        \varphi_\bP(f) = \int_\calX e^{if(x)} \bP(\d x) = \int_\calX e^{itu(x)} \bP(\d x),
    \end{equation*}
    and by~\eqref{eq:equality_of_functional_over_sphere}, we arrive at the equality of characteristic functionals, $\varphi_\bP(f) = \varphi_\bQ(f)$. By~\Cref{res:char_fnal_is_char} characteristic functionals uniquely determine the underlying distribution, and therefore $\bP = \bQ$.
\end{proof}

For the sake of clarity, we give the proof of the original result.

\begin{proof}[Proof of~\Cref{res:cramer-wold}]
Suppose $\{\rho_\bP^{\alpha,u} : \alpha \in [0, 1], u \in S_\calH\} = \{\rho_\bQ^{\alpha,u} : \alpha \in [0, 1], u \in S_\calH\}$ for some Borel probability measures $\bP, \bQ$. For any fixed $u$, since every quantile of $u\#\bP$ and $u\#\bQ$ coincide, the measures coincide as well, $u\#\bP=u\#\bQ$. As that holds for every $u$, by~\Cref{res:projections_determine_distribution}, $\bP = \bQ$.
\end{proof}

As discussed in the main text, the assumptions in~\Cref{res:cramer-wold} are much weaker than those typically used to establish injectivity of kernel mean embeddings. \citet{Bonnier2023} prove that their alternative RKHS embeddings, kernelised cumulants, are injective when the kernel is continuous, bounded, and point-separating, and the input space is Polish. These conditions are already far more permissive than the usual KME requirements, but they are still stronger than those in~\Cref{res:cramer-wold}. We now show how our assumptions can be weakened further.

Provided $\calX$ is a Tychonoff space, i.e., a completely regular Hausdorff space, part (b) of Theorem 2.2 in~\citet{vakhania1987probability} says the following.
\begin{theorem}[Theorem 2.2(b) in \citet{vakhania1987probability} for RKHS]
\label{res:char_fnal_is_char_b}
    Suppose $\calX$ is Tychonoff,~\Cref{as:kernel} holds, and for Radon probability measures $\bP, \bQ$ on $\calX$, it holds that $\varphi_\bP(f) = \varphi_\bQ(f)$ for every $f \in \calH$. Then, $\bP = \bQ$.
\end{theorem} 
Therefore, when~\Cref{as:input_space} is replaced with $\calX$ being Tychonoff,~\Cref{res:cramer-wold} continues to hold, but only for Radon $\bP, \bQ$ rather than any Borel $\bP, \bQ$. Radon probability measures can be intuitively seen as the "non-pathological" Borel measures, a restriction employed in order to drop the regularity assumptions of $\calX$ being separable and $\sigma$-compact.

\subsection{Proof of \Cref{res:if_meanchar_then_quantchar}}
\label{appendix:proof_if_meanchar_then_quantchar}
We prove that every mean-characteristic kernel is quantile-characteristic, and give an example quantile-characteristic kernel that is not mean-characteristic.

\paragraph{mean-characteristic $\Rightarrow$ quantile-characteristic.} Suppose $k$ on $\calX$ is mean-characteristic, and $\bP \neq \bQ$ are any probability measures on $\calX$. We will identify a unit-norm $u$ for which the sets of quantiles of $u \# \bP$ and $u \# \bQ$ differ.

Since $k$ is mean characteristic, $\mu_\bP \neq \mu_\bQ$, and $\MMD^2(\bP, \bQ) = \|\mu_\bP - \mu_\bQ \|^2_\calH > 0$. Recall that MMD can be expressed as
\begin{equation*}
        \MMD^2(\bP, \bQ) = \sup_{u \in \calH, \|u\|_\calH \leq 1}\left| \bE_{X \sim \bP} u(X) - \bE_{Y \sim \bQ} u(Y)\right|^2,
\end{equation*}
and the supremum is attained at $u^* = (\mu_\bP - \mu_\bQ) / \|\mu_\bP - \mu_\bQ \|_\calH$~\citep{gretton2012kernel}. In other words, $\bE_{X \sim \bP} u^*(X) \neq \bE_{Y \sim \bQ} u^*(Y)$: the means of $u^*\#\bP$ and $u^*\#\bQ$ do not coincide. Therefore, the measures $u^*\#\bP$ and $u^*\#\bQ$ do not coincide, or equivalently $\{\rho_{u^*\#\bP}^\alpha : \alpha \in [0,1]\} \neq \{\rho_{u^*\#\bQ}^\alpha : \alpha \in [0,1]\}$. Then, $\{\rho_\bP^{u, \alpha} : \alpha \in [0,1], u \in S_\calH\} \neq \{\rho_\bQ^{u, \alpha} : \alpha \in [0,1], u \in S_\calH\}$. And since this holds for any arbitrary $\bP \neq \bQ$, the kernel $k$ is quantile-characteristic.

\paragraph{quantile-characteristic $\not\Rightarrow$ mean-characteristic.}

To show the converse implication does not hold, we provide an example when $k$ is quantile-characteristic but not mean-characteristic. Take $\calX = \bR^d$, and let $k$ be a degree $T$ polynomial kernel, $k(x, x') = (x^\top x' + 1)^T$. Since~\Cref{as:input_space,as:kernel} hold ($\bR^d$ is Polish, and $k$ is trivially continuous and separating), by~\Cref{res:cramer-wold} the kernel $k$ is quantile-characteristic.

Now, we show $k$ is not mean-characteristic. Suppose $\bP$ and $\bQ$ are such that $\bE_{X \sim \bP} X^t = \bE_{Y \sim \bQ} Y^t$ for $t \in \{1, \dots, T\}$: for example, the Gaussian and Laplace distribution with matching expectation and variance and $T=2$, as is done in~\Cref{sec:experimental_results}. Then, $\bE_{X \sim \bP} (X^\top x')^t = \bE_{Y \sim \bP} (Y^\top x')^t$ for any $x' \in \bR^d$, and since
\begin{align*}
    \mu_\bP(x') \coloneqq \bE_{X\sim \bP} k(X, x') = \bE_{X\sim \bP} [(X^\top x' + 1)^T] &= \bE_{X\sim \bP} \left[\sum_{t=0}^T \begin{pmatrix} T \\ t \end{pmatrix} (X^\top x')^t \right] \\
    &= \sum_{t=0}^T \begin{pmatrix} T \\ t \end{pmatrix} \bE_{X\sim \bP} \left[(X^\top x')^t \right],
\end{align*}
it holds that $\mu_\bP=\mu_\bQ$. The kernel is not mean-characteristic.

\subsection{Proof of \Cref{res:consistency_KQE}}\label{appendix:proof_consistency_KQE}
By the Theorem in~\citet[Section 2.3.2]{serfling2009approximation}, for any $\varepsilon>0$ it holds that
\begin{equation*}
    \bP(| \rho^\alpha_{u \# \bP_N} - \rho^\alpha_{u \# \bP}| > \varepsilon) \leq 2 e^{-2 N \delta_\varepsilon^2},
\end{equation*}
for
\begin{equation*}
    \delta_\varepsilon \coloneqq \min\left\{\int_{\rho^\alpha_{u \# \bP}}^{\rho^\alpha_{u \# \bP} + \varepsilon} f_{u \# \bP}(t) \d t, \int_{\rho^\alpha_{u \# \bP}-\varepsilon}^{\rho^\alpha_{u \# \bP}} f_{u \# \bP}(t) \d t\right\}.
\end{equation*}
Since it was assumed $f_{u\#\bP}(t) \geq c_u > 0$, it holds that $\delta_\varepsilon \geq c_u \varepsilon$, and $\bP(| \rho^\alpha_{u \# \bP_N} - \rho^\alpha_{u \# \bP}| > \varepsilon) \leq 2 e^{-2 N c_u^2 \varepsilon^2}$, or equivalently,
\begin{equation*}
    \bP(| \rho^\alpha_{u \# \bP_N} - \rho^\alpha_{u \# \bP}| \leq \varepsilon)
    \geq 1 - 2 e^{-2 N c_u^2 \varepsilon^2}.
\end{equation*}
Take $\delta \coloneqq 2 e^{-2 N c_u^2 \varepsilon^2}$. Then,
\begin{equation*}
    \bP(| \rho^\alpha_{u \# \bP_N} - \rho^\alpha_{u \# \bP}| \leq C(\delta, u) N^{-1/2})
    \geq 1 - \delta, \qquad \text{for} \qquad C(\delta, u)=\sqrt{\frac{\log(2/\delta)}{2 c_u^2}}.
\end{equation*}
Since $\| \rho^{\alpha,u}_{\bP_N} - \rho^{\alpha,u}_\bP\|_\calH=| \rho^\alpha_{u \# \bP_N} - \rho^\alpha_{u \# \bP}|$, the proof is complete.

\subsection{Proof of \Cref{res:KQE_characterise_dists}}\label{appendix:proof_quantile_characteristic}
We prove $\ekqd$ and $\supkqd$, defined in~\eqref{eq:general_distances} as
\begin{equation*}
\begin{split}
    \ekqd_p(\bP, \bQ; \nu, \gamma) &= \left(\bE_{u \sim \gamma} \tau_p^p\left(\bP, \bQ; \nu, u \right) \right)^{\nicefrac{1}{p}},\\
    \supkqd_p(\bP, \bQ; \nu) &= \big(\sup_{u \in S_\calH} \tau_p^p\left(\bP, \bQ; \nu, u \right)\big)^{\nicefrac{1}{p}},
\end{split}
\end{equation*}
are probability metrics on the set of Borel probability measures on $\calX$. Symmetry and non-negativity hold trivially.
\paragraph{Triangle inequality.} By Minkowski inequality, for any $\bP, \bP', \bQ$,
\begin{align*}
    \int_0^1 \big|\rho_\bP^\alpha - \rho_{\bP'}^\alpha \big| ^p \nu(\d \alpha) &\leq \Bigg(\left(\int_0^1 \big|\rho_\bP^\alpha - \rho_\bQ^\alpha \big| ^p \nu(\d \alpha) \right)^{1/p} \\
    &\hspace{2cm}+ \left(\int_0^1 \big|\rho_\bQ^\alpha - \rho_{\bP'}^\alpha \big| ^p \nu(\d \alpha) \right)^{1/p} \Bigg)^p.
\end{align*}
Plugging this in and using Minkowski inequality again on the outermost integral, we get
\begin{align*}
    \ekqd_p(\bP, \bP'; \nu, \gamma)
    &= \left(\bE_{u \sim \gamma} \int_0^1 \big|\rho_\bP^\alpha - \rho_{\bP'}^\alpha \big| ^p \nu(\d \alpha) \right)^{1/p} \\
    &\leq \Bigg(\bE_{u \sim \gamma} \Bigg(\left(\int_0^1 \big|\rho_\bP^\alpha - \rho_\bQ^\alpha \big| ^p \nu(\d \alpha) \right)^{1/p} \\
    &\hspace{2cm}+ \left(\int_0^1 \big|\rho_\bQ^\alpha - \rho_{\bP'}^\alpha \big| ^p \nu(\d \alpha) \right)^{1/p} \Bigg)^p \Bigg)^{1/p} \\
    &\leq \Bigg(\bE_{u \sim \gamma} \int_0^1 \big|\rho_\bP^\alpha - \rho_\bQ^\alpha \big| ^p \nu(\d \alpha) \Bigg)^{1/p} \\
    &\hspace{2cm}+ \Bigg( \bE_{u \sim \gamma} \int_0^1 \big|\rho_\bQ^\alpha - \rho_{\bP'}^\alpha \big| ^p \nu(\d \alpha) \Bigg)^{1/p} \\
    &= \ekqd_p(\bP, \bQ; \nu, \gamma) + \ekqd_p(\bQ, \bP'; \nu, \gamma).
\end{align*}
Similarly, since $\sup_x f^p(x) = (\sup_x |f(x)|)^p$ for any $f$,
\begin{align*}
    \supkqd_p(\bP, \bP'; \nu, \gamma)
    &= \left(\sup_{u \in S_\calH} \int_0^1 \big|\rho_\bP^\alpha - \rho_{\bP'}^\alpha \big| ^p \nu(\d \alpha) \right)^{1/p} \\
    &\leq \Bigg(\sup_{u \in S_\calH} \Bigg(\left(\int_0^1 \big|\rho_\bP^\alpha - \rho_\bQ^\alpha \big| ^p \nu(\d \alpha) \right)^{1/p} \\
    &\hspace{2cm}+ \left(\int_0^1 \big|\rho_\bQ^\alpha - \rho_{\bP'}^\alpha \big| ^p \nu(\d \alpha) \right)^{1/p} \Bigg)^p \Bigg)^{1/p} \\
    &= \left( \sup_{u \in S_\calH} \int_0^1 \big|\rho_\bP^\alpha - \rho_\bQ^\alpha \big| ^p \nu(\d \alpha) \right)^{1/p} \\
    &\hspace{2cm}+ \left( \sup_{u \in S_\calH} \int_0^1 \big|\rho_\bQ^\alpha - \rho_{\bP'}^\alpha \big| ^p \nu(\d \alpha) \right)^{1/p} \\
    &= \supkqd_p(\bP, \bQ; \nu, \gamma) + \supkqd_p(\bQ, \bP'; \nu, \gamma).
\end{align*}

\paragraph{Identity of indiscernibles.} In the rest of this section, we show that
\begin{equation*}
    \ekqd_p(\bP, \bQ; \nu, \gamma) = 0 \iff \bP = \bQ,
\end{equation*}
and
\begin{equation*}
    \supkqd_p(\bP, \bQ; \nu, \gamma) = 0 \iff \bP = \bQ.
\end{equation*}
Necessity (i.e., the $\Leftarrow$ direction) holds trivially: quantiles of identical measure are identical. To prove sufficiency, we only need to show that both discrepancies aggregate the directions in a way that preserves injectivity, meaning
\begin{align*}
    \ekqd_p(\bP, \bQ)=0 &\Rightarrow \rho_\bP^{\alpha,u}=\rho_\bQ^{\alpha,u} \text{ for all } \alpha, u,\\
    \supkqd_p(\bP, \bQ)=0 &\Rightarrow \rho_\bP^{\alpha,u}=\rho_\bQ^{\alpha,u} \text{ for all } \alpha, u.
\end{align*}
Together with~\Cref{res:cramer-wold}, this will complete the proof of sufficiency.

First, we show that for any pair of probability measures, a $\nu$-aggregation over the quantiles is injective.

\begin{lemma}
\label{res:nu_aggregation_is_injective}
Let $\nu$ have full support, i.e., $\nu(A)>0$ for any open $A \subset [0, 1]$. For any Borel probability measures $\bP', \bQ'$,
\begin{equation*}
    \int_0^1 | \rho_{\bP'}^\alpha - \rho_{\bQ'}^\alpha |^2 \nu(\d \alpha) = 0 \qquad \Rightarrow \qquad \rho_{\bP'}^\alpha = \rho_{\bQ'}^\alpha \quad \text{for all} \quad \alpha \in [0, 1].
\end{equation*}
\end{lemma}
\begin{proof}
    Suppose $\int_0^1 | \rho_{\bP'}^\alpha - \rho_{\bQ'}^\alpha |^2 \nu(\d \alpha) = 0$, but there is an $\alpha_0$ such that $\rho_{\bP'}^{\alpha_0} \neq \rho_{\bQ'}^{\alpha_0}$. We will show that this implies the existence of an open set (containing $\alpha_0$) over which $|\rho_{\bP'}^{\alpha_0} - \rho_{\bQ'}^{\alpha_0}|^2 > 0$, which will contradict $\nu$ having full support.
    
    Since $|\rho_{\bP'}^{\alpha_0} - \rho_{\bQ'}^{\alpha_0}|^2 > 0$ and the quantile function $\alpha \mapsto q_\bP^\alpha$ is left-continuous (by definition) for any probability measure $\bP$, there is a $\alpha_1<\alpha_0$ such that $|\rho_{\bP'}^{\alpha} - \rho_{\bQ'}^{\alpha}|^2 > 0$ for all $\alpha \in (\alpha_1, \alpha_0]$. Take some $\alpha_2 \in (\alpha_1, \alpha_0)$. Then, for all $\alpha \in (\alpha_1, \alpha_2)$, we have $|\rho_{\bP'}^{\alpha} - \rho_{\bQ'}^{\alpha}|^2 > 0$. We arrive at a contradiction. Such $\alpha_0$ cannot exist, and therefore $\rho_{\bP'}^\alpha = \rho_{\bQ'}^\alpha$ for all $\alpha \in [0, 1]$.
\end{proof}
This result applies directly to the directional differences $\tau_p$. Provided $\nu$ has full support,
\begin{align*}
    \tau_p(\bP, \bQ;\nu, u) = 0 \ \qquad \Rightarrow \qquad \rho_{\bP'}^\alpha = \rho_{\bQ'}^\alpha \quad \text{for all} \quad \alpha \in [0, 1].
\end{align*}
Since supremum aggregation simply considers $u$ that corresponds to the largest $\tau^p_p(\bP, \bQ;\nu, u)$, this concludes the proof for $\supkqd$.
Expectation aggregation over the directions $u$ needs an extra result, given below.
\begin{lemma}
Let $\gamma$ have full support on $S_\calH$, and $\nu$ have full support on $[0, 1]$. For any Borel probability measures $\bP, \bQ$ on $\calX$,
\begin{equation*}
    \bE_{u \sim \gamma} \tau^p_p(\bP, \bQ;\nu, u) = 0 \qquad \Rightarrow \qquad \bP = \bQ.
\end{equation*}
\end{lemma}
\begin{proof}
    Same as in the proof~\Cref{res:cramer-wold}, we will use the technique of characteristic functionals $\varphi_\bP, \varphi_\bQ$, to carefully prove equality almost everywhere with respect to a full support measure $\gamma$ implies full equality. Consider the function
    \begin{equation*}
        f \mapsto \varphi_\bP(f) - \varphi_\bQ(f),
    \end{equation*}
    which is continuous by continuity of characteristic functionals. Define $f_0 \equiv 0$, the zero function in $\calH$. The set
    \begin{align*}
        \calH^{\setminus 0} 
        &\coloneqq \{f \in \calH \setminus \{f_0\}: \varphi_\bP(f) - \varphi_\bQ(f) \in \bR \setminus \{0\}\} \\
        &= \{f \in \calH \setminus \{f_0\}: \varphi_\bP(f) \neq \varphi_\bQ(f)\}
    \end{align*}
    is open, as a preimage of an open set $\bR \setminus \{0\}$, intersected with an open set $\{\calH \setminus \{f_0\}\}$. Since the projection map $f \mapsto f/\|f\|_\calH$ is open on $\calH \setminus \{f_0\}$, the projection of $\calH_{\setminus 0}$ onto $S_\calH$ is open. In other words, the set
    \begin{equation*}
        S_\calH^{\setminus 0} \coloneqq \{u \in S_\calH: \varphi_\bP(t_u u) \neq \varphi_\bQ(t_u u) \text{ for some } t_u \in \bR \}
    \end{equation*}
    is open in $S_\calH$. Then, by definition of characteristic functionals, for $u \in S_\calH^{\setminus 0}$ it holds that
    \begin{equation*}
        \varphi_{u\#\bP}(t_u) = \varphi_\bP(t_u u) \neq \varphi_\bQ(t_u u) = \varphi_{u\#\bQ}(t_u),
    \end{equation*}
    which implies the characteristic functions of $u\#\bP$ and $u\#\bQ$ are not identical, and therefore $u\#\bP \neq u\#\bQ$. Since $\nu$ has full support on $[0, 1]$, it follows that
    \begin{equation*}
        \tau^p_p(\bP, \bQ;\nu, u)=\int_0^1 | \rho_{u\#\bP}^\alpha - \rho_{u\#\bQ}^\alpha |^p \nu(\d \alpha) > 0,\qquad \text{ for all } u \in S_\calH^{\setminus 0}
    \end{equation*}
    We arrive at a contradiction: since $\gamma$ has full support on $S_\calH$ and $S_\calH^{\setminus 0} \subseteq S_\calH$ was shown to be an open set, it holds that
    \begin{equation*}
        \bE_{u \sim \gamma} \tau^p_p(\bP, \bQ;\nu, u) \geq \int_{S_\calH^{\setminus 0}} \tau^p_p(\bP, \bQ;\nu, u) \gamma(\d u) > 0.
    \end{equation*}
    Therefore, for $\bE_{u \sim \gamma} \tau^p_p(\bP, \bQ;\nu, u)$ to be zero, $S_\calH^{\setminus 0}$ must be empty, which, by construction, can only happen when $\calH^{\setminus 0}$ is empty, i.e., $\varphi_\bP(f) = \varphi_\bQ(f)$ for all $f \in \calH \setminus f_0$, where $f_0 \equiv 0$. Since $\varphi_\bP(f_0) = \varphi_\bQ(f_0)$ holds trivially for any $\bP, \bQ$, the characteristic functionals of $\bP$ and $\bQ$ are identical. By~\Cref{res:char_fnal_is_char}, $\bP = \bQ$. This concludes the proof.
\end{proof}

\subsection{Proof of \Cref{res:consistency_KQD}}\label{appendix:proof_consistency_KQD}
We start with two auxiliary lemmas that, when combined, bound $\ekqd$ approximation error due to replacing $\bP, \bQ$ with $\bP_N, \bQ_N$ in $N^{-1/2}$. This will be crucial in showing the convergence of the approximate e-KQD to the true e-KQD.
\begin{lemma}
\label{res:lemma_for_consistency}
For any measure $\nu$ on $[0, 1]$ and any measure $\gamma$ on $S_\calH$, it holds that
\begin{align*}
    &|\ekqd_1(\bP_N, \bQ_N; \nu, \gamma) - \ekqd_1(\bP, \bQ; \nu, \gamma)| \\
    &\hspace{2cm}\leq \ekqd_1(\bP_N, \bP; \nu, \gamma) + \ekqd_1(\bQ_N, \bQ; \nu, \gamma).
\end{align*}
\end{lemma}
\begin{proof}
By the definition of $\ekqd_1$ and Jensen inequality for the absolute value,
\begin{align*}
    &|\ekqd_1(\bP_N, \bQ_N; \nu, \gamma) - \ekqd_1(\bP, \bQ; \nu, \gamma)| \\
    &\hspace{3cm}= \left|\bE_{u \sim \gamma} \left[ \int_0^1 \left( |\rho^\alpha_{u\#\bP_N} - \rho^\alpha_{u\#\bQ_N}| - |\rho^\alpha_{u\#\bP} - \rho^\alpha_{u\#\bQ}| \right) \d \alpha\right]\right| \\
    &\hspace{3cm}\leq \bE_{u \sim \gamma} \left[ \int_0^1 \left| |\rho^\alpha_{u\#\bP_N} - \rho^\alpha_{u\#\bQ_N}| - |\rho^\alpha_{u\#\bP} - \rho^\alpha_{u\#\bQ}| \right| \d \alpha\right]
\end{align*}
By the reverse triangle inequality followed by the triangle inequality,
\begin{equation}
\label{eq:rev_tr_then_tr}
\begin{split}
    \left|| \rho_{u\#\bP_N}^\alpha - \rho_{u\#\bQ_N}^\alpha | - | \rho_{u\#\bP}^\alpha - \rho_{u\#\bQ}^\alpha | \right|
    &\leq |\rho_{u\#\bP_N}^\alpha - \rho_{u\#\bP}^\alpha + \rho_{u\#\bQ}^\alpha - \rho_{u\#\bQ_N}^\alpha| \\
    &\leq |\rho_{u\#\bP_N}^\alpha - \rho_{u\#\bP}^\alpha|+|\rho_{u\#\bQ_N}^\alpha - \rho_{u\#\bQ}^\alpha|,
\end{split}
\end{equation}
and the statement of the lemma follows.
\end{proof}
\begin{lemma}
\label{res:lemma2_for_consistency}
Let $\nu$ be a measure on $[0, 1]$ with density $f_\nu$ bounded above by $C_\nu>0$.
With probability at least $1-\delta/4$, for $C'(\delta) = 2 C_\nu \sqrt{\log(8/\delta)/2}$, it holds that
\begin{align*}
    \ekqd_1(\bP_N, \bP; \nu, \gamma) \leq \frac{C'(\delta)}{2} N^{-1/2}
\end{align*}
\end{lemma}
\begin{proof}
Recall that
\begin{align*}
    \ekqd_1(\bP_N, \bP; \nu, \gamma) &= \bE_{u \sim \gamma} \left[\tau_1(\bP_N, \bP; \nu, u) \right], \\
    \tau_1(\bP_N, \bP; \nu, u) &= \int_0^1 |\rho_{u\#\bP_N}^\alpha - \rho_{u\#\bP}^\alpha| \nu(\d \alpha) .
\end{align*}
Let $F_{u\#\bP}$ and $F_{u\#\bP_N}$ be the CDFs of $u\#\bP$ and $u\#\bP_N$ respectively. Then,
\begin{align*}
    \int_0^1 |\rho_{u\#\bP_N}^\alpha - \rho_{u\#\bP}^\alpha| \nu(\d \alpha) 
    &\leq C_\nu \int_0^1 |\rho_{u\#\bP_N}^\alpha - \rho_{u\#\bP}^\alpha| \d \alpha \\
    &= C_\nu \int_{u(\calX)} |F_{u\#\bP_N}(t) - F_{u\#\bP}(t) | \d t \\
    &\leq C_\nu \sup_{t \in u(\calX)} |F_{u\#\bP_N}(t) - F_{u\#\bP}(t) |,
\end{align*}
where the last equality is the well known fact that integrated difference between quantiles is equal to integrated difference between CDFs (see, for instance,~\citet[Theorem 2.9]{bobkov2019one}). By the Dvoretzky-Kiefer-Wolfowitz inequality, with probability at least $1-\delta/4$ it holds that,
\begin{equation*}
    \sup |F_{u\#\bP_N}(t) - F_{u\#\bP}(t)| < \sqrt{ \log(8/\delta) / 2} N^{-1/2},
\end{equation*}
and therefore, with probability at least $1-\delta/4$ for $C'(\delta) = 2 C_\nu \sqrt{ \log(8/\delta) / 2}$,
\begin{equation*}
    \tau_1(\bP_N, \bP; \nu, u) = \int_0^1 |\rho_{u\#\bP_N}^\alpha - \rho_{u\#\bP}^\alpha| \nu(\d \alpha) \leq \frac{C'(\delta)}{2} N^{-1/2}.
\end{equation*}
In other words, the random variable $\tau_1(\bP_N, \bP; \nu, u)$ is sub-Gaussian with sub-Gaussian constant $C_\tau \coloneqq C_\nu^2/(2n)$, meaning
\begin{equation*}
    \mathrm{Pr} \left[\tau_1(\bP_N, \bP; \nu, u) \geq \varepsilon\right] \leq 2 \exp\{- \varepsilon^2 / C_\tau^2\}
\end{equation*}
One of the equivalent definitions for a sub-Gaussian random variable is the moment condition: for any $p \geq 1$,
\begin{equation*}
    \bE_{x_1, \dots, x_N} \left[\tau_1(\bP_N, \bP; \nu, u)^p \right] \leq 2 C_\tau^p \Gamma(p/2 + 1).
\end{equation*}
An application of Jensen inequality and Fubini's theorem shows that the moment condition holds for $\bE_{u \sim \gamma} \tau_1(\bP_N, \bP; \nu, u)$,
\begin{align*}
    \bE_{x_1, \dots, x_N} \left[ \left(\bE_{u \sim \gamma} \tau_1(\bP_N, \bP; \nu, u)\right)^p \right] 
    &\leq \bE_{x_1, \dots, x_N} \bE_{u \sim \gamma} \left[\tau_1(\bP_N, \bP; \nu, u)^p \right]\\
    &= \bE_{u \sim \gamma} \bE_{x_1, \dots, x_N} \left[\tau_1(\bP_N, \bP; \nu, u)^p \right] \\
    &\leq 2 C_\tau^p \Gamma(p/2 + 1).
\end{align*}
Therefore, $\bE_{u \sim \gamma} \tau_1(\bP_N, \bP; \nu, u)$ is sub-Gaussian with constant $C_\tau = C_\nu^2/(2n)$, meaning it holds with probability at least $1 - \delta/4$ that
\begin{equation*}
    \ekqd_1(\bP_N, \bP; \nu, \gamma) = \bE_{u \sim \gamma} \tau_1(\bP_N, \bP; \nu, u) \leq \frac{C'(\delta)}{2} N^{-1/2}.
\end{equation*}
\end{proof}

We are now ready to prove the full result.
\begin{proof}[Proof of~\Cref{res:consistency_KQD}]
Let $C_\nu$ be an upper bound on the density of $\nu$.
By triangle inequality, the full error can be upper bounded by $R_L$, the error due to approximation of $\gamma$ with $\gamma_L$, plus $R_N$, the error due to approximation of $\bP, \bQ$ with $\bP_N, \bQ_N$,
\begin{align*}
    &| \ekqd_1(\bP_N, \bQ_N;\nu, \gamma_L) - \ekqd_1(\bP, \bQ;\nu, \gamma) | \\
    &\hspace{2cm}\leq | \ekqd_1(\bP_N, \bQ_N;\nu, \gamma_L) - \ekqd_1(\bP_N, \bQ_N;\nu, \gamma) | \\
    &\hspace{3cm}+ | \ekqd_1(\bP_N, \bQ_N;\nu, \gamma) - \ekqd_1(\bP, \bQ;\nu, \gamma) | \\
    &\hspace{2cm}\eqcolon R_L + R_N.
\end{align*}
We bound $R_L$ in $L^{-1/2}$, and $R_N$ in $N^{-1/2}$, with high probability.

\paragraph{Bounding $R_L$.}
Recall that 
\begin{equation*}
    \ekqd_1(\bP_N, \bQ_N; \nu, \gamma) = \bE_{u \sim \gamma} \left[ \int_0^1 | \rho_{u\#\bP}^\alpha - \rho_{u\#\bQ}^\alpha | \nu(\d \alpha) \right].
\end{equation*}
Therefore, we may apply McDiarmid's inequality provided for any $u, u' \in S_\calH$ we upper bound the difference
\begin{equation*}
    \left| \int_0^1 \left| \rho_{u\#\bP}^\alpha - \rho_{u\#\bQ}^\alpha \right| - \left| \rho_{u'\#\bP}^\alpha - \rho_{u'\#\bQ}^\alpha \right| \nu(\d \alpha) \right|.
\end{equation*}
We have that
\begin{align*}
    &\left| \int_0^1 \left| \rho_{u\#\bP}^\alpha - \rho_{u\#\bQ}^\alpha \right| - \left| \rho_{u'\#\bP}^\alpha - \rho_{u'\#\bQ}^\alpha \right| \nu(\d \alpha) \right|
    \\
    &\hspace{2cm}\stackrel{(A)}{\leq} \int_0^1 \left| \rho_{u\#\bP}^\alpha - \rho_{u\#\bQ}^\alpha \right| \nu(\d \alpha) + \int_0^1 \left| \rho_{u'\#\bP}^\alpha - \rho_{u'\#\bQ}^\alpha \right| \nu(\d \alpha) \\
    &\hspace{2cm}\stackrel{(B)}{\leq} 2 C_\nu \sup_{u \in S_\calH} W_1(u\#\bP, u\#\bQ) \\
    &\hspace{2cm}\stackrel{(C)}{\leq} 2 C_\nu \sup_{u \in S_\calH} \bE_{X \sim \bP} \bE_{Y \sim \bQ} |u(X) - u(Y)|
    \\
    &\hspace{2cm}\stackrel{(D)}{\leq} 2 C_\nu \bE_{X \sim \bP} \bE_{Y \sim \bQ} \sqrt{k(X, X) - 2 k(X, Y) + k(Y, Y)}
\end{align*}
where $(A)$ holds by Jensen's and triangle inequalities; $(B)$ uses boundedness of the density of $\nu$ by $C_\nu$ and the property of the Wasserstein distance in $\bR$ from~\eqref{eq:wasserstein_as_quantiles}; $(C)$ uses the infimum definition of the Wasserstein distance; and $(D)$ holds by the reasoning we employed multiple times through the chapter, via reproducing property, Cauchy-Schwarz, and having $u,u' \in S_\calH$.
So we arrive at a bound
\begin{align*}
    &\left| \int_0^1 \left| \rho_{u\#\bP}^\alpha - \rho_{u\#\bQ}^\alpha \right| - \left| \rho_{u'\#\bP}^\alpha - \rho_{u'\#\bQ}^\alpha \right| \nu(\d \alpha) \right| \\
    &\hspace{2cm}\leq 2 C_\nu \bE_{X \sim \bP} \bE_{Y \sim \bQ} \sqrt{k(X, X) - 2 k(X, Y) + k(Y, Y)} \eqcolon 2 C_\nu C_k.
\end{align*}
Now that boundedness of the difference has been established, by McDiarmid's inequality, with probability at least $1 - \delta/2$ and for $C''(\delta) = \sqrt{2 C_\nu C_k \log(4 / \delta)}$ it holds that
\begin{equation*}
    | \ekqd_1(\bP_N, \bQ_N;\nu, \gamma_L) - \ekqd_1(\bP_N, \bQ_N;\nu, \gamma)| \leq C''(\delta)L^{-1/2}.
\end{equation*}
\paragraph{Bounding $R_N$.} By~\Cref{res:lemma_for_consistency},
\begin{align*}
    &|\ekqd_1(\bP_N, \bQ_N; \nu, \gamma) - \ekqd_1(\bP, \bQ; \nu, \gamma)| \\
    &\hspace{2cm}\leq \ekqd_1(\bP_N, \bP; \nu, \gamma) + \ekqd_1(\bQ_N, \bQ; \nu, \gamma)
\end{align*}
By~\Cref{res:lemma2_for_consistency} and the union bound, with probability at least $1-\delta/2$ and for $C'(\delta) = 2C_\nu \sqrt{\log(8/\delta)/2}$, it holds that
\begin{align*}
    R_N = |\ekqd_1(\bP_N, \bQ_N; \nu, \gamma) - \ekqd_1(\bP, \bQ; \nu, \gamma)| \leq C'(\delta) N^{-1/2}.
\end{align*}
\paragraph{Combining bounds.} By applying the union bound again, to $R_L + R_N$, we get that, with probability at least $1-\delta$,
\begin{align*}
    | \ekqd_1(\bP_N, \bQ_N;\nu, \gamma_L) - \ekqd_1(\bP, \bQ;\nu, \gamma) | 
    &\leq R_L + R_N \\
    &\leq C''(\delta)L^{-1/2} + C'(\delta) N^{-1/2} \\
    &\leq C(\delta)(L^{-1/2} + N^{-1/2}),
\end{align*}
for $C(\delta) = \max\{C'(\delta), C''(\delta)\} = \bigo(\sqrt{\log(1/\delta)})$. This completes the proof.
\end{proof}
As pointed out in the main text, 
\begin{equation*}
    \bE_{X \sim \bP} \bE_{Y \sim \bQ} \sqrt{k(X, X) - 2 k(X, Y) + k(Y, Y)}<\infty
\end{equation*}
holds immediately when $\bE_{X \sim \bP}\sqrt{k(X, X)}$ and $\bE_{X \sim \bQ}\sqrt{k(X, X)}$ are finite, and even more specifically, when the kernel $k$ is bounded. Unbounded $k$ and finite expectations, for example, happens when the tails of both $\bP$ and $\bQ$ decay fast enough to "compensate" for the growth of $k(x, x)$. For instance, when $k$ is a polynomial kernel of any order (which is unbounded), and $\bP$ and $\bQ$ are laws of sub-exponential random variables. For clarity, note that $\bE_{X \sim \bP} \bE_{Y \sim \bQ} \sqrt{k(X, X) - 2 k(X, Y) + k(Y, Y)}$ does not compare to MMD, which integrates $k(X, X')$ rather than $k(X, X)$ (see~\eqref{eq:MMD2_exact}).

For integer $p>1$, proving the $N^{-1/2}$ convergence rate is feasible if more involved, primarily because we can no longer reduce the problem to the convergence of empirical CDFs to true CDFs. In general, for $p>1$,
\begin{align*}
    \int_0^1 |\rho_{u\#\bP_N}^\alpha - \rho_{u\#\bP}^\alpha|^p \d \alpha \neq \int_{u(\calX)} |F_{u\#\bP_N}(t) - F_{u\#\bP}(t) |^p \d t.
\end{align*}
The following result, restated in our notation, makes the added complexity explicit.
\begin{lemma}[Theorem 5.3 in~\citet{bobkov2019one}]
\label{res:wasserstein_convergence}
Suppose $k: \bR^d \times \bR^d \to \bR$ is a bounded kernel, and $\nu$ has a density $0<c_\nu \leq f_\nu \leq C_\nu$ on $[0,1]$. Then, for any $u \in S_\calH$, and for any $p \geq 1$ and $N \geq 1$,
\begin{align*}
    \bE_{x_1, \dots, x_N \sim \bP} \left[ \tau_p^p(\bP_N, \bP; \nu, u) \right] \leq \left(\frac{5p C_\nu}{\sqrt{N + 2}}\right)^p J_p(u\#\bP)
\end{align*}
for
\begin{align*}
    J_p(u\#\bP) = \int_{u(\calX)} \frac{\left( F_{u\#\bP}(t) (1 - F_{u\#\bP}(t)) \right)^{p/2}}{f^{p-1}_{u\#\bP}(t)}\d t.
\end{align*}
Further, it holds that $\bE_{x_1, \dots, x_N \sim \bP} \left[\tau_p^p(\bP_N, \bP; \nu, u)\right] = \bigo(N^{-p/2})$ if and only if $J_p(u\#\bP) < \infty$.
\end{lemma}

We now state a likely result for $p>1$ as a conjecture, and outline the proof.
\begin{conjecture}[\textbf{Finite-Sample Consistency for Empirical KQDs for $p>1$}]
\label{res:consistency_KQD_p}
    Let $\calX \subseteq \bR^d$, $\nu$ have a density, $\bP, \bQ$ be measures on $\calX$ with densities bounded away from zero, $f_\bP(x) \geq c_\bP>0$ and $f_\bQ(x) \geq c_\bQ>0$. Suppose $\bE_{X \sim \bP} [k(X, X)^{\nicefrac{p}{2}}]<\infty$ and $\bE_{Y \sim \bQ} [ k(Y, Y)^{\nicefrac{p}{2}}]<\infty$, and $x_1, \dots, x_N \sim \bP, y_1, \dots, y_N \sim \bQ$. Then,
    \begin{align*}
        \bE_{\substack{x_1, \dots, x_N \sim \bP \\ y_1, \dots, y_N \sim \bQ}}| \ekqd_p(\bP_N, \bQ_N;\nu, \gamma_L) - \ekqd_p(\bP, \bQ;\nu, \gamma) | =\bigo(L^{-\nicefrac{1}{2}} + N^{-\nicefrac{1}{2}}).
    \end{align*}
\end{conjecture}
\begin{proof}[Sketch proof]
    Analogously to the proof of~\Cref{res:consistency_KQD}, we can decompose the term of interest as
    \begin{align*}
        &\bE_{\substack{x_1, \dots, x_N \sim \bP \\ y_1, \dots, y_N \sim \bQ}}| \ekqd_p(\bP_N, \bQ_N;\nu, \gamma_L) - \ekqd_p(\bP, \bQ;\nu, \gamma) | \\
        &\hspace{2cm}\leq 
        \bE_{\substack{x_1, \dots, x_N \sim \bP \\ y_1, \dots, y_N \sim \bQ}}| \ekqd_p(\bP_N, \bQ_N;\nu, \gamma_L) - \ekqd_p(\bP_N, \bQ_N;\nu, \gamma) |\\
        & \hspace{3cm} +\left(\bE_{x_1, \dots, x_N\sim \bP} \ekqd^p_p(\bP_N, \bP; \nu, \gamma) \right)^{\nicefrac{1}{p}} \\
        & \hspace{3cm}+ \left(\bE_{y_1, \dots, y_N\sim \bQ}\ekqd^p_p(\bQ_N, \bQ; \nu, \gamma)\right)^{\nicefrac{1}{p}}
    \end{align*}
    The first term can be, same as in the proof of~\Cref{res:consistency_KQD}, bounded by McDiarmid's inequality. The second term (to the power $p$) takes the form 
    \begin{equation*}
        \bE_{x_1, \dots, x_N\sim \bP} \ekqd^p_p(\bP_N, \bP; \nu, \gamma) = \bE_{x_1, \dots, x_N\sim \bP} \bE_{u\sim \gamma} \tau_p^p(\bP_N, \bP; \nu, u).
    \end{equation*}
    Then, by~\Cref{res:wasserstein_convergence} (possibly modified to account for an extra expectation), to get the result we will need to show that $\bE_{u \sim \gamma} J_p(u\#\bP) < \infty$,
    \begin{equation*}
        \bE_{u \sim \gamma} J_p(u\#\bP) = \bE_{u \sim \gamma} \left[\int_{u(\calX)} \frac{\left( F_{u\#\bP}(t) (1 - F_{u\#\bP}(t)) \right)^{p/2}}{f^{p-1}_{u\#\bP}(t)}\d t \right] < \infty
    \end{equation*}
    The numerator is upper bounded by $2^{-p}$. The denominator may get arbitrarily small without the numerator getting arbitrarily small: when the PDF $f^{p-1}_{u\#\bP}(t)$ is small, the CDF $F_{u\#\bP}(t)$ need not be close to zero or one. Therefore, it is necessary and sufficient to show
    \begin{equation}
    \label{eq:one_over_density}
        \bE_{u \sim \gamma} \left[\int_{u(\calX)} \frac{1}{f^{p-1}_{u\#\bP}(t)}\d t\right] < \infty.
    \end{equation}
    We proceed to outline key elements of the proof of such a result, and leave a rigorous proof for future work. By the coarea formula, and since $f_\bP(x) \geq c_\bP > 0$,
    \begin{equation*}
        f_{u\#\bP}(t) = \int_{u^{-1}(t)} \frac{f_\bP(x)}{|\nabla u(x)|} H^{d-1}(\d x) \geq c_0 \int_{u^{-1}(t)} \frac{1}{|\nabla u(x)|} H^{d-1}(\d x),
    \end{equation*}
    for
    \begin{equation*}
        |\nabla u(x)| = \sqrt{\sum_{i=1}^d \left(\frac{\partial u(x)}{\partial x_i} \right)^2},
    \end{equation*}
    where $u^{-1}(t)=\{x \in \calX : u(x)=t\}$, and $H^{d-1}$ is the $d-1$-dimensional Hausdorff measure, which within $\calX \subseteq \bR^d$ is equal to $d-1$ dimensional Lebesgue measure, scaled by a constant that only depends on $d-1$.
    
    Therefore, the integral in~\eqref{eq:one_over_density} may diverge if the integral
    \begin{equation}
    \label{eq:volume_of_level_set}
        \int_{u^{-1}(t)} \frac{1}{|\nabla u(x)|} H^{d-1}(\d x)
    \end{equation}
    gets very small over "large" parts of $u(\calX)$, on average over $u \sim \gamma$. Trivially, if $u$ is constant over some interval (or more generally, $u$ has infinitely many critical points), the integral diverges. Fortunately, the more general condition is easy to control: if $u$ is a \emph{Morse function} and $\calX$ is compact, then $u$ has only a finite number of critical points. It is a classic result (see, for instance,~\citet[Theorem 1.2]{Hirsch1976}) that Morse functions form a dense open subset of twice differentiable real-valued functions on $\bR^d$, denoted $C^2(\bR^d)$. Therefore, if $\calH \subset C^2(\calX)$ (which can be reduced to smoothness of the kernel $k$; it holds, for instance, for the Mat\'ern-5/2 kernel), we get that $u \sim \gamma$ has a finite number of critical points almost surely under mild regularity assumptions on $\gamma$.
    
    The final ingredient is to use the Morse lemma to lower bound~\eqref{eq:volume_of_level_set} in the epsilon-ball of each critical point. Morse lemma says $u$ is quadratic around each critical point, which yields bounds on both the volume of $u^{-1}(t)$ and $1/|\nabla u(x) |$ in terms of the eigenvalues of the Hessian. Careful analysis of the eigenvalues will be needed to ensure the expectation with respect to $u \sim \gamma$ is finite.
\end{proof}

\subsection{Proof of \Cref{res:connections_slicedwasserstein,res:connections_maxslicedwasserstein}}
\label{appendix:proof_connections_slicedwasserstein}

The equality in~\eqref{eq:wasserstein_as_quantiles} immediately gives the connection of $\ekqd$ and $\supkqd$ to the expected-SW and max-SW, respectively, previously only defined on $\calX=\bR^d$.

Further, for $\calX=\bR^d$, viewing $x \mapsto k(x, \cdot)$ as a transformation on $\calX$ reveals a connection to generalised sliced Wasserstein (GSW, \citet{kolouri_generalized_2022}). In particular, the polynomial kernel $k(x, x') = (x^\top x' + 1)^T$ of odd degree $T$ recovers the polynomial transformation for which GSW was proven to be a probability metric. Outside of the case of the polynomial case, proving that GSW is a metric is highly challenging. This is easier under the kernel framework, as we showed in~\Cref{res:KQE_characterise_dists}.
In~\citet{kolouri_generalized_2022}, the authors investigate learning transformations with neural networks (NNs). An interesting direction for future work is the relationship between said NNs and the kernels they induce.

\subsection{Proof of~\Cref{res:sampling_from_gm}}
\label{sec:proof_sampling_from_gm}
Recall that by definition of Gaussian measures in Hilbert spaces~\citep{kukush2020gaussian}, a random element $f \in \calH$ has the law of a Gaussian measure $\calN(0, C_M)$ on $\calH$ when for any $g \in \calH$,
\begin{equation}
\label{eq:defn_of_gm}
    \langle f, g \rangle_\calH \sim \calN(0, \langle C_M[g], g\rangle).
\end{equation}
Since $C_M[g](x) = \nicefrac{1}{M} \sum_{m=1}^M g(z_m) k(z_m, x)$, by the reproducing property,
\begin{equation}
\label{eq:empirical_integral_op_explicit}
    \langle C_M[g], g\rangle = \frac{1}{M}\sum_{m=1}^M g(z_m)^2.
\end{equation}
Take $f(x) = \nicefrac{1}{\sqrt M} \sum_{m=1}^M \lambda_m k(z_m, x)$, for $\lambda_1, \dots, \lambda_M \sim \calN(0, 1)$. Then, for any $g \in \calH$, by the reproducing property it holds that
\begin{equation*}
    \langle f, g \rangle_\calH = \frac{1}{\sqrt M} \sum_{m=1}^M \lambda_m g(z_m) \sim \calN\left(0, \frac{1}{M} \sum_{m=1}^M g(z_m)^2\right),
\end{equation*}
which is exactly the Gaussian measure with covariance operator $C_M$, as per~\cref{eq:defn_of_gm,eq:empirical_integral_op_explicit}.

\section{Additional Numerical Results}\label{appendix:experiments}

\subsection{Type I control}

\begin{figure}
    \centering
    \includegraphics[width=0.7\linewidth]{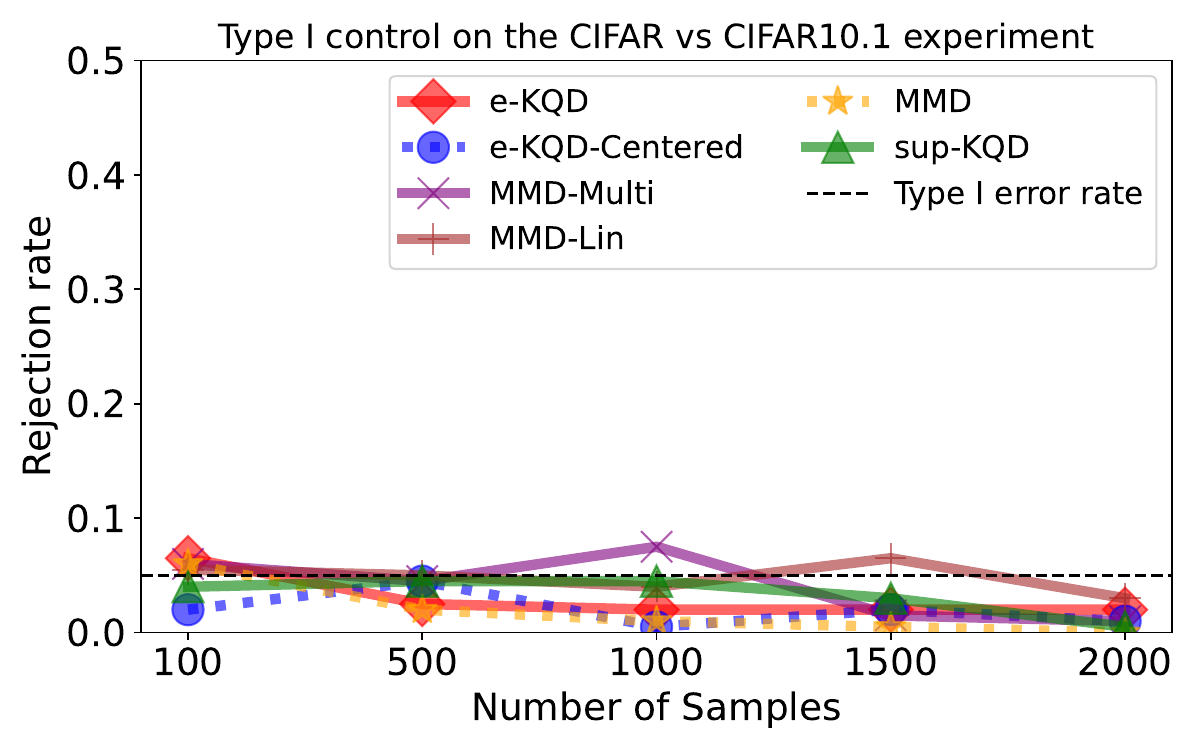}
    \caption[Type I control on CIFAR-10 vs. CIFAR-10.1.]{Type I control results for our experiment on CIFAR-10 vs. CIFAR-10.1. We see all methods control their Type I error around or below the specified Type I error rate $0.05$, thus confirming our tests in the main text are valid testing procedures.}
    \label{fig:type-1-result}
\end{figure}

We report the Type I control experiments for the CIFAR-10 vs. CIFAR-10.1 experiment. Results are shown in \Cref{fig:type-1-result}.

\subsection{\Cref{fig:main_experiments} for $\ekqd_1$}
\label{sec:exp_for_p1}

It is common in power $p$-parameterised methods to select $p=2$, to balance out sensitivity to outliers (which is higher for larger $p$, to the point of methods becoming brittle for $p > 2$), and robustness (which tends to be highest for $p=1$); this trade-off, for instance, inspired the introduction of the Huber loss~\citep{huber1964robust}. However, for completeness, we now repeat experiments in the chapter for $p=1$. The relationship to baseline approaches (MMD, MMD-Multi, and MMD-Lin) remains the same as observed for $p=2$. However, it is evident that $\ekqd_1$ performed better than $\ekqd_2$ at the power decay and galaxy MNIST experiments, but the centered $\ekqd_1$ performed worse than centered $\ekqd_2$ at the Laplace vs. Gaussian experiment. The implications of choosing $p$ warrant a deeper investigation, left to future work.

\begin{figure*}[t]
    \centering
    \includegraphics[width=\linewidth]{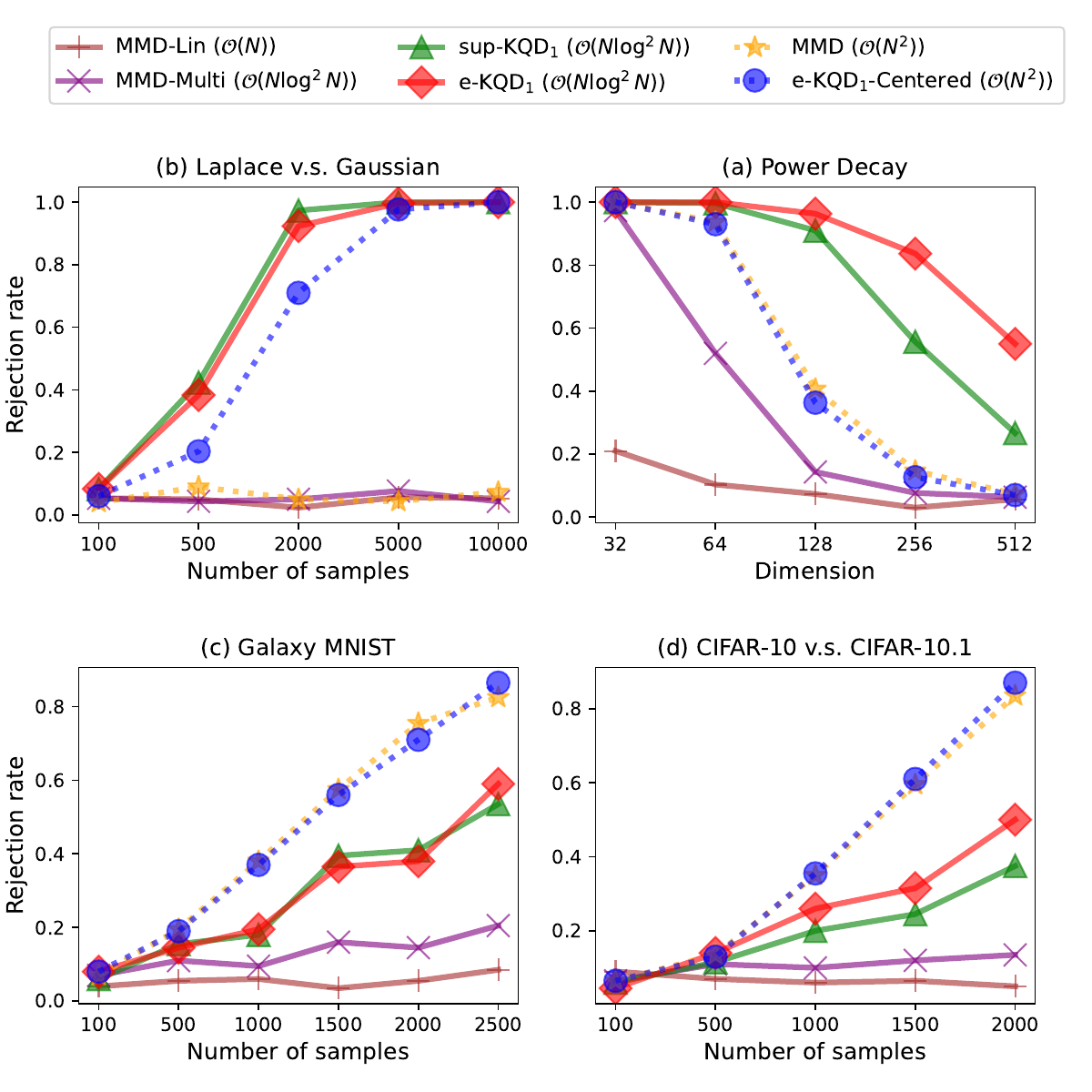}
    \caption[Experimental results comparing KQDs with baseline approaches, for $p=1$]{The experiments in~\Cref{fig:main_experiments} repeated for $p=1$. Experimental results comparing our proposed methods with baseline approaches. A higher rejection rate indicates better performance in distinguishing between distributions. \textbf{Same as for $p=2$, quadratic-time quantile-based estimators perform comparably to quadratic-time MMD estimators, while near-linear time quantile-based estimators often outperform their MMD-based counterparts.}}
    \label{fig:main_experiments_for_p1}
\end{figure*}

\subsection{Comparison of weighting measures}

The Gaussian Kernel Quantile Discrepancy introduced in~\Cref{sec:estimator} has multiple weighting measures that determine properties of the distance: the measure $\nu$ on the quantile levels, the measure $\xi$ within the covariance operator, and the measure $\gamma$ on the unit sphere $S_\calH$. We investigate the impact of varying these.

\begin{figure*}[t]
    \centering
    \includegraphics[width=\linewidth]{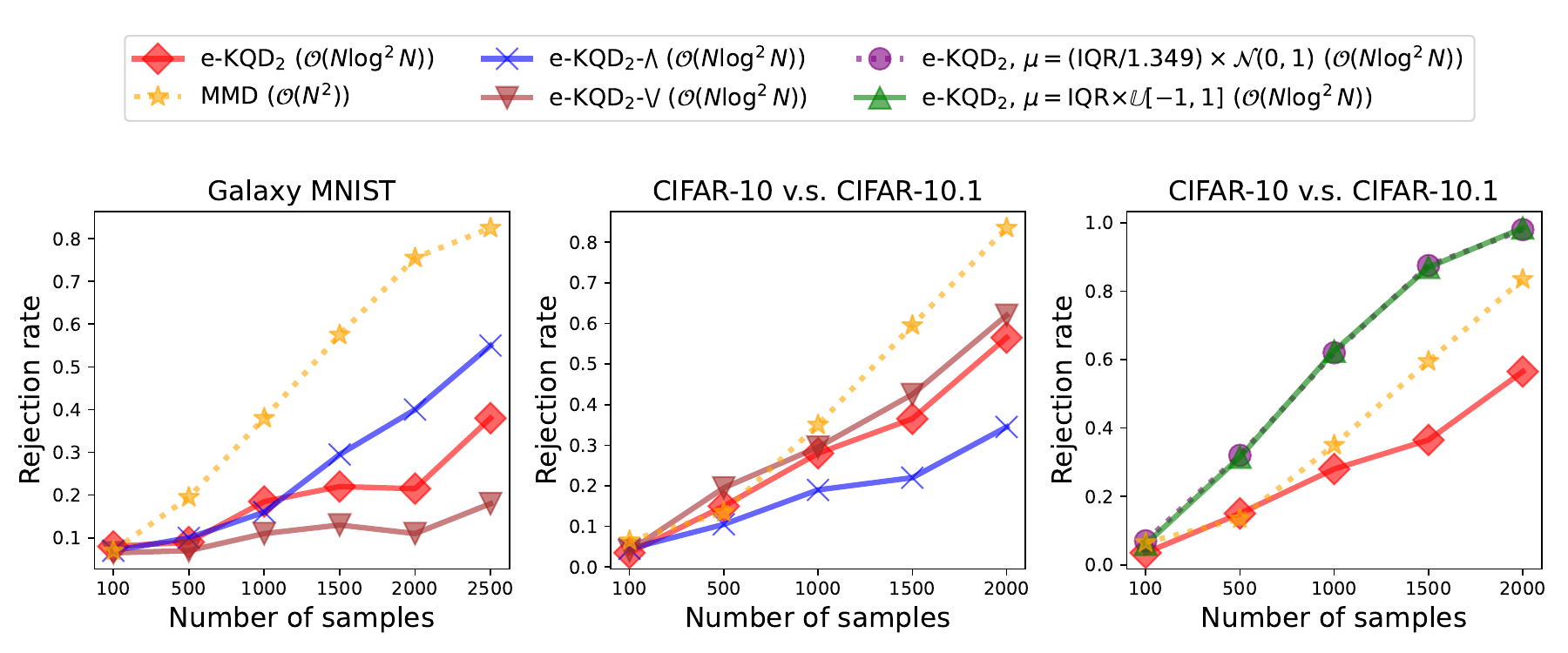}
    \caption[Gaussian KQD test power under different weighting measures.]{Gaussian KQD test power under different weighting measures. \textit{Left, middle:} Varying measure $\nu$: down-weighing ($/\setminus$) extremes boosts power on Galaxy MNIST, while up-weighing ($\setminus/$) them helps on CIFAR. Uniform weighting remains a strong default, with optimal $\nu$ depending on the dataset.
    \textit{Right:} Varying measure $\xi$: using an IQR-scaled Gaussian or uniform default reference measure $\xi$ both outperform MMD, indicating potential advantage of a "default" $\xi$ over the problem-based $\xi=(\bP_N + \bQ_N)/2$.}
    \label{fig:varying_combined}
\end{figure*}

\paragraph{Varying $\nu$.} We conducted the following experiment using the Galaxy MNIST and CIFAR datasets. We varied $\nu$, from assigning more weight to the extreme quantiles to down-weighting them. The results are presented in~\Cref{fig:varying_combined}, where the reverse triangle $\setminus/$ stands for up-weighing extreme quantiles, and the triangle $/\setminus$ stands for down-weighing them. We observed some improvement over the uniform $\nu$: for Galaxy MNIST, test power improved when $\nu$ assigned less weight to extremes, whereas for CIFAR, the opposite was true, with higher test power when more weight was given to extremes. Uniform weighting of the quantiles remained a good choice. This suggests that tuning $\nu$ beyond the uniform is problem-dependent and can enhance performance. The difference likely arises from the nature of the problems: CIFAR datasets, where samples are expected to be similar, benefit from emphasising extremes, while Galaxy MNIST, which contains fundamentally different galaxy images, performs better when “robustified,” i.e., focusing on differences away from the tails. Exploring this further presents an exciting avenue for future work.

\paragraph{Varying $\xi$.} The reference measure $\xi$ in the covariance operator $C$ serves to "cover the input space" and is typically set to a "default" measure on the space; for $\bR^d$, the standard Gaussian measure. The choice $(\bP_N + \bQ_N)/2$ made in the chapter aims to adhere to the most general setting, when no default measure is available other than $\bP_N$ and $\bQ_N$.

We report a comparison on performance when the reference measure is: (1) $(\bP_N + \bQ_N)/2$; (2) a standard Gaussian measure, scaled by IQR/1.349 to match the spread of the data, where IQR is the interquartile range of $\bP_N + \bQ_N$, and 1.349 is the interquartile range of the standard Gaussian; and
(3) a uniform measure on $[-1, 1]^d$, scaled by IQR.

The results, presented in~\Cref{fig:varying_combined}, show performance superior to MMD for the standard/uniform $\xi$. This indicates value in picking a "default" measure when one is available.

\paragraph{Varying $\gamma$.} Varying the measure on the sphere beyond a Gaussian is extremely challenging in infinite-dimensional spaces due to the complexity of both its theoretical definition and practical sampling. Since no practically relevant alternative has been proposed, we leave this direction unexplored.

\subsection{Comparison to sliced Wasserstein distances}

\begin{figure*}[t]
    \centering
    \includegraphics[width=\linewidth]{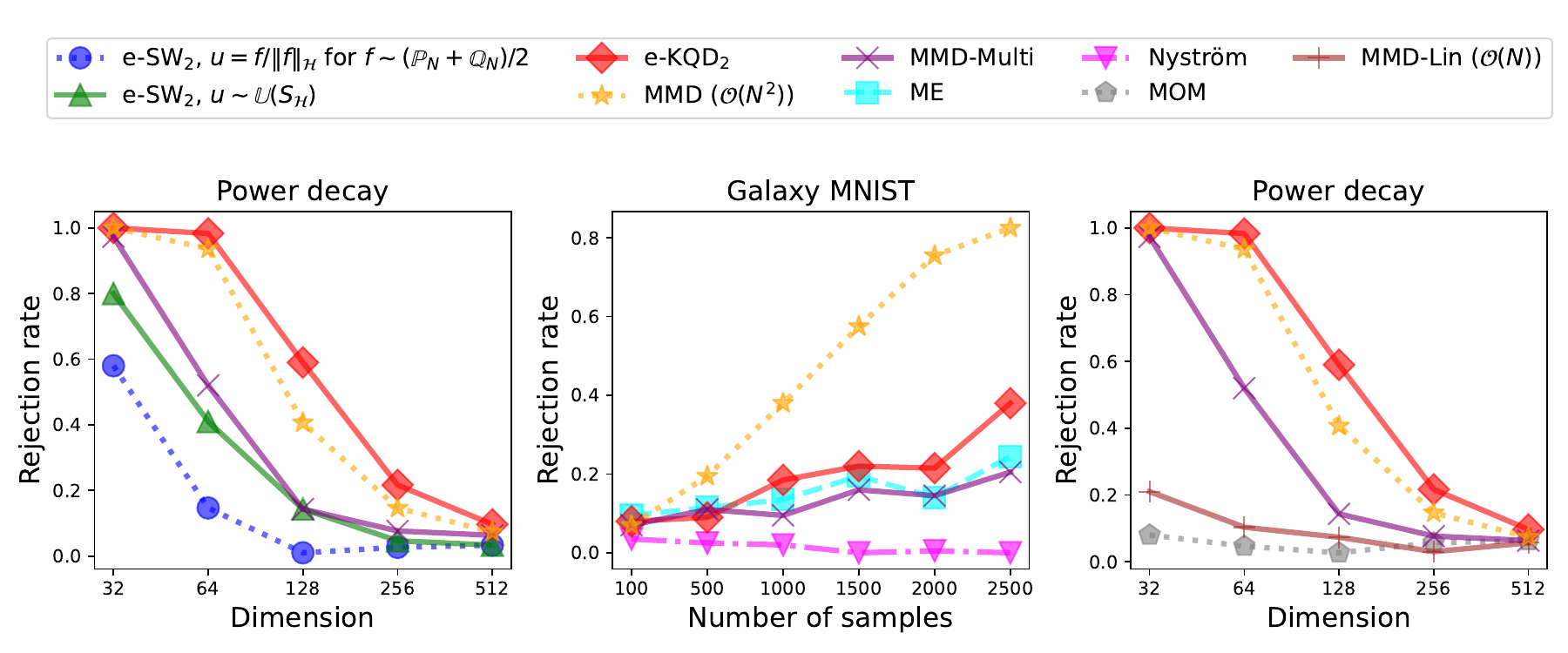}
    \caption[KQD vs. MMD based on other KME approximations]{All methods are cost $\bigo(N \log^2 N)$ unless specified otherwise. \textit{Left:} Gaussian KQD compared with sliced Wasserstein with uniform or data-driven directions, on the power decay problem. Sliced Wasserstein fall well below KQD, consistent with their equivalence to KQD using the less expressive linear kernel.
    \textit{Middle:} Comparison with alternative approximate KME methods, at matching cost. ME matches MMD-multi power, while Nyström-MMD suffers high Type II error.
    \textit{Right:} Comparison with Median-of-Means (MOM) KME approximation, at matching cost. MOM is primarily a robustness-enforcing method, not a cheap-approximation method, and does not perform well at set cost of $\bigo(N \log^2 N)$.}
    \label{fig:sw_and_kme_approx}
\end{figure*}

We extend the power decay experiment to include sliced Wasserstein and max-sliced Wasserstein distances, with directions (1) sampled uniformly on the sphere, and (2) sampled from $(\bP_N + \bQ_N)/2$ and projected onto the sphere. The results are plotted in~\Cref{fig:sw_and_kme_approx}, and show that sliced Wasserstein distances perform significantly worse than $\ekqd$. This outcome is expected, as noted in~\Cref{res:connections_slicedwasserstein,res:connections_maxslicedwasserstein}, sliced Wasserstein is equivalent to $\ekqd$ with the linear kernel, which is less expressive than the Gaussian kernel.

\subsection{Comparison with MMD based on Other KME Approximations}

There are several efficient kernel mean embedding methods available in the literature, and no single approach has emerged as definitively superior. To complement experiments in the chapter, we compare the $\ekqd$ (at matching cost) with (1) The Mean Embedding (ME) approximation of MMD of~\citet{chwialkowski2015fast}, which was identified as the best-performing method in their numerical study; (2) the Nystr\"om-MMD method of~\citet{chatalic2022nystrom}, and (3) the Median-of-Means (MOM) approximation of~\citet{lerasle2019monk}, specifically, their faster method (MONK BCD-Fast) that achieves matching cost to our e-KQD at the number of blocks $N/\log N$.

The results are presented in~\Cref{fig:sw_and_kme_approx}. ME performs at the level of MMD-multi, while Nystr\"om has extremely high Type II error, likely due to sensitivity to hyperparameters. Due to Median-of-Means still being considerably slower than e-KQD (with the number of optimiser iterations set to $T=100$), we apply it to a cheaper Power Decay problem (rather than the larger and more complicated Galaxy MNIST), where it performs at the level of the linear approximation of MMD. This may be due to MOM primarily being a robustness-enforcing method, rather than a method aiming to build an efficient approximation of MMD.

\end{document}